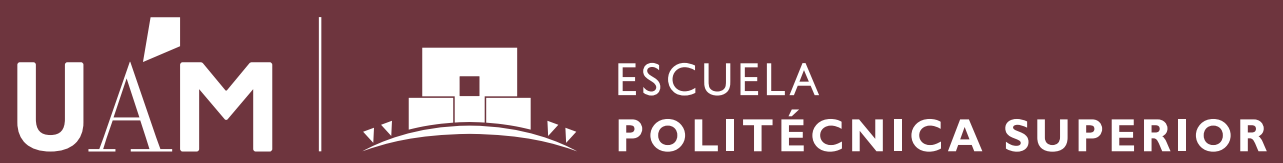

# Uncertainty Estimation and Generalization Bounds for Modern Deep Learning

## Advances in Function-Space Variational Inference, Linearized Laplace Approximation, Deep Ensembles, and Chernoff-Based Generalization Bounds

**Luis Antonio Ortega Andrés**

Department of Computer Science
Machine Learning Group

Madrid, June 2026

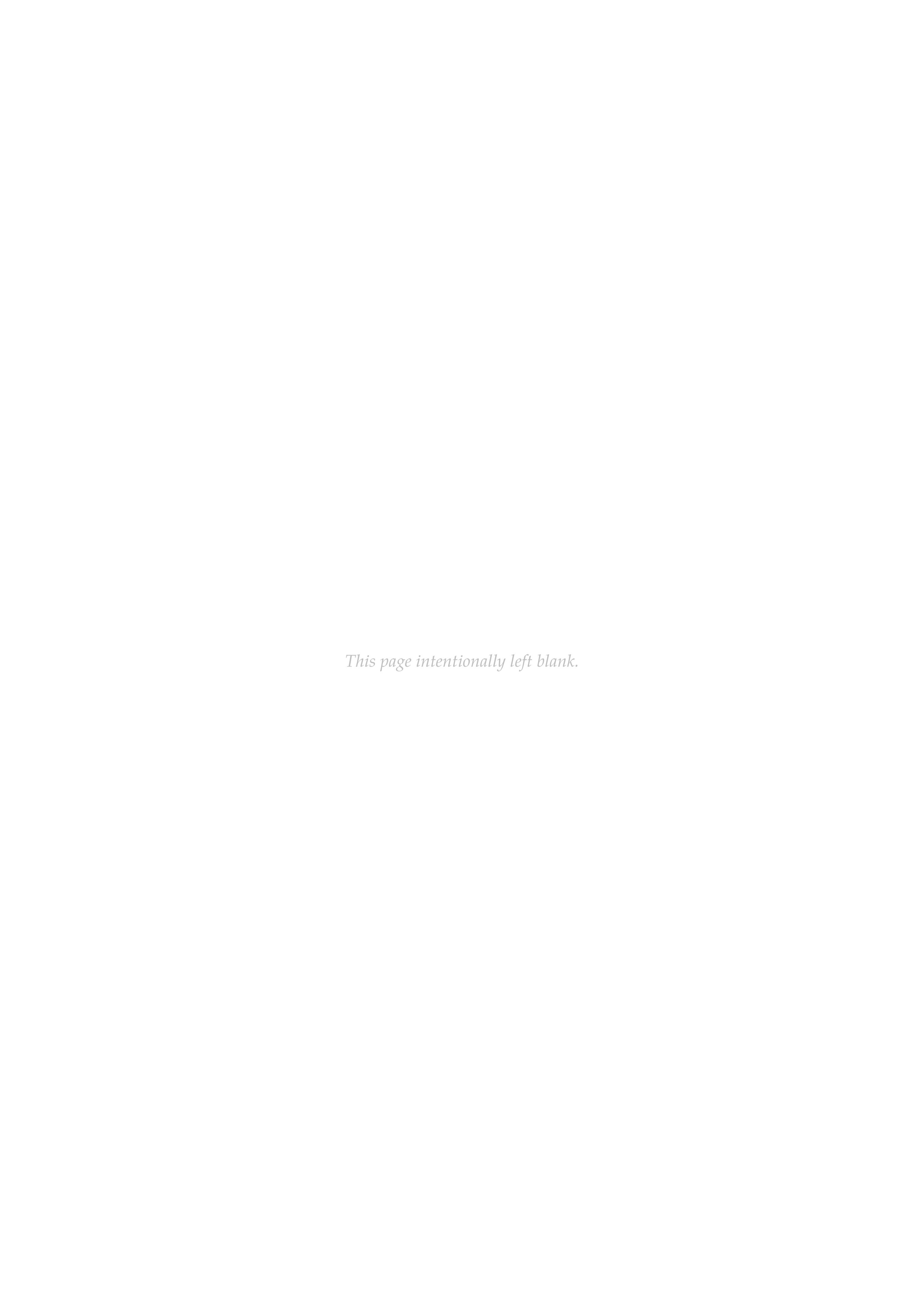

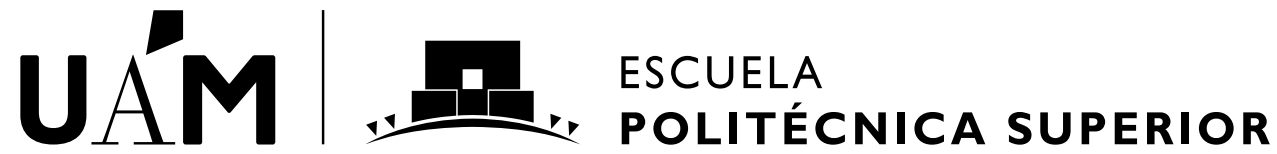

# Uncertainty Estimation and Generalization Bounds for Modern Deep Learning

## Advances in Function-Space Variational Inference, Linearized Laplace Approximation, Deep Ensembles, and Chernoff-Based Generalization Bounds

**Luis Antonio Ortega Andrés**

**Supervisor:** Daniel Hernández Lobato
*Associate Professor, Universidad Autónoma de Madrid*

Department of Computer Science
Machine Learning Group

*Dissertation*

Madrid, June 2026

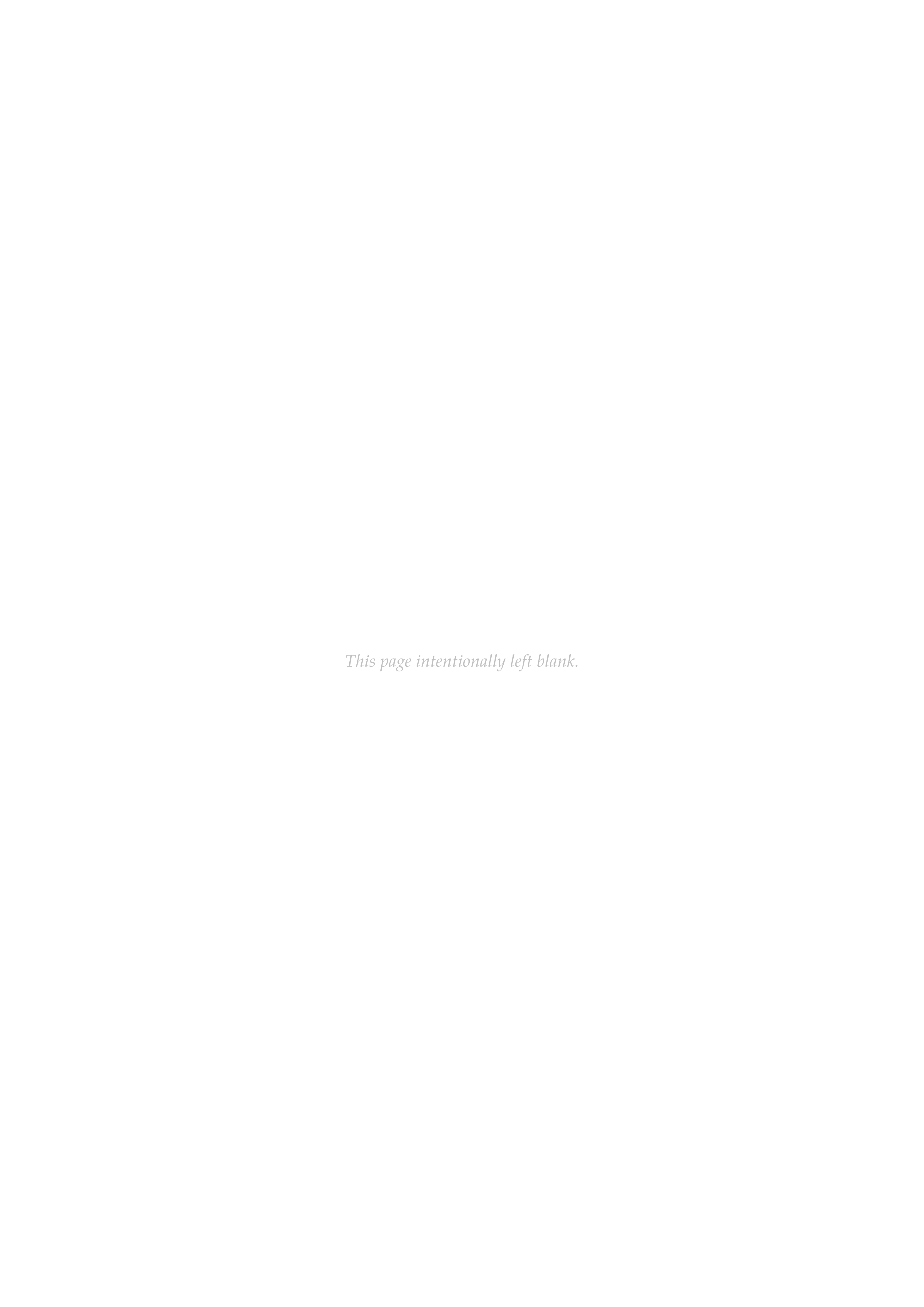
This page intentionally left blank.

**Uncertainty Estimation and Generalization Bounds for Modern Deep Learning**

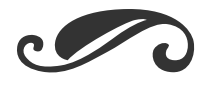

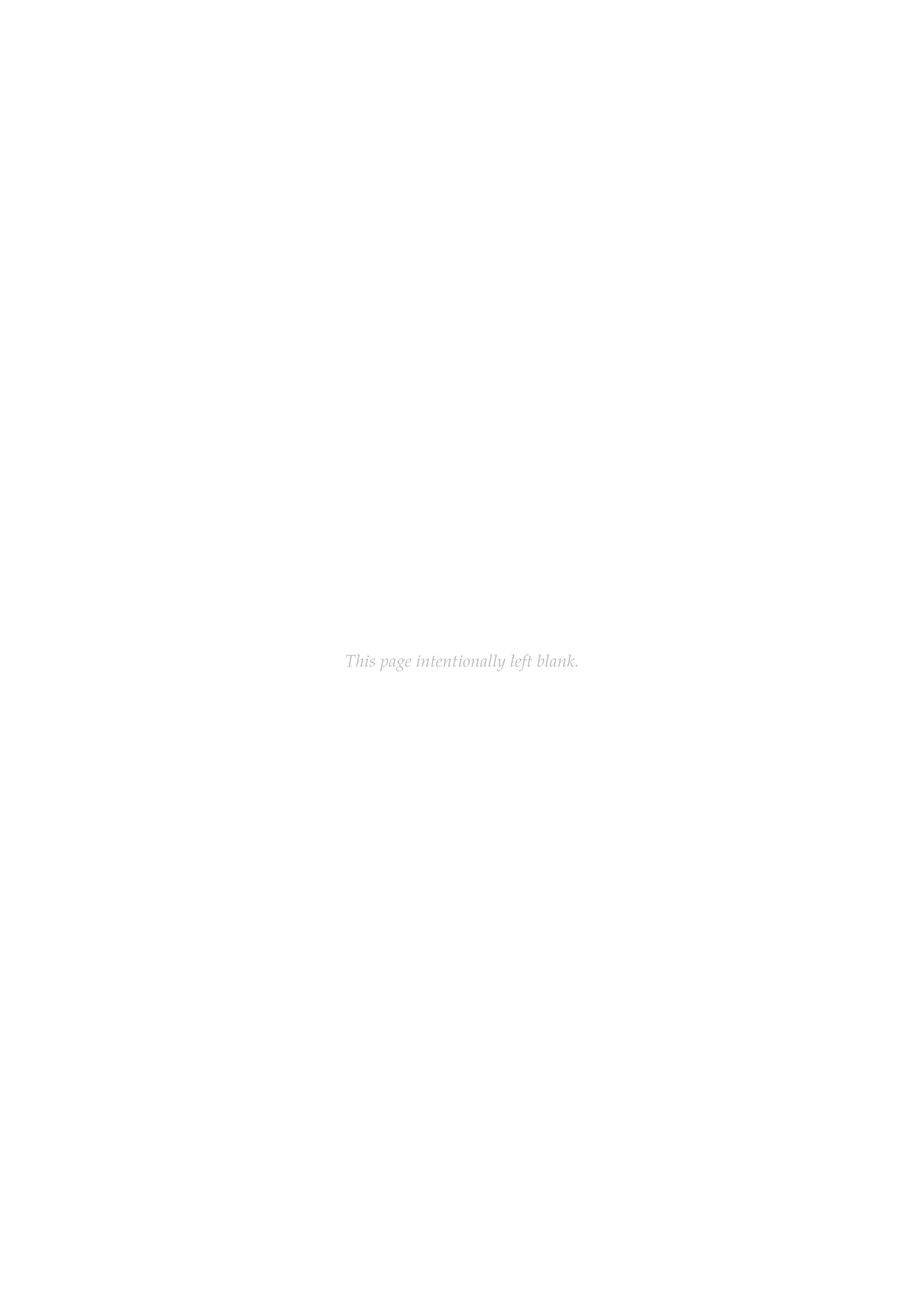

To Elizabeth,
the little light that was already shining
while these pages were slowly aligning.

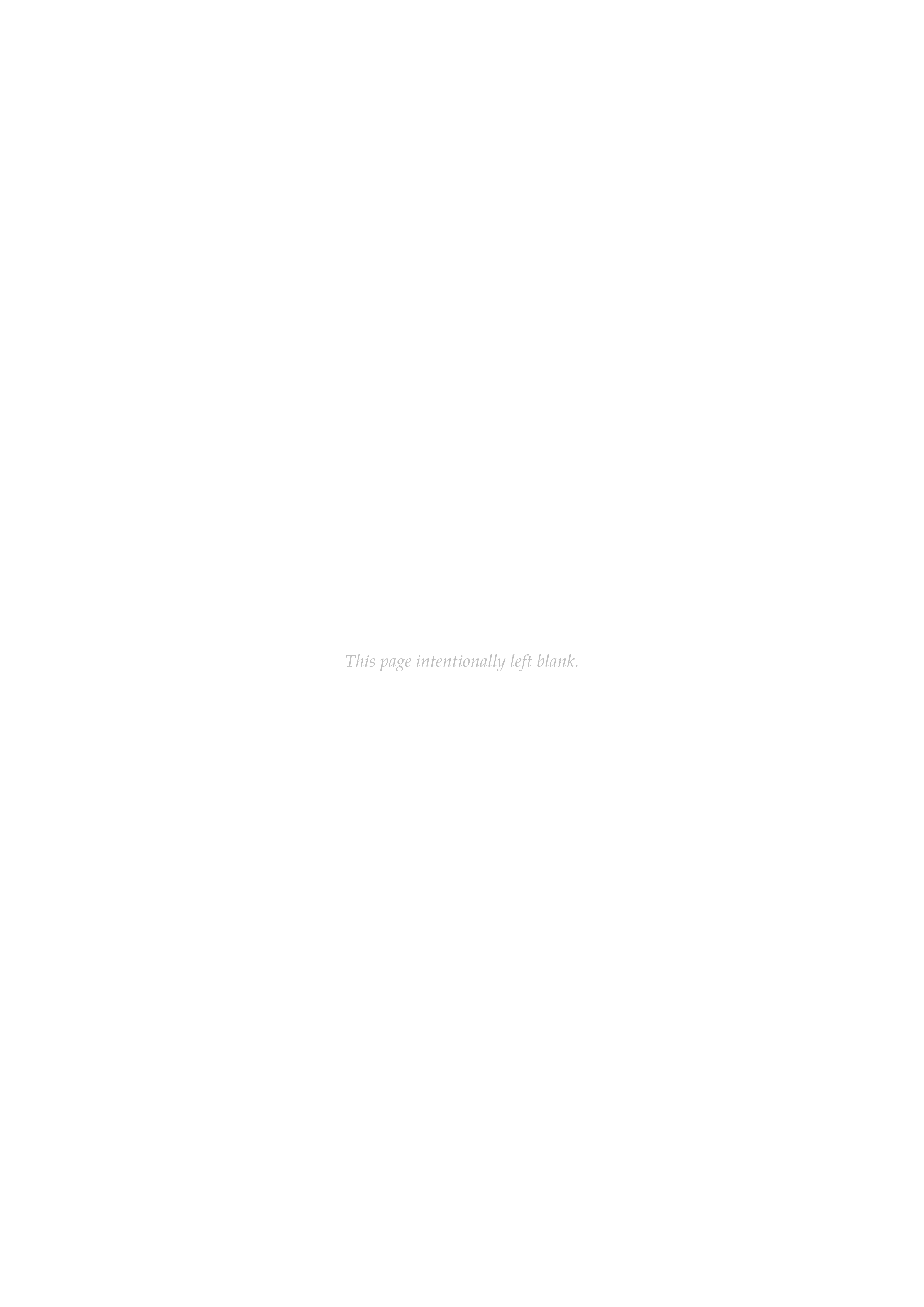
*This page intentionally left blank.*

# Acknowledgements

First and foremost, I would like to express my deepest gratitude to my supervisor, Daniel Hernández-Lobato. Thank you for taking me on as a student and for trusting me. Your way of asking simple, precise questions at exactly the right moment has profoundly influenced the way I think about research. I am especially grateful for your patience during the slow phases of this thesis, for your detailed comments on drafts (no matter how rough), and for your door always being open for a quick question that usually turned into a long discussion.

I am equally grateful to Andrés Masegosa; our meetings before deadlines were always incredibly intense, but equally helpful. Thank you for your honest feedback and for pointing out the weak spots in my arguments. I have learned a lot from both your scientific insight and your way of working.

I would like to thank the members of my thesis committee, Fran Ruiz, Isabel Valera, Antonio Artés-Rodríguez, Ángela Fernández, and Carlos Alaíz, for agreeing to read and evaluate this thesis. I know how busy you all are, and I truly appreciate the time and attention you have devoted to this work.

To my flatmates Mike, Victor, Alexis, and Antonio, thank you for being the everyday support behind this thesis: for the late-night talks, for the calm and the coffee, for the laughter, and for cooking and cleaning for me when a deadline was near. Thank you so much for all those years of joy.

To Antonio and Saez, thank you for always being available when I reached out to discuss any mathematical difficulties that arose over these years.

To Gloria, thank you for enduring countless dinners with me while my mind was busy thinking about how to efficiently implement things in PyTorch.

To Efi and Jesús, thank you for becoming two of the most important people in my life and for putting up with me while I spent months writing this dissertation.

Finally, I would like to thank my parents, who have given me every opportunity.

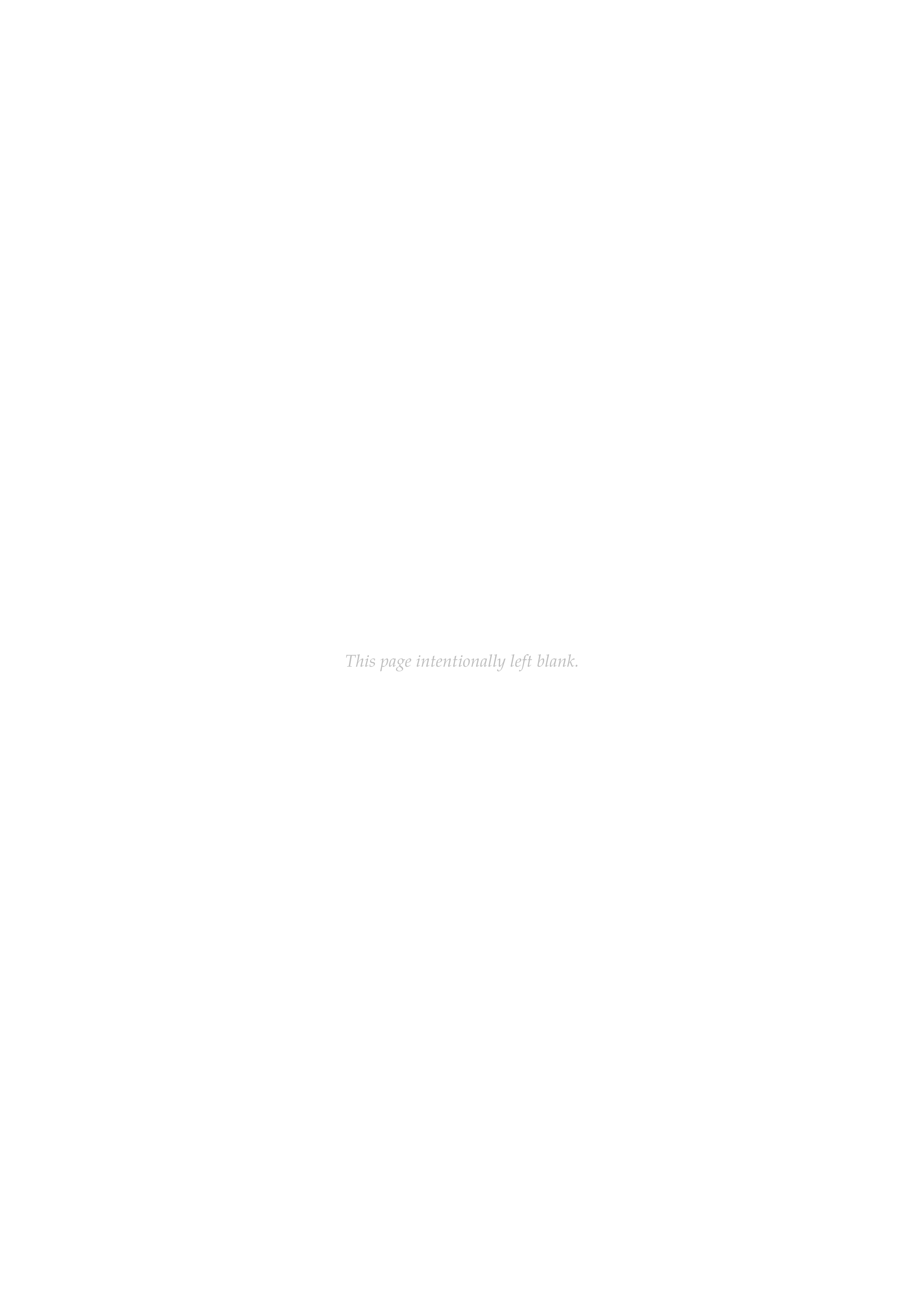
This page intentionally left blank.

# Abstract

This thesis investigates how Bayesian principles can deepen our understanding of modern deep learning systems. While neural networks achieve remarkable predictive performance, their ability to generalize and to quantify uncertainty remains only partly understood. This thesis approaches this challenge from both methodological and theoretical angles: unifying Bayesian inference, function-space modeling, and large-deviation theory under a common probabilistic perspective.

On the methodological side, the thesis introduces the *Deep Variational Implicit Process* (DVIP), a scalable Bayesian framework that extends implicit processes to deep architectures. DVIP models distributions over functions that are easy to sample from but lack explicit densities, enabling expressive, non-Gaussian priors and efficient variational inference in function space. The model achieves competitive performance with deep Gaussian processes at a fraction of the computational cost. Complementing this, two post-hoc methods—the *Variational Linearized Laplace Approximation* (VaLLA) and the *Fixed-Mean Gaussian Process* (FMGP)—are proposed to equip pretrained deterministic networks with calibrated uncertainty estimates. Both approaches deliver well-calibrated predictions on large-scale tasks, bridging deterministic and Bayesian deep learning.

The theoretical contributions focus on one of the central open questions in modern machine learning: why do large, over-parameterized neural networks generalize so well? To address this, the thesis develops a unified probabilistic framework that connects three key mechanisms—*diversity*, *smoothness*, and *stochasticity*—within the language of PAC–Bayesian and large-deviation theory. The framework formalizes how ensemble diversity reduces generalization error by encouraging functional independence among predictors, and how smoothness—captured through the curvature of the loss landscape—can be interpreted as enlarging the rate function that governs the concentration of empirical loss. PAC–Chernoff bounds derived within this setting remain meaningful even in interpolation regimes, providing a quantitative, distribution-dependent explanation for double-descent behaviour. Finally, stochasticity in optimization, particularly in stochastic gradient descent (SGD), is analyzed as an implicit form of regularization. The noise introduced by mini-batch sampling acts as a probabilistic mechanism that biases the learning process toward flatter minima and more stable solutions. This connection between optimization dynamics and generalization, grounded in large-deviation principles, offers a probabilistic account of how random-ness and structure interact in deep learning. Altogether, this theoretical line of work unifies seemingly distinct explanations of generalization under a single mathematical framework, clarifying how probabilistic structure, model diversity, and stochastic training jointly shape predictive performance.

Taken together, these contributions provide both practical tools for scalable uncertainty estimation and theoretical insight into the probabilistic structure of deep learning. This thesis argues that reliable generalization and calibrated uncertainty emerge naturally when learning systems are viewed—and designed—through the lens of Bayesian reasoning.

**Keywords:** Bayesian deep learning, Gaussian process models, variational inference, implicit processes, uncertainty quantification, PAC–Bayesian generalization bounds, large-deviation analysis, stochastic optimization, implicit regularization, probabilistic learning theory

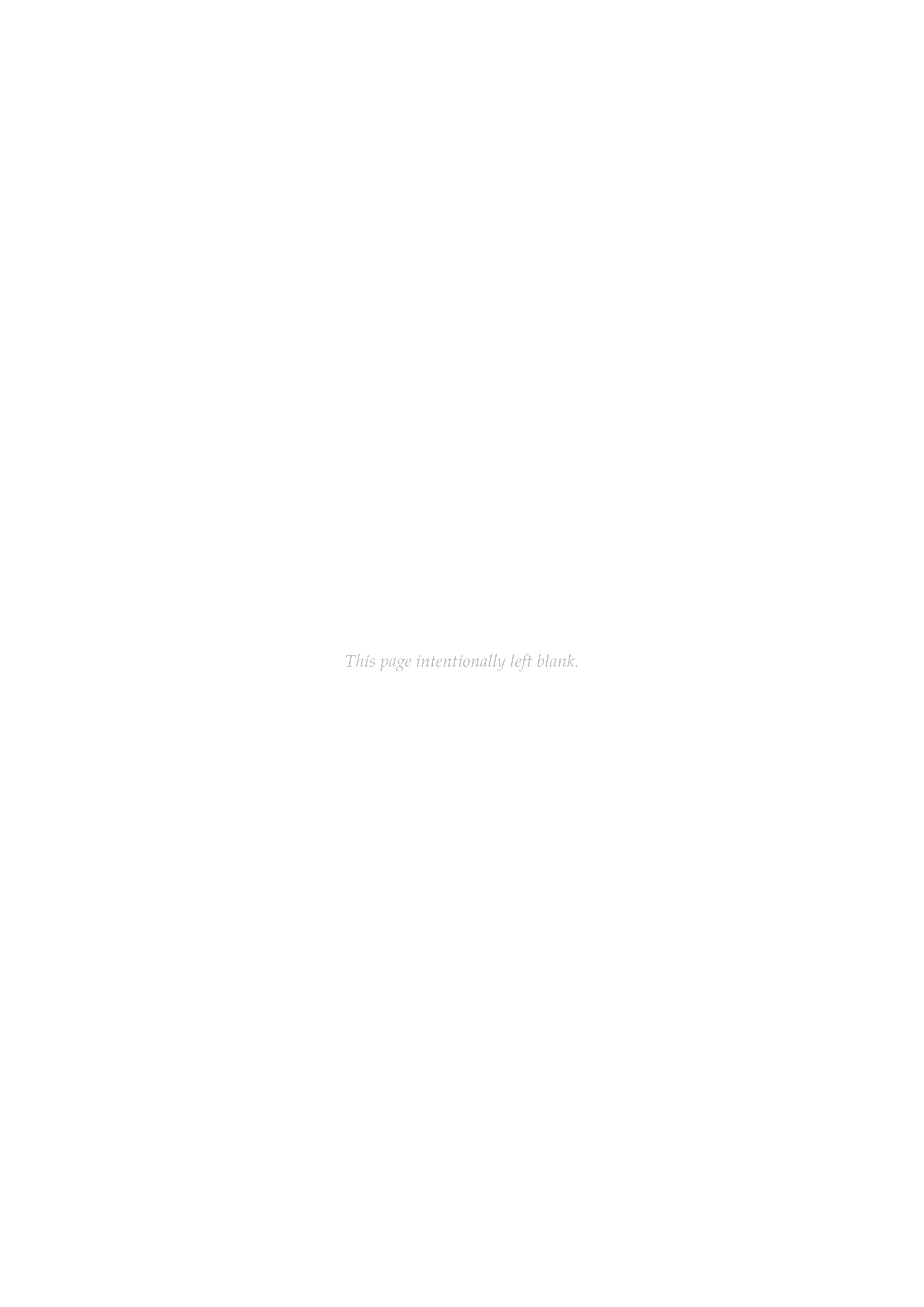
This page intentionally left blank.

# Contents

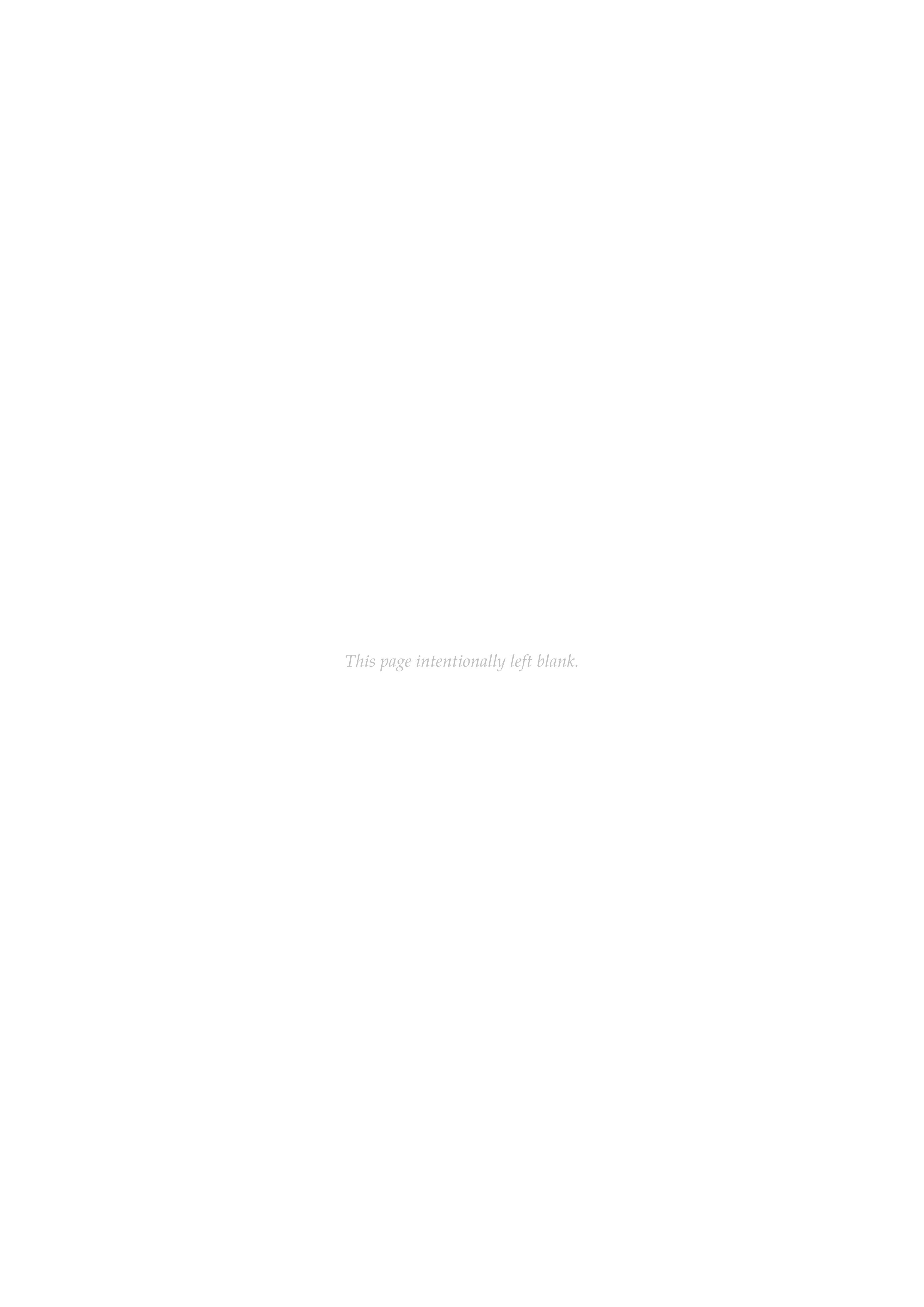
This page intentionally left blank.

# 1 Introduction

## 1.1 Thesis Outline and Contributions

Machine learning is a field of computer science that studies how to build algorithms able to *automatically* discover regularities in data (Bishop, 2006; Hastie et al., 2001). Throughout this thesis, such a regularity, or *pattern*, is understood as any recurrent structure that can be captured by a predictive rule with minimal human guidance. The focus of this thesis narrows to the supervised setting. A training dataset is denoted by

$$\mathcal{D} = \{(\mathbf{x}_1, y_1), \ldots, (\mathbf{x}_n, y_n)\} \subset \mathbb{R}^D \times \mathbb{R}^C , \tag{1.1}$$

where each $\mathbf{x}_i$ represents an input (attribute) vector and $y_i$ its associated target value. *Supervised learning* aims to infer an *unknown function* $f : \mathbb{R}^D \to \mathbb{R}^C$ that defines the underlying process of the observed data, that is, $y \approx f(\mathbf{x})$. The goal is to learn or approximate such mapping $f$. When $y$ ranges over a continuous domain (e.g. $\mathbb{R}$) the problem is termed *regression*; when $y$ takes values in a finite set such as $\{-1, 1\}$ it is called *classification*, and $y$ is then the *class label*.

In practice, a hypothesis space $\mathcal{F}$ is chosen, and the objective is to determine the element in such space that is closer to the unknown function $f$. The usual approach is to define a *parametric family of functions*, that is, a hypothesis space that solely depends on a set of vector parameters $\boldsymbol{\theta} \in \boldsymbol{\Theta} \subset \mathbb{R}^P$. An example of this characterization is Neural Networks (NNs), where once an architecture is fixed, a family of functions is defined for each configuration of parameters.

It is common to use the subscript notation to denote a predictor that is fully determined by its parameters as $f_{\boldsymbol{\theta}}$. The aim is then to find the set of parameters $\hat{\boldsymbol{\theta}}$, using the given samples $\mathcal{D}$ so that $f_{\boldsymbol{\theta}}$ approximates $f$ as closely as possible (Bishop, 1995; Hastie et al., 2001). The resulting predictor should not only reproduce the training targets but also *generalize*—that is, maintain its accuracy on new, unseen observations.

Two practical difficulties complicate this endeavor. Firstly, supervised datasets are often limited in size and may be corrupted by random noise (Bishop, 2006), which inflates the variance of any parameter estimate. Secondly, distinguishing intrinsic regularities from spurious fluctuations produced by chance is difficult. Consequently, selecting $\mathcal{F}$ is critical: if the class of candidate functions is too restrictive, the learned rule *under-fits* fails to capture genuine structure; if it is overly flexible, the rule *over-fits* compromises by adapting to noise. Both extremes deteriorate predictive performance on future data.

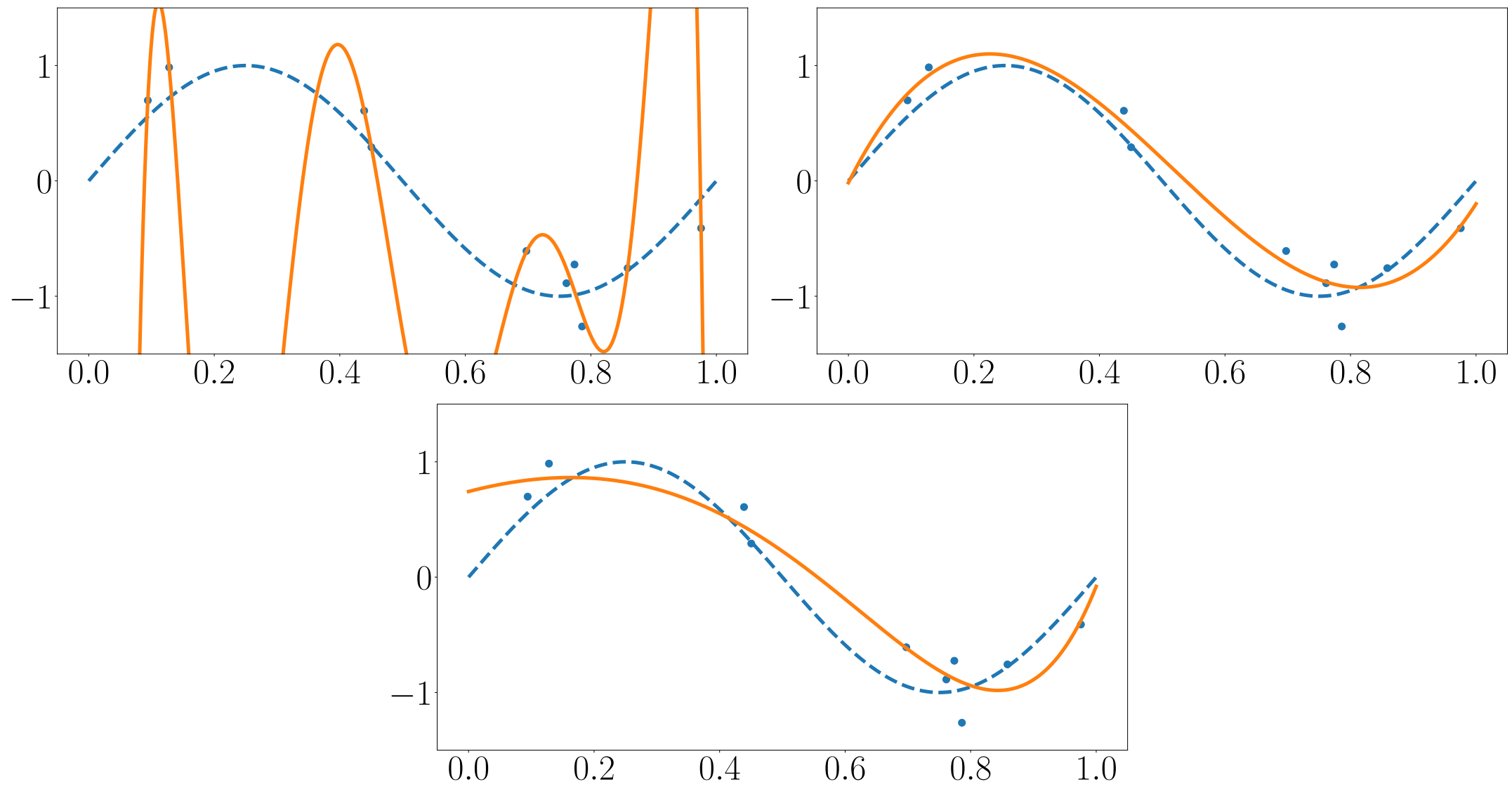

**Figure 1.1:** ***Illustrative polynomial curve–fitting experiment showing the perils of model complexity and the benefits of regularization.*** *Orange curves are fitted polynomials, blue dots noisy training samples, and dashed blue line the true function $f(x) = \sin(2\pi x)$.* Top left: *an unregularized ninth-degree polynomial interpolates the data but oscillates wildly, a classic instance of over-fitting.* Top right: *a cubic model ($M = 3$) captures the main trend yet misses some curvature, exhibiting mild under-fitting.* Bottom: *the same ninth-degree model with $\ell_2$ regularization ($\lambda > 0$) suppresses extreme oscillations, striking a better bias–variance balance.*

### Illustrative toy example

To make the notions of model complexity, underfitting, and overfitting concrete, a simple regression experiment is presented; see Figure 1.1. The task is to recover an unknown input–output relationship from examples. In supervised learning, a collection of input–output pairs

$$\mathcal{D} = \{(x_i, y_i)\}_{i=1}^{N} \tag{1.2}$$

is observed, where $x_i \in [0, 1]$ is an input (here, a scalar) and $t_i \in \mathbb{R}$ is a noisy observation of an underlying target function. In this example, the data are generated by

$$y = \sin(2\pi x) + \varepsilon, \qquad \varepsilon \sim \mathcal{N}(0, \sigma^2), \quad \sigma = 0.3, \qquad x \sim \mathcal{U}(0, 1). \tag{1.3}$$

The goal of learning is to produce a predictive rule $x \mapsto f(x)$ that maps new, unseen inputs to accurate outputs.

A widely used approach is to restrict attention to a family (*hypothesis class*) of candidate functions and to select within that family the function that best matches the observed data. Here the family of univariate polynomials of maximum degree $M$ is considered:

$$\mathcal{F}_M \;=\; \Big\{\, f_{\mathbf{w}}(x) \;=\; \sum_{m=0}^{M} w_m\, x^m \;\Big|\; \mathbf{w} \in \mathbb{R}^{M+1} \Big\}. \tag{1.4}$$

Where the set of parameters $\boldsymbol{\theta}$ correspond to the set of coefficients $\mathbf{w}$. Define the basis vector $\boldsymbol{\phi}(x) = (1, x, \ldots, x^M)^\top$ and the design matrix,

$$\boldsymbol{\Phi} \;=\; \big[\boldsymbol{\phi}(x_1)\ \cdots\ \boldsymbol{\phi}(x_N)\big]^\top \in \mathbb{R}^{N\times(M+1)}, \qquad \mathbf{y} = (y_1, \ldots, y_N)^\top. \tag{1.5}$$

The set of coefficients that minimizes the difference between predictions and true labels is the ordinary least-squares (OLS) estimator:

$$\hat{\mathbf{w}} \;=\; \arg\min_{\mathbf{w}} \|\boldsymbol{\Phi}\mathbf{w} - \mathbf{y}\|_2^2 \;=\; \big(\boldsymbol{\Phi}^\top\boldsymbol{\Phi}\big)^{-1}\boldsymbol{\Phi}^\top\mathbf{y}. \tag{1.6}$$

To mitigate sensitivity to noise and numerical instabilities when $M$ is large, a quadratic penalty (*ridge* or Tikhonov regularization) is commonly added:

$$\hat{\mathbf{w}} \;=\; \arg\min_{\mathbf{w}} \|\boldsymbol{\Phi}\mathbf{w} - \mathbf{y}\|_2^2 \;+\; \lambda\,\|\mathbf{w}\|_2^2 \;=\; \big(\boldsymbol{\Phi}^\top\boldsymbol{\Phi} + \lambda\mathbf{I}\big)^{-1}\boldsymbol{\Phi}^\top\mathbf{y}, \qquad \lambda > 0. \tag{1.7}$$

Intuitively, the penalty discourages overly large coefficients, which in turn discourages rapid oscillations of the fitted polynomial.

**Model complexity, under-fitting, and over-fitting.** The degree $M$ controls the expressive power (*capacity*) of the model class. Small $M$ yields simple functions that may be unable to capture the true sinusoidal shape; this is *under-fitting*. Large $M$ admits highly flexible polynomials that can interpolate the training data, including the noise; this is *over-fitting*. In the experiment (Figure 1.1), $M = 3$ produces a smooth curve that misses some curvature (underfitting), whereas $M = 9$ fits the $N$ points almost exactly and exhibits pronounced oscillations between them (overfitting). Introducing the $\ell_2$ penalty ($\lambda > 0$) or reducing $M$ produces a smoother curve that tracks the underlying $\sin(2\pi x)$ more faithfully and performs better on unseen inputs.

**Training vs. generalization.** Performance on the observed sample (the *training error*) does not necessarily reflect performance on new data (the *generalization error*). Overfitted models often achieve near-zero training error while performing poorly on new inputs because they have modeled the noise. Regularization ($\lambda > 0$) or reduced complexity ($M$ smaller) improves generalization by trading a small increase in training error for a larger decrease in prediction error on new data.

**Bias–variance trade-off.** This example illustrates the classical bias–variance trade-off. Low-degree polynomials have high bias (they cannot represent the sinusoid well) but low variance (predictions are stable across datasets), whereas high-degree polynomials have low bias (they can represent complex shapes) but high variance (predictions change markedly with the sampled noise). Regularization shifts the solution toward lower variance without excessively increasing bias, often yielding the lowest total prediction error.

**Interpretation.** Although polynomial regression in one dimension is deliberately simple, the same phenomena occur broadly in machine learning: richer models can interpolate, but careful control of complexity (via architecture choices, regularization, or data) is required to attain strong generalization.

### 1.1.1 Bayesian Machine Learning

In the Bayesian paradigm, probability is employed to encode a *prior* degree of confidence in each candidate hypothesis (Bishop, 2006; MacKay, 2003). Once a likelihood function has been posited—i.e. a mathematical description of how the observed targets could be generated—Bayes' rule transforms these priors into *posterior* probabilities in light of the training evidence. Prediction then amounts to averaging the individual hypotheses' outputs, each weighted by its posterior mass.

A convenient parameterization introduces a parameter vector $\mathbf{w}$ indexing the hypotheses. Let $\mathbf{X} = [\mathbf{x}_1, \ldots, \mathbf{x}_N]^\top \in \mathbb{R}^{N\times d}$ denote the design matrix of inputs and $\mathbf{y} = (y_1, \ldots, y_N)^\top \in \mathbb{R}^N$ the vector of targets. Given a prior $P(\mathbf{w})$ and a likelihood $P(\mathbf{y}|\mathbf{X}, \mathbf{w}) = \prod_{i=1}^N P(y_i|\mathbf{x}_i, \mathbf{w})$, Bayes' rule yields the posterior

$$P(\mathbf{w}|\mathcal{D}) = \frac{P(\mathbf{y}|\mathbf{X}, \mathbf{w})\, P(\mathbf{w})}{P(\mathbf{y}|\mathbf{X})} \propto P(\mathbf{y}|\mathbf{X}, \mathbf{w})\, P(\mathbf{w}), \qquad \mathcal{D} = \{(\mathbf{x}_i, y_i)\}_{i=1}^N. \tag{1.8}$$

The normalizing constant $P(\mathbf{t}|\mathbf{X}) = \int P(\mathbf{y}|\mathbf{X}, \mathbf{w})\, P(\mathbf{w})\, \mathrm{d}\mathbf{w}$ is the *marginal likelihood* (model evidence), central to Bayesian model comparison. A point estimate such as the maximum a posteriori (MAP) solution maximizes $\log P(\mathbf{w}|\mathcal{D}) = \log P(\mathbf{y}|\mathbf{X}, \mathbf{w}) + \log P(\mathbf{w}) + \text{const}$. Prediction averages over parameter uncertainty via the posterior predictive:

$$P(y^\star|\mathbf{x}^\star, \mathcal{D}) = \int p(y^\star|\mathbf{x}^\star, \mathbf{w})\, P(\mathbf{w} \mid \mathcal{D})\, \mathrm{d}\mathbf{w}, \tag{1.9}$$

which embodies the Bayesian "weighted averaging" described above.

Bayesian procedures offer a number of well-known benefits over many alternative learning strategies. First, because it penalizes unnecessarily elaborate models, the framework implements an intrinsic form of Occam's razor, thereby facilitating principled model selection (Bishop, 2006; MacKay, 2003). Second, assuming that an appropriate model class has been chosen, Bayesian inference retains the full distribution of plausible parameter values instead of collapsing uncertainty into a single point estimate; this built-in accounting for parameter variability often translates into more reliable out-of-sample predictions. Third, expert domain knowledge can be incorporated in a transparent way via the choice of prior, which is particularly advantageous when only limited training data are available.

These theoretical virtues come at a computational price. Exact posteriors typically require evaluating high-dimensional integrals or combinatorial sums, which are rarely feasible for realistic problems (Bishop, 2006; MacKay, 2003). Markov Chain Monte Carlo (MCMC) methods (Neal, 1993) sidestep the analytic intractability by drawing samples from a carefully constructed chain whose stationary distribution equals the desired posterior. However, long chains and careful convergence diagnostics render MCMC expensive in practice. A variety of faster, deterministic approximations—most notably variational techniques (Jaakkola, 2001; Thomas P. Minka, 2001)—replace the true posterior with a simpler, tractable family of distributions, though their applicability may be limited and certain parameter subsets can remain difficult to approximate. In such cases one often resorts to Type-II maximum likelihood (empirical Bayes), which optimizes those hard-to-integrate parameters by repeatedly invoking the approximate inference algorithm, thereby driving up the total training cost (Bishop, 2006).

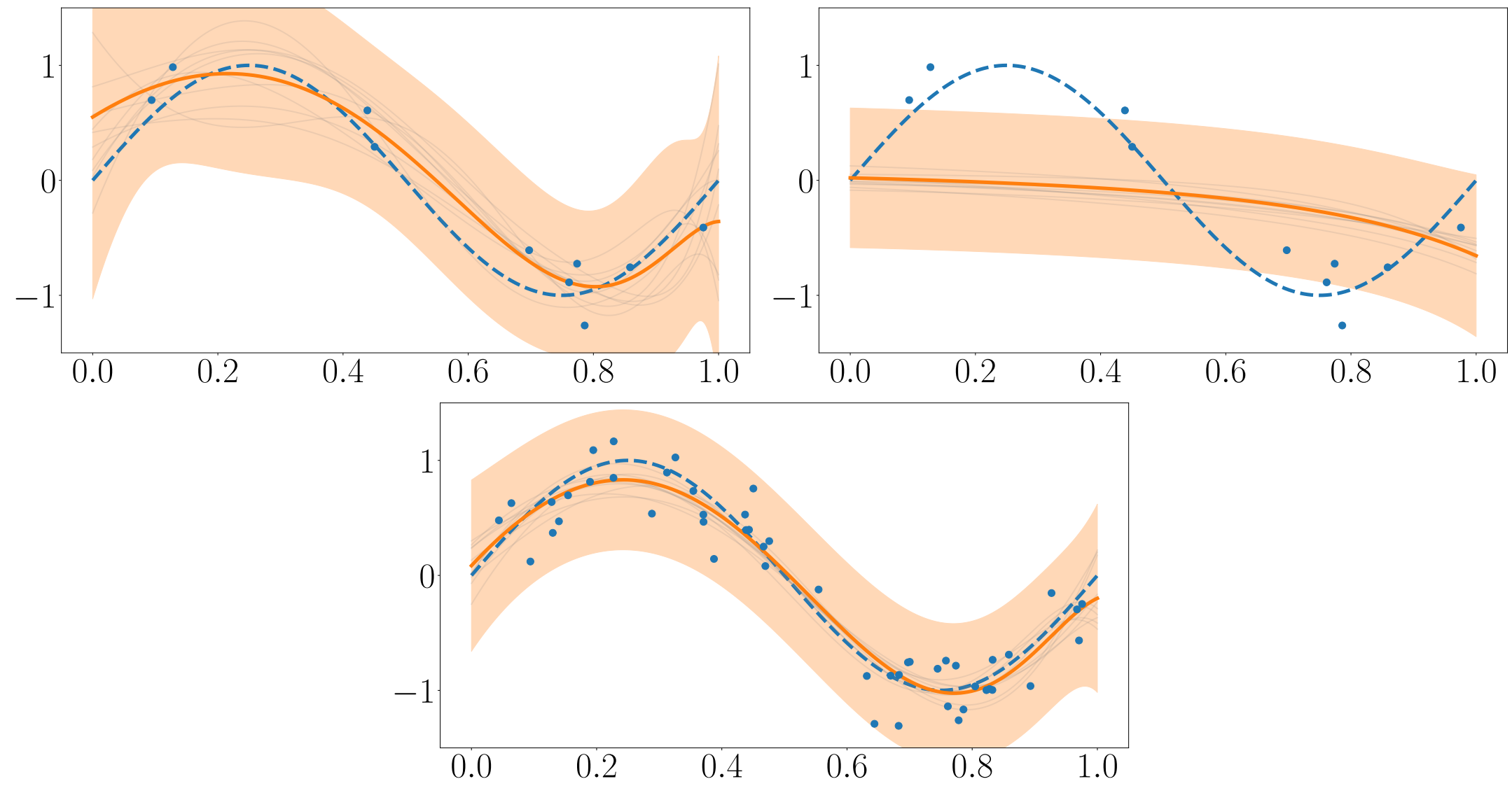

**Figure 1.2:** ***Interplay of model capacity, prior strength, and sample size in Bayesian polynomial regression.*** *Each panel shows the predictive mean (orange), a 95% credible band (shaded), ten posterior sample curves (thin grey), the noisy observations (blue dots), and the ground-truth function* $y = \sin(2\pi x)$ *(dashed blue).* Top left: *a twelfth-degree polynomial under an ultra-weak weight prior* ($\alpha = 0.001$) *over-fits dramatically, with oscillations between the sparse data points.* Top right: *restoring the ninth-degree model but tightening the prior to* $\alpha = 100$ *shrinks the coefficients toward zero; the predictive collapses toward a near-flat line and the credible band narrows markedly, illustrating prior dominance.* Bottom: *keeping the weak prior and ninth-degree model but increasing the sample size to* $N = 50$ *lets the likelihood override the prior; oscillations diminish and uncertainty contracts wherever data are abundant, demonstrating how evidence tempers over-fitting and reduces epistemic risk.*

### Bayesian Toy Study: Polynomial Regression

To visualize how Bayesian learning balances model complexity, prior beliefs, and data volume, the previous sinusoidal regression experiment is revisited under a Bayesian linear model; see Figure 1.2. Details on statistical properties, distributions and derivations can be found in Chapter 2. A zero-mean isotropic Gaussian prior encodes beliefs about coefficient magnitudes,

$$P(\mathbf{w}|\alpha) = \mathcal{N}(\mathbf{w}|\mathbf{0}, \alpha^{-1}\mathbf{I}), \tag{1.10}$$

where $\alpha > 0$ (the prior precision) controls the degree of shrinkage toward $0$. With Gaussian observations, Bayes' Theorem yields a Gaussian posterior:

$$P(\mathbf{w}|\mathbf{y}, \alpha, \sigma) = \mathcal{N}(\mathbf{w}|\mathbf{m}, \mathbf{S}_N), \quad \mathbf{S}^{-1} = \alpha\mathbf{I} + \sigma^{-2}\mathbf{\Phi}^\top\mathbf{\Phi}, \quad \mathbf{m} = \sigma^{-2}\mathbf{S}\mathbf{\Phi}^\top\mathbf{y}. \tag{1.11}$$

For a new input $x^*$, the predictive distribution integrates parameter uncertainty:

$$P(y^\star|x^\star, \mathcal{D}, \alpha, \sigma) = \mathcal{N}\left(y^\star|\boldsymbol{\phi}(x^\star)^\top\mathbf{m}, \sigma^2 + \boldsymbol{\phi}(x^\star)^\top\mathbf{S}\boldsymbol{\phi}(x^\star)\right). \tag{1.12}$$

The variance decomposes into irreducible (aleatoric) noise $\sigma^2$ and an epistemic term $\boldsymbol{\phi}(x^*)^\top\mathbf{S}_N\boldsymbol{\phi}(x^*)$ that shrinks as informative data accumulate or as the prior becomes more concentrated. The following settings highlight canonical Bayesian behaviors:

1. *High capacity with a weak prior:* a degree-$12$ polynomial with $\alpha = 10^{-3}$ (weak shrinkage). Excess flexibility combined with a lax prior allows the model to track noise, producing oscillatory fits and large predictive uncertainty—an instance of Bayesian over-fitting under weak regularization.
2. *Strong prior under limited data:* a degree-$9$ polynomial with $\alpha = 100$ (strong shrinkage). The prior heavily down-weights large coefficients, yielding a smoother fit but also noticeable bias when data are scarce; the prior dominates the likelihood, resulting in under-fitting.
3. *More data with the same strong prior:* the same degree-$9$ model and $\alpha = 100$, but trained on $5$ times as many samples. The increased sample size strengthens the likelihood relative to the prior, taming oscillations and reducing epistemic uncertainty; the posterior mean tracks the sinusoid more closely and the predictive bands contract.

The posterior covariance $\mathbf{S}$ and its induced predictive variance quantify the interplay between model capacity, prior strength, and data volume. Weak priors paired with high-capacity models can yield posterior distributions that admit wiggly functions consistent with noisy samples. Strong priors can curb such behavior but risk high bias when data are limited. Increasing $N$ concentrates the posterior around functions supported by the data, causing the likelihood to outweigh the prior and improving generalization while shrinking uncertainty. These effects are visible in Figure 1.2 through the smoothness of the posterior mean and the width of the predictive intervals.

Despite the elegance and conceptual appeal of Bayesian learning, applying it to modern deep networks presents major challenges. Exact inference is intractable, and existing approximations often collapse uncertainty or become computationally prohibitive. During this thesis, these issues motivated the exploration of scalable alternatives that preserve the core Bayesian principles of uncertainty quantification and regularization. In particular, this research has contributed new approaches that reformulate inference directly in function space—avoiding ill-posed parameter posteriors—and methods that retrofit uncertainty around pre-trained deterministic models without retraining. Together, these advances bring Bayesian reasoning closer to practical deployment in modern architectures, enabling calibrated predictions, better uncertainty estimates, and deeper theoretical understanding of Bayesian deep learning.

### 1.1.2 Generalization

A predictive model is regarded as useful only insofar as it *generalizes*, i.e. maintains high predictive accuracy on data that were *not* available during training (Bishop, 2006; Vapnik, 1998). The quest to understand and to control generalization therefore sits at the heart of statistical learning theory. Classical analyses treat the training sample as a random draw from an unknown generating distribution; learning is then framed as choosing, from a hypothesis class $\mathcal{F}$, a function whose expected risk is close to the minimum achievable risk within $\mathcal{F}$ (Shalev-Shwartz and Ben-David, 2014). Guarantees are expressed in terms of capacity measures such as the Vapnik–Chervonenkis (VC) dimension, Rademacher complexity, or covering numbers, all of which quantify how readily the hypothesis class can fit random noise.

From a practical standpoint, limited data and noisy observations translate into a *bias–variance* trade-off (Hastie et al., 2009). If the hypothesis space is too restrictive, the learned predictor exhibits high bias, failing to capture important structure; if the space is overly flexible, it possesses high variance, adapting to incidental fluctuations in the training set and thereby overfitting. Modern deep networks add a further twist: they can possess more parameters than training samples and still generalize well, a phenomenon sometimes called the "double descent" curve (Nakkiran et al., 2020; C. Zhang et al., 2017). Understanding these regimes has become an active area of contemporary research.

A repertoire of algorithmic techniques has emerged to foster generalization in practice. Regularization—whether explicit, as in $\ell_2$ or $\ell_1$ penalties, or implicit, as in the stochasticity of mini-batch gradient descent—constrains the effective capacity of the model. Data augmentation synthetically enlarges the sample, while cross-validation supplies an unbiased estimate of out-of-sample error for model selection (Bishop, 2006). Ensemble methods such as bagging and boosting aggregate multiple hypotheses, often reducing variance without a commensurate increase in bias. Collectively, these ideas form the methodological backbone that allows contemporary machine learning systems to move beyond the training corpus and perform robustly in the real world.

**A unifying visual example (Figure 1.3).** In the classical setting (left panel) test error follows the familiar U-curve: as model capacity grows, bias falls but variance rises, producing an optimal "sweet spot". In stark contrast, the right-hand panel shows that over-parameterized learners can suffer a sharp risk explosion at the interpola-tion threshold and yet *recover* low error once capacity increases further—the so-called double-descent phenomenon that has reignited interest in modern capacity measures. These contrasting behaviors underscore the need for a theory that simultaneously accounts for bias, variance, and the surprising generalization of highly over-parameterized models.

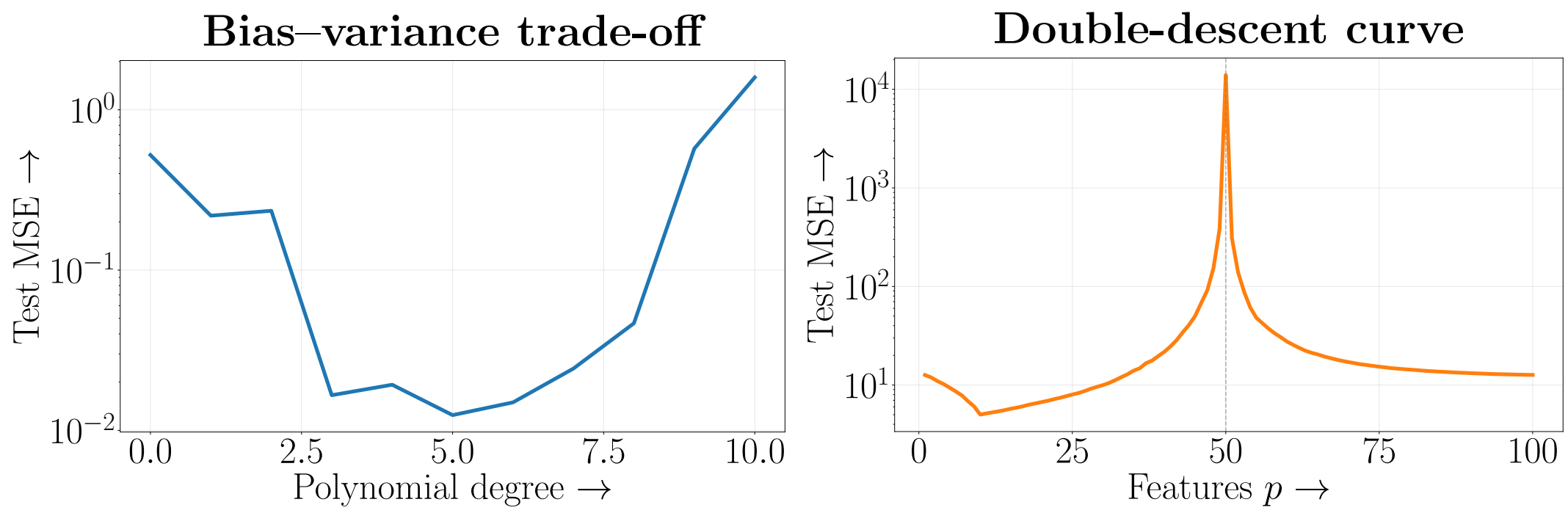

**Figure 1.3:** ***Two canonical generalization landscapes obtained from synthetic experiments.*** Left: *Average test mean–squared error (MSE) when fitting polynomials of degree* 0–10 *to* $N_{train} = 40$ *noisy samples from* $f(x) = \sin(2\pi x)$ $(\sigma = 0.1)$. *The characteristic U-shape illustrates the classical bias–variance trade-off: low-capacity models under-fit (high bias), while high-capacity models over-fit (high variance).* Right: *Test MSE for min-norm linear regression with* $p$ *random Gaussian features and a fixed training set of* $N_{train} = 50$ *examples corrupted by strong noise* $(\sigma = 2)$. *The pronounced spike exactly at* $p = N_{train}$ *marks the interpolation threshold; error falls again for* $p > N_{train}$, *producing the modern double-descent curve. Each point aggregates the results of 1 000 (left) or 200 (right) independent Monte-Carlo repeats.*

Despite substantial progress, fundamental questions remain open: what precise properties of optimization algorithms and network architectures explain the empirical generalization behavior of over-parameterized models? How can one derive capacity measures that capture those properties yet remain analytically tractable? This thesis contributes to that line of inquiry by proposing a novel theoretical framework that explains the relationship between ensemble generalization and diversity. Furthermore, a smoothness-based complexity term that remains non-vacuous even in the interpolation regime is introduced, allowing one to create a unifying explanation of a wide range of learning techniques used in modern machine learning.

### 1.1.3 Contributions

**C1. A Bayesian perspective for modern deep learning**

*Motivation and scope.* State-of-the-art deep networks are typically deployed as deterministic predictors that output point estimates. In many settings—risk-aware decision making, clinical or scientific inference, autonomous systems, and active data acquisition—the object of interest is a *predictive distribution* that quantifies both epistemic and aleatoric uncertainty. Obtaining such distributions for modern architectures is difficult for three reasons. First, exact Bayesian posteriors over the parameters of deep networks are analytically intractable and computationally prohibitive to approximate at scale. Second, common approximations (e.g., sim-ple variational families or naïve ensembles) can be miscalibrated, sensitive to optimization, and degrade accuracy if applied end-to-end. Third, parameter-space posteriors are often ill-posed surrogates for predictive uncertainty because many distinct parameter values induce near-identical functions, making inference geometry unfavorable.

*Approach and main results.* The difficulties outlined above are addressed by adopting a function–space formulation of learning and by developing inference procedures that remain tractable at modern scales.

- *Function-space variational inference via Deep Variational Implicit Processes (DVIP).* Deep networks are represented as compositions of implicit processes, which induces a stochastic process prior over functions without requiring explicit parametric densities in weight space. A variational family is specified directly over functions; an evidence lower bound is derived in function space; and unbiased stochastic estimators are constructed using pathwise sampling through the implicit layers. This yields calibrated predictive distributions and competitive accuracy while avoiding pathologies of parameter-space posteriors. (**Chapter 3**)
- *Diversity-aware ensemble objectives with theoretical support.* An upper bound on ensemble risk is proved that decomposes into (i) the $\rho$-average member error and (ii) a negative diversity term measuring disagreement among members' predictions. A PAC–Bayesian analysis then links the empirical diversity term to generalization error, yielding a regularized training objective that promotes beneficial diversity while preserving consistency guarantees across regression and classification losses. (**Chapter 5.2**)

Collectively, these developments provide routes to calibrated predictive distributions: either by performing inference directly in function space (DVIP) or by training ensembles with a diversity term justified by generalization bounds.

**C2. Post-hoc uncertainty for pre-trained networks**

*Motivation and scope.* In practical pipelines, a high-performing deterministic backbone is often already available after substantial engineering and training effort. Replacing this model with a bespoke Bayesian architecture is undesirable due to engineering cost, potential accuracy loss, and computational expense. The central need is a *post-hoc* mechanism that (i) attaches calibrated predictive uncertainty to an existing model without end-to-end retraining, (ii) preserves the backbone's accuracy and inductive biases, and (iii) scales to modern datasets and architectures (e.g., ImageNet-class problems and large molecular libraries). Traditional posterior approximations rarely meet these constraints simultaneously: Markov chain Monte Carlo is generally too slow, Laplace approximations may be local and unstable, and naïve temperature scaling or dropout-based methods can be poorly calibrated, especially under distribution shift. Methods are needed that retrofit uncertainty in a principled way while controlling computational and statistical trade-offs.

*Approach and main results.* Two post-hoc methodologies are introduced that attach Bayesian uncertainty to existing predictors while preserving accuracy and controlling computational cost.

- *Variational Linearized Laplace Approximation (VaLLA).* A linearized Laplace approximation is taken around a pre-trained network, producing a local Gaussian posterior in function space whose covariance encodes curvature information. This covariance is reinterpreted as a kernel for a sparse variational Gaussian process with decoupled inducing variables. The surrogate GP is trained to match the local behavior of the backbone, so that the predictive mean aligns with the base model while the predictive variance reflects local epistemic uncertainty.(**Chapter 4.2**)
- *Fixed-Mean Gaussian Processes (FMGPs).* The mean of the GP is fixed to the output of a pre-trained network; covariance hyperparameters (and inducing variables) are learned by maximizing a variational bound. This decoupling simplifies optimization, preserves the backbone's inductive bias, and enables transparent control of uncertainty through kernel choice. Predictive intervals are computed in closed form and remain well-calibrated under moderate distribution shift. Large-scale experiments demonstrate feasibility and strong calibration in computer vision and chem-informatics benchmarks. (**Chapter 4.3**)

Both approaches are architecture-agnostic, require no end-to-end retraining of the backbone, and provide calibrated uncertainty suitable for downstream decision-making under risk.

**C3. Distribution-dependent generalization for interpolating models**

*Motivation and scope.* Over-parameterized models that *interpolate* the training data routinely generalize well and can exhibit double-descent risk curves, yet classical uniform-convergence bounds (e.g., VC/Rademacher) often become vacuous in these regimes. Two gaps arise. First, capacity-only bounds ignore structure in the data-generating distribution that strongly influences generalization, leading to bounds that fail to tighten with increasing sample size. Second, widely used practices—$\ell_2$ penalties, distance-from-initialization and input-gradient constraints, architectural invariances, data augmentation, and ensemble diversity—lack a unified theory that quantifies when they reduce test error at or beyond the interpolation threshold. A *distribution-dependent* framework is needed that (i) remains informative for interpolators, (ii) yields a complexity measure aligned with observed concentration of empirical losses, and (iii) explains why and when added capacity or induced invariances *improve* generalization rather than harm it.

*Approach and main results.* A distribution-dependent analysis is developed that re-mains informative at interpolation and unifies several empirical practices through a single complexity measure.

- *PAC–Chernoff bounds that remain tight at interpolation.* A bound is derived that upper-bounds the population risk in terms of the inverse of a *rate function* governing the large-deviation behavior of the empirical loss. Unlike uniform-convergence bounds, the resulting guarantee depends on the data-generating distribution and is *exact* (perfectly tight) for any interpolator. This property holds even in over-parameterized regimes, thereby covering models at or beyond the interpolation threshold. The rate function quantitatively captures the concentration of the empirical loss and explains observed double-descent curves. (**Chapter 5.3**)
- *Smoothness via the rate function and consequences for regularization.* The inverse rate function is used as a smoothness measure for interpolators. It is shown that standard regularizers—$\ell_2$ penalties, constraints on distance from initialization, and control of input-gradient norms—act as proxies that decrease the inverse rate function, thereby lowering the bound on test error at interpolation. (**Chapter 5.3.6**)
- *Data augmentation and architectural invariances.* The analysis shows that augmentation schemes can increase the rate function (i.e., increase concentration of the empirical loss), yielding smaller generalization error for interpolators under the PAC–Chernoff bound. Analogously, architectures with built-in invariances (e.g., convolutional networks for translations) are shown to avoid the concentration degradation induced by transformed inputs, again improving the bound. Conditions are identified under which these mechanisms are effective and when departures from ideal group structure still confer benefits. (**Chapter 5.3.7**)
- *Over-parameterization requirements and double descent.* Conditions are derived under which additional parameters are required to realize smoother interpolators (in the rate-function sense). This explains why, beyond the interpolation threshold, enlarging the model can reduce test error and accounts for the decreasing branch of double descent through improved concentration properties captured by the rate function. (**Chapters 5.3.8 and 5.3.5**)
- *Implicit bias of stochastic gradient descent.* A large-deviation perspective is introduced to analyze the implicit bias of stochastic gradient descent (SGD). It is shown that the stochasticity inherent to mini-batch updates induces an implicit regularization effect, promoting convergence toward flat minima that exhibit higher concentration of the empirical loss and, consequently, better generalization. The analysis connects the noise statistics of SGD to the rate function of the loss landscape, providing a distribution-dependent explanation of why SGD tends to favor solutions with superior generalization properties. (**Chapter 5.4**)

The resulting framework specifies when interpolators generalize, provides theory-backed training objectives (including diversity-aware ensembles), and guides the design of regularization, augmentation, and architectural inductive biases that improve test performance at and beyond interpolation.

## 1.2 Full list of Publications

This section provides a comprehensive list of the publications produced during my PhD.

### 1.2.1 Main publications

In this subsection, I present the principal articles that form the core contributions of this dissertation.

1. **Ortega, L.A.**, Cabañas, R. &, Masegosa, A.. (2022). Diversity and Generalization in Neural Network Ensembles. *Proceedings of The 25th International Conference on Artificial Intelligence and Statistics* (CORE A), in *Proceedings of Machine Learning Research* 151:11720-11743. Available from https://proceedings.mlr.press/v151/ortega22a.html.
   Detailed in Chapter 5.2.
2. **Ortega, L.A.**, Rodríguez-Santana, S., & Hernández-Lobato, D.. (2023). Deep Variational Implicit Processes. *The Eleventh International Conference on Learning Representations* (CORE A+). Available from https://iclr.cc/virtual/2023/poster/12053.
   Detailed in Chapter 3.
3. **Ortega, L.A.**, Rodriguez Santana, S., & Hernández-Lobato, D.. (2024). Variational Linearized Laplace Approximation for Bayesian Deep Learning. *Proceedings of the 41st International Conference on Machine Learning* (CORE A+), in *Proceedings of Machine Learning Research* 235:38815-38836. Available from https://proceedings.mlr.press/v235/ortega24a.html.
   Detailed in Chapter 4.2.
4. Masegosa, A., & **Ortega, L.A.**. (2025). PAC-Chernoff Bounds: Understanding Generalization in the Interpolation Regime. *Journal of Artificial Intelligence Research* (Q2). Presented at the *European Conference on Artificial Intelligence* (CORE A) as a spotlight presentation. Available from https://doi.org/10.1613/jair.1.17036.
   Detailed in Chapter 5.3.
5. **Ortega, L.A.**, Rodriguez Santana, S., & Hernández-Lobato, D.. (2025). Fixed-mean Gaussian Processes for ad-hoc Bayesian Deep Learning. Submitted to Journal on Machine Learning Research. Available from https://arxiv.org/abs/2412.04177.
   Detailed in Chapter 4.3.

6. **Ortega, L.A.** & Masegosa, A.. (2025). A Large Deviation Theory Analysis on the Implicit Bias of SGD. Neurocomputing (Q1). Available from https://www.sciencedirect.com/science/article/abs/pii/S0925231226003577.
   Detailed in Chapter 5.4.

### 1.2.2 Other Publications

In this subsection, I present other publications produced during my PhD that are not directly included as part of this dissertation.

1. Zhang, Y., Wu, Y., **Ortega, L.A.**, & Masegosa, A.. (2024). The Cold Posterior Effect Indicates Underfitting, and Cold Posteriors Represent a Fully Bayesian Method to Mitigate It. *Transactions for Machine Learning Research*. Available from https://openreview.net/forum?id=GZORXGxHHT.
2. Casado, I., **Ortega, L.A.**, Pérez, A., & Masegosa, A.. (2024). PAC-Bayes-Chernoff bounds for unbounded losses. *The Thirty-eighth Annual Conference on Neural Information Processing Systems* (CORE A+). Available from https://openreview.net/forum?id=CyzZeND3LB.
3. **Ortega, L.A.**, Rodriguez Santana, S., & Hernández-Lobato, D.. (2025). Scalable Linearized Laplace Approximation via Surrogate Neural Kernel. Submitted to ESANN 2026 (CORE B).

## 1.3 Chapters Summary

**Chapter 2 – Bayesian Inference, Gaussian Processes and Generalization Bounds.** This chapter develops the mathematical and probabilistic foundations required for the thesis. It begins with a measure-theoretic treatment of probability, introducing $\sigma$-algebras, probability measures, conditional probability, independence, and random variables. Bayesian inference is then presented as a principled framework for uncertainty estimation, with emphasis on approximate inference techniques such as variational inference, the mean-field family, and the black-box $\alpha$-energy objective. Gaussian processes (GPs) are introduced as non-parametric Bayesian models for functions, covering their predictive distributions, hyper-parameter learning, scalability via inducing-point approximations and random features, and a dual interpretation through Gaussian measures in Hilbert spaces. The chapter concludes with generalization theory in the form of Probably Approximately Correct (PAC) bounds, contrasting classical uniform convergence results with the more general PAC–Bayesian framework, which provides distribution-dependent guarantees. Together, these elements establish the theoretical backbone for the subsequent chapters on scalable Bayesian deep learning and generalization in modern neural networks.

**Chapter 3 – Deep Variational Implicit Processes.** This chapter introduces *Deep Variational Implicit Processes* (DVIPs), a flexible Bayesian framework for uncertainty-aware deep learning. Building upon the concept of implicit processes (IPs), which generalize Gaussian processes by defining priors through "samplable" but density-free stochastic processes, DVIPs extend variational implicit processes (VIPs) to multi-layer hierarchies. Each layer employs IP-based priors approximated by GPs, enabling scalable variational inference with Monte Carlo sampling and stochastic optimization. Unlike VIPs, DVIPs yield non-Gaussian predictive distributions, improving flexibility and calibration. Extensive experiments on regression benchmarks, image classification, and large-scale datasets demonstrate that DVIPs match or surpass deep GPs while being more computationally efficient. The chapter highlights the role of depth, prior adaptation, and domain-specific priors (e.g., CNN-based) in enhancing expressivity and predictive performance.

**Chapter 4 – Post-hoc Uncertainty Estimation for Pre-trained Networks.** This chapter addresses post-hoc uncertainty estimation for pre-trained deep networks through two complementary approaches: the Variational Linearized Laplace Approximation (VaLLA) and Fixed-Mean Gaussian Processes (FMGPs). VaLLA reformulates the linearized Laplace approximation (LLA) in function space, recasting its covariance as a GP kernel and enabling scalable surrogates with sparse variational GPs. FMGPs take this further by treating any pre-trained model as the deterministic mean of a GP, training only the covariance while keeping the mean fixed, thereby converting deterministic predictors into calibrated Bayesian ones. Both methods allow uncertainty estimation without retraining the original network and scale to large datasets such as ImageNet or molecular property prediction tasks. Experiments confirm that these approaches achieve competitive calibration with modest computational overhead, establishing practical recipes for retrofitting uncertainty into modern deep learning models.

**Chapter 5 – Generalization in Neural Networks.** Brings together two complementary lines of research that tackle why modern deep-learning models can generalize even when they contain far more parameters than training points.

- **Chapter 5.2**. Although practitioners know that ensembles work best when their members err differently, there has been no consensus definition of "diversity" nor a theory that links it to test error in deep nets. A unified diversity measure is introduced and, via an upper-bound decomposition, show how ensemble error splits into two terms: average individual error and a negative contribution from diversity. A subsequent PAC-Bayesian analysis makes this link rigorous and distribution-dependent, revealing how correlation between members controls the bound. Experiments on CIFAR-10/100 classifiers and a Wine-Quality regressor confirm that higher measured diversity correlates with larger "gap" between the mean member loss and the ensemble loss, exactly as the theory predicts.
- **Chapter 5.3**. Classic VC- or Rademacher-style bounds become vacuous for modern, over-parameterized interpolators. The chapter, therefore, seeks bounds that depend on the data-generating distribution rather than the finite training sample.

The result is tight for any model that perfectly fits the training set. It reproduces the double-descent curve by showing how widening a network can increase the rate function after the interpolation threshold, thus lowering test error.
Classical regularizers ($\ell_2$ decay, distance-from-init, input-gradient penalties) as well as data-augmentation and invariant architectures all act by enlarging the rate function, i.e. making the model smoother.

- **Chapter 5.4**. Explores the implicit bias of stochastic gradient descent (SGD) through the lens of large-deviation theory. It demonstrates that the stochasticity of mini-batch updates induces an implicit regularization effect that biases learning toward flatter minima with higher loss concentration, thereby improving generalization. The analysis connects the noise covariance of SGD to the rate function governing the loss landscape, offering a distribution-dependent explanation of why SGD tends to converge to solutions with superior generalization performance and how batch size, learning rate, and noise structure influence this behaviour.

**Appendix A — Miscellaneous.** This section provides the precise synthetic setups behind Figure 1.3 and collects closed-form Gaussian identities that will be used later in the thesis.

**Appendix B — Deep Variational Implicit Processes: full regression results.** Gathers the complete UCI regression tables and additional figures underpinning Chapter 3 (VIP, DVIP with varying depth, sparse GPs, and deep GPs). Exact values referenced in the chapter as Table B.1 are reported here.

**Appendix C — Mathematical Proofs.** Collects formal statements and proofs for all theoretical results in all the chapters of this dissertation.

**Appendix D — *BayesiPy*: Post-hoc Bayesian Inference for Pre-trained Neural Networks.** Introduces *BayesiPy*, a Python library that wraps a trained backbone with a Bayesian posterior approximation to obtain calibrated predictive means and uncertainties without end-to-end retraining. It catalogues the included methods in the library—Linearised Laplace (LLA), Accelerated LLA (ELLA), Variational LLA (VaLLA), mean-field variational inference (MFVI), Spectral-Normalised GP (SNGP), and Fixed-Mean GP (FMGP)—and provides usage examples (e.g., FMGP over a PyTorch model), highlighting when each approach is preferable

# 2

# Bayesian Inference, Gaussian Processes and Generalization Bounds

In this chapter, the basics of Bayesian inference, Gaussian processes, and theoretical bounds for the generalization error of deep learning models are introduced.

## 2.1 Probability–Theoretic Foundations

The statistical methodologies developed in this thesis rest on the rigorous measure–theoretic formulation of probability introduced by Kolmogorov (1956) and further elaborated in standard monographs such as Billingsley (1995) and Durrett (2019). In this chapter, the key concepts and notational conventions that will be employed throughout this dissertation are introduced. The presentation is self–contained but deliberately concise; readers seeking a comprehensive treatment are referred to the aforementioned texts as well as Klenke (2013) for additional details.

### 2.1.1 Measurable Spaces and Probability Measures

Let $\Omega$ denote the *sample space*: the collection of all possible outcomes of a well–defined experiment or data–generating mechanism. A subset $A \subseteq \Omega$ is called an *event*. In order to assign probabilities consistently, a suitable family of events that is stable under the usual set operations is required.

**Definition 2.1** ($\sigma$–algebra)**.** Let $\mathcal{P}(\Omega)$ be the power set of $\Omega$. A non–empty collection $\mathcal{F} \subseteq \mathcal{P}(\Omega)$ is called a *$\sigma$–algebra* if

1. $\Omega \in \mathcal{F}$; (contains the whole space)
2. $A \in \mathcal{F} \;\Rightarrow\; A^{c} \in \mathcal{F}$; (closed under complementation)
3. $\{A_n\}_{n\in\mathbb{N}} \subset \mathcal{F} \;\Rightarrow\; \bigcup_{n\in\mathbb{N}} A_n \in \mathcal{F}$. (closed under countable unions)

The pair $(\Omega, \mathcal{F})$ is called a *measurable space*. It follows from the definition that $\emptyset \in \mathcal{F}$ and that $\mathcal{F}$ is closed under countable intersections.

**Definition 2.2** (Probability measure)**.** Given a measurable space $(\Omega, \mathcal{F})$, a mapping $P : \mathcal{F} \to [0, 1]$ is called a *probability measure* if

1. **Non–negativity:** $P(A) \geq 0$ for all $A \in \mathcal{F}$;
2. **Normalization:** $P(\Omega) = 1$;
3. **Countable additivity:** for any countable collection of pairwise disjoint events $\{A_n\}_{n\in\mathbb{N}} \subset \mathcal{F}$, it verifies that $P\big(\bigcup_{n\in\mathbb{N}} A_n\big) = \sum_{n\in\mathbb{N}} P(A_n)$.

The triple $(\Omega, \mathcal{F}, P)$ is termed a *probability space*. From the axioms, one immediately derives $P(\emptyset) = 0$ and the inclusion–exclusion identity $P(A \cup B) = P(A) + P(B) - P(A \cap B)$ for all $A, B \in \mathcal{F}$.

### 2.1.2 Conditional Probability and Independence

**Definition 2.3** (Conditional probability)**.** Let $A, B \in \mathcal{F}$ with $P(B) > 0$. The *conditional probability* of $A$ given $B$ is defined by

$$P(A|B) = \frac{P(A \cap B)}{P(B)}. \tag{2.1}$$

**Theorem 2.4** (Bayes' theorem)**.** *For any $A, B \in \mathcal{F}$ with $P(B) > 0$,*

$$P(A|B) = \frac{P(B|A)P(A)}{P(B)}. \tag{2.2}$$

Bayes' theorem underpins modern Bayesian statistics (Gelman et al., 2013) by enabling the systematic update of prior beliefs $P(A)$ in the light of new evidence $B$.

**Definition 2.5** (Independence)**.** Events $A, B \in \mathcal{F}$ are called *independent* if $P(A \cap B) = P(A)P(B)$, equivalently $P(A|B) = P(A)$ whenever $P(B) > 0$. More generally, $A$ and $B$ are *conditionally independent* given a third event $C$ with $P(C) > 0$ if $P(A \cap B|C) = P(A|C)P(B|C)$; this is noted as $A \perp B \mid C$.

### 2.1.3 Random Variables

A random variable formalizes the idea of a numerical quantity generated by a random experiment. Following Athreya and Lahiri (2006), the measure–theoretic definition is adopted.

**Definition 2.6** (Random variable)**.** Let $(\Omega, \mathcal{F}, P)$ be a probability space and $(E, \mathcal{E})$ a measurable space. A measurable function $X : \Omega \to E$ is called a *random variable*. The *push–forward* measure $P_X := P \circ X^{-1}$ is the *distribution* of $X$.

Throughout the thesis, random variables are denoted by uppercase letters (e.g., $X$), whereas realizations or deterministic values are rendered in lowercase (e.g., $x$). Boldface symbols (e.g., $\mathbf{X}$) indicate random vectors. Usually, $\boldsymbol{\theta}$ denotes the full parameter vector (or collection of parameters), while $\theta$ denotes a single scalar parameter—either the parameter of a single-parameter model or an individual component $\theta_j$ of $\boldsymbol{\theta}$.

#### 2.1.3.1 Distribution Functions

**Definition 2.7** (Cumulative distribution function)**.** For a real–valued random variable $X$, the mapping $F_X : \mathbb{R} \to [0, 1]$ given by $F_X(x) = P(X \leq x)$ is called the *cumulative distribution function* (CDF).

If $X$ takes values in a countable set, it is termed *discrete* and its probability mass function is $p_X(x) = P(X = x)$. When $X$ has an uncountable support and there exists a density $f_X$ with respect to Lebesgue measure such that $F_X(x) = \int_{-\infty}^{x} f_X(u)\,\mathrm{d}u$, $X$ is said to be *continuous* with *probability density function* $f_X$. Mixed–type variables can be represented as combinations of discrete and continuous components (Grimmett and Stirzaker, 2001).

For the rest of this thesis, the integral notation will be used uniformly for discrete and continuous cases. For discrete, the integral is understood with respect to the *counting measure* $\#(\mathrm{d}x) = \sum_{n \in \mathcal{I}} \delta(x - n)\,\mathrm{d}x$.

#### 2.1.3.2 Joint, Marginal, and Conditional Distributions

Let $X$ and $Y$ be random variables on the same probability space. The *joint distribution* $P_{X,Y}$ is defined on the product $\sigma$–algebra and satisfies $P_{X,Y}(A \times B) = P(X \in A, Y \in B)$. The *marginal* of $X$ is obtained by integration (or summation) over $Y$:

$$P_X(A) = \int_y P_{X,Y}(A, \mathrm{d}y). \tag{2.3}$$

Conditional distributions $P_{X|Y}$ are defined analogously whenever $P_Y$ assigns positive probability to the conditioning event or set. Two random variables are *independent* if $P_{X,Y} = P_X P_Y$; conditional independence given a third variable $Z$ is denoted by $X \perp Y \mid Z$ and is central to graphical–model semantics (Lauritzen, 1996).

A sequence $\{X_n\}_{n=1}^N$ is said to be *independent and identically distributed* (*i.i.d.*) if the joint CDF factorizes $F_{X_1,\ldots,X_N}(x_1, \ldots, x_N) = \prod_{n=1}^N F_{X_1}(x_n)$. Random vectors are collected into column form $\mathbf{X} = (X_1, \ldots, X_N)^T$; no ambiguity should arise between vector notation and sets of variables.

## 2.2 Bayesian Inference

Effective generalization and reliable uncertainty estimation remain pivotal challenges in contemporary machine learning research. Bayesian inference addresses these challenges by systematically incorporating uncertainty through probabilistic modeling (Bishop, 2006; MacKay, 2003). In Bayesian supervised learning, observed data are linked to model parameters via a likelihood function, while prior distributions quantify initial beliefs about these parameters. Using Bayes' theorem, these beliefs are updated in light of empirical evidence, yielding posterior distributions over parameters. However, exact Bayesian inference is generally infeasible for complex models, prompting the adoption of approximate methods (Blei et al., 2017; Neal, 1993).

In contemporary machine learning, a Bayesian interpretation of probability offers a principled way to capture both our ignorance about the true model and the uncertainty in its parameters. Consider a model governed by parameters $\boldsymbol{\theta}$ and a training set $\mathcal{D}$. Equipped with a prior distribution $P(\boldsymbol{\theta})$, Bayes' rule transforms this prior into the posterior distribution that incorporates the information contained in the data:

$$P(\boldsymbol{\theta}|\mathcal{D}) = \frac{P(\mathcal{D}|\boldsymbol{\theta})\, P(\boldsymbol{\theta})}{P(\mathcal{D})}\,, \tag{2.4}$$

where $P(\mathcal{D}|\boldsymbol{\theta})$ is the *likelihood*, measuring how well a particular choice of parameters explains the observed data, and

$$P(\mathcal{D}) = \int P(\mathcal{D}|\boldsymbol{\theta})P(\boldsymbol{\theta})\, \mathrm{d}\boldsymbol{\theta}\,, \tag{2.5}$$

is the marginal likelihood (or evidence) ensuring that the posterior is a proper probability distribution.

**Beyond parameter estimation** Once computed, the posterior can serve multiple roles: it yields predictive distributions for new data, supplies calibrated credible intervals, and underpins hierarchical extensions in which additional latent variables are introduced to simplify computation or enrich the model (Bishop, 2006; MacKay, 2003).

### 2.2.1 Bayesian vs. Frequentist Viewpoints

This Bayesian treatment differs fundamentally from a frequentist approach (Hastie et al., 2001). In frequentist statistics, the parameters are considered fixed but unknown; one first produces a point estimate and then quantifies its sampling variability. In the Bayesian framework, by contrast, $\boldsymbol{\theta}$ is regarded as uncertain, and the entire posterior expresses our updated beliefs after observing $\mathcal{D}$. Importantly, assigning a distribution to $\boldsymbol{\theta}$ is a statement about *epistemic* uncertainty, not a claim that the parameters physically fluctuate at random.

*Epistemic* (model) uncertainty arises from limited data, incomplete knowledge, or model misspecification; it reflects uncertainty about the parameters or the hypothesis class and is, in principle, reducible with additional informative data or a better model. It tends to be large in regions of the input space that are underrepresented (including out-of-distribution inputs) and is commonly quantified by variability across plausible models, e.g., Bayesian posterior dispersion or the spread of predictions from deep ensembles. By contrast, *aleatoric* (data) uncertainty stems from intrinsic randomness in the data-generating process—such as measurement noise, label noise, or inherent class overlap—and is irreducible even with infinite data. Aleatoric uncertainty may be *homoscedastic* (constant across inputs) or *heteroscedastic* (input-dependent), and is typically modeled via an explicit noise/variance term in the likelihood or through distributional/quantile losses. Conceptually, total predictive uncertainty decomposes into these two components (via the law of total variance): epistemic captures uncertainty

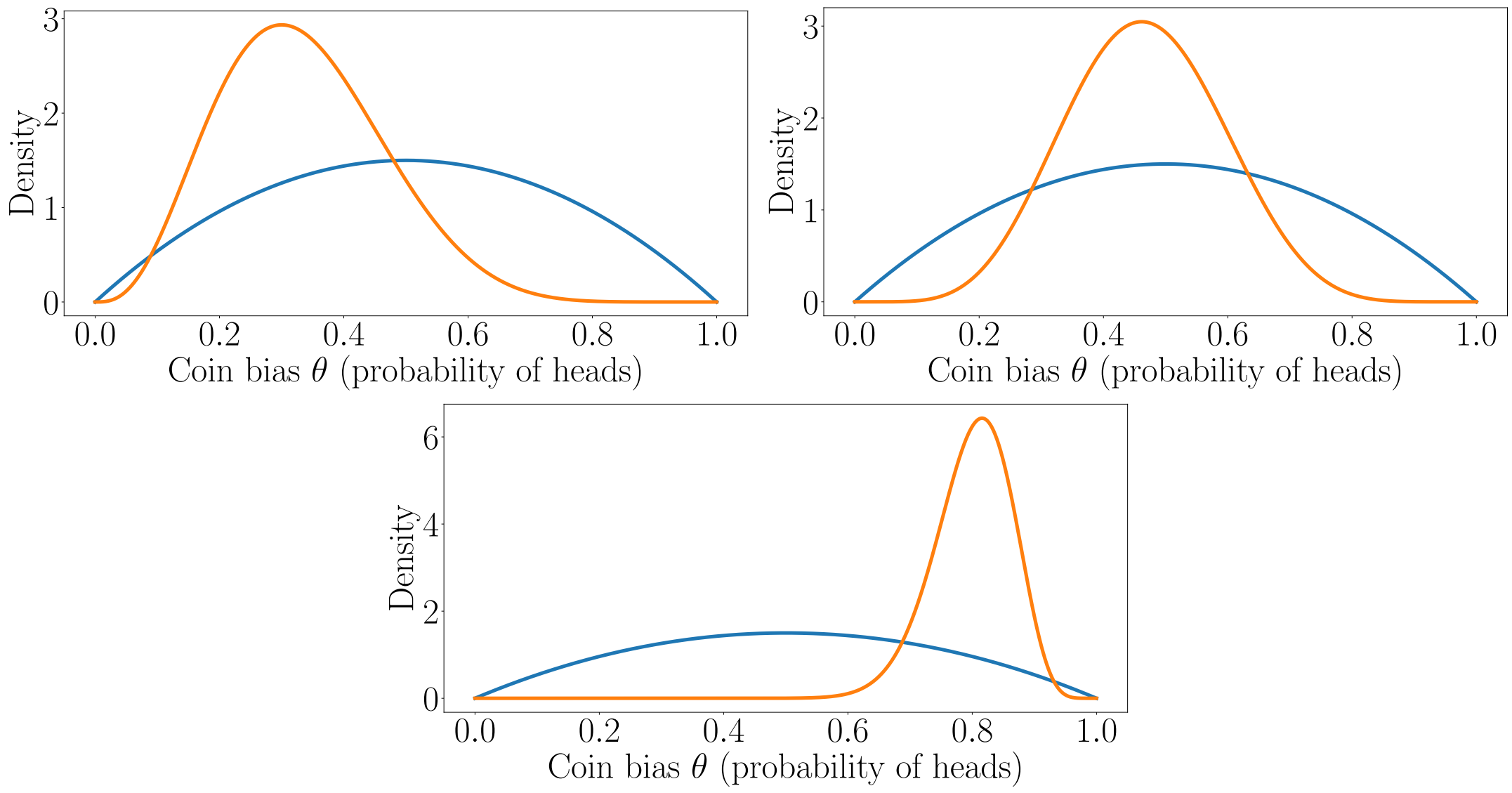

**Figure 2.1:** ***Sequential Bayesian updates of a coin–bias model.*** *Starting from the same mild* prior Beta$(2, 2)$ (*blue curve, identical in all panels*), *the figure shows how the* posterior *distribution tightens and shifts as more heads H are observed while the number of tails remains fixed at 6.* (*Top left*) *After only* $H = 2$ (Beta$(4, 8)$, *mean* $\approx 0.33$) *the posterior is still broad and centered below* 0.5. (*Top right*) *With* $H = 5$ (Beta$(7, 8)$, *mean* $\approx 0.47$) *the mass has moved toward fairness and the variance has decreased.* (*Bottom*) *With* $H = 30$ (Beta$(32, 8)$, *mean* $\approx 0.80$) *the distribution is peaked, reflecting increased confidence in a head–biased coin. The sequence illustrates the Bayesian principle that accumulating evidence both* re-positions *the belief about the parameter and* reduces *uncertainty.*

about the model, while aleatoric captures uncertainty in the observations themselves.

The Bayesian treatment of a simple Bernoulli process is described next. Let $\mathcal{D}$ denote a dataset of $n$ independent coin tosses $\{x_1, \ldots, x_n\}$, where each $x_i \in \{0, 1\}$ equals 1 ("heads") with probability $\theta$ and 0 ("tails") otherwise. The likelihood of the entire sequence, conditioned on the fixed but unknown bias of the coin $\theta$, is

$$P(\mathcal{D}|\theta) = \prod_{i=1}^{n} \theta^{x_i} (1-\theta)^{1-x_i} = \theta^k (1-\theta)^{n-k}, \tag{2.6}$$

where $k = \sum_{i=1}^{n} x_i$ is the number of heads.

A prior distribution is said to be *conjugate* to a likelihood family when, after observing any data sample, the resulting posterior distribution belongs to the same parametric family as the prior, differing only in its hyper-parameters. A conjugate prior for $\theta \in [0, 1]$ is the Beta distribution, $\theta \sim \text{Beta}(\alpha_0, \beta_0)$, with density proportional to $\theta^{\alpha_0 - 1}(1-\theta)^{\beta_0 - 1}$. The hyper-parameters $\alpha_0$ and $\beta_0$ encode prior pseudo-counts of heads and tails respectively; choosing $\alpha_0 = \beta_0 = 1$ yields a uniform prior. By Bayes' theorem, the posterior is

$$P(\theta|\mathcal{D}) = \frac{P(\mathcal{D}|\theta)\, P(\theta)}{P(\mathcal{D})} = \text{Beta}\left(\alpha_0 + k,\ \beta_0 + n - k\right). \tag{2.7}$$

From Equation (2.7), closed-form expressions for the first two moments can be ob-

tained:

$$\mathbb{E}[\theta|\mathcal{D}] = \frac{\alpha_0 + k}{\alpha_0 + \beta_0 + n}, \quad \mathrm{Var}[\theta|\mathcal{D}] = \frac{(\alpha_0 + k)(\beta_0 + n - k)}{(\alpha_0 + \beta_0 + n)^2\,(\alpha_0 + \beta_0 + n + 1)}. \tag{2.8}$$

See Figure 2.1 for an illustration on how the prior distribution is updated into the posterior. Equation (2.7) shows that each observed head increments the shape parameter $\alpha_0$, while each tail increments $\beta_0$, updating prior pseudo-counts with empirical counts. Consequently,

1. as $n \to \infty$ the posterior mean tends to the empirical proportion $k/n$, and the variance $\mathrm{Var}[\theta|\mathcal{D}] \to 0$, increasing certainty about $\theta$;
2. for small samples, the prior exerts appreciable influence, pulling the posterior toward the prior mean $\alpha_0/(\alpha_0 + \beta_0)$ and preventing over-confident estimates when $k$ is near $0$ or $n$;
3. the additive form of Equation (2.7) reveals that the Beta prior contributes exactly $\alpha_0 + \beta_0$ virtual observations, providing an intuitive interpretation of the hyperparameters as prior evidence.

With the uniform prior $(\alpha_0, \beta_0) = (1, 1)$, the posterior reduces to $\mathrm{Beta}(1 + k, 1 + n - k)$; for $n = 0$ this coincides with the prior, reflecting complete ignorance before any data are observed. This conjugate Beta–Binomial model affords analytical tractability, facilitating closed-form credible intervals, model comparison via the marginal likelihood $P(\mathcal{D}) = \mathrm{Beta}(\alpha_0 + k, \beta_0 + n - k)/\mathrm{Beta}(\alpha_0, \beta_0)$, and principled incorporation of expert knowledge through the choice of $(\alpha_0, \beta_0)$.

Given $n$ Bernoulli trials with $k$ heads, the frequentist treats the success probability $\theta$ as a fixed but unknown constant and employs maximum-likelihood estimation:

$$\hat{\theta}_{\mathrm{MLE}} = \frac{k}{n}. \tag{2.9}$$

Sampling uncertainty is quantified via a confidence interval. Using the (asymptotic) Wald interval, one obtains

$$\hat{\theta}_{\mathrm{MLE}} \;\pm\; z_{1-\alpha/2}\sqrt{\frac{\hat{\theta}_{\mathrm{MLE}}\,(1 - \hat{\theta}_{\mathrm{MLE}})}{n}}, \tag{2.10}$$

where $z_{1-\alpha/2}$ is the standard–normal quantile (e.g. $1.96$ for $95\%$ confidence). Exact alternatives such as the Clopper–Pearson or Wilson intervals avoid the poor small-sample performance of the Wald form but require numerical evaluation.

The fundamental differences between the frequentist and Bayesian paradigms can be summarized as follows:

1. **Ontological stance**: the frequentist regards $\theta$ as fixed and randomness arises solely from repeated sampling; the Bayesian treats $\theta$ itself as random, attributing epistemic uncertainty to it through the Beta prior.

2. **Estimation**: both approaches yield point estimates that converge to $k/n$ as $n \to \infty$, but the Bayesian posterior mean $\mathbb{E}[\theta|\mathcal{D}] = (\alpha_0 + k)/(\alpha_0 + \beta_0 + n)$ is a shrinkage estimator, drawing extreme sample proportions toward the prior mean when $n$ is small.
3. **Uncertainty quantification**: a $100(1 - \alpha)\%$ *confidence* interval contains the true $\theta$ in repeated experiments with long-run frequency $1 - \alpha$; a $100(1 - \alpha)\%$ *credible* interval from the Beta posterior assigns probability $1-\alpha$ to the statement "$\theta$ lies in this set" *conditioned on the observed data*. The interpretive gap is especially evident for small samples, where the frequentist interval may be empty or degenerate (e.g. $k = 0$ or $k = n$), whereas the Bayesian interval remains well-behaved via the prior's pseudo-counts.
4. **Role of prior information**: frequentist inference eschews prior knowledge, aiming for procedures with guaranteed coverage irrespective of subjective beliefs; the Bayesian framework naturally incorporates expert opinion or previous studies through $(\alpha_0, \beta_0)$, with transparent impact that diminishes as $n$ grows.
5. **Computational tractability**: while both the Bayesian and frequentist analyses are analytically solvable for the simple Bernoulli–Binomial model, this tractability does not extend to more realistic problems. Bayesian inference is only analytically feasible in conjugate or highly simplified cases; in general, it requires approximate or sampling-based methods such as variational inference or MCMC, which are substantially more computationally demanding than frequentist estimators.
6. **Model comparison and prediction**: the Bayesian marginal likelihood,

$$P(\mathcal{D}) = \mathrm{Beta}(\alpha_0 + k, \beta_0 + n - k)/\mathrm{Beta}(\alpha_0, \beta_0)\,, \tag{2.11}$$

   furnishes a principled basis for model selection and naturally integrates parameter uncertainty into predictive distributions. Frequentist counterparts often rely on information criteria (AIC, BIC) or cross-validation, treating fitted parameters as point estimates.

These differences exemplify the broader philosophical and methodological divergence between Bayesian and frequentist paradigms while also highlighting their agreement in large-sample limits, where prior influence fades and both approaches tend toward the empirical proportion $k/n$.

### 2.2.2 Variational Inference

Modern Bayesian models in machine learning typically involve high-dimensional parameter spaces, non-conjugate likelihoods, complex latent structures, and large datasets. In such settings, the posterior distribution is rarely available in closed form, and exact marginalization requires integrals that are analytically intractable and computationally prohibitive. Posterior landscapes are often multi-modal and strongly correlated, further complicating integration and rendering naive quadrature or exact dynamic programming infeasible. These obstacles motivate *approximate inference* methods that deliver tractable surrogates to the true posterior while respecting the computational constraints of modern practice (e.g., mini-batch training, accelerators, and streaming data) (Bishop, 2006; Blei et al., 2017; MacKay, 2003).

Approximate Bayesian inference is often required because exact posteriors are analytically intractable for most realistic models. A widely used deterministic approach is Variational Inference (VI) (Jordan, 1999), which recasts inference as an optimization problem. VI approximates the true posterior with a simpler, tractable distribution and minimizes a divergence—typically the Kullback–Leibler divergence—between them, thus trading exactness for computational efficiency (Blei et al., 2017; Jordan, 1999).

VI (Jordan, 1999) exchanges the inference issues related to the Bayesian approach with an optimization problem. In VI, a family of distributions $\mathcal{Q}$ over the latent variables $\boldsymbol{\theta}$ needs to be fixed. The aim is to find the element that minimizes its Kullback-Leibler divergence with the intractable posterior $P(\boldsymbol{\theta}|\mathcal{D})$ within that family:

$$Q^\star = \arg\min_{Q \in \mathcal{Q}} KL(Q(\boldsymbol{\theta})|P(\boldsymbol{\theta}|\mathcal{D}))\,. \tag{2.12}$$

**Definition 2.8** (Kullback-Leibler divergence)**.** Let $P$ and $Q$ be two probability distributions over the same probability space $\Omega$, the Kullback-Leibler divergence $\mathrm{KL}(Q|P)$ measures the *dissimilarity* between both distributions as

$$\mathrm{KL}(Q|P) = \mathbb{E}_Q\left[\log Q(x) - \log P(x)\right]. \tag{2.13}$$

The Kullback-Leibler divergence is defined if and only if for all $x \in \Omega$ such that $P(x) = 0$, then $Q(x) = 0$. In measure terms, $Q$ is absolutely continuous with respect to $P$.

**Proposition 2.9.** *The Kullback-Leibler divergence is always non-negative.* [Proof in Appendix C.1]

*Variational Bayesian methods* or simply *variational methods* are primarily used for two purposes:

1. Perform statistical inference over the unobserved variables by providing an analytical approximation to their posterior probability.
2. Derive and compute a lower bound for the observed data marginal likelihood $\log P(\mathcal{D})$. This is commonly used to perform model selection, where a model with a higher marginal likelihood has a greater probability for being the model that generated the data.

These $Q$ distributions are typically referred to as *variational distributions*—the term comes from *calculus of variations*, a field that studies optimization over functions—of the optimization problem. The Kullback-Leibler divergence between the variational distribu-tion $Q(\boldsymbol{\theta})$ and the real distribution $P(\boldsymbol{\theta}|\mathcal{D})$ may be decomposed in the following way:

$$\begin{aligned}\mathrm{KL}(Q(\boldsymbol{\theta})|P(\boldsymbol{\theta}|\mathcal{D})) &= \mathbb{E}_{Q(\boldsymbol{\theta})}[\log Q(\boldsymbol{\theta})] - \mathbb{E}_{Q(\boldsymbol{\theta})}[\log P(\boldsymbol{\theta}|\mathcal{D})] \\ &= \mathbb{E}_{Q(\boldsymbol{\theta})}[\log Q(\boldsymbol{\theta})] - \mathbb{E}_{Q(\boldsymbol{\theta})}[\log P(\mathcal{D},\boldsymbol{\theta}) - \log P(\mathcal{D})] \\ &= \mathbb{E}_{Q(\boldsymbol{\theta})}[\log Q(\boldsymbol{\theta})] - \mathbb{E}_{Q(\boldsymbol{\theta})}[\log P(\mathcal{D},\boldsymbol{\theta})] + \log P(\mathcal{D})\,.\end{aligned} \tag{2.14}$$

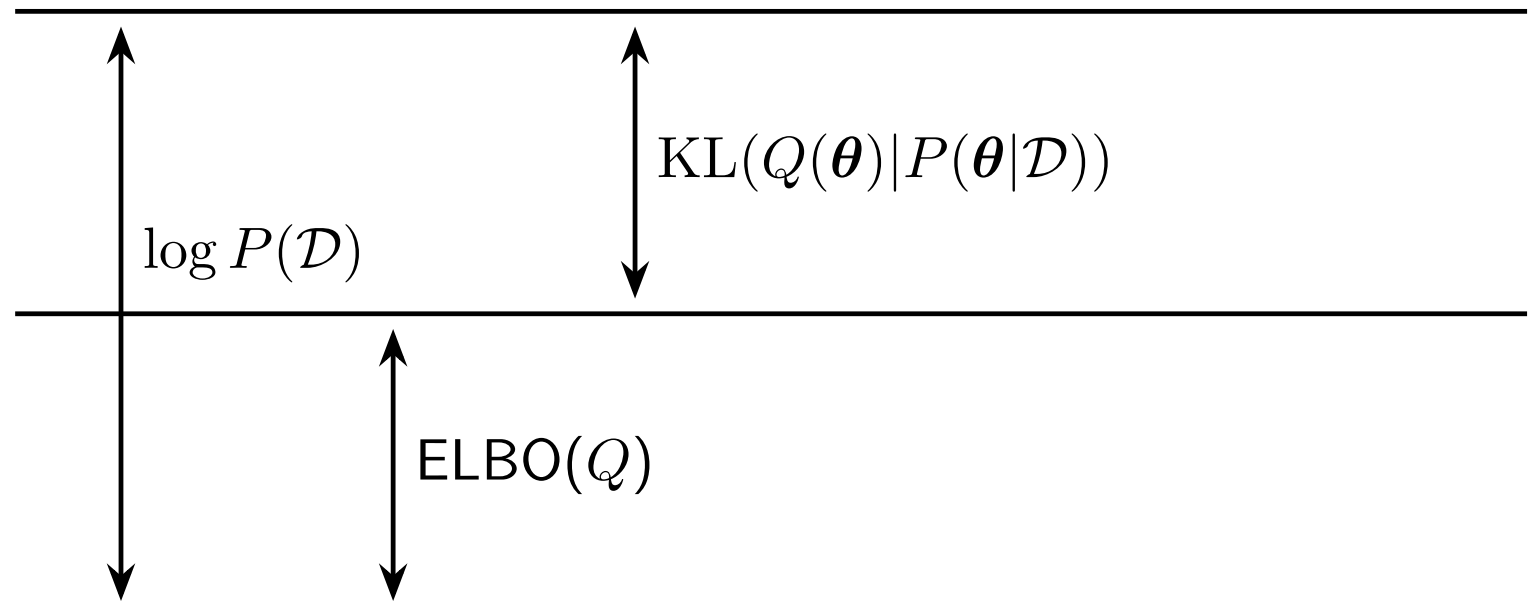

**Figure 2.2:** ***Graphical decomposition of the log evidence into its two fundamental terms.*** *The top horizontal line represents the log-likelihood* $\log P(\mathcal{D})$*, the middle line shows the variational lower bound (ELBO)* $\mathrm{ELBO}(Q)$*, and the gap between them—indicated by the shorter vertical arrow—is the Kullback–Leibler divergence* $\mathrm{KL}(Q(\boldsymbol{\theta})|P(\boldsymbol{\theta}|\mathcal{D}))$*. The double arrows emphasize that the ELBO and the KL add exactly to recover the total evidence, underscoring that maximizing the ELBO is equivalent to minimizing the distance between the variational distribution q and the true posterior.*

Although the Kullback-Leibler divergence cannot be computed since $P(\boldsymbol{\theta}|\mathcal{D})$ is unknown, an equivalent objective can be optimized; its positiveness can be used to set the following lower bound to the evidence, defined as *Evidence Lower Bound* or *ELBO*:

$$\log P(\mathcal{D}) \geq -\underbrace{\mathbb{E}_{Q(\boldsymbol{\theta})}[\log Q(\boldsymbol{\theta})]}_{\text{Entropy}} + \underbrace{\mathbb{E}_{Q(\boldsymbol{\theta})}[\log P(\mathcal{D},\boldsymbol{\theta})]}_{\text{Energy}}\,. \tag{2.15}$$

*Energy* and *Entropy* terms come from statistical physics terminology (Barber, 2012). As $P(\mathcal{D})$ does not depend on $Q$, minimizing the Kullback-Leibler divergence is equivalent to maximizing the ELBO, where equality holds if and only if $Q(\boldsymbol{\theta}) = P(\boldsymbol{\theta}|\mathcal{D})$. The ELBO may be written as

$$\begin{aligned}\mathrm{ELBO}(Q) &= \mathbb{E}_{Q(\boldsymbol{\theta})}[\log P(\boldsymbol{\theta})] + \mathbb{E}_{Q(\boldsymbol{\theta})}[\log P(\mathcal{D}|\boldsymbol{\theta})] - \mathbb{E}_{Q(\boldsymbol{\theta})}[\log Q(\boldsymbol{\theta})] \\ &= \mathbb{E}_{Q(\boldsymbol{\theta})}[\log P(\mathcal{D}|\boldsymbol{\theta})] - \mathrm{KL}(Q(\boldsymbol{\theta})|P(\boldsymbol{\theta}))\,,\end{aligned} \tag{2.16}$$

where it is expressed as the sum of the log likelihood of the observations and the Kullback-Leibler divergence between the prior $P(\boldsymbol{\theta})$ and the variational distribution $Q(\boldsymbol{\theta})$. See Figure 2.2 for an illustration of the decomposition. This formula can be further examined: a variational distribution that maximizes the first term is the one that makes the observed data most probable; in contrast, the variational distribution that maximizes the second term is $Q(\boldsymbol{\theta}) = P(\boldsymbol{\theta})$, the prior. This gives the intuitive idea that the KL divergence in the second term works as a regularizer for the variational distribution, i.e., this distribution needs to explain the data (maximize the first term) without diverging from the prior distribution (maximize the second term).

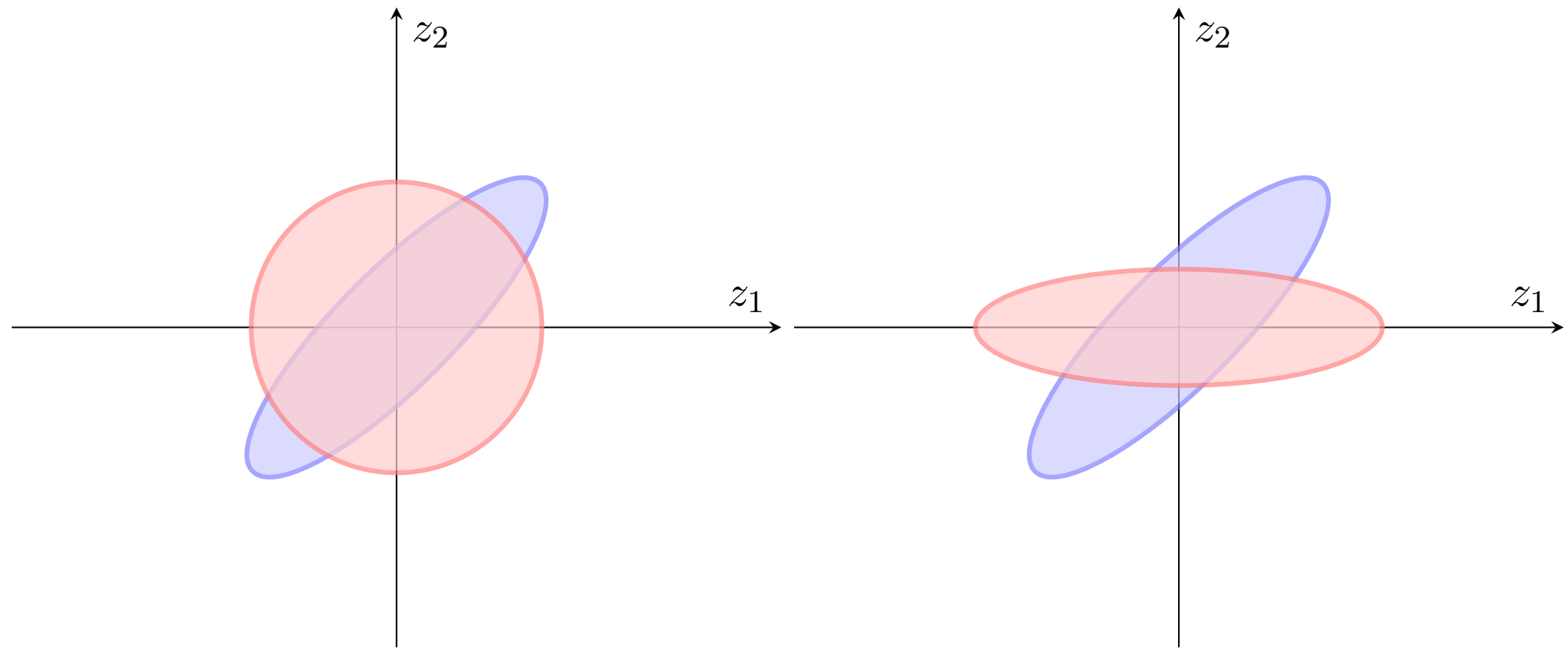

**Figure 2.3:** ***Exact bi-variate Gaussian posterior (blue $1\sigma$ ellipse) and illustrative mean-field Gaussian approximations (orange discs).*** *Both panels share axes $z_1$ (horizontal) and $z_2$ (vertical). The blue ellipse shows the one–standard-deviation contour of the true (correlated) posterior. The orange discs depict mean-field Gaussians chosen for illustration only—they are not optimized to approximate the posterior.* Left: *a mean-field approximation whose variance is visibly larger than the posterior's in both coordinates.* Right: *a mean-field approximation whose variance is visibly smaller* in one coordinate (*and mismatched overall*), *highlighting how independence and marginal-variance constraints can distort the posterior geometry.*

### 2.2.3 Mean Field Family

In most cases, the optimization of the ELBO is intractable if a complex variational family of distributions is selected. A common assumption is that $\mathcal{Q}$ is the *mean-field variational family*. There, $\mathcal{Q}$ is defined as the family of distributions where the variables are mutually independent, i.e, any $Q \in \mathcal{Q}$ verifies

$$Q(\boldsymbol{\theta}) = \prod_{j=1}^{P} Q_j(\theta_j), \tag{2.17}$$

where $\boldsymbol{\theta} = \{\theta_1, \dots, \theta_P\}$ is the considered set of latent variables. Notice that each *factor* $Q_j$ can be different and this family does not depend on the observed data.

The mean-field family can capture any marginal of the latent variables but not the correlation between them, as it assumes they are independent. For example, consider a two-dimensional Gaussian distribution where most of the density is concentrated inside the blue ellipse shown in Figure 2.3. A mean-field approximation would factorize as a product of two Gaussian distributions, condensing its density in a circle or flat ellipse as shown in light red. Both cases in the figure highlight the structural limitation of classical mean-field factorizations—namely, their inability to represent off-diagonal covariance—thereby motivating richer variational families for correlated posteriors.

Notice the parametric form of each factor $Q_j$ is not specified, and the appropriate configuration depends on the variable. For example, a continuous variable might have a Gaussian factor, and a categorical variable might have a categorical factor.

### 2.2.4 Black-box $\alpha$-energy

Classical variational inference (VI) frames posterior computation as an optimization problem by choosing a tractable family $\mathcal{Q}$ and minimizing $\mathrm{KL}(Q(\boldsymbol{\theta})|P(\boldsymbol{\theta}|\mathcal{D}))$. Although convenient, the asymmetric Kullback–Leibler divergence penalizes *overlap* rather than *coverage*, leading to under-dispersed approximations in multi–modal settings (Thomas P Minka, 2005). The *black-box $\alpha$-energy* framework (BB-$\alpha$) (Hernandez-Lobato et al., 2016) generalizes VI by replacing the KL with the $\alpha$–divergence of Zhu and Rohwer (1995):

$$D_\alpha(p|q) = \frac{1}{\alpha(1-\alpha)}\left(1 - \int P(\boldsymbol{\theta})^\alpha Q(\boldsymbol{\theta})^{1-\alpha}\,\mathrm{d}\boldsymbol{\theta}\right), \qquad \alpha \in \mathbb{R}\setminus\{0,1\}. \tag{2.18}$$

**Proposition 2.10.** *The $\alpha$–divergence interpolates between the two asymmetric KLs:*

$$\lim_{\alpha\to 0} D_\alpha(P|Q) = \mathrm{KL}(Q|P), \quad \textit{and,} \quad \lim_{\alpha\to 1} D_\alpha(P|Q) = \mathrm{KL}(P|Q). \tag{2.19}$$

[Proof in Appendix C.1]

Exact minimization of $D_\alpha(p|q)$ is analytically intractable because the true posterior $P(\boldsymbol{\theta}|\mathcal{D})$ is unknown. For i.i.d. data $\mathcal{D} = \{(\mathbf{x}_n, y_n)\}_{n=1}^N$ the objective decomposes as (Y. Li and Gal, 2017)

$$\mathcal{L}_\alpha(q) = R_\beta(Q|P) - \frac{1}{\alpha}\sum_{n=1}^N \log \mathbb{E}_Q[P(y_n|\mathbf{x}_n,\boldsymbol{\theta})^\alpha] \tag{2.20}$$

where $R_\beta$ is the Rényi divergence of order $\beta = N/(N-\alpha)$ (Rényi, 1961):

$$R_\beta(Q|P) := (\beta-1)^{-1}\log\int Q(\boldsymbol{\theta})^\beta P(\boldsymbol{\theta})^{1-\beta}\,\mathrm{d}\boldsymbol{\theta}\,. \tag{2.21}$$

For $\alpha \ll N$ the Rényi term approaches the familiar $\mathrm{KL}(Q|P)$, yielding the *approximate $\alpha$–energy*

$$\tilde{\mathcal{L}}_\alpha(Q) := \mathrm{KL}(Q|P) - \frac{1}{\alpha}\sum_{n=1}^N \log \mathbb{E}_Q[P(y_n|\mathbf{x}_n,\boldsymbol{\theta})^\alpha]. \tag{2.22}$$

The expectation inside the logarithm in Equation (2.22) is rarely analytic. BB-$\alpha$ employs Monte Carlo samples $\{\boldsymbol{\theta}^{(s)}\}_{s=1}^S \sim q(\boldsymbol{\theta})$ and the reparameterization trick to obtain an *unbiased* estimate of the energy but a *biased* estimate of its gradient due to Jensen's inequality:

$$\log \mathbb{E}_Q[h(\boldsymbol{\theta})] \approx \log\left(\frac{1}{S}\sum_{s=1}^S h(\boldsymbol{\theta}^{(s)})\right). \tag{2.23}$$

The bias decays as $\mathcal{O}(S^{-1})$ (Hernandez-Lobato et al., 2016) and is negligible in practice for $S \geq 10$.

Setting $\alpha = 0$ recovers standard VI, $\alpha = 1$ yields power–EP, and $\alpha = \frac{1}{2}$ reproduces the Hellinger distance. Values $\alpha \in (0,1)$ encourage the variational posterior to amortize between mode–seeking (VI) and mass–covering (power–EP) behavior, often giving tighter marginal likelihood estimates and better calibrated predictive variances (Buntine and Jaakkola, 2022; Hernandez-Lobato et al., 2016). Refer to Figure 2.4 for an illustration on the impact of the value on $\alpha$. For $\alpha \to -\infty$ — the solution collapses onto the narrowest mode, reproducing the extreme *mode–seeking* behavior of the reverse KL; $\alpha = 0$ — reverse KL (standard variational Bayes), which still concentrates on a single mode but with finite variance; $\alpha = 0.5$ — a symmetric choice that balances mode fidelity and mass coverage; $\alpha = 1$ — forward KL (power–EP limit), yielding a *mass–covering* approximation that sacrifices peak height to explain both modes; $\alpha \to \infty$ — the approximation centers on the broader–variance mode, emphasizing coverage of high–mass regions. Together, the panels demonstrate the continuous transition from mode–seeking to mass–covering behavior as $\alpha$ moves from negative to positive infinity, thereby motivating the use of $\alpha$–divergence objectives to trade off accuracy and calibration in approximate Bayesian inference.

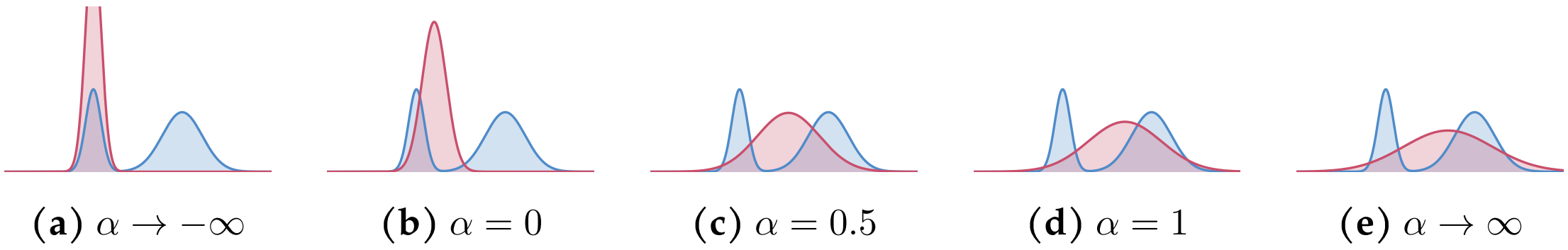

**Figure 2.4:** ***Overlay of a bimodal target density*** $p(\theta)$ ***(blue) and its variational approximation*** $q(\theta)$ ***(red) for five divergence parameters*** $\alpha$. *Each panel shows both densities with semi–transparent fills to emphasize the probability mass. The sequence from left to right demonstrates the transition from extreme mode–seeking* ($\alpha \to -\infty$) *through the reverse-KL* ($\alpha = 0$)*, an intermediate trade-off* ($\alpha = 0.5$)*, the forward-KL* ($\alpha = 1$)*, to strong mass-covering behavior* ($\alpha \to \infty$).

## 2.3 Gaussian Processes

GPs furnish a principled, non-parametric Bayesian framework for function approximation in which prior assumptions are encoded through a mean function $m$ and a positive-definite covariance (kernel) function $K$ (Rasmussen and Williams, 2006). The GP prior, combined with a suitable likelihood, yields closed-form predictive distributions that quantify epistemic uncertainty—an ability that is indispensable in safety-critical and uncertainty-sensitive applications (Deisenroth et al., 2015; Girard et al., 2003). Let $(X_1, \ldots, X_n)$ be a vector of real-valued random variables with joint distribution:

$$(X_1, \ldots, X_n) \ \sim \ \mathcal{N}(\boldsymbol{\mu}, \boldsymbol{\Sigma}), \tag{2.24}$$

where $\boldsymbol{\mu} \in \mathbb{R}^n$ and $\boldsymbol{\Sigma} \in \mathbb{R}^{n \times n}$ denote the mean vector and positive-definite covariance matrix, respectively. For any index set $\mathcal{J} \subseteq \{1, \ldots, n\}$ with $m := |\mathcal{J}|$, define the sub-vector and sub-matrix

$$\boldsymbol{\mu}_{\mathcal{J}} \ := \ (\mu_j)_{j \in \mathcal{J}} \in \mathbb{R}^m, \qquad \boldsymbol{\Sigma}_{\mathcal{J}} \ := \ (\Sigma_{jk})_{j,k \in \mathcal{J}} \in \mathbb{R}^{m \times m}. \tag{2.25}$$

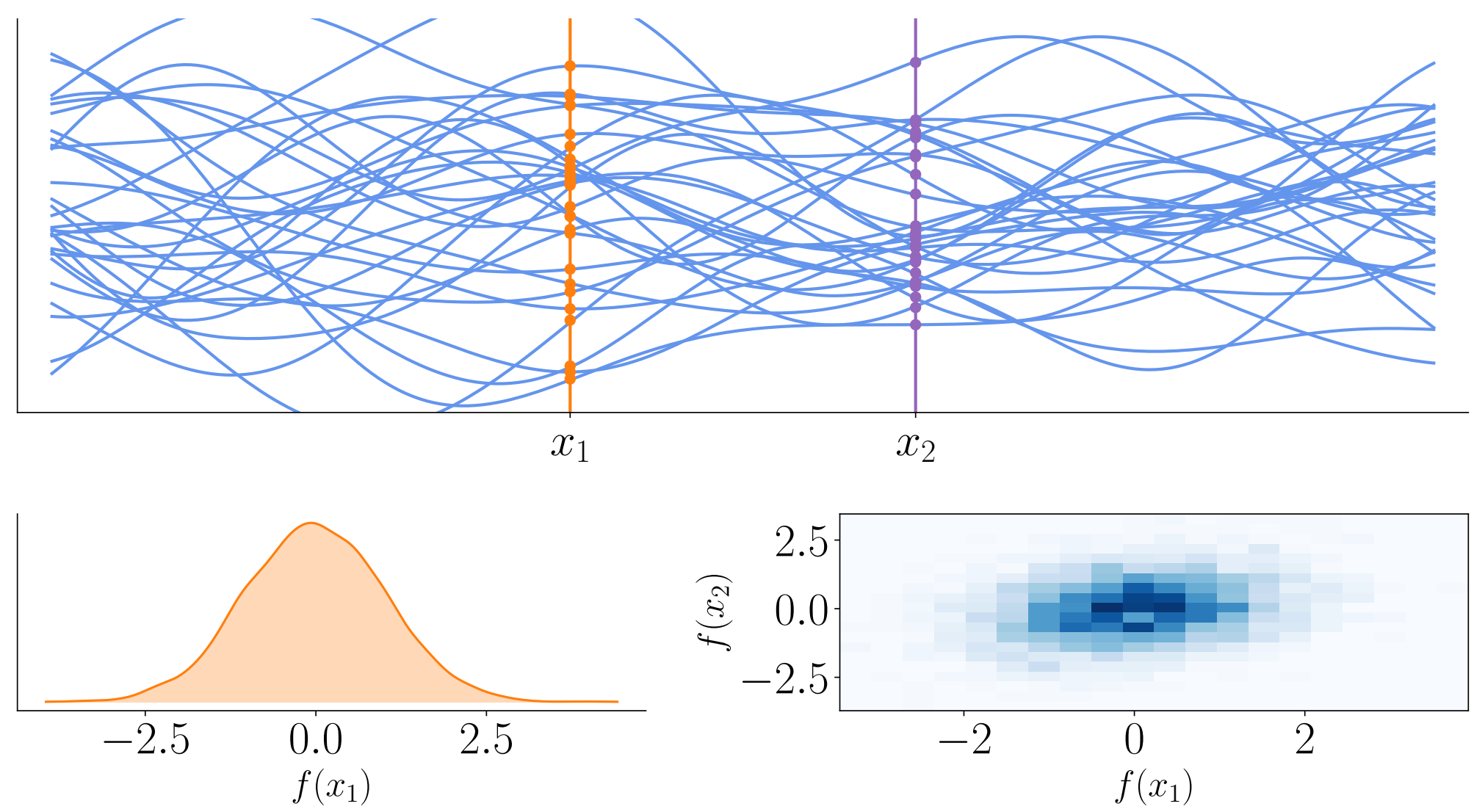

**Figure 2.5:** ***Illustrative properties of a zero–mean Gaussian process (GP) with a squared–exponential covariance,*** $k(x, x') = \sigma^2 \exp\big[-(x - x')^2/(2\ell^2)\big]$ ***where*** $\sigma^2 = 1$ ***and*** $\ell = 0.30$**.** Top: *Thirty independent sample paths on the interval* $[-1, 1]$*; the vertical orange and purple lines mark two input locations,* $x_1 = -0.25$ *and* $x_2 = +0.25$*, with the corresponding function values indicated by dots.* Bottom left: *Empirical marginal density of* $f(x_1)$*, displayed as a filled histogram with a kernel–density estimate overlaid.*Bottom right: *Empirical joint density of* $\big(f(x_1), f(x_2)\big)$ *shown as a two–dimensional histogram, revealing the positive correlation induced by the RBF kernel. Together, the panels visualize* $(i)$ *the smoothness of GP sample paths,* $(ii)$ *Gaussian marginals at individual inputs, and* $(iii)$ *the elliptical joint structure characteristic of bi-variate Gaussian projections.*

Marginalization of a multivariate Gaussian preserves Gaussianity; hence

$$(X_j)_{j \in \mathcal{J}} \sim \mathcal{N}(\boldsymbol{\mu}_{\mathcal{J}}, \boldsymbol{\Sigma}_{\mathcal{J}}) \quad \text{for every } \mathcal{J} \subseteq \{1, \ldots, n\}. \tag{2.26}$$

Conversely, any finite set of evaluations drawn from a Gaussian process must satisfy this same marginal consistency property.

**Definition 2.11** (Gaussian process)**.** A *Gaussian process* is a collection $\{f(\mathbf{x})\}_{\mathbf{x} \in \mathbb{R}^M}$ of random variables such that every finite sub-collection is jointly Gaussian. For a given mean $m : \mathbb{R} \to \mathbb{R}$ and covariance $K : \mathbb{R} \times \mathbb{R} \to \mathbb{R}$ functions, it is denoted as

$$f \sim \mathcal{GP}(m(\cdot), K(\cdot, \cdot)).$$

Given an input design matrix $\mathbf{X} = (\mathbf{x}_1, \ldots, \mathbf{x}_N)^\mathsf{T} \in \mathbb{R}^{N \times M}$, let $\mathbf{f} = (f(\mathbf{x}_1), \ldots, f(\mathbf{x}_N))^\mathsf{T}$. By Definition 2.11,

$$\mathbf{f} \sim \mathcal{N}(m(\mathbf{X}), K(\mathbf{X}, \mathbf{X})), \tag{2.27}$$

where $m(\mathbf{X}) := [m(\mathbf{x}_1), \ldots, m(\mathbf{x}_N)]^\mathsf{T}$ and $[K(\mathbf{X}, \mathbf{X})]_{ij} = K(\mathbf{x}_i, \mathbf{x}_j)$.

Learning the parameters of a finite-dimensional Gaussian distribution requires $N + N(N+1)/2$ degrees of freedom—a regime that is typically *over-determined* in classical parametric models such as linear regression. In contrast, GP regression assumes that the latent function $f$ is a realization of an *infinite-dimensional* stochastic process, mak-ing the inference problem *under-determined*. Practical inference usually proceeds by restricting the kernel to a parametric family and, without loss of generality, setting $m = 0$.

A widely used choice is the squared-exponential or radial-basis-function (RBF) kernel:

$$K_{\text{RBF}}(\mathbf{x}, \mathbf{x}') = \sigma^2 \exp\left[-\frac{\|\mathbf{x} - \mathbf{x}'\|_2^2}{2\ell^2}\right], \tag{2.28}$$

parameterized by a characteristic length-scale $\ell > 0$ and marginal variance $\sigma^2 > 0$. The kernel in Equation (2.28) is infinitely differentiable, thereby encoding strong prior beliefs about the smoothness of $f$ (Stein, 1999). Figure 2.5 visualizes typical sample paths and induced marginal/joint densities for the hyper-parameters $\ell = 0.30$ and $\sigma^2 = 1$.

### 2.3.1 Gaussian Likelihood and Posterior Prediction

In supervised regression, one observes noisy targets $y_i = f(\mathbf{x}_i) + \varepsilon_i$ with i.i.d. noise $\varepsilon_i \sim \mathcal{N}(0, \sigma_\varepsilon^2)$. Denote by $\mathbf{y} = (y_1, \dots, y_N)^\top$ the training outputs. For a collection of test inputs $\mathbf{X}^\star$ the joint prior over training and test function values is

$$\begin{bmatrix} \mathbf{f} \\ \mathbf{f}^\star \end{bmatrix} \sim \mathcal{N}\left(\begin{bmatrix} m(\mathbf{X}) \\ m(\mathbf{X}^\star) \end{bmatrix}, \begin{bmatrix} K(\mathbf{X}, \mathbf{X}) & K(\mathbf{X}, \mathbf{X}^\star) \\ K(\mathbf{X}^\star, \mathbf{X}) & K(\mathbf{X}^\star, \mathbf{X}^\star) \end{bmatrix}\right). \tag{2.29}$$

Because the convolution of Gaussians remains Gaussian (Murphy, 2012), marginalizing the latent $\mathbf{f}$ yields the joint distribution

$$\begin{bmatrix} \mathbf{y} \\ \mathbf{f}^\star \end{bmatrix} \sim \mathcal{N}\left(\begin{bmatrix} m(\mathbf{X}) \\ m(\mathbf{X}^\star) \end{bmatrix}, \begin{bmatrix} K(\mathbf{X}, \mathbf{X}) + \sigma_\varepsilon^2 \mathbf{I} & K(\mathbf{X}, \mathbf{X}^\star) \\ K(\mathbf{X}^\star, \mathbf{X}) & K(\mathbf{X}^\star, \mathbf{X}^\star) \end{bmatrix}\right). \tag{2.30}$$

Conditioning on the observed data delivers the GP predictive distribution (Rasmussen and Williams, 2006):

$$\mathbf{f}^\star | \mathbf{y} \ \sim \ \mathcal{N}(\boldsymbol{\mu}^\star, \, \boldsymbol{\Sigma}^\star), \tag{2.31}$$

where

$$\boldsymbol{\mu}^\star = m(\mathbf{X}^\star) + K(\mathbf{X}^\star, \mathbf{X})[K(\mathbf{X}, \mathbf{X}) + \sigma_\varepsilon^2 \mathbf{I}]^{-1}(\mathbf{y} - m(\mathbf{X})), \tag{2.32}$$

$$\boldsymbol{\Sigma}^\star = K(\mathbf{X}^\star, \mathbf{X}^\star) - K(\mathbf{X}^\star, \mathbf{X})[K(\mathbf{X}, \mathbf{X}) + \sigma_\varepsilon^2 \mathbf{I}]^{-1} K(\mathbf{X}, \mathbf{X}^\star). \tag{2.33}$$

Figure 2.6 illustrates the effect of observing five noisy data points on the posterior mean and variance under an RBF kernel.

Although the squared-exponential kernel is a convenient default, its strong smoothness assumption can lead to model mismatch in the presence of high-frequency or non-stationary phenomena (Rasmussen and Williams, 2006).

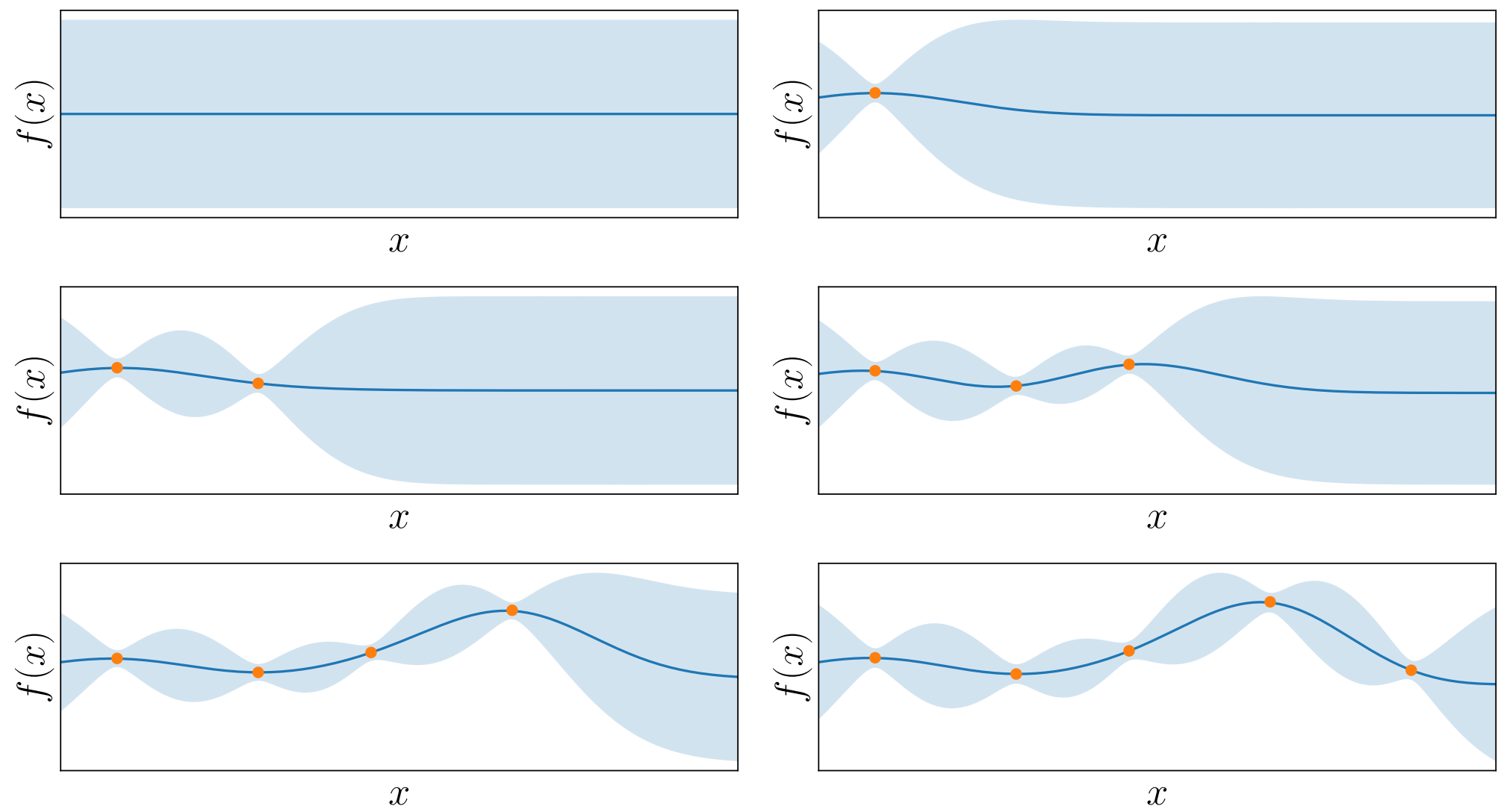

**Figure 2.6:** ***The six panels are arranged chronologically from* top-left *(prior) to* bottom-right *(posterior after five observations).*** *A zero-mean GP prior with squared-exponential kernel $k(x, x') = \sigma^2 \exp[-(x - x')^2/(2\ell^2)]$ ($\ell = 0.3$, $\sigma^2 = 1$) is paired with independent Gaussian observation noise $\varepsilon \sim \mathcal{N}(0, \sigma_\varepsilon^2)$ where $\sigma_\varepsilon = 0.10$. Each panel shows the posterior mean curve (solid blue), its 95 % credible band (shaded), and all data points assimilated up to that stage. As observations are sequentially incorporated, the credible band contracts in the neighborhood of each new input while regions far from data retain the prior level of uncertainty, illustrating how the GP smoothly reconciles prior assumptions with empirical evidence.*

### 2.3.2 Hyper–Parameter Learning

Let $\boldsymbol{\theta} = (\sigma_\varepsilon^2, \vartheta)$ denote the collection of *hyper-parameters*, comprising the observation-noise variance $\sigma_\varepsilon^2$ together with the kernel parameters $\vartheta$ (length–scales, spectral amplitudes, etc.). Conditioning on a data set $\mathcal{D} = \{(\mathbf{x}_i, y_i)\}_{i=1}^N$ and adopting the Gaussian–likelihood model $y_i = f(\mathbf{x}_i) + \varepsilon_i$ with $\varepsilon_i \overset{\text{iid}}{\sim} \mathcal{N}(0, \sigma_\varepsilon^2)$, the *marginal likelihood* (or *evidence*) is obtained by integrating out the latent function values $\mathbf{f} := f(\mathbf{X})$:

$$\begin{aligned} \log P(\mathbf{y}|\boldsymbol{\theta}) &= \log \int P(\mathbf{y}|\mathbf{f}, \sigma_\varepsilon^2)\, P(\mathbf{f}|\vartheta)\, \mathrm{d}\mathbf{f} \\ &= -\tfrac{1}{2}\mathbf{y}^\mathsf{T}\mathbf{C}^{-1}\mathbf{y} - \tfrac{1}{2}\log\det\mathbf{C} - \tfrac{N}{2}\log 2\pi, \end{aligned} \tag{2.34}$$

where $\mathbf{C} = \mathbf{K}_\vartheta + \sigma_\varepsilon^2\mathbf{I}$ and $[\mathbf{K}_\vartheta]_{ij} = K_\vartheta(\mathbf{x}_i, \mathbf{x}_j)$. Equation (2.34) decomposes into a *data–fit* term (first), a complexity penalty proportional to the model capacity (second), and a constant normalization term. Maximizing the evidence embodies *Occam's razor* by trading off goodness of fit against model complexity (MacKay, 1992b; Rasmussen and Williams, 2006). Gradients with respect to $\boldsymbol{\theta}$ can be computed analytically,

$$\frac{\partial \log p(\mathbf{y}|\boldsymbol{\theta})}{\partial \theta_j} = \frac{1}{2}\,\mathrm{tr}\left[\left(\alpha\alpha^T - \mathbf{C}^{-1}\right)\frac{\partial \mathbf{C}}{\partial \theta_j}\right] \tag{2.35}$$

with $\alpha = \mathbf{C}^{-1}\mathbf{y}$, making L-BFGS or conjugate-gradient optimization straightforward (Nocedal and Wright, 2006).

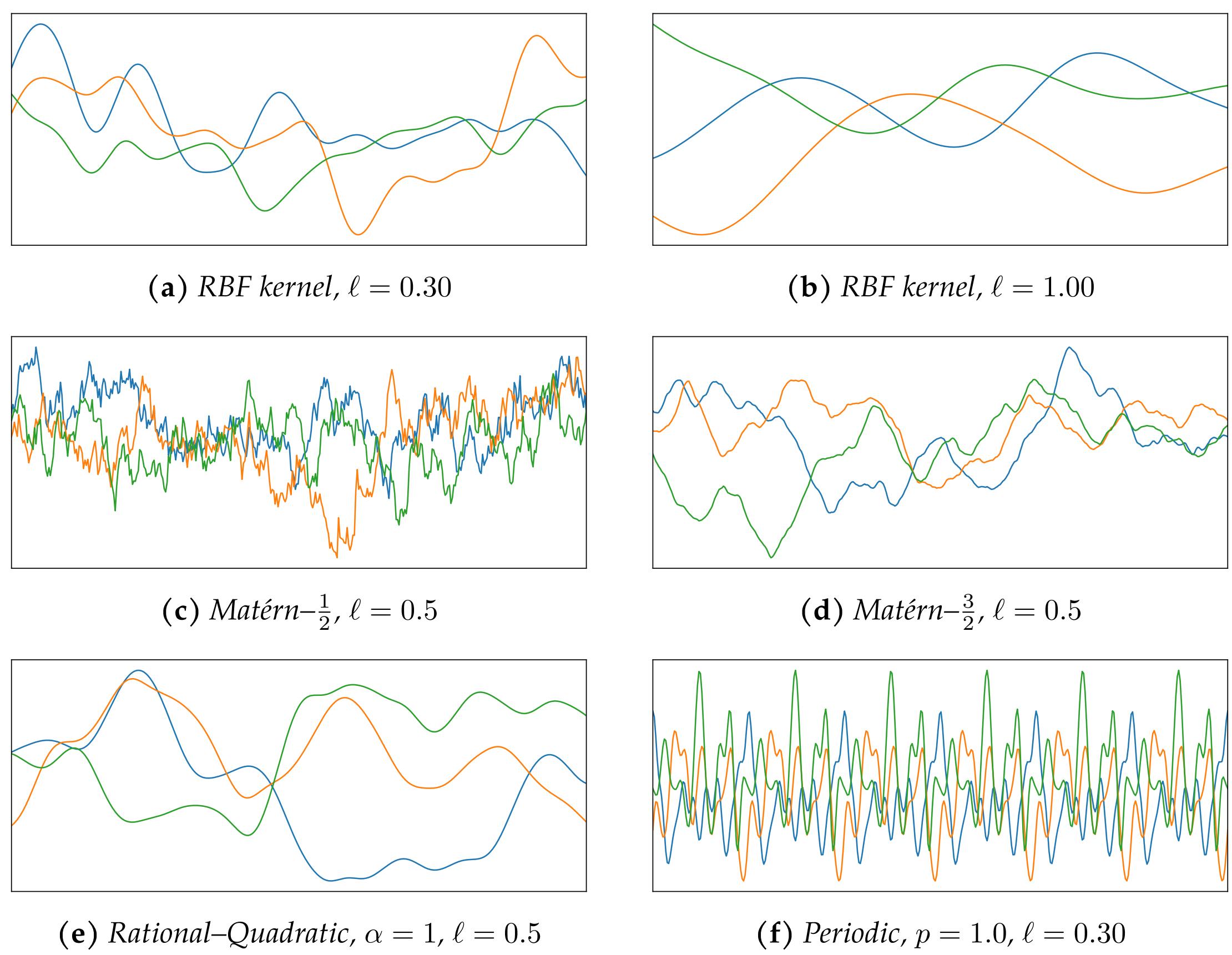

(a) *RBF kernel*, $\ell = 0.30$

(b) *RBF kernel*, $\ell = 1.00$

(c) *Matérn*–$\frac{1}{2}$, $\ell = 0.5$

(d) *Matérn*–$\frac{3}{2}$, $\ell = 0.5$

(e) *Rational–Quadratic*, $\alpha = 1$, $\ell = 0.5$

(f) *Periodic*, $p = 1.0$, $\ell = 0.30$

**Figure 2.7:** ***Prior variability induced by six common Gaussian-process covariance functions, each with unit marginal variance and three independent sample paths.*** *Short-length-scale RBF (top-left) yields very smooth but rapidly varying trajectories, whereas a longer length-scale (top-right) produces gently varying functions. Matérn kernels with* $\nu = \frac{1}{2}$ *and* $\nu = \frac{3}{2}$ *illustrate increasing mean-square differentiability. The rational-quadratic kernel behaves as a scale mixture of RBFs, admitting both long- and short-range variation, while the periodic kernel generates exactly repeating structure. Visual inspection highlights how kernel choice encodes prior beliefs about smoothness, roughness and periodicity before any data are observed.*

In practice, squared–exponential kernels are commonly equipped with Automatic Relevance Determination (ARD) length-scales, assigning a separate smoothness parameter to each input coordinate; maximizing the *marginal likelihood* $\log P(\mathbf{y}|\boldsymbol{\theta})$ then performs data-driven feature weighting and, in effect, feature selection (Rasmussen and Williams, 2006). Figure 2.7 complements this discussion by showing samples drawn from Gaussian processes with different kernels, illustrating how kernel choice (and its hyper-parameters) controls the qualitative properties of functions—e.g., smoothness, periodicity, or roughness. This qualitative view connects to hyperparameter learning because the marginal likelihood can be used not only to tune $\boldsymbol{\theta}$ within a kernel family but also to *compare* kernel families and select the one expected to generalize best. Alternative Bayesian treatments place a prior on $\boldsymbol{\theta}$ and sample from the hyper-posterior via Markov chain Monte Carlo (Filippone and Engler, 2015; Gilks et al., 1995). When gradients are unreliable or the objective is multi-modal, Bayesian optimization of the marginal likelihood itself is an effective strategy (Snoek et al., 2012).

### 2.3.3 Computational Considerations

Exact inference requires solving a linear system and evaluating a log-determinant for the $N \times N$ matrix $\mathbf{C}$, incurring $\mathcal{O}(N^3)$ floating-point operations and $\mathcal{O}(N^2)$ memory. Consequently, *vanilla* GPs are limited to a few thousand training points on commodity hardware (Rasmussen and Williams, 2006). Research over the past two decades has produced two broad solution classes:

1. **Structured algebra** exploits Kronecker, Toeplitz or inducing Kronecker product structure to obtain nearly linear scaling (Gardner et al., 2018; Wilson and Nickisch, 2015).
2. **Low-rank or sparse approximations** replace the full covariance by a rank-$M$ surrogate with $M \ll N$, leading to $\mathcal{O}(NM^2)$ training cost. Within this family, variational inducing-point methods dominate due to their solid theoretical footing and favorable empirical accuracy (Hensman et al., 2013; Titsias, 2009).

### 2.3.4 Inducing–Point Approximations

Exact GP inference scales cubically with the number of training cases $N$ and therefore becomes prohibitive for data sets that nowadays easily reach $10^5$–$10^6$ points. A fruitful line of research mitigates this cost by introducing a much smaller set of $M \ll N$ *inducing inputs* (or *pseudo–inputs*) $\mathbf{Z} = \{\mathbf{z}_m\}_{m=1}^{M}$ together with their function values $\mathbf{u} = f(\mathbf{Z}) = (f(\mathbf{z}_1), \dots, f(\mathbf{z}_M))^\top$ (Snelson and Ghahramani, 2006). This section provides a comprehensive discussion of the resulting sparse Gaussian–process (SGP) framework and its variational reformulation.

#### 2.3.4.1 Historical Perspective and Deterministic Schemes

The earliest sparse models such as the *subset of regressors* (SoR) approximation replaced the full covariance $\mathbf{K}_{NN}$ by the Nyström factor $\mathbf{K}_{NM}\mathbf{K}_{MM}^{-1}\mathbf{K}_{MN}$, reducing storage to $\mathcal{O}(NM)$ but at the cost of severe underestimation of predictive uncertainty (Smola and Schölkopf, 2000). The *deterministic training conditional* (DTC), *partially independent training conditional* (PITC) and the *fully independent training conditional* (FITC) successively restored more of the original covariance structure while retaining $\mathcal{O}(NM^2)$ complexity (Quiñonero-Candela and Rasmussen, 2005). These methods can be interpreted as exact inference in a *modified* probabilistic model; hence they do not optimise a true marginal likelihood with respect to $\mathbf{Z}$.

**Subset of regressors (SoR).** The original sparse–GP idea is to retain only a *small* subset $\mathbf{Z} = \{\mathbf{z}_m\}_{m=1}^{M}$ with $M \ll N$ and to build the covariance of the full data set from these $M$ points via the Nyström approximation

$$\tilde{\mathbf{K}}_{\text{SoR}} = \mathbf{K}_{NM}\,\mathbf{K}_{MM}^{-1}\,\mathbf{K}_{MN}, \qquad \mathbf{K}_{AB} := K(\mathbf{A}, \mathbf{B}), \tag{2.36}$$

first proposed by Smola and Schölkopf (2000). Because all computations now factor through the $M \times M$ matrix $\mathbf{K}_{MM}$, training requires only $\mathcal{O}(NM^2)$ time and $\mathcal{O}(NM)$ memory, while each test prediction costs $\mathcal{O}(M^2)$. The price of this efficiency is that $\operatorname{rank}(\tilde{\mathbf{K}}_{\text{SoR}}) \leq M$; consequently the predictive variance vanishes as the test input $\mathbf{x}^\star$ moves far from all inducing points, yielding overly confident forecasts at the fringes of the input space.

**Deterministic training conditional (DTC).** DTC corrects this problem by retaining the exact marginal variance on the training set: $\tilde{\mathbf{K}}_{\text{DTC}} = \tilde{\mathbf{K}}_{\text{SoR}} + \operatorname{diag}(\mathbf{K}_{NN} - \tilde{\mathbf{K}}_{\text{SoR}})$ (Seeger, 2003). Predictive variances are now sensible away from data, yet DTC preserves only diagonal residuals and therefore fails to model inter-point correlations.

**Partially and fully independent training conditionals.** PITC partitions the data into $B$ disjoint blocks $\{\mathcal{I}_b\}_{b=1}^{B}$ and adds back block-diagonal residuals:

$$\tilde{\mathbf{K}}_{\text{PITC}} = \tilde{\mathbf{K}}_{\text{SoR}} + \operatorname{blkdiag}_{b=1}^{B}\Big(\mathbf{K}_{\mathcal{I}_b\mathcal{I}_b} - \tilde{\mathbf{K}}_{\text{SoR},\mathcal{I}_b}\Big). \tag{2.37}$$

For $B = N$ PITC reduces to the *fully independent training conditional* (FITC) of Snelson and Ghahramani (2006), which adds back all *diagonal* residual terms. Both schemes require $\mathcal{O}(NM^2)$ time and $\mathcal{O}(NM)$ memory, and critically they restore the *exact* predictive variance at the training inputs while maintaining tractable algebra:

$$\begin{aligned}
\mu_{\text{FITC}}(\mathbf{x}^\star) &= \mathbf{q}_{x^\star}^{\top}\big(\mathbf{Q}_{NN} + \mathbf{\Lambda} + \sigma_\varepsilon^2\mathbf{I}\big)^{-1}\mathbf{y},\\
\sigma^2_{\text{FITC}}(\mathbf{x}^\star) &= k_{x^\star x^\star} - \mathbf{q}_{x^\star}^{\top}\big(\mathbf{Q}_{NN} + \mathbf{\Lambda} + \sigma_\varepsilon^2\mathbf{I}\big)^{-1}\mathbf{q}_{x^\star},
\end{aligned}$$

with $\mathbf{q}_{x^\star} = \mathbf{K}_{MM}^{-1}\mathbf{K}_{Mx^\star}$, $\mathbf{Q}_{NN} = \mathbf{K}_{NM}\mathbf{K}_{MM}^{-1}\mathbf{K}_{MN}$ and $\mathbf{\Lambda} = \operatorname{diag}(\mathbf{K}_{NN} - \mathbf{Q}_{NN})$. Empirically, FITC attains the lowest root-mean-square error among deterministic sparse schemes but exhibits a pronounced tendency to inflate length-scales during evidence maximization (Bauer et al., 2016).

**Interpretation as exact inference in an amended model.** All of the above approximations can be written $p_{\text{approx}}(\mathbf{f}) = \mathcal{N}(\mathbf{0}, \tilde{\mathbf{K}})$ with $\tilde{\mathbf{K}}$ *not* equal to the original kernel matrix; therefore, performing exact GP inference in a *different* probabilistic model. Hyperparameter learning optimizes the marginal likelihood of this amended model, which can lead to systematic bias in the inferred length-scales and noise variance (Titsias, 2009). The variational framework of the following section addresses this shortcoming by optimizing a genuine lower bound on the true evidence.

**Summary of deterministic approximations.** Table 2.1 contrasts algebraic form, computational cost, and known pathologies. Although superseded in most applications by variational inducing-point methods, deterministic schemes remain useful for debugging and as initializers because they require no optimization over covariance matrices.

**Table 2.1:** ***Comparison of single-pass sparse Gaussian Process (GP) approximation methods that employ deterministic algebraic formulations.*** *For each method, the table reports the structure imposed on the surrogate covariance matrix* $\tilde{\mathbf{K}}$*, the resulting asymptotic training cost in terms of the number of data points* $N$ *and inducing points* $M$*, and the primary limitations that have been identified for practical use.*

| **Method** | $\tilde{\mathbf{K}}$ structure | Train cost | Known issues |
|---|---|---|---|
| SoR | rank-$M$ Nyström | $NM^2$ | variance collapse |
| DTC | rank-$M$ + diag | $NM^2$ | no off-diag residuals |
| FITC | rank-$M$ + diag residual | $NM^2$ | length-scale inflation |
| PITC | rank-$M$ + block-diag residual | $NM^2$ | block selection heuristic |

#### 2.3.4.2 Variational Formulation

A principled alternative casts sparsification as variational inference (Titsias, 2009). Select a set of *inducing locations* $\mathbf{Z} \subset \mathcal{X}$, then augment the GP prior with $\mathbf{u} := f(\mathbf{Z})$,

$$P(\mathbf{f}, \mathbf{u}) = P(\mathbf{f}|\mathbf{u})\, P(\mathbf{u}) = \mathcal{N}(\mathbf{f}; \mathbf{K}_{NM}\mathbf{K}_{MM}^{-1}\mathbf{u},\, \mathbf{K}_{NN} - \mathbf{Q}_{NN})\, \mathcal{N}\left(\mathbf{u}; \mathbf{0}, \mathbf{K}_{MM}\right), \tag{2.38}$$

where $\mathbf{Q}_{NN} = \mathbf{K}_{NM}\mathbf{K}_{MM}^{-1}\mathbf{K}_{MN}$. Introducing a variational joint density $Q(\mathbf{f}, \mathbf{u}) = P(\mathbf{f}|\mathbf{u})Q(\mathbf{u})$, with $Q(\mathbf{u}) = \mathcal{N}(\boldsymbol{\mu}, \mathbf{S})$ yields the evidence lower bound:

$$\mathcal{L} = \sum_{i=1}^{N} \mathrm{E}_{Q(f_i)}[\log P(y_i|f_i)] - \mathrm{KL}[Q(\mathbf{u})|P(\mathbf{u})], \qquad Q(\mathbf{f}) = \int P(\mathbf{f}|\mathbf{u})Q(\mathbf{u})\, \mathrm{d}\mathbf{u}\,. \tag{2.39}$$

Optimizing $\mathcal{L}$ with respect to $\boldsymbol{\mu}$ and $\mathbf{S}$ in closed form collapses the bound to the FITC marginal likelihood plus an information–theoretic correction term that discourages degenerate covariance approximations (Bauer et al., 2016). Hyper–parameters $\boldsymbol{\theta}$ and inducing locations $\mathbf{Z}$ are then learned by gradient ascent on Equation (2.39). Critically, $\mathcal{O}(NM^2)$ time and $\mathcal{O}(NM)$ memory make mini-batch stochastic optimization possible, paving the way for the *Stochastic Variational GP* (SVGP) of Hensman et al. (2013).

The predictive distribution at test inputs $\mathbf{X}^\star$ remains analytic, being Gaussian with:

$$\begin{aligned} \mathbb{E}_q[f(\mathbf{X}^\star)] &= \mathbf{K}_{\star M}\mathbf{K}_{MM}^{-1}\boldsymbol{\mu}\,, \\ \mathrm{Var}_q[f(\mathbf{X}^\star)] &= \mathbf{K}_{\star} - \mathbf{K}_{\star M}\mathbf{K}_{MM}^{-1}(\mathbf{K}_{MM} - \mathbf{S})\mathbf{K}_{MM}^{-1}\mathbf{K}_{M\star}\,. \end{aligned}$$

The computational cost is now $\mathcal{O}(M^2)$ per test point, independent of $N$. See Figure 2.8 for an illustration of this method.

The number $M$ and the placement of $\mathbf{Z}$ heavily influence predictive accuracy and computational cost. Common heuristics initialize $\mathbf{Z}$ with $k$–means centroids, random subsets of data, Latin–hypercube designs, or condition–number–based strategies (Rubin and Stein, 2016). Empirical studies show that subsequent gradient refinement is essential; otherwise, the optimizer partly compensates by inflating length–scales, leading to over–smooth predictions (Burt and Rasmussen, 2020).

Table 2.2 summarizes the asymptotic trade–offs. For modern GPU-accelerated li-braries, $M$ in the low thousands is common, permitting data sets two orders of magnitude larger than those tractable by direct Cholesky factorization while retaining high fidelity to the exact GP posterior (Burt and Rasmussen, 2020).

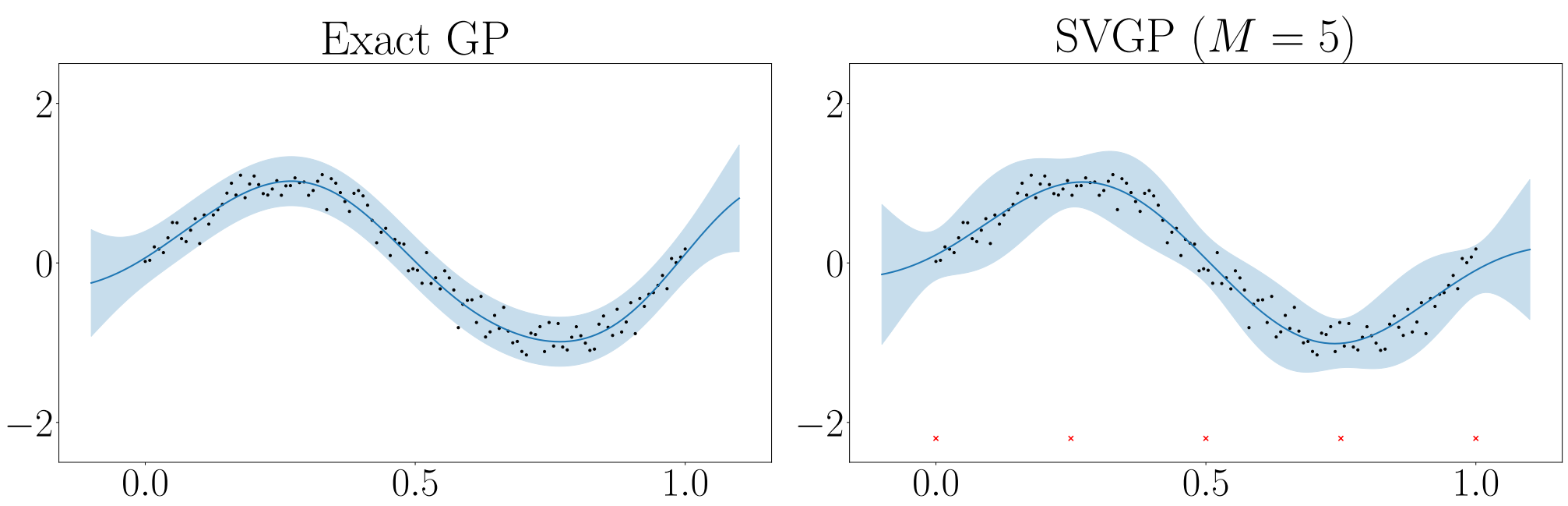

**Figure 2.8:** ***Side-by-side comparison of an exact Gaussian process and a sparse variational Gaussian process (SVGP) with five inducing points.*** *The left panel shows the predictive mean (solid blue curve) and 95% confidence band (shaded) of the exact GP fitted to the training data (black dots). The right panel depicts the SVGP approximation using $M = 5$ inducing locations (red $\times$ along the bottom); its mean function closely follows the exact solution, while the uncertainty band widens slightly between inducing points, illustrating how a small set of inducing variables can capture global structure with only a modest loss of fidelity.*

**Table 2.2:** ***Asymptotic computational complexity of exact and representative sparse Gaussian-process (GP) methods for a training set of size $N$ and $P$ test points.*** *For sparse variants, $M$ denotes the number of inducing points and $B$ the mini-batch size used during stochastic optimization. Reported costs cover the dominant terms for training time, prediction time, and peak memory usage under typical implementations.*

| Method | Train time | Predict time | Memory |
|---|---|---|---|
| Exact GP | $\mathcal{O}(N^3)$ | $\mathcal{O}(PN^2)$ | $\mathcal{O}(N^2)$ |
| FITC / PITC | $\mathcal{O}(NM^2)$ | $\mathcal{O}(PM^2)$ | $\mathcal{O}(NM)$ |
| SVGP (mini-batch) | $\mathcal{O}(BM^2)$ | $\mathcal{O}(PM^2)$ | $\mathcal{O}(NM)$ or $\mathcal{O}(M^2)$ with caching |

### 2.3.5 Bayesian Linear–Regression Surrogates for Gaussian Processes

Exact GP regression inherits the cubic computational cost of inverting the $N \times N$ kernel matrix $\mathbf{K}$. A complementary route to scalability is to *approximate the GP itself* by a Bayesian linear model in a *finite* feature space. Early instances include *relevance–vector machines* (Quiñonero-Candela and Rasmussen, 2005; Tipping, 2001), while modern treatments exploit random–feature expansions (Balog et al., 2016; Gal and Turner, 2015; Rahimi and Recht, 2007).

Let $K : \mathcal{X} \times \mathcal{X} \to \mathbb{R}$ be a continuous, symmetric, positive–definite kernel whose diagonal $K(\mathbf{x}, \mathbf{x})$ is integrable. By Mercer's theorem (Mercer, 1909), there exist non-negative eigenvalues $\{\lambda_i\}_{i=1}^{\infty}$ and an orthonormal set of eigen-functions $\{\psi_i\}_{i=1}^{\infty} \subset L^2(\mathcal{X})$ such that

$$K(\mathbf{x}, \mathbf{x}') = \sum_{i=1}^{\infty} \lambda_i \psi_i(\mathbf{x}) \psi_i(\mathbf{x}') \,. \tag{2.40}$$

Define the (countably infinite) *feature map*

$$\begin{aligned} \phi &: \mathcal{X} \to \ell^2 \\ \mathbf{x} &\mapsto (\sqrt{\lambda_1}\psi_1(\mathbf{x}), \sqrt{\lambda_2}\psi_2(\mathbf{x}), \dots)^\mathsf{T}. \end{aligned} \tag{2.41}$$

Then $K$ is the inner product in feature space, $K(\mathbf{x}, \mathbf{x}') = \phi(\mathbf{x})^\mathsf{T}\phi(\mathbf{x}')$. Using a zero prior mean (extensions to $m(\cdot) \neq 0$ are straightforward) and substituting Equation (2.40) into the standard GP predictive equations yields

$$\mu^\star = \phi(\mathbf{x}^\star)^\mathsf{T}\mathbf{\Phi}(\mathbf{X})\left[\mathbf{\Phi}(\mathbf{X})^\mathsf{T}\mathbf{\Phi}(\mathbf{X})\right]^{-1}\mathbf{y}, \tag{2.42}$$

$$\Sigma^\star = \phi(\mathbf{x}^\star)^\mathsf{T}\phi(\mathbf{x}^\star) - \phi(\mathbf{x}^\star)^\mathsf{T}\mathbf{\Phi}(\mathbf{X})\left[\mathbf{\Phi}(\mathbf{X})^\mathsf{T}\mathbf{\Phi}(\mathbf{X})\right]^{-1}\mathbf{\Phi}(\mathbf{X})^\mathsf{T}\phi(\mathbf{x}^\star), \tag{2.43}$$

with $\mathbf{\Phi}(\mathbf{X}) := [\phi(\mathbf{x}_1), \dots, \phi(\mathbf{x}_N)]^\mathsf{T}$. Equations (2.42)–(2.43) coincide with the posterior of a Bayesian *linear* model $f(\mathbf{x}) = \phi(\mathbf{x})^\mathsf{T}\mathbf{w}$, $\mathbf{w} \sim \mathcal{N}(\mathbf{0}, \mathbf{I})$. The obstacle is that the feature map is infinite–dimensional.

For *shift-invariant* kernels $K(\mathbf{x}, \mathbf{x}') = k(\mathbf{x} - \mathbf{x}')$ Bochner's theorem states that $k$ is the Fourier transform of a finite non–negative measure $\tau$. Drawing $\{\boldsymbol{\omega}_s\}_{s=1}^S \sim \tau$ and random phases $\{b_s\}_{s=1}^S \sim \mathcal{U}[0, 2\pi]$ one obtains the *random Fourier feature* (RFF) map

$$\hat{\phi}_{\text{RFF}}(\mathbf{x}) = \frac{1}{\sqrt{S}}\left[\cos(\boldsymbol{\omega}_0^\mathsf{T}\mathbf{x}+b_0),\ \sin(\boldsymbol{\omega}_0^\mathsf{T}\mathbf{x}+b_0),\ \dots,\ \cos(\boldsymbol{\omega}_S^\mathsf{T}\mathbf{x}+b_S),\ \sin(\boldsymbol{\omega}_S^\mathsf{T}\mathbf{x}+b_S)\right]. \tag{2.44}$$

Then $\mathbb{E}[\hat{\phi}_{\text{RFF}}(\mathbf{x})^\mathsf{T}\hat{\phi}_{\text{RFF}}(\mathbf{x}')] = K(\mathbf{x}, \mathbf{x}')$ and the Monte–Carlo error decays as $\mathcal{O}(S^{-1/2})$ (Rahimi and Recht, 2007). A *Bayesian* linear model with these features retains closed form predictive moments and reduces training from $\mathcal{O}(N^3)$ to $\mathcal{O}(NS^2)$.

Bach (2017) shows that, for kernels with spectral density bounded away from zero and infinity, an RFF GP with $S = \mathcal{O}(d\,\epsilon^{-2}\log d)$ features approximates the full GP within $\epsilon$ in operator norm with high probability. Burt and Rasmussen (2020) further proves that the marginal- likelihood error decays as $\mathcal{O}(S^{-1})$.

Kronecker and circulant structures can reduce the cost of sampling and multiplication with $\hat{\phi}(\mathbf{x})$ from $\mathcal{O}(S)$ to $\mathcal{O}(\log S)$ without degrading accuracy (Le et al., 2013; Wilson and Nickisch, 2015). Tensor-train and Gaussian-quadrature features extend RFFs to Matérn kernels and low-dimensional manifolds (Novikov and Izmailov, 2018).

If the spectral measure $\tau$ is replaced by the empirical distribution on a subset of training points, the Bayesian linear surrogate recovers Nyström inducing-point GPs (Section 2.3.4.1). Placing an independent AR(1) hyper-prior on the weights yields the RVM sparsity-promoting posterior (Tipping, 2001). Hence, random-feature GPs unify earlier sparse approaches under a single linear–regression view.

With $S \ll N$ the dominant cost of Bayesian random-feature regression is the inversion of the $S \times S$ weight covariance, $\mathcal{O}(S^3)$, plus $\mathcal{O}(NS^2)$ to compute the design matrix $\hat{\mathbf{\Phi}}(\mathbf{X})$. Stochastic optimisation and conjugate gradients reduce both costs to $\mathcal{O}(BS)$ per iteration, making GPs feasible for $N \gtrsim 10^6$ on commodity GPUs.

### 2.3.6 Dual Interpretation of Gaussian Processes Via Gaussian Measures

This section takes the familiar idea of a Gaussian process (GP) and translates it into the language of functional analysis. After briefly reminding the reader what inner-product and Hilbert spaces are, it shows how every GP can be seen as a special kind of Gaussian *measure* living in an infinite-dimensional function space. Thinking in measures rather than kernels may feel abstract, but it pays off later: it allows the thesis to borrow powerful operator-theoretic tools and provides a unifying viewpoint for the variational approach in Chapter 4. In short, this section builds the mathematical "bridge" that the rest of the thesis will repeatedly cross whenever it needs to move between the concrete world of kernels and the more flexible world of measures.

Throughout this thesis, the language of functional analysis—most notably, the theory of *Hilbert spaces*—plays a central role. This section collects the essential concepts and clarifies the notation used henceforth. Standard references include Conway (1990), Reed and Simon (1980), and Rudin (1991). For further information on abstract Wiener spaces and their construction, refer to Bogachev (1998), Gross (1967), Kuo (2006), and Van Neerven et al. (2010).

**Definition 2.12** (Inner–product spaces)**.** A (real) *inner–product space* is a vector space $(\mathcal{V}, +, \cdot)$ over $\mathbb{R}$ equipped with a bilinear form $\langle \cdot, \cdot \rangle_{\mathcal{V}} : \mathcal{V} \times \mathcal{V} \to \mathbb{R}$ such that, for all $u, v \in \mathcal{V}$ and $\lambda \in \mathbb{R}$,

1. *symmetry:* $\langle u, v \rangle_{\mathcal{V}} = \langle v, u \rangle_{\mathcal{V}}$;
2. *positive definiteness:* $\langle v, v \rangle_{\mathcal{V}} \geq 0$ with equality if and only if $v = 0$.

The inner product induces the norm $\|v\|_{\mathcal{V}} := \sqrt{\langle v, v \rangle_{\mathcal{V}}}$ and the associated metric $d(u, v) := \|u - v\|_{\mathcal{V}}$.

**Definition 2.13** (Hilbert spaces)**.** A Hilbert space $(\mathcal{H}, \langle \cdot, \cdot \rangle_{\mathcal{H}})$ is an inner–product space that is *complete*: every Cauchy sequence converges in the norm. Completeness ensures the existence of orthonormal bases and enables orthogonal projection onto closed subspaces, paralleling Euclidean–space geometry.

**Notation.**

- Inner products appear as $\langle f, g \rangle_{\mathcal{H}}$. When the space is clear from context, the subscript is omitted.
- Norms are written $\|f\|_{\mathcal{H}} := \sqrt{\langle f, f \rangle_{\mathcal{H}}}$.
- $\mathcal{L}(\mathcal{H})$ denotes the space of bounded linear operators on $\mathcal{H}$; $\mathcal{L}^+(\mathcal{H})$ is the cone of positive, self–adjoint operators:

$$\mathcal{L}^+(\mathcal{H}) := \{A : \mathcal{H} \to \mathcal{H},\ A = A^\star\ ,\ A \geq 0\}\,. \tag{2.45}$$

**Canonical examples.**

1. $\mathbb{R}^d$ with the Euclidean inner product $\langle u, v \rangle = \sum_{i=1}^d u_i v_i$.

2. The sequence space $\ell^2 := \{a = (a_1, a_2, \dots) : \sum_i a_i^2 < \infty\}$ with $\langle a, b\rangle = \sum_i a_i b_i$.
3. The function space $L^2(\mathcal{X}, \mu)$ consisting of (equivalence classes of) square–integrable functions with inner product $\langle f, g\rangle = \int_{\mathcal{X}} f(x)g(x)\,\mathrm{d}\mu(x)$.

**Definition 2.14** (Reproducing–kernel Hilbert spaces)**.** A Hilbert space $\mathcal{H} \subset \mathbb{R}^{\mathcal{X}}$ of functions on $\mathcal{X}$ is a *reproducing–kernel Hilbert space* (RKHS) if the point–evaluation functional $\delta_{\mathbf{x}} : f \mapsto f(\mathbf{x})$ is continuous for every $\mathbf{x} \in \mathcal{X}$. By the Riesz representation theorem, there exists a unique element $\phi_{\mathbf{x}} \in \mathcal{H}$ such that $f(\mathbf{x}) = \langle \phi_{\mathbf{x}}, f\rangle_{\mathcal{H}}$; the kernel $K(\mathbf{x}, \mathbf{x}') := \langle \phi_{\mathbf{x}}, \phi_{\mathbf{x}'}\rangle_{\mathcal{H}}$ is symmetric, positive–definite, and *reproduces* $\mathcal{H}$.

By Moore–Aronszajn theorem (Aronszajn, 1950), if $K$ is a positive definite kernel on $\mathcal{X}$, then, there exists a unique Hilbert space of functions on $\mathcal{H}$ for which $K$ is a reproducing kernel. More precisely, let $\mathcal{H}_0(\mathcal{X})$ be the linear span of $K$ on $\mathcal{X}$ defined as

$$\mathcal{H}_0(\mathcal{X}) := \Big\{ \sum_{i=1}^{n} a_i K(\cdot, \mathbf{x}_i) : n \in \mathbb{N},\ a_i \in \mathbb{R},\ \mathbf{x}_i \in \mathcal{X} \Big\}. \tag{2.46}$$

By the Moore–Aronszajn theorem, the completion of $\mathcal{H}_0(\mathcal{X})$ with respect to

$$\Big\langle \sum_{i=1}^{n} a_i\, K(\,\cdot\,, \mathbf{x}_i),\ \sum_{j=1}^{m} b_j\, K(\,\cdot\,, \mathbf{x}'_j) \Big\rangle_{\mathcal{H}_0} := \sum_{i=1}^{n}\sum_{j=1}^{m} a_i b_j\, K(\mathbf{x}_i, \mathbf{x}'_j)\,, \tag{2.47}$$

is a Hilbert space $\mathcal{H}$ verifying the reproducing property with $\phi_{\mathbf{x}} = K(\cdot, \mathbf{x}),\ \forall \mathbf{x} \in \mathcal{X}$.

**Definition 2.15** (Real Banach space)**.** Let $\mathcal{X}$ be a set, then, a Banach space of functions is a pair $(\mathcal{B}, \|\cdot\|_{\mathcal{B}})$ such that

1. $\mathcal{B}$ is a vector space of (equivalence classes of) functions $f : \mathcal{X} \to \mathbb{R}$.
2. $\|\cdot\|_{\mathcal{B}} : \mathcal{B} \to [0, \infty)$ is a norm, i.e.,
   (a) *Positive definiteness:* $\|f\|_{\mathcal{B}} = 0 \iff f = 0$;
   (b) *Homogeneity:* $\|\lambda f\|_{\mathcal{B}} = |\lambda|\, \|f\|_{\mathcal{B}}$ for all $\lambda \in \mathbb{R}$ and $f \in \mathcal{B}$;
   (c) *Triangle inequality:* $\|f + g\|_{\mathcal{B}} \le \|f\|_{\mathcal{B}} + \|g\|_{\mathcal{B}}$ for all $f, g \in \mathcal{B}$;
3. $(\mathcal{B}, \|\cdot\|_{\mathcal{B}})$ is *complete*: every Cauchy sequence $(f_n)_{n\in\mathbb{N}} \subset \mathcal{B}$ converges in norm to some $f \in \mathcal{B}$, i.e. $\lim_{n\to\infty} \|f_n - f\|_{\mathcal{B}} = 0$.

It is worth mentioning that every Hilbert space is a Banach space.

In any separable (contains a countable, dense subset) Banach or Hilbert space, every Borel probability measure is a Radon measure (i.e., inner regular by compact sets and outer regular by open sets), which guarantees that probability mass can be approximated arbitrarily well by compact "core" sets—an essential property when approximating expectations numerically or proving convergence theorems (Bogachev, 2007). The formal definition of Radon measures is skipped, as it is unnecessary for the discussion of this thesis, where every Banach and Hilbert space is assumed to be separable.

**Definition 2.16** (Gaussian measure, (Stroock, 2010))**.** A Borel probability measure $P$ on a Hilbert space $\mathcal{H}$ is *Gaussian* with mean $\mu \in \mathcal{H}$ and covariance operator $\Sigma\colon \mathcal{H} \to \mathcal{H}$ if every continuous linear functional $L : \mathcal{H} \to \mathbb{R}$ satisfies $L_\sharp P = \mathcal{N}(L\mu,\, L\Sigma L^*)$. Where $L_\# P$ denotes the *push-forward* (image) measure of $P$ under the map $L$. Formally, for any Borel set $A \subset \mathbb{R}$,

$$(L_\# P)(A) = P(L^{-1}(A)). \tag{2.48}$$

Thus, the statement $L_\# P = \mathcal{N}(L\mu,\, L\Sigma L^*)$ means that $L(F)$ is Gaussian with mean $L\mu$ and variance $L\Sigma L^*$ whenever $F \sim P$.

Informally, the above definition states that applying $L$ to a random draw $F \sim P$; the result is a real Gaussian random variable. Requiring $L_\sharp P$ to be Gaussian for *every* such $L : \mathcal{H} \to \mathbb{R}$ forces the entire random element to be "Gaussian in all directions", exactly the infinite-dimensional analogue of a multivariate normal vector. A Gaussian random element should have finite second moment (which equals the trace of the covariance operator (Da Prato and Zabczyk, 2014, Proposition 2.16)):

$$\mathbb{E}[\|F\|^2_{\mathcal{H}}] = \operatorname{tr}(\Sigma)\,. \tag{2.49}$$

If $\operatorname{tr}(\Sigma) = \infty$ the "expected norm" blows up, then no $\sigma$-additive probability measure on $\mathcal{H}$ can realise those one-dimensional Gaussians simultaneously. Bogachev's theorem (Bogachev, 1998) therefore states that a centred Gaussian *Radon* measure $\gamma = \mathcal{N}(0, \Sigma)$ exists *if and only* if the covariance operator satisfies $\operatorname{tr}(\Sigma) < \infty$. Intuitively, trace-class requires the eigenvalues of $\Sigma$ to decay fast enough so that the random element does not wander off to infinity along an ever-increasing number of orthogonal directions. In finite dimensions, any symmetric, positive-definite matrix defines a bona-fide Gaussian probability measure via its quadratic form. The situation changes dramatically in an infinite–dimensional Banach space $\mathcal{B}$. Many natural covariance forms, such as the identity or other non–compact operators, fail this trace–class test; yet their finite coordinate projections remain perfectly meaningful. One therefore introduces the notion of a *cylindrical Gaussian measure*: a finitely additive set function on the algebra generated by linear functionals $\{h \mapsto \langle e_i, h\rangle_H\}_{i\in\mathbb{N}}$ that assigns multivariate normal laws to every finite sub-collection (Bogachev, 1998). Cylindrical Gaussians always exist—no trace-class is required—and serve as indispensable "weak" probability objects when modelling phenomena such as space–time white noise or when working with unbounded covariance operators. They provide the correct finite-dimensional distributions while acknowledging that no $\sigma$–additive extension to the whole Borel $\sigma$–algebra of $\mathcal{H}$ is possible in the non–trace-class regime.

**Definition 2.17** (Cylindrical $\sigma$–algebra, (Bogachev, 1998))**.** Let $(\mathcal{H}, \langle\cdot,\cdot\rangle)$ be a (real) Hilbert space. For each $h \in \mathcal{H}$ define the coordinate map $L_h : \mathcal{H} \to \mathbb{R}$ by $L_h(x) = \langle x, h\rangle$. The *cylindrical $\sigma$–algebra* on $\mathcal{H}$, written $\mathscr{C}(\mathcal{H})$, is the smallest $\sigma$–algebra that makes all $L_h$ measurable; equivalently,

$$\mathscr{C}(\mathcal{H}) = \sigma\Big\{(L_{h_1}, \ldots, L_{h_k})^{-1}(B) :\ k \in \mathbb{N},\ h_1, \ldots, h_k \in \mathcal{H},\ B \in \mathscr{B}(\mathbb{R}^k)\Big\}. \tag{2.50}$$

Sets of the form $(L_{h_1}, \dots, L_{h_k})^{-1}(B)$ are called *cylinders*; they depend only on finitely many linear coordinates of $\mathcal{H}$.

**Definition 2.18** (Cylindrical Gaussian measure)**.** Let $\mathcal{H}$ be a separable Hilbert space, $m \in \mathcal{H}$, and $S \in \mathcal{L}^+(\mathcal{H})$ a bounded, self–adjoint, non–negative operator. Denote by $\mathscr{C}(\mathcal{H})$ the cylindrical $\sigma$–algebra. A set function $P : \mathscr{C}(\mathcal{H}) \to [0,1]$ is called a *cylindrical Gaussian measure with mean $m$ and covariance operator $S$*, $\mathrm{Gauss}_{\mathrm{cyl}}(m, S)$, if, for every finite family of vectors $h_1, \dots, h_k \in \mathcal{H}$, one has

$$(L_{h_1}, \dots, L_{h_k})_\sharp P \;=\; \mathcal{N}\big((\langle m, h_1\rangle, \dots, \langle m, h_k\rangle), [\langle h_j, S(h_i)\rangle]\big)\,. \tag{2.51}$$

#### 2.3.6.1 From Gaussian Processes to Gaussian Measures

Here, the construction transitions from the object most practitioners know—a GP defined by a mean function and a kernel—to its measure-theoretic avatar. A *cylindrical* Gaussian distribution in the underlying Hilbert space is built. The takeaway is practical: *once you have a kernel, you automatically have a measure, so all the machinery of Gaussian measures becomes available for analysing or approximating your model later on.*

The operator viewpoint adopted in this chapter allows us to move back and forth between the *kernel definition* of a Gaussian process and the *measure–theoretic* definition that underlies the theory of random elements in Banach spaces. Consider the zero–mean case and show how the kernel alone determines a cylindrical Gaussian distribution in the RKHS.

**Theorem 2.19.** *Let $K : \mathcal{X} \times \mathcal{X} \to \mathbb{R}$ be a positive-definite kernel, $\mathcal{H}$ its RKHS, and $\phi_x := K(\,\cdot\,, x) \in \mathcal{H}$. Define the centred cylindrical Gaussian measure*

$$P_{\mathrm{cyl}} \;=\; \mathrm{Gauss}_{\mathrm{cyl}}(0, I_{\mathcal{H}}) \quad \textit{on the cylindrical } \sigma\textit{-algebra } \mathscr{C}(\mathcal{H}). \tag{2.52}$$

*For every finite index set $X = \{x_1, \dots, x_m\} \subset \mathcal{X}$,*

$$(\delta_{x_1}, \dots, \delta_{x_m})_\# P_{\mathrm{cyl}} \;=\; \mathcal{N}(0,\, K(X, X))\,, \tag{2.53}$$

*so $P_{\mathrm{cyl}}$ reproduces exactly the finite-dimensional marginals of the process $f \sim \mathcal{GP}(0, K)$.* [Proof in Appendix C.1]

The measure $P_{\mathrm{cyl}}$ lives only on the cylindrical $\sigma$-algebra generated by finite collections of continuous linear functionals on $\mathcal{H}$; it need not be countably additive on all Borel sets of $\mathcal{H}$. It verifies that $I_{\mathcal{H}} \in L^+(\mathcal{H})$, but its trace is not finite when $\dim \mathcal{H} = \infty$. As a result, since $I_{\mathcal{H}} \notin L^+_{\mathrm{tr}}(\mathcal{H})$ (is not trace-class), $P_{\mathrm{cyl}}$ remains purely cylindrical unless $\mathcal{H}$ is embedded into a larger Banach space (Gross, 1967).

**Theorem 2.20.** *Let $(\mathcal{H}, \langle\cdot,\cdot\rangle_{\mathcal{H}})$ be a separable infinite-dimensional Hilbert space. Then there exist:*

1. *a separable Banach space $(\mathcal{B}, \|\cdot\|_{\mathcal{B}})$ that contains $\mathcal{H}$ as a dense subspace under $\|\cdot\|_{\mathcal{B}}$,*

2. *a continuous, injective embedding $i : \mathcal{H} \hookrightarrow \mathcal{B}$ such that the push-forward of the canonical cylindrical Gaussian on $\mathcal{H}$ with covariance $I_{\mathcal{H}}$,* $\text{Gauss}_{\text{cyl}}(0, I_{\mathcal{H}})$*, under $i$ extends* uniquely *to a centred Gaussian Radon probability measure $\gamma = \mathcal{N}_{\mathcal{B}}(0, J)$ on $\mathcal{B}$ with covariance operator*
$$J = i \circ i^* : \mathcal{B}^* \longrightarrow \mathcal{B}, \tag{2.54}$$
*where $i^* : \mathcal{B}^* \to \mathcal{H}$ is the adjoint operator of $i$.*

*such that the triple $(\mathcal{B}, \gamma, \mathcal{H})$ is an* abstract Wiener space*; that is, $\mathcal{H}$ is the Cameron–Martin space of $\gamma$.* [Proof in Appendix C.1]

The abstract Wiener space $(\mathcal{B}, \gamma, \mathcal{H})$ built in Theorem 2.20 provides a universal "background": the Radon measure $\gamma = \mathcal{N}_{\mathcal{B}}(0, J)$ is nothing but *white noise* on $\mathcal{H}$, now made $\sigma$–additive by relaxing the norm. Shifting this noise by any Cameron–Martin element and using the associated Paley-Wiener map (Bogachev, 1998) produces a Gaussian measure whose finite–dimensional marginals coincide with those of a GP.

**Theorem 2.21.** *Let $f \sim \mathcal{GP}(m, K)$, with $m : \mathcal{X} \to \mathbb{R}$ and $K : \mathcal{X} \times \mathcal{X} \to \mathbb{R}$. Let $m \in \mathcal{H}$, where $\mathcal{H}$ is the RKHS generated by $K$ with sections $\phi_x := K(\,\cdot\,, x)$. Let $(\mathcal{B}, \gamma, \mathcal{H})$ be the abstract Wiener space constructed in Theorem 2.20; write $i : \mathcal{H} \hookrightarrow \mathcal{B}$ for the embedding and let $J = i \circ i^*$ be its covariance operator so that $\gamma = \mathcal{N}_{\mathcal{B}}(0, J)$. Let $Z \sim \gamma$ and define $F := i(m) + Z$. Then:*

1. *$F$ is a $\mathcal{B}$-valued Gaussian random element with law*
$$F \sim \mathcal{N}_{\mathcal{B}}(i(m),\ J). \tag{2.55}$$

2. *Let $\mathcal{W} : \mathcal{H} \to L^2(\gamma)$ denote the Paley–Wiener map associated with $(\mathcal{B}, \gamma, \mathcal{H})$, i.e. for $h \in \mathcal{H}$, $\mathcal{W}(h)$ is the $\gamma$-measurable linear functional on $\mathcal{B}$ satisfying $\mathcal{W}(h)(i(g)) = \langle h, g \rangle_{\mathcal{H}}$ for all $g \in \mathcal{H}$. Then, for every finite set $X = \{x_1, \ldots, x_n\} \subset \mathcal{X}$,*
$$\big(\mathcal{W}(\phi_{x_1}), \ldots, \mathcal{W}(\phi_{x_n})\big)_{\sharp} \mathcal{N}_{\mathcal{B}}(i(m),\ J) \;=\; \mathcal{N}\big(m(X),\ K(X, X)\big). \tag{2.56}$$
*Equivalently, $\big(\mathcal{W}(\phi_{x_1})(F), \ldots, \mathcal{W}(\phi_{x_n})(F)\big) \sim \mathcal{N}\big(m(X),\ K(X, X)\big)$.*

Remark. *If, in addition, the point evaluations $\delta_x$ are continuous on $\mathcal{B}$ (i.e. $\delta_x \in \mathcal{B}^*$ for all $x \in \mathcal{X}$), then $\mathcal{W}(\phi_x) = \delta_x$ $\gamma$-a.s., and the display in* (2) *coincides with the push-forward by $(\delta_{x_1}, \ldots, \delta_{x_n})$.* [Proof in Appendix C.1]

In summary, every Gaussian process $\mathcal{GP}(m, K)$ can be represented as a Gaussian measure $\mathcal{N}_{\mathcal{B}}(i(m), J)$ on a Banach space obtained by embedding its RKHS and shifting the resulting white noise by the mean element $m$.

#### 2.3.6.2 From Gaussian Measures to Gaussian Processes

The companion subsection completes the round-trip. Starting with nothing more than abstract "white-noise" in a Banach space, it shows that colouring this noise with a positive operator $\Sigma$ and optionally shifting it by a mean function $m$ yields a random element whose finite-dimensional evaluations follow exactly the GP distribution one would write down with the kernel

$$K_\Sigma(x, x') \;=\; \langle \varphi_x, \Sigma(\varphi_{x'}) \rangle. \tag{2.57}$$

This construction is extremely flexible—tweaking $\Sigma$ lets you generate whole families of kernels while staying within the same Hilbert geometry—and it underpins the variational families and kernel-learning methods developed later in the thesis. Whenever Chapter 4.3 "optimizes $\Sigma$", the covariance of this Gaussian measure is tuned, confident that the result is still a legitimate GP.

From now on, we *begin* with a Gaussian measure—nothing more than white noise on the abstract Wiener space introduced in Theorem 2.20—and show how every choice of a positive operator $\Sigma$ and a Cameron–Martin shift $m$ turns that measure into a fully fledged Gaussian process. The construction is extremely flexible: by varying $\Sigma$, you obtain an entire family of covariance kernels while remaining within the same Hilbert geometry.

**Theorem 2.22.** *Let $K$ be a positive-definite kernel with RKHS $\mathcal{H}$ (sections $\phi_x := K(\cdot, x)$). Let $(\mathcal{B}, \gamma, \mathcal{H})$ be an abstract Wiener space with continuous embedding $i : \mathcal{H} \hookrightarrow \mathcal{B}$. Fix $m \in \mathcal{H}$ and $\Sigma \in \mathcal{L}^+(\mathcal{H})$.*

1. *$P := \mathcal{N}_\mathcal{B}(i(m),\; i\Sigma i^*)$ is a well-defined (Radon) Gaussian measure on $\mathcal{B}$ with mean $i(m)$ and covariance operator $C = i\Sigma i^*$.*
2. *Let $\mathcal{W} : \mathcal{H} \to L^2(\gamma)$ denote the Paley–Wiener map associated with $(\mathcal{B}, \gamma, \mathcal{H})$, i.e. for $h \in \mathcal{H}$, $\mathcal{W}(h)$ is the $\gamma$-measurable linear functional on $\mathcal{B}$ satisfying $\mathcal{W}(h)(i(g)) = \langle h, g \rangle_\mathcal{H}$ for all $g \in \mathcal{H}$. Then, for every finite set $X = \{x_1, \dots, x_n\} \subset \mathcal{X}$,*

$$\big(\mathcal{W}(\phi_{x_1}), \dots, \mathcal{W}(\phi_{x_n})\big)_\sharp P \;=\; \mathcal{N}\big(m(X),\, K_\Sigma(X, X)\big). \tag{2.58}$$

Remark. *If, in addition, the point evaluations $\delta_x$ are continuous on $\mathcal{B}$ (i.e. $\delta_x \in \mathcal{B}^*$ for all $x \in \mathcal{X}$), then $\mathcal{W}(\phi_x) = \delta_x$ $\gamma$-a.s., and the display in (2) coincides with the push-forward by $(\delta_{x_1}, \dots, \delta_{x_n})$.* [Proof in Appendix C.1]

Theorem 2.22 therefore shows that *coloring abstract white noise with $\Sigma$ and adding a shift $m$ is sufficient to construct the familiar Gaussian process $\mathcal{GP}(m, K_\Sigma)$*. Observe that the kernel

$$K_\Sigma(x, x') = \langle \phi_x, \Sigma(\phi_{x'}) \rangle_\mathcal{H} \tag{2.59}$$

need not coincide with the original kernel $K$; by varying the operator $\Sigma$ one obtains an entire family of Gaussian processes while working inside the *same* Hilbert space $\mathcal{H}$.

As an abuse of notation, instead of $\mathcal{N}_\mathcal{B}(i(m), i\Sigma i^*)$, we will use $\mathcal{N}(f|m, \Sigma)$ to denote the Gaussian law.

*Remark.* The zero-mean GP prior $\mathcal{GP}(0, K)$ is obtained from the dual formulation using $\mathcal{N}(f|0, I)$. Furthermore, given a set of observations $(\mathbf{X}, \mathbf{y})$ from a regression task with Gaussian noise $\sigma^2$, the GP posterior is obtained from the (posterior) Gaussian measure $P(f|\mathbf{y}) = \mathcal{N}(f|\mu^\star, \Sigma^\star)$ where:

$$\mu^\star = \textstyle\sum_{i=1}^N \alpha_i \phi_{\mathbf{x}_i} \,, \tag{2.60}$$

$$\Sigma^\star(\phi) = \phi - \textstyle\sum_{i=1}^N \sum_{j=1}^N \phi_{\mathbf{x}_i} \Lambda_{ij} \left\langle \phi_{\mathbf{x}_j}, \phi \right\rangle_{\mathcal{H}} \,, \tag{2.61}$$

with $\mathbf{\Lambda} = (K(\mathbf{X}, \mathbf{X}) + \sigma^2 \boldsymbol{I})^{-1} \in \mathbb{R}^{N\times N}$ and $\boldsymbol{\alpha} = \mathbf{\Lambda}\boldsymbol{y} \in \mathbb{R}^N$.

As seen in this chapter, the pair $(\mu, \Sigma)$ provides a complete parameterisation of Gaussian processes. We use $\mathcal{P}_{\mathcal{H}}$ to denote the set of Gaussian measures:

$$\mathcal{P}_{\mathcal{H}} = \Big\{ \mathcal{N}(f|\mu, \Sigma) \;:\; \mu \in \mathcal{H},\; \Sigma \in \mathcal{L}^+(\mathcal{H}) \Big\} \,. \tag{2.62}$$

## 2.4 Probably Approximately Correct Bounds

Rigorous performance guarantees are indispensable when one seeks to provide statistical–learning algorithms with the full status of *scientific* tools. Within the *Probably Approximately Correct* (PAC) framework of Vapnik and Chervonenkis, such guarantees are delivered by *PAC bounds*—inequalities that hold with high probability over the random draw of the training sample and relate the *true* (generalization) risk of a predictor to its *empirical* risk estimated on finite data. The aim of this section is to give a formally self-contained introduction to PAC bounds.

Let $(\mathbf{x}, y) \sim \nu$ be an unknown distribution on $\mathcal{X} \times \mathcal{Y}$, where $\mathcal{X} \subset \mathbb{R}^d$ denotes the input space and $\mathcal{Y}$ the label space. A *predictor* is a measurable map $f_{\boldsymbol{\theta}} \colon \mathcal{X} \to \mathcal{Y}$ indexed by a parameter $\boldsymbol{\theta} \in \boldsymbol{\Theta} \subseteq \mathbb{R}^m$. Performance is assessed through a bounded loss $\ell \colon \mathcal{Y}^2 \to [0, C]$, so that the (expected) *risk* is

$$L(\boldsymbol{\theta}) = \mathbb{E}_{(\mathbf{x},y)\sim\nu}\big[\ell(f_{\boldsymbol{\theta}}(\mathbf{x}), y)\big], \tag{2.63}$$

while the *empirical* risk on a sample $D = \{(\mathbf{x}_i, y_i)\}_{i=1}^n \overset{\text{i.i.d.}}{\sim} \nu^n$ is

$$\hat{L}(D, \boldsymbol{\theta}) = \tfrac{1}{n} \sum_{i=1}^n \ell(f_{\boldsymbol{\theta}}(\mathbf{x}_i), y_i). \tag{2.64}$$

A learning algorithm is a (possibly randomized) estimator $\hat{\boldsymbol{\theta}} = \hat{\boldsymbol{\theta}}(D)$. In empirical-risk minimization (ERM) one takes $\hat{\boldsymbol{\theta}}_{\text{ERM}} \in \arg\min_{\boldsymbol{\theta}\in\boldsymbol{\Theta}} \hat{L}(D, \boldsymbol{\theta})$.

For a finite hypothesis set $\boldsymbol{\Theta} = \{\boldsymbol{\theta}_1, \ldots, \boldsymbol{\theta}_M\}$, Hoeffding's inequality combined with the union bound yields, for any $\delta \in (0, 1)$, the foundational result of McAllester (1999) and Seeger (2002):

$$\mathbb{P}_{D\sim\nu^n}\left( L(\hat{\boldsymbol{\theta}}_{\text{ERM}}) \le \min_{\boldsymbol{\theta}\in\Theta}\; \hat{L}(D, \boldsymbol{\theta}) + C\sqrt{\tfrac{\log(M/\delta)}{2n}} \right) \ge 1 - \delta. \tag{2.65}$$

Although historically important, this bound is unsatisfactory in high- or infinite-dimensional settings because the complexity term scales with the logarithm of the number of hy-pothesis, $\log M$, and the argument treats ERM only implicitly via a worst-case analysis.

### 2.4.1 Uniform Convergence for Finite Hypothesis Classes

In order to build intuition for the more sophisticated PAC–Bayesian machinery that will occupy the remainder of this chapter, the analysis begins with the classical setting in which the hypothesis class is *finite*. Although this special case is arguably of limited practical relevance in modern large-scale learning, it is historically foundational and constitutes the simplest instance where one observes, in a crystalline manner, the tension between approximation and estimation.

Throughout this thesis the PAC–Bayes template "*with probability at least* $1-\delta$ *over the draw of the sample* $D \sim \nu^n$*, for all* $\boldsymbol{\theta} \in \boldsymbol{\Theta}$*, simultaneously ...*". will be used. Formally, this means that for any confidence level $\delta \in (0,1)$, the corresponding event $\textbf{Event}(\boldsymbol{\theta}, D, \delta)$ will hold for every $\boldsymbol{\theta} \in \boldsymbol{\Theta}$ unless you have drawn a "bad" data set $D \sim \nu^n$; that is, the event holds with probability at least $1-\delta$. Mathematically:

$$\forall \delta \in (0,1), \quad \mathbb{P}_{D\sim\nu^n}\Big(\bigcap_{\boldsymbol{\theta}\in\boldsymbol{\Theta}} \textbf{Event}(\boldsymbol{\theta}, D, \delta)\Big) \geq 1-\delta\,.$$

The same reading applies to other instances in the text where the "event" may be an inequality (e.g. $A \leq B$) or an implication (e.g. $A \Rightarrow B$).

**Theorem 2.23.** *Let* $\boldsymbol{\Theta}$ *be a finite hypothesis class with cardinality* $M$ *and let* $\ell$ *be a loss bounded in* $[0, C]$*. Then for any* $\delta \in (0,1)$*, with probability at least* $1-\delta$ *over the random draw of the sample* $D \sim \nu^n$*, the following* simultaneous *concentration inequality holds:*

$$\forall \boldsymbol{\theta} \in \boldsymbol{\Theta}: \qquad L(\boldsymbol{\theta}) \;\leq\; \hat{L}(D, \boldsymbol{\theta}) + C\sqrt{\frac{\log(M/\delta)}{2n}}. \tag{2.66}$$

[Proof in Appendix C.1]

Figure 2.9 shows an illustration of Theorem 2.23; the left panel shows how the union–bound complexity term $C\sqrt{\log(M/\delta)/(2n)}$ decays as $n^{-1/2}$ (with larger $M$ shifting the curve upward by $\sqrt{\log M}$), while the right panel depicts the uniform-convergence geometry in which, with probability at least $1-\delta$, all hypotheses lie within a diagonal band of width $C\sqrt{\log(M/\delta)/(2n)}$ around the identity line. Fix accuracy $\varepsilon \in (0, C)$ and confidence $\delta \in (0,1)$. From Hoeffding's inequality

$$\mathbb{P}\left(L(\boldsymbol{\theta}) - \hat{L}(D, \boldsymbol{\theta}) > \varepsilon\right) \leq \exp\left(-\frac{2n\varepsilon^2}{C^2}\right). \tag{2.67}$$

Solving for $n$ shows that it suffices to draw

$$n \;\geq\; \frac{2C^2}{\varepsilon^2}\left[\log M + \log(2/\delta)\right] \tag{2.68}$$

examples in order to guarantee $L(\hat{\boldsymbol{\theta}}_{\mathrm{ERM}}) \leq \hat{L}(D, \hat{\boldsymbol{\theta}}_{\mathrm{ERM}}) + \varepsilon$ with probability at least $1-\delta$. Inverting the relationship highlights the *logarithmic* dependence on $M$, a fact whose importance cannot be overstated: the magnitude of $\log M$ constitutes a statistical-complexity measure of the hypothesis class.

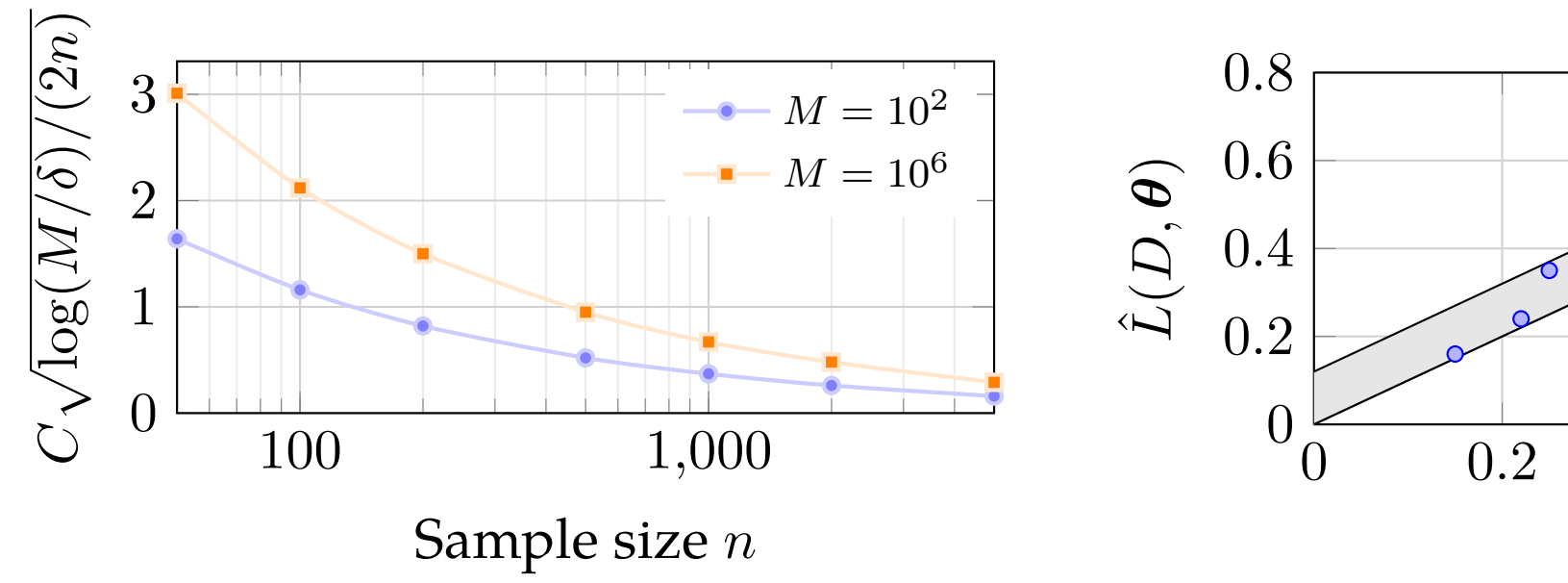

**Figure 2.9:** ***Finite-class learning curves and uniform-convergence geometry illustrate the classical PAC setting.*** *(left) For fixed class size $M$, the union-bound complexity term decays like $n^{-1/2}$; increasing $M$ merely raises the curve by a factor of $\sqrt{\log M}$. (right) Uniform convergence guarantees that, with probability $\geq 1-\delta$, every hypothesis lies inside a diagonal band of width $C\sqrt{\log(M/\delta)/(2n)}$ over the identity line.*

Despite its elegance, Theorem 2.23 suffers from two notorious shortcomings in light of contemporary machine learning:

1. it is vacuous in the ubiquitous scenario where $M$ is astronomically large or even countably (or uncountably) infinite, and
2. it treats $\hat{\boldsymbol{\theta}}_{\mathrm{ERM}}$ only *implicitly* through a worst-case argument, thereby failing to exploit any *data-dependent* structure of the empirical-risk landscape.

These deficiencies motivate a transition from naive union-bound reasoning to the more nuanced *PAC–Bayesian* paradigm—introduced independently by McAllester (1999) and Seeger (2002)—which replaces the combinatorial term $\log M$ with an *information-theoretic* complexity involving the Kullback–Leibler divergence between a posterior and a prior distribution over $\boldsymbol{\Theta}$.

### 2.4.2 The PAC–Bayesian Paradigm

The logarithmic-cardinality bound of Section 2.4.1 becomes uninformative once $|\boldsymbol{\Theta}|$ is large or infinite. The *PAC-Bayesian* paradigm circumvents this by trading the crude union bound for an information–theoretic control. Fix any *prior* distribution $\pi \in \mathcal{P}(\boldsymbol{\Theta})$ that is chosen independently of the sample $D$. After observing $D$ select an arbitrary *posterior* $\rho = \rho(D) \in \mathcal{P}(\boldsymbol{\Theta})$.

**Theorem 2.24.** [Catoni, 2007] *Let the loss satisfy $0 \leq \ell(\cdot,\cdot) \leq C$ and fix $\delta \in (0,1)$. For every $\lambda > 0$ the following event holds with probability at least $1-\delta$ over the draw of $D \sim \nu^n$:*

$$\forall \rho \in \mathcal{P}(\Theta): \quad \mathbb{E}_{\boldsymbol{\theta}\sim\rho}[L(\boldsymbol{\theta})] \;\leq\; \mathbb{E}_{\boldsymbol{\theta}\sim\rho}[\hat{L}(D,\boldsymbol{\theta})] + \frac{\lambda C^2}{8n} + \frac{\mathrm{KL}(\rho|\pi) + \log(1/\delta)}{\lambda}. \tag{2.69}$$

[Proof in Appendix C.1]

Theorem 2.24 states that for any prior $\pi$ on $\mathbf{\Theta}$, any data-dependent posterior $\rho$, and any $\lambda > 0$, with probability at least $1 - \delta$ over the draw of the sample $D \sim \nu^n$,

$$\mathbb{E}_{\boldsymbol{\theta}\sim\rho}[L(\boldsymbol{\theta})] \;\leq\; \mathbb{E}_{\boldsymbol{\theta}\sim\rho}[\hat{L}(D,\boldsymbol{\theta})] + \frac{\lambda C^2}{8n} + \frac{\mathrm{KL}(\rho|\pi) + \log(1/\delta)}{\lambda}. \tag{2.70}$$

Thus, the *true* (population) loss averaged under $\rho$ is, with high probability, no greater than the *empirical* loss averaged under $\rho$ plus a complexity penalty. The penalty has two parts: (i) a variance-like term $\lambda C^2/(8n)$, which decays as $1/n$, and (ii) a data-dependent term $(\mathrm{KL}(\rho|\pi) + \log(1/\delta))/\lambda$, which shrinks when $\rho$ stays close to the prior $\pi$ (small KL) and when looser confidence ($\delta$ larger) is acceptable. The parameter $\lambda$ balances these two terms; optimizing the bound in $\lambda$ yields

$$\lambda^\star \;=\; \frac{2\sqrt{2n(\mathrm{KL}(\rho|\pi) + \log(1/\delta))}}{C}, \tag{2.71}$$

and

$$\mathbb{E}_\rho[L(\boldsymbol{\theta})] \leq \mathbb{E}_\rho[\hat{L}(D,\boldsymbol{\theta})] + C\sqrt{\frac{\mathrm{KL}(\rho|\pi) + \log(1/\delta)}{2n}}. \tag{2.72}$$

Consequently, the *generalization gap* $\mathbb{E}_\rho[L(\boldsymbol{\theta})] - \mathbb{E}_\rho[\hat{L}(D,\boldsymbol{\theta})]$ scales as

$$O\big(\sqrt{(\mathrm{KL} + \log(1/\delta))/n}\big)\,. \tag{2.73}$$

Practically, the theorem justifies learning by minimizing the right-hand side over $\rho$: it leads to objectives of the form $\mathbb{E}_\rho[\hat{L}(D,\boldsymbol{\theta})] + \frac{1}{\lambda}\mathrm{KL}(\rho|\pi)$, i.e., empirical risk plus a PAC–Bayesian regularizer that keeps $\rho$ close to the prior. Tighter bounds are obtained when (a) $n$ is large, (b) the loss bound $C$ is small, and (c) the posterior $\rho$ concentrates near $\pi$ (small KL), reflecting low model complexity relative to the prior.

As the bound in Theorem 2.24 holds for every $\rho$, a principled data-dependent choice is the *Gibbs* posterior

$$\rho_\lambda(\mathrm{d}\boldsymbol{\theta}) := \frac{\exp[-\lambda\,\hat{L}(D,\boldsymbol{\theta})]\,\pi(\mathrm{d}\boldsymbol{\theta})}{\int_{\mathbf{\Theta}} \exp[-\lambda\,\hat{L}(D,\vartheta)]\,\pi(\mathrm{d}\vartheta)}.$$

Owing to the Donsker–Varadhan identity, $\rho_\lambda$ minimizes the right-hand side of the theorem inequality, so that

$$\mathbb{E}_{\rho_\lambda}[L(\boldsymbol{\theta})] \leq \inf_{\rho\in\mathcal{P}(\mathbf{\Theta})} \left\{ \mathbb{E}_\rho[\hat{L}(D,\boldsymbol{\theta})] + \frac{\lambda C^2}{8n} + \frac{\mathrm{KL}(\rho|\pi) + \log(1/\delta)}{\lambda} \right\}. \tag{2.74}$$

If $\mathbf{\Theta}$ is finite and $\pi$ is uniform, set $\lambda^\star := C^{-1}\sqrt{2n\log(2M/\delta)}$. Plugging $\lambda^\star$ into Equation (2.74) gives

$$\mathbb{E}_{\rho_{\lambda^\star}}[L(\boldsymbol{\theta})] \;\leq\; \min_{\boldsymbol{\theta}\in\mathbf{\Theta}} \hat{L}(D,\boldsymbol{\theta}) + C\sqrt{\frac{\log(2M/\delta)}{2n}}, \tag{2.75}$$

matching Theorem 2.23 up to the harmless two-sided-tail factor $\sqrt{2}$.

## 2.5 Conclusions

This chapter has established the theoretical and probabilistic foundations upon which the rest of the thesis builds. Beginning from a measure-theoretic formulation of probability, the basic language of random variables, distributions, and conditional dependencies that underlie all Bayesian inference is introduced. Within this framework, Bayes' theorem provides a mechanism to update prior beliefs in light of data, transforming uncertainty about model parameters into posterior distributions that drive predictive reasoning. The discussion contrasted Bayesian and frequentist perspectives, emphasizing the decomposition of predictive uncertainty into epistemic and aleatoric components, as well as the role of conjugate priors and computational approximations such as variational inference.

Subsequently, Gaussian processes were presented as non-parametric Bayesian models that define distributions directly over functions, offering a principled way to quantify uncertainty in predictions. Their dual interpretation—both as Bayesian linear models in feature space and as Gaussian measures in Hilbert space—reveals their centrality in unifying kernel methods, Bayesian inference, and function-space learning. Practical considerations such as hyperparameter learning, scalability via inducing points, and the use of approximate inference schemes were discussed, highlighting both their flexibility and computational challenges.

Finally, the chapter connected Bayesian inference to the theory of generalization through Probably Approximately Correct (PAC) and PAC–Bayesian bounds. These frameworks formalize how predictive performance can be bounded in expectation and how prior knowledge can regularize learning in a principled manner. Together, the concepts introduced here provide the mathematical and conceptual foundation for the methods developed in subsequent chapters, which extend Bayesian reasoning to deep architectures and analyze their generalization behavior in modern over-parameterized regimes.

# 3 Deep Variational Implicit Processes

Building on the theoretical background introduced in the previous chapter—particularly the Bayesian treatment of learning and the role of Gaussian process priors—this chapter explores a practical and scalable realization of function-space inference: the Deep Variational Implicit Process (DVIP). This framework extends the idea of implicit stochastic processes to hierarchical, deep architectures, allowing one to perform variational inference directly in function space without requiring explicit parametric densities. The result is a principled Bayesian approach that combines the flexibility of deep networks with the calibration and uncertainty-awareness of Gaussian processes.

## 3.1 Introduction

The Bayesian approach has become popular for capturing the uncertainty associated with the predictions made by models that otherwise provide point-wise estimates, such as NNs (Gal, 2016; Gelman et al., 2013; Murphy, 2012). However, when carrying out Bayesian inference, obtaining the posterior distribution in the space of parameters can become a limiting factor since it is often intractable. Symmetries and strong dependencies between parameters make the approximate inference problem much more complex. This is precisely the case in large deep NNs. Nevertheless, all these issues can be alleviated by carrying out approximate inference in the space of functions, which presents certain advantages due to the simplified problem. This makes the approximations obtained in this space more precise than those obtained in parameter space, as shown in the literature (Ma and Hernández-Lobato, 2021; Ma et al., 2019; Santana et al., 2021; Sun et al., 2019).

A recent method for function-space approximate inference is the Variational Implicit Process (VIP) (Ma et al., 2019). VIP considers an Implicit Process (IP) as the prior distribution over the target function. IPs constitute a very flexible family of priors over functions that generalize Gaussian processes (Ma et al., 2019). Specifically, IPs are processes that may lack a closed-form expression but that are easy to sample from. Examples include Bayesian neural networks (BNN), neural samplers, and warped GPs, among others (Santana et al., 2021). Nevertheless, the posterior process of an IP is intractable most of the time (except in particular cases as in GPs). VIP addresses this issue by approximating the posterior using the posterior of a GP with the same mean and covariances as the prior IP. Thus, the approximation used in VIP results in a Gaussian predictive distribution, which may be too restrictive.

Recently, the concatenation of stochastic processes has been used to produce models of increased flexibility. An example is DGPs in which a GP is used as the input of another GP, systematically (Damianou and Lawrence, 2013). Based on the success of DGPs, it is natural to consider the concatenation of IPs to extend their capabilities in a similar fashion to DGPs. Therefore, DVIP is introduced, a multi-layer extension of VIP that provides increased expressive power, enables more accurate predictions, gives better calibrated uncertainty estimates, and captures more complex patterns in the data. Each layer contains several IPs that are approximated using VIPs. Importantly, the flexibility of the IP-based prior formulation enables numerous models as the prior over functions, leveraging the benefits of *e.g.* convolutional NNs, that increase the performance on image datasets. Critically, DVIPs can adapt the prior IPs to the observed data, resulting in improved performance. When GP priors are considered, DVIPs are equivalent to a DGP. Thus, it can be seen as a generalization of DGPs.

Approximate inference in DVIPs is done via VI; where computational scalability is achieved in each unit using a linear approximation of the GP that approximates the prior IP, as in VIP (Ma et al., 2019). The predictive distribution of a VIP is Gaussian. However, since the inputs in the second and following layers are random in DVIPs, the final predictive distribution is non-Gaussian. This predictive distribution is intractable. Nevertheless, one can easily sample from it by propagating samples through the IP network. This also enables a Monte Carlo approximation of the VI objective which can be optimized using stochastic techniques, as in DGPs (Salimbeni and Deisenroth, 2017). Generating the required samples is straightforward given that the variational posterior depends only on the output of the previous layers. This results in an iterative sampling procedure that can be conducted in a scalable manner. Importantly, the direct evaluation of covariances is not needed in DVIPs, further reducing its cost compared to that of DGPs. The predictive distribution is a mixture of Gaussians (non-Gaussian), more flexible than that of VIP.

DVIPs are evaluated in several experiments, both in regression and classification. They show that DVIPs outperform a single-layer VIP with a more complex IP prior. DVIPs give results similar to and often better than those of DGPs (Salimbeni and Deisenroth, 2017), while having a lower cost and improved flexibility (due to the more general IP prior). The conducted experiments also show that adding more layers in DVIPs does not overfit and often improves results.

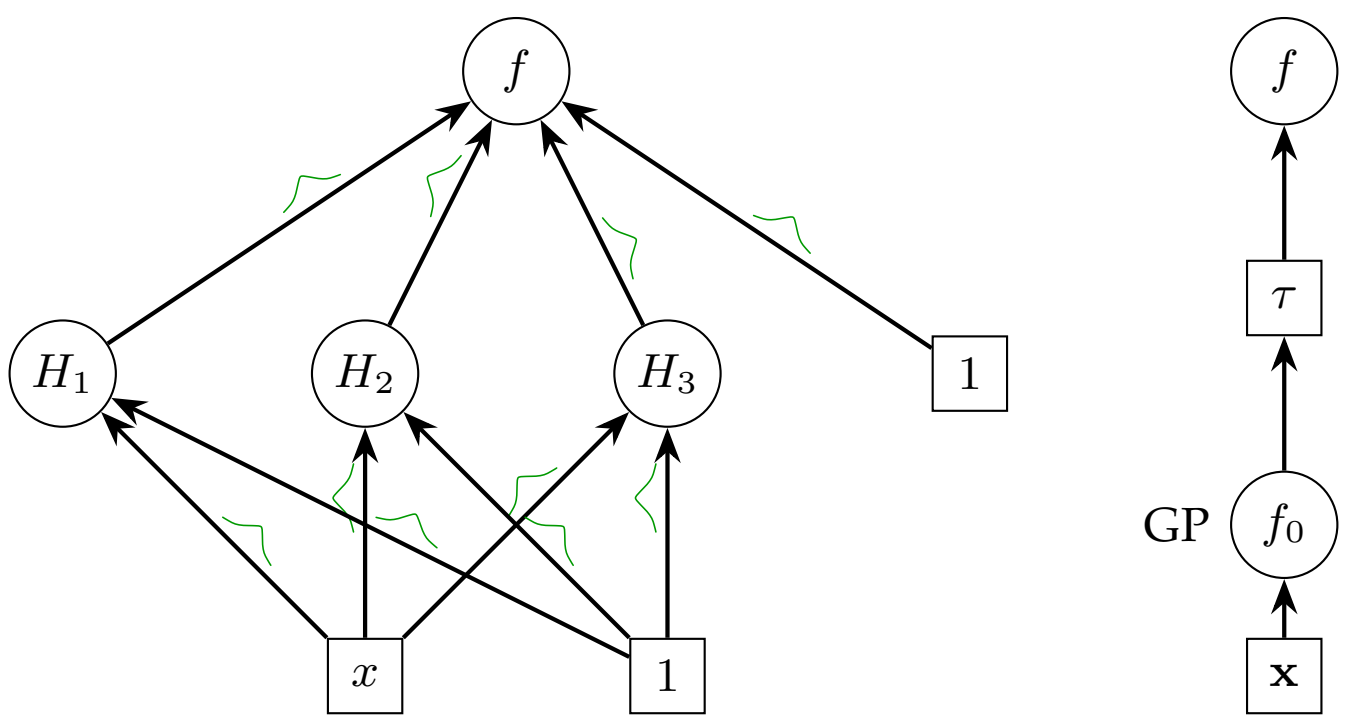

**Figure 3.1:** ***Graphical fragments of two prototypical implicit-process priors.*** Left: *A fully-connected neural-network. Circles (e.g. $H_1$–$H_3$, $f$) are latent random variables; squares are deterministic inputs or biases. Black arrows indicate conditional mappings, and a tiny green bell curve is superimposed on each edge to emphasize the underlying Gaussian component of the stochastic mapping.* Right: *A warped Gaussian process. The latent GP $f_0 \sim \mathrm{GP}(0, \kappa)$ receives the input $\mathbf{x}$ and is deterministically transformed by a non-linear warping $\tau : \mathbb{R} \to \mathbb{R}$, yielding the non-Gaussian output $f = \tau(f_0(\mathbf{x}))$. The label "GP" highlights that only the first stage is Gaussian; the composite process inherits the flexibility of the warp.*

## 3.2 Implicit Processes

The Gaussian assumption underlying classical GPs is arguably restrictive in modern machine–learning settings that involve highly non-linear, strongly heteroscedastic, or structured outputs. *Implicit Processes* (IPs) proposed by Ma et al. (2019) constitute a broad non-parametric family whose sample paths are obtained by transforming a *base* stochastic process through a measurable mapping. Formally,

**Definition 3.1** (Implicit process)**.** Let $(\Omega, \mathcal{F}, P)$ be a probability space and $g : \Omega \times \mathcal{X} \to \mathbb{R}$ a measurable map that is jointly continuous in the second argument. The collection of random variables $\{f(\mathbf{x}) = g(\omega, \mathbf{x}) : \mathbf{x} \in \mathcal{X}\}$ is an *implicit process*, written as $f \sim \mathrm{IP}(g, P)$.

Typical examples of implicit processes include neural networks and warped GPs. Figure 3.1 shows graphical representations of these two cases.

1. **Neural Networks**: where $g$ is the function that represents the neural network, $\omega$ is the set of weights and $\omega \sim P$ a distribution over them.
2. **Warped GPs**: where $g(\omega, \mathbf{x}) = \tau(f_0(\mathbf{x}))$ with $f_0 \sim \mathcal{GP}(0, K)$ and $\tau : \mathbb{R} \to \mathbb{R}$ a non-linear warping. Where the stochasticity of the process usually comes solely from the GP.

Unlike GPs, an IP generally lacks closed-form finite-dimensional distributions, yet Kolmogorov's extension theorem (Kolmogorov, 1956) guarantees the existence of a unique probability measure on the path space provided the consistency axioms hold. In practice, IPs are manipulated through *samples*; the absence of an explicit density requires approximate inference.

**Theorem 3.2** (Kolmogorov Extension Theorem (Kolmogorov, 1956))**.** *Let $\mathcal{X}$ be an arbitrary index set. For every finite subset $\mathcal{J} \subset \mathcal{X}$ let $P_{\mathcal{J}}$ be a probability measure on* $(\mathbb{R}^{\mathcal{J}}, \mathcal{B}(\mathbb{R})^{\otimes \mathcal{J}})$ *such that for all finite* $\mathcal{I} \subset \mathcal{J} \subset \mathcal{X}$*,*

$$\pi_{\mathcal{I}\mathcal{J}\,\sharp} P_{\mathcal{J}} = P_{\mathcal{I}}, \qquad \pi_{\mathcal{I}\mathcal{J}} : \mathbb{R}^{\mathcal{J}} \longrightarrow \mathbb{R}^{\mathcal{I}} \text{ the natural coordinate projection.} \tag{3.1}$$

*Then there exists a* unique *probability measure* $P$ *on* $(\mathbb{R}^{\mathcal{X}}, \mathcal{B}(\mathbb{R})^{\otimes \mathcal{X}})$ *whose finite-dimensional marginals coincide with the given family* $\{P_{\mathcal{J}}\}_{\mathcal{J} \subset \mathcal{X}}$*.*

Recall the definition of an implicit process: $f(\mathbf{x}) = g(\omega, \mathbf{x})$ with $\omega \sim P$ and measurable $g : \Omega \times \mathcal{X} \to \mathbb{R}$. For any finite $\mathcal{J} = \{\mathbf{x}_1, \ldots, \mathbf{x}_k\}$ define the random vector

$$\mathbf{f}_{\mathcal{J}}(\omega) := \big(g(\omega, \mathbf{x}_1), \ldots, g(\omega, \mathbf{x}_k)\big) \in \mathbb{R}^{\mathcal{J}}. \tag{3.2}$$

Its distribution is $P_{\mathcal{J}} := P \circ \mathbf{f}_{\mathcal{J}}^{-1}$, well-defined because $g$ is measurable.

**Lemma 3.3.** *The family* $\{P_{\mathcal{J}}\}_{\mathcal{J} \subset \mathcal{X}}$ *satisfies the Kolmogorov Extension Theorem; that is, for every pair of finite index sets* $\mathcal{I} \subset \mathcal{J}$ *it verifies that* $\pi_{\mathcal{I}\mathcal{J}\,\sharp} P_{\mathcal{J}} = P_{\mathcal{I}}$*.* [Proof in Appendix C.2]

There exists a unique probability measure $P_{\text{IP}}$ on $\mathbb{R}^{\mathcal{X}}$ such that the finite-dimensional distributions of the implicit process $f(\cdot)$ coincide with $\{P_{\mathcal{J}}\}$. Consequently, $f$ is a stochastic process.

### 3.2.1 Variational Implicit Processes

VIPs build a GP prior distribution based on samples taken from the implicit process formulation. The resulting construction relies on a set of linear regression coefficients that encode how the IP samples are *combined*. More precisely, given $S$ i.i.d. draws $\{f_s\}_{s=1}^{S} \sim \text{IP}(g, P)$, empirical features can be created as:

$$\hat{\phi}(\mathbf{x}) = \frac{1}{\sqrt{S}} \big(f_0(\mathbf{x}) - \hat{m}(\mathbf{x}), \ldots, f_S(\mathbf{x}) - \hat{m}(\mathbf{x})\big)^T \tag{3.3}$$

with empirical mean $\hat{m}(\mathbf{x}) := S^{-1} \sum_s f_s(\mathbf{x})$. Define a Bayesian linear model using those centered features as $f(\mathbf{x}) = \hat{m}(\mathbf{x}) + \hat{\phi}(\mathbf{x})^{\mathsf{T}}\mathbf{a}$, $\mathbf{a} \sim \mathcal{N}(\mathbf{0}, \mathbf{I})$. The resulting underlying prior model is then a GP with mean and covariance functions defined as

$$m^{\star}(\mathbf{x}) = \frac{1}{S} \sum_{s=1}^{S} f_s(\mathbf{x})\,, \qquad \mathcal{K}^{\star}(\mathbf{x}_1, \mathbf{x}_2) = \phi(\mathbf{x}_1)^{\mathsf{T}} \phi(\mathbf{x}_2)\,. \tag{3.4}$$

The true posterior $P(\boldsymbol{a}|\mathcal{D})$ is intractable in classification problems and has a computational cost of $\mathcal{O}(NS^2)$ for regression. However, VI can be used to optimize a variational posterior $Q(\boldsymbol{a})$ by maximizing the ELBO.

$$\mathcal{L}_{\text{VIP}} = \mathbb{E}_{Q(\mathbf{f})}[\log P(\mathbf{y}|\mathbf{f})] - \text{KL}[Q(\mathbf{a})|P(\mathbf{a})], \qquad Q(\mathbf{a}) = \mathcal{N}(\boldsymbol{\mu}_{\mathbf{a}}, \mathbf{S}_{\mathbf{a}})\,, \tag{3.5}$$

by stochastic gradient ascent using mini-batches of size $B$; each iteration scales as $\mathcal{O}(BS^2)$. The optimal $(\boldsymbol{\mu}_{\mathbf{a}}, \mathbf{S}_{\mathbf{a}})$ minimizes the KL divergence between the surrogate and the intractable IP posterior. Empirically, VIP delivers calibrated predictive uncertainty on regression and Bayesian optimization benchmarks with $S \approx 100$ features (Ma et al., 2019).

Critically, the samples $f_s(\mathbf{x})$ keep the dependence w.r.t. the IP prior parameters $\boldsymbol{\omega}$ (e.g. the BNN parameters), which enables prior adaptation to the observed data in VIP (Ma et al., 2019). Given $Q(\boldsymbol{a})$, it is straightforward to make predictions. A limitation is, however, that the predictive distribution for $y$ is Gaussian in regression. More precisely, given a test point $\mathbf{x}^\star$, the induced distribution over function evaluations $Q(\mathbf{f})$ is Gaussian given by

$$Q(\mathbf{f}^\star) = \mathcal{N}(m(\mathbf{x}^\star) + \phi(\mathbf{x}^\star)^T \boldsymbol{m}\,, \boldsymbol{\phi}(\mathbf{x}^\star)^T \boldsymbol{S} \boldsymbol{\phi}(\mathbf{x}^\star))\,, \tag{3.6}$$

which is straightforward given the linear approximation $f(\mathbf{x}) = \boldsymbol{\phi}(\mathbf{x})^{\mathrm{T}}\boldsymbol{a} + m^\star(\mathbf{x})$ and the variational distribution $Q(\boldsymbol{a}) \sim \mathcal{N}(\boldsymbol{m}, \boldsymbol{S})$. Using a Gaussian likelihood $P(y|\mathbf{f}) = \mathcal{N}(0, \sigma^2)$, it leads to the predictive distribution

$$\begin{aligned} P(y^\star|\mathbf{x}^\star) &= \int P(y^\star|\mathbf{f}^\star) Q(\mathbf{f}^\star)\, \mathrm{d}\mathbf{f}^\star \\ &= \mathcal{N}(y^\star; m(\mathbf{x}^\star) + \phi(\mathbf{x}^\star)^T \boldsymbol{m}\,, \boldsymbol{\phi}(\mathbf{x}^\star)^T \boldsymbol{S} \boldsymbol{\phi}(\mathbf{x}^\star) + \sigma^2)\,. \end{aligned} \tag{3.7}$$

Although a Gaussian predictive distribution has the advantage of being tractable in closed form, the main drawback is that Gaussian distributions may not be flexible enough to explain complex data patterns.

### 3.2.2 Computational Complexity

If not for the linear regression approximation, the computational complexity of VIPs would match that of standard GPs, being cubic on training and quadratic on prediction. However, this approach is inapplicable to datasets with thousands of points.

The usage of the linear approximation simplifies the training complexity using stochastic optimization and mini-batches to be $\mathcal{O}(BS^2D + BSC)$ where $B$ denotes the size of the mini-batch, $S$ the number of prior samples, $D$ the data targets dimensionality, and $C$ the cost of taking a prior sample. In most cases, the second term (linear on $S$) would be much smaller than the quadratic term. However, given that in practice $S$ is relatively small, using more complex prior functions would lead to the linear term being the most significant.

## 3.3 Deep Variational Implicit Processes

Although sparse Gaussian process approximations alleviate the $\mathcal{O}(N^3)$ computational burden, a single kernel with a fixed similarity metric rarely captures the heterogeneous structure present in real–world data. One remedy is to *compose* multiple GPs in a feed-forward hierarchy, yielding a *deep Gaussian process* (DGP) (Damianou and Lawrence, 2013). Such compositions adapt the notion of similarity at each depth, but introduce strong nonlinearities and destroy analytic tractability—the predictive distribution need no longer be Gaussian. Approximate inference therefore relies on (i) expectation propagation with sparse inducing variables (T. Bui et al., 2016) or (ii) doubly-stochastic variational Bayes, which couples Monte–Carlo sampling with stochastic optimization (Salimbeni and Deisenroth, 2017).

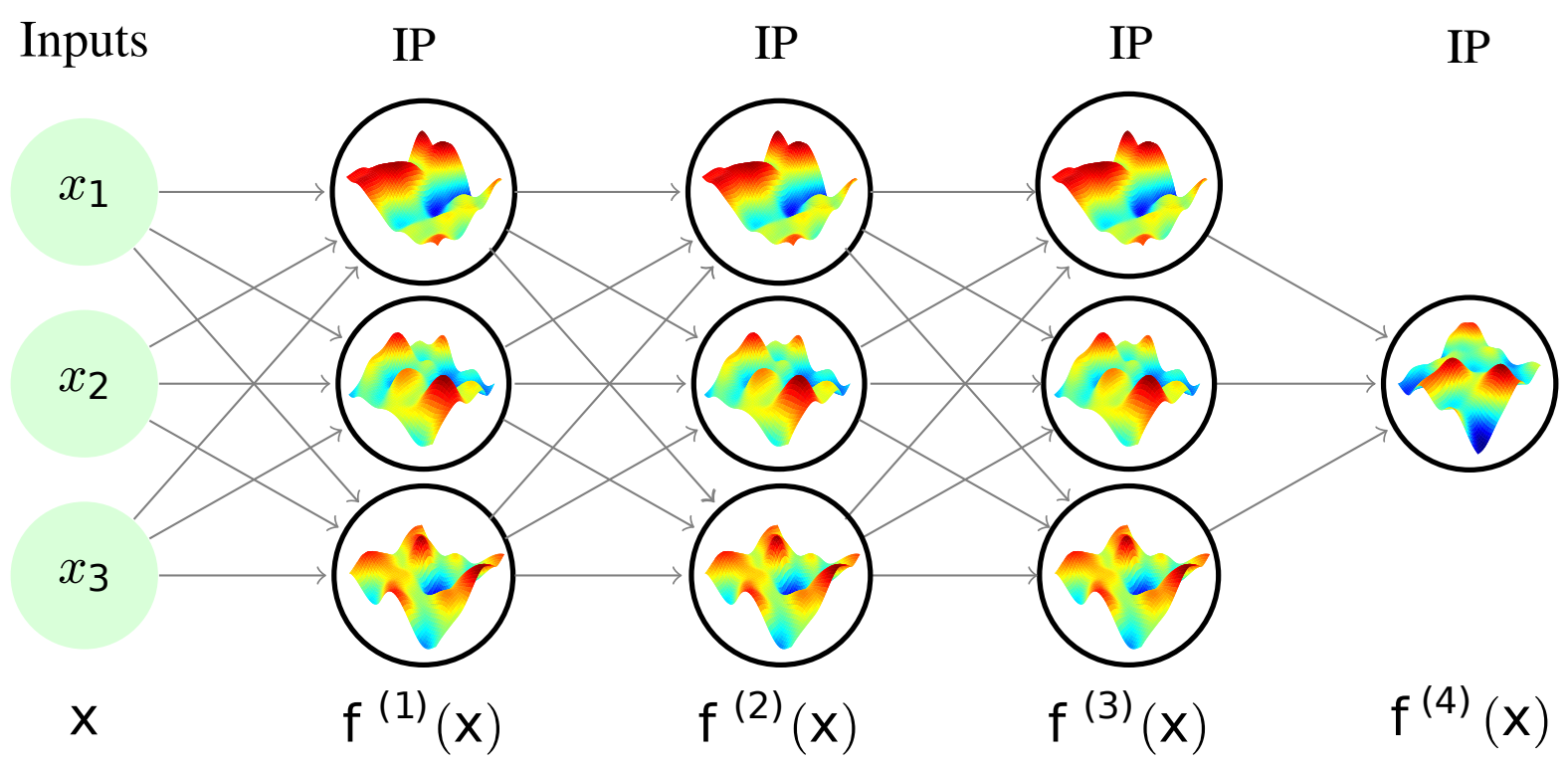

**Figure 3.2:** ***Fully-connected Deep Variational Implicit Process*** **(*DVIP*)**. *An input vector* $\mathbf{x}$ *is successively transformed by* $L$ *hidden layers, each comprising* $H_l$ *independent implicit-process (IP) units. Within a layer, every unit receives the entire output of the previous layer, so the composite mapping* $\mathbf{x} \mapsto \mathbf{f}^L(\mathbf{x})$ *forms a hierarchical, non-parametric prior whose representation capacity grows with depth.*

The IP prior of Ma et al. (2019) can be elevated to a deep setting. A *deep implicit process* (DIP) concatenates $L$ independent IP layers, thereby defining a hierarchical, non-parametric prior over latent functions; the resulting model is termed a *deep variational implicit process* (DVIP) once variational inference is applied. Figure 3.2 depicts the fully connected architecture considered in this work.

Let $H_l$ be the width of layer $l$. Denote by $\mathbf{f}^l_{\cdot,n} \in \mathbb{R}^{H_l}$ the layer-$l$ activation for input $\mathbf{x}_n$, and write $f^l_h(\cdot)$ for the $h$-th unit at that layer.

**Definition 3.4** (Deep implicit process)**.** Fix $L \in \mathbb{N}$ and widths $(H_1, \ldots, H_L) \in (\mathbb{N}^L$. A *deep implicit process* is the collection of random variables $\{f^l_{h,n} : l = 1, \ldots, L,\ h = 1, \ldots, H_l,\ n = 1, \ldots, N\}$ defined recursively by

$$\mathbf{f}^0_{\cdot,n} = \mathbf{x}_n, \qquad f^l_{h,n} = f^l_h(\mathbf{f}^{l-1}_{\cdot,n}), \tag{3.8}$$

where each latent function $f^l_h(\cdot) \sim \mathcal{IP}(g^l_h, P^l_h)$ is an *independent* implicit process driven by generator $g^l_h$ and latent distribution $P^l_h$.

As in VIPs, GP approximations are considered for all the IPs in the deep IP prior defined above. This provides an expression for $f^l_{h,n}$ given the previous layer's output $\mathbf{f}^{l-1}_{\cdot,n}$ and $\boldsymbol{a}^l_h$, the coefficients of the linear model for the unit $h$ at layer $l$. Namely,

$$f^l_{h,n} = \boldsymbol{\phi}^l_h(\mathbf{f}^{l-1}_{\cdot,n})^T \boldsymbol{a}^l_h + m^\star_{h,l}(\mathbf{f}^{l-1}_{\cdot,n}), \tag{3.9}$$

where $\boldsymbol{\phi}^l_h(\cdot)$ and $m^\star_{h,l}(\cdot)$ depend on the prior IP parameters $\boldsymbol{\theta}^l_h$. To increase the flexibility of the model, latent Gaussian noise is added around each inner $f^l_{h,n}$ with variance $\sigma^2_{l,h}$ for $l \in \{1, \ldots, L-1\}$, that is, all inner layers. From now on $\boldsymbol{\sigma}^2_l$ will denote the set of unit noises at layer $l$, thus, $\boldsymbol{\sigma}_l^2 = \{\sigma_{l,h}^2\}_{h=1}^{H_l}$. As a result of these approaches, $P(f_{h,n}^l|\mathbf{f}_{\cdot,n}^{l-1}, \boldsymbol{a}_h^l)$ is Gaussian with a linear mean in terms of $\boldsymbol{a}_h^l$: More precisely, $\forall l \in \{1,\dots,L-1\}$,

$$P(f_{h,n}^l|\mathbf{f}_{\cdot,n}^{l-1}, \boldsymbol{a}_h^l) = \mathcal{N}\left(f_{h,n}^l|\, m_{h,l}^\star(\mathbf{f}_{\cdot,n}^{l-1}) + \boldsymbol{\phi}_h^l(\mathbf{f}_{\cdot,n}^{l-1})^{\mathrm{T}}\boldsymbol{a}_h^l\,, \sigma_{l,h}^2\right) \tag{3.10}$$

and linearly deterministic for the last layer

$$f_{h,n}^L = m_{h,L}^\star(\mathbf{f}_{\cdot,n}^{L-1}) + \boldsymbol{\phi}_h^L(\mathbf{f}_{\cdot,n}^{L-1})^{\mathrm{T}}\boldsymbol{a}_h^L\,. \tag{3.11}$$

Let $\boldsymbol{a}^l = \{\boldsymbol{a}_1^l,\dots,\boldsymbol{a}_{H_l}^l\}$ and $\mathbf{f}^{\,l} = \{\mathbf{f}_{\cdot,1}^{\,l},\dots,\mathbf{f}_{\cdot,N}^{\,l}\}$. The joint distribution of all the variables (observed and latent) in DVIP is

$$P(\mathbf{y},\{\mathbf{f}^{\,l},\boldsymbol{a}^l\}_{l=1}^L) = \underbrace{\prod_{n=1}^N P(y_n|\mathbf{f}_{\cdot,n}^{\,L})}_{\text{Likelihood}}\underbrace{\prod_{n=1}^N\prod_{l=1}^L\prod_{h=1}^{H_l} P(f_{h,n}^l|\mathbf{f}_{\cdot,n}^{\,l-1},\boldsymbol{a}_h^l)P(\boldsymbol{a}_h^l)}_{\text{DVIP prior}}\,, \tag{3.12}$$

where the prior for the linear coefficients is set to a standard Gaussian, i.e, $P(\boldsymbol{a}_h^l) = \mathcal{N}(\boldsymbol{a}_h^l|\mathbf{0},\mathbf{I})$. It may seem that the prior assumes independence across points. However, dependencies arise by marginalizing out each $\boldsymbol{a}_h^l$, which is tractable since the model is linear in $\boldsymbol{a}_h^l$ (this marginalization can be seen as a convolution of Gaussians).

The posterior over the latent variables $P\left(\{\mathbf{f}^{\,l},\boldsymbol{a}^l\}_{l=1}^L|\mathbf{y}\right)$ is approximated using a similar approximation to that of deep GPs (Salimbeni and Deisenroth, 2017). This approximation has a fixed part and a tunable part, simplifying dependencies among layer units, but maintaining dependencies between layers:

$$Q(\{\mathbf{f}^{\,l},\boldsymbol{a}^l\}_{l=1}^L) = \prod_{n=1}^N\prod_{l=1}^L\prod_{h=1}^{H_l} P(f_{h,n}^l|\mathbf{f}_n^{l-1},\boldsymbol{a}_h^l)Q(\boldsymbol{a}_h^l)\,,\quad Q(\boldsymbol{a}_h^l) = \mathcal{N}(\boldsymbol{a}_h^l|\boldsymbol{m}_h^l,\boldsymbol{S}_h^l)\,, \tag{3.13}$$

where the factors $P(f_{h,n}^l|\mathbf{f}_n^{l-1},\boldsymbol{a}_h^l)$ are fixed to be the same factors as those in Equation (3.12), and the factors $Q(\boldsymbol{a}_h^l)$ are the ones being specifically tuned. Computing $Q(\mathbf{f}_{\cdot,n}^{\,L})$, as specified by Equation (3.13) is intractable. However, one can easily sample from $Q(\mathbf{f}_{\cdot,n}^{\,L})$ by propagating samples through the network.

**Proposition 3.5.** *Let $\boldsymbol{\Omega} = \{\boldsymbol{m}_h^l,\boldsymbol{S}_h^l : l=1,\dots,L\,, h=1,\dots,H_l\}$ denote the parameters of $Q$ and $\boldsymbol{\Theta} = \{\boldsymbol{\theta}_h^l : l=1,\dots,L\,, h=1,\dots,H_l\}$ the deep IP prior parameters. Then, a variational lower bound can be derived; at whose maximum the KL divergence between $Q(\{\mathbf{f}^{\,l},\boldsymbol{a}^l\}_{l=1}^L)$ and $P\left(\{\mathbf{f}^{\,l},\boldsymbol{a}^l\}_{l=1}^L|\mathcal{D}\right)$ is minimized*

$$\mathcal{L}\left(\boldsymbol{\Omega},\boldsymbol{\Theta},\{\boldsymbol{\sigma}_l^2\}_{l=1}^{L-1}\right) = \sum_{n=1}^N \mathbb{E}_{Q(\mathbf{f}_{\cdot,n}^{\,L})}[\log P(y_n|\mathbf{f}_{\cdot,n}^{\,L})] - \sum_{l=1}^L\sum_{h=1}^{H_l} KL(Q(\boldsymbol{a}_h^l)|P(\boldsymbol{a}_h^l))\,. \tag{3.14}$$

[Proof in Appendix C.2]

In the above proposition, $\mathrm{KL}(Q(\boldsymbol{a}_h^l)|P(\boldsymbol{a}_h^l))$ involves Gaussian distributions and can be evaluated analytically. The expectations $\mathbb{E}_{Q(\mathbf{f}_{\cdot,n}^{\,L})}\left[\log P(y_n|\mathbf{f}_{\cdot,n}^{\,L})\right]$ are intractable. How-ever, they can be approximated via Monte Carlo, using the reparameterization trick (Kingma and Welling, 2014). Moreover, $\sum_{n=1}^{N} \mathbb{E}_{Q(\mathbf{f}_{\cdot,n}^{L})}\left[\log P(y_n|\mathbf{f}_{\cdot,n}^{L})\right]$ can be approximated using mini-batches of size $B$:

$$\mathcal{L}\left(\mathbf{\Omega}, \mathbf{\Theta}, \{\boldsymbol{\sigma}_l^2\}_{l=1}^{L-1}\right) \approx \frac{N}{B}\sum_{b=1}^{B} \mathbb{E}_{Q(\mathbf{f}_{\cdot,b}^{L})}[\log P(y_b|\mathbf{f}_{\cdot,b}^{L})] - \sum_{l=1}^{L}\sum_{h=1}^{H_l} \mathrm{KL}(Q(\boldsymbol{a}_h^l)|P(\boldsymbol{a}_h^l))\,. \quad (3.15)$$

The consequence is that Equation (3.15) can be maximized w.r.t. $\mathbf{\Omega}$, $\mathbf{\Theta}$ and $\{\boldsymbol{\sigma}_l^2\}_{l=1}^{L-1}$, using stochastic optimization methods. Maximization w.r.t. $\mathbf{\Theta}$ allows for prior adaptation to the observed data, which is a key factor when considering IP priors.

### 3.3.1 Sampling from the Marginal Posterior Approximation

The approximation of the variational ELBO in Equation (3.15) requires samples from $Q(\mathbf{f}_{\cdot,n}^{L})$ for all the instances in a mini-batch. This marginal only depends on the variables of the inner layers and units $f_{h,n}^{l}$corresponding to the $n^{th}$ instance. Therefore, samples from $Q(\mathbf{f}_{\cdot,n}^{L})$ can be drawn by recursively propagating samples from the first to the last layer, using $\mathbf{x}_n$ as the network input. Specifically, $Q(f_{h,n}^{l}|\mathbf{f}_{\cdot,n}^{l-1})$ is Gaussian as

$$\begin{aligned} Q(f_{h,n}^{l}|\mathbf{f}_{\cdot,n}^{l-1}) &= \int P(f_{h,n}^{l}|\mathbf{f}_{\cdot,n}^{l-1}, \boldsymbol{a}_h^l)Q(\boldsymbol{a}_h^l)\,\mathrm{d}\boldsymbol{a}_h^l \\ &= \mathcal{N}\left(f_{h,n}^{l}|\,\hat{m}_{h,n}^{l}(\mathbf{f}_{\cdot,n}^{l-1}), \hat{v}_{h,n}^{l}(\mathbf{f}_{\cdot,n}^{l-1})\right)\,. \end{aligned} \quad (3.16)$$

with mean and variance:

$$\begin{aligned} \hat{m}_{h,n}^{l}(\mathbf{f}_{\cdot,n}^{l-1}) &= \boldsymbol{\phi}_h^l(\mathbf{f}_{\cdot,n}^{l-1})^{\mathrm{T}}\boldsymbol{m}_h^l + m_{h,l}^{\star}(\mathbf{f}_{\cdot,n}^{l-1})\,, \\ \hat{v}_{h,n}^{l}(\mathbf{f}_{\cdot,n}^{l-1}) &= \boldsymbol{\phi}_h^l(\mathbf{f}_{\cdot,n}^{l-1})^{\mathrm{T}}\boldsymbol{S}_h^l\boldsymbol{\phi}_h^l(\mathbf{f}_{\cdot,n}^{l-1}) + \sigma_{l,h}^2\mathbb{I}[l \neq L]\,, \end{aligned} \quad (3.17)$$

where $\boldsymbol{m}_h^l$ and $\boldsymbol{S}_h^l$ are the parameters of $Q(\boldsymbol{a}_h^l)$. This decomposition can be easily shown as the joint variational posterior is

$$\begin{aligned} Q(\{\mathbf{F}^l\}_{l=1}^{L}) &= \prod_{n=1}^{N}\prod_{l=1}^{L}\prod_{h=1}^{H_L} Q(f_{h,n}^{l}|\mathbf{f}_{\cdot,n}^{l-1}) \\ &= \prod_{n=1}^{N}\prod_{l=1}^{L}\prod_{h=1}^{H_L} \int P(f_{h,n}^{l}|\mathbf{f}_{\cdot,n}^{l-1}, \mathbf{a}_h^l)Q(\mathbf{a}_h^l)\,\mathrm{d}\mathbf{a}_h^l \\ &= \prod_{n=1}^{N}\prod_{l=1}^{L}\prod_{h=1}^{H_L} \mathcal{N}\left(f_{h,n}^{l}|\hat{m}_{h,n}^{l}(\mathbf{f}_{\cdot,n}^{l-1}), \hat{v}_{h,n}^{l}(\mathbf{f}_{\cdot,n}^{l-1})\right)\,. \end{aligned} \quad (3.18)$$

As a result, the $n^{th}$ marginal of the final layer depends only on the $n^{th}$ marginals of the other layers. That is,

$$Q(f_n^L) = \int \prod_{l=1}^{L}\prod_{h=1}^{H_L} \mathcal{N}\left(f_{h,n}^{l}|\hat{m}_{h,n}^{l}(\mathbf{f}_{\cdot,n}^{l-1}), \hat{v}_{h,n}^{l}(\mathbf{f}_{\cdot,n}^{l-1})\right)\mathrm{d}\mathbf{f}_{\cdot,n}^{1}, \ldots, \mathrm{d}\mathbf{f}_{\cdot,n}^{L-1}\,. \quad (3.19)$$

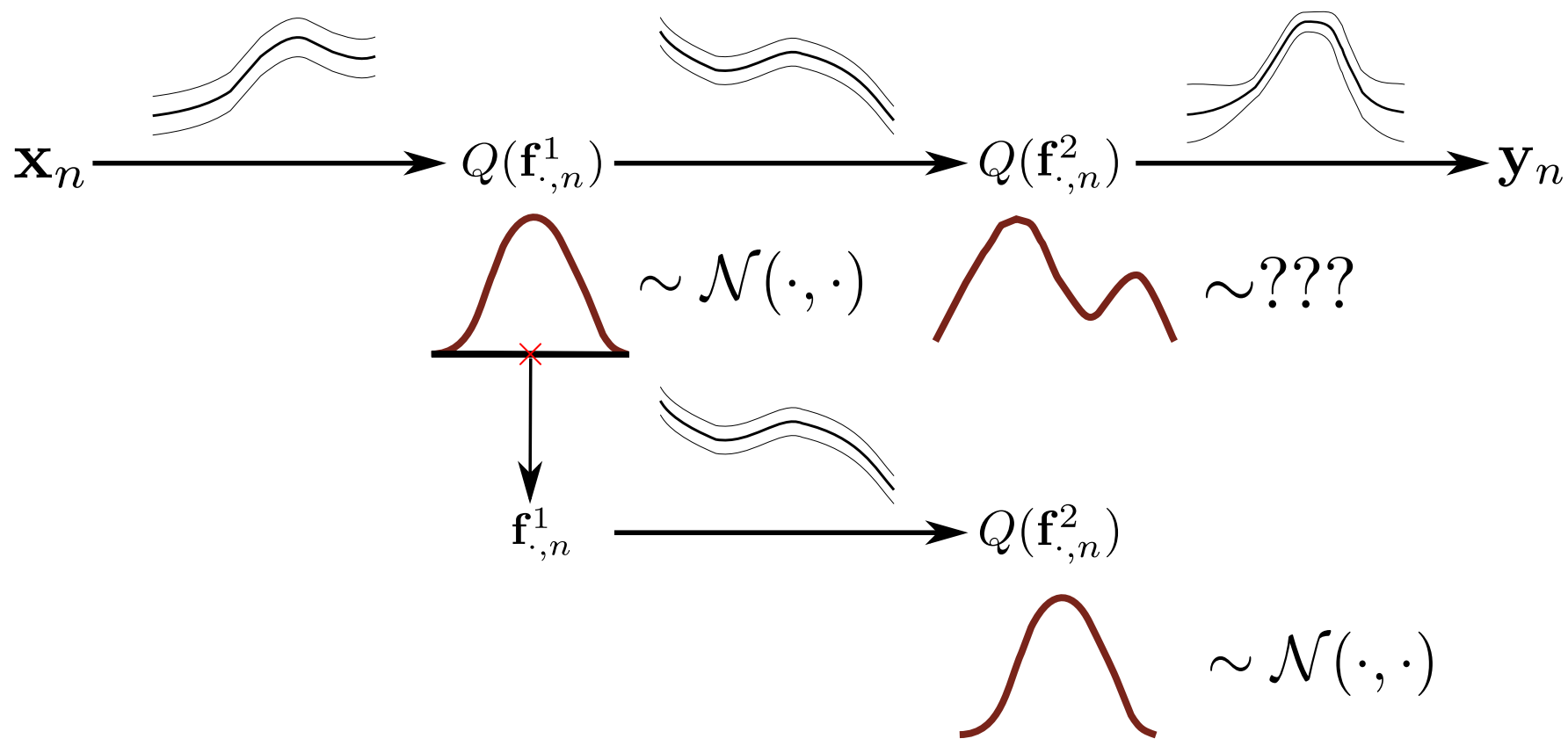

**Figure 3.3:** ***Sampling from DVIPs marginal posterior.*** *Starting with an input $\mathbf{X}_n$, compute the variational posterior of the first layer $Q(\mathbf{f}^1_{\cdot,n})$, and draw a latent function $\mathbf{f}^1_{\cdot,n}$. That sample is propagated through the next implicit layer, yielding a new posterior $Q(\mathbf{f}^2_{\cdot,n})$ and, ultimately, the predictive distribution over the observation $\mathbf{y}_n$.*

Initially, consider $l = 1$. Setting $\hat{\mathbf{f}}^0_{\cdot,n} = \mathbf{x}_n$ and generating a sample from $Q(f^l_{h,n}|\hat{\mathbf{f}}^0_{\cdot,n})$ for $h = 1, \ldots, H_l$. Let $\hat{\mathbf{f}}^l_{\cdot,n}$ be that sample. Then, use $\hat{\mathbf{f}}^l_{\cdot,n}$ as the input for the next layer. This process repeats for $l = 2, \ldots, L$, until obtaining $\hat{\mathbf{f}}^L_{\cdot,n} \sim Q(\mathbf{f}^L_{\cdot,n})$. Figure 3.3 showcases the employed sampling procedure.

### 3.3.2 Making Predictions for New Instances

Let $\mathbf{f}^L_{\cdot,\star}$ be the values at the last layer for the new instance $\mathbf{x}_\star$. The distribution $Q(\mathbf{f}^L_{\cdot,\star})$ is approximated by propagating $R$ Monte Carlo samples through the network. Then,

$$Q(\mathbf{f}^L_{\cdot,\star}) \approx \frac{1}{R}\sum_{r=1}^{R}\prod_{h=1}^{H_L}\mathcal{N}\left(f^L_{h,\star}|\hat{m}^L_{h,\star}(\hat{\mathbf{f}}^{L-1,r}_{\cdot,\star}), \hat{v}^L_{h,\star}(\hat{\mathbf{f}}^{L-1,r}_{\cdot,\star})\right), \tag{3.20}$$

where $\hat{\mathbf{f}}^{L-1,r}_{\cdot,\star}$ is the $r^{th}$ sample arriving at layer $L-1$ and $\hat{m}^L_{h,\star}(\cdot)$ and $\hat{v}^L_{h,\star}(\cdot)$ are given by Equation (3.17). Notice that Equation (3.20) is a mixture of $R$ Gaussians, which is expected to be more flexible than the predictive distribution of VIPs. In the conducted experiments $R$ is set to 100 for testing and to 1 for training. Computing, $P(y_\star) = \mathbb{E}_Q[P(y_\star|\mathbf{f}^L_{\cdot,\star})]$ is tractable in regression, and can be approximated using 1-dimensional quadrature in binary and multi-class classification, as in the case of DGPs (Salimbeni and Deisenroth, 2017). Assume a Gaussian likelihood for regression problems, $P(y_\star|\mathbf{f}^L_{\cdot,\star}) = \mathcal{N}(y_\star|\mathbf{f}^L_{\cdot,\star}, \boldsymbol{\sigma}^2_L\boldsymbol{I})$, where $\boldsymbol{\sigma}^2_L = (\sigma^2_{L,1}, \ldots, \sigma^2_{L,H_L})^T$, the expectation takes the form:

$$\begin{aligned} P(y_\star|\mathbf{x}_\star) &= \mathbb{E}_{Q(\mathbf{f}^L_{\cdot,\star})}[P(y_\star|\mathbf{f}^L_{\cdot,\star})] = \int Q(\mathbf{f}^L_{\cdot,\star})P(y_\star|\mathbf{f}^L_{\cdot,\star})\, d\mathbf{f}^L_{\cdot,\star} \\ &= \frac{1}{R}\sum_{r=1}^{R}\prod_{h=1}^{H_L}\mathcal{N}\Big(f^L_{h,\star}|\hat{m}^L_{h,\star}(\hat{\mathbf{f}}^{L-1,r}_{\cdot,\star}), \hat{v}^L_{h,\star}(\hat{\mathbf{f}}^{L-1,r}_{\cdot,\star}) + \sigma^2_{L,h}\Big). \end{aligned} \tag{3.21}$$

*Remark.* Using a Gaussian Likelihood is equivalent to considering Gaussian noise at the final layer similarly to the noise employed in the inner layers. The noise on the Gaussian likelihood is denoted as $\boldsymbol{\sigma}_L^2$ to highlight this fact.

### 3.3.3 Input Propagation

Inspired by the *skip layer* approach of deep convolutional networks, *e.g.* ResNet (He et al., 2016b), and the addition of the original input to each layer on DGPs (Duvenaud et al., 2014; Salimbeni and Deisenroth, 2017), the same approach is implemented on DVIPs. For this, the previous input is added to the mean in Equation (3.17) if the input and output dimensions of the layer are the same, except in the last layer where the mean does not change. In short,

$$\hat{m}^l_{h,n}(\mathbf{f}^{l-1}_{\cdot,n}) = \boldsymbol{\phi}^l_h(\mathbf{f}^{l-1}_{\cdot,n})^{\mathrm{T}} \boldsymbol{m}^l_h + m^{\star}_{h,l}(\mathbf{f}^{l-1}_{\cdot,n}) + f^{l-1}_{h,n} \mathbb{I}[l \neq L] \quad \forall l = 1, \ldots, L\,. \tag{3.22}$$

### 3.3.4 Computational Complexity

The computational cost at layer $l$ in DVIPs is $\mathcal{O}(BS^2H_l + BSC_l)$, where $S$ is the number of samples from the prior IPs, $B$ is the size of the mini-batch, $H_l$ is the output dimension of the layer, and $C_l$ is the cost of sampling from the prior. Notice the same number of prior samples is used for all the layers for simplicity. Moreover, only the diagonal of the covariance matrices is being computed, as in VIPs.

The first term of the cost is similar to that of a DGPs, which have a squared cost in terms of the number of inducing points (T. Bui et al., 2016; Hensman et al., 2013; Salimbeni and Deisenroth, 2017), notice that in that case, only the diagonal of the covariance matrix is computed; otherwise, the cost would be cubic. However, the number of prior IP samples $S$ is smaller than the typical number of inducing points in DGPs. In fact, the experiments carried out $S = 20$, as suggested for VIPs (Ma et al., 2019). Considering a DVIP with $L$ layers, the total cost is in $\mathcal{O}(BS^2(H_1 + \cdots + H_L) + BS(C_1 + \cdots + C_L))$. Where the second term is negligible unless heavy models are used as prior. Experiments show that, despite the generation of the prior IP samples using BNNs, DVIPs are faster compared to DGPs, and the gap between the two becomes bigger as $L$ increases.

## 3.4 Related Work

The relationship between GPs and IPs, such as BNNs, has been extensively studied recently. A single-layer BNN with cosine activations and infinite width is equivalent to a GP with RBF kernel (Hensman et al., 2017). A deep BNN is equivalent to a GP with a compositional kernel (Cho and Saul, 2009), as shown in Jaehoon Lee et al., 2018. These methods make it possible to create expressive kernels for GPs. An inverse reasoning is used in Flam-Shepherd et al., 2017 where the properties of a GP prior are encoded into the weight priors of a BNN.

As described in Section 3.2.1, VIPs (Ma et al., 2019) arise from the treatment of BNNs as instances of IPs. For this, an approximate GP is used to assist inference. Specif-ically, a prior GP is built with mean and covariance function given by the prior IP, a BNN. VIPs can make use of the more flexible IP prior, whose parameters can be inferred from the data, resulting in improved results over GPs (Ma et al., 2019). However, it is limited by the fact that VIP's predictive distribution is Gaussian, as a consequence of the GP approximation. Deep variational implicit processes (DVIPs) overcome this problem by providing a non-Gaussian predictive distribution. It is also expected to lead to a more flexible model with better calibrated uncertainty estimates.

Besides VIP there are other methods that have tried to make inference using IPs. In Sun et al., 2019 *functional Bayesian neural networks* (fBNN) are introduced, where a second IP is used to approximate the posterior of the first IP. This is a more flexible approximation than that of VIP. However, because both the prior and the posterior are implicit, the noisy gradient of the variational ELBO is intractable and has to be approximated. For this, a spectral gradient estimator is used (Shi et al., 2018). To ensure that the posterior IP resembles the prior IP in data-free regions, fBNN relies on uniformly covering the input space. In high-dimensional spaces this can lead to poor results. As a consequence of the spectral gradient estimator, fBNN cannot tune the prior IP parameters to the data. In the particular case of a GP prior, fBNN simply maximizes the marginal likelihood of the GP w.r.t. the prior parameters. However, a GP prior implies a GP posterior. This questions using a second IP for posterior approximation.

*Sparse implicit processes* (SIPs) consider an inducing point based approach for approximate inference in the context of IPs (Santana et al., 2021). SIP does not have the limitations of either VIP or fBNN, given that it produces flexible predictive distributions (Gaussian mixtures) and it can adjust the prior parameters to the data. SIP, however, relies on a classifier to estimate the KL-term in the variational ELBO, which adds computational cost. SIP's improvements over VIP are orthogonal to those of DVIP over VIP and, in principle, SIP could also be used as the building blocks of DVIP, leading to even better results.

*Functional variational inference* (FVI, Ma and Hernández-Lobato, 2021) minimizes the KL-divergence between stochastic processes for approximate inference using IPs. Specifically, between the model's IP posterior and a second IP, such as fBNN. This is done efficiently by approximating first the IP prior using a stochastic process generator (SPG). Then, a second SPG is used to efficiently approximate the posterior of the previous SPG.Both SPGs share key features that make this an easy task. However, FVI is also limited, like fBNN, in the sense that it cannot adjust the prior to the data, which questions its practical utility.

As shown in Santana et al., 2021, adjusting the prior IP to the observed data is key for accurate predictions. This questions using fBNN and FVI as the building blocks of a model using deep IP priors on the target function. Moreover, these methods do not consider deep architectures such as the one shown in Figure 3.2. For this reason, comparative results focus on comparing with VIP, as DVIP generalizes VIP.

The concatenation of IPs with the goal of describing priors over functions has not been studied previously in the literature. However, the concatenation of GPs, follow-ing a multi-layer architecture and resulting in deep GPs (DGPs), has received a lot of attention from the community (T. Bui et al., 2016; Cutajar et al., 2017; Havasi et al., 2018; Lawrence and Moore, 2007; Salimbeni and Deisenroth, 2017; H. Yu et al., 2019). In principle, DVIP is a generalization of DGPs in the same way as IPs generalize GPs. Namely, the IP prior of each layer's unit can simply be a GP. Samples from such a GP prior can be efficiently obtained using, *e.g.*, a 1-layer BNN with cosine activation functions that is wide enough (Cutajar et al., 2017; Rahimi and Recht, 2007). DGPs are hence restricted to GP priors, while DVIP has the extra flexibility of considering a wider range of IP priors that need not be GPs. As a matter of fact, in the conducted experiments, DVIP significantly outperforms DGPs in image-related datasets, where using specific IP priors based on convolutional neural networks, can give a significant advantage. Experiments also show that DVIP is faster to train than a DGP, and the difference becomes larger with the number of layers $L$.

## 3.5 Conducted Experiments

DVIPs have been evaluated on several tasks, including time series interpolation, regression, and classification problems. Unless stated otherwise, the number of linear regression coefficients is set to $S = 20$ and a BNN as the IP prior for each unit for all the layers. These BNNs have two hidden layers of $10$ units each with *tanh* as the activation function, as in Ma et al., 2019. The prior mean and variance of each weight and bias in the BNN have been tuned. As no regularizer is used for the prior parameters, the prior mean and variances are constrained to be the same in a layer of the BNN. This configuration avoids overfitting and leads to improved results. See Section 3.5.6.1 for further details on the matter.

To speed up computations, at each layer $l$ the generative function that defines the IP prior is shared across units. That is, the function $g_h^l$ is the same for every unit $h$ in that layer. As a consequence, the prior IP samples only need to be generated once per layer, as in Ma et al., 2019. This assumption is similar to that of DGPs, which assume shared GP prior parameters for each layer (Salimbeni and Deisenroth, 2017). Under these assumptions and simplifications, each VIP layer contains $12$ tunable parameters in its prior, which are the mean and variance of each weight and bias of the three layers in its BNN.

To evaluate the performance of DVIPs, the model is compared with VIPs (Ma et al., 2019) and DGPs, closely following Salimbeni and Deisenroth, 2017. In DGPs $100$, shared inducing points in each layer are considered. The employed optimization algorithm is ADAM (Kingma and Ba, 2015) with a learning rate $10^{-3}$, as in Ma et al., 2019. In DGPs, the learning rate is set to $10^{-2}$, as in Salimbeni and Deisenroth, 2017.

Following Salimbeni and Deisenroth, 2017, unless indicated otherwise, in DVIPs and DGPs the input dimension is used as the layer dimensionality for all the inner layers, *i.e* $H_l = M$, for $l = 1, \ldots, L-1$. In DGPs the kernel employed is RBF with ARD (Rasmussen and Williams, 2006). The batch size is $100$. All methods are trained for 150 000 iterations unless indicated otherwise.

The marginal likelihood regularizer described in Ma et al., 2019 is not employed in VIPs nor VIP layers since the authors indicated that they did not use it in practice. Black-box $\alpha$-divergences are used for VIPs with $\alpha = 0.5$, as suggested in Ma et al., 2019.

Results are not compared with fBNNs nor FVIs, described in Section 3.4, because these methods cannot tune the prior IP parameters to the data nor do they consider deep architectures such as the one shown in Figure 3.2 (right). The source code can be accessed at https://github.com/Ludvins/2023-ICLR-DVIP.

### 3.5.1 Regression UCI benchmarks

DVIPs are compared with DGPs and VIPs on 8 different regression datasets from the UCI Repository (Dua and Graff, 2019). Following common practice, the performance is validated on each dataset using 20 different train-test partitions of the data with 10% test size (J. M. Hernández-Lobato and Adams, 2015; Salimbeni and Deisenroth, 2017).

Both DVIPs and DGPs are tested using 2, 3, 4, and 5 layers. Moreover, VIPs, which are equivalent to DVIPs with $L = 1$, are tested with the same prior as DVIPs and using a bigger BNN of 200 units per layer. Furthermore, results are compared with a single sparse GPs, which is equivalent to DGPs for $L = 1$. Figure 3.4 shows the results obtained in terms of the negative test log-likelihood. Exact results are found in Table B.1.

Results show that DVIPs with at least 3 layers perform best on 4 out of the 8 datasets (Boston, Energy, Concrete and Power), having comparable results on Winered and Naval (all methods have zero RMSE on this dataset). DGPs perform best on 2 datasets (Protein and Kin8nm), but the differences are small. Adding more layers in DVIPs does not lead to overfitting and it gives similar and often better results (more notably on larger datasets: Naval, Protein, Power and Kin8nm). DVIPs also perform better than VIPs most of the time. By contrast, using a more flexible BNN prior in VIPs (*i.e.*, 200 units) does not improve results. Figure 3.5 shows the training time in seconds of each method. These execution times show that DVIPs are faster than DGPs and faster than VIPs with the 200 units BNN prior. Summing up, DVIPs achieve similar results to those of DGPs, but at a smaller cost.

### 3.5.2 Interpolation Results

As part of the empirical evaluation, experiments on the CO2 time-series dataset (https://scrippsco2.ucsd.edu) are carried out. This dataset consists of $CO_2$ measurements from the Mauna Loa Observatory, Hawaii, in 1978. The dataset is split into five consecutive and equal parts, with the 2nd and 4th parts used as missing testing data. All models are trained for 100 000 iterations. Figure 3.6 shows the predictive distribution of DVIPs and DGPs with $L = 2$ on the data. Results show that DVIPs can capture the data trend in the missing gaps. For DVIPs, samples from the learned prior are shown, which are very smooth. By contrast, a DGP with RBF kernels fails to capture the data trend, leading to mean reversion and over-estimation of the prediction uncertainty. Thus, the BNN prior considered by DVIP could be a better choice here. This issue of DGPs can be overcome using compositional kernels (Duvenaud et al., 2014), but that requires using kernel search algorithms.

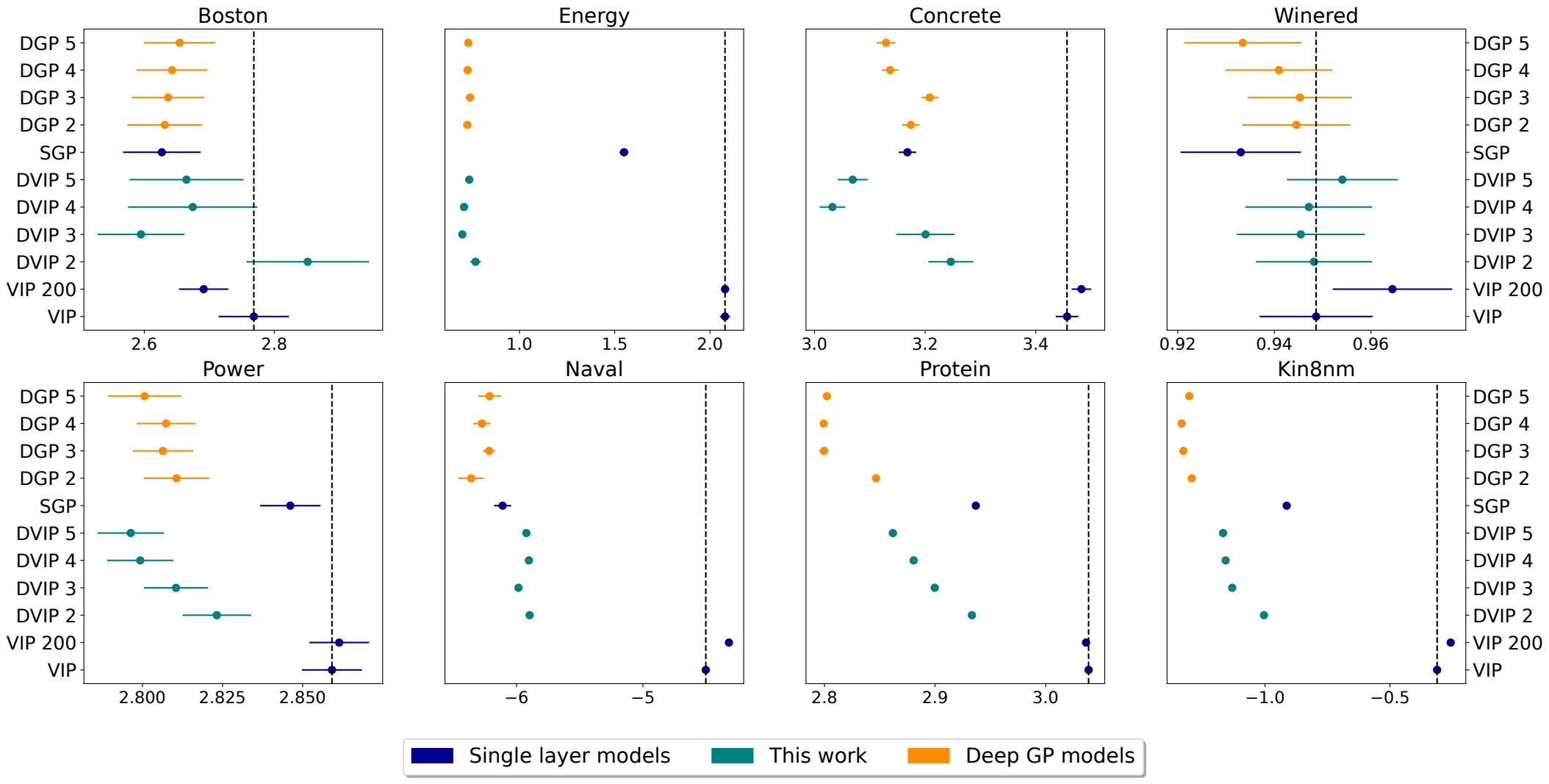

**Figure 3.4:** ***Negative test log–likelihood (NLL) on eight UCI regression benchmarks.*** *Each panel corresponds to one dataset (Boston, Energy, Concrete, WineRed, Power, Naval, Protein, and Kin8nm). Dots mark the mean NLL across 20 random train–test splits; horizontal bars indicate one standard error. Colors distinguish* single–layer models (*blue*), the proposed method (*green*), *and* deep GP baselines (*orange*). *The dashed vertical line in each panel shows VIP as baseline for reference. Lower values (further to the left) indicate better predictive performance.*

### 3.5.3 Large Scale Regression

Each method is evaluated on 3 large regression datasets. First, the Year dataset (UCI) (Bertin-Mahieux, 2011) with 515 345 instances and 90 features, where the original train/test splits are used. Second, the US flight delay (Airline) dataset (Dutordoir et al., 2020; Hensman et al., 2017), where following Salimbeni and Deisenroth, 2017 the first 700 000 instances are used for training and the next 100 000 for testing. A total of 8 features are considered: *month, day of month, day of week, plane age, air time, distance, arrival time and departure time*. Lastly, data recorded in January 2015 from the Taxi dataset is used. In this dataset 10 attributes are considered: *time of day, day of week, day of month, month, pickup latitude, pickup longitude, drop-off longitude, drop-off latitude, trip distance and trip duration*. Trips with a duration lower than 10 seconds and larger than 5 hours are removed as in Salimbeni and Deisenroth, 2017, leaving 12 695 289 instances. Results are averaged over 20 train/test splits with 90 % and 10 % of the data. Here, each method is trained for 500 000 iterations. The results obtained are shown in Table 3.1. The last column shows the best result by DGPs, which is achieved for $L = 3$ on each dataset.

It can be observed that DVIPs outperform VIPs on all datasets, and on Airline and Taxi, the best methods are DVIPs. In Taxi, SGPs and DGPs give similar results, while DVIPs improve over VIPs. The best method on Year, however, is DGPs. Since the dif-ference between DVIPs and DGPs is found in the prior, one could hypothesize that GPs with an RBF kernel may be a better prior on this dataset than the BNN IP prior considered by DVIPs. Therefore, since DVIPs generalize DGPs, DVIPs using GP priors should give similar results to those of DGPs on Year. This is further analyzed in Section 3.5.6.4.

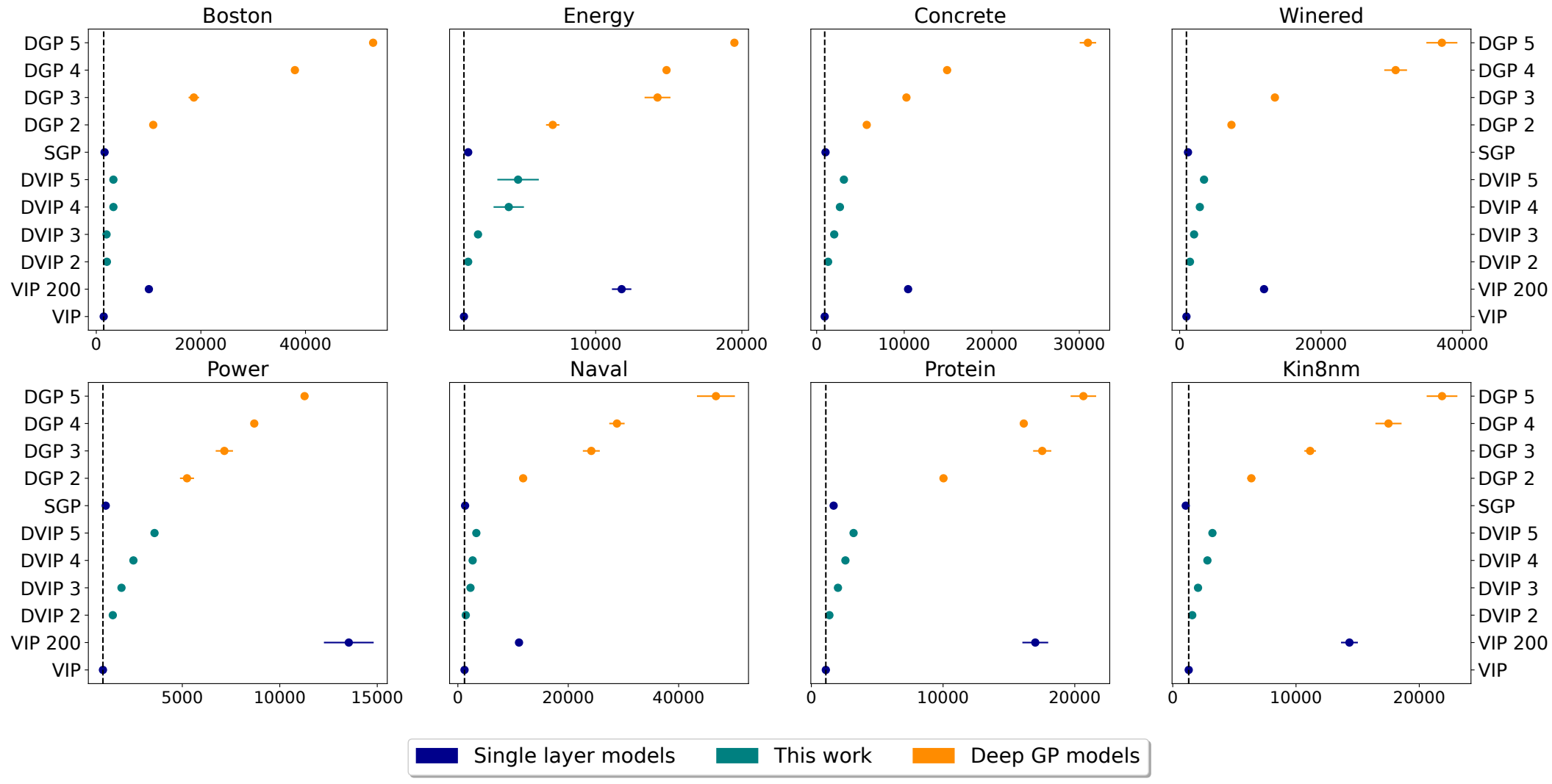

**Figure 3.5:** ***CPU training time on eight UCI regression benchmarks.*** *Each panel corresponds to one dataset (Boston, Energy, Concrete, WineRed, Power, Naval, Protein, and Kin8nm). Dots mark the mean NLL across 20 random train–test splits; horizontal bars indicate one standard error. Colors distinguish* single–layer models (*blue*), the proposed method (*green*), *and* deep GP baselines (*orange*). *The dashed vertical line in each panel shows VIP as baseline for reference. Lower values (further to the left) indicate lower training time.*

### 3.5.4 Image Classification

Binary classification experiments were carried out using the Rectangles dataset (Salimbeni and Deisenroth, 2017) and the multi-class dataset MNIST (L. Deng, 2012). Each dataset has $28 \times 28$ pixel images. The Rectangles dataset has 12 000 images of a (non-square) rectangle. The task is to determine if the height is larger than the width. For this dataset, a probit likelihood is employed for all the models. The MNIST dataset has 70 000 images of handwritten digits. The labels correspond with each digit. In this case, the robust-max multi-class likelihood (D. Hernández-Lobato et al., 2011) is used. Given the size of Rectangles, 20 000 iterations are enough to ensure convergence. Results on this dataset are averaged over 10 different random seeds, but the standard errors are omitted as they are always lower than 0.1. The original train-test splits are used for both datasets. Examples of images from these datasets are shown in Figure 3.7.

Critically, DVIP's capability to use more flexible priors is exploited. Specifically, in the first layer, a convolutional NN (CNN) prior with two layers of 4 and 8 channels, respectively, is employed. The inner dimensions of DVIPs and DGPs are fixed to 30, as in Salimbeni and Deisenroth, 2017. No input propagation is used in the first layer. The results obtained are shown in Table 3.2. These show that DVIPs obtain much better results than those of DGPs and VIPs. In particular, DVIPs increase the accuracy by 11% on Rectangles compared to DGPs, probably as a consequence of the CNN prior considered in the first layer of the network being more suited for image-based datasets.

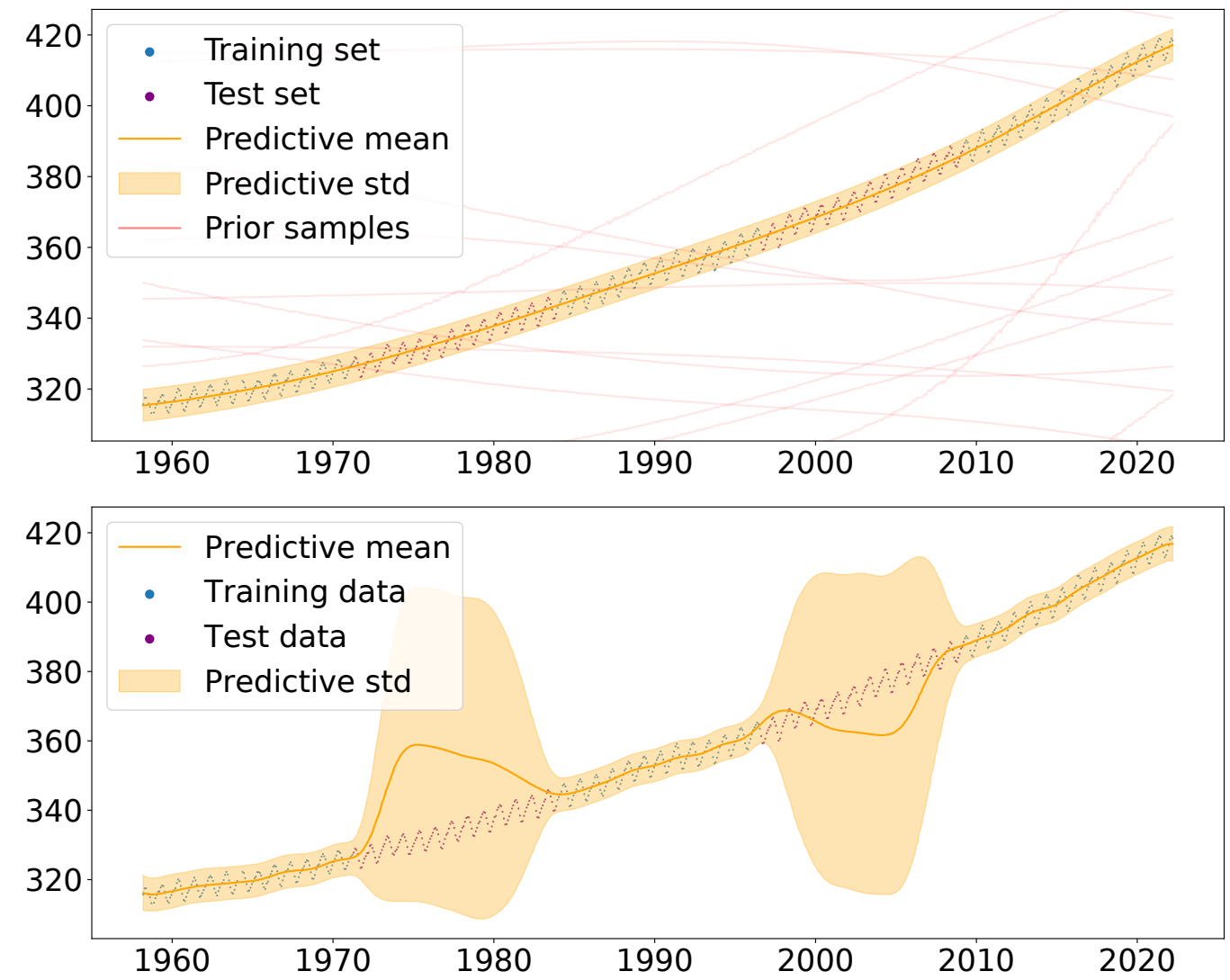

**Figure 3.6:** ***Missing–value interpolation on the Mauna Loa $CO_2$ time series.*** *Monthly $CO_2$ concentrations (1958–2022) are divided into five consecutive, equal–length segments; the 2nd and 4th segments (black crosses) are withheld as test gaps. Both models are trained for 100 000 iterations and use $L = 2$ layers.* Upper*: the proposed DVIP. The dashed red curve is the predictive mean, the orange band shows $\pm 2\sigma$, and faint red lines are samples from the learned (smooth) prior.* Lower*: a DGP with RBF kernels. The model suffers from mean-reversion in the gaps between the training data.*

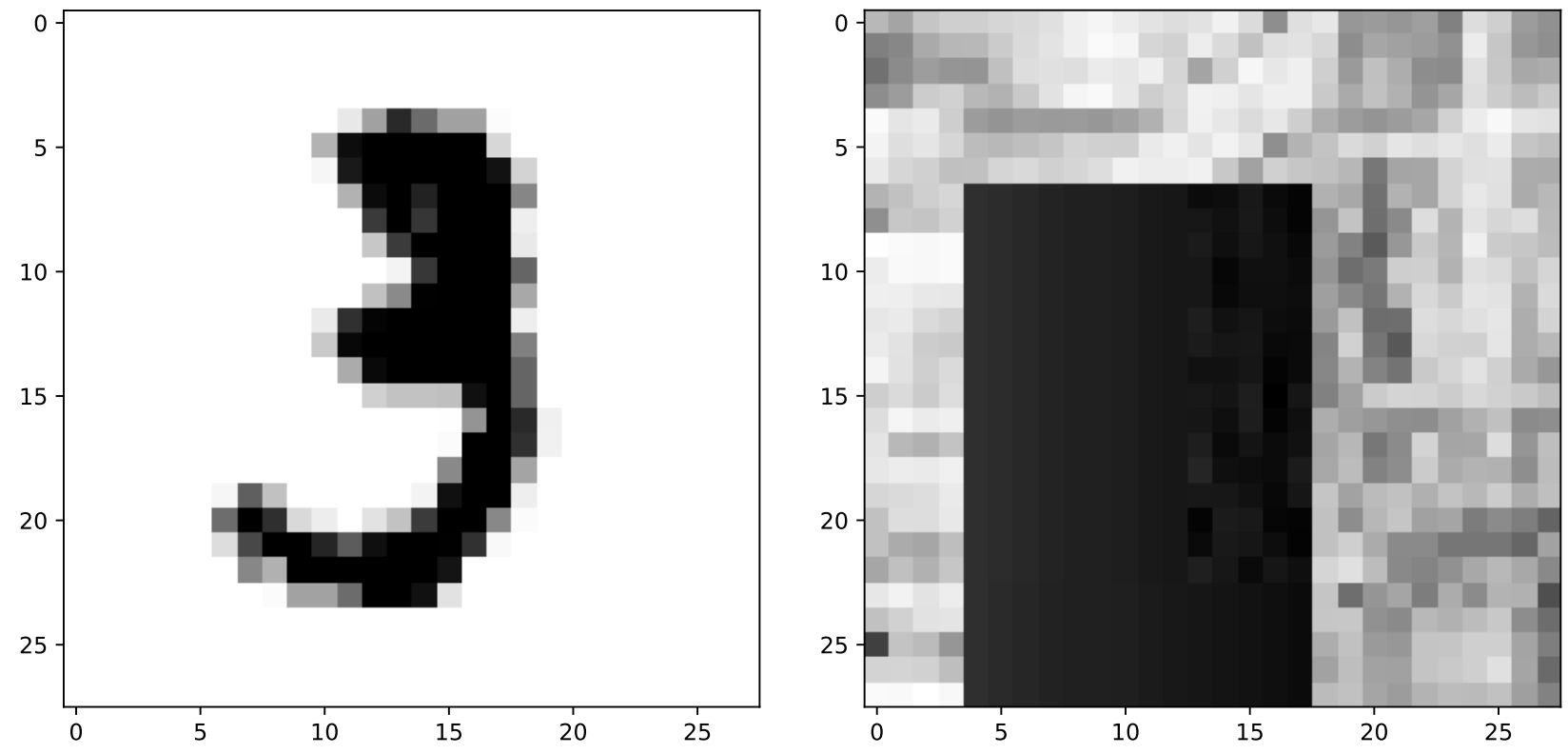

**Figure 3.7:** ***Example inputs from the image–classification benchmarks.*** Left*: a $28 \times 28$ grayscale MNIST digit belonging to class "3".* Right*: a $28 \times 28$ sample from the Rectangles dataset, where the inner dark rectangle is* taller *than it is wide. Pixel intensities are rendered in grayscale for clarity.*

### 3.5.5 Large Scale Classification

Each method is evaluated on two massive binary datasets: SUSY and HIGGS, with 5.5 million and 10 million instances, respectively. These datasets contain Monte Carlo physics simulations to detect the presence of the Higgs boson and super-symmetry (Baldi et al., 2014). The original train/test splits of the data are used, and trained for 500 000 iterations. The AUC metric is reported for comparison with Baldi et al., 2014 and Salimbeni and Deisenroth, 2017. Results are shown in Table 3.3, averaged over 10 different random seed initializations. In the case of DGPs, the best results are shown, which correspond to $L = 4$ and $L = 5$, respectively. DVIP achieves the highest performance on SUSY (AUC of 0.8756) which is comparable to that of DGPs (0.8751) and to the best reported results in Baldi et al., 2014. Namely, shallow NNs (NN, 0.875), deep NN (DNN, 0.876) and boosted decision trees (BDT, 0.863). On HIGGS, despite seeing a steady improvement over VIP by using additional layers, the performance is worse than that of DGP (AUC 0.8324). Again, GPs with an RBF kernel may be a better prior here, and DVIP using inducing points and a GP prior should give similar results to those of DGP. However, the high computational cost of approximately sampling from the GP prior will make this too expensive.

**Table 3.1:** ***Large–scale regression: Year, Airline, and Taxi datasets.*** *Three error metrics—root mean-squared error (RMSE), negative log-likelihood (NLL), and continuous ranked probability score (CRPS)— are reported for single-layer baselines (SGP and VIP), deep variational implicit processes (DVIP) of depth $L = 2$–5, and the best deep GP $L = 3$. Lower values indicate better performance; The best metric is highlighted in* ***purple****; the second-best in* ***teal****.*

| **RMSE** | Single-layer | | Ours | | | | Salimbeni |
|---|---|---|---|---|---|---|---|
| | SGP | VIP | DVIP 2 | DVIP 3 | DVIP 4 | DVIP 5 | Best DGP |
| Year | **9.15 ± 0.01** | 10.27 ± 0.01 | 9.61 ± 0.03 | 9.34 ± 0.02 | 9.30 ± 0.03 | 9.27 ± 0.03 | **8.94 ± 0.03** |
| Airline | 38.61 ± 0.05 | 38.90 ± 0.06 | 37.96 ± 0.03 | 37.91 ± 0.05 | **37.83 ± 0.03** | **37.80 ± 0.05** | 37.95 ± 0.04 |
| Taxi | 554.22 ± 0.32 | 554.60 ± 0.19 | 549.28 ± 0.59 | **531.42 ± 1.59** | 547.33 ± 1.03 | **538.94 ± 2.23** | 552.90 ± 0.33 |

| **NLL** | Single-layer | | Ours | | | | Salimbeni |
|---|---|---|---|---|---|---|---|
| | SGP | VIP | DVIP 2 | DVIP 3 | DVIP 4 | DVIP 5 | Best DGP |
| Year | **3.62 ± 0.00** | 3.74 ± 0.00 | 3.68 ± 0.00 | 3.64 ± 0.00 | 3.64 ± 0.00 | 3.63 ± 0.00 | **3.59 ± 0.00** |
| Airline | 5.10 ± 0.00 | 5.11 ± 0.00 | 5.08 ± 0.00 | **5.07 ± 0.00** | **5.07 ± 0.00** | **5.06 ± 0.00** | **5.07 ± 0.00** |
| Taxi | 7.73 ± 0.00 | 7.73 ± 0.00 | 7.72 ± 0.00 | **7.69 ± 0.00** | 7.72 ± 0.00 | **7.70 ± 0.00** | 7.73 ± 0.00 |

| **CRPS** | Single-layer | | Ours | | | | Salimbeni |
|---|---|---|---|---|---|---|---|
| | SGP | VIP | DVIP 2 | DVIP 3 | DVIP 4 | DVIP 5 | Best DGP |
| Year | **4.83 ± 0.01** | 5.45 ± 0.01 | 5.13 ± 0.02 | 4.96 ± 0.01 | 4.93 ± 0.01 | 4.91 ± 0.02 | **4.68 ± 0.01** |
| Airline | 17.90 ± 0.05 | 17.93 ± 0.04 | 17.53 ± 0.05 | 17.54 ± 0.07 | 17.51 ± 0.05 | **17.45 ± 0.04** | **17.47 ± 0.03** |
| Taxi | 283.79 ± 0.19 | 284.22 ± 0.20 | 282.09 ± 0.32 | **274.65 ± 0.68** | 281.28 ± 0.44 | **277.60 ± 0.90** | 282.99 ± 0.21 |

**Table 3.2:** ***Image–classification benchmarks: MNIST and Rectangles.*** *Test accuracy (in %), average log–likelihood (higher is better, i.e. less negative), and area under the ROC curve (AUC, Rectangles only) are reported for single–layer baselines (SGP and VIP), deep variational implicit processes with $L = 2$–5 layers (DVIP $k$), and deep Gaussian processes (DGP 2–3) from Salimbeni and Deisenroth (2017). Each entry shows the mean ± one standard error. The best metric is highlighted in* ***purple****, the second best in* ***teal****.*

| **MNIST** | Single-layer | | Ours | | Salimbeni | |
|---|---|---|---|---|---|---|
| | SGP | VIP | DVIP 2 | DVIP 3 | DGP 2 | DGP 3 |
| Acc. (%) | 96.25 ± 0.04 | 97.99 ± 0.03 | **98.39 ± 0.05** | **98.36 ± 0.04** | 97.75 ± 0.04 | 97.86 ± 0.05 |
| NLL | 0.146 ± 0.00 | 0.144 ± 0.00 | **0.074 ± 0.00** | 0.080 ± 0.00 | 0.082 ± 0.00 | **0.072 ± 0.00** |

| **Rectangles** | Single-layer | | Ours | | | | Salimbeni |
|---|---|---|---|---|---|---|---|
| | SGP | VIP | DVIP 2 | DVIP 3 | DVIP 4 | DVIP 5 | DGP 3 |
| Acc. (%) | 72.54 ± 0.14 | 85.63 ± 0.18 | 87.84 ± 0.20 | **88.21 ± 0.12** | **87.43 ± 0.20** | 86.49 ± 0.17 | 75.16 ± 0.16 |
| NLL | 0.518 ± 0.00 | 0.348 ± 0.00 | **0.306 ± 0.00** | **0.295 ± 0.00** | 0.309 ± 0.00 | 0.320 ± 0.00 | 0.470 ± 0.00 |
| AUC | 0.828 ± 0.00 | 0.930 ± 0.00 | **0.950 ± 0.00** | **0.953 ± 0.00** | 0.947 ± 0.00 | 0.939 ± 0.00 | 0.858 ± 0.00 |

**Table 3.3:** ***Large–scale binary classification: SUSY and HIGGS.*** *Single–layer baselines (SGP and VIP), DVIP with $L = 2$–$5$ layers (DVIP $k$), and the best deep Gaussian process from Salimbeni and Deisenroth (2017) (DGP 4 for SUSY and DGP 5 for HIGGS). Metrics are test accuracy (in %), negative log–likelihood (NLL; lower is better), and area under the ROC curve (AUC). Entries show the mean ± one standard error. Best is* ***purple****; second best is* ***teal****.*

| **SUSY** | Single-layer | | Ours | | | | Salimbeni |
|---|---|---|---|---|---|---|---|
| | SGP | VIP | DVIP 2 | DVIP 3 | DVIP 4 | DVIP 5 | DGP 4 |
| Acc. (%) | $79.75 \pm 0.02$ | $78.68 \pm 0.02$ | $80.11 \pm 0.03$ | $80.13 \pm 0.01$ | $\mathbf{80.22 \pm 0.01}$ | $\mathbf{80.24 \pm 0.02}$ | $80.06 \pm 0.01$ |
| NLL | $0.436 \pm 0.00$ | $0.456 \pm 0.00$ | $\mathbf{0.429 \pm 0.00}$ | $\mathbf{0.429 \pm 0.00}$ | $\mathbf{0.427 \pm 0.00}$ | $\mathbf{0.427 \pm 0.00}$ | $0.432 \pm 0.00$ |
| AUC | $0.872 \pm 0.00$ | $0.857 \pm 0.00$ | $0.874 \pm 0.00$ | $0.874 \pm 0.00$ | $\mathbf{0.875 \pm 0.00}$ | $\mathbf{0.875 \pm 0.00}$ | $0.875 \pm 0.00$ |
| **SUSY** | Single-layer | | Ours | | | | Salimbeni |
| | SGP | VIP | DVIP 2 | DVIP 3 | DVIP 4 | DVIP 5 | DGP 5 |
| Acc. (%) | $69.95 \pm 0.03$ | $57.42 \pm 0.03$ | $66.09 \pm 0.02$ | $69.85 \pm 0.02$ | $70.43 \pm 0.01$ | $\mathbf{72.01 \pm 0.02}$ | $\mathbf{74.92 \pm 0.01}$ |
| NLL | $0.573 \pm 0.00$ | $0.672 \pm 0.00$ | $0.611 \pm 0.00$ | $0.575 \pm 0.00$ | $0.565 \pm 0.00$ | $\mathbf{0.542 \pm 0.00}$ | $\mathbf{0.501 \pm 0.00}$ |
| AUC | $0.769 \pm 0.00$ | $0.624 \pm 0.00$ | $0.719 \pm 0.00$ | $0.770 \pm 0.00$ | $0.778 \pm 0.00$ | $\mathbf{0.796 \pm 0.00}$ | $\mathbf{0.832 \pm 0.00}$ |

### 3.5.6 Ablation Studies

The results presented in the previous sections highlight the overall performance and calibration properties of Deep Variational Implicit Processes across multiple benchmarks. To better understand the internal behavior of the model, a closer analysis of its components is conducted through a set of targeted ablation experiments. These experiments quantify the individual contribution of each architectural and inference choice to the final predictive performance, shedding light on which design aspects are most critical for stability and accuracy.

#### 3.5.6.1 Impact of the Constrained Prior

Figure 3.8 shows, on a toy problem, the obtained predictive distributions and learned prior samples of VIP, for $\alpha = 0$ (see Section 2.2.4 for further information about $\alpha$-divergences). This corresponds to a 1-layer particular case of DVIP. Two cases are considered: (1) using a full unconstrained prior BNN, in which the mean and variance of each weight and bias can be independently tuned (shown above), and (2) the considered constrained BNN in which prior means and variances are shared across layers (shown below). The second approach considerably reduces the number of parameters in the prior, and it generates smoother prior functions. The predictive distribution is also smoother than when the prior is unconstrained. However, despite providing better results by avoiding overfitting, there might be datasets where using the full unconstrained parameterization of the BNN leads to improved results. For example, in problems where a more flexible prior may be beneficial to obtain good generalization properties on unseen data.

Consider the Boston and the Energy datasets. To highlight that prior over-fitting is a dataset-specific matter, Table 3.4 shows the obtained results using an unconstrained prior and a constrained prior for VIP on the aforementioned datasets. As one may see, significantly over-fitting is taking place on Boston. More precisely, the training error improves a lot when using the unconstrained prior. By contrast, test error and other performance metrics deteriorate on the test set. In Energy, however, a more flexible prior results in better test RMSE and CRPS, but worse test log-likelihood.

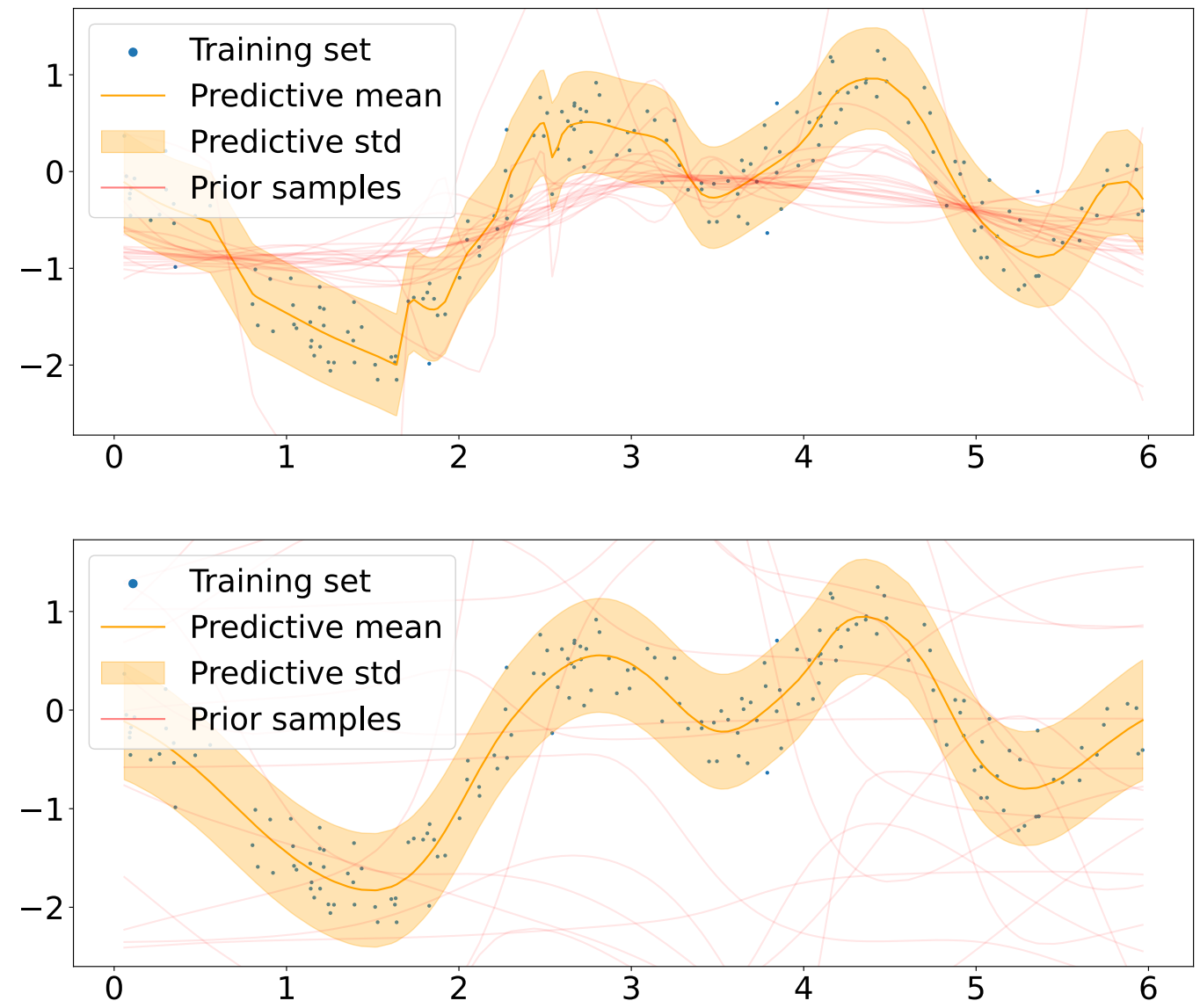

**Figure 3.8:** ***Toy regression: effect of the prior on predictive behavior.*** *Training inputs (blue dots) are fitted with a two–layer BNN.* Top*: model trained with a* full *(unconstrained) weight prior.* Bottom*: model trained with a* constrained *weight prior that encourages smoother functions. In each panel the dashed orange curve is the predictive mean, the shaded band shows $\pm 2$ predictive standard deviations, and faint red curves are draws from the prior. The constrained prior yields a markedly smoother posterior and more moderate uncertainty outside the data range compared with the full prior.*

**Table 3.4:** ***Effect of the BNN prior on VIP performance (Boston and Energy).*** *Unconstrained and constrained Bayesian-NN priors are compared within the VIP framework ($\alpha = 0$). Shown are training and test values for root mean-squared error (RMSE), negative log-likelihood (NLL), and continuous ranked probability score (CRPS). The unconstrained prior fits the training data extremely well but generalizes poorly—most notably on Boston, where the test NLL explodes. Imposing the constraint regularizes the prior, yielding lower test RMSE, NLL, and CRPS on both datasets, at the cost of higher training error. Numbers are means $\pm$ one standard error over 20 random splits.*

| **Unconstrained** | RMSE Train | RMSE Test | NLL Train | NLL Test | CRPS Train | CRPS Test |
|---|---|---|---|---|---|---|
| Boston | $0.05 \pm 0.00$ | $5.85 \pm 0.14$ | $-1.21 \pm 0.19$ | $5126.07 \pm 274.79$ | $0.03 \pm 0.00$ | $4.31 \pm 0.08$ |
| Energy | $0.14 \pm 0.00$ | $0.57 \pm 0.01$ | $-0.51 \pm 0.01$ | $6.52 \pm 0.42$ | $0.079 \pm 0.00$ | $0.36 \pm 0.01$ |
| **Constrained** | RMSE Train | RMSE Test | NLL Train | NLL Test | CRPS Train | CRPS Test |
| Boston | $3.90 \pm 0.02$ | $4.73 \pm 0.24$ | $2.77 \pm 0.00$ | $23.03 \pm 0.07$ | $2.06 \pm 0.01$ | $2.40 \pm 0.08$ |
| Energy | $2.35 \pm 0.03$ | $2.57 \pm 0.08$ | $2.27 \pm 0.01$ | $2.07 \pm 0.02$ | $1.28 \pm 0.01$ | $1.27 \pm 0.04$ |

#### 3.5.6.2 Impact of the Number of Prior Samples

The impact of the number of prior samples $S$ has on the performance and training time of the proposed method is explored. Table 3.5 shows the results obtained using DVIP with 3 layers on Protein and Power datasets (UCI). These results show how increasing the number of prior samples produces better results in terms of performance compared to using lower values of $S$. However, it is important to consider that this value scales quadratically the computational complexity of evaluating the ELBO, heavily influencing the training cost of the model.

**Table 3.5:** ***Impact of the number of prior samples $S$ on three–layer DVIP.*** *For the UCI Power and Protein datasets test root mean–squared error (RMSE), negative log–likelihood (NLL), continuous ranked probability score (CRPS), and total training time (CPU seconds) as the number of prior samples varies from $S = 10$ to $S = 50$ are reported. Increasing $S$ yields modest yet consistent gains in predictive metrics, at the cost of an approximately linear rise in computation. Entries are means $\pm$ one standard error over 20 random splits.*

| **Power** | $S = 10$ | $S = 20$ | $S = 30$ | $S = 40$ | $S = 50$ |
|---|---|---|---|---|---|
| RMSE | $4.03 \pm 0.04$ | $4.01 \pm 0.04$ | $3.94 \pm 0.04$ | $3.92 \pm 0.04$ | $3.94 \pm 0.04$ |
| NLL | $2.81 \pm 0.01$ | $2.81 \pm 0.00$ | $2.79 \pm 0.01$ | $2.78 \pm 0.01$ | $2.79 \pm 0.01$ |
| CRPS | $2.19 \pm 0.01$ | $2.18 \pm 0.01$ | $2.14 \pm 0.01$ | $2.11 \pm 0.01$ | $2.13 \pm 0.01$ |
| CPU Time (s) | $2693 \pm 19$ | $2806 \pm 22$ | $3152 \pm 20$ | $3451 \pm 63$ | $3742 \pm 55$ |
| **Protein** | $S = 10$ | $S = 20$ | $S = 30$ | $S = 40$ | $S = 50$ |
| RMSE | $4.53 \pm 0.01$ | $4.40 \pm 0.01$ | $4.28 \pm 0.01$ | $4.27 \pm 0.01$ | $4.21 \pm 0.01$ |
| NLL | $2.92 \pm 0.00$ | $2.90 \pm 0.00$ | $2.87 \pm 0.00$ | $2.86 \pm 0.00$ | $2.85 \pm 0.00$ |
| CRPS | $2.52 \pm 0.00$ | $2.43 \pm 0.00$ | $2.36 \pm 0.00$ | $2.34 \pm 0.00$ | $2.31 \pm 0.00$ |
| CPU Time (s) | $2334 \pm 28$ | $2734 \pm 19$ | $3616 \pm 11$ | $3727 \pm 35$ | $4330 \pm 52$ |

**Table 3.6:** ***Influence of the prior BNN architecture on DVIP performance.*** *Two priors—a deeper but narrower three–layer network (10–10–10) and a wider two–layer network (20–20)—within DVIPs of depth $L = 2$–$4$ are compared. Test root mean–squared error (RMSE), negative log–likelihood (NLL), and continuous ranked probability score (CRPS) are reported for the UCI Power and Protein datasets. Performance differences across priors are modest: the wider 20–20 prior slightly improves several metrics on Power, while results on Protein remain largely unchanged. Entries are means $\pm$ one standard error over 20 random splits.*

| | BNN 10-10-10 | | | BNN 20-20 | | |
|---|---|---|---|---|---|---|
| **Power** | DVIP 2 | DVIP 3 | DVIP 4 | DVIP 2 | DVIP 3 | DVIP 4 |
| RMSE | $4.02 \pm 0.04$ | $3.94 \pm 0.04$ | $3.97 \pm 0.04$ | $4.02 \pm 0.04$ | $3.99 \pm 0.04$ | $3.93 \pm 0.04$ |
| NLL | $2.81 \pm 0.01$ | $2.79 \pm 0.00$ | $2.80 \pm 0.01$ | $2.81 \pm 0.01$ | $2.80 \pm 0.01$ | $2.79 \pm 0.01$ |
| CRPS | $2.18 \pm 0.01$ | $2.13 \pm 0.01$ | $2.15 \pm 0.01$ | $2.18 \pm 0.01$ | $2.16 \pm 0.01$ | $2.13 \pm 0.01$ |
| **Protein** | DVIP 2 | DVIP 3 | DVIP 4 | DVIP 2 | DVIP 3 | DVIP 4 |
| RMSE | $4.57 \pm 0.01$ | $4.37 \pm 0.01$ | $4.29 \pm 0.01$ | $4.55 \pm 0.01$ | $4.43 \pm 0.01$ | $4.31 \pm 0.01$ |
| NLL | $2.93 \pm 0.00$ | $2.89 \pm 0.00$ | $2.87 \pm 0.00$ | $2.93 \pm 0.00$ | $2.90 \pm 0.00$ | $2.87 \pm 0.00$ |
| CRPS | $2.55 \pm 0.00$ | $2.42 \pm 0.00$ | $2.36 \pm 0.00$ | $2.54 \pm 0.00$ | $2.45 \pm 0.00$ | $2.37 \pm 0.00$ |

#### 3.5.6.3 Robustness over the Prior BNN Architecture

In this section, the impact that changing the prior BNN structure has over the performance of the proposed method is studied. Table 3.6 shows the results obtained using DVIP with 2, 3, and 4 layers on Protein and Power datasets (UCI) with two different BNN structures, 2 hidden layers with 20 units (20-20) and 3 hidden layers with 10 units (10-10-10). These results (that are to be compared with the original ones obtained using 2 hidden layers and 10 units on Table B.1) show that changing the structure of the BNN does not heavily affect the obtained results, given that it is capable of learning similar function distributions with the different BNN architectures.

#### 3.5.6.4 Using a GP as Prior in DVIP and VIP

In this section, the use of a GP prior in DVIP to approximate GPs and hence DGPs is explored. From a theoretical point of view, GP samples could be used in VIPs prior, ensuring that VIP does converge to a GP when the number of prior samples (and linear regression coefficients) is large enough. As a result, DVIP can converge to a DGP. However, DVIP needs to evaluate covariances among the process values at the training points. This requires taking continuous samples from a GP, something that is not possible in practice unless one samples the process values at all the training points, something that is intractable for big datasets. To surpass this limitation, a BNN with a single layer of cosine activation functions can be used to approximate a GP with RBF kernel (Rahimi and Recht, 2007). Generating continuous samples from this BNN is easy. One only has to sample the weights from their corresponding prior distribution. However, in order to correctly approximate the GP, a large number of hidden units are needed in the BNN, increasing the computational cost of taking such samples.

Furthermore, in many situations, the predictive distribution of a sparse GP is very different from that of a full GP. Meaning that even when using an approximate GP prior in VIP, by means of a BNN with enough hidden units, it may not be enough to accurately approximate a sparse GP. Specifically, this is the case in the Year dataset. In this dataset, the difference in performance between DVIP and DGP is not only a consequence of the different prior, but also the posterior approximation. More precisely, DVIP uses a linear regression approximation to the GP, while DGP uses an inducing points based approximation. To show this, an inducing points approximation is proposed on VIP. For this, the required covariance matrices are estimated using a large number of prior samples. The obtained results can be seen in Table 3.7. There, the results of VIP using an approximated GP prior, using both the linear regression approximation and an approximation based on inducing points. Results of the sparse GP and the average training time of each method in seconds are also reported. The table shows that VIP can be used with inducing points to approximate a GP. Specifically, the inducing points approximation of VIP gives very similar results to those of a sparse GP. However, this is achieved at a prohibitive training time. The computational bottlenecks are the GP approximation using a wide single layer BNN (2 000 units in the hidden layer), and the generation of a large number of prior samples from the BNN to approximate the covariance matrix (2 000 samples). Given the high computational cost of training a VIP model on this dataset, considering DVIP with more than 1 layer is too expensive.

Extra experiments in the smaller datasets Protein and Kin8nm are conducted to assess if using a GP prior in the context of DVIP generates results that are closer to those of a DGP. These are the regression datasets from the UCI repository where DGPs performed better than DVIP. In this case, inducing points approximation of the GP is not considered, as in the previous paragraph, but the linear regression approximation of VIP. Results include (i) DVIP using the initially proposed BNN prior with 2 layers of 10 hidden units and tanh activation functions; (ii) DVIP using the wide BNN that approximates the prior GP; (iii) DGPs using sparse GPs in each layer based on inducing points. The results obtained are shown in Tables 3.8 and 3.9. In both cases using a GP prior in DVIP often improves results and performs similarly to a DGP, especially for a large number of layers $L$, even when there are differences in the approximation of the posterior distribution, *i.e.*, DVIP uses a linear regression approximation to the GP and DGP uses inducing points. Again, using a wide BNN to approximate the prior GP results in a significant increment in the training time, making DVIP slower than DGP.

**Table 3.7:** ***VIP versus GP–based variants on the Year dataset.*** *The baseline VIP employs a two–layer BNN prior (width = 10,* `tanh` *activations) and a linear–regression approximation to the GP. VIP-GP (linear regression) replaces the BNN by a wide single–layer cosine network that approximates an RBF GP but keeps the linear–regression posterior. VIP-GP (inducing points) further switches to an inducing–point posterior with* 100 *pseudo–inputs. A sparse GP (SGP) with the same* 100 *inducing points is included for reference. While the inducing–point version of VIP nearly matches the SGP in predictive metrics (RMSE, NLL, CRPS), it incurs a 25× larger training time owing to (i) the very wide cosine network (*2 000 *hidden units) needed to mimic the GP prior and (ii) the* 2 000 *Monte Carlo samples required to approximate its covariance matrix. All VIP variants are trained with* $\alpha = 0$.

| **Year** | VIP | VIP-GP (linear regression) | VIP-GP (inducing points) | SGP |
|---|---|---|---|---|
| RMSE | $10.27 \pm 0.01$ | $10.23 \pm 0.01$ | $9.28 \pm 0.01$ | $9.15 \pm 0.01$ |
| NLL | $3.74 \pm 0.00$ | $3.77 \pm 0.00$ | $3.64 \pm 0.00$ | $3.62 \pm 0.00$ |
| CRPS | $5.45 \pm 0.01$ | $5.45 \pm 0.02$ | $4.85 \pm 0.01$ | $4.83 \pm 0.01$ |
| CPU Time (s) | $1217 \pm 257$ | $1687 \pm 271$ | $30867 \pm 326$ | $1874 \pm 265$ |

**Table 3.8:** ***Protein (UCI) — effect of replacing the BNN prior by an approximate GP prior in DVIP.*** *Top block: results with the original two–layer BNN prior. Bottom block: results after substituting the BNN by a wide single–layer cosine network that approximates an RBF GP ("GP Prior"). For each prior, the reported methods include a vanilla VIP, DVIP models with* $L = 2$–5 *layers, sparse-GP and deep-GP baselines from Salimbeni and Deisenroth (2017). Using the GP prior narrows the gap between DVIP and DGP—especially for deeper models—yet increases training time by roughly a factor of two, underscoring the computational cost of wide networks and large prior sample counts.*

| **BNN Prior** | VIP | DVIP 2 | DVIP 3 | DVIP 4 | DVIP 5 |
|---|---|---|---|---|---|
| RMSE | $4.76 \pm 0.01$ | $4.24 \pm 0.01$ | $4.14 \pm 0.01$ | $4.14 \pm 0.01$ | $\mathbf{4.09 \pm 0.01}$ |
| NLL | $2.98 \pm 0.00$ | $2.86 \pm 0.00$ | $2.84 \pm 0.00$ | $2.84 \pm 0.00$ | $\mathbf{2.83 \pm 0.00}$ |
| CRPS | $2.68 \pm 0.00$ | $2.34 \pm 0.00$ | $2.26 \pm 0.00$ | $2.25 \pm 0.00$ | $\mathbf{2.21 \pm 0.00}$ |
| CPU time (s) | $3086 \pm 173$ | $3981 \pm 182$ | $8604 \pm 774$ | $9931 \pm 616$ | $12568 \pm 327$ |
| **GP Prior** | VIP | DVIP 2 | DVIP 3 | DVIP 4 | DVIP 5 |
| RMSE | $4.89 \pm 0.01$ | $4.26 \pm 0.01$ | $4.07 \pm 0.01$ | $4.02 \pm 0.01$ | $\mathbf{4.01 \pm 0.01}$ |
| NLL | $3.00 \pm 0.00$ | $2.87 \pm 0.00$ | $2.82 \pm 0.00$ | $2.81 \pm 0.00$ | $\mathbf{2.81 \pm 0.00}$ |
| CRPS | $2.77 \pm 0.00$ | $2.35 \pm 0.00$ | $2.21 \pm 0.00$ | $2.17 \pm 0.00$ | $\mathbf{2.16 \pm 0.00}$ |
| CPU time (s) | $5880 \pm 249$ | $12293 \pm 564$ | $20351 \pm 1110$ | $18514 \pm 575$ | $25835 \pm 1259$ |
| Salimbeni | SGP | DGP 2 | DGP 3 | DGP 4 | DGP 5 |
| RMSE | $4.56 \pm 0.01$ | $4.17 \pm 0.01$ | $\mathbf{4.00 \pm 0.01}$ | $4.01 \pm 0.01$ | $4.02 \pm 0.01$ |
| NLL | $2.93 \pm 0.00$ | $2.84 \pm 0.00$ | $\mathbf{2.79 \pm 0.00}$ | $\mathbf{2.79 \pm 0.00}$ | $2.80 \pm 0.00$ |
| CRPS | $2.56 \pm 0.00$ | $2.31 \pm 0.00$ | $\mathbf{2.19 \pm 0.00}$ | $\mathbf{2.19 \pm 0.00}$ | $2.20 \pm 0.00$ |
| CPU time (s) | $2690 \pm 114$ | $10031 \pm 129$ | $17528 \pm 689$ | $16128 \pm 190$ | $20653 \pm 969$ |

As a conclusion, DVIPs' general and flexible definition allows the usage of (approximate) GP priors and inducing points. This enables performing (nearly) equally to a sparse GP or a DGP. However, the computational overhead of doing these approximations is prohibitive. Thus, if a GP prior is to be considered, it is a much better approach to just use the DGP framework, and leave DVIP to cases where flexible but easy-to-sample-from priors can be used.

**Table 3.9:** ***Kin8nm (UCI) — DVIP with BNN versus GP priors.*** *Replacing the BNN prior by a wide cosine network that approximates an RBF GP (GP Prior) systematically improves the predictive metrics (RMSE, NLL, CRPS) of DVIP and often matches or surpasses the performance of deep Gaussian processes with inducing points. The gain, however, comes at the expense of significantly longer training times—up to* $\sim 2.3\times$ *slower for* $L = 5$ *layers—because thousands of hidden units and prior samples are required to emulate the GP covariance structure. All VIP models are trained with* $\alpha = 0$.

| **BNN Prior** | VIP | DVIP 2 | DVIP 3 | DVIP 4 | DVIP 5 |
|---|---|---|---|---|---|
| RMSE | $0.15 \pm 0.00$ | $\mathbf{0.07 \pm 0.00}$ | $\mathbf{0.07 \pm 0.00}$ | $\mathbf{0.07 \pm 0.00}$ | $\mathbf{0.07 \pm 0.00}$ |
| NLL | $-0.47 \pm 0.00$ | $-1.13 \pm 0.00$ | $\mathbf{-1.18 \pm 0.00}$ | $-1.16 \pm 0.00$ | $-1.17 \pm 0.00$ |
| CRPS | $0.08 \pm 0.00$ | $\mathbf{0.04 \pm 0.00}$ | $\mathbf{0.04 \pm 0.00}$ | $\mathbf{0.04 \pm 0.00}$ | $\mathbf{0.04 \pm 0.00}$ |
| CPU time (s) | $2109 \pm 57$ | $5086 \pm 232$ | $6927 \pm 27$ | $9459 \pm 77$ | $11763 \pm 141$ |
| **GP Prior** | VIP | DVIP 2 | DVIP 3 | DVIP 4 | DVIP 5 |
| RMSE | $0.14 \pm 0.00$ | $0.07 \pm 0.00$ | $\mathbf{0.06 \pm 0.00}$ | $\mathbf{0.06 \pm 0.00}$ | $\mathbf{0.06 \pm 0.00}$ |
| NLL | $-0.43 \pm 0.00$ | $-1.20 \pm 0.00$ | $-1.25 \pm 0.00$ | $\mathbf{-1.26 \pm 0.00}$ | $-1.25 \pm 0.00$ |
| CRPS | $0.08 \pm 0.00$ | $0.04 \pm 0.00$ | $\mathbf{0.03 \pm 0.00}$ | $\mathbf{0.03 \pm 0.00}$ | $\mathbf{0.03 \pm 0.00}$ |
| CPU time (s) | $6672 \pm 442$ | $14300 \pm 847$ | $17573 \pm 1033$ | $22561 \pm 980$ | $22669 \pm 657$ |
| Salimbeni | SGP | DGP 2 | DGP 3 | DGP 4 | DGP 5 |
| RMSE | $0.09 \pm 0.00$ | $\mathbf{0.06 \pm 0.00}$ | $\mathbf{0.06 \pm 0.00}$ | $\mathbf{0.06 \pm 0.00}$ | $\mathbf{0.06 \pm 0.00}$ |
| NLL | $-0.91 \pm 0.00$ | $-1.29 \pm 0.00$ | $-1.32 \pm 0.00$ | $\mathbf{-1.33 \pm 0.00}$ | $-1.30 \pm 0.00$ |
| CRPS | $0.05 \pm 0.00$ | $\mathbf{0.03 \pm 0.00}$ | $\mathbf{0.03 \pm 0.00}$ | $\mathbf{0.03 \pm 0.00}$ | $\mathbf{0.03 \pm 0.00}$ |
| CPU time (s) | $2053 \pm 81$ | $6375 \pm 331$ | $11147 \pm 472$ | $17502 \pm 1060$ | $21846 \pm 1246$ |

## 3.6 Conclusions

Deep Variational Implicit Processes are defined as a model based on the concatenation of implicit processes (IPs) as the prior for the latent function. The conducted experiments demonstrate that DVIPs can be used on a variety of regression and classification problems with no need of hand-tuning. Results show that DVIPs surpass or match the performance of single layer VIPs and GPs. It also gives similar and sometimes better results than those of deep GPs (DGPs). However, the main comparative advantage of DVIPs is that they have less computational cost than DGPs. Experiments have also demonstrated that DVIPs are both effective and scalable on a wide range of tasks. DVIPs do not seem to over-fit on small datasets by increasing the depth, and on large datasets, extra layers often improve performance. It is also shown that increasing the number of layers is far more effective than increasing the complexity of the prior of single-layer VIPs, which heavily affects the training time of the model. Besides the added computation time, which is rather minor, no drawbacks to the use of DVIPs instead of single-layer VIPs are seen, but rather significant benefits.

The use of domain specific priors, e.g. considering CNNs in the first layer, has demonstrated to give outstanding results in image-based datasets compared to other GP methods. This establishes a new use of IPs with not-so-general prior functions. The employment of these priors on other domain specific tasks, such as forecasting or data encoding, is foreseen as an emerging field of study. The prior flexibility also results in a generalization of DGPs. As a matter of fact, DVIPs should give similar results to those of DGPs if a GP is considered as the IP prior for each unit.

Despite the good results, DVIPs present some limitations: first of all, the implicit prior works as a black box from the interpretability point of view. The prior parameters do not represent a clear property of the model in comparison to the original GP's kernel parameters, such as the length and amplitude. Furthermore, even though using 20 samples from the prior has shown to give remarkable results in most cases, there might be situations where this number must be increased, having a big impact on the model's training time.

# 4

# Post-hoc Uncertainty Estimation for Bayesian Deep Learning

The methods developed in the previous chapter introduced scalable Bayesian inference directly in function space through Deep Variational Implicit Processes. However, in many real-world scenarios, a high-performing deterministic network is already available, and retraining it within a full Bayesian framework is impractical. This chapter therefore shifts focus from end-to-end Bayesian training to post-hoc uncertainty estimation—attaching calibrated predictive distributions to existing neural networks without modifying their original parameters. Two complementary approaches are proposed: the Variational Linearized Laplace Approximation (VaLLA) and the Fixed-Mean Gaussian Process (FMGP), both of which provide principled and computationally efficient routes to retrofit uncertainty into modern deep learning models.

## 4.1 Introduction

Deep neural networks (DNNs) have become the *de facto* solution for a broad range of pattern–recognition tasks, consistently achieving state-of-the-art accuracy in areas such as computer vision and natural language processing (He et al., 2016a; Vaswani et al., 2017). However, conventional (deterministic) DNNs deliver only point predictions; they are frequently miscalibrated (Guo et al., 2017) and provide poor representations of epistemic uncertainty (Blundell et al., 2015). These shortcomings are unacceptable in risk-sensitive applications—for example, autonomous driving (Kendall and Gal, 2017) and medical decision support (Leibig et al., 2017)—where a faithful quantification of predictive confidence is essential.

A principled remedy is to cast the network weights as random variables and perform Bayesian inference, yielding a *Bayesian neural network* (BNN) (Graves, 2011; MacKay, 1992c; Neal, 2012). Because exact posterior inference is intractable in modern architectures, one resorts to approximations such as variational inference (VI) (Blundell et al., 2015), Markov chain Monte Carlo (MCMC) (T. Chen et al., 2014), or the Laplace approximation (LA) (MacKay, 1992a; Ritter et al., 2018). Empirically, VI and MCMC often underperform plain back-propagation (Wenzel et al., 2020a), whereas LA strikes a favorable balance between accuracy and cost: starting from the maximum-a-posteriori (MAP) solution obtained by standard training, it fits a Gaussian posterior with mean at the MAP estimate and covariance given by the (negative inverse) Hessian of the log-posterior.

Forming and inverting the full Hessian is infeasible in large networks, so practitioners substitute the Generalised Gauss–Newton (GGN) matrix (Martens and Grosse, 2015). Although this approximation may under-fit (Lawrence, 2001), the effect is alleviated when the model is linearized at the MAP parameters. This insight underlies the *Linearized Laplace Approximation* (LLA) (Foong et al., 2019; Immer et al., 2021), which applies LA to the first-order Taylor expansion of the network. LLA preserves the original predictive mean—thus retaining baseline performance—while attaching principled error bars, and its GGN is guaranteed positive-definite, allowing post-hoc application at arbitrary weight configurations.

Despite these advantages, LLA is not yet turnkey for industrial-scale problems: computing the Jacobian of the network output with respect to millions of parameters for every training instance is prohibitively expensive. Consequently, further structural approximations—e.g. diagonal or Kronecker-factored (KFAC) representations of the GGN—or data sub-sampling are employed, sacrificing some statistical fidelity for tractability. Designing methods that match the calibration quality of Bayesian approaches while scaling gracefully to contemporary architectures, therefore, remains an active and important research challenge.

**Contributions of this chapter.** To overcome the scalability bottleneck of LLA while preserving its favorable calibration properties, two recent lines of work are examined:

(i) **Variational LLA (VaLLA).** LLA's predictive distribution is reinterpreted as that of a Gaussian process (GP) (Khan et al., 2019). A sparse variational GP with inducing points (Titsias, 2009) is then employed *only* for the covariance, leaving the mean *exactly equal* to the original DNN output by working in the dual, RKHS representation (see Section 2.3.6) VaLLA supports stochastic mini-batch optimization and attains sub-linear cost in the data size, outperforming alternative scalable-LA surrogates such as Kronecker/diagonal GGN factorizations, Nyström-based ELLA (Z. Deng et al., 2022), and sample-then-optimize schemes (Antorán et al., 2023).

(ii) **Fixed-Mean Gaussian Processes (FMGPs).** Extending the RKHS perspective (see Section 2.3.6) further, a new family of sparse GPs is introduced, whose posterior *mean* is fixed to an arbitrary continuous function—e.g. the predictions of a pre-trained high-performance DNN—while the posterior covariance is learned via VI using decoupled inducing points (Cheng and Boots, 2016). FMGP thus converts *any* pre-trained network into a Bayesian predictor without requiring Jacobians or direct access to internal parameters. Scalability to ImageNet-scale vision models and to molecular-property prediction (QM9) is shown, achieving uncertainty estimates competitive with Hamiltonian Monte Carlo on small problems but at orders-of-magnitude lower cost.

Together, these two methodologies illustrate how LLA's theoretical appeal can be retained in large-scale, practical settings through judicious kernel approximations and RKHS duality, paving the way for reliable uncertainty quantification in modern deep learning.

## 4.2 Variational Linearized Laplace Approximation

Building on the Gaussian-process (GP) foundations laid out in Section 2.3, *Variational Linearized Laplace Approximation* (VaLLA) is introduced. The method inherits (i) the scalability of variational sparse GPs and (ii) the post-hoc nature of linearized Laplace approximation (LLA), allowing for enhanced uncertainty estimation of a pre-trained deep model employing a sparse approximation to LLA.

As reviewed in Section 2.3.4.2, Variational sparse GPs approximate the GP posterior using a GP parameterized by $M$ inducing points $\mathbf{Z}$, each in $\mathbb{R}^D$, and associated process values $\mathbf{u} = f(\mathbf{Z})$ (Titsias, 2009),

$$P(\mathbf{f}, \mathbf{u}|\mathbf{y}) \approx Q(\mathbf{f}, \mathbf{u}) = P(\mathbf{f}|\mathbf{u})Q(\mathbf{u})\,, \tag{4.1}$$

where $Q(\mathbf{u}) = \mathcal{N}(\mathbf{u}|\hat{\boldsymbol{m}}, \hat{\boldsymbol{S}})$, $\mathbf{f} = f(\mathbf{X})$, and $P(\mathbf{f}|\mathbf{u})$ are fixed. The approximate posterior distribution $Q(\mathbf{u})$ is obtained by minimizing the KL-divergence $\mathrm{KL}\left(Q(\mathbf{f}, \mathbf{u})|P(\mathbf{f}, \mathbf{u}|\mathbf{y})\right)$. In practice, the minimization problem is transformed into the maximization of the lower bound of the log-marginal likelihood.

$$\log P(\mathbf{y}) \geq \max_{\mathbf{Z}, \hat{\boldsymbol{m}}, \hat{\boldsymbol{S}}} \int Q(\mathbf{f}, \mathbf{u}) \log \frac{P(\mathbf{y}|\mathbf{f})P(\mathbf{f}|\mathbf{u})P(\mathbf{u})}{Q(\mathbf{f}, \mathbf{u})}\, \mathrm{d}(\mathbf{f}, \mathbf{u})\,, \tag{4.2}$$

which has cost $\mathcal{O}(NM^2{+}M^3)$ due to the cancellation of the factor $P(\mathbf{f}|\mathbf{u})$ since $Q(\mathbf{f}, \mathbf{u}) = P(\mathbf{f}|\mathbf{u})Q(\mathbf{u})$. The following result establishes a connection between SVGPs and a subset of Gaussian measures in the Hilbert space. This is built on the RKHS theory presented in Section 2.3.6.

**Theorem 4.1.** *For any kernel function $K$ and its corresponding RKHS $\mathcal{H}$ —with feature maps defined as $\phi_{\mathbf{x}} := K(\cdot, \mathbf{x}) \in \mathcal{H}$—, any $\boldsymbol{a} \in \mathbb{R}^M$, $\boldsymbol{A} \in \mathbb{R}^{M\times M}$ with $\boldsymbol{A} \succeq 0$ and $\mathbf{Z} = (\mathbf{z}_1, \ldots, \mathbf{z}_M) \in \mathcal{X}^M$, let $\tilde{\mu} \in \mathcal{H}$ and $\tilde{\Sigma} : \mathcal{H} \to \mathcal{H}$ be defined as*

$$\tilde{\mu} = \sum_{m=1}^{M} a_m \phi_{\mathbf{z}_m}, \qquad \text{and} \qquad \tilde{\Sigma} = \left( I + \sum_{i=1}^{M}\sum_{j=1}^{M} \phi_{\mathbf{z}_i} A_{i,j} \phi_{\mathbf{z}_j}^T \right)^{-1}, \tag{4.3}$$

*then, $\tilde{\Sigma} \in L^+(\mathcal{H})$, and the Gaussian measure defined by $(\tilde{\mu}, \tilde{\Sigma})$ is equivalent to a SVGP with*

*variational distribution* $Q(\mathbf{u}) = \mathcal{N}(\mathbf{u}|\boldsymbol{m}, \boldsymbol{S})$, *defined as*

$$\boldsymbol{m} = K(\mathbf{Z}, \mathbf{Z})\boldsymbol{a}, \quad \textit{and} \quad \boldsymbol{S} = K(\mathbf{Z}, \mathbf{Z}) - K(\mathbf{Z}, \mathbf{Z})(\boldsymbol{A}^{-1} + K(\mathbf{Z}, \mathbf{Z}))^{-1}K(\mathbf{Z}, \mathbf{Z}) \,. \tag{4.4}$$

[Proof in Appendix C.3]

This defines a family of Gaussian measures $\mathcal{Q} \subset \mathcal{P}_{\mathcal{H}}$ as

$$\mathcal{Q} := \left\{\mathcal{N}(f|\tilde{\mu}_{\boldsymbol{a}}, \tilde{\Sigma}_{\boldsymbol{A}}) : \boldsymbol{a} \in \mathbb{R}^M, \boldsymbol{A} \in \mathbb{R}^{M\times M}, \boldsymbol{A} \succeq 0, \mathbf{Z} \in \mathcal{X}^M\right\} , \tag{4.5}$$

where $\mathcal{H}$ and $M$ are omitted from $\mathcal{Q}'s$ notation for simplicity as more sub-indexes will be used later.

### 4.2.1 Decoupled Sparse Variational Gaussian Processes

Theorem 4.1 establishes that the variational inference procedure introduced by Titsias (2009) conducts optimization over a Gaussian measure whose mean, $\tilde{\mu}$, and covariance operator, $\tilde{\Sigma}$, are *both* expressed in terms of a single finite set of inducing basis functions $\{\phi_{\mathbf{z}} \in \mathcal{H} \,:\, \mathbf{z} \in \mathbf{Z}\} \subset \mathcal{H}$, where $\mathcal{H}$ denotes the RKHS generated by the kernel.

Cheng and Boots (2017) extend this construction by allowing *different* inducing bases for the mean and the covariance. Let $M_\alpha, M_\beta \in \mathbb{N}$, and

$$\mathbf{Z}_\alpha = \{\mathbf{z}_{\alpha,1}, \dots, \mathbf{z}_{\alpha,M_\alpha}\} \in \mathcal{X}^{M_\alpha}, \qquad \mathbf{Z}_\beta = \{\mathbf{z}_{\beta,1}, \dots, \mathbf{z}_{\beta,M_\beta}\} \in \mathcal{X}^{M_\beta} \,. \tag{4.6}$$

Define the Gaussian measure as

$$\tilde{\mu} = \sum_{m=1}^{M_\alpha} a_m \phi_{\mathbf{z}_{\alpha,m}}, \qquad \text{and} \qquad \tilde{\Sigma} = \left(I + \sum_{i=1}^{M_\beta}\sum_{j=1}^{M_\beta} \phi_{\mathbf{z}_{\beta,i}} A_{i,j} \phi_{\mathbf{z}_{j,\beta}}^T\right)^{-1} , \tag{4.7}$$

where $\boldsymbol{A} \succeq 0$. This decoupled parameterization is a clear generalization from standard SVGPs and cannot be obtained using the approach of Titsias, 2009 unless $\mathbf{Z}_\alpha = \mathbf{Z}_\beta$. The decoupled space of Gaussian measures is defined then as:

$$\mathcal{Q}^+ = \left\{\mathcal{N}(f|\tilde{\mu}_{\alpha,\boldsymbol{a}}, \tilde{\Sigma}_{\beta,\boldsymbol{A}}) : \boldsymbol{a} \in \mathbb{R}^{M_\alpha}, \boldsymbol{A} \in \mathbb{R}^{M_\beta\times M_\beta}, \boldsymbol{A} \succeq 0, \mathbf{Z}_\alpha \in \mathcal{X}^{M_\alpha}, \mathbf{Z}_\beta \in \mathcal{X}^{M_\beta}\right\} , \tag{4.8}$$

where it is verified that $\mathcal{Q} \subset \mathcal{Q}^+ \subset \mathcal{P}_{\mathcal{H}}$.

This two-basis parameterization *strictly* generalizes the single-basis formulation of Titsias (2009); there exists no shared inducing set capable of reproducing every pair $(\mathbf{Z}_\alpha, \mathbf{Z}_\beta)$ when $M_\alpha \neq M_\beta$.

For any $Q(f) = \mathcal{N}(f|\tilde{\mu}, \tilde{\Sigma})$ drawn from the decoupled Gaussian measure family, the evidence lower bound (ELBO) is (Cheng and Boots, 2016)

$$\mathcal{L}(Q) = \mathbb{E}_Q\left[\log P(\mathbf{y}|f)\right] - \mathrm{KL}(Q|P) \,, \tag{4.9}$$

where $p(f)$ denotes the prior Gaussian measure (typically $\mathcal{N}(f|0, I)$) and $P(\mathbf{y}|f)$ the likelihood. Because $Q$ and $P$ are Gaussian, the Kullback–Leibler divergence admits a closed-form expression (Cheng and Boots, 2016):

$$\mathrm{KL}(Q|P) = \frac{1}{2}\boldsymbol{a}^T\boldsymbol{K}_\alpha\boldsymbol{a} + \frac{1}{2}\log\det(I + \boldsymbol{K}_\beta\boldsymbol{A}) - \frac{1}{2}\mathrm{tr}[\boldsymbol{K}_\beta(\boldsymbol{A}^{-1} + \boldsymbol{K}_\beta)^{-1}], \tag{4.10}$$

with $(\boldsymbol{K}_\alpha)_{i,j} = K(\mathbf{z}_{\alpha,i}, \mathbf{z}_{\alpha,j})$ and $(\boldsymbol{K}_\beta)_{i,j} = K(\mathbf{z}_{\beta,i}, \mathbf{z}_{\beta,j})$. The expectation term in Equation (4.9) depends on the choice of likelihood (e.g. Gaussian, Bernoulli) and is typically handled via Gaussian quadrature or the reparameterization trick. Gradients of $\mathcal{L}(q)$ with respect to $\boldsymbol{a}$, $\boldsymbol{A}$, and the inducing locations can thereafter be obtained in closed form or by automatic differentiation.

The decoupling of inducing sets introduces two notable advantages:

1. **Expressiveness.** Allowing $M_\alpha \neq M_\beta$ enables the mean to be represented with a smaller (or larger) set of points than the covariance, yielding a more parsimonious yet accurate approximation.
2. **Computational trade-off.** Evaluating $\tilde{\mu}$ scales with $M_\alpha$, whereas computing the log-determinant in Equation (4.10) scales as $O(M_\beta^3)$. Choosing $M_\beta < M_\alpha$ can therefore reduce the dominant cubic cost while retaining a rich mean function.

Consequently, the framework of Cheng and Boots (2017) contains the sparse-GP variational method of Titsias (2009) as the special case $\mathbf{Z}_\alpha = \mathbf{Z}_\beta$ and $M_\alpha = M_\beta$, yet affords superior flexibility for large-scale or heteroscedastic data sets encountered in contemporary applications.

### 4.2.2 Linearized Laplace Approximation (LLA)

Consider the task of inferring an unknown target function $f : \mathbb{R}^D \to \mathbb{R}$ from noisy observations $\mathbf{y} = (y_1, \ldots, y_N)^\top$ at inputs $\mathbf{X} = (\mathbf{x}_1, \ldots, \mathbf{x}_N)$. In deep learning, a neural network

$$g : \mathbb{R}^D \times \mathbb{R}^P \to \mathbb{R},$$

with parameters $\boldsymbol{\theta} \in \mathbb{R}^P$, is used to approximate $f$ such that there exists a parameter configuration $\hat{\boldsymbol{\theta}}$ satisfying $f(\cdot) \approx g(\cdot, \hat{\boldsymbol{\theta}})$. After training, the function $g(\cdot, \hat{\boldsymbol{\theta}})$ acts as a deterministic predictor of the data-generating process. However, standard neural networks do not quantify predictive uncertainty, often yielding overconfident predictions in regions without training data.

From a Bayesian standpoint, uncertainty over the model parameters is introduced by specifying a prior $P(\boldsymbol{\theta})$ and forming the posterior

$$P(\boldsymbol{\theta}|\mathbf{y}) \propto P(\mathbf{y}|\boldsymbol{\theta})\, P(\boldsymbol{\theta}),$$

where the likelihood is induced by the network evaluations $\mathbf{g} = (g(\mathbf{x}_1, \boldsymbol{\theta}), \ldots, g(\mathbf{x}_N, \boldsymbol{\theta}))^\top$. In regression, the likelihood is typically Gaussian, while in classification problems $y_i \in \{1, \ldots, C\}$ with $C$ denoting the number of classes, and the likelihood is categorical with probabilities given by a softmax link function. In this latter case,

$$g : \mathbb{R}^D \times \mathbb{R}^P \to \mathbb{R}^C,$$

produces $C$ logits, one per class label.

Exact posterior inference is impossible for modern deep networks because the likelihood is highly non-linear, so one typically introduces a tractable *approximate* posterior $Q(\boldsymbol{\theta}) \approx P(\boldsymbol{\theta}|\mathbf{y})$. Predictions for a new input $\mathbf{x}^\star$ are then obtained by Monte Carlo marginalization,

$$P(y^\star|\mathbf{x}^\star, \mathbf{y}) = \mathbb{E}_{P(\boldsymbol{\theta}|\mathbf{y})}\left[P(y^\star|\mathbf{x}^\star, \boldsymbol{\theta})\right] \approx \frac{1}{S}\sum_{s=1}^{S} P(y^\star|\mathbf{x}^\star, \boldsymbol{\theta}_s), \qquad \boldsymbol{\theta}_s \sim Q(\boldsymbol{\theta}), \tag{4.11}$$

where the empirical average exposes both *aleatoric* and *epistemic* uncertainty (Bishop, 2006).

The Laplace approximation (LA) replaces the true posterior by a Gaussian centered at the maximum–a–posteriori (MAP) solution $\hat{\boldsymbol{\theta}} = \arg\max_{\boldsymbol{\theta}}\left[\log P(\mathbf{y}|\boldsymbol{\theta}) + \log P(\boldsymbol{\theta})\right]$:

$$Q(\boldsymbol{\theta}) = \mathcal{N}(\boldsymbol{\theta}|\hat{\boldsymbol{\theta}}, \boldsymbol{\Sigma}), \qquad \boldsymbol{\Sigma}^{-1} = -\nabla^2_{\boldsymbol{\theta}\boldsymbol{\theta}}\Big[\log P(\mathbf{y}|\boldsymbol{\theta}) + \log P(\boldsymbol{\theta})\Big]_{\boldsymbol{\theta}=\hat{\boldsymbol{\theta}}}. \tag{4.12}$$

Assuming an isotropic Gaussian prior $P(\boldsymbol{\theta}) = \mathcal{N}(\boldsymbol{\theta}|\mathbf{0}, \sigma_0^2\boldsymbol{I})$ adds the identity term $\boldsymbol{I}/\sigma_0^2$ to the precision,

$$\boldsymbol{\Sigma}^{-1} = -\nabla^2_{\boldsymbol{\theta}\boldsymbol{\theta}} \log P(\mathbf{y}|\boldsymbol{\theta})_{\boldsymbol{\theta}=\hat{\boldsymbol{\theta}}} + \frac{\boldsymbol{I}}{\sigma_0^2}. \tag{4.13}$$

For deep networks, the Hessian is intractable and not guaranteed positive-definite. A common remedy is to replace it by the positive semi-definite *Generalized Gauss–Newton* (GGN) matrix (Immer et al., 2021),

$$\boldsymbol{\Sigma}^{-1} \approx \sum_{n=1}^{N} J_{\hat{\boldsymbol{\theta}}}(\mathbf{x}_n)^\top \Lambda(\mathbf{x}_n, y_n) J_{\hat{\boldsymbol{\theta}}}(\mathbf{x}_n) + \frac{\boldsymbol{I}}{\sigma_0^2}, \tag{4.14}$$

with $J_{\hat{\boldsymbol{\theta}}}(\mathbf{x}) := \nabla_{\boldsymbol{\theta}}\, g(\mathbf{x}, \boldsymbol{\theta})\big|_{\boldsymbol{\theta}=\hat{\boldsymbol{\theta}}}$ the Jacobian of the network outputs and

$$\Lambda(\mathbf{x}_n, y_n) := -\nabla^2_{\mathbf{g}\mathbf{g}} \log P(y_n|\mathbf{g})\big|_{\mathbf{g}=g(\mathbf{x}_n, \hat{\boldsymbol{\theta}})} \tag{4.15}$$

the Hessian of the negative log-likelihood with respect to the network outputs $\mathbf{g}$. Because the GGN is exactly the Hessian of the *linearized* network

$$g^{\text{lin}}_{\hat{\boldsymbol{\theta}}}(\mathbf{x}, \boldsymbol{\theta}) = g(\mathbf{x}, \hat{\boldsymbol{\theta}}) + J_{\hat{\boldsymbol{\theta}}}(\mathbf{x})\,(\boldsymbol{\theta} - \hat{\boldsymbol{\theta}}), \tag{4.16}$$

the approximate posterior is centered on $\hat{\boldsymbol{\theta}}$ but its curvature is governed by the linear surrogate, a mismatch that can lead to underfitting (Lawrence, 2001).

The *linearized* Laplace approximation (LLA) resolves the mismatch by also making predictions through the linear surrogate:

$$P_{\text{LLA}}(y^\star|\mathbf{x}^\star,\mathbf{y}) = \mathbb{E}_{Q(\boldsymbol{\theta})}[P(y^\star|g^{\text{lin}}_{\hat{\boldsymbol{\theta}}}(\mathbf{x}^\star,\boldsymbol{\theta}))]\,. \tag{4.17}$$

LLA therefore uses a Gaussian posterior over a *linear* model, for which the GGN is exact. The main computational bottleneck is now the inversion of the precision matrix, whose cost scales cubically with the number of parameters. A dual Gaussian-process formulation trades this for a cubic dependence on the number of training points instead.

A linear model endowed with a Gaussian prior on its weights induces a Gaussian process (GP) over functions (Rasmussen and Williams, 2006). Adopting the isotropic prior $p(\boldsymbol{\theta}) = \mathcal{N}(\boldsymbol{\theta}|\mathbf{0},\sigma_0^2\mathbf{I}_P)$, the linearized Bayesian neural network $g^{\text{lin}}_{\hat{\boldsymbol{\theta}}}(\mathbf{x},\boldsymbol{\theta})$ defines a GP whose mean and covariance functions are:

$$m(\mathbf{x}) = g^{\text{lin}}_{\hat{\boldsymbol{\theta}}}(\mathbf{x},\mathbf{0}), \qquad \text{and} \qquad K_{\text{prior}}(\mathbf{x},\mathbf{x}') = \sigma_0^2\, J_{\hat{\boldsymbol{\theta}}}(\mathbf{x})^\top J_{\hat{\boldsymbol{\theta}}}(\mathbf{x}'). \tag{4.18}$$

Replacing the prior $p(\boldsymbol{\theta})$ by the Laplace posterior $q(\boldsymbol{\theta}) = \mathcal{N}(\boldsymbol{\theta}|\hat{\boldsymbol{\theta}},\boldsymbol{\Sigma})$ simply shifts the GP mean to the non-linear network output while re-scaling the covariance:

$$m(\mathbf{x}) = g(\mathbf{x},\hat{\boldsymbol{\theta}}), \qquad K(\mathbf{x},\mathbf{x}') = J_{\hat{\boldsymbol{\theta}}}(\mathbf{x})^\top\,\boldsymbol{\Sigma}\,J_{\hat{\boldsymbol{\theta}}}(\mathbf{x}'). \tag{4.19}$$

Under the Generalized Gauss–Newton (GGN) approximation $\boldsymbol{\Sigma}^{-1} = \sum_n J_n^\top \Lambda_n J_n + \sigma_0^{-2}\mathbf{I}_P$ the Woodbury matrix identity yields

$$\boldsymbol{\Sigma} = \sigma_0^2\Big(\mathbf{I}_P - \mathbf{J}^\top\big(\sigma_0^{-2}\boldsymbol{\Lambda}^{-1}_{\mathbf{X},\mathbf{y}} + \mathbf{J}\mathbf{J}^\top\big)^{-1}\mathbf{J}\Big), \tag{4.20}$$

where $\mathbf{J}$ stacks all Jacobians $J_{\hat{\boldsymbol{\theta}}}(\mathbf{x}_n)$ row-wise and $\boldsymbol{\Lambda}_{\mathbf{X},\mathbf{y}} = \operatorname{diag}(\Lambda_1,\ldots,\Lambda_N)$. Substituting this expression into the covariance formula above leads to a purely *function-space* representation. The (scaled) Neural Tangent Kernel (NTK)

$$\kappa(\mathbf{x},\mathbf{x}') = \sigma_0^2\,J_{\hat{\boldsymbol{\theta}}}(\mathbf{x})^\top J_{\hat{\boldsymbol{\theta}}}(\mathbf{x}'), \qquad \mathbf{Q} = \boldsymbol{\Lambda}^{-1}_{\mathbf{X},\mathbf{y}} + \kappa(\mathbf{X},\mathbf{X}). \tag{4.21}$$

The posterior GP covariance simplifies to

$$K(\mathbf{x},\mathbf{x}') \;=\; \kappa(\mathbf{x},\mathbf{x}') - \kappa(\mathbf{x},\mathbf{X})\,\mathbf{Q}^{-1}\,\kappa(\mathbf{X},\mathbf{x}'). \tag{4.22}$$

Interpreting the Linearized Laplace Approximation (LLA) as a GP shifts the computational bottleneck from parameter space to data space: evaluating $\mathbf{Q}^{-1}$ costs $\mathcal{O}(N^3 + N^2P)$ operations. For $C$-class classification the kernel matrices have dimension $NC \times NC$; the cubic dependence on both the sample size $N$ and output dimension $C$ must therefore be addressed by further approximations.

### 4.2.3 Using Decoupled SGP and LLA

The decoupled reparameterization of sparse GPs establishes a model where the mean of the approximated posterior distribution is anchored to a pre-trained MAP solution. This method is named *variational LLA* (VaLLA).

**Proposition 4.2.** *For any kernel $K : \mathcal{X} \times \mathcal{X} \to \mathbb{R}$ and its corresponding RKHS $\mathcal{H}$, if $g(\cdot, \hat{\boldsymbol{\theta}}) \in \mathcal{H}$, then $\forall \epsilon > 0$, $\exists M_\alpha \in \mathbb{N}^+$, $\mathbf{Z}_\alpha \in \mathcal{X}^{M_\alpha}$, $\boldsymbol{a} \in \mathbb{R}^{M_\alpha}$ such that, for any $M_\beta \in \mathbb{N}$, $\mathbf{Z}_\beta \in \mathcal{X}^{\mathcal{M}_\beta}$, $\boldsymbol{A} \in \mathbb{R}^{M_\beta \times M_\beta}$ with $\boldsymbol{A} \succeq 0$, the Gaussian measure $Q(f) = \mathcal{N}(f|\tilde{\mu}_{\alpha,\boldsymbol{a}}, \tilde{\Sigma}_{\beta,\boldsymbol{A}}) \in \mathcal{Q}^+$ has a corresponding GP with mean and covariance functions defined as*

$$m^\star(\mathbf{x}) = \tilde{\mu}_{\alpha,\boldsymbol{a}}(\mathbf{x}) \,, \quad K^\star(\mathbf{x}, \mathbf{x}') = K(\mathbf{x}, \mathbf{x}') - K_{\mathbf{x},\mathbf{Z}_\beta}(\boldsymbol{A}^{-1} + \boldsymbol{K}_\beta)^{-1} K_{\mathbf{Z}_\beta,\mathbf{x}'} \,, \tag{4.23}$$

*where $d_{\mathcal{H}}(g(\cdot, \hat{\boldsymbol{\theta}}), \tilde{\mu}_{\alpha,\boldsymbol{a}}) \leq \epsilon$, with $d_{\mathcal{H}}(\cdot, \cdot)$.* [Proof in Appendix C.3]

As a result, the predictive covariance function is

$$K^\star(\mathbf{x}, \mathbf{x}') = K(\mathbf{x}, \mathbf{x}') - K_{\mathbf{x},\mathbf{Z}_\beta}(\boldsymbol{A}^{-1} + \boldsymbol{K}_\beta)^{-1} K_{\mathbf{Z}_\beta,\mathbf{x}'} \,. \tag{4.24}$$

Proposition 4.2 implies that if $g(\cdot, \hat{\boldsymbol{\theta}}) \in \mathcal{H}$, there exist values for $\boldsymbol{a}$ and inducing points for the mean $\mathbf{Z}_\alpha$ such that $d_{\mathcal{H}}(g(\cdot, \hat{\boldsymbol{\theta}}), \tilde{\mu}_{\alpha,\boldsymbol{a}})$ can be made as small as desired. For a sufficiently small $\epsilon$, $m^\star(\cdot) \approx g(\cdot, \hat{\boldsymbol{\theta}})$, and $g(\cdot, \hat{\boldsymbol{\theta}})$ can be used for prediction instead of $m^\star(\mathbf{x})$. Thus, there is no need to optimize $\boldsymbol{a}$ and $\mathbf{Z}_\alpha$ in Equation (4.9), and the posterior distribution of VaLLA uses $g(\cdot, \hat{\boldsymbol{\theta}})$ as its mean function. The optimal parameters $\mathbf{Z}_\beta$ and $\boldsymbol{A}$ can be found by optimizing Equation (4.9) with $\boldsymbol{a}$ and $\mathbf{Z}_\alpha$ held constant. From the following proposition, computing the optimal value of $\mathbf{A}$ has cost $\mathcal{O}(NM_\beta^2 + M_\beta^3)$.

**Proposition 4.3.** *The value of $\boldsymbol{A}$ that minimizes Equation (4.9) is*

$$\boldsymbol{A} = \frac{1}{\sigma^2} \boldsymbol{K}_\beta^{-1} \boldsymbol{K}_{\boldsymbol{Z}_\beta,\boldsymbol{X}} \boldsymbol{K}_{\boldsymbol{X},\boldsymbol{Z}_\beta} \boldsymbol{K}_\beta^{-1} \,, \tag{4.25}$$

*where $\sigma^2$ is the noise variance and $\boldsymbol{K}_{\boldsymbol{X},\boldsymbol{Z}_\beta}$ is a matrix with the prior covariances between $f(\mathbf{X})$ and $f(\mathbf{Z}_\beta)$. If $\boldsymbol{Z}_\beta = \mathbf{X}$, the covariance function of the predictive distribution in Equation (4.24) is equal to that of the full GP.* [Proof in Appendix C.3]

Proposition 4.2 assumes that $g(\cdot, \hat{\boldsymbol{\theta}}) \in \mathcal{H}$. In practice, this may not be the case. Covariance functions such as squared exponential or Matérn are recognized for spanning the entire space of continuous functions. However, whether $g(\cdot, \hat{\boldsymbol{\theta}}) \in \mathcal{H}$ holds in general remains unknown. From here onward, if $g(\cdot, \hat{\boldsymbol{\theta}}) \notin \mathcal{H}$, then $\mathcal{H}$ is assumed to be sufficiently expressive to include a close approximation to $g(\cdot, \hat{\boldsymbol{\theta}})$. Consequently, $g(\cdot, \hat{\boldsymbol{\theta}})$ can be used as the sparse GP posterior mean.

#### 4.2.3.1 Hessian Approximation in VaLLA

Despite the formulation using GPs function-space duality, VaLLA can also be understood as a Hessian approximation method. Note that the predictive covariances of VaLLA are given in Equation (4.24). From this equation, a connection with the exact posterior variances given by the full Hessian in Equation (4.19) can be made. From there, one can conclude that VaLLA's inverse negative Hessian approximation is given by:

$$\sigma_0^2 \boldsymbol{I}_P - \sigma_0^2 \Phi_{\mathbf{Z}_\beta} (\boldsymbol{A}^{-1} + \sigma_0^2 \Phi_{\mathbf{Z}_\beta}^T \Phi_{\mathbf{Z}_\beta})^{-1} \Phi_{\mathbf{Z}_\beta}^T \sigma_0^2 \,, \tag{4.26}$$

which is equal to $(\boldsymbol{I}_P/\sigma_0^2 + \Phi_{\mathbf{Z}_\beta}^T \boldsymbol{A} \Phi_{\mathbf{Z}_\beta})^{-1}$ using the Woodbury formula, where $\boldsymbol{A}$ is a free parameter adjusted by VaLLA by minimizing the KL divergence between stochastic processes, as described above. VaLLA's negative Hessian approximation is the inverse of the previous quantity. Namely,

$$\boldsymbol{I}_P/\sigma_0^2 + \Phi_{\mathbf{Z}_\beta}^T \boldsymbol{A} \Phi_{\mathbf{Z}_\beta} \,, \tag{4.27}$$

which is similar in structure to the GGN approximation in Equation (4.14) considered by LLA, where the data-dependent term has been replaced by the inducing points and the matrix $\boldsymbol{A}$.

VaLLA's predictive mean is anchored to the DNN output, making the maximization of the ELBO in Equation (4.9) unsuitable for tuning hyper-parameters like the prior variance $\sigma_0^2$. Specifically, in a regression scenario with Gaussian noise with variance $\sigma^2$ the first term in the r.h.s. of Equation (4.9) becomes:

$$\sum_{i=1}^{N} -\frac{\log(2\pi\sigma^2)}{2} - \frac{(y_i - g(\mathbf{x}_i, \hat{\boldsymbol{\theta}}))^2}{2\sigma^2} - \frac{K(\mathbf{x}_i, \mathbf{x}_i)}{2\sigma^2} \,, \tag{4.28}$$

where $(y_i - g(\mathbf{x}_i, \hat{\boldsymbol{\theta}}))^2$ is constant. Maximizing Equation (4.28) w.r.t. $\sigma_0^2$ results in the prior covariances, $\sigma_0^2 J_{\hat{\boldsymbol{\theta}}}(\mathbf{x})^{\text{T}} J_{\hat{\boldsymbol{\theta}}}(\mathbf{x}')$, tending to $0$. This makes posterior covariances $K(\mathbf{x}_i, \mathbf{x}_i)$ also tend to $0$, effectively canceling the last term in Equation (4.28). The KL term in Equation (4.9) is also optimal and $0$ if $\sigma_0^2 \to 0$. The reasoning is that, in sparse GPs, tuning hyper-parameters involves a trade-off between fitting the mean to the training data and reducing the predictive variance of the model. Therefore, in VaLLA's setting, where the predictive mean is fixed, the optimal predictive variance tends to be zero.

To address these issues, an alternative objective to Equation (4.9) is proposed, making use of $\alpha$-divergences. This new objective facilitates hyper-parameter optimization:

$$\max_{Q(f)} \sum_{i=1}^{N} \frac{1}{\alpha} \log \mathbb{E}_Q \left[ P(y_i|f)^\alpha \right] - \text{KL}\left( Q|P \right) \,. \tag{4.29}$$

Here $\alpha \in (0, 1]$ is a parameter. Instead of minimizing $\text{KL}(q(f)|P(f|\mathbf{y}))$, this objective minimizes the $\alpha$-divergence between $P(f|\mathbf{y})$ and $Q(f)$ (Y. Li and Gal, 2017). Remarkably, this can be achieved by simply changing the data-dependent term in the objective of Equation (4.9).

The use of $\alpha$-divergences (see Section 2.2.4 for more details) for approximate inference has been extensively studied (T. D. Bui et al., 2017; Rodríguez-Santana and Hernández-Lobato, 2022; Villacampa-Calvo and Hernández-Lobato, 2020), with observations suggesting that values of $\alpha \approx 0$ result in better predictive mean estimation. Conversely, values of $\alpha \approx 1$ provide superior predictive distributions in terms of the log-likelihood. In all the conducted experiments, $\alpha = 1$. In this case, Equation (4.29) does not promote $\sigma_0^2 \to 0$, unlike Equation (4.9), as the data-dependent term is the log-likelihood of the training data. An unexpected behavior, however, is that Equation (4.29) may lead to overfitting. To alleviate this, the proposed solution is the use of early-stopping with a validation set (see Section 4.2.6.2 for further details). Early stopping is also used in other LLA approximations *e.g.* ELLA (Z. Deng et al., 2022).

The objective in Equation (4.29) supports mini-batch optimization with cost $\mathcal{O}(M_\beta^3)$. For $\alpha = 1.0$,

$$N|\mathcal{B}|^{-1}\sum_{b\in\mathcal{B}} \log \mathbb{E}_Q\left[P(y_b|f(\mathbf{x}_b))\right] - \mathrm{KL}\left(Q|P\right) . \tag{4.30}$$

Here, $\mathcal{B}$ denotes a mini-batch, and the expectation can be computed in closed form in regression problems. In classification, an approximation is available via using the softmax approximation of Daxberger et al. (2021b). This sub-linear cost of VaLLA enables its use in very large datasets.

Predictions for test points $(y^\star, \mathbf{x}^\star)$ are computed using Equation (4.24) with the DNN output $g(\mathbf{x}^\star, \hat{\boldsymbol{\theta}})$ as the mean. $P(y^\star|\mathbf{x}^\star) \approx \mathbb{E}_Q[P(y^\star|f(\mathbf{x}^\star)]$ is evaluated as in training.

The locations of the inducing points $\mathbf{Z}_\beta$ are found by optimizing Equation (4.30) with K-means initialization.

#### 4.2.3.2 Limitations of VaLLA

VaLLA is limited by three factors: (i) Computing the predictive distribution at each training iteration involves inverting $\mathbf{A}^{-1} + K_{\mathbf{Z}_\beta}$ in Equation (4.24), with cubic cost in the number of inducing points $M_\beta$. Therefore, VaLLA cannot accommodate a very large number of inducing points. (ii) The objective in Equation (4.29) requires a validation set and early stopping for effective optimization of the prior variance $\sigma_0^2$, thus further increasing training time. (iii) VaLLA requires additional training compared to other LLA approximations. However, in this regard, early stopping can also reduce the training time by cutting down the number of iterations. In the Taxi experiments performed in Section 4.2.5, early stopping is triggered when only $16.6\%$ of the training data has been seen. (iii) Mini-batch optimization in Equation (4.30) involves evaluating $K_{\mathbf{x},\mathbf{Z}_\beta}$ $\forall \mathbf{x} \in \mathcal{B}$ and $\boldsymbol{K}_\beta$. Hence, VaLLA requires efficient evaluation of the (scaled) Neural Tangent Kernel, $\kappa(\cdot,\cdot) = \sigma^2 J_{\hat{\boldsymbol{\theta}}}(\cdot)^{\mathrm{T}} J_{\hat{\boldsymbol{\theta}}}(\cdot)$ and its gradients to find $\mathbf{Z}_\beta$. While there are libraries that use structure in the derivatives for the efficient computation of $\kappa(\cdot,\cdot)$, these are limited to a few DNN models (Novak et al., 2022). A simple but inefficient approach to evaluate $\kappa(\cdot,\cdot)$ involves computing and storing all full Jacobians in memory, for each mini-batch instance and inducing point. This is tractable in many problems, but makes VaLLA infeasible for very large problems, *e.g.*, ImageNet. Appendix 4.2.6.1 shows a very efficient *layer-by-layer* method to obtain $K_{\mathbf{x},\mathbf{Z}_\beta}$ $\forall \mathbf{x} \in \mathcal{B}$ and $\boldsymbol{K}_\beta$. However, this requires computing each layer's contribution to the Jacobian at hand, which is difficult for large and complex DNNs.

### 4.2.4 Related Work

LA for DNNs was originally introduced by MacKay (1992a), applying it to small networks using the full Hessian. MacKay (1992b) also proposed an approximation similar to the generalized Gauss-Newton (GGN). The combination of scalable factorizations or diagonal Hessian approximations (Botev et al., 2017; Martens and Grosse, 2015) with the GGN approximation (Martens, 2020) played a crucial role in the resurgence of LA for modern DNNs (Khan et al., 2019; Ritter et al., 2018). Recent works aim to relax the Gaussian assumption of LLA, adopting a Riemannian-Laplace approximation, where samples naturally fall into weight regions with low negative log-posterior (Bergamin et al., 2023).

To address the underfitting issue associated with LA (Lawrence, 2001), particularly when combined with the GGN approximation, Ritter et al. (2018) proposed a Kronecker factored (KFAC) LLA approximation. This approach outperforms LA with a diagonal Hessian matrix.

The GP interpretation of LLA (Khan et al., 2019) allows using GP approximate methods to speed up the computations. Immer et al. (2021) propose to use a subset of the training dataset as a scalable alternative to the true GP. Jongseok Lee et al. (2022) propose a Mixture of Experts approach where each expert is trained on a different soft-margin cluster. However, the proposed clustering algorithm, although more efficient than Kernel-K-means, has linear cost w.r.t. the training set size. VaLLA, on the other hand, has sub-linear training time w.r.t. training set size due to mini-batch training. Moreover, it is not clear how to consider neighboring clusters in high dimensional input spaces. The authors only provide code for a 1-dimensional problem. Third, fitting a local GP using the data of the corresponding cluster and its neighbors is expected to overestimate the predictive variance since the model has been trained with a smaller number of training instances (see Figure 4.9). This is particularly the case in datasets with millions of training instances such as Taxi. This problem is also described by Immer et al. (2021).

Z. Deng et al. (2022) proposed a Nyström approximation of the true GP covariance matrix by using $M \ll N$ points chosen at random from the training set. The method, called ELLA, has cost $\mathcal{O}(NM^3)$. ELLA also requires computing the costly Jacobian vectors required in VaLLA, but does not need their gradients. Unlike VaLLA, the Nyström approximation needs to visit each instance in the training set. However, as stated by Z. Deng et al. (2022), ELLA suffers from overfitting. An early-stopping strategy, using a validation set, is proposed to alleviate it. In this case, ELLA only considers a subset of the training data. ELLA does not allow for hyperparameter optimization, unlike VaLLA. The prior variance $\sigma_0^2$ must be tuned using grid search and a validation set, which increases training time significantly.

The recent work of Scannell et al. (2024) proposes a similar approach to VaLLA, where an inducing point sparse approach is used to construct a GP from a pre-trained DNN. However, two main points differentiate this work from the proposed approach: (i) the pre-trained DNN is not kept as the posterior mean of the model, potentially losing prediction performance and also departing from LLA's post-hoc nature and goal; (ii) instead of using mini-batches to optimize variational parameters, they perform a full iteration over the training data to find optimal variational parameters. Thus, this results in a potentially slower method than VaLLA, which, due to early stopping and stochastic optimization, can avoid iterating over the full dataset.

Samples from a GP posterior can be efficiently computed using stochastic optimization, eluding the explicit inversion of the kernel matrix (Lin et al., 2024). This approach can be extended to LLA to generate samples from the GP posterior, avoiding the $\mathcal{O}(N^3)$ cost (Antorán et al., 2023). However, this method cannot provide an estimate of the log-marginal likelihood for hyper-parameter optimization. To address this limitation, Antorán et al. (2023) propose using the *EM-algorithm*, where samples are generated (E-step) and hyper-parameters are optimized afterwards (M-step) iteratively. The EM algorithm significantly increases computational cost, as generating a single sample is as expensive as training the original DNN on the full data. Finally, the method of Antorán et al. (2023) only considers classification problems.

Another GP-based approach for obtaining prediction uncertainty in the context of DNNs is the Spectral-normalized Neural Gaussian Process (SNGP) (J. Z. Liu et al., 2023), where the last layer of a DNN is replaced by a GP. This approach allows us to either (i) fine-tune a pre-trained DNN model, or (ii) train a full DNN model from scratch. The former method is considered for the conducted experiments. However, replacing the last layer with a GP often reduces the prediction performance of the initial DNN. This is also observed in the results of J. Z. Liu et al. (2023). As a result, this method also lies outside LLA-based methods' main objective, which is to preserve the initial DNN predictions.

### 4.2.5 Experiments

VaLLA is evaluated against the canonical LLA baselines implemented in the `Laplace` library of Daxberger et al. (2021a). VaLLA is trained with a fixed mini-batch size of 100. For the regression, MNIST, and Fashion-MNIST benchmarks, a conventional multilayer perceptron (MLP) is employed; the resulting model checkpoints are archived to guarantee full reproducibility. For CIFAR-10, the same ResNet backbone employed by Z. Deng et al. (2022) is adopted, where their published metrics are quoted for all competing methods to maintain strict architectural parity. The prior variance of every LLA variant—diagonal, Kronecker-factored (KFAC), and last-layer—is tuned by maximizing the approximate log marginal likelihood. Specifically, the `log_marginal_likelihood` routine in the `Laplace` library is used to optimize them for 40 000 Adam steps with a learning rate of $10^{-3}$.

The complete VaLLA implementation is publicly available at https://github.com/Ludvins/Variational-LLA.

#### 4.2.5.1 Synthetic Regression

The predictive distribution of VaLLA is compared with that of LLA (which is considered the optimal method), other LLA variants, and ELLA, on the 1-D regression problem of Izmailov et al. (2020). In ELLA and VaLLA, the optimal hyper-parameters from LLA are employed for this experiment. The results in Figure 4.1 illustrate that VaLLA's predictive distribution closely aligns with that of LLA. In the ablation studies, Figure 4.7 depicts the predictive distributions of VaLLA and ELLA for varying numbers of inducing points and points in the Nyström approximation, respectively. It shows that VaLLA converges to the true posterior faster than ELLA, with VaLLA tending to overestimate the predictive variance while ELLA underestimates it. In Figure 4.6, the effect of tuning the prior variance in VaLLA is observed in another toy 1-D problem, with and without early stopping. Notably, early stopping, using a validation set, prevents overly small predictive variances in VaLLA. Finally, given that VaLLA estimates the prior variance by maximizing Equation (4.29), it tends to underestimate LLA's predictive variance.

#### 4.2.5.2 Airline, Year and Taxi Regression Problems

For regression problems (Year, Airline, and Taxi datasets), a 3-layer fully connected NN with $200$ units per layer was used. Optimal weights were obtained by minimizing the RMSE over $20{,}000$ iterations with a batch size of $100$ and the Adam optimizer (Kingma and Ba, 2015) with learning rate $10^{-2}$ and weight decay $10^{-2}$.

Three large regression datasets are used to validate the performance of VaLLA: (i) the *Year* dataset (UCI) with $515\ 345$ instances and $90$ features, using the original train/test splits; (ii) the *US flight delay* (*Airline*) dataset (Dutordoir et al., 2020). Following Luis A. Ortega et al. (2023), the first $700\ 000$ instances are used for training and the next $100\ 000$ for testing. Eight features are considered: month, day of the month, day of the week, plane age, air time, distance, arrival time, and departure time; (iii) the *Taxi* dataset, with data recorded in January 2023 (Salimbeni and Deisenroth, 2017). Nine attributes are considered: time of day, day of week, day of month, month, `PULocationID`, `DOLocationID`, distance, and duration. Trips shorter than 10 seconds and longer than 5 hours are filtered out, resulting in 3 million instances. The first $80\%$ is used as training data, the next $10\%$ as validation data, and the last $10\%$ as test data. In all experiments, a 3-layer DNN with 200 units, *tanh* activations, and $\ell_2$ regularization is employed. VaLLA and ELLA use $100$ inducing points and $100$ random points, respectively. A total of $40\ 000$ iterations with mini-batch size $100$ are performed for VaLLA. However, for the Taxi dataset, which contains nearly 3 million instances, early stopping terminates training at $5\ 000$ iterations for one of the random seed initializa-tions (this value differs across seeds). Thus, 500 000 points are visited during training for that seed, corresponding to only 16.6% of the complete dataset.

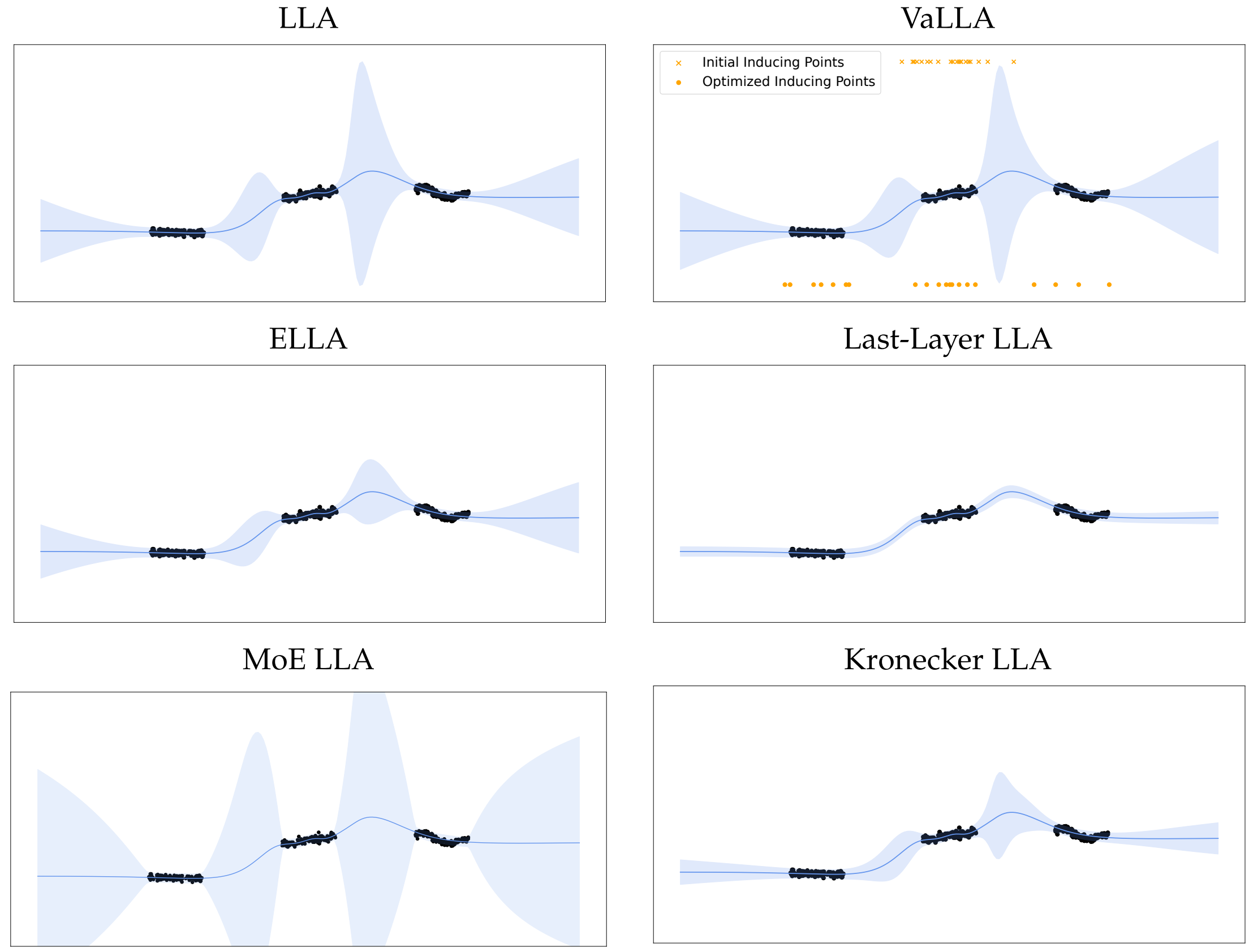

**Figure 4.1:** ***Predictive mean and 95 % credible regions produced by six Laplace–based approximations on a toy one-dimensional regression problem.*** *Each panel shows the posterior mean (blue line) and its* $\pm 2$ *s.d. envelope (blue shade) obtained with Last-Layer Laplace Approximation (LLA), Variational LLA (VaLLA), Empirical LLA (ELLA), diagonal LLA, mixture-of-experts LLA (MoE–LLA, 200 clusters) and Kronecker-factorized LLA. All models share a two-layer MLP with 50 hidden units whose noise and prior hyper-parameters are fixed by maximizing the marginal log-likelihood under standard LLA; those values are reused by VaLLA and ELLA. VaLLA represents predictive covariances with 20 inducing points, whereas ELLA relies on 20 random input locations and 20 random Fourier features. Across the input space VaLLA yields uncertainty bands that are at least as well-calibrated as, and often sharper than, those of competing Laplace variants.*

Table 4.1 presents the averaged results over 5 random seeds. LLA is not considered here due to intractability. Negative log likelihood (NLL), continuous ranked probability score (CRPS) (Gneiting and Raftery, 2007) and a centered quantile metric (CQM), described below, are reported. VaLLA performs best according to NLL and CQM, while it gives worse results in terms of CRPS compared to the other methods.

**Centered Quantile Metric (CQM).** The *Centered Quantile Metric* (CQM) is introduced as a calibration measure for regression models whose predictive distributions are Gaussian *with a common mean*. CQM extends the well-known *Expected Calibration Error* from classification to this setting by comparing the nominal and empirical cover-age of centered quantile intervals.

**Table 4.1: *Performance of VaLLA and alternative Laplace–based regressors on three tabular benchmarks.*** *The table reports the test Negative Log-Likelihood (NLL), Continuous Ranked Probability Score (CRPS) and Calibrated Quantile Metric (CQM) for the Airline, Year and Taxi datasets, averaged over five random seeds (all standard deviations $< 10^{-4}$ and therefore omitted). Competing methods include the maximum-a-posteriori baseline (MAP); diagonal, Kronecker-factorized (KFAC) and last-layer ($^\star$) variants of the Laplace approximation (LLA); the accelerated Laplace approximation (ELLA); and VaLLA with 100 or 200 inducing points. The best value for each metric–dataset pair is highlighted in* ***purple****, and the second best in* ***teal****.*

| | Airline | | | Year | | | Taxi | | |
|---|---|---|---|---|---|---|---|---|---|
| Model | NLL | CRPS | CQM | NLL | CRPS | CQM | NLL | CRPS | CQM |
| MAP | 5.087 | 18.436 | 0.158 | 3.674 | 5.056 | 0.164 | 3.763 | **3.753** | **0.227** |
| LLA Diag | 5.096 | **18.317** | 0.144 | 3.650 | 4.957 | 0.122 | 3.714 | 3.979 | 0.270 |
| LLA KFAC | 5.097 | **18.317** | 0.144 | 3.650 | **4.955** | 0.121 | 3.705 | 3.977 | 0.270 |
| LLA$^*$ | 5.097 | **18.319** | 0.144 | 3.650 | **4.954** | 0.120 | 3.718 | 3.975 | 0.270 |
| LLA$^*$ KFAC | 5.097 | **18.317** | 0.144 | 3.650 | **4.954** | 0.120 | 3.718 | 3.976 | 0.270 |
| ELLA | 5.086 | 18.437 | 0.158 | 3.674 | 5.056 | 0.164 | 3.753 | **3.754** | **0.227** |
| VaLLA 100 | **4.923** | 18.610 | **0.109** | **3.527** | 5.071 | **0.084** | **3.287** | 3.968 | **0.188** |
| VaLLA 200 | **4.918** | 18.615 | **0.107** | **3.493** | 5.026 | **0.076** | **3.280** | 3.993 | **0.188** |

For a test input $\mathbf{x}$ with predictive distribution $\mathcal{N}(\mu(\mathbf{x}), \sigma^2(\mathbf{x}))$, define the $\alpha$–central interval

$$I(\mathbf{x}, \alpha) = (\lambda(-\alpha), \lambda(\alpha)), \qquad \lambda(\alpha) = \Phi^{-1}_{\mu(\mathbf{x}),\sigma^2(\mathbf{x})}\left(\frac{1+\alpha}{2}\right), \quad \alpha \in (0,1), \tag{4.31}$$

where $\Phi^{-1}_{\mu,\sigma^2}$ denotes the inverse CDF of a Gaussian with mean $\mu$ and variance $\sigma^2$. Let

$$\gamma(\alpha) = \mathbb{P}_{(\mathbf{x},y)}\left[y \in I(\mathbf{x}, \alpha)\right] \tag{4.32}$$

be the fraction of test observations whose targets fall inside that interval. A perfectly calibrated model satisfies $\gamma(\alpha) = \alpha$ for every $\alpha$. CQM aggregates the deviation from this ideal across all confidence levels:

$$\text{CQM} = \int_0^1 |\gamma(\alpha) - \alpha| \, d\alpha. \tag{4.33}$$

All LLA methods retain the maximum-a-posteriori estimate as predictive mean, so differences in $I(\mathbf{x}, \alpha)$ arise solely from their predictive variances. The integral in Equation (4.33) is approximated on an eleven-point grid over $\alpha$, which simultaneously yields a scalar summary and the calibration curve $\gamma(\alpha)$ (Figure 4.2). Figure 4.2 shows the evolution of $\mathbb{P}_{(\mathbf{x}^\star,y^\star)}\left[y^\star \in I(\mathbf{x}^\star, \alpha)\right]$ w.r.t. $\alpha$ for the best performing models in the regression problems. CQM corresponds to the area between the shown curve and $y = x$ (black line). This figure allows us to argue that (in general) all methods are over-estimating the predictive variance as they are giving values above the diagonal. That is, for a specific value of $\alpha \in (0, 1)$, the reported probabilities are higher than $\alpha$, meaning that, on average, there are more points in $I(\mathbf{x}, \alpha)$ than they should. That is, the predicted inter-val is larger than it should be, which can only mean that the variance is over-estimated. From a geometrical perspective, it is clear that CQM is always greater than 0 and lower than 0.5; independently of the model and dataset used.

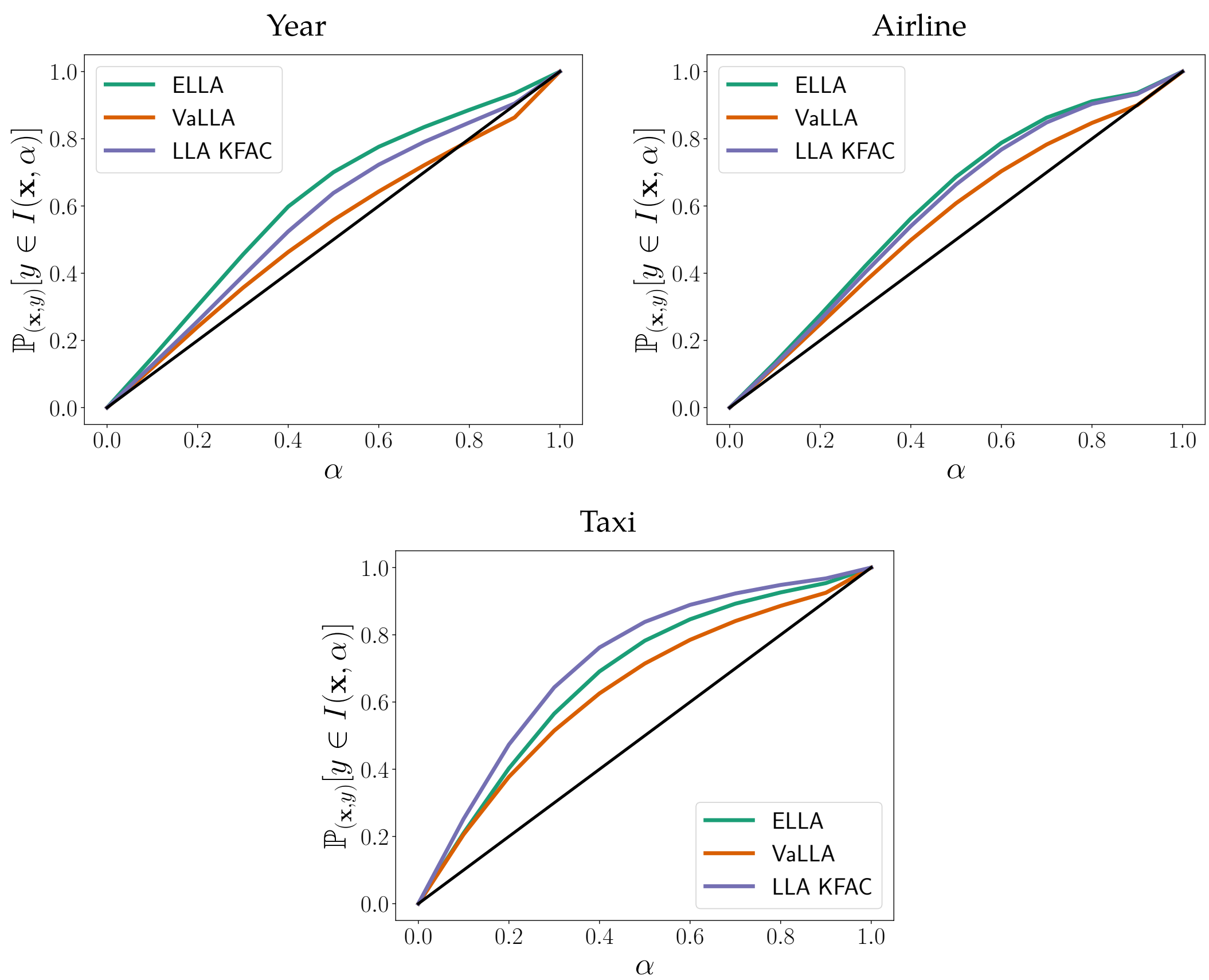

**Figure 4.2:** ***Centered Quantile Metric (CQM) calibration curves.*** *Empirical coverage $\gamma(\alpha)$ (vertical axis) is plotted against the nominal coverage $\alpha$ (horizontal axis) for the Year (top left), Airline (top right) and Taxi (bottom) regression tasks. Curves correspond to the Accelerated Laplace (ELLA), Variational LLA (VaLLA) and LLA–KFAC variants, while the diagonal $y = \alpha$ (black) marks perfect calibration. All methods reuse the same MAP predictive mean given by a fully–connected network with three hidden layers of 200 TANH units; discrepancies therefore arise solely from their predictive variances. Lines lying above the diagonal indicate over–estimation of uncertainty, whereas those below indicate under–estimation.*

In fact, this figure allows us to visually study the level of over/infra-estimation of the prediction uncertainty, for each degree of confidence $\alpha$. For example, in the Year dataset, VaLLA slightly over-estimates the uncertainty for $\alpha \in (0, 0.7)$ while it infra-estimates it for larger values of $\alpha$.

#### 4.2.5.3 Image Classification Problems

**MNIST and FMNIST.** For MNIST and FMNIST experiments, a 2-layer fully connected NN was used with 200 units in each layer. The optimal weights are obtained by minimizing the NLL using 20 000 iterations of batch size 100 and Adam optimizer (Kingma and Ba, 2015) with learning rate $10^{-3}$ and weight decay $10^{-3}$.

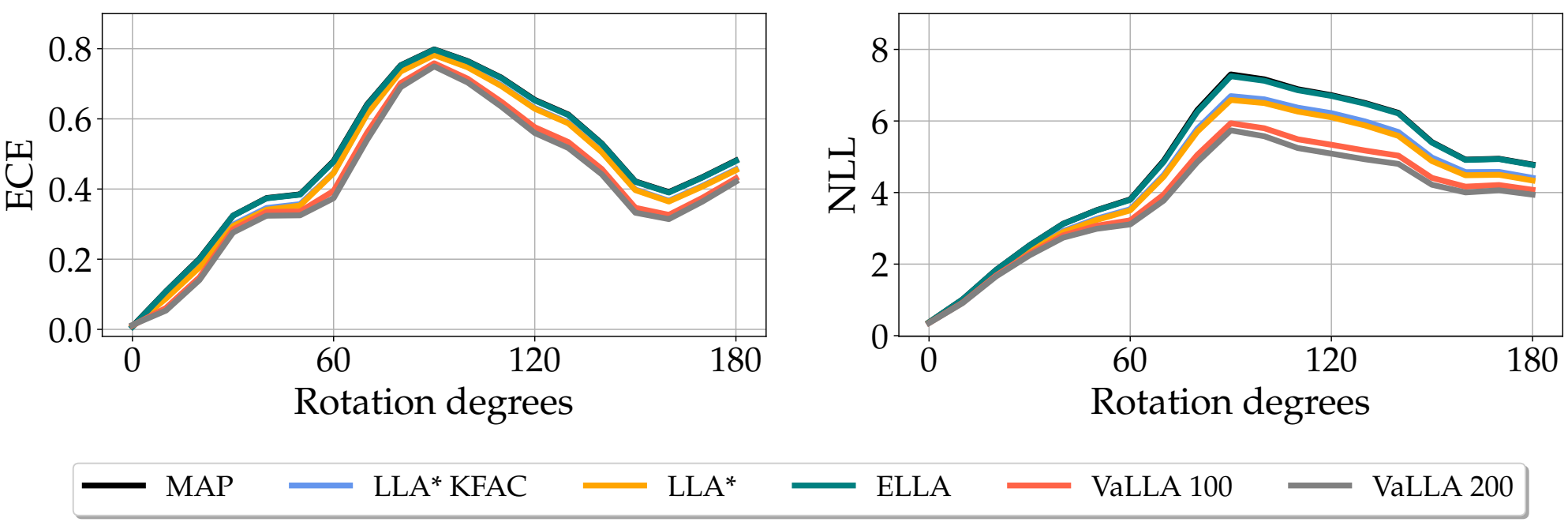

**Figure 4.3:** ***Expected calibration error and negative log-likelihood for rotated FMNIST as a function of input rotation angle.*** *The left panel plots the expected calibration error (ECE), and the right panel plots the negative log-likelihood (NLL), for rotation angles from 0° to 180°. Curves are shown for six inference schemes: maximum-a-posteriori (MAP), Kronecker-factorized last-layer Laplace (LLA* KFAC), last-layer Laplace (LLA*), accelerated Laplace (ELLA), and variational Laplace with 100 or 200 inducing points (VaLLA 100 / VaLLA 200).*

**Table 4.2:** ***Performance comparison on the MNIST (left) and Fashion-MNIST (right) datasets.*** *Results are averaged over five random seeds (standard deviations $< 10^{-4}$ and therefore omitted). The best metric is highlighted in* ***purple****; the second-best in* ***teal****. The asterisk (*) denotes variants that restrict the Laplace approximation to the last layer.*

**(a)** *MNIST*

| Model | ACC | NLL | ECE | BRIER | OOD-AUC |
|---|---|---|---|---|---|
| MAP | **97.6** | **0.076** | **0.008** | **0.036** | 0.905 |
| LLA Diag | 97.4 | 0.143 | 0.072 | 0.053 | 0.922 |
| LLA KFAC | 97.5 | 0.094 | 0.029 | 0.041 | **0.949** |
| LLA* | **97.6** | 0.081 | 0.015 | 0.037 | 0.909 |
| LLA* KFAC | **97.6** | 0.081 | 0.015 | 0.037 | 0.909 |
| ELLA | **97.6** | **0.076** | **0.008** | **0.036** | 0.905 |
| Sampled LLA | **97.6** | 0.087 | 0.026 | 0.040 | **0.954** |
| VaLLA 100 | **97.7** | **0.076** | **0.010** | **0.036** | 0.916 |
| VaLLA 200 | **97.7** | **0.075** | **0.010** | **0.035** | 0.921 |

**(b)** *FMNIST*

| Model | ACC | NLL | ECE | BRIER | OOD-AUC |
|---|---|---|---|---|---|
| MAP | 86.6 | 0.373 | **0.009** | 0.193 | 0.874 |
| LLA Diag | 86.2 | 0.397 | 0.043 | 0.201 | 0.914 |
| LLA KFAC | 86.5 | 0.377 | 0.014 | 0.194 | **0.932** |
| LLA* | 86.6 | 0.373 | **0.008** | 0.193 | 0.882 |
| LLA* KFAC | 86.6 | 0.373 | **0.008** | 0.193 | 0.880 |
| ELLA | 86.6 | 0.373 | **0.008** | 0.193 | 0.874 |
| VaLLA 100 | **87.4** | **0.335** | 0.011 | **0.182** | 0.923 |
| VaLLA 200 | **87.6** | **0.332** | 0.013 | **0.181** | **0.933** |

A fully connected DNN with 200 units in each layer and *tanh* activations is employed. In VaLLA, 100 and 200 inducing points are considered, whereas ELLA uses 2 000 random points. The out-of-distribution (OOD) detection ability of each method is evaluated using the entropy of the predictive distribution as the score. The area under the ROC curve (AUC) is computed for the binary problem that distinguishes between instances from the dataset pairs MNIST/FMNIST and FMNIST/MNIST (Immer et al., 2021). Moreover, on FMNIST, the robustness of the predictive distribution is assessed by rotating the test images by up to 180° and computing the ECE and NLL on the rotated images (Ovadia et al., 2019) (Figure 4.3).

Table 4.2 shows the results on MNIST. VaLLA gives better uncertainty estimates in terms of NLL and the Brier score but performs less effectively in terms of ECE. Remarkably, VaLLA improves prediction accuracy (ACC) due to the approximation of Daxberger et al. (2021b) to compute class probabilities in multi-class problems. In terms of OOD-AUC, VaLLA outperforms the MAP solution but lags behind other methods *s.a.* Sampled-LLA or LLA with Kronecker approximations. Figure 4.4 illustrates the training times for each method, with VaLLA being faster than ELLA, Sampled-LLA, or Last-Layer LLA.

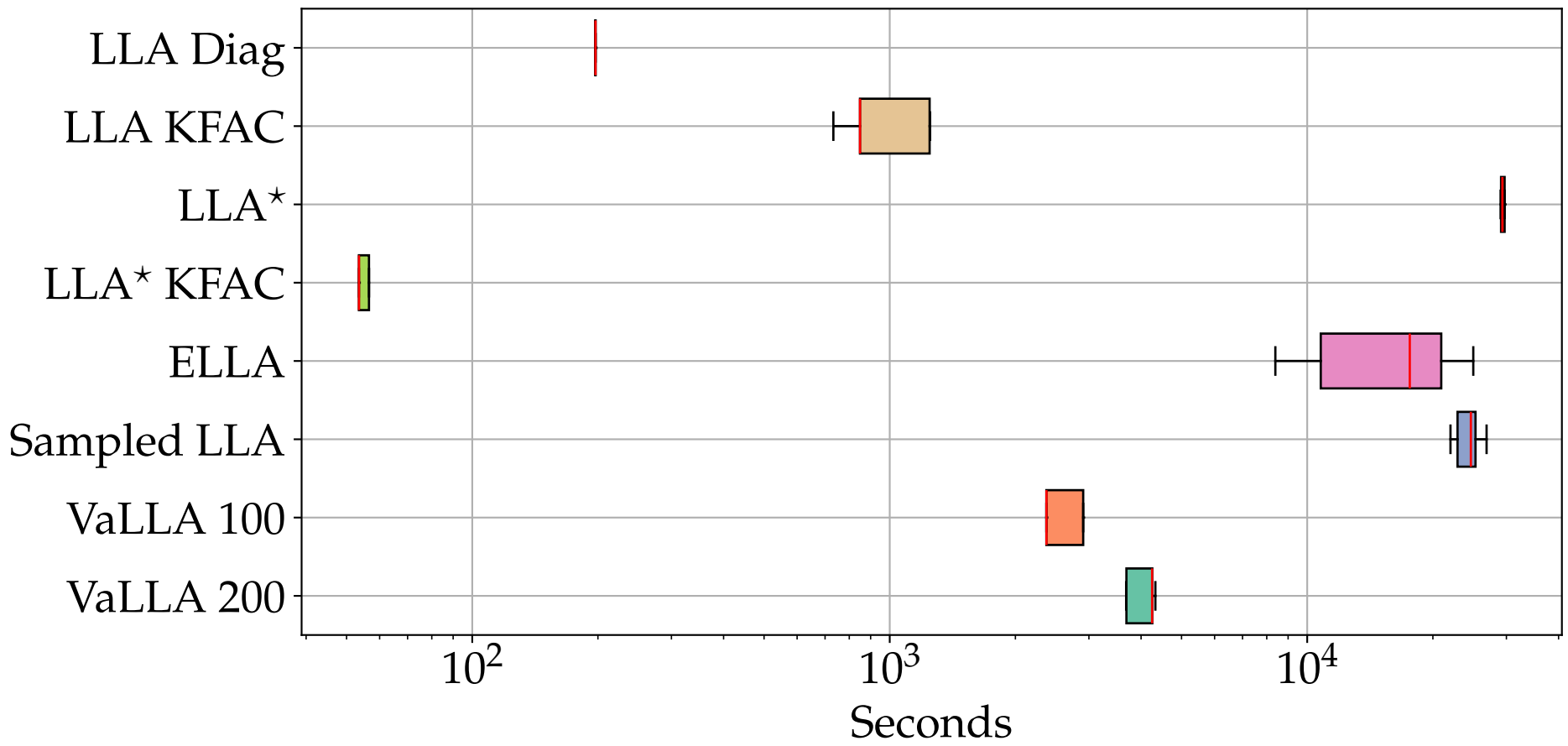

**Figure 4.4:** ***Box-plots of wall-clock training time for eight Laplace-based inference schemes.*** *The plot shows the distribution of total training time in seconds (logarithmic $x$-axis) across repeated runs for LLA Diag, LLA KFAC, last-layer Laplace (LLA*), last-layer KFAC (LLA* KFAC), accelerated Laplace (ELLA), sampled LLA, and VaLLA with 100 or 200 inducing points. For ELLA, ten candidate prior variances are evaluated on a validation set; Sampled-LLA is executed with eight EM iterations and 32 Monte-Carlo samples per step. The asterisk (*) denotes variants that restrict the Laplace approximation to the last layer.*

Finally, Table 4.2 displays the results on FMNIST. Here, VaLLA excels in prediction accuracy and provides the best uncertainty estimates in terms of NLL and the Brier score. Although it does not perform as well in terms of ECE, the differences are small. VaLLA also achieves the best results in OOD-AUC. Figure 4.3 shows VaLLA holds better performance in terms of ECE and NLL as the test images' corruption increases (rotation level), indicating the greater robustness of VaLLA's predictive distribution.

**CIFAR10 and ResNet.** Various ResNet architectures are used, and the corresponding pre-trained models are those of Z. Deng et al. (2022) (accessible at https://github.com/chenyaofo/pytorch-cifar-models). Table 4.3 reports ACC, NLL, and ECE for each method, including LLA variants, a mean-field VI approach (Z. Deng and Zhu, 2023), fine-tuned SNGP (J. Z. Liu et al., 2023), and the approach of Immer et al. (2021) that uses the GP interpretation of LLA with a random subset of 500 training instances (GP-Subset). For the latter, the prior parameter was set to the weight decay value used when training the MAP solution, $\sigma_0^2 = 0.04$, as suggested by Immer et al. (2021). The prior was not scaled by the subset size, as scaling resulted in worse performance. For this experiment, VaLLA is trained for $40\,000$ iterations or until Early-Stopping triggers. Since the same pre-trained models are used, the results of all other methods are consistent with those reported by Z. Deng et al. (2022). VaLLA with $M_\beta = 100$ outperforms other methods in most cases, consistently ranking as the best or second-best method. Figure 4.5 shows the NLL on the perturbed test set with five increasing levels of 19 image corruptions (Z. Deng et al., 2022). Each box plot summarizes the test NLL for each intensity level across all 19 corruptions. The results again highlight VaLLA's robust predictive distribution, achieving lower NLL compared to the other methods.

**Table 4.3:** ***Predictive accuracy, negative log-likelihood and expected calibration error for four ResNet depths on CIFAR-10.*** *The table reports Monte-Carlo estimates (64 samples) of test accuracy (ACC ↑), negative log-likelihood (NLL ↓) and expected calibration error (ECE ↓) for ResNet-20, -32, -44 and -56, together with a mean rank computed from NLL and ECE only. Methods include maximum-a-posteriori (MAP), mean-field variational inference (MF-VI), Spectral-normalized Neural Gaussian Process (SNGP), Gaussian-process with a subset of data points (GP–Subset), diagonal and Kronecker-factorized Laplace (LLA Diag / LLA KFAC), last-layer Laplace variants (*) with and without KFAC, accelerated LLA (ELLA, $M = 2\,000$, $K = 20$), sampled LLA (64 samples) and variational Laplace (VaLLA). All entries are averaged over five random seeds (standard deviations $< 10^{-3}$ and therefore omitted). The best value per metric–architecture pair is highlighted in* ***purple*** *and the second-best in* ***teal****; the asterisk (*) denotes last-layer Laplace approximations.*

| | ResNet-20 | | | ResNet-32 | | | ResNet-44 | | | ResNet-56 | | | Rank |
|---|---|---|---|---|---|---|---|---|---|---|---|---|---|
| Method | ACC | NLL | ECE | ACC | NLL | ECE | ACC | NLL | ECE | ACC | NLL | ECE | |
| MAP | **92.6** | 0.282 | 0.039 | **93.5** | 0.292 | 0.041 | **94.0** | 0.275 | 0.039 | **94.4** | 0.252 | 0.037 | – |
| MF-VI | **92.7** | **0.231** | 0.016 | **93.5** | 0.222 | 0.020 | **93.9** | 0.206 | 0.018 | **94.4** | 0.188 | 0.016 | – |
| SNGP | 92.4 | 0.266 | 0.024 | 93.2 | 0.256 | 0.025 | 93.8 | 0.242 | 0.028 | 93.8 | 0.229 | 0.022 | – |
| GP - Subset | **92.6** | 0.555 | 0.299 | **93.4** | 0.462 | 0.247 | 93.6 | 0.424 | 0.225 | **94.4** | 0.403 | 0.221 | – |
| LLA Diag | 92.2 | 0.728 | 0.404 | 92.7 | 0.755 | 0.430 | 92.8 | 0.778 | 0.445 | **92.9** | 0.843 | 0.480 | – |
| LLA KFAC | 92.0 | 0.852 | 0.467 | 91.8 | 1.027 | 0.547 | 91.4 | 1.091 | 0.566 | 89.8 | 1.174 | 0.579 | – |
| LLA* | **92.6** | 0.269 | 0.034 | **93.5** | 0.259 | 0.033 | **94.0** | 0.237 | 0.028 | **94.4** | 0.213 | 0.022 | – |
| LLA* KFAC | **92.6** | 0.271 | 0.035 | **93.5** | 0.260 | 0.033 | **94.0** | 0.232 | 0.028 | **94.4** | 0.202 | 0.024 | – |
| ELLA | 92.5 | 0.233 | 0.009 | **93.5** | **0.215** | **0.008** | **93.9** | 0.204 | **0.007** | **94.4** | 0.187 | **0.007** | 2.37 |
| Sampled LLA | 92.5 | **0.231** | **0.006** | **93.5** | 0.217 | **0.008** | **94.0** | **0.200** | **0.007** | **94.4** | **0.185** | 0.015 | **2.00** |
| VaLLA | **92.6** | **0.228** | **0.007** | **93.5** | **0.211** | **0.007** | **94.0** | **0.198** | **0.008** | **94.4** | **0.183** | **0.009** | **1.37** |

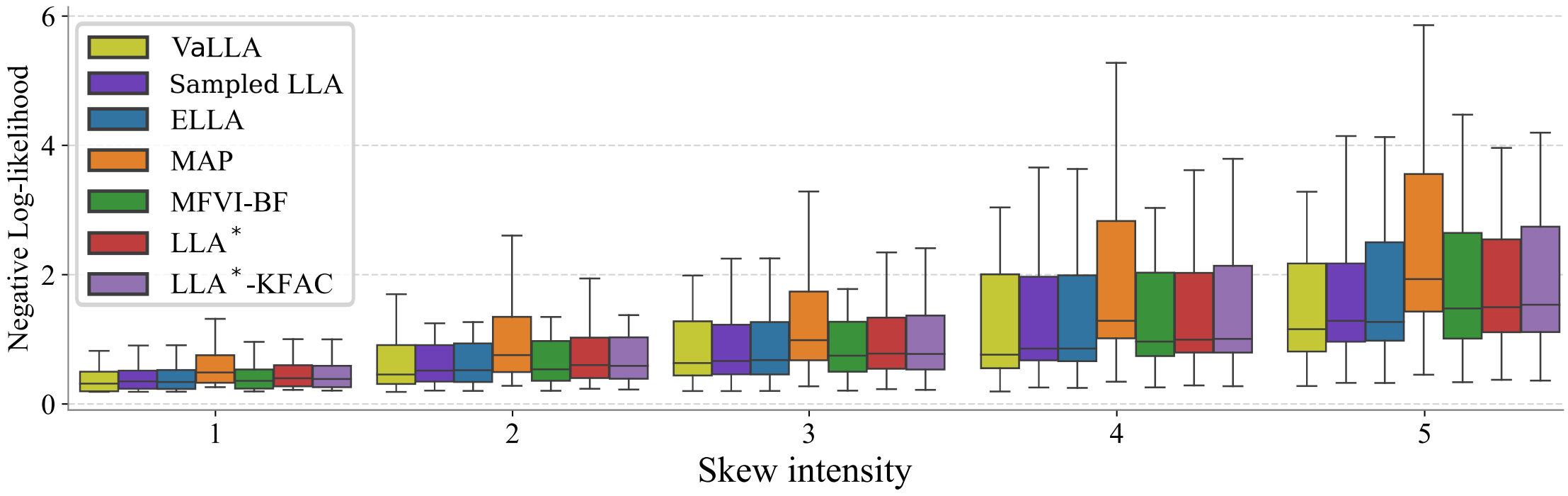

**Figure 4.5:** ***Negative log-likelihood box-plots for corrupted CIFAR-10 evaluated with ResNet-56 at five skew-intensity levels.*** *For each skew intensity $s \in \{1, \ldots, 5\}$ the figure shows the distribution of test NLL across random seeds for seven inference schemes—VaLLA, Sampled LLA (64 Monte-Carlo samples), ELLA ($M$=2000, $K$=20), MAP, MFVI-BF, last-layer Laplace (LLA*) and its Kronecker-factorized variant (LLA* KFAC). Each box denotes the inter-quartile range with the median indicated by a horizontal line; whiskers extend to the minimum and maximum observed values.*

### 4.2.6 Ablation Studies

To better understand the contribution of each design choice in the proposed approach, the next subsection presents a detailed ablation analysis of its main components.

#### 4.2.6.1 Efficient Kernel Computation for MLP

In this section, an efficient implementation for computing the Neural Tangent Kernel $\kappa(\mathbf{x}, \mathbf{x}')$ is reviewed. First of all, consider that the computation of the kernel can be reduced to a summation on the number of parameters of the model:

$$\kappa(\mathbf{x}, \mathbf{x}') = \sigma_0^2 J_{\hat{\boldsymbol{\theta}}}(\mathbf{x})^T J_{\hat{\boldsymbol{\theta}}}(\mathbf{x}') = \sigma_0^2 \sum_{\theta_s \in \hat{\boldsymbol{\theta}}} \frac{\partial}{\partial \theta_s} g(\mathbf{x}, \hat{\boldsymbol{\theta}}) \frac{\partial}{\partial \theta_s} g(\mathbf{x}', \hat{\boldsymbol{\theta}}). \tag{4.34}$$

One of the limitations of computing the kernel is storing $J_{\hat{\boldsymbol{\theta}}}(\mathbf{x})$ in memory, which is a 3-dimensional tensor of (batch size, number of classes, number of parameters). Computing the kernel as a sum allows us to simplify the required computations significantly (no longer have to store in memory the Jacobians). Consider now a MLP as

$$g(\mathbf{x}, \hat{\boldsymbol{\theta}}) = h_L \circ a \circ H_{L-1} \circ \cdots \circ a \circ h_1(\mathbf{x}) \,, \tag{4.35}$$

where each function $a$ is a non-linear activation function and each function $h$ is a linear function of the form

$$h_l(\mathbf{x}) = \boldsymbol{W}_l^T \mathbf{x} + \boldsymbol{b}_l \,. \tag{4.36}$$

With this, $g$ is supposed to be a fully-connected neural network of $L$ layers. Each of the partial derivatives of the neural network is

$$\frac{\partial}{\partial W_{l,j,i}} g(\mathbf{x}, \hat{\boldsymbol{\theta}}) \quad \text{and} \quad \frac{\partial}{\partial b_{l,j}} g(\mathbf{x}, \hat{\boldsymbol{\theta}}) \quad \forall l = 1, \dots, L \,, \tag{4.37}$$

and the kernel is computed simply by adding the product of these derivatives. Here, $i$ is a sub-index denoting input $i$-th to layer $l$. Similarly, $j$ is a sub-index denoting each component of the bias vector parameter at layer $l$, or similarly, each output of that layer.

In fact, using the structure of the model and the chain rule, the derivative of the $o^{th}$ output of the network w.r.t. the $j^{th}, i^{th}$ weight parameter of the $l^{th}$ layer is:

$$\frac{\partial}{\partial W_{l,j,i}} g_o(\mathbf{x}, \hat{\boldsymbol{\theta}}) = \left( \frac{\partial}{\partial h_l} g_o(\mathbf{x}, \hat{\boldsymbol{\theta}}) \right)^T \left( \frac{\partial}{\partial W_{l,j,i}} h_l \right) \,, \tag{4.38}$$

where each of the two vectors in the r.h.s. has length equal to the number of units in the layer $l$. In fact,

$$\frac{\partial}{\partial W_{l,j,i}} h_l = \mathbf{1}_l \cdot a(h_{l-1})_i \,, \tag{4.39}$$

where $a(h_{l-1})_i$ corresponds to the inputs of the $l^{th}$ layer. Moreover, $\frac{\partial}{\partial h_l} g_o(\mathbf{x}, \hat{\boldsymbol{\theta}})$ can also be computed using the chain rule:

$$\frac{\partial}{\partial h_l} g_o(\mathbf{x}, \hat{\boldsymbol{\theta}}) = \frac{\partial}{\partial h_{l+1}} g_o(\mathbf{x}, \hat{\boldsymbol{\theta}}) \frac{\partial}{\partial h_l} h_{l+1} = \frac{\partial}{\partial h_{l+1}} g_o(\mathbf{x}, \hat{\boldsymbol{\theta}}) \boldsymbol{W}_l^T \mathrm{diag}(a'(h_l)) \,, \tag{4.40}$$

which can be computed by back-propagating the derivatives. The same derivations apply to the biases of each layer $b_{l,j}$. As a result, the derivatives depend only on a back-propagating term $\frac{\partial}{\partial h_l} g_o(\mathbf{x}, \hat{\boldsymbol{\theta}})$ for each layer, the parameter values $\boldsymbol{W}_l, \boldsymbol{b}_l$, and the propagated outputs at each layer $h_1, \dots, h_{L-1}$, evaluated at the non-linear activation $a(\cdot)$ and its derivative $a'(\cdot)$. Consequently, if the intermediate outputs of each layer $(h_1, \dots, h_{L-1})$ are stored during the forward pass, a single backward pass suffices to compute $\frac{\partial}{\partial h_l} g_o(\mathbf{x}, \hat{\boldsymbol{\theta}})$ for each layer.

Critically, given each $\frac{\partial}{\partial h_l} g_o(\mathbf{x}, \hat{\boldsymbol{\theta}})$, the contribution of each layer to the kernel can be added using Equation (4.38). In this process, computations can be accelerated by exploiting structure in the derivatives. For example, in Equation (4.39) the derivative has a simple form—a vector of ones scaled by a scalar. Furthermore, there is no dependence on $j$, the output unit corresponding to the weight $W_{l,j,i}$. Therefore, for two instances $\mathbf{x}$ and $\mathbf{x}'$, the kernel contribution (ignoring the prior variance parameter) corresponding to outputs $o$ and $o'$ is:

$$
\begin{aligned}
\frac{\partial}{\partial W_{l,j,i}} g_o(\mathbf{x}, \hat{\boldsymbol{\theta}}) \frac{\partial}{\partial W_{l,j,i}} g'_o(\mathbf{x}', \hat{\boldsymbol{\theta}}) &= \left(\frac{\partial}{\partial h_l} g_o(\mathbf{x}, \hat{\boldsymbol{\theta}})\right)^T \left(\frac{\partial}{\partial W_{l,j,i}} h_l\right) \left(\frac{\partial}{\partial h_l} g_o(\mathbf{x}', \hat{\boldsymbol{\theta}})\right)^T \left(\frac{\partial}{\partial W_{l,j,i}} h_l\right) \\
&= \left(\frac{\partial}{\partial h_l} g_o(\mathbf{x}, \hat{\boldsymbol{\theta}})\right)^T \left(\frac{\partial}{\partial W_{l,j,i}} h_l\right) \left(\frac{\partial}{\partial W_{l,j,i}} h_l\right)^T \left(\frac{\partial}{\partial h_l} g'_o(\mathbf{x}', \hat{\boldsymbol{\theta}})\right) \\
&= \left(\frac{\partial}{\partial h_l} g_o(\mathbf{x}, \hat{\boldsymbol{\theta}})\right)^T \mathbf{1}_l \cdot a(h_{l-1}(\mathbf{x}))_i \cdot a(h_{l-1}(\mathbf{x}'))_i \mathbf{1}_l^T \left(\frac{\partial}{\partial h_l} g'_o(\mathbf{x}', \hat{\boldsymbol{\theta}})\right) \\
&= s^l_{o,\mathbf{x}} a(h_{l-1}(\mathbf{x}))_i \cdot a(h_{l-1}(\mathbf{x}'))_i s^l_{o',\mathbf{x}'} \\
&= s^l_{o,\mathbf{x}} \mathbf{s}^l_{o',\mathbf{x}'} a(h_{l-1}(\mathbf{x}))_i a(h_{l-1}(\mathbf{x}'))_i \,,
\end{aligned}
$$

with $s^l_{o,\mathbf{x}} = \left(\frac{\partial}{\partial h_l} g_o(\mathbf{x}, \hat{\boldsymbol{\theta}})\right)^T \mathbf{1}_l$ a scalar. Similar simplifications occur in the case of *e.g.*, a convolutional layer.

Summing up, by using this method, all the required kernel matrices can be easily and efficiently computed, for a mini-batch of data points and a set of inducing points, with a similar cost as that of letting the mini-batch or the inducing points go through the DNN. A disadvantage is, however, that the described computations will have to be manually coded for each different DNN architecture. This becomes tedious in the case of very big DNNs with complicated layers, as described in Section 4.2.3.2.

#### 4.2.6.2 Over-fitting and Early Stopping

Standard maximization of the ELBO does not permit optimization of the prior variance. In summary, the optimal prior variance becomes infinite because the mean is fixed at the MAP solution. As discussed, this issue is circumvented by applying $\alpha$-divergences, which are well-defined in this learning setup and allow optimization of the prior. However, this objective is not perfect and tends to overfit the prior variance to the training data. The middle column of Figure 4.6 shows the predictive distribution (twice the standard deviation) learned by VaLLA using the black points as training data. The MAP solution is obtained with a two hidden-layer MLP with 50 hidden units and *tanh* activation, optimized to minimize the RMSE of the training data for 10 000 iterations using Adam with learning rate $10^{-3}$. VaLLA, in contrast, is trained for 20 000 iterations. As shown in the figure, the prior variance is overfit to the data, such that the uncertainty does not increase in the central gap of the data.

In this experiment, VaLLA optimizes hyper-parameters jointly with the variational objective. LLA optimizes the prior variance and the likelihood variance by maximizing

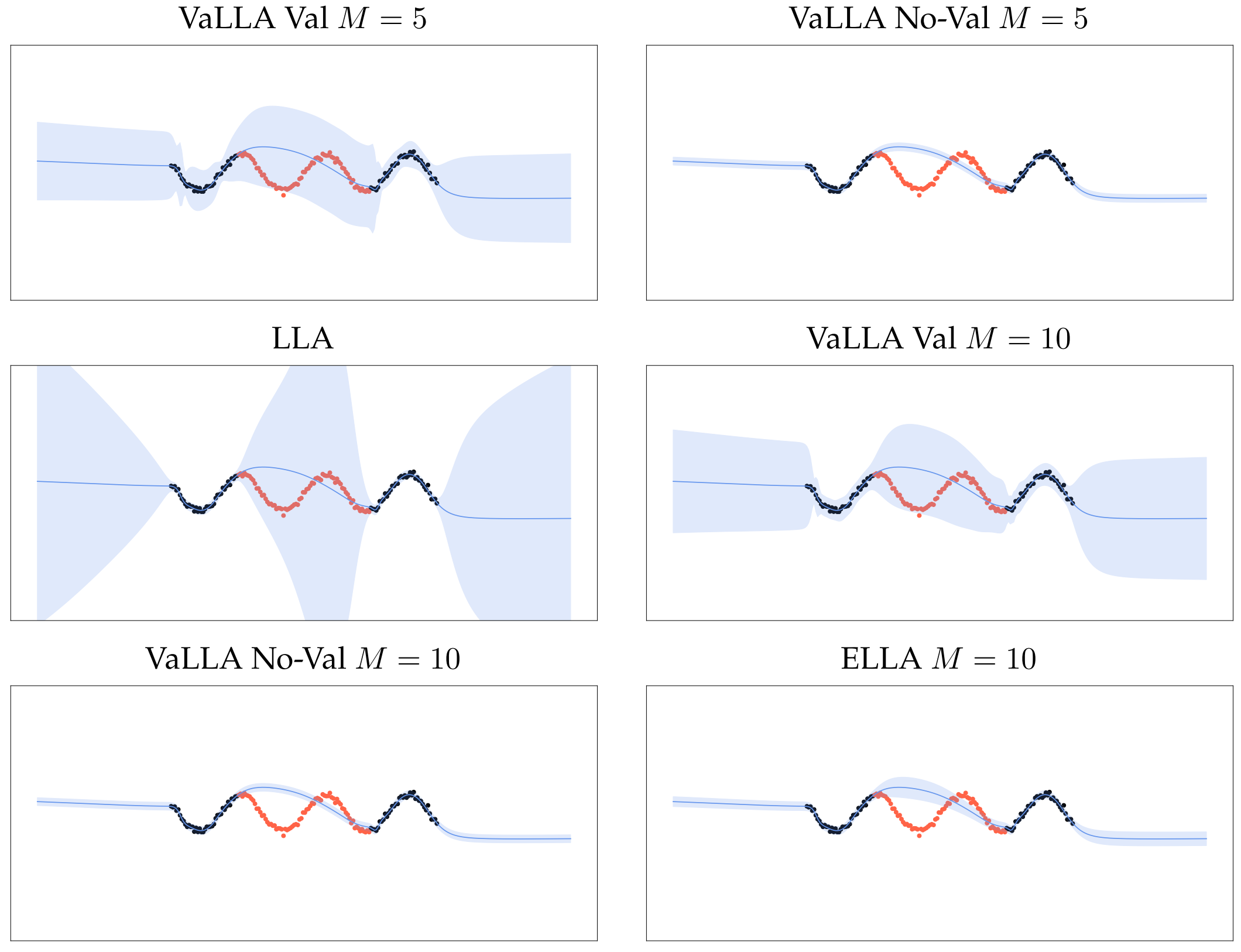

**Figure 4.6:** ***Predictive mean and $\pm 2$ s.d. bands obtained with six Laplace-based setups on a toy one-dimensional regression task.*** *The figure shows posterior predictive functions for a two-layer MLP with 50 hidden units. Top row, left to right: VaLLA with early stopping and $M$=5 inducing points (VaLLA Val $M = 5$); and VaLLA without early stopping ($M$=5 VaLLA No-Val $M = 5$). Middle row: full-network Laplace approximation (LLA) and VaLLA with early stopping and $M$=10 (VaLLA Val $M = 10$). Bottom row: VaLLA without early stopping ($M$=10; VaLLA No-Val $M = 10$); and the accelerated Laplace approximation with $M$=10 (ELLA $M = 10$). In every panel the blue line denotes the posterior mean, the blue shade spans twice the posterior standard deviation, black dots mark training inputs and orange dots indicate validation inputs.*

the marginal log-likelihood, and ELLA uses the optimal hyper-parameters obtained by LLA.

Two simple courses of action are available: (i) return to the original ELBO and select the prior variance via cross-validation; or (ii) perform early stopping with the $\alpha$-divergences objective, using a validation set to halt training before the prior variance overfits. The latter approach may not always succeed, as it assumes the existence of a training point at which the prior variance captures the underlying data without overfitting. However, when the prior variance is set to a relatively large value compared to the small, overfitting-prone optimum, this method yields strong performance for VaLLA while avoiding the cost of cross-validation. The left column of Figure 4.6 shows the predictive distribution (twice the standard deviation) learned by VaLLA in this setting, using the black points as training data and the orange points as the validation set. In the experiments, the validation NLL was computed every 100 training iterations, and training was stopped when this metric worsened, thereby reducing

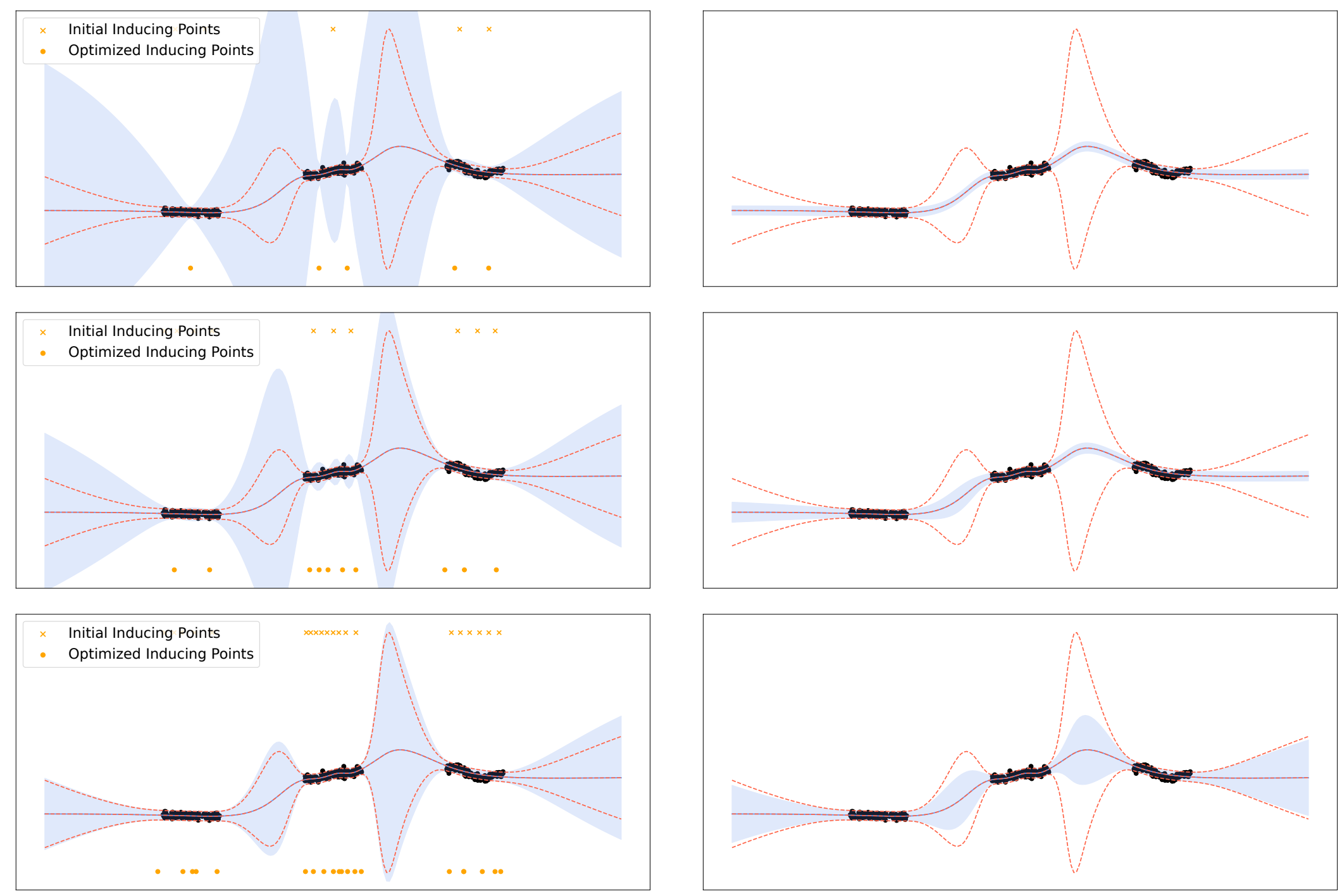

**Figure 4.7:** ***Posterior predictive mean and*** $\pm 2$ ***s.d. envelopes for three inducing-point budgets on a toy one-dimensional regression task.*** *Rows correspond to* $M \in \{5, 10, 20\}$ *inducing points for a two-layer MLP with 50 hidden units. Left column: Variational Laplace approximation (VaLLA) with initial inducing locations (yellow crosses) and optimized locations (yellow dots). Right column: accelerated Laplace approximation (ELLA) using the same values of* $M$*. In every panel the blue curve denotes the predictive mean and the blue shade spans twice the predictive standard deviation; black dots mark training inputs. A dotted orange line provides the predictive mean of the full-network Laplace approximation (LLA) as a common reference.*

computational time.

#### 4.2.6.3 Increasing Inducing Points

Using the optimal covariance in Proposition 4.3, if the set of inducing points equals the training points $\mathbf{Z} = \mathbf{X}$, the posterior distribution of VaLLA coincides with that of the exact LLA Gaussian Process. This observation suggests that increasing the number of inducing points leads to improved uncertainty estimates. This section examines how closely the predictive distribution of VaLLA matches that of LLA as the number of inducing points in $\mathbf{Z}$ increases.

Figure 4.7 shows the obtained predictive distribution of VaLLA (first column) and ELLA (second column) for $M = 5$, $M = 10$ and $M = 20$ inducing points/samples. The initial and final locations of the inducing points are also shown for VaLLA. The posterior distribution obtained by LLA is shown in dotted orange. The MAP solution is obtained using a 2 hidden layer MLP with 50 hidden units and *tanh* activation, optimized to minimize the RMSE of the training data for 12 000 iterations with Adam and learning rate $10^{-3}$. VaLLA, on the other hand, is trained for 30 000 iterations. For this experiment, VaLLA and ELLA use the optimal prior variance and likelihood variance obtained by optimizing LLA's marginal log likelihood. As one may see in the image, it is clear that one of the main differences between the two methods is that VaLLA tends to overestimate the variance, whereas ELLA tends to underestimate it, compared to LLA. Furthermore, the value of $M$ for which the model is closer to the LLA posterior is lower for VaLLA than for ELLA. As $M$ increases, VaLLA's predictive distribution becomes closer and closer to that of LLA.

The initial and final positions of the inducing locations are also shown in the figure. For this experiment, the initial values are computed using K-Means. It can be seen how VaLLA is capable of tuning the inducing locations and moving them from one cluster of points to another as needed. This is one of the main advantages of this method compared to ELLA.

## 4.3 Fixed-Mean Gaussian Processes

A novel family of GPs, *Fixed-Mean Gaussian Processes* (FMGPs), is presented. This family of function-space distributions is defined using the dual formulation of sparse variational GPs. First, consider a subset of the input space $\mathcal{Z} \subset \mathcal{X}$. For any kernel function $K$, the *kernel section* is defined as the closure of all linear combinations of kernel evaluations on $\mathcal{Z}$; that is,

$$K(\mathcal{Z}) := \overline{\left\{ \sum_{i=1}^{n} a_i K(\cdot, \mathbf{z}_i) : n \in \mathbb{N},\ a_i \in \mathbb{R},\ \mathbf{z}_i \in \mathcal{Z} \right\}}. \tag{4.41}$$

**Definition 4.4.** A kernel function is said to be universal if for any compact subset $\mathcal{Z}$ of the input space $\mathcal{X}$, the kernel section $K(\mathcal{Z})$ is dense in $C(\mathcal{Z})$ with the infinity norm. That is, for any $f \in C(\mathcal{Z})$ and any $\epsilon > 0$, there exists $g_\epsilon \in K(\mathcal{Z})$ such that $\|g_\epsilon - f\|_\infty \leq \epsilon$.

Given a universal kernel, building on the dual representation of GPS (Section 2.3.6) the decoupled family of measures $\mathcal{Q}^+$ is employed to show that the mean function of the corresponding GP can be *fixed* to any continuous function on compact subsets.

**Proposition 4.5.** *Let $\mathcal{Z} \subset \mathcal{X}$ be compact and let $K\colon \mathcal{X} \times \mathcal{X} \to \mathbb{R}$ be a* universal *kernel in the sense of Definition 4.4. Then, for every function $f \in C(\mathcal{Z})$ and for every $\varepsilon > 0$ there exist*

$$M_\alpha \in \mathbb{N}, \qquad \{\mathbf{z}_1, \ldots, \mathbf{z}_{M_\alpha}\} \subset \mathcal{Z}, \quad \textit{and} \quad \{a_1, \ldots, a_{M_\alpha}\} \subset \mathbb{R}, \tag{4.42}$$

*such that the finite kernel expansion*

$$m(\mathbf{x}) \;:=\; \langle \phi_{\mathbf{x}}, \tilde{\mu}_{\alpha,\mathbf{a}} \rangle_{\mathcal{H}_K} \;=\; \sum_{m=1}^{M_\alpha} a_m\, K(\mathbf{x}, \mathbf{z}_m), \qquad \mathbf{x} \in \mathcal{Z}, \tag{4.43}$$

*satisfies the uniform approximation bound*

$$\|f - m\|_{\infty,\mathcal{Z}} \;\leq\; \varepsilon. \tag{4.44}$$

[Proof in Appendix C.3]

As a result, for any error rate $\epsilon > 0$, the posterior mean of a decoupled GP can be set to *any* continuous function on any compact subset of the input space. More precisely, the decoupled formulation of GPs can be used to *fix* the posterior mean to the output $g(\cdot)$ of a given pre-trained DNN.

**Definition 4.6.** For any compact subset of the input space $\mathcal{Z} \subset \mathcal{X}$, continuous function $g \in C(\mathcal{Z})$, error $\epsilon > 0$ and universal kernel $K : \mathcal{X} \times \mathcal{X} \to \mathbb{R}$, the set of $g$-mean Gaussian measures $\mathcal{Q}^g_{\mathcal{Z},\epsilon} \subset \mathcal{Q}^+$ is defined as

$$\mathcal{Q}^g_{\mathcal{Z},\epsilon} := \left\{ \mathcal{N}(f|\tilde{\mu}_{\alpha,\boldsymbol{a}}, \tilde{\Sigma}_{\beta,\boldsymbol{A}}) \ : \ \boldsymbol{A} \in \mathbb{R}^{M_\beta \times M_\beta}, \boldsymbol{A} \succeq 0, \mathbf{Z}_\beta \in \mathcal{X}^{M_\beta} \right\}, \tag{4.45}$$

where $\tilde{\mu}_{\alpha,\boldsymbol{a}}$ verifies $\|g(\mathbf{x}) - \langle \phi_{\mathbf{x}}, \tilde{\mu}_{\alpha,\boldsymbol{a}} \rangle_{\mathcal{H}} \|_{\mathcal{Z}} \leq \epsilon$.

Thus, for any $Q(f) \in \mathcal{Q}^g_{\mathcal{Z},\epsilon}$, the corresponding GP $f \sim \mathcal{GP}(m^\star, K^\star_{\mathbf{Z}_\beta,\boldsymbol{A}})$ verifies that

$$\|g(\mathbf{x}) - m^\star(\mathbf{x})\|_{\mathcal{Z}} \leq \epsilon, \tag{4.46}$$

$$K^\star_{\mathbf{Z}_\beta,\boldsymbol{A}}(\mathbf{x},\mathbf{x}') = K(\mathbf{x},\mathbf{x}') - K(\mathbf{x},\mathbf{Z}_\beta)(\boldsymbol{A} + K(\mathbf{Z}_\beta,\mathbf{Z}_\beta))^{-1} K(\mathbf{Z}_\beta,\mathbf{x}'). \tag{4.47}$$

This set of GPs is referred to as *fixed-mean Gaussian processes* (FMGPs). By Proposition 4.5, it is clear that for any $g \in C(\mathcal{Z})$, it is verified that $\mathcal{Q}^g_{\mathcal{Z},\epsilon} \neq \emptyset$ and its corresponding set of FMGPs exists. Again, VI can be used to find the optimal $q$ from this parametric family. That is,

$$\underset{Q \in \mathcal{Q}^g_{\mathcal{Z},\epsilon}}{\arg\min} \ \mathrm{KL}(Q(f)|P(f|\mathbf{y})) = \underset{Q \in \mathcal{Q}^+}{\arg\max} \ \mathbb{E}_{Q(f)}[\log P(\mathbf{y}|f)] - \mathrm{KL}(Q(f)|P(f)), \tag{4.48}$$

where the KL term is, setting $p(f) = \mathcal{N}(f|0, I)$:

$$\mathrm{KL}(Q(f)|P(f)) = \frac{1}{2} \log |\boldsymbol{I} - \boldsymbol{K}_\beta(\boldsymbol{A}^{-1} + \boldsymbol{K}_\beta)^{-1}| - \frac{1}{2} \mathrm{tr}\,(\boldsymbol{K}_\beta \boldsymbol{A}) + \text{constant}. \tag{4.49}$$

Only $\mathbf{A}$ and $\mathbf{Z}_\beta$ require optimization. The method that solves Equation (4.48) is referred to as the *fixed-mean Gaussian process* (FMGP).

Figure 4.8 shows a set representation of the different families of Gaussian measures considered in this section. Observe that $\mathcal{Q}^+$, in which there are different inducing points for the mean and the covariances, is the largest family. This family includes both $\mathcal{Q}$, in which there is only a single set of inducing points for the mean and the covariances, and $\mathcal{Q}^g_{\mathcal{Z},\epsilon}$, in which the posterior mean is fixed to approximate $g$ in $\mathcal{Z}$ with error at most $\epsilon$. Note that there could potentially be some overlap between the Gaussian measures in $\mathcal{Q}$ and in $\mathcal{Q}^g_{\mathcal{Z},\epsilon}$.

### 4.3.1 Application to *Post-hoc* Bayesian Deep Learning

FMGP enables the conversion of DNN into approximate Bayesian models while maintaining the DNN output as the predictive mean. The process is straightforward and

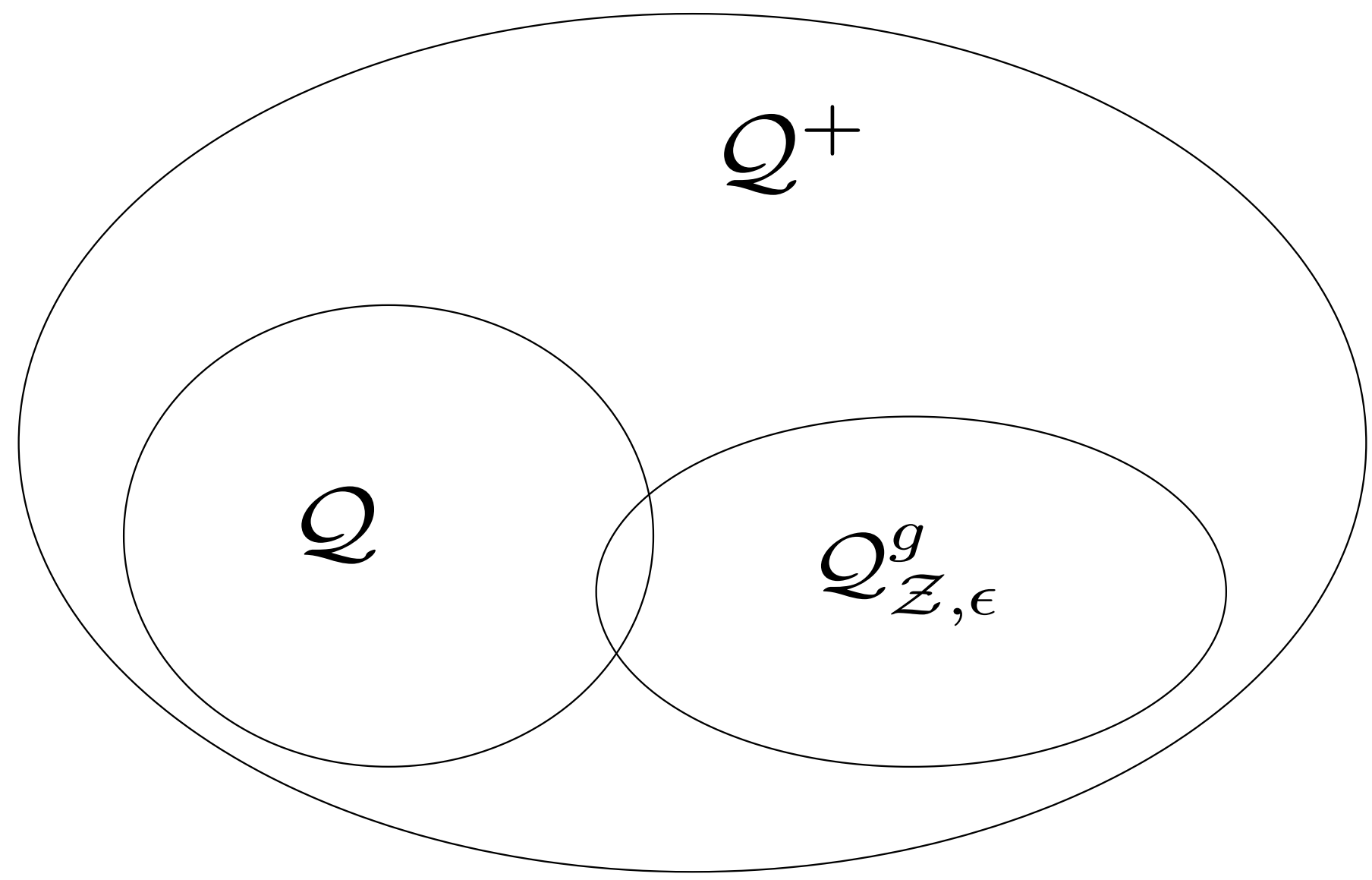

**Figure 4.8:** ***Schematic inclusion relations among three variational Gaussian families for fixed-mean Gaussian processes.*** *The outer ellipse represents the largest family $\mathcal{Q}^+$; the left circular region depicts the baseline mean-field family $\mathcal{Q}$; and the right ellipse shows the structured family $\mathcal{Q}^g_{\mathcal{Z},\varepsilon}$, which partly overlaps with $\mathcal{Q}$ and is fully contained within $\mathcal{Q}^+$.*

can be summarized as follows:

1. Given a pre-trained model $g \in C(\mathcal{X})$, choose a parametric family of kernels that defines an RKHS and a family of Gaussian measures $\mathcal{P}_{\mathcal{H}}$, *e.g.* squared exponential kernels.
2. Ensure that there exists a compact set $\mathcal{Z} \subset \mathcal{X}$ where inputs are expected. The parametric family of Gaussian measures $\mathcal{Q}^g_{\mathcal{Z},\epsilon}$ exists for any $\epsilon > 0$.
3. Initialize a measure in this family, *e.g.* initialize $\mathbf{Z}_\beta$ using K-Means (Lloyd, 1982) and $\boldsymbol{A}$ as the identity matrix.
4. Perform VI to optimize the variational measure ($\boldsymbol{A}$ and $\mathbf{Z}_\beta$), along with the kernel hyper-parameters $\Omega$ and the noise variance $\sigma^2$, using Equation (4.48). The predictive variance is computed as in Equation (4.47), and, if $\mathcal{Z}$ is large and $\epsilon$ small, the predictive mean $m^\star$ approximates the pre-trained model $g$. Thus, in practice, $m^\star$ can be replaced by $g$ in the computations.

### 4.3.2 Regularization and Loss Function

In standard sparse GPs, tuning hyper-parameters entails balancing the fit of the mean to the training data against reducing the model's predictive variance. In FMGPs, however, the predictive mean is fixed, eliminating this trade-off. Consequently, the kernel hyper-parameters $\Omega$ adjust only the predictive variance without affecting the mean. Maximizing the variational-inference ELBO in Equation (4.48) with respect to $\Omega$ can therefore yield undesirable solutions in which the predictive variance collapses to zero. To counteract this, a regularization technique is introduced using an additional variational Gaussian measure. Specifically, an auxiliary Gaussian measure $Q^\star \in \mathcal{Q}$ is con-sidered that shares $Q$'s parameters ($\boldsymbol{A}$, $\mathbf{Z}_\beta$, and the kernel hyper-parameters) while also incorporating $\boldsymbol{a} \in \mathbb{R}^{M_\beta}$ and $\mathbf{Z} = \mathbf{Z}_\beta$ as additional parameters for its predictive mean. This yields the following loss function:

$$\mathcal{L}(\Gamma) = \underbrace{\mathbb{E}_{Q(f)}[\log P(\mathbf{y}|f)] - \mathrm{KL}(Q|P)}_{\mathrm{ELBO}(Q)} + \underbrace{\mathbb{E}_{Q^\star(f)}[\log P(\mathbf{y}|f)] - \mathrm{KL}(Q^\star|P)}_{\mathrm{ELBO}(Q^\star)} . \tag{4.50}$$

with $\Gamma = \{\boldsymbol{a}, \boldsymbol{A}, \mathbf{Z}_\beta, \Omega, \sigma^2\}$. This loss function implies that the predictive variances must account for training data under two predictive means: the pre-trained one, $g(\cdot)$, and the one defined by $\boldsymbol{a}$. Additionally, $\Omega$ cannot be adjusted solely to fit the variances, as it also affects $q^\star$'s predictive mean.

Moreover, the use of $\alpha$-divergences (see Section 2.2.4 for more details) for approximate inference has been widely explored (T. D. Bui et al., 2017; Hernandez-Lobato et al., 2016; Rodríguez-Santana and Hernández-Lobato, 2022; Villacampa-Calvo and Hernández-Lobato, 2020), with findings indicating that values of $\alpha \approx 0$ enhance predictive mean estimation, while $\alpha \approx 1$ improve predictive distributions, reflected in higher test log-likelihood performance. In consequence, instead of minimizing $\mathrm{KL}(Q(f)|P(f|\mathbf{y}))$, our objective is changed using a generalized view of VI (Knoblauch et al., 2022) to minimize the $\alpha$-divergence between $P(f|\mathbf{y})$ and $Q(f)$, in an approximate way, for $\alpha \approx 1.0$ (Y. Li and Gal, 2017). This can be achieved by changing the data-dependent term of the loss (Section 2.2.4):

$$\mathcal{L}(\Gamma) = \log \mathbb{E}_{Q(f)}[P(\mathbf{y}|f)] - \mathrm{KL}(Q|P) + \log \mathbb{E}_{Q^\star(f)}[P(\mathbf{y}|f)] - \mathrm{KL}(Q^\star|P) , \tag{4.51}$$

where now the expectation is inside the logarithm function.

The objective in Equation (4.51) supports mini-batch optimization with a cost in $\mathcal{O}(M_\beta^3 + |\mathcal{B}|M_\beta^2)$:

$$\begin{aligned} \mathcal{L}(\Gamma) \approx & \frac{N}{|\mathcal{B}|} \sum_{b \in \mathcal{B}} \log \mathbb{E}_{Q(f)} \left[P(y_b|f)\right] - \mathrm{KL}\left(Q|P\right) \\ & + \frac{N}{|\mathcal{B}|} \sum_{b \in \mathcal{B}} \log \mathbb{E}_{Q^\star(f)} \left[P(y_b|f)\right] - \mathrm{KL}\left(Q^\star|P\right) , \end{aligned} \tag{4.52}$$

where $\mathcal{B}$ is a mini-batch of points. The expectation can be computed in closed form in regression. In classification, an approximation is available via the softmax method in Daxberger et al., 2021b.

### 4.3.3 Limitations

FMGP is limited by three factors:

1. Computing the predictive distribution at each training iteration involves inverting $\boldsymbol{A}^{-1} + \boldsymbol{K}_{\mathbf{Z}_\beta}$, with cubic cost in the number of inducing points $M_\beta$. Therefore, FMGP cannot accommodate a very large number of inducing points. However, as shown in the experiments, this number can be set to a very low value, such as 20, even for classification tasks with a thousand classes.

2. FMGP requires additional optimization steps compared to other *post-hoc* approximations, *e.g.*, Z. Deng et al., 2022. However, these other methods often rely on visiting *every training point* to compute specific updates. As a result, FMGP training can be faster in large datasets featuring millions of instances such as ImageNet.
3. The construction of FMGP requires choosing a (parametric) kernel. This is both an advantage, as it may better capture the underlying data patterns in the modeling process, and a disadvantage, as it may be difficult to efficiently use an effective kernel in some tasks, such as image classification.

### 4.3.4 Related Work

Due to its *post-hoc* nature, FMGP is highly related to the Linearized Laplace Approximation (LLA) for deep learning. In MacKay, 1992b, the Laplace Approximation (LA) was introduced by applying it to small DNNs. LA simply approximates the DNN posterior in parameter space with a Gaussian centered at its mode matching the posterior Hessian. LA can be made more scalable by considering a Generalized Gauss-Newton (GGN) approximation of the Hessian at the mode, which is equivalent to linearizing the DNN. When this linearization is used at prediction time, LA becomes a *post-hoc* method known as LLA (Ritter et al., 2018). Moreover, LLA addresses the underfitting issues associated with LA (Lawrence, 2001). Despite this, the GGN approximate Hessian of LLA is still intractable in real problems and has to be further approximated using, *e.g.*, Kronecker factored (KFAC) approximations. Recently, many approaches have been developed to make LLA more scalable and accurate, trying to match LLA with the GGN approximate Hessian, including Nyströn approximations (Z. Deng et al., 2022), variational approaches (Luis A. Ortega et al., 2024a; Scannell et al., 2024) or sample-based approximations (Antorán et al., 2023).

Among LLA methods, FMGP is most closely related to Variational LLA (VaLLA) (Luis A. Ortega et al., 2024a). VaLLA interprets the intractable LLA approximation as a GP (Immer et al., 2021; Khan et al., 2019), which is possible due to the DNN linearization. Then, it uses a VI sparse GP to approximate the posterior variances. If the Neural Tangent Kernel (NTK) is considered (which is simply given by the DNN Jacobian, *i.e.* $\phi_{\mathbf{x}} = \mathrm{d}g(\mathbf{x})/\mathrm{d}\boldsymbol{\theta}$), VaLLA can be recovered as a specific case of the FMGP formulation under certain hypotheses. The key differences between FMGP and VaLLA are: (i) FMGP does not rely on the NTK, which allows it to make use of more suitable kernels for specific tasks. Furthermore, the NTK demands computing the Jacobian of the neural network at each iteration, which drastically increases prediction costs, making it impractical to apply VaLLA to large problems (*e.g.* ImageNet). Furthermore, as it will be shown in the experiments, the FMGP kernel flexibility allows both for enhanced predictive distributions as well as more efficiently computed predictions. (ii) FMGP does not rely on a LLA approximation. Therefore, LLA with the GGN Hessian approximation need not be *optimal* under this framework. (iii) The NTK kernel is not guaranteed to be *universal*, and VaLLA relies on the hypothesis that the DNN output $g(\cdot)$ is in $\mathcal{H}$, which may not be the case. (iv) VaLLA does not consider a regularization technique to avoid overfitting, requiring early stopping and a validation set. In short, VaLLA is constructed by performing a variational approximation to the GP resulting from LLA. By contrast, FMGP uses VI to find the optimal measure within a set of Gaussian measures with a fixed mean.

In Z. Deng et al., 2022, the authors propose a Nyström approximation of the GGN Hessian approximation of LLA using $M \ll N$ points chosen at random from the training set. The method, called ELLA, has cost $\mathcal{O}(NM^3)$. ELLA also requires computing the costly Jacobian vectors required in VaLLA, but does not need their gradients. Unlike VaLLA, the Nyström approximation needs to visit each instance in the training set. However, as stated in Z. Deng et al., 2022, ELLA suffers from overfitting. Again, an early-stopping strategy using a validation set is proposed to alleviate it. In this case, ELLA only considers a subset of the training data. ELLA does not allow for hyper-parameter optimization, unlike VaLLA. The prior variance $\sigma_0^2$ must be tuned using grid search and a validation set, increasing the required training time significantly.

Samples from LLA's corresponding GP posterior can be efficiently computed using stochastic optimization, without inverting the kernel matrix (Antorán et al., 2023; Lin et al., 2024). This approach avoids LLA's $\mathcal{O}(N^3)$ cost. However, this method does not provide an estimate of the log-marginal likelihood for hyper-parameter optimization. Thus, in Antorán et al., 2023 it is proposed to use the *EM-algorithm* for this task, where samples are generated (E-step) and hyper-parameters are optimized afterwards (M-step) iteratively. This significantly increases training cost, as generating a single sample is as expensive as fitting the original DNN on the full data. A limitation of this approach is that in Antorán et al., 2023 only classification problems are considered and there is empirical evidence showing that VaLLA is faster and gives better results.

Another GP-based approach for obtaining predictive uncertainty in DNNs is the Spectral-normalized Neural Gaussian Process (SNGP) (J. Z. Liu et al., 2023), which replaces the last layer of the DNN with a GP. SNGP allows either (i) fine-tuning a pre-trained DNN or (ii) training a full DNN from scratch. The comparison in the experiments focuses on the former. In practice, replacing the last layer with a GP does not preserve the predictive mean of the pre-trained DNN and often leads to a drop in predictive performance; similar behavior was also reported by J. Z. Liu et al. (2023).

Another simple option to transform a pre-trained DNN model to a Bayesian one is to consider a mean-field VI approximation of the DNN posterior where the means are initialized to the pre-trained optimal solution weights and kept fixed. This is known as *mean-field VI fine-tuning* (Z. Deng and Zhu, 2023) and, as demonstrated in the conducted experiments, it can achieve good results in terms of both prediction performance and uncertainty estimation. However, this method demands full training of the variance of each weight, which can be very costly and may require several training epochs. Furthermore, this method provides no closed-form predictive distribution. It relies on generating Monte Carlo samples to make predictions. As a result, further approximations must be considered to reduce the training time, such as *Flipout Trick* (Wen et al., 2018). Even though these techniques successfully reduce the training time, the required Monte Carlo samples significantly increase prediction time.

### 4.3.5 Experiments

The proposed method, FMGP, is compared with other methods including: last-layer LLA with and without KFAC approximation, ELLA (Z. Deng et al., 2022), VaLLA (Luis A. Ortega et al., 2024a), a mean-field VI fine-tuning approach (Z. Deng and Zhu, 2023) and SNGP (J. Z. Liu et al., 2023). FMGP and VaLLA use $100$ inducing points, as in Luis A. Ortega et al., 2024a. ELLA employs $2\,000$ random points and $20$ random features as in Z. Deng and Zhu, 2023. All the timed experiments are executed on an Nvidia A100 graphics card. Finally, an implementation of FMGP is publicly available at https://github.com/Ludvins/FixedMeanGaussianProcesses.

#### 4.3.5.1 Synthetic Experiment

The experiment in Figure 4.9 illustrates the predictive distributions of commonly used Bayesian approaches on the synthetic 1-dimensional dataset from Izmailov et al., 2020. It compares the predictive distribution of FMGPs against other methods, including the pre-trained DNN with optimized Gaussian noise (MAP), the linearized Laplace approximation (LLA) with prior precision and Gaussian noise optimized to maximize the marginal likelihood estimate, mean-field VI (MFVI) fine-tuning of the pre-trained model with Gaussian noise optimized on the training data, a GP with a squared exponential kernel and hyper-parameters that maximize the marginal likelihood, and Hamiltonian Monte Carlo (HMC) using a uniform prior for the variance of both the Gaussian noise and the Gaussian prior over the DNN's weights $\boldsymbol{\theta}$.

In this simple problem, HMC's predictions serve as the *gold standard* for assessing the predictive variances of other methods. Note, however, that HMC does not scale to large problems. Figure 4.9 shows that MAP and MFVI tend to underestimate the predictive variance, while LLA tends to overestimate it by interpolating between data clusters. On the other hand, FMGP and the GP produce predictive variances comparable to those of HMC, with the GP yielding slightly larger variances.

However, the GP's predictive mean does not align with the DNN output (given by the predictive mean of the MAP output) and suffers from a prior mean reversion problem, where the GP mean reverts to the prior mean between the second and third point clusters, which is expected to worsen the resulting predictive performance. Moreover, the GP does not scale well to large problems. By contrast, FMGP not only produces predictive variances similar to those of HMC but also retains the predictive mean equal to the DNN's output, which is expected to result in improved prediction accuracy.

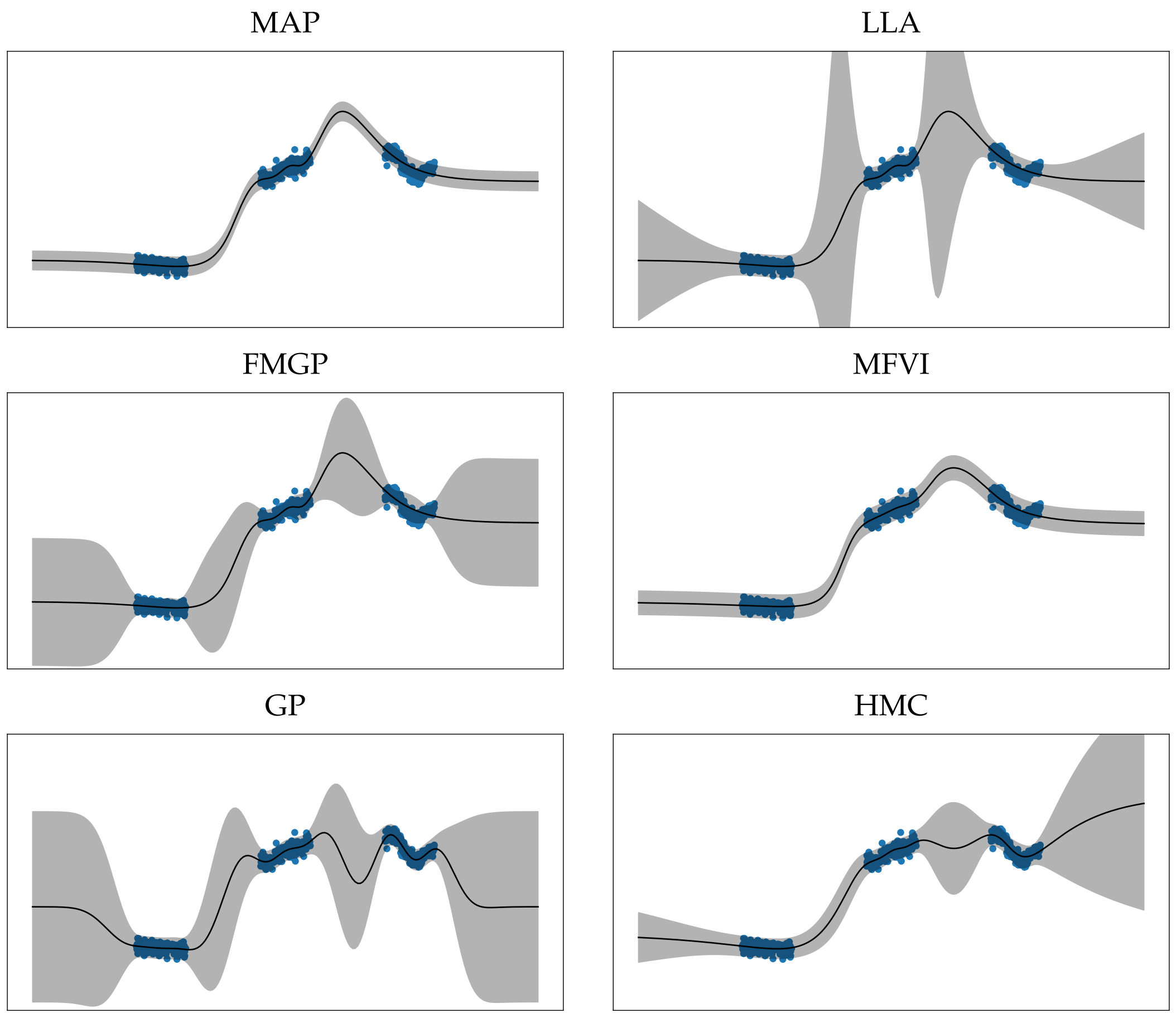

**Figure 4.9:** ***Predictive distribution (mean in black, $2\sigma$ shaded region) on a toy 1D regression dataset.*** *The considered approaches include a 2 hidden layer MLP with 50 units trained using back-propagation (MAP), linearized Laplace approximation (LLA), fixed-mean Gaussian process (FMGP) with squared exponential kernel, mean-field variational inference (MFVI) for DNN fine-tuning, Gaussian process (GP) with squared exponential kernel, and Hamilton Monte Carlo (HMC). All methods' hyper-parameters are optimized using training data except HMC, which uses uniform hyper-priors.*

#### 4.3.5.2 Regression Problems

As part of the experimental evaluation, consider three different large regression datasets:

1. The *Year* dataset (Bertin-Mahieux, 2011) with 515 345 instances and 90 features. The data is divided as: the first 400 000 instances as train subset and the following 63 715 for validation. The rest of instances are taken for the test set.
2. The *US flight delay* (*Airline*) dataset (Dutordoir et al., 2020). Following Luis A. Ortega et al., 2024a, use the first 600 000 instances for training, the following 100 000 instances for validation and the next 100 000 for testing. Here, 8 features are considered: *month*, *day of the month*, *day of the week*, *plane age*, *air time*, *distance*, *arrival time* and *departure time*.
3. The *Taxi dataset*, with data recorded on January, 2023 (Luis A. Ortega et al., 2023). For this dataset, 9 attributes are considered: *time of day*, *day of week*, *day of month*, *month*, *PULocationID*, *DOLocationID*, *distance* and *duration*; while the predictive variable is the *price*. Following Luis A. Ortega et al., 2024a, filter trips shorter than 10 seconds and larger than 5 hours, resulting in 3 050 311 million instances. The first 80% is used as train data, the next 10% as validation data, and the last 10% as testing data.

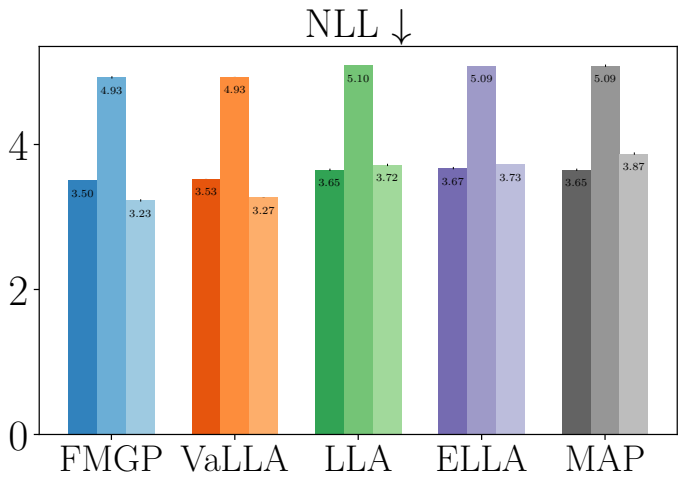

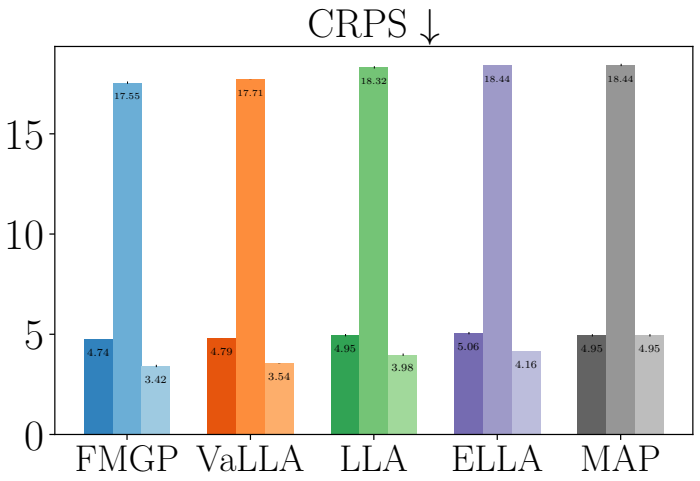

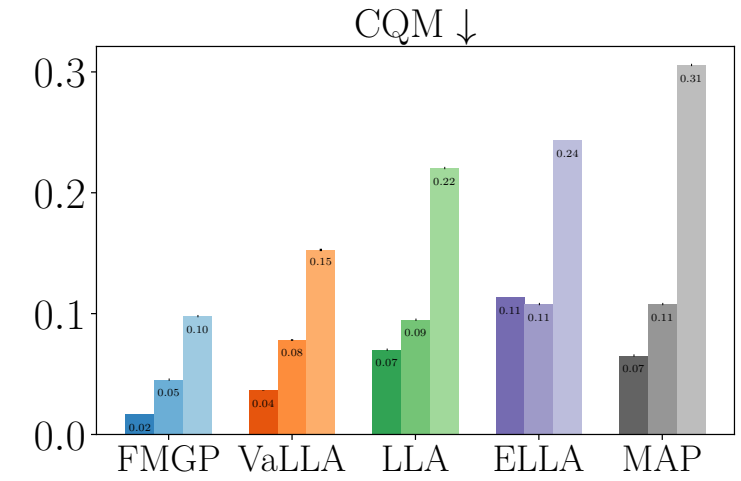

**Figure 4.10:** ***Triple-bar plots of NLL, CRPS and CQM for five post-hoc uncertainty methods on three tabular regression datasets.*** *The left, center and right panels display test negative log-likelihood (NLL↓), continuous ranked probability score (CRPS↓) and calibrated quantile metric (CQM↓), respectively. Within each panel, the color groups correspond to FMGP, VaLLA, LLA, ELLA and MAP; each group contains three bars that represent the Year, Airline and Taxi datasets (ordered left to right). Bar height indicates the mean across five independent repetitions, and thin vertical lines denote the associated standard errors, which are visually negligible in most cases.*

In all experiments, a pre-trained 3-layer DNN with 200 units with *tanh* activations is employed, following Luis A. Ortega et al., 2024a. ELLA is trained without *early-stopping* as overfitting is not observed in these regression problems. Hyper-parameters are chosen using a grid search and the validation set. FMGP employs the squared-exponential kernel with hyper-parameters given by kernel amplitude and one length scale per input feature. MAP results are obtained by learning the optimal Gaussian noise using a validation set. A last-layer Kronecker approximation is used for LLA.

Figure 4.10 shows average results for each method over 5 different random seeds. The quality of the predictive distribution is measured in terms of the negative log likelihood (NLL), the continuous ranked probability score (CRPS) (Gneiting and Raftery, 2007) and a centered quantile metric (CQM) (Luis A. Ortega et al., 2024a). Intuitively, CRPS can be understood as a generalization of the mean absolute error to predictive distributions. As seen in Section 4.2.5, CQM measures the *difference* between the model's quantiles and the data quantiles under the same predictive mean, which is always the case here for each method. CQM is like a generalization of expected calibration error for regression problems. It is defined as:

$$\text{CQM} = \int_0^1 \left| \mathbb{P}_{(\mathbf{x}^\star, y^\star)} \left[ y^\star \in I(\mathbf{x}^\star, \alpha) \right] - \alpha \right| d\alpha \,, \tag{4.53}$$

where $I(\mathbf{x}, \alpha) = (\lambda(-\alpha), \lambda(\alpha))$, $\lambda(\alpha) = \Phi^{-1}_{\mu(\mathbf{x}),\sigma^2(\mathbf{x})}(\frac{1+\alpha}{2})$ and $\Phi_{\mu(\mathbf{x}),\sigma^2(\mathbf{x})}$ is the CDF of a Gaussian with mean $\mu(\mathbf{x})$ and variance $\sigma^2(\mathbf{x})$, specified by each model's predictive distribution.

Figure 4.10 shows that FMGP performs best according to all three metrics (the lower the better), where the biggest difference is obtained in terms of CQM. As a result, we can argue that FMGP provides better uncertainty estimates (in terms of NLL) and cal-ibration (both in terms of CRPS and CQM) compared to state-of-the-art LLA variants in regression settings.

#### 4.3.5.3 CIFAR10 Dataset and ResNet Architectures

Several experiments with various ResNet architectures (He et al., 2016a) on the CIFAR10 dataset (Krizhevsky, Hinton, et al., 2009) are performed. To facilitate reproducibility, the considered pre-trained models are publicly available and accessible at https://github.com/chenyaofo/pytorch-cifar-models. The considered models are the following: ResNet20 ($272\,474$ parameters), ResNet32 ($466\,906$ parameters), ResNet44 ($661\,338$ parameters) and ResNet56 ($855\,770$ parameters). Following Z. Deng et al., 2022 and Luis A. Ortega et al., 2024a, ELLA and VaLLA use as a validation set a data-augmented subset of $5\,000$ training points from the train set. This validation set is obtained by performing random image crops of the training images of sizes in $[0.5, 1]$.

In multi-class classification problems, the kernel used in FMGP should model dependencies among the different DNN outputs, one for each class label. Therefore, the following simple kernel in FMGP is employed in that setting:

$$K((\mathbf{x}, c), (\mathbf{x}', c')) = B_{c,c'} \times K_{RBF}(\mathbf{x}, \mathbf{x}') \times (\psi(\mathbf{x})^T \psi(\mathbf{x}') + \delta_{\mathbf{x}=\mathbf{x}'})\,,$$

which includes a p.s. matrix $\boldsymbol{B} \in \mathbb{R}^{C\times C}$ to model output dependencies, a squared exponential kernel in the input space, and a linear kernel plus noise in the high-level features $\psi(\cdot)$, that correspond to the output of the pre-trained model up to the second-to-last layer. The trainable hyper-parameters are the squared exponential amplitude and length scales of the RBF kernel (one per input feature), along with the matrix $\boldsymbol{B}$, parameterized by its Cholesky decomposition. This simple kernel gives good results in the conducted experiments. More sophisticated kernels are possible, potentially leading to even better results. The inducing points are randomly assigned to a class label.

Figure 4.11 shows the negative log-likelihood (NLL), expected calibration error (ECE), and Brier score of each method. Furthermore, the out-of-distribution AUC of each method is also reported in a binary classification problem with the SVHN dataset as the out-of-distribution data (Netzer et al., 2011). In each method, the predictive entropy is used as the threshold for classification between in and out-of-distribution. The training and evaluation times for each method are also reported. Recent work Mucsányi et al., 2024 shows how different uncertainty quantification metrics tend to *cluster* and the importance of measuring prediction uncertainty using as many as possible. Accuracy is not shown here as most methods barely change the pre-trained DNN accuracy. Notwithstanding, it is worth mentioning that SNGP tends to lower the accuracy of the model, as shown in J. Z. Liu et al., 2023, while MFVI tends to increase it slightly, as noticed in Z. Deng et al., 2022 and Luis A. Ortega et al., 2024a.

Figure 4.11 shows that FMGP, MFVI, VaLLA, and ELLA provide the highest performance in terms of NLL and Brier scores (the lower the better). However, in terms of ECE (also the lower the better), SNGP, VaLLA, and FMGP provide better-calibrated uncertainties. As a result, FMGP and VaLLA seem to provide better uncertainty quantification with better-calibrated predictive distributions. However, for out-of-distribution detection, the best AUC is obtained by MFVI, ELLA, VaLLA, and SNGP. Figure 4.12 shows histograms of the entropy of the predictive distribution of each method for each type of test data (in and out-of-distribution). The poor results of FMGP in this task may be due to the kernel choice. More sophisticated kernels may improve FMGP's results in this setting as well.

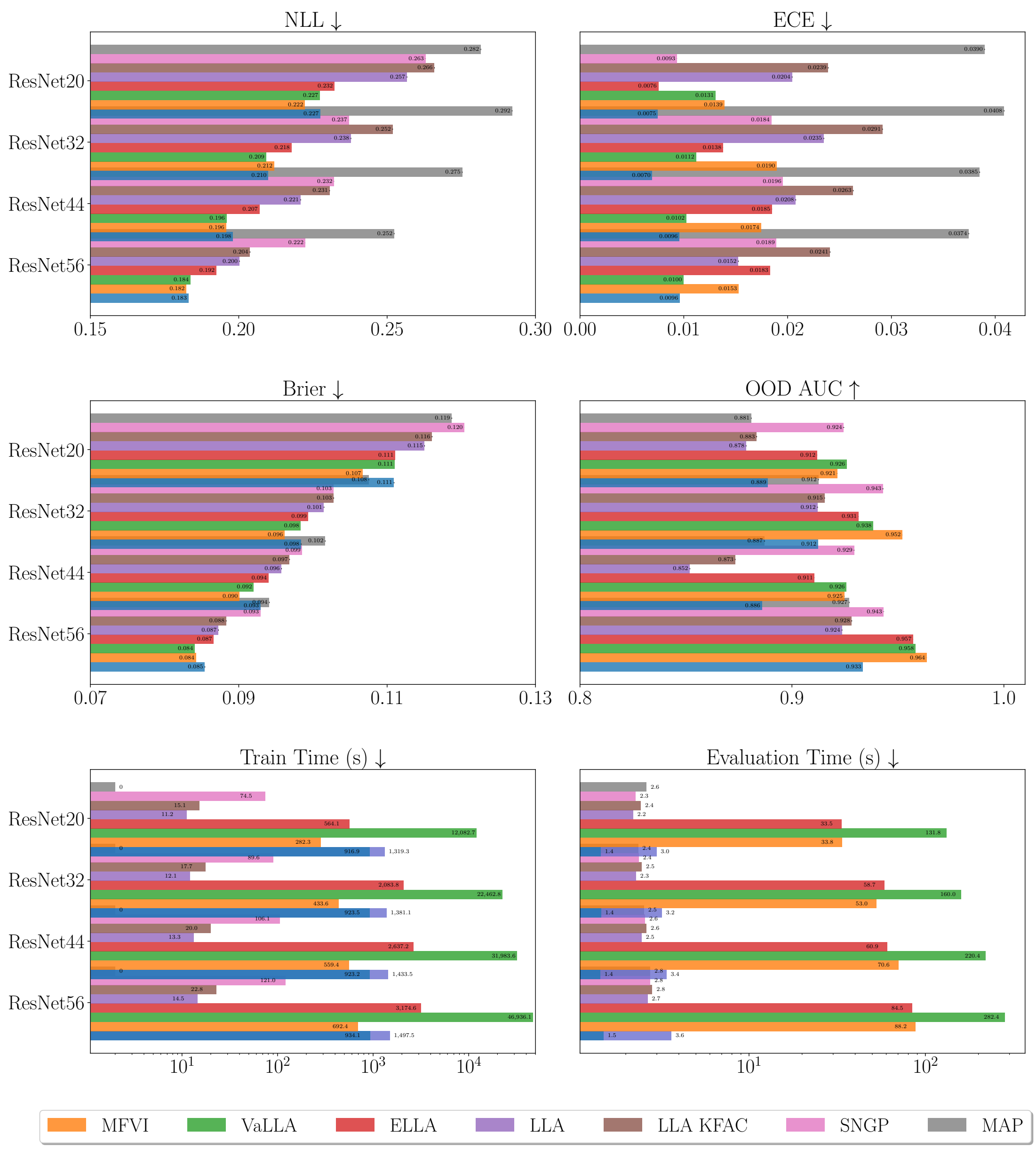

**Figure 4.11:** ***Test-set predictive metrics, out-of-distribution detection performance and computational cost for eight inference schemes across four ResNet architectures on CIFAR-10.*** *The figure contains six horizontal bar plots: (top row, left to right) negative log-likelihood (NLL) and expected calibration error (ECE), and (middle row) Brier score, and area under the ROC curve for out-of-distribution detection between CIFAR-10 and SVHN (OOD AUC), and (bottom row) wall-clock training time and per-example evaluation time (both on logarithmic scales). Within each panel, bars are grouped by architecture—ResNet-20, -32, -44 and -56—and colored by inference method: FMGP, MFVI, VaLLA, ELLA, last-layer Laplace (LLA), last-layer Kronecker-factorized Laplace (LLA KFAC), SNGP and a maximum-a-posteriori baseline (MAP). Heights correspond to the mean of five random-seed repetitions, with thin error bars indicating the associated standard errors (visually negligible in most cases).*

Regarding training time, Figure 4.11 shows that last-layer LLA approaches are the fastest to train, with VaLLA being the slowest method. At prediction time, SNGP, last-layer LLA, and FMGP are quite similar to the pre-trained model. By contrast, VaLLA, ELLA, and MFVI take larger prediction times. In VaLLA and ELLA, this is due to the computation of the Jacobians, while in MFVI this is due to Monte-Carlo sampling. Since FMGP is agnostic of the pre-trained model architecture, it only uses the DNN's predictions. Therefore, all the model outputs are precomputed and used directly when training FMGP and making predictions using this model. As a result, a second bar is shown for FMGP indicating the training and evaluation time when pre-computing the outputs for both training and evaluation sets. In such a setting, the speed-up of FMGP is approximately $\times 1.5$ for training time and $\times 2.2$ for evaluation time.

Regarding predictive robustness, in Figure 4.13 the NLL and ECE of each method on rotated images of the CIFAR10 test set are shown, as in Luis A. Ortega et al., 2024a. These results indicate that FMGP is the most robust method in terms of NLL, while it lies around the middle ground in terms of ECE. ELLA and VaLLA achieve the best results in this regard.

#### 4.3.5.4 ImageNet Dataset and Extra ResNet Architectures

Several experiments with more ResNet architectures (He et al., 2016a) on the ImageNet 1k dataset (Russakovsky et al., 2015) are performed. This dataset has $1\,000$ different classes and over 1 million data instances. As pre-trained models, those from TorchVision (Maintainers and Contributors, 2016) available at https://pytorch.org/vision/main/models/resnet.html are considered. Specifically, the considered models are ResNet18 ($11\,689\,512$ parameters), ResNet34 ($21\,797\,672$ parameters), ResNet50 ($25\,557\,032$ parameters), ResNet101 ($44\,549\,160$ parameters) and ResNet152 ($60\,192\,808$ parameters). Importantly, due to the size of the DNNs and dataset, many methods became infeasible in these experiments. Specifically, LLA cannot be used even with last-layer approximations due to memory limitations. Furthermore, Monte Carlo sampling for MFVI testing takes longer than 1 day for models larger than ResNet18. For this reason, MFVI is only tested on the ResNet18 architecture. SNGP is not evaluated as it requires a training time of several days on the smallest architecture.

Table 4.4 shows the results obtained for each method on each ResNet architecture. The best method is highlighted in red, and the second-to-best method is highlighted in green. Overall, FMGP obtains the best performance (NLL and ECE) while remaining the second-to-best in terms of computational time, only behind the MAP solution for the bigger models. As an additional detail, ELLA's validation set is computed using the same data-augmentation strategy proposed in Z. Deng et al., 2022.

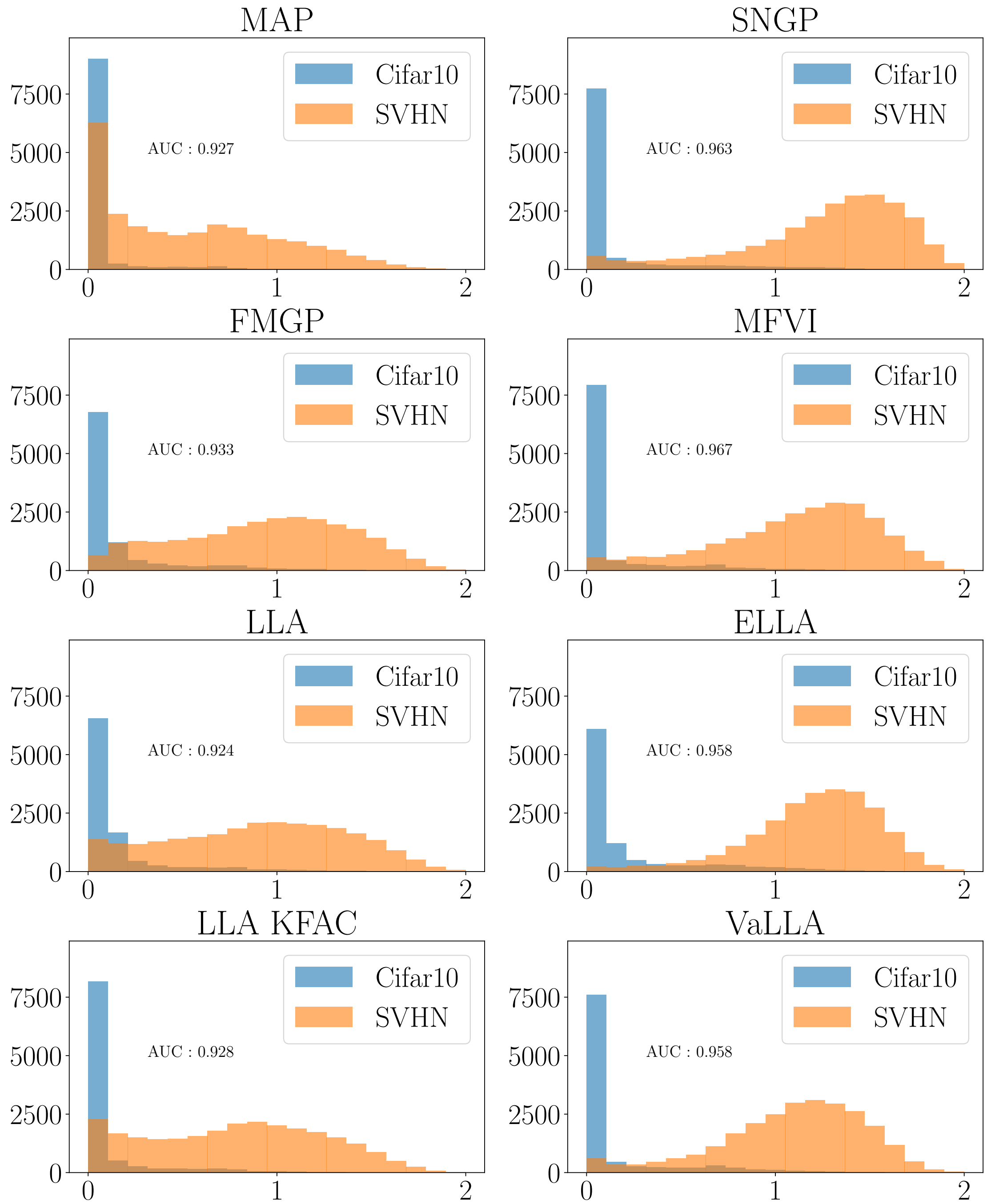

**Figure 4.12:** ***Histograms of predictive-entropy values for CIFAR-10 (blue) and SVHN (orange) inputs under eight inference schemes.*** *Each panel corresponds to one method—MAP, FMGP, last-layer Laplace (LLA), last-layer Kronecker-factorized Laplace (LLA KFAC), SNGP, mean-field variational inference (MFVI), empirical Laplace (ELLA) and variational Laplace (VaLLA)—applied to a ResNet-56 classifier. Bar heights count test examples, aggregated over five independent repetitions. The horizontal axis denotes entropy (in nats) and the vertical axis the number of samples; the number printed within each subplot is the mean area under the ROC curve computed from the displayed histograms.*

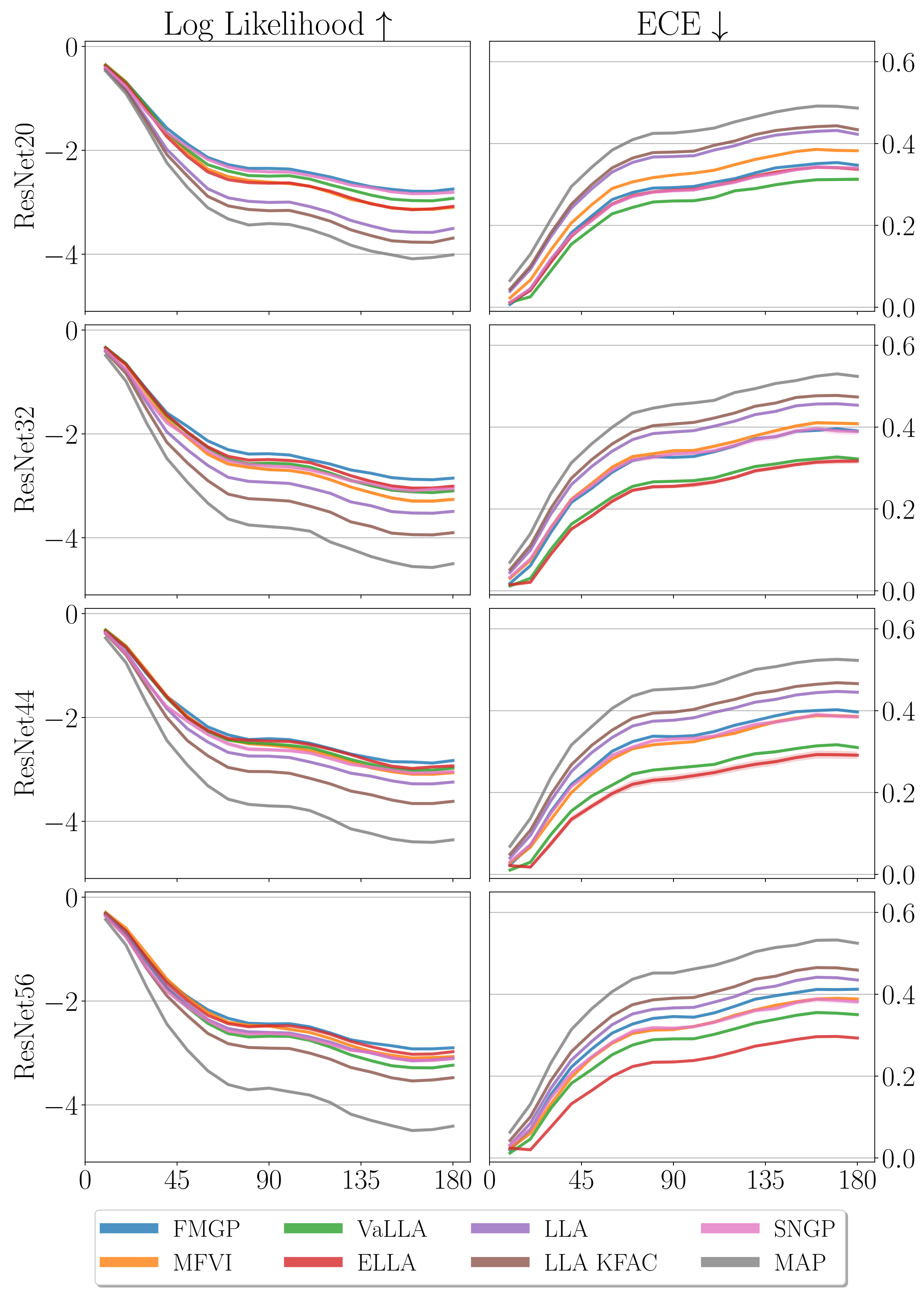

**Figure 4.13:** ***Log-likelihood and expected calibration error as functions of image-rotation angle for four ResNet architectures on CIFAR-10.*** *Columns correspond to ResNet-20, -32, -44 and -56; the top row plots test log-likelihood and the bottom row plots expected calibration error (ECE) for rotation angles from 0° to 180°. Coloured curves identify the eight inference schemes indicated in the legend—FMGP, MFVI, VaLLA, ELLA, last-layer Laplace (LLA), last-layer Kronecker-factorized Laplace (LLA KFAC), SNGP and a maximum-a-posteriori baseline (MAP). All curves are obtained from repeated runs with different random seeds.*

**Table 4.4:** ***Predictive performance, calibration and computational cost for four inference methods applied to five ImageNet-pretrained ResNet architectures.*** *For each backbone (ResNet18, 34, 50, 101 and 152) the table reports the test negative log-likelihood (NLL↓), expected calibration error (ECE↓), total wall-clock training time (seconds) and per-example evaluation time (seconds), expressed as mean ± standard error over five random seeds. Methods include a deterministic maximum-a-posteriori baseline (MAP), ELLA, with $M = 2\,000$ candidate points and $K = 20$ random features, FMGP and mean-field variational inference (MFVI). Zero training times indicate that no additional optimization beyond the pretrained weights was required. Best value is highlighted in purple and second to best in teal.*

| Model | Method | NLL | ECE | Train Time | Test Time |
|---|---|---|---|---|---|
| ResNet18 | MAP | **1.247±0.000** | 0.026±0.000 | **0.0±0.0** | **5.058±0.029**$\times 10^2$ |
| | ELLA | 1.248±0.000 | **0.025**±0.000 | **7.890±0.275**$\times 10^3$ | 8.060±0.010$\times 10^2$ |
| | FMGP | 1.248±0.001 | **0.015±0.001** | 1.835±0.099$\times 10^4$ | **7.324±0.001**$\times 10^2$ |
| | MFVI | **1.242±0.001** | 0.040±0.000 | 7.602±0.032$\times 10^4$ | 3.773±0.308$\times 10^4$ |
| ResNet34 | MAP | **1.081±0.000** | 0.035±0.000 | **0.0±0.0** | **5.088±0.004**$\times 10^2$ |
| | ELLA | 1.082±0.000 | **0.034±0.000** | **1.201±0.373**$\times 10^4$ | 1.087±0.018$\times 10^3$ |
| | FMGP | **1.077±0.000** | **0.016±0.000** | 1.942±0.103$\times 10^4$ | **8.563± 0.011**$\times 10^2$ |
| ResNet50 | MAP | **0.962±0.000** | 0.037±0.000 | **0.0±0.0** | **4.954±0.010**$\times 10^2$ |
| | ELLA | **0.962±0.000** | **0.036±0.000** | 2.997±1.215$\times 10^4$ | 1.954±0.018$\times 10^3$ |
| | FMGP | **0.958±0.001** | **0.018±0.001** | **2.543±0.046**$\times 10^4$ | **1.100±0.010**$\times 10^3$ |
| ResNet101 | MAP | **0.912±0.000** | 0.049±0.000 | **0.0±0.0** | **5.059±0.001**$\times 10^2$ |
| | ELLA | 0.913±0.000 | **0.048±0.000** | 4.464±1.649$\times 10^4$ | 2.808±0.001$\times 10^3$ |
| | FMGP | **0.900±0.000** | **0.030±0.001** | **2.654±0.064**$\times 10^4$ | **1.134±0.001**$\times 10^3$ |
| ResNet152 | MAP | **0.876±0.000** | 0.050±0.000 | **0.0±0.0** | **6.324±0.004**$\times 10^2$ |
| | ELLA | 0.877±0.000 | **0.048±0.000** | 6.820±0.526$\times 10^4$ | 3.877±0.007$\times 10^3$ |
| | FMGP | **0.865±0.001** | **0.024±0.001** | **2.973±0.069**$\times 10^4$ | **1.267±0.002**$\times 10^3$ |

The obtained results for ELLA are slightly different from those reported in Z. Deng et al., 2022 since ELLA's performance highly depends on the particular data augmentation performed to create the validation set. Despite using the same hyper-parameters for this step, using the current PyTorch versions leads to different results.

#### 4.3.5.5 Protein Feature Prediction Dataset

QM9 is a dataset that provides quantum chemical properties (at the DFT level) for a relevant, consistent, and comprehensive chemical space of around $130\,000$ small organic molecules (Ruddigkeit et al., 2012). In this experiment, a small convolutional neural network with message passing is trained following the Torch-Geometric (Fey and Lenssen, 2019) tutorial available at https://github.com/pyg-team/pytorch_geometric/blob/master/examples/qm9_nn_conv.py. The model is trained to predict the *dipole moment* target.

In this regression setting, the input space consists of molecular graphs rather than the tabular data typically used in supervised learning. Accordingly, for each evaluated method, the network up to the last two linear layers is treated as a feature embedding of the graphs, and the data are assumed to live in that embedding space. The first 10 000 instances are used for testing, the next 10 000 for validation, and the remaining 110 000 for training. ELLA is trained without *early stopping*, and hyper-parameters are selected via grid search on the validation set. FMGP employs the squared-exponential kernel with hyper-parameters comprising the amplitude and one length scale per dimension. MAP results are obtained by estimating the Gaussian noise on the validation set.

**Table 4.5:** ***Negative log-likelihood and continuous ranked probability score for the QM9 dipole-moment prediction task.*** *The table reports the mean ± standard error of test NLL (lower is better) and CRPS across multiple random seeds for four inference schemes: a maximum-a-posteriori baseline (MAP), the last-layer Laplace approximation (LLA), ELLA and FMGP. Best value is highlighted in purple and second to best in teal.*

| Method | NLL | CRPS |
|---|---|---|
| MAP | -1.76±0.016 | 0.0221±0.00 |
| LLA | -1.78±0.021 | **0.0218±0.00** |
| ELLA | **-1.80±0.013** | 0.0219±0.00 |
| FMGP | **-1.85**±**0.017** | **0.0216**±**0.00** |

The results are shown in Table 4.5 for MAP, last-layer Kronecker LLA, ELLA, and FMGP in terms of negative log-likelihood (NLL) and CRPS. Average results across 5 repetitions are reported. The best result is highlighted in red and the second-best in green. FMGP achieves the best performance (lower is better) in both NLL and CRPS among the considered methods.

## 4.4 Conclusions

In this chapter, two complementary *post-hoc* uncertainty estimation methods were introduced: the **Variational Linearized Laplace Approximation (VaLLA)** and the **Fixed-Mean Gaussian Process (FMGP)**. Both approaches stem from a generalized sparse GP formulation that enables fixing the predictive mean to any desired function within the RKHS, typically set to the output of a pre-trained DNN. This design allows them to retrofit calibrated predictive distributions onto existing networks without retraining.

VaLLA offers robust uncertainty estimates with computational costs independent of the number of training instances, efficiently scaling to deep models with millions of parameters and large datasets. It outperforms existing sampling-based and Nyström-approximation methods in speed and stability, while maintaining predictive robustness under input perturbations. FMGP extends this idea through a dual variational formulation, providing a simple yet effective framework that relies solely on DNN outputs—without requiring Jacobians or architectural details—thereby achieving minimal evaluation time and broad applicability across regression and classification tasks.

Both methods demonstrate that accurate, scalable uncertainty quantification can be achieved even for large deterministic networks, bridging the gap between practical deep learning and Bayesian modeling. Moreover, VaLLA's and FMGP's efficiency and reliability suggest promising applications in **Bayesian optimization**, where fast and accurate uncertainty estimation can significantly improve the search for optimal solutions.

# 5 Generalization in Neural Networks

The previous chapter focused on methods for *post-hoc uncertainty estimation* in pretrained deep networks, introducing the Variational Linearized Laplace Approximation (VaLLA) and Fixed-Mean Gaussian Processes (FMGPs). These approaches demonstrated how probabilistic reasoning can be incorporated into deterministic neural architectures to produce well-calibrated predictive distributions without retraining. Having established a principled framework for estimating uncertainty, we now turn to the complementary question of *why* deep learning models—often highly over-parameterized—are nevertheless capable of generalizing effectively. Understanding this phenomenon requires moving beyond the estimation of predictive uncertainty to the study of the fundamental mechanisms that govern generalization itself.

## 5.1 Introduction

The success of modern deep neural networks (DNNs) is contradictory when viewed through the lens of classical statistical learning theory. Despite being heavily over-parameterized—often containing far more parameters than training samples—these models not only fit their training data perfectly but also generalize remarkably well to unseen inputs. This empirical observation stands in stark contrast to the predictions of traditional uniform-convergence frameworks, such as VC or Rademacher bounds, and their PAC and PAC–Bayesian generalizations reviewed in Section 2.4, which suggest that interpolation should lead to poor generalization. Bridging this theoretical gap has become one of the central challenges in contemporary machine learning research.

This chapter unifies and extends two complementary lines of work aimed at understanding why deep networks generalize in the over–parameterized regime. The first investigates *explicit and ensemble–based mechanisms* that promote generalization by encouraging diversity among models, while the second develops a *distribution–dependent theoretical framework*—based on large–deviation and Chernoff–style bounds—that ex-plains generalization even when interpolation occurs. Together, these perspectives provide a cohesive view that links the geometry of the loss landscape, the stochasticity of training dynamics, and the concentration of empirical losses to the observed generalization behavior of deep models.

A natural connection between this chapter and the previous one lies in viewing generalization as a form of *predictive uncertainty control*. While post-hoc uncertainty estimation methods such as VaLLA and FMGPs quantify uncertainty *after* training, the mechanisms that enable generalization can be seen as constraining uncertainty *during* training, by limiting how model predictions vary across unseen data. From this perspective, generalization arises not only from architectural design or optimization choices, but from how the learning dynamics shape the effective posterior over model hypotheses. Concepts such as flat minima, ensemble diversity, and loss concentration can be interpreted within this probabilistic framework: each corresponds to a preference for solutions whose predictive distributions remain stable under small perturbations of data, parameters, or initialization. This viewpoint provides a coherent transition from uncertainty quantification to the theoretical study of generalization developed in this chapter.

Section 5.2 introduces a rigorous theory of *diversity and generalization* in deep ensembles. It formalizes the long-standing intuition that ensembles perform best when their constituent members make diverse errors. A unified diversity measure is proposed, leading to a decomposition of the ensemble generalization error into two interpretable components: the average individual error and a diversity-dependent correction term. This decomposition is then grounded in the PAC-Bayesian framework, yielding distribution-dependent generalization guarantees and practical regularization objectives that encourage beneficial diversity. Empirical results on image classification and regression tasks confirm that the theoretical diversity-error relationship aligns with observed ensemble performance.

Section 5.3 develops *PAC–Chernoff bounds*, a new class of generalization bounds that remain meaningful in the interpolation regime. By expressing the generalization error in terms of a *rate function* derived from large-deviation theory, the analysis captures how smoothness, regularization, and architectural invariances influence generalization. The resulting bounds are tight for any model that perfectly fits its training data and naturally reproduce the *double–descent* phenomenon observed in practice. Classical regularizers such as weight decay, distance–from–initialization penalties, or input–gradient control are all shown to act by enlarging the rate function, effectively increasing model smoothness and improving generalization.

Finally, Section 5.4 examines the *implicit bias of stochastic gradient descent* (*SGD*) through the same large–deviation perspective. The stochasticity of mini–batch updates is shown to induce an implicit regularization effect, biasing learning toward flatter minima with higher loss concentration. This provides a distribution–dependent explanation for SGD's empirical tendency to find solutions with superior generalization properties, and clarifies how optimization hyperparameters—such as learning rate, batch size, and noise structure—affect this bias.

In summary, this chapter demonstrates that generalization in modern neural networks can be understood as an interplay between *diversity*, *smoothness*, and *concentration*. Whether through explicit ensemble training, distribution-dependent PAC-Chernoff bounds, or the stochastic dynamics of SGD, these mechanisms collectively define how deep learning models achieve robust performance beyond classical theoretical limits.

## 5.2 Diversity and Generalization in Deep Neural Network Ensembles

Ensemble methods are one of the most widely used and studied techniques in machine learning (Breiman, 1996; Breiman, 2001; Hansen and Salamon, 1990). They have been successfully applied in many real-world problems (Girshick et al., 2014; J. Wang et al., 2012; Ykhlef and Bouchaffra, 2017; X. Zhou et al., 2014) and are usually part of the winning strategies in many machine learning competitions (T. Chen and Guestrin, 2016; Hoch, 2015; Puurula et al., 2014; Stallkamp et al., 2012). Recently, ensembles have also become very popular to improve uncertainty modeling in deep neural networks (Lakshminarayanan et al., 2017; Maddox et al., 2019; Wen et al., 2019; Wenzel et al., 2020b).

Ensembles are created by combining several individual predictors. It is widely accepted (Dietterich, 2000; Lu et al., 2010) that the prediction performance of an ensemble jointly depends on the individual performance and the diversity of its individual members. Intuitively speaking, a set of predictors is diverse when their predictions do not coincide on all the samples. It is known that when classifiers are diverse, they tend to make independent errors; therefore, when they are aggregated, their errors tend to cancel out (Berend and Kontorovich, 2016), which improves the ensemble prediction. For this reason, diversity has long been recognized as a key factor in ensemble performance (Brown et al., 2005; Cunningham and Carney, 2000; Kuncheva and Whitaker, 2003). The same cancellation of errors effect happens in the case of neural network ensembles (Hansen and Salamon, 1990; Lakshminarayanan et al., 2017; S. Lee et al., 2016) where heuristic measures of diversity are usually analyzed to get insights into the ensemble learning algorithms (Fort et al., 2019; Wen et al., 2019; Wenzel et al., 2020b).

Unfortunately, there is a lack of consensus surrounding the underlying theory that can explain the role of diversity in the generalization performance of ensembles. The error rate of an ensemble and an individual predictor, for example, is well defined by the use of a loss function, but there is no well-established definition of diversity (Kuncheva and Whitaker, 2003). It is not clear how exactly the diversity among ensemble members affects the generalization error of the ensemble.

In this work, a novel theoretical framework that explains the relationship between diversity and the generalization performance of an ensemble is introduced. This theoretical framework is derived from previously published results with no direct connec-tion among them (Krogh and Vedelsby, 1994; Masegosa, 2020; Masegosa et al., 2020). The main contribution is finding a theoretically sound way to combine these previous results in a single theoretical framework that explains the role of diversity in the generalization performance of a wide range of different ensembles. This general framework could potentially help the machine learning community to have a better understanding of the underlying trade-offs that have to be considered when designing novel ensemble learning algorithms, especially in the context of neural networks. The detailed contributions of this framework are the following:

1. A general measure of ensemble diversity,
2. A theoretical analysis that shows how the correlation among ensemble members affects diversity,
3. The exact trade-off that exists between this diversity measure, the performance of the individual predictors, and the generalization error of the ensemble,
4. An analysis of the strategies used by most of the current neural network ensemble learning algorithms to promote diversity,
5. An empirical evaluation of this theoretical framework.

This analysis covers model averaging and weighted majority vote ensembles under the cross-entropy loss, square error, and 0-1 loss.

### 5.2.1 Related Work

The concept of diversity has been described under various labels, including ambiguity (Krogh and Vedelsby, 1994), dependency (Z.-H. Zhou, 2012), orthogonality (Kuncheva and Whitaker, 2003), and disagreement (Masegosa et al., 2020), among others. An extensive literature proposes numerous measures of diversity, most of which are defined directly from the predictions of individual models (Buschjäger et al., 2020; Chandra and Yao, 2004; Kuncheva and Whitaker, 2003; Roli et al., 2001; Tang et al., 2006; Z.-H. Zhou and Li, 2010). Nevertheless, none of these works introduces a generic measure that is applicable across heterogeneous ensemble types while maintaining a formal connection to the ensemble generalization error.

#### Diversity and Generalization

The earliest theoretical attempts to explain why diversity reduces ensemble error date back to Krogh and Vedelsby (1994) and Geman et al. (1992) in the context of regression ensembles. These analyses demonstrate that higher diversity can lower prediction error; however, they do not characterize how the empirical diversity of a given ensemble relates to its generalization error. This thesis extends the results of Krogh and Vedelsby (1994) and establishes such a link for regression ensembles via PAC-Bayesian bounds (McAllester, 1998). The analysis further carries over to ensembles of classifiers.

Several studies have sought to adapt the squared-error decomposition used for regression ensembles (Krogh and Vedelsby, 1994) to classification settings (Brown, 2009;

Z. Jiang et al., 2017; Y. Yu et al., 2011; Z.-H. Zhou, 2012). Building on PAC-Bayesian upper bounds on the ensemble risk, the present work derives a new decomposition applicable to a broad class of loss functions; the decomposition explicitly separates a term due to the errors of individual members from a diversity component.

In the setting of majority-vote ensembles and PAC-Bayesian theory, Laviolette et al. (2011) and Germain et al. (2015) proposed bounds that depend on the error rate and the variance of the individual classifiers, where the variance term admits an interpretation as a diversity measure. However, this analysis is restricted to binary classification.

More recently, Masegosa et al. (2020) developed a PAC-Bayesian analysis of majority-vote ensembles for multiclass classification. Nonetheless, the resulting bound includes an explicit diversity term only in the binary case, and that term coincides with the one of Germain et al. (2015) and Laviolette et al. (2011).

This thesis extends the results of Masegosa et al. (2020) by providing a PAC-Bayesian bound for multiclass classification that explicitly incorporates a novel diversity term. Additional connections are established with regression ensembles and with model averaging for probabilistic classifiers.

In the context of model averaging for probabilistic classifiers, Masegosa (2020) introduced a PAC-Bayesian analysis whose bound explicitly contains a diversity term. However, that work primarily addresses Bayesian model averaging under model misspecification; the role of diversity in ensemble learning is only briefly discussed and is neither theoretically nor empirically examined. The present thesis undertakes a substantially more comprehensive theoretical and empirical study of how diversity affects the generalization performance of several types of ensembles.

**Diversity and Ensemble Learning**

Virtually all ensemble methods promote diversity among their constituent models, either implicitly or explicitly. Classical techniques such as Bagging (Breiman, 1996; Breiman, 2001) and Boosting (Freund and Schapire, 1996) encourage diversity implicitly by generating varied training datasets. Deep ensembles likewise rely on implicit mechanisms, including random initialization (Lakshminarayanan et al., 2017; Wen et al., 2019), modifications to the optimizer (Maddox et al., 2019; Wenzel et al., 2020b; R. Zhang et al., 2019), and heterogeneous hyperparameter configurations (Wenzel et al., 2020b). This thesis provides a theoretical account of why such diversity-inducing strategies lead to superior ensemble performance.

A growing body of ensemble-learning methods optimizes loss functions augmented with terms that explicitly induce diversity (Buschjäger et al., 2020; Jain et al., 2020; Z. Jiang et al., 2017; Y. Liu and Yao, 1999; Pang et al., 2019). However, these formulations generally lack a formal link to the ensemble's generalization error, in contrast to the approach developed here. Masegosa (2020) proposed an ensemble-learning algorithm for the cross-entropy loss derived from PAC-Bayesian bounds that explicitly promote diversity, but its evaluation was limited to ensembles of simple neural net-works. The present thesis substantially extends that empirical study and applies the analysis to regression ensembles and weighted majority-vote ensembles.

### 5.2.2 Preliminaries

Before introducing the main theoretical results, it is convenient to establish notation and clarify what is meant by an *ensemble* and a *ρ-weighted* predictor.

In general terms, an *ensemble* is a collection of predictors—also called *members* or *base models*—that are trained independently (or semi-independently) and whose predictions are combined to form a single, aggregate output. Ensembles are widely used in machine learning to reduce variance, improve robustness, and obtain better-calibrated uncertainty estimates compared to any individual model. Formally, if $\boldsymbol{\Theta}$ denotes the parameter space of individual models and $h_{\boldsymbol{\theta}} = h(\cdot; \boldsymbol{\theta})$ is a predictor parameterized by $\boldsymbol{\theta} \in \boldsymbol{\Theta}$, an ensemble corresponds to a finite or continuous collection of such predictors indexed by $\boldsymbol{\theta}$. The ensemble prediction is then obtained by averaging or voting over its members according to a weighting scheme.

Let $\rho$ be a probability distribution (or discrete probability mass function) defined over the parameter space $\boldsymbol{\Theta}$. This distribution specifies how much *weight* or *importance* is assigned to each individual predictor $h_{\boldsymbol{\theta}}$ within the ensemble. The expectation $\mathbb{E}_{\rho}[\,\cdot\,]$ thus denotes an average with respect to this distribution. In the simplest case, $\rho$ is uniform over $K$ members, $\rho(\boldsymbol{\theta}_k) = 1/K$, corresponding to an equally weighted ensemble. More generally, $\rho$ may assign higher weight to better-performing or more diverse members. The term *ρ-weighted* therefore refers to any ensemble quantity (e.g., prediction, loss, or risk) that is defined as an expectation with respect to $\rho$.

Formally, let $D = \{(\mathbf{x}_1, y_1), \dots, (\mathbf{x}_n, y_n)\}$ denote a set of independent and identically distributed samples drawn from an unknown distribution $\nu$ over $\mathcal{X} \times \mathcal{Y}$, where $\mathcal{X}$ and $\mathcal{Y}$ denote the input and output spaces, respectively. For a single model $h_{\boldsymbol{\theta}}$, its prediction for an input $\mathbf{x} \in \mathcal{X}$ is $h_{\boldsymbol{\theta}}(\mathbf{x}) = h(\mathbf{x}; \boldsymbol{\theta})$. An ensemble predictor is then defined by integrating (or summing) the predictions of individual members under $\rho$, yielding the *ρ-weighted ensemble prediction*

$$h_{\rho}(\mathbf{x}) = \mathbb{E}_{\rho}[\, h_{\boldsymbol{\theta}}(\mathbf{x}) \,]. \tag{5.1}$$

In the common case of a finite ensemble composed of $K$ predictors $\{h_{\boldsymbol{\theta}_1}, \dots, h_{\boldsymbol{\theta}_K}\}$, the distribution $\rho$ reduces to a discrete probability mass function over these members, $\rho(\boldsymbol{\theta}_k) = w_k$, where $w_k \geq 0$ and $\sum_{k=1}^{K} w_k = 1$. The ensemble output is therefore the weighted average of the individual predictions:

$$h_{\rho}(\mathbf{x}) = \sum_{k=1}^{K} w_k \, h_{\boldsymbol{\theta}_k}(\mathbf{x}). \tag{5.2}$$

When all members are assigned equal weights, $w_k = 1/K$, this simplifies to the uniform

average:

$$h_\rho(\mathbf{x}) = \frac{1}{K}\sum_{k=1}^{K} h_{\boldsymbol{\theta}_k}(\mathbf{x}), \tag{5.3}$$

which corresponds to the standard ensemble used in deep learning practice, such as in bagging or deep ensembles; and the main focus of this thesis. In this sense, $\rho$ encodes both the composition of the ensemble and the relative influence of each member in forming the aggregate prediction.

Depending on the type of learning problem, an ensemble combines its members' predictions in different ways, leading to distinct forms of the ensemble predictor $h_\rho(\mathbf{x})$ and its associated loss function $\ell(\rho, \mathbf{x}, y)$. The three most common cases are as follows:

- **Regression ensembles** use the $\rho$-weighted model average predictor and the *squared error loss* (denoted as the $sq$-loss). The prediction of the ensemble for a given input $\mathbf{x}$ is the $\rho$-weighted average of the individual model predictions:

  $$h_\rho(\mathbf{x}) = \mathbb{E}_\rho[h(\mathbf{x}; \boldsymbol{\theta})]. \tag{5.4}$$

  For a specific data point $(\mathbf{x}, y)$, the loss of an individual regressor is

  $$\ell_{sq}(\boldsymbol{\theta}, \mathbf{x}, y) = (y - h(\mathbf{x}; \boldsymbol{\theta}))^2, \tag{5.5}$$

  and the ensemble loss is

  $$\ell_{sq}(\rho, \mathbf{x}, y) = \left(y - \mathbb{E}_\rho[h(\mathbf{x}, \boldsymbol{\theta})]\right)^2. \tag{5.6}$$

  In other words, the ensemble predicts the mean output across its members and measures the squared deviation from the true value.
- **Weighted majority voting ensembles** are defined for classification tasks using the $\rho$-weighted majority vote predictor and the *zero–one loss* (denoted as $0/1$-loss). Each individual classifier $h_{\boldsymbol{\theta}}$ outputs a discrete class label in $\mathcal{Y}$. The ensemble prediction is obtained by computing, for each possible class $y'$, the $\rho$-weighted probability that the individual classifiers predict $y'$, and selecting the class with the highest average support:

  $$h_\rho(\mathbf{x}) = \arg\max_{y' \in \mathcal{Y}} \mathbb{E}_\rho\big[\mathbb{I}(h(\mathbf{x}, \boldsymbol{\theta}) = y')\big]. \tag{5.7}$$

  For a finite ensemble, this corresponds to the familiar majority vote:

  $$h_\rho(\mathbf{x}) = \arg\max_{y' \in \mathcal{Y}} \sum_{k=1}^{K} w_k\, \mathbb{I}(h(\mathbf{x}; \boldsymbol{\theta}_k) = y')\,. \tag{5.8}$$

  The individual loss for a data point $(\mathbf{x}, y)$ is

  $$\ell_{0/1}(\boldsymbol{\theta}, \mathbf{x}, y) = \mathbb{I}[\, h(\mathbf{x}, \boldsymbol{\theta}) \neq y\,], \tag{5.9}$$

and the ensemble loss is

$$\ell_{0/1}(\rho, \mathbf{x}, y) = \mathbb{I}\left[\arg\max_{y' \in \mathcal{Y}} \mathbb{E}_\rho\left[\mathbb{I}(h(\mathbf{x}, \boldsymbol{\theta}) = y')\right] \neq y\right]. \quad (5.10)$$

Thus, the ensemble prediction corresponds to a weighted vote across models, and the loss indicates whether the majority prediction matches the ground truth.

- **Model averaging ensembles** are used for probabilistic classifiers, where each model outputs a predictive distribution over the labels, $h(\mathbf{x}; \boldsymbol{\theta}) = P(\cdot|\mathbf{x}, \boldsymbol{\theta})$. The ensemble prediction is the $\rho$-weighted average of these probability distributions:

$$P_\rho(y|\mathbf{x}) = \mathbb{E}_\rho[P(y|\mathbf{x}, \boldsymbol{\theta})], \quad (5.11)$$

which, in the finite case, becomes

$$P_\rho(y|\mathbf{x}) = \sum_{k=1}^{K} w_k \, P(y|\mathbf{x}, \boldsymbol{\theta}_k). \quad (5.12)$$

This is the standard approach used in Bayesian model averaging and deep ensembles, where the final predicted probability is the mean of the members' probabilities. For a specific data point $(\mathbf{x}, y)$, the individual and ensemble cross-entropy losses are

$$\ell_{ce}(\boldsymbol{\theta}, \mathbf{x}, y) = -\log P(y|\mathbf{x}, \boldsymbol{\theta}), \quad \ell_{ce}(\rho, \mathbf{x}, y) = -\log \mathbb{E}_\rho[P(y|\mathbf{x}, \boldsymbol{\theta})]. \quad (5.13)$$

The ensemble prediction is thus a mixture of individual predictive distributions, penalized by the negative log-likelihood of the true label.

For any of these loss functions, denoted generically as $\ell(\boldsymbol{\theta}, \mathbf{x}, y)$, the empirical loss of an individual model $h_{\boldsymbol{\theta}}$ over a dataset $D$ is

$$\hat{L}(\boldsymbol{\theta}, D) = \frac{1}{n} \sum_{i=1}^{n} \ell(\boldsymbol{\theta}, \mathbf{x}_i, y_i), \quad (5.14)$$

and its expected population loss is

$$L(\boldsymbol{\theta}) = \mathbb{E}_\nu[\ell(\boldsymbol{\theta}, \mathbf{x}, y)], \quad (5.15)$$

where the expectation $\mathbb{E}_\nu$ is taken over $(\mathbf{x}, y) \sim \nu$. Analogously, the expected loss of an ensemble is

$$L(\rho) = \mathbb{E}_\nu[\ell(\rho, \mathbf{x}, y)]. \quad (5.16)$$

Throughout the remainder of this chapter, the quantities $L(\rho)$, $L(\boldsymbol{\theta})$, and $\hat{L}(\boldsymbol{\theta}, D)$ are written without explicit subscripts when the context is clear. Unless otherwise stated, all analyses apply to any of the three ensemble settings above. When an explicit subscript is provided (i.e., $sq$, $0/1$, or $ce$), it refers to the corresponding ensemble model.

### 5.2.3 Diversity and Generalization

This section develops a PAC-Bayesian bound for multiclass classification that explicitly includes a novel term quantifying the diversity of the ensemble.

#### 5.2.3.1 Decomposing the Loss of an Ensemble via an Upper Bound

The following result formalizes the intuition that the generalization ability of an ensemble depends not only on the average accuracy of its members but also on how *diverse* their predictions are. The bound shows that the expected ensemble loss $L(\rho)$ can be upper-bounded by the average individual loss $\mathbb{E}_\rho[L(\boldsymbol{\theta})]$ minus a nonnegative diversity term $\mathbb{D}(\rho)$. In other words, even if the constituent models have similar average performance, the ensemble can achieve a strictly lower overall loss whenever its members make *uncorrelated* or *complementary* errors.

The factor $\alpha$ accounts for the scaling properties of different loss functions, ensuring the bound holds uniformly across regression, probabilistic, and classification settings. For the squared-error loss, the bound becomes exact, confirming that the ensemble's improvement arises precisely from the variance reduction due to averaging. The diversity measures $\mathbb{D}_{sq}(\rho)$, $\mathbb{D}_{ce}(\rho)$, and $\mathbb{D}_{0/1}(\rho)$ quantify, for each case, the expected variability in predictions across ensemble members under the data distribution $\nu$. Each can be written as the expected variance (with respect to $\rho$) of a function $f(y, \mathbf{x}; \boldsymbol{\theta})$ determined by the loss type, unifying all three formulations under a common framework.

In summary, the theorem provides a principled decomposition of the ensemble's error into two interpretable components:

1. the *average loss* of the individual models, reflecting their collective accuracy, and
2. a *diversity term* that captures how much the models disagree.

A higher diversity $\mathbb{D}(\rho)$ therefore implies a tighter upper bound—and potentially a lower ensemble loss—offering a theoretical justification for the well-known empirical benefit of combining diverse models.

**Theorem 5.1.** *For any distribution $\rho$ over $\boldsymbol{\Theta}$, and any of the three considered loss functions for ensembles, there exists a function $\mathbb{D}(\rho)$ such that*

$$L(\rho) \leq \alpha(\mathbb{E}_\rho[L(\boldsymbol{\theta})] - \mathbb{D}(\rho)), \tag{5.17}$$

*where $\alpha$ equals $1$ if the $sq$-loss or the $ce$-loss are considered, and $4$ for the $0/1$-loss. Furthermore, for the $sq$-loss, this inequality becomes an equality. The expression of the diversity measure for each of these loss functions is:*

$$\mathbb{D}_{sq}(\rho) \;:=\; \mathbb{E}_\nu\Big[\mathbb{V}_\rho(h_R(\mathbf{x};\boldsymbol{\theta}))\Big], \tag{5.18}$$

$$\mathbb{D}_{ce}(\rho) \;:=\; \mathbb{E}_\nu\left[\mathbb{V}_\rho\left(\frac{P(y|\mathbf{x},\boldsymbol{\theta})}{\sqrt{2}\max_{\boldsymbol{\theta}} P(y|\mathbf{x},\boldsymbol{\theta})}\right)\right], \tag{5.19}$$

$$\mathbb{D}_{0/1}(\rho) \;:=\; \mathbb{E}_\nu\Big[\mathbb{V}_\rho\Big(\mathbb{I}(h_W(\mathbf{x};\boldsymbol{\theta}) \neq y)\Big)\Big], \tag{5.20}$$

*where* $\mathbb{V}_\rho(\cdot)$ *denotes the variance of a function w.r.t.* $\rho$*, and for* $\mathbb{D}_{ce}(\rho)$ *to be well-defined it must verify that* $0 < \max_{\boldsymbol{\theta}\in\boldsymbol{\Theta}} P(y|\mathbf{x},\boldsymbol{\theta}) \leq 1$ *for every* $(\mathbf{x}, y) \in supp(\nu)$*. Finally, note that all diversity terms described above can be written as*

$$\mathbb{D}(\rho) = \mathbb{E}_\nu\Big[\mathbb{V}_\rho\left(f(y,\mathbf{x};\boldsymbol{\theta})\right)\Big], \tag{5.21}$$

*with a specific function* $f$ *for each of the loss functions.* [Proof in Appendix C.4]

The $sq$-version of Theorem 5.1 is equivalent to the well-known decomposition of the squared error of a regression ensemble (Krogh and Vedelsby, 1994). On the other hand, the $0/1$-version of Theorem 5.1 is novel and based on the analysis given by Masegosa et al., 2020, which, in turn, is based on second-order Markov inequalities. Lastly, the $ce$-version is equivalent to the one previously proposed by Masegosa, 2020 based on second-order Jensen inequalities (Becker, 2012; Liao and Berg, 2019). More precisely, the following result demonstrates how to define a tighter second-order Jensen bound for the cross-entropy error using the Jensen inequality stated in Liao and Berg, 2019.

**Theorem 5.2.** *Any distribution* $\rho$ *over* $\boldsymbol{\Theta}$ *satisfies the following inequality,*

$$L_{ce}(\rho) \leq \mathbb{E}_\rho[L_{ce}(\boldsymbol{\theta})] - \mathbb{V}^T_{ce}(\rho), \tag{5.22}$$

*where* $\mathbb{V}^T_{ce}(\rho)$ *is the normalized variance of* $P(y|\mathbf{x},\boldsymbol{\theta})$ *w.r.t.* $\rho(\boldsymbol{\theta})$*,*

$$\mathbb{V}^T_{ce}(\rho) := \mathbb{E}_\nu\Big[h(m,\mu)\mathbb{E}_\rho\Big[(P(y|\mathbf{x},\boldsymbol{\theta}) - P(y))^2\Big]\Big]. \tag{5.23}$$

*Where* $\mu := \mathbb{E}_\rho[P(y|\mathbf{x},\boldsymbol{\theta})]$*,* $m := \max_{\boldsymbol{\theta}} P(y|\mathbf{x},\boldsymbol{\theta})$ *and* $h(m,\mu) := \frac{\ln\mu - \ln m}{(m-\mu)^2} + \frac{1}{\mu(m-\mu)}$. [Proof in Appendix C.4]

Theorem 5.2 refines the previous decomposition result by providing a tighter upper bound for the cross-entropy loss based on a *second-order* version of Jensen's inequality (Liao and Berg, 2019). While Theorem 5.1 establishes a general first-order relationship between the ensemble loss $L(\rho)$, the mean individual loss $\mathbb{E}_\rho[L(\boldsymbol{\theta})]$, and a variance-based diversity term $\mathbb{D}(\rho)$, the present theorem introduces an additional correction factor that accounts for the local curvature of the logarithmic function underlying the cross-entropy loss.

In the earlier bound, the diversity term $\mathbb{D}_{ce}(\rho)$ captures the average dispersion of the predictive probabilities $P(y|\mathbf{x},\boldsymbol{\theta})$ across the ensemble members. However, this measure depends linearly on the variance and treats all deviations equally, regardless of how sensitive the logarithmic loss is to those deviations. In contrast, Theorem 5.2 introduces the term $\mathbb{V}^T_{ce}(\rho)$, which incorporates a multiplicative weighting function $h(m,\mu)$ that depends on both the mean probability $\mu$ and its maximum value $m$. This weighting modulates the contribution of the variance according to the curvature of the log function, yielding a *tighter and more accurate* approximation of the ensemble cross-entropy loss.

Intuitively, while Theorem 5.1 provides a general, interpretable decomposition valid for multiple loss types, Theorem 5.2 specializes this idea to the cross-entropy setting, enhancing precision by taking into account second-order information about how averaging probabilities affects the log-likelihood. The result is a sharper characterization of the ensemble's improvement, where the benefit of diversity is not only proportional to disagreement among models but also weighted by how much those disagreements matter in terms of predictive confidence.

#### 5.2.3.2 How to Measure the Diversity of an Ensemble?

In this thesis, the use of the diversity term $\mathbb{D}(\rho)$ given in Theorem 5.1 is proposed as a diversity measure of an ensemble. This diversity measure satisfies some intuitive properties.

**Lemma 5.3.** *The diversity terms $\mathbb{D}(\rho)$ defined in Theorem 5.1 satisfy the following properties:*

1. *If all the ensemble members provide the same predictions, or if $\rho$ outs all its probability mass on a single predictor, then $\mathbb{D}(\rho)$ is null.*
2. $0 \leq \mathbb{D}(\rho) \leq \mathbb{E}_\rho[L(\boldsymbol{\theta})]$.
3. *$\mathbb{D}(\rho)$ is invariant to reparametrizations.*

[Proof in Appendix C.4]

Every point in Lemma 5.3 can be discussed from an empirical point of view:

1. It is clear that if all models make the same prediction or place the same probabilities, the empirical diversity is zero.
2. The same arguments used for the theoretical diversity may be applied to the empirical one, using an empirical version of Theorem 5.1 that reduces to take the expectation over the empirical distribution at each step.
3. To show that invariance beholds, the same procedure can be followed using the empirical expectation. In this case, given that the empirical probability is compactly supported, the variable change theorem holds.

The above properties show that $\mathbb{D}(\rho)$ can be considered as a measure of the diversity of an ensemble. The first property follows the intuition of diversity as a measure of the difference among ensemble members' errors, while the second property is something that has been empirically found in the literature: the diversity of an ensemble usually decreases when the predictive error of individual members is reduced (Fort et al., 2019). The last property is a desirable result for any diversity measure.

Another common knowledge about diversity is that it decreases when the predictors are highly correlated (Berend and Kontorovich, 2016; Brown, 2009; Y. Yu et al., 2011; Z.-H. Zhou, 2012). The following result shows how this diversity measure nicely captures this relationship of diversity and correlation among predictors:

**Theorem 5.4.** *The diversity terms $\mathbb{D}(\rho)$ defined in Theorem 5.1 can be written as*

$$\mathbb{D}(\rho) = \mathbb{V}_{\nu\times\rho}(f(y,\mathbf{x};\boldsymbol{\theta})) - \mathbb{E}_{\rho\times\rho}[Cov_\nu(f(y,\mathbf{x};\boldsymbol{\theta}), f(y,\mathbf{x};\boldsymbol{\theta}'))], \tag{5.24}$$

*where $\rho\times\rho$ denotes the joint distribution over $\boldsymbol{\theta}\times\boldsymbol{\theta}$, $\rho\times\nu$ denotes the joint distribution over $\boldsymbol{\theta}\times(\mathcal{X},\mathcal{Y})$, and $Cov_\nu(\cdot,\cdot)$ is the covariance between two models with respect to the data generating distribution $\nu$.* [Proof in Appendix C.4]

This result states that the proposed diversity measure increases as the correlation among ensemble members is reduced. Eventually, a much higher diversity will be obtained if ensembles are anti-correlated (i.e. negative covariance). However, Theorem 5.4 also introduces a novel insight: ensemble diversity is not only about the correlation among individual models. The above decomposition of the diversity also shows, through the first term ($\mathbb{V}_{\nu\times\rho}(f(y,\mathbf{x};\boldsymbol{\theta}))$), that ensembles with high diversity should provide different predictions across the different individual models and across the different data samples. Although this is out of the scope of this thesis, this could potentially be used to study why randomization approaches to build ensembles, for example, random forests (Breiman, 2001), result in highly diverse ensembles, as they directly maximize this variance, which is positively related to diversity.

#### 5.2.3.3 How is Diversity Related to the Performance of an Ensemble?

The role that diversity plays in the performance of an ensemble is described by Theorem 5.1. According to this result, the generalization error of an ensemble $L(\rho)$ should be reduced as $\mathbb{D}(\rho)$ is increased, which is a measure of the diversity of the ensemble as shown in the previous section. However, Theorem 5.1 provides additional novel insights.

The following result formalizes an empirically observed phenomenon (Dietterich, 2000; Lu et al., 2010) that higher ensemble diversity induces a higher gap between the average loss of the individual models ($\mathbb{E}_\rho[L(\boldsymbol{\theta})]$) and the expected loss of the ensemble ($L(\rho)$). In other words, the higher the diversity, the higher the advantage of combining these models. The proof is omitted as it is a direct consequence of Theorem 5.1.

**Corollary 5.5.** *For any distribution $\rho$ over $\boldsymbol{\Theta}$, it verifies that*

$$\mathbb{D}(\rho) \leq \mathbb{E}_\rho[L(\boldsymbol{\theta})] - \tfrac{1}{\alpha}L(\rho), \tag{5.25}$$

*where $\alpha$ is equal to $1$ for the $sq$-loss or the $ce$-loss, and $4$ for the $0/1$-loss. Furthermore, for the $sq$-loss, this inequality becomes an equality.*

Another open question in the ensemble's literature is under which situations an ensemble of models outperforms a single model. The following result establishes that this occurs when the ensemble's diversity is large enough.

**Corollary 5.6.** *For any distribution* $\rho$ *over* $\mathbf{\Theta}$, *it verifies that an ensemble of models weighted according to a distribution* $\rho$ *performs better than a single model* $\boldsymbol{\theta}^\star \in \mathbf{\Theta}$, *that is,* $L(\rho) < L(\boldsymbol{\theta}^\star)$, *if*

$$\mathbb{E}_\rho[L(\boldsymbol{\theta})] - \tfrac{1}{\alpha}L(\boldsymbol{\theta}^\star) < \mathbb{D}(\rho) \tag{5.26}$$

*where* $\alpha$ *is equal to* 1 *for the* sq*-loss or the* ce*-loss, and* 4 *for the* 0/1*-loss. For the* sq*-loss, the inverse implication also holds.*

However, $\mathbb{D}(\rho)$ is defined in terms of the unknown data-generating distribution $\nu$. As a result, $\mathbb{D}(\rho)$ cannot be computed. To address this issue, the use of the empirical version of $\mathbb{D}(\rho)$, denoted by $\hat{\mathbb{D}}(\rho, D)$ is proposed. This quantity directly depends on the empirical distribution defined by the data sample $D$. $\hat{\mathbb{D}}(\rho, D)$ satisfies the same properties as $\mathbb{D}(\rho)$; more precisely, it verifies every point in Lemma 5.3:

1. It is clear that if all models make the same prediction or place the same probabilities, the empirical diversity is zero.
2. In order to show this, the same arguments used for the theoretical diversity must be applied to the empirical one, using an empirical version of Theorem 1 that reduces to take the expectation over the empirical distribution at each step.
3. To show that invariance beholds, the same procedure followed in the proof can be used again. In this case, given that the empirical probability is compactly supported, the variable change theorem holds.

Moreover, $\hat{\mathbb{D}}(\rho, D)$ quantifies the diversity of an ensemble on a given sample $D$. A central question is how this empirical diversity measure relates to the ensemble's generalization performance. To address this question, the analysis relies on PAC-Bayesian bounds.

PAC-Bayesian analysis (Langford and Shawe-Taylor, 2002; McAllester, 1998; Seeger, 2002; see also Section 2.4 for a self-contained introduction) provides *Probably Approximately Correct* (PAC) upper bounds on a model's generalization error. The resulting PAC-Bayesian bounds depend on empirical quantities. Specifically, consider a family of bounds that upper-bound the generalization error of an ensemble, $L(\rho)$, in terms of the $\rho$-weighted empirical errors of the individual models, $\mathbb{E}_\rho[\hat{L}(\boldsymbol{\theta}, D)]$, together with the empirical diversity of their predictions, $\hat{\mathbb{D}}(\rho, D)$. These PAC-Bayesian bounds hold with high probability over random draws of the training sample.

**Theorem 5.7.** *For any prior distribution* $\pi$ *over* $\boldsymbol{\theta}$ *independent of* $D$, *any* $\xi \in (0, 1)$, *and any* $\lambda > 0$, *with probability at least* $1 - \xi$ *over draws of training data* $D \sim \nu^n$, *for all distributions* $\rho$ *over* $\boldsymbol{\theta}$, *simultaneously,*

$$L(\rho) \leq \alpha\left(\mathbb{E}_\rho[\hat{L}(\boldsymbol{\theta}, D)] - \hat{\mathbb{D}}(\rho, D) + \frac{2\mathrm{KL}(\rho|\pi) + \epsilon(\nu, \pi, \lambda, n, \xi)}{\lambda n}\right) \tag{5.27}$$

*where* $\alpha$ *is equal to* 1 *for the* sq*-loss or the* ce*-loss, and* 4 *for the* 0/1*-loss. Furthermore,* $\epsilon$ *is a positive function that is independent of* $\rho$ *but also depends on the specific loss* (*the functional forms of these* $\epsilon$ *terms can be found in the proof*). [Proof in Appendix C.4]

The $sq$ and $0/1$ versions of the PAC-Bayesian bound of Theorem 5.7 are novel. While the $ce$-version of Theorem 5.7 was previously proposed in Masegosa, 2020.

The preceding result provides a clear theoretical explanation for the widely observed phenomena of accurate and diversified models leading to ensemble methods with low generalization error (Dietterich, 2000). High accurate models imply lower values of $\mathbb{E}_\rho[\hat{L}(\boldsymbol{\theta}, D)]$, while highly diverse models imply higher values of $\hat{\mathbb{D}}(\rho, D)$. Consequently, the combined effect of both, as described by the second-order PAC-Bayesian bounds of Theorem 5.7, induces a lower upper bound on the generalization error of the ensemble.

#### 5.2.3.4 How to Exploit Diversity to Learn Ensembles?

The PAC-Bayesian bounds of Theorem 5.7 provide a rigorous framework for ensemble learning. Because these bounds hold simultaneously for all distributions $\rho$, it is possible to select the distribution that minimizes the resulting high-probability upper bounds on the ensemble's generalization error (Langford and Shawe-Taylor, 2002; McAllester, 1998; Seeger, 2002). Section 5.2.5 details how this approach applies to ensembles of neural networks (i.e., how each distribution $\rho$ specifies an ensemble of neural networks), following the ideas introduced by Masegosa (2020).

As discussed in Section 5.2.1, a growing number of ensemble-learning methods employ objectives that, in addition to favoring members with low error (small $\mathbb{E}_\rho[\hat{L}(\boldsymbol{\theta}, D)]$), explicitly encourage diversity. The present analysis furnishes a theoretical justification for such approaches: promoting diversity can improve the ensemble's generalization performance. Further elaboration is provided in Section 5.2.5.

The situation differs for ensembles of deep neural networks. In this setting, individual models are typically large and operate in the interpolation regime (C. Zhang et al., 2017); consequently, the term $\mathbb{E}_\rho[\hat{L}(\boldsymbol{\theta}, D)]$ appearing in Theorem 5.7 is nearly zero, if not exactly so. Moreover, by Lemma 5.3, the empirical diversity term $\hat{\mathbb{D}}(\rho, D)$ also becomes negligible. As a result, algorithms that directly minimize the bound in Theorem 5.7 tend to be ineffective, since the diversity component $\hat{\mathbb{D}}(\rho, D)$ exerts little influence on the learning dynamics. A formal statement of this issue is given in Section 5.2.5, and the next section provides empirical evidence.

This observation appears contradictory: the analysis indicates that high diversity is crucial for generalization, yet deep ensembles exhibit near-zero empirical diversity alongside strong generalization. The apparent paradox is resolved by noting that low diversity on the training data (i.e., $\hat{\mathbb{D}}(\rho, D) \approx 0$) does not imply low diversity on the test distribution (i.e., $\mathbb{D}(\rho) > 0$), which is the relevant quantity for generalization according to Theorem 5.1.

Most of the current state-of-the-art deep ensemble learning algorithms follow this general scheme: they independently learn each neural network of the ensemble by minimizing the provided loss function (usually the $ce$-loss or the $sq$-loss) using some randomization method (e.g. random initialization of the parameters (Lakshminarayanan et al., 2017; Wen et al., 2019), or different hyper-parameters for the gradient descent algorithm (Wenzel et al., 2020b)) in order to force the gradient descent algorithm to converge to different local minima of the loss function. Current state-of-the-art deep ensemble learning algorithms exploit the highly multi-modal landscape of the loss function (Fort et al., 2019) to achieve that.

In consequence, when the ensemble is composed of $K$ models $\{\boldsymbol{\theta}_1, \ldots, \boldsymbol{\theta}_K\}$ defining different predictive functions, then the expected diversity $\mathbb{D}(\rho)$ will be positive, as stated in the following result:

**Lemma 5.8.** *If there exists $\boldsymbol{\theta}_i \neq \boldsymbol{\theta}_j$ and an input sample $\mathbf{x} \in supp(\nu)$, where $supp(\nu)$ denotes the support of the data generating function, such that $h(\mathbf{x}; \boldsymbol{\theta}_i) \neq h(\mathbf{x}; \boldsymbol{\theta}_j)$, it verifies that $\mathbb{D}(\rho) > 0$.*

This analysis shows that ensembles of deep neural networks promote diversity by learning neural networks which induce different predictive functions. And this is achieved, in general, by using randomization strategies that exploit the highly multi-modal landscape of the loss function of deep neural networks. The existence evidence in some previous works (Fort et al., 2019) clearly aligns with these conclusions.

### 5.2.4 Experimental Evaluation

The conducted experimentation is applied to the three kinds of ensembles considered in this thesis. Note that regression ensembles are associated to the $sq$-loss, weighted majority vote ensembles are associated to the $0/1$-loss, and model averaging ensembles to the $ce$-loss. These losses will be used as a way to refer to the different ensembles (for example, the generalization error of a regression ensemble will be denoted by $L_{sq}(\rho)$).

The regression ensemble is evaluated on the Wine-Quality (Cortez et al., 2009) data set using a multilayer perceptron with one layer containing 50 hidden units and a dropout layer; this model is denoted as MLP50.

The majority vote and the model averaging ensembles are evaluated on two standard data sets, CIFAR-10 and CIFAR-100 (Krizhevsky, Hinton, et al., 2009), using two networks: LeNet5 (LeCun et al., 1989) and Resnet20 (He et al., 2016a). LeNet5 is chosen for its simplicity, meaning it does not operate on the interpolation regime (i.e. the empirical error is not close to zero). On the other hand, ResNet20 operates in (or close to) the interpolation regime for the CIFAR-10 and the CIFAR-100 data sets. More complex networks could have been employed, but for the aim of this empirical evaluation, ResNet20 is powerful enough.

All ensembles are made of four individual models. This is not an arbitrary amount, as this setting is the default one in the widely used neural network ensemble library *Uncertainty Baselines*.

Two ensemble learning algorithms are considered: *ensemble* (Lakshminarayanan et al., 2017), in which each randomly initialized model is optimized independently via gradient descent, and *P2B-Ensemble* (Masegosa, 2020), in which randomly initialized models are learned jointly by minimizing the PAC-Bayesian bound of Theorem 5.7. The former is a representative approach to neural-network ensembles based on randomization (Lakshminarayanan et al., 2017). In both cases, the $ce$-loss is used for CIFAR-10 and CIFAR-100, while the $sq$-loss is used for Wine-Quality. For majority-vote ensembles, the $ce$-loss is optimized in place of the non-differentiable $0/1$-loss, as $ce$ provides a suitable surrogate. In all settings, error and diversity measures are computed post hoc, enabling an analysis of their relationship.

**Table 5.1:** ***Training hyper-parameters employed for LeNet-5, ResNet-20 and MLP-50.*** *Shown are the base learning rate, total number of training epochs, $\ell_2$ regularization coefficient, milestones of the piece-wise constant learning-rate decay schedule* $[60, 120, 160]$*, and the per-core mini-batch size used for each architecture.*

| Hyper-parameter | LeNet5 | ResNet20 | MLP50 |
|---|---|---|---|
| base learning rate | 0.001 | 0.1 | 0.001 |
| epochs | 200 | 250 | 250 |
| $\ell_2$ regularization | $2 \cdot 10^{-4}$ | $2 \cdot 10^{-4}$ | $2 \cdot 10^{-4}$ |
| learning rate decay | [60,120,160] | [60,120,160] | [60,120,160] |
| per core batch size | 64 | 64 | 32 |

The test set is used to approximate the generalization error of the ensemble, $L(\rho)$, the error of individual models, $L(\boldsymbol{\theta})$, and the expected diversity of the ensemble, $\mathbb{D}(\rho)$. For experiments employing the $ce$-loss, a tighter diversity measure introduced by Masegosa (2020) (Theorem 5.2) is adopted. Out-of-distribution benchmarks (Snoek et al., 2019) are not considered, as the objective is to empirically assess the theoretical analysis under the assumption that training and test data are drawn from the same distribution.

Each experiment is repeated five times with different random seeds. Reproducibility code is available at https://github.com/PGM-Lab/2022-AISTATS-diversity. The experimental evaluation is conducted on hardware with 8 TPU cores. The set of hyper-parameters considered is reported in Table 5.1.

#### 5.2.4.1 Experimental Evaluation

The evaluation begins by assessing the tightness of the upper bounds provided by Theorem 5.1. Figure 5.1 reports the results. Each point corresponds to an ensemble specified by a distribution $\rho_\delta$. The distance of a point to the identity line quantifies the slack of the bound in Theorem 5.1. For the $sq$-version, the bound is exact, as stated in the theorem. For the $ce$-version, the bound is also fairly tight. For the $0/1$-version, two variants are considered: the original with $\alpha = 4$, which is noticeably loose, and an illustrative variant with $\alpha = 1$ (not an upper bound) suggesting that bounds with constants closer to 1 may be attainable, noting that $\alpha = 4$ is a worst-case factor (Masegosa et al., 2020).

Section 5.2.3.3 argues that diversity is directly linked to the gains obtained by combining models. More precisely, as stated in Corollary 5.5, ensembles with larger diversity $\mathbb{D}(\rho)$ should exhibit a larger gap between the average performance of individual models $\mathbb{E}_\rho[L(\boldsymbol{\theta})]$ and the ensemble performance $L(\rho)$. Figure 5.2 shows that higher di-versity is consistently associated with a larger performance gap, i.e., $\mathbb{E}_\rho[L(\boldsymbol{\theta})] - L(\rho)$, across all settings. The figure also indicates that *P2B-Ensemble*, which explicitly promotes diversity, consistently induces higher-diversity ensembles than the standard *Ensemble* algorithm.

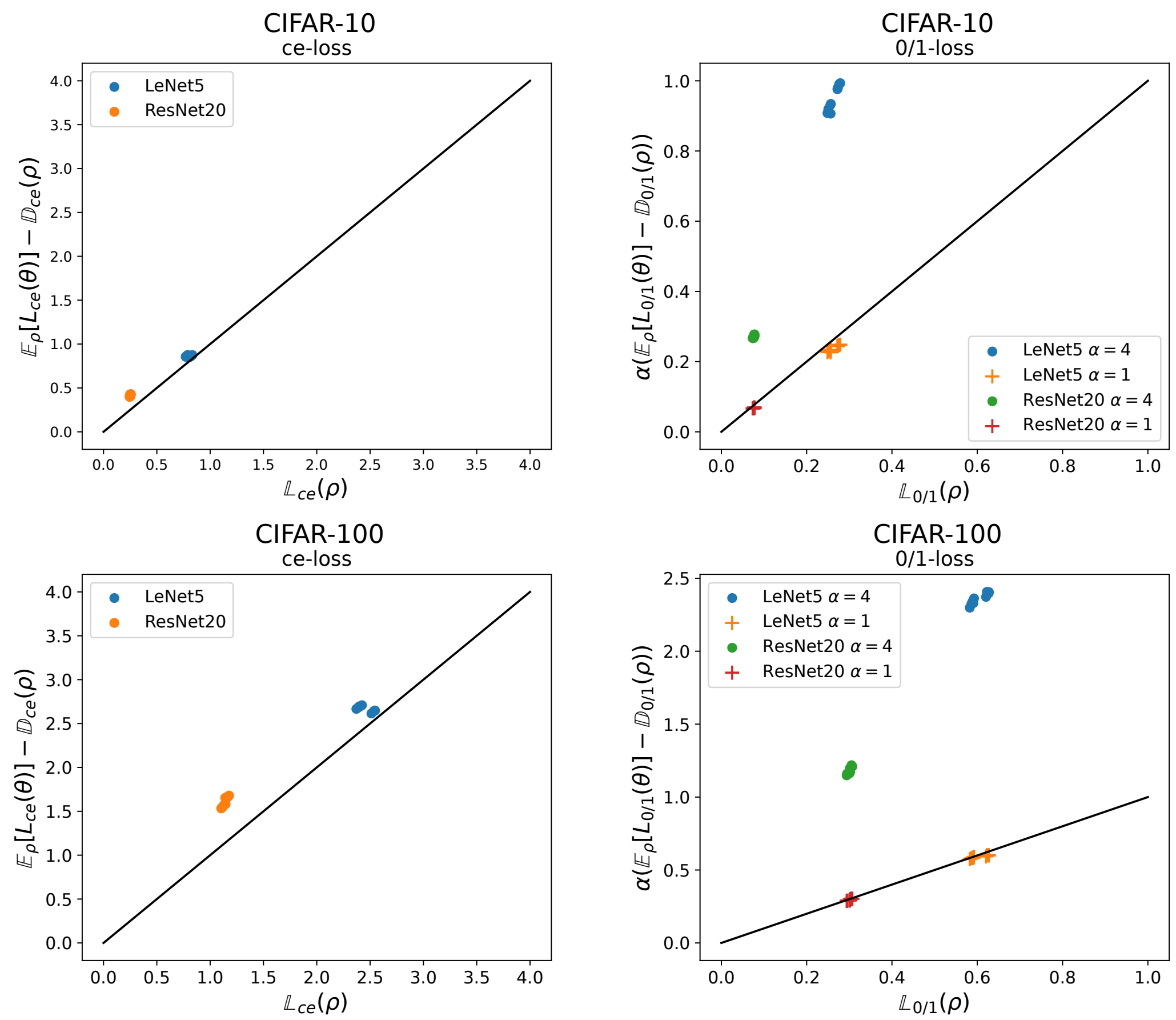

**Figure 5.1:** ***Scatter of empirical ensemble loss against the theoretical upper bound of Theorem 5.1.*** *Each dot represents a single ensemble obtained with either the* Ensemble *or* PAC2B-Ensemble *algorithm; the coordinates are* $(\alpha(\mathbb{E}_\rho[L(\theta)] - D(\rho)), L(\rho))$. *The solid diagonal denotes equality; vertical departures from this line quantify how tight the bound is for that ensemble.*

Section 5.2.3.4 explains why the *P2B-Ensemble* algorithm—obtained by minimizing the PAC-Bayesian bound of Theorem 5.7—yields superior ensembles when the individual networks do not operate in the interpolation regime. The empirical results in Figure 5.3 corroborate this analysis.

- The *P2B-Ensemble* algorithm effectively induces ensembles with higher diversity that have better generalization performance than the *Ensemble* algorithm for MLP50 and LetNet5.
- However, *P2B-Ensemble* on ResNet20 does not induce ensembles with much higher diversity than the *Ensemble* algorithm. Moreover, it performs similar or worst than the *Ensemble* algorithm.

Figure 5.4 shows that the empirical error and empirical diversity terms of the bound of Theorem 5.7 for the ResNet20 ensemble are much smaller than for the LeNet5 ensem-ble. In that case, ResNet20 is operating close to the interpolation regime for CIFAR-10 and close to the interpolation regime for CIFAR-100.

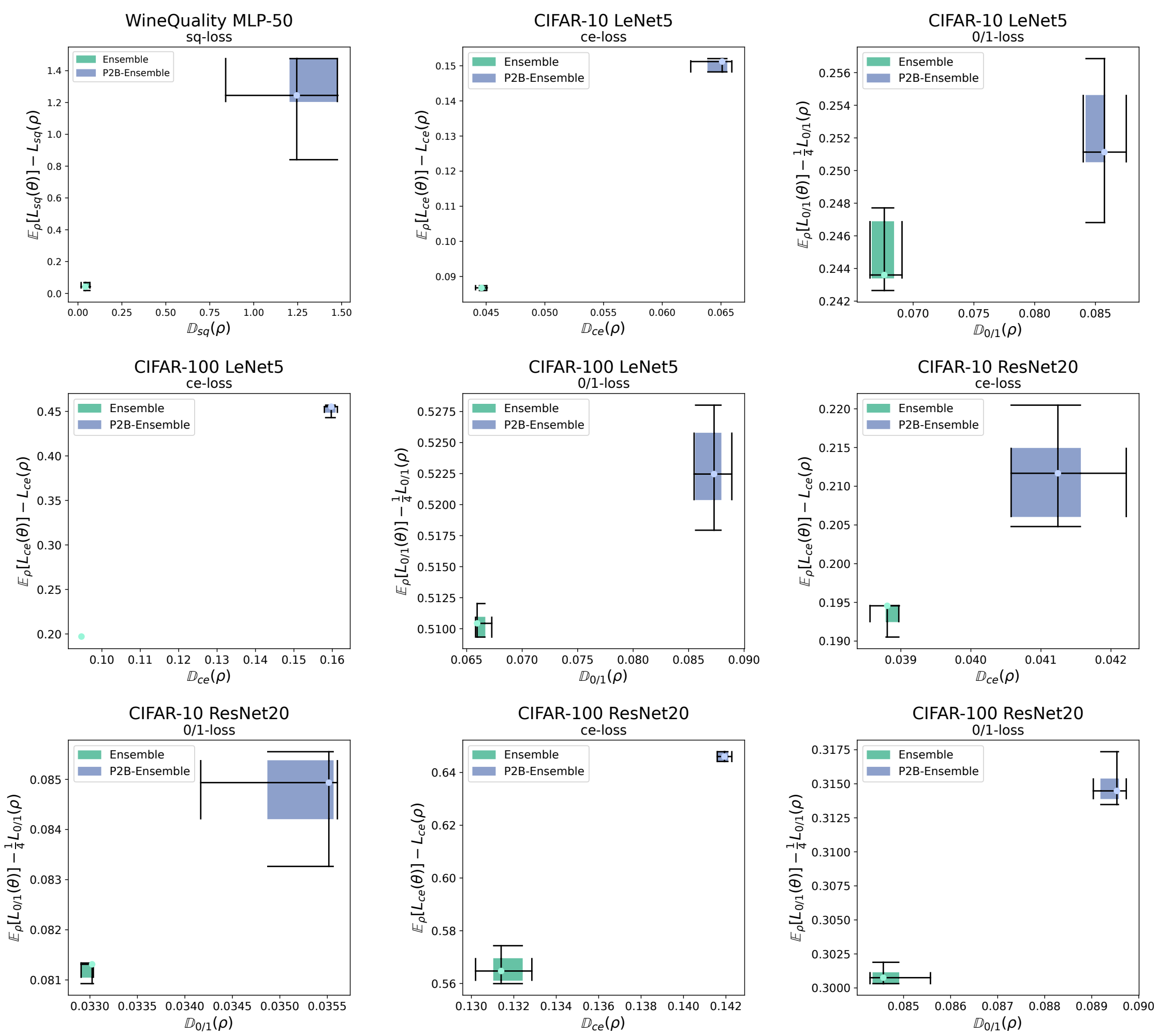

**Figure 5.2:** ***Box-plots of ensemble diversity versus gap between individual and ensemble losses.*** *For each learning algorithm a box-plot summarizes the ensembles it produced; the abscissa is the empirical diversity $D(\rho)$, while the ordinate shows $\mathbb{E}_\rho[L(\theta)] - L(\rho)$ (positive values indicate a lower ensemble loss).*

#### 5.2.4.2 Experiments with different ensemble sizes

Figure 5.5 extends Figure 5.2 by considering multiple ensemble sizes. Consistent with Corollary 5.5, ensembles exhibiting larger diversity $\mathbb{D}(\rho)$ also display a larger gap between the average performance of the individual models, $\mathbb{E}_\rho[L(\boldsymbol{\theta})]$, and the performance of the ensemble, $L(\rho)$. The figure further shows that, for LeNet5 and MLP-50, *Ensemble* does not obtain a substantial increase in diversity when the ensemble size grows, whereas for ResNet20, *Ensemble* steadily increases diversity as the ensemble size increases. A plausible explanation is that random initialization is less effective at exploring distinct modes for simpler architectures than for more complex ones. By contrast, *P2B-Ensemble* consistently achieves higher diversity as the ensemble size increases.

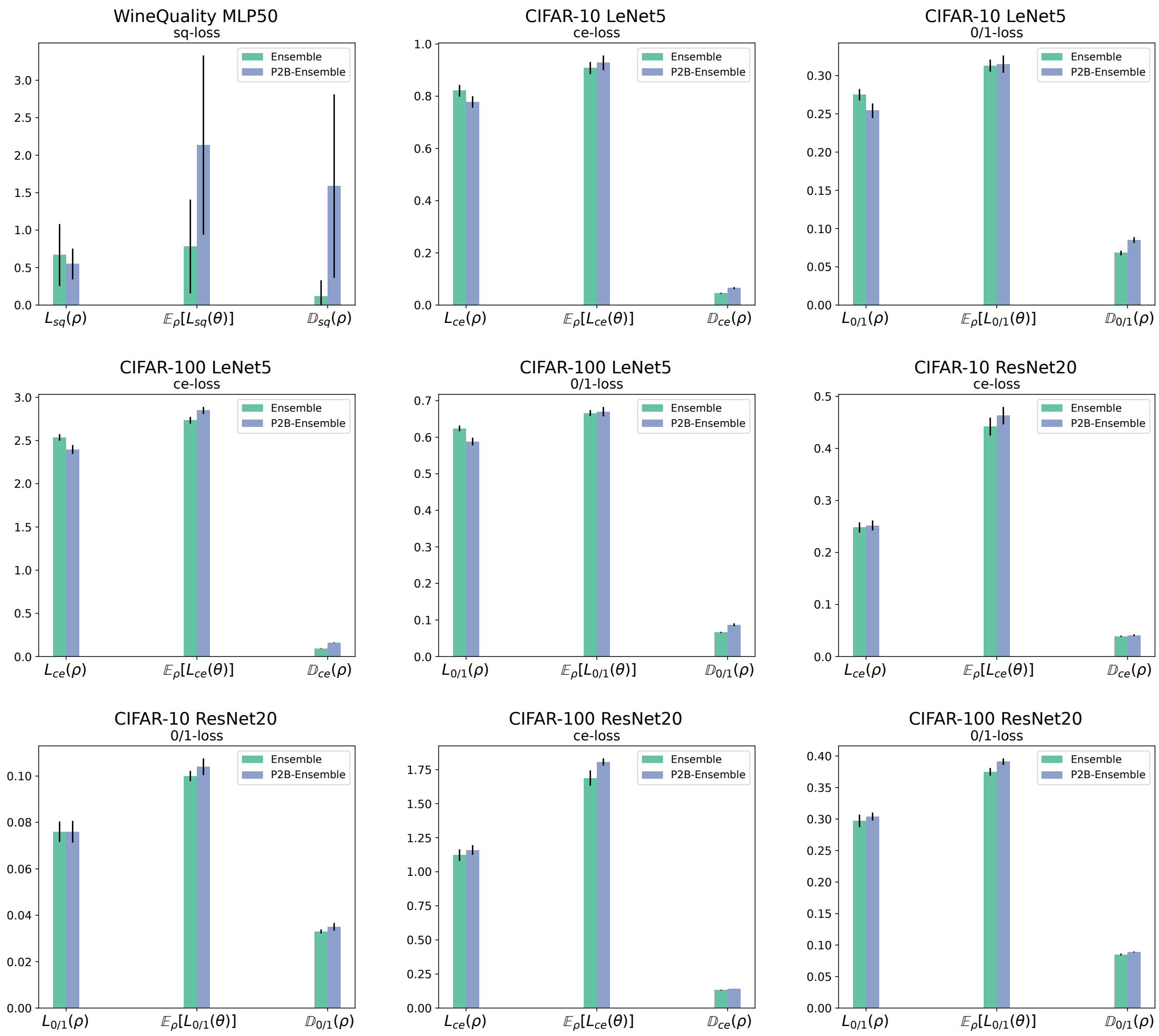

**Figure 5.3:** ***Bar-plots of ensemble error, mean individual error and diversity for nine dataset–architecture–loss combinations.*** *The figure is arranged as a* $3 \times 3$ *grid. Each panel title specifies the dataset, network architecture and loss type—WineQuality/MLP-50 with squared loss (top left); CIFAR-10 or CIFAR-100 with LeNet-5 or ResNet-20 using cross-entropy (ce-loss) or* 0–1 *loss (remaining panels). Within every panel, three bar groups depict (from left to right) the ensemble test error* $L_{sq/ce/0\text{-}1}(\rho)$*, the average test error of the ensemble members* $\mathbb{E}_\rho[L_{sq/ce/0\text{-}1}(\theta)]$*, and the ensemble diversity* $\mathbb{D}_{sq/ce/0\text{-}1}(\rho)$*. For each quantity two colored bars are shown: green for* Ensemble *training and blue for* P2B-Ensemble*. Bar heights give the mean over five random seeds; thin vertical whiskers indicate* $\pm 3$ *standard deviations.*

Figure 5.6 reports the generalization performance of ensembles learned with *Ensemble* and *P2B-Ensemble* for ensemble sizes from two to five. For LeNet5, increasing the number of models degrades ensemble performance, in contrast to ResNet20 and MLP-50, where performance improves with larger ensembles. A convincing explanation for this phenomenon is not currently available.

#### 5.2.4.3 Evaluation of Corollary 5.6

Finally, evidence is provided regarding the usefulness of Corollary 5.6 for determining when an ensemble surpasses a single model. Figure 5.7 reports how often the condition of Corollary 5.6 is satisfied by ensembles trained with *Ensemble*, the standard ensemble-learning algorithm. Consequently, those ensembles meeting the condition outperform the best individual model. Points below the line correspond to ensembles that satisfy the condition. As shown for the $ce$-loss on CIFAR-10 and CIFAR-100, all *Ensemble* models meet the condition and outperform the best individual model, indicating that random initialization constitutes an effective mechanism for learning high-quality ensembles. By contrast, for the $sq$-loss on Wine-Quality with MLP-50, *Ensemble* often fails to produce ensembles that generalize better than the best individual model. As illustrated in Figure 5.2, this shortfall arises because *Ensemble* yields ensembles with very low diversity.

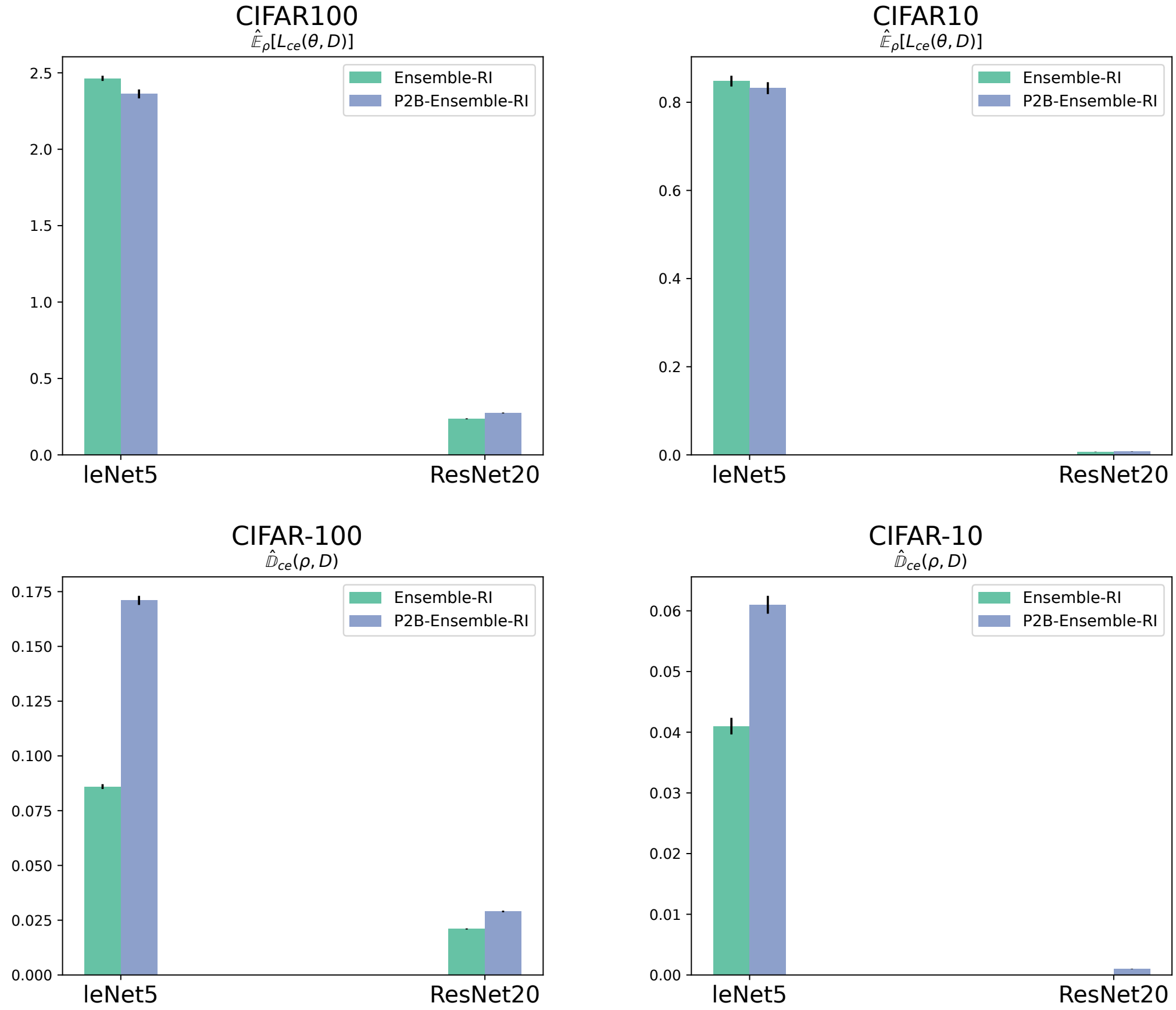

**Figure 5.4:** ***Bar-charts of empirical individual error and empirical diversity for LeNet-5 and ResNet-20 on CIFAR-10 and CIFAR-100.*** *The figure contains four panels whose titles specify the dataset and the metric used* ($\mathbb{E}_\rho[\hat{L}_{\text{ce}}(\boldsymbol{\theta}, D)]$ *or* $\hat{D}_{\text{ce}}(\rho, D)$). *Within each panel the horizontal axis lists the two architectures, and for every architecture two coloured bars report the mean across five random seeds* (Ensemble-RI *in green and* P2B-Ensemble-RI *in blue*)*; thin vertical lines denote* $\pm 3$ *standard deviations.*

For the 0-1 loss, the same ensembles trained with the $ce$-loss are evaluated (recall that majority-vote ensembles are trained with $ce$). In this case, no model satisfies the condition of Corollary 5.6 under the bound with $\alpha = 4$. Nevertheless, the ensemble consistently outperforms the best single model empirically. The worst-case constant $\alpha = 4$ substantially weakens the bound and limits its applicability. Setting $\alpha = 1$—acknowledging that this choice does not yield an upper bound—resolves the issue in the sense that all models then satisfy the corollary's condition, suggesting the possibility of tighter constants closer to 1.

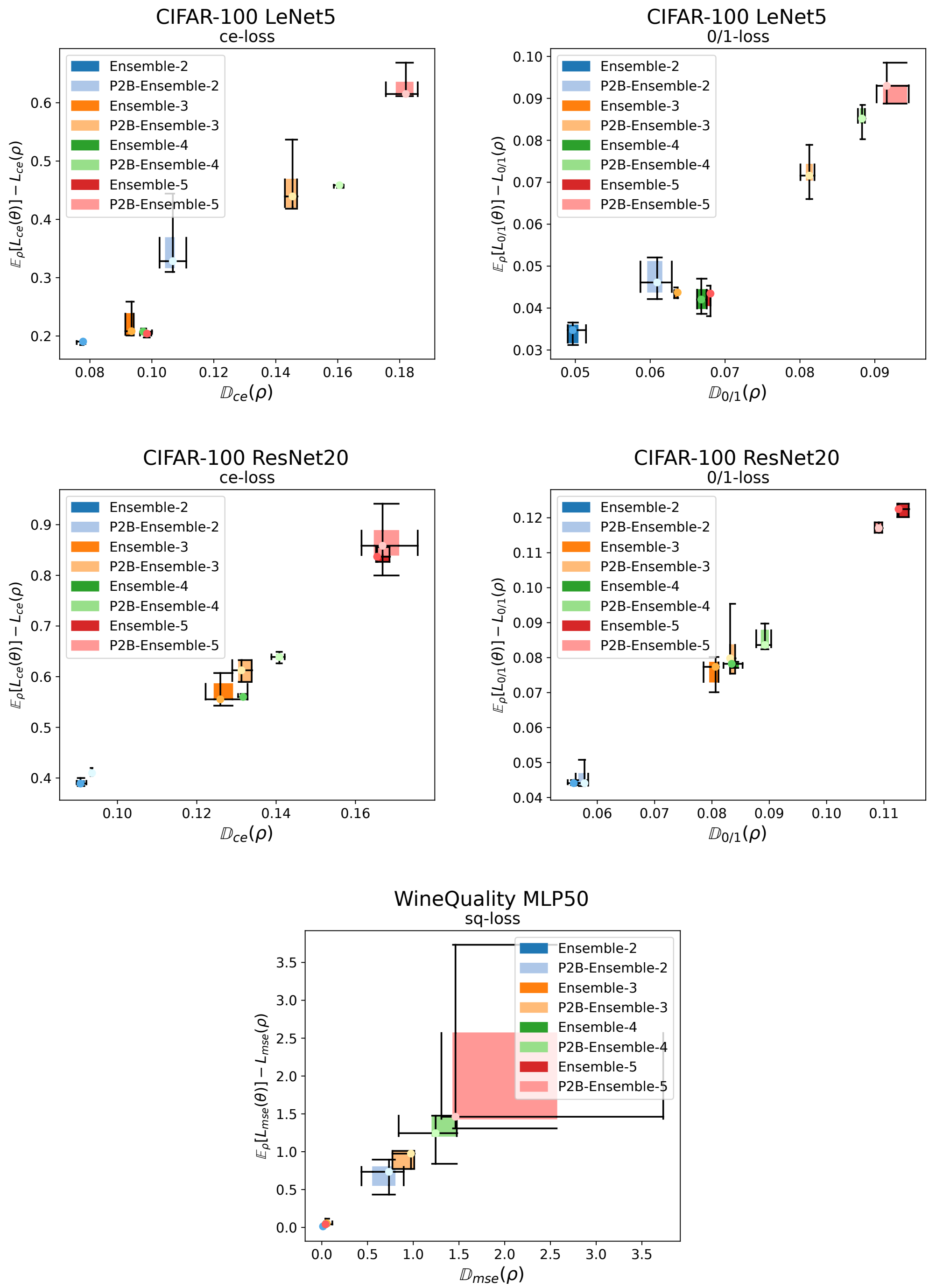

**Figure 5.5:** ***Box-plots of ensemble diversity versus gap between member and ensemble losses across datasets, networks and ensemble sizes.*** *The figure comprises five panels—CIFAR-100/LeNet-5 with cross-entropy loss, CIFAR-100/LeNet-5 with 0-1 loss, CIFAR-100/ResNet-20 with cross-entropy loss (top row), and CIFAR-100/ResNet-20 with 0-1 loss plus WineQuality/MLP-50 with squared loss (bottom row). Within every panel each colored box-plot represents a collection of ensembles trained with the same algorithm (*Ensemble *or* P2B-Ensemble*) and the same size as indicated in the legend. The abscissa records the empirical diversity $\mathbb{D}(\rho)$ computed under the loss specified in the panel title, while the ordinate shows the difference $\mathbb{E}_\rho[L(\boldsymbol{\theta})] - L(\rho)$ between the average individual test loss and the ensemble test loss. For each ensemble set the box spans the inter-quartile range, the central line marks the median, and whiskers extend to the minimum and maximum values observed over five random seeds.*

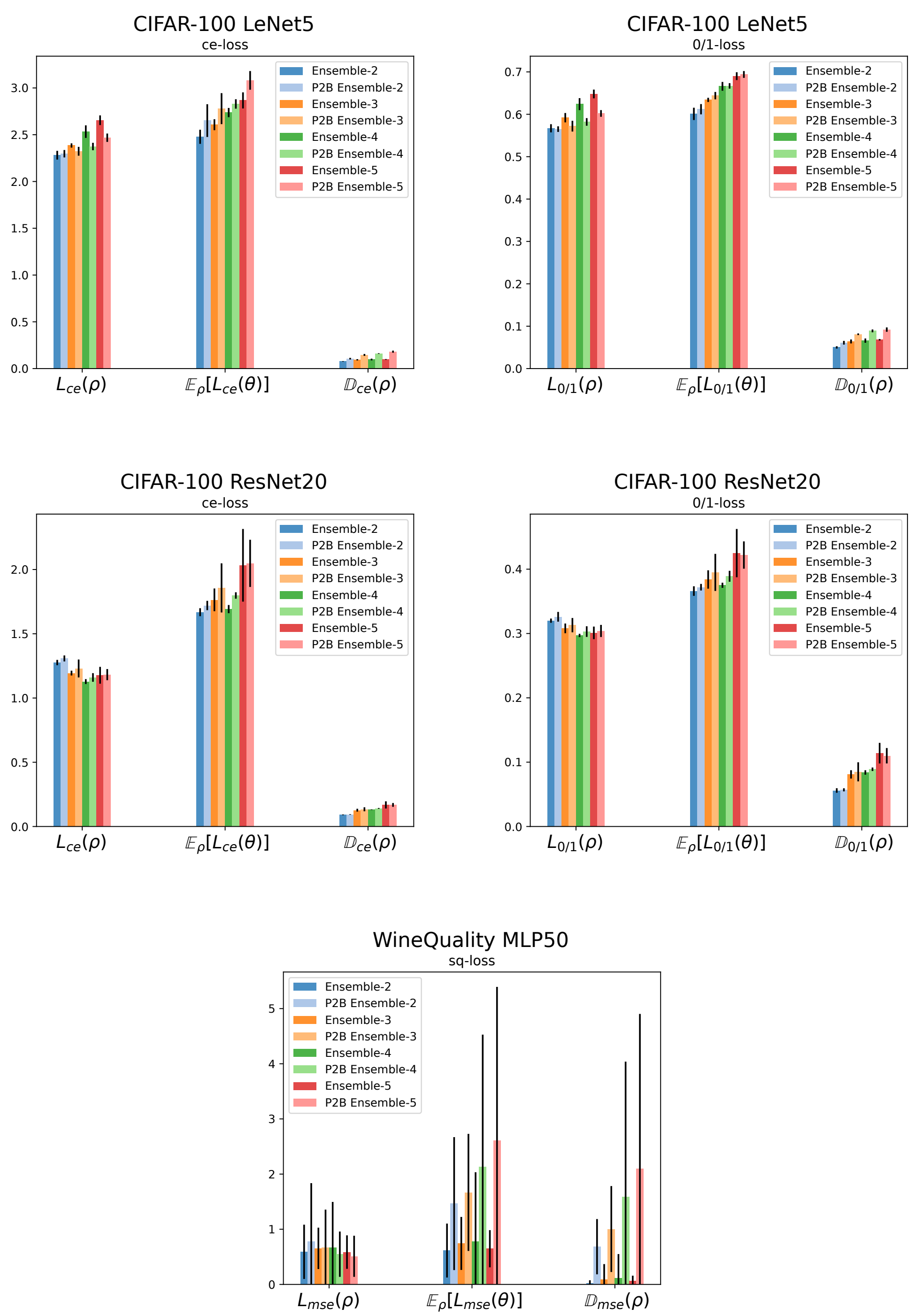

**Figure 5.6:** ***Bar-plots of ensemble, member and diversity statistics for multiple ensemble sizes.*** *The figure comprises five subplots: (top row, left to right) CIFAR-100 with LeNet-5 evaluated under cross-entropy loss and CIFAR-100 with LeNet-5 under 0-1 loss; (middle row, left to right) CIFAR-100 with ResNet-20 under cross-entropy loss and CIFAR-100 with ResNet-20 under 0-1 loss; (bottom row) WineQuality with MLP-50 under squared loss. Within every subplot three groups of bars visualize (from left to right) the ensemble test loss $L(\rho)$, the average test loss of the ensemble members $\mathbb{E}_\rho[L(\theta)]$ and the ensemble diversity $\mathbb{D}(\rho)$. For each quantity eight colored bars are shown-one pair for every ensemble size: the solid bar corresponds to the* Ensemble *algorithm and the adjacent hatched bar to* P2B-Ensemble*. Bar heights give the mean over five random seeds, and black whiskers indicate $\pm 3$ standard deviations.*

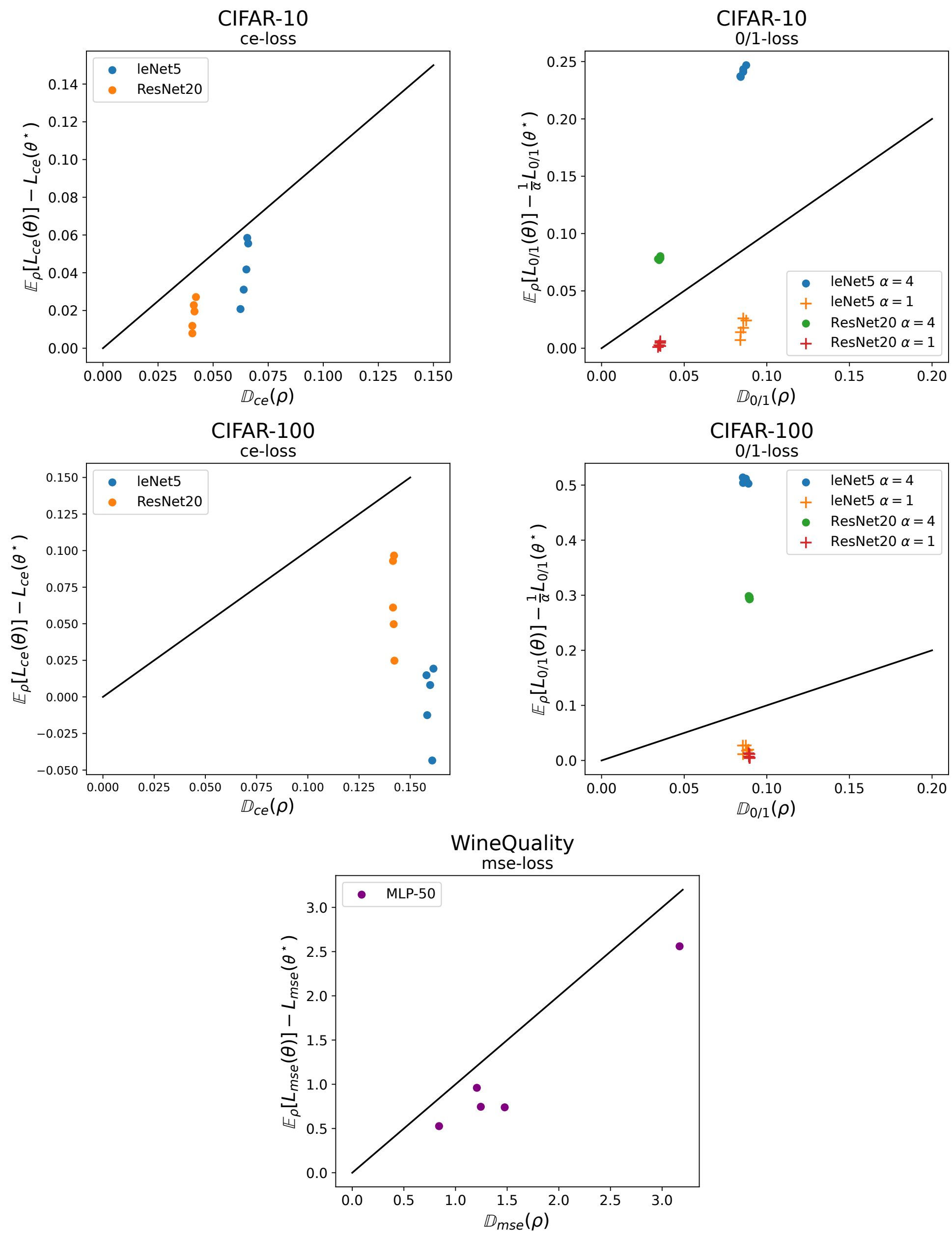

**Figure 5.7:** ***Scatter-plots of empirical diversity versus average-to-best individual loss gap for five dataset–loss settings.*** *The figure comprises five panels: CIFAR-10 with cross-entropy loss, CIFAR-10 with* $0-1$ *loss (top row, left to right), CIFAR-100 with cross-entropy loss, and CIFAR-100 with* $0-1$ *loss (middle row, left to right), WineQuality with mean-squared error (bottom row). In every panel each dot corresponds to a single ensemble learned with the* Ensemble *algorithm; the horizontal axis is the empirical diversity* $\mathbb{D}(\rho)$ *under the plotted loss, while the vertical axis shows the difference* $\mathbb{E}_\rho[L(\theta)] - L(\theta^\star)$*, where* $\theta^\star$ *denotes the lowest-loss individual model within that ensemble. Colors distinguish network architectures (LeNet-5, ResNet-20 or MLP-50) and marker shapes encode ensemble sizes indicated in the legend. The solid black diagonal represents the line* $y = \mathbb{D}(\rho)$*; points relative to this reference illustrate where the measured gap equals the empirical diversity.*

### 5.2.5 How to Exploit Diversity to Learn Ensembles

This section explains how to leverage diversity—guided by PAC-Bayesian analysis—to construct and optimize ensemble models.

#### 5.2.5.1 Working with a Finite Parameter Space

The assumption of a finite parameter space is not restrictive in this setting. Let $\boldsymbol{\Theta} = \{\boldsymbol{\theta}_1, \ldots, \boldsymbol{\theta}_K\}$ with potentially very large $K$. In principle, $\boldsymbol{\Theta}$ could contain all finite-precision vectors of dimension $M$. In any case, the distribution $\rho$ assigns positive probability only to those models that constitute the ensemble, so that only a sparse subset of $\boldsymbol{\Theta}$ receives nonzero mass.

#### 5.2.5.2 Working with a Continuous Parameter Space

The only point at which adopting a continuous parameter space affects the argumentation is in Lemma 5.3, specifically in the reparameterization-invariance property of the diversity measures. If $\boldsymbol{\Theta}$ is a continuous, non-compact set such as $\mathbb{R}^M$ for some $M \in \mathbb{N}$, the change-of-variables theorem cannot be invoked directly. To circumvent this issue, consider the compact truncation:

$$\boldsymbol{\Theta} = \{\boldsymbol{\theta} \in \mathbb{R}^M : \|\boldsymbol{\theta}\|_2 \leq N\} \subset \mathbb{R}^M, \tag{5.28}$$

where $N$ denotes the largest $\ell_2$ norm attainable under the chosen finite-precision representation. This restriction renders the requisite transformations valid while leaving the set of implementable models unchanged in practice.

#### 5.2.5.3 Mixtures of multivariate Gaussian Distributions approximation

As detailed in Section 5.2.3.4, consider a uniformly weighted Gaussian mixture, denoted by $\rho_\delta$, to represent an ensemble of $K$ models. The distribution over parameters, given a fixed set of means $(\boldsymbol{\theta}_1, \ldots, \boldsymbol{\theta}_K)$, is

$$\rho_\delta(\boldsymbol{\theta}) = \frac{1}{K} \sum_{k=1}^{K} \mathcal{N}(\boldsymbol{\theta};\, \boldsymbol{\theta}_k,\, \epsilon I). \tag{5.29}$$

Under this distribution, the expected empirical loss is

$$\mathbb{E}_{\rho_\delta}[\hat{L}(\boldsymbol{\theta}, D)] = \int \rho_\delta(\boldsymbol{\theta})\, \hat{L}(\boldsymbol{\theta}, D)\, d\boldsymbol{\theta} \tag{5.30}$$

$$= \frac{1}{K}\sum_{k=1}^{K} \int \mathcal{N}(\boldsymbol{\theta};\, \boldsymbol{\theta}_k,\, \epsilon I)\, \hat{L}(\boldsymbol{\theta}, D)\, d\boldsymbol{\theta} \tag{5.31}$$

$$= \frac{1}{K}\sum_{k=1}^{K} \mathbb{E}_{\mathcal{N}(\boldsymbol{\theta}_k, \epsilon I)}\big[\hat{L}(\boldsymbol{\theta}, D)\big]. \tag{5.32}$$

When $\epsilon$ is sufficiently small and $\hat{L}$ is smooth, the expectation under the sharply concentrated Gaussian can be approximated by the loss evaluated at the mean. More precisely,

$$\int_{\boldsymbol{\theta}} \mathcal{N}(\boldsymbol{\theta}; \boldsymbol{\theta}_k, \epsilon I) f(\boldsymbol{\theta}) \approx f(\boldsymbol{\theta}_k) \quad \forall k = 1, \ldots, K. \tag{5.33}$$

Using that, the expected value of the loss can be approximated as

$$\mathbb{E}_{\rho_\delta}[\hat{L}(\boldsymbol{\theta}, D)] = \frac{1}{K}\sum_{k=1}^{K} \int_{\boldsymbol{\theta}} \mathcal{N}(\boldsymbol{\theta};\ \boldsymbol{\theta}_k, \epsilon I) \hat{L}(\boldsymbol{\theta}, D) \approx \frac{1}{K}\sum_{k=1}^{K} \hat{L}(\boldsymbol{\theta}_k, D). \tag{5.34}$$

The same argument applies to the KL regularizer. Using a non-overlapping approximation for the Gaussian mixture components,

$$\sum_{i=1}^{K} \mathcal{N}(\boldsymbol{\theta}_k;\, \boldsymbol{\theta}_i,\, \epsilon I) \;\approx\; \mathcal{N}(\boldsymbol{\theta}_k;\, \boldsymbol{\theta}_k,\, \epsilon I) = \frac{1}{(2\pi)^{M/2} \epsilon^{M/2}}, \quad \forall\, k \in \{1, \ldots, K\}, \tag{5.35}$$

where $M$ is the dimensionality of $\boldsymbol{\theta}$. The rationale is that the means $(\boldsymbol{\theta}_1, \ldots, \boldsymbol{\theta}_K)$ are sufficiently separated so that, for any distinct pair $(\boldsymbol{\theta}_i, \boldsymbol{\theta}_k)$, evaluating the Gaussian centered at $\boldsymbol{\theta}_i$ with covariance $\epsilon I$ at $\boldsymbol{\theta}_k$ is negligible. This assumption is mild because $\epsilon$ can be chosen small; for example, enforcing $\min_{i \neq k} \|\boldsymbol{\theta}_i - \boldsymbol{\theta}_k\|_2 \geq c\sqrt{\epsilon}$ with a constant $c$ (e.g., $c = 3$) renders component overlap negligible (under the "$3\sigma$" heuristic, over $99.7\%$ of the mass lies within $3\sqrt{\epsilon}$ along any coordinate).

As a result, the following approximation arises for the regularization term:

$$KL(\rho_\delta|\pi) = \int_{\boldsymbol{\theta}} \rho_\delta(\boldsymbol{\theta}) \log \frac{\rho_\delta(\boldsymbol{\theta})}{\pi(\boldsymbol{\theta})} \tag{5.36}$$

$$= \frac{1}{K}\sum_{k=1}^{K} \int_{\boldsymbol{\theta}} \mathcal{N}(\boldsymbol{\theta};\ \boldsymbol{\theta}_i, \epsilon I) \left( \log \frac{1}{K}\sum_{i=1}^{K} \mathcal{N}(\boldsymbol{\theta};\ \boldsymbol{\theta}_i, \epsilon I) - \log \pi(\boldsymbol{\theta}) \right) \tag{5.37}$$

$$\text{(by Equation (5.33))} \approx \frac{1}{K}\sum_{k=1}^{K} \left( \log \frac{1}{K}\sum_{i=1}^{K} \mathcal{N}(\boldsymbol{\theta}_k;\ \boldsymbol{\theta}_i, \epsilon I) - \log \pi(\boldsymbol{\theta}_k) \right) \tag{5.38}$$

$$\text{(by Equation (5.35))} \approx \frac{1}{K}\sum_{k=1}^{K} \left( \log \frac{1}{K}\mathcal{N}(\boldsymbol{\theta}_k;\ \boldsymbol{\theta}_k, \epsilon I) - \log \pi(\boldsymbol{\theta}_k) \right) \tag{5.39}$$

$$= -\frac{1}{K}\sum_{k=1}^{K} \log \pi(\boldsymbol{\theta}_k) + \frac{1}{K}\sum_{k=1}^{K} \log \frac{1}{K} \frac{1}{\sqrt{(2\pi)^M \epsilon^M}}\,. \tag{5.40}$$

The same approximation given by Equation (5.33) can be applied to the general variance formula:

$$\hat{\mathbb{V}}_{\rho_\delta}(f(\boldsymbol{\theta})) = \mathbb{E}_{\rho_\delta^2}\left[f(\boldsymbol{\theta})^2 - f(\boldsymbol{\theta})f(\boldsymbol{\theta}')\right] = \mathbb{E}_{\rho_\delta}\left[f(\boldsymbol{\theta})^2\right] - \mathbb{E}_{\rho_\delta^2}\left[f(\boldsymbol{\theta})f(\boldsymbol{\theta}')\right] \tag{5.41}$$

$$= \frac{1}{K}\sum_{k=1}^{K} f(\boldsymbol{\theta}_k)^2 - \frac{1}{K^2}\sum_{i=1}^{K}\sum_{j=1}^{K} f(\boldsymbol{\theta}_i)f(\boldsymbol{\theta}_j). \tag{5.42}$$

Given this, it is easy to approximate each of the diversity terms defined in Theorem 5.1. Where $f$ is determined by the considered loss function: for the $sq$-loss, $f(\boldsymbol{\theta}) = h_R(\mathbf{x};\boldsymbol{\theta})$, the $ce$-loss, $f(\boldsymbol{\theta}) = p(y|\mathbf{x},\boldsymbol{\theta})$, and the $0/1$-loss $f(\boldsymbol{\theta}) = \mathbb{I}(h(\mathbf{x};\boldsymbol{\theta}) \neq y)$.

#### 5.2.5.4 Ensemble Learning Algorithms Which Explicitly Promote Diversity

**Negative Correlation Learning (Y. Liu and Yao, 1999).** This method minimizes

$$\mathbb{E}_\rho[\hat{L}_{sq}(\boldsymbol{\theta}, D)] \;-\; \lambda\,\mathrm{NC}(\rho, D), \tag{5.43}$$

where $\lambda \in [0,1]$, $\mathbb{E}_D[\cdot]$ denotes expectation with respect to the empirical data distribution, and

$$\mathrm{NC}(\rho, D) := \mathbb{E}_D\left[\frac{1}{K}\sum_{k=1}^{K}(h_R(\mathbf{x};\boldsymbol{\theta}_k) - h_R(\mathbf{x};\rho))\sum_{j\neq k}(h_R(\mathbf{x};\boldsymbol{\theta}_j) + h_R(\mathbf{x};\rho))\right] \tag{5.44}$$

is the empirical negative correlation term that explicitly promotes diversity. By straightforward algebra,

$$\mathrm{NC}(\rho, D) \;=\; -\hat{\mathbb{D}}_{sq}(\rho, D). \tag{5.45}$$

Hence, the objective becomes $\mathbb{E}_\rho[\hat{L}_{sq}(\boldsymbol{\theta}, D)] + \lambda\hat{\mathbb{D}}_{sq}(\rho, D)$, which coincides with the learning objective derived from the PAC-Bayesian analysis for the squared loss. This connection yields a PAC-Bayesian interpretation of the negative correlation ensemble learning algorithm (Y. Liu and Yao, 1999).

**Generalized Ambiguity Decomposition (Z. Jiang et al., 2017).** This approach extends the decomposition of Krogh and Vedelsby (1994) to more general loss functions, expressing ensemble risk in terms of individual-model errors and an ensemble "diversity" component. Unlike the present PAC-Bayesian treatment, the resulting decomposition is not derived from upper bounds and coincides with the formulation developed here only for the squared loss. Moreover, it does not relate the ensemble's generalization error to empirical diversity, does not address multiclass classification, and the decomposition for the $0/1$-loss does not include a diversity term. Weighted majority votes are likewise not considered.

**Generalized Negative Correlation Learning (Buschjäger et al., 2020).** Building on Z. Jiang et al. (2017), this approach proposes a learning objective that incorporates a diversity term, although the specific form of the term differs from those introduced here. The analysis does not treat the 0-1 loss or weighted majority votes. Ultimately, the authors recommend the alternative objective:

$$\lambda \hat{L}(\rho, D) + (1-\lambda)\, \mathbb{E}_{\rho}\Big[\hat{L}(\boldsymbol{\theta}, D)\Big], \qquad \lambda \in [0,1], \tag{5.46}$$

which trades off the empirical ensemble loss against the average empirical loss of the constituent models without explicitly retaining a diversity regularizer.

#### 5.2.5.5 Standard Learning Algorithms Do Not Explicitly Promote Diversity

Standard ensemble learning algorithms can be interpreted as methods trying to minimize the following objective function:

$$\mathbb{E}_{\rho_\delta}[\hat{L}(\boldsymbol{\theta}, D)] + \frac{KL(\rho|\pi)}{\lambda n} \tag{5.47}$$

where either the $ce$-loss or the $sq$-loss is employed.

This learning objective does not include the diversity term, $\hat{\mathbb{D}}(\rho, D)$, encouraging diversity. In fact, under the approximations discussed in Section 5.2.5.3 and discarding constant terms, this learning objective can be expressed as:

$$\frac{1}{K}\sum_k \hat{L}(\boldsymbol{\theta}_k, D) - \frac{\ln \pi(\boldsymbol{\theta}_k)}{\lambda n} \tag{5.48}$$

where each $\boldsymbol{\theta}_k$ can be learned independently from the rest due to the presence of $\hat{\mathbb{V}}(\rho, D)$.

### 5.2.6 Discussion

Despite the widely recognized importance of diversity for ensemble performance, prior work has not provided a theoretically grounded, broadly applicable account of the relationship between diversity and generalization across a wide range of ensemble models. The present study contributes toward filling this gap.

Theorem 5.1 demonstrates that an upper-bound-based decomposition of ensemble error furnishes a general and practical approach applicable to multiple ensemble formulations. In particular, the analysis clarifies how correlations among predictors influence ensemble diversity. This decomposition also enables a PAC-Bayesian treatment that establishes a direct link between an ensemble's empirical diversity and its generalization performance, and it sheds light on why ensembles of deep neural networks can effectively promote diversity through randomization strategies.

Although several ingredients of the analysis are known in the literature, the main contribution lies in unifying these elements into a coherent framework for reasoning about diversity and generalization in neural network ensembles.

The theoretical development has limitations that suggest avenues for further research. First, the upper bounds in Theorem 5.1 may admit tighter variants; for example, C-bounds (Germain et al., 2015) are known to improve upon the $0/1$-bound (see Masegosa et al., 2020 for discussion). A similar observation applies to the PAC-Bayesian bounds used in Theorem 5.7, which inherit looseness from the underlying decomposition. Moreover, the analysis focuses on a general form of PAC-Bayesian bounds; more specialized and tighter alternatives could be adopted, as exemplified by Masegosa et al. (2020).

## 5.3 Generalization Error and Chernoff Bounds

In modern machine learning, model classes have such large capacity that optimizers virtually always retrieve a model *interpolating* the training data; that is, with null training error (C. Zhang et al., 2017). These models are called *interpolators* and it is well known that there are many within a large model class (Livni et al., 2014) and that some of them have a remarkably small generalization error (difference between "training error" and "expected or test error") while others do not (Feldman and Zhang, 2020).

The machine learning community has made a great effort during the last years to understand why modern learning algorithms retrieve, most of the time, interpolators with a small generalization error (Bartlett et al., 2020; Nagarajan and Kolter, 2019b). Most of these attempts often focus on providing generalization bounds. These bounds provide an upper limit on the expected error, tying it to variables related to the training dataset and the model produced by a learning algorithm. Such bounds typically resemble the following form,

$$L(\mathcal{A}(D)) \leq \hat{L}(\mathcal{A}(D), D) + \mathcal{C}(\mathcal{A}(D), D, \delta)\,, \tag{5.49}$$

where $D$ represents the training dataset, independently and identically distributed (i.i.d.) from a data-generating distribution $\nu$; $\mathcal{A}$ denotes the learning algorithm, which generates a hypothesis (a model) from a given dataset (for example, via the gradient descent method); $L$ and $\hat{L}$ indicate the expected and empirical errors of a hypothesis, respectively; $\mathcal{C}$ denotes a complexity measure; and the inequality holds with high probability, at least $1-\delta$ over draws of datasets from the distribution $\nu$. A detailed introduction to classical PAC and PAC–Bayesian bounds of this type is given in Section 2.4.

Examples of bounds falling inside this scheme include Vapnik–Chervonenkis (VC) bounds (Bartlett et al., 2019; Vapnik and Chervonenkis, 2015), Rademacher bounds (Mohri et al., 2018), (Norm and margin)-based bounds (Nagarajan and Kolter, 2017; Neyshabur et al., 2015a) and Sharpness-based measures (Keskar et al., 2017; Nagarajan and Kolter, 2019a). Even PAC-Bayes with a fixed prior (McAllester, 1999) can be included here as algorithms that retrieve a distribution over the models.

However, many recent works are increasingly suggesting that these bounds are probably vacuous in the over-parameterized regime and, in consequence, unable to explain generalization in this setting. C. Zhang et al., 2017 was the first one to provide quite convincing empirical evidence of this in the context of deep neural networks. Nagarajan and Kolter, 2019b later provided a series of experiments and theoretical results in the same direction. These works also show that even in the case where the complexity measure $\mathcal{C}$ depends directly on the algorithm $\mathcal{A}$ rather than simply on its output $\mathcal{A}(D)$ (for example, assuming the algorithm always retrieves models whose parameters' norm is lower than a given constant) it is also vacuous. Recently, Y. Wang et al., 2024 showed how any near-interpolator exhibits rapid parameter norm growth, which implies that existing data-dependent parameter-norm-based bounds are necessarily loose, concluding that *"explaining the generalization capability of near-interpolators will require new tools"*. More generally, Gastpar et al., 2024 managed to show that, under some conditions resembling over-parameterization, there is no generalization bound like Equation (5.49) tight for all data-generating distributions, and that algorithm-dependent bounds are also provably vacuous. For all these reasons, the emerging conclusion is that: *bounds that solely depend on the training data are provably vacuous for over-parameterized model classes and are unable to explain generalization.*

The aforementioned works (Gastpar et al., 2024; Nagarajan, 2021; Nagarajan and Kolter, 2019b) directly or indirectly advocate for the need of exploring alternative bounds that not solely depend on the training data but also use information from the data-generating distribution. These bounds are usually referred to as *distribution-dependent bounds*; where the complexity term $\mathcal{C}$ explicitly depends on the data-generating distribution $\nu$. Distribution-dependent bounds have been used before in different contexts (Catoni, 2007; T. Zhang, 2006); however, to the best of current knowledge, none of these bounds are known to be tight or, at least, non-vacuous for over-parameterized interpolators. Even more importantly, these kinds of bounds have not been used before to analyze the generalization of over-parameterized interpolators.

### Contribution

In this thesis, the use of distribution-dependent bounds is studied within a finite hypothesis space. The goal is to understand how modern learning techniques manage to produce models that both interpolate the data and have a small generalization error. More precisely, Theorem 5.17 introduces a distribution-dependent PAC-Chernoff bound that is *perfectly tight* for any algorithm retrieving models interpolating the training data, even when the model class is over-parameterized. This bound has the following standard form:

$$L(\mathcal{A}(D)) \leq \hat{L}(\mathcal{A}(D), D) + \mathcal{C}(\mathcal{A}(D), n, \nu, \delta)\,, \tag{5.50}$$

where, now, the complexity term $\mathcal{C}$ also depends on the data-generating distribution $\nu$. Proposition 5.20 shows that with high probability $1-\delta$,

$$\mathcal{C}(\mathcal{A}(D), n, \nu, \delta) \leq L(\mathcal{A}(D)) \leq \hat{L}(\mathcal{A}(D), D) + \mathcal{C}(\mathcal{A}(D), n, \nu, \delta)\,, \tag{5.51}$$

which ensures that the proposed bound is *perfectly tight for interpolators*, that is to say, for models whose empirical loss is null or small enough to be considered negligible or within an acceptable error margin, denoted as $\hat{L}(\mathcal{A}(D), D) \leq \epsilon$.

As shown in this thesis, these bounds are directly connected to Large Deviation Theory (LDT) (Ellis, 2012) because their complexity measure $\mathcal{C}(\mathcal{A}(D), n, \nu, \delta)$ directly depends on the so-called *rate function* (also known as the Cramér-Chernoff function), which is the central element of LDT. The *rate function* is used to present a new characterization of the *smoothness* of a model using distribution-dependent measures. According to Theorem 5.23, this approach enables a precise characterization of *which interpolators better generalize*, addressing an outstanding open question in machine learning.

Despite being based on oracle quantities, the theoretical framework built around this complexity measure based on the rate function allows creating a unified explanation of a wide range of learning techniques used in modern machine learning. Namely, $\ell_2$-norm, distance from initialization, and input-gradient regularization (Section 5.3.6), invariant architectures, data augmentation (Section 5.3.7) and over-parameterization (Section 5.3.8). Under this framework, this thesis shows why each of these learning techniques produces interpolators that generalize well and why and when they complement each other (for example, why $\ell_2$-norm regularization is also effective in combination with the use of invariant architectures).

The thesis *does not claim* that the proposed theoretical analyses of each of these learning techniques are better than existing ones at the individual level. For example, it is *not claimed* that this explanation about why $\ell_2$-norm promotes generalization is better than existing ones, for example, in the context of linear regression (Bartlett et al., 2020). However, it *is claimed* that using the rate function and PAC-Chernoff bounds, which are distribution-dependent quantities, it is possible to jointly explain, up to some degree, the generalization of interpolators, the role of over-parameterization, and why many widely used modern learning techniques find interpolators that generalize. To the best of current knowledge, this is the first theoretical framework capable of achieving insights that cover such a wide range of learning techniques. At the same time, this work shows that distribution-dependent bounds are a promising research direction for understanding the generalization of modern machine learning methods.

### 5.3.1 Related Work

In the context of uniform convergence bounds, many efforts have been made towards obtaining tighter generalization bounds (Bartlett et al., 2017; Golowich et al., 2018; Kawaguchi et al., 2022; Liang et al., 2019; Neyshabur et al., 2017a), by, for example, only considering the models effectively visited by the optimization algorithm. However, it is clear that these bounds are not actually meaningful (see Nagarajan and Kolter, 2019b for further references and discussions). The PAC-Bayes framework has also adapted to this new paradigm by exploiting properties of the models induced by the training data set (such as low spectral norm (Neyshabur et al., 2017b), noise stability (Arora et al., 2018b), de-randomization (Negrea et al., 2020), and compression (Arora et al., 2018b; W. Zhou et al., 2019)), but none of these bounds are shown to be empirically tight in the over-parameterized model classes and, for example, are unable to describe the interplay between invariant architectures and over-parameterization. In fact, very recently, Gastpar et al., 2024 proved that, for over-parameterized model classes, none of these bounds are tight for all data-generating distributions.

This thesis mainly differs from these related works in the sense that an *oracle complexity measure* is used, that is, PAC-Chernoff bounds and the rate function assume access to the data-generating distribution. Consequently, it is possible to tightly bound the generalization error of an interpolator, as elaborated in Section 5.3.4. This capability stems from robust results such as Theorem 5.23 even under unbounded losses. Furthermore, Theorem 5.23 and the PAC-Chernoff bound of Theorem 5.17 enable us to leverage architectural invariances with respect to transformations inherent in the data-generating distribution.

Many other recent theoretical studies try to understand specific learning techniques in deep learning in an isolated and independent way to simplify the problem (e.g., (Arora et al., 2016; Arora et al., 2018a; Bubeck and Sellke, 2023; S. Chen et al., 2020; Gilbert et al., 2017; Gunasekar et al., 2018; Hardt and Ma, 2017; Patel et al., 2016; Poole et al., 2016; Schoenholz et al., 2017; Soudry et al., 2018; Tishby and Zaslavsky, 2015; Vidal et al., 2020)). As a result of such an isolated approach, they are not fully able, most of the time, to establish links among different learning techniques, as done in this work. Furthermore, relying on the insights provided by the field of geometric deep learning (Bronstein et al., 2021), this thesis provides a theoretical explanation, from a statistical inference point of view, of why such invariances induce smoother model classes and better generalization.

### 5.3.2 Preliminaries

To fix notation and assumptions, let $D = \{(\mathbf{x}_i, \mathbf{y}_i)\}_{i=1}^n$ denote a *training dataset* of size $n \geq 1$ drawn i.i.d. from an unknown distribution $\nu(\mathbf{y}, \mathbf{x})$. Consider a model class parameterized by $\boldsymbol{\theta} \in \boldsymbol{\Theta}$. For any $\boldsymbol{\theta} \in \boldsymbol{\Theta}$, let $\ell(\mathbf{y}, \mathbf{x}, \boldsymbol{\theta})$ denote the loss, the expected loss (risk) be $L(\boldsymbol{\theta}) = \mathbb{E}_\nu[\ell(\mathbf{y}, \mathbf{x}, \boldsymbol{\theta})]$, and the empirical loss be $\hat{L}(\boldsymbol{\theta}, D) = \frac{1}{n}\sum_{i=1}^n \ell(\mathbf{y}_i, \mathbf{x}_i, \boldsymbol{\theta})$. Define the subset of zero-variance models as $\boldsymbol{\Theta}_0 := \{\boldsymbol{\theta} \in \boldsymbol{\Theta} : \mathbb{V}_\nu(\ell(\mathbf{y}, \mathbf{x}, \boldsymbol{\theta})) = 0\}$.

Computations are carried out on a finite-precision machine. If the model is represented by $p$ parameters and each parameter has numerical precision $\log_2(k)$ bits, the parameter space can be viewed as a finite grid in $\mathbb{R}^p$ with at most $k^p$ distinct models (e.g., $k = 2^{32}$ for single precision and $k = 2^{64}$ for double precision). This discretization resides in $\mathbb{R}^p$ and does not restrict the definition of the loss, which may be specified on a continuous domain. Section 5.3.10 discusses how the results may extend to infinite model classes using recently proposed PAC-Bayes–Chernoff bounds (Casado et al., 2024). The specific assumptions adopted in this work are as follows:

**Assumption 5.9.** The loss is lower-bounded; more precisely, the essential infimum of the loss is finite and positive. That is, $\forall \theta \in \Theta$,

$$
\begin{aligned}
m_{\boldsymbol{\theta}} &:= \operatorname*{ess\,inf}_{(\mathbf{x},\mathbf{y})\,\in\,\mathrm{supp}(\nu)} \ell(\mathbf{y}, \mathbf{x}, \boldsymbol{\theta}) \\
&= \sup\Big\{a \in \mathbb{R} \;:\; \nu\big(\{(\mathbf{x}, \mathbf{y}) \in \mathcal{X} \times \mathcal{Y} \mid \ell(\mathbf{y}, \mathbf{x}, \boldsymbol{\theta}) < a\}\big) = 0\Big\} \geq 0
\end{aligned}
\tag{5.52}
$$

In words, this is the greatest lower bound on $\ell(\mathbf{y}, \mathbf{x}, \boldsymbol{\theta})$ that holds *almost everywhere* under the data distribution $\nu$ is finite and positive. Furthermore, it is assumed that the expected loss is always finite, $\forall \boldsymbol{\theta} \in \boldsymbol{\Theta}\,, L(\boldsymbol{\theta}) < \infty$.

This assumption assures that the loss function is lower-bounded and the generalization error is finite (not necessarily upper-bounded). The first part of the assumption is naturally satisfied in many standard problems; for example, in multiclass classification problems with softmax activation and the cross-entropy loss and regression problems with mean squared error, as it is clear that $m_{\boldsymbol{\theta}} = 0$. The second part of the assumption $(L(\boldsymbol{\theta}) < \infty)$ is mainly introduced for the sake of simplicity in the mathematical exposition. In fact, this could be relaxed in many of the theoretical results of this work, under some considerations. However, under the scope of this thesis, that is studying the generalization error of interpolators, assuming a finite expected error, and thus, a finite generalization error, is a reasonable simplification for the exposition of the theoretical results.

Some extra considerations regarding this assumption raise in the case of a Gaussian likelihood and log-loss, which can be considered another *usual setting* in machine learning. In this setup, the density has to be lower than one for any data sample and the variance of the Gaussian distribution can not be null in order to satisfy Assumption 5.9. However, this is not restrictive; lower than one Gaussian densities is usual in high dimensional Gaussians and can be ultimately imposed with a restriction on the variances. On the other hand, non-zero variances are usually *desirable* to ensure the stability of machine learning models. In fact, variances are typically restricted to be positive or computed as the exponential of a logarithm-scaled variable, ensuring their positiveness.

As an example of a case not considered under Assumption 5.9, the loss function cannot be Gaussian-distributed, as $m_{\boldsymbol{\theta}} = -\infty$ in this case. In this regard, an exponentially-distributed loss could be used verifying Assumption 5.9.

### 5.3.3 The Rate Function

The *rate function*, denoted by $\mathcal{I}_{\boldsymbol{\theta}}(a)$, plays a central role in Large Deviation Theory (LDT) (Ellis, 2012), a branch of probability theory tightly connected to statistical mechanics, that deals with understanding the behavior of rare events or large fluctuations in random systems. The rate function is normally used to understand the underlying structure of these events and how their probabilities change as they move away from the typical or average behavior.

Mathematically, the rate function is defined as the *Legendre transform* of the *cumulant-generating function*, denoted by $J_{\boldsymbol{\theta}}(\lambda)$, which is the natural logarithm of the *moment-generating function* of a random variable, in this case $L(\boldsymbol{\theta})-\ell(\mathbf{y},\mathbf{x},\boldsymbol{\theta})$ with $\mathbf{y},\mathbf{x}\sim\nu(\mathbf{y},\mathbf{x})$.

**Definition 5.10** (Rate Function)**.** For any model $\boldsymbol{\theta}\in\boldsymbol{\Theta}$, its rate function is a real-valued function $\mathcal{I}_{\boldsymbol{\theta}} : [0, L(\boldsymbol{\theta}) - m_{\boldsymbol{\theta}}) \to \mathbb{R}_0^+$, defined as

$$\mathcal{I}_{\boldsymbol{\theta}}(a) = \sup_{\lambda>0}\ \lambda a - J_{\boldsymbol{\theta}}(\lambda) \quad \forall a \in [0, L(\boldsymbol{\theta}) - m_{\boldsymbol{\theta}})\,, \tag{5.53}$$

where the cumulant-generating function is a real-valued function $J_{\boldsymbol{\theta}} : \mathbb{R}_0^+ \to \mathbb{R}_0^+$, defined as

$$J_{\boldsymbol{\theta}}(\lambda) = \log \mathbb{E}_\nu\left[e^{\lambda(L(\boldsymbol{\theta})-\ell(\mathbf{y},\mathbf{x},\boldsymbol{\theta}))}\right] \quad \forall\lambda\geq 0\,. \tag{5.54}$$

The inverse of the rate function, denoted $\mathcal{I}_{\boldsymbol{\theta}}^{-1}(s)$, will also play a relevant role in this thesis.

**Definition 5.11** (Inverse Rate Function)**.** For any model $\boldsymbol{\theta}\in\boldsymbol{\Theta}$, the inverse rate function is a real-valued function $\mathcal{I}_{\boldsymbol{\theta}}^{-1} : \mathbb{R}_0^+ \to [0, L(\boldsymbol{\theta}) - m_{\boldsymbol{\theta}}]$, defined as

$$\mathcal{I}_{\boldsymbol{\theta}}^{-1}(s) = \inf_{\lambda>0} \frac{J_{\boldsymbol{\theta}}(\lambda)+s}{\lambda} \quad \forall s\geq 0\,. \tag{5.55}$$

Note that, when $\mathbb{P}_\nu(\ell(\mathbf{y},\mathbf{x},\boldsymbol{\theta}) = m_{\boldsymbol{\theta}}) \neq 0$, for $\forall s \geq -\log\mathbb{P}_\nu(\ell(\mathbf{y},\mathbf{x},\boldsymbol{\theta}) = m_{\boldsymbol{\theta}})$, $\mathcal{I}_{\boldsymbol{\theta}}^{-1}(s)$ is constantly equal to $L(\boldsymbol{\theta}) - m_{\boldsymbol{\theta}}$ and, in consequence, $\mathcal{I}_{\boldsymbol{\theta}}^{-1}(s)$ is a *generalized inverse* of $\mathcal{I}_{\boldsymbol{\theta}}(a)$ (Rockafellar, 1970). For the sake of simplicity, it is assumed throughout the rest of this thesis that for $a > L(\boldsymbol{\theta}) - m_{\boldsymbol{\theta}}$, $\mathcal{I}_{\boldsymbol{\theta}}(a) = \infty$. Furthermore, it is assumed that binary operators apply to this value following common sense. According to the following proposition, both the rate function and its inverse are well-defined for models satisfying Assumption 5.9.

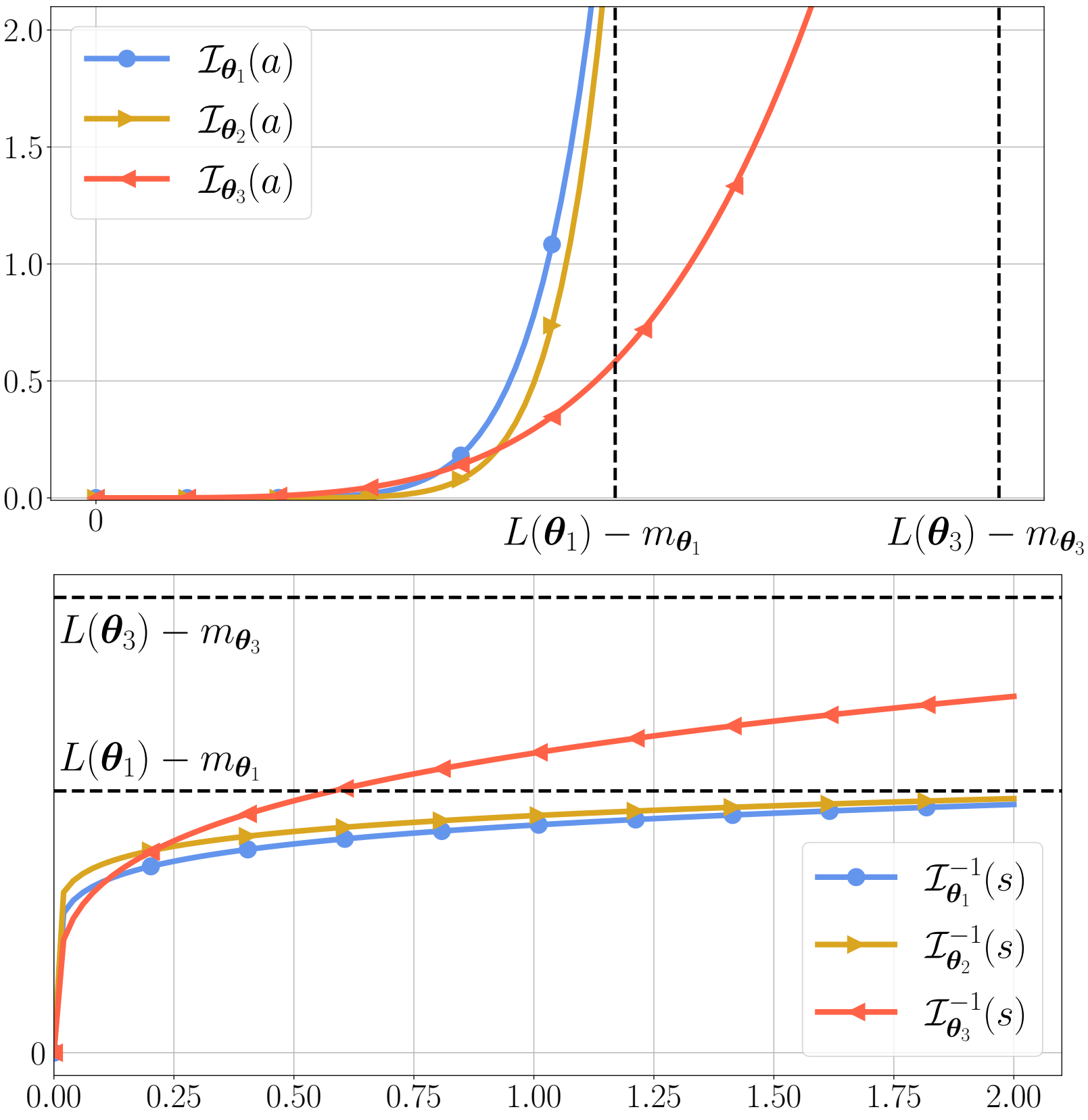

**Figure 5.8:** ***Illustration on different* rate function *(upper) and* inverse rate function *(lower) with the same or different domains of definition.*** *Three different models $\theta_1, \theta_2, \theta_3 \in \Theta$ are shown where $\theta_1$ and $\theta_2$ share the same definition interval for their rate functions.*

**Proposition 5.12.** *Under Assumption 5.9, $\forall \theta \in \Theta$, $\mathcal{I}_\theta(\cdot)$ and $\mathcal{I}^{-1}_\theta(\cdot)$, are well defined. That is, $\forall a \in [0, L(\theta) - m_\theta)$, $\mathcal{I}_\theta(a) < \infty$ and $\forall s \in \mathbb{R}^+_0$, $\mathcal{I}^{-1}_\theta(s) < \infty$.* [Proof in Appendix C.5]

The following result states some properties of the rate and inverse rate function, shedding some light on their monotony, curvature, and shape. In combination with Figure 5.8, this result should provide some intuition on how these two functions behave in general.

**Proposition 5.13** (Rockafellar, 1970)**.** *For any $\theta \in \Theta$, the rate function $\mathcal{I}_\theta(\cdot)$ and the inverse rate function $\mathcal{I}^{-1}_\theta(\cdot)$ satisfy the following properties,*

*(i)* *$\mathcal{I}_\theta(\cdot)$ is convex and $\mathcal{I}^{-1}_\theta(\cdot)$ is concave; both monotonically increasing.*

*(ii)* *Their derivatives at the origin are characterized as:*

$$\lim_{a\to 0} \frac{\partial}{\partial a} \mathcal{I}_\theta(a) = 0 \quad \textit{and} \quad \lim_{s\to 0} \frac{\partial}{\partial s} \mathcal{I}^{-1}_\theta(s) = +\infty\,. \tag{5.56}$$

*(iii)* *In case $\log \mathbb{P}(\ell(\mathbf{y}, \mathbf{x}, \theta) < +\infty$, it verifies that*

$$\lim_{a\to (L(\theta)-m_\theta)^-} \mathcal{I}_\theta(a) = -\log \mathbb{P}(\ell(\mathbf{y}, \mathbf{x}, \theta) = m_\theta)\,, \tag{5.57}$$

$$\lim_{s\to -\log \mathbb{P}(\ell(\mathbf{y},\mathbf{x},\theta)=m_\theta)} \mathcal{I}^{-1}_\theta(s) = L(\theta) - m_\theta\,. \tag{5.58}$$

*Otherwise, the same limits hold treating the negative logarithm as infinite.*

*(iv)* $\mathcal{I}_{\boldsymbol{\theta}}(\cdot)$ *and* $\mathcal{I}_{\boldsymbol{\theta}}^{-1}(\cdot)$ *are invariant to reparameterizations.*

Note that the rate function of a Gaussian random variable, like many other standard random variables, does not have a vertical asymptote. However, if the loss follows a Gaussian distribution, it will not satisfy the first condition in Assumption 5.9, as the essential infimum is not finite.

The relevance of the rate function is a consequence of the following results; firstly, the classic Chernoff bound defines how likely it is to observe, over different i.i.d. data sets, an empirical loss $\hat{L}(D,\boldsymbol{\theta})$ that deviates from the expected loss $L(\boldsymbol{\theta})$ by a positive quantity $a > 0$.

**Theorem 5.14.** [Chernoff, 1952] *For any fixed* $\boldsymbol{\theta} \in \boldsymbol{\Theta}$ *and* $a > 0$*, it satisfies*

$$\mathbb{P}_{D\sim\nu^n}\Big(L(\boldsymbol{\theta}) - \hat{L}(D,\boldsymbol{\theta}) \geq a\Big) \leq e^{-n\mathcal{I}_{\boldsymbol{\theta}}(a)}\,. \tag{5.59}$$

[Proof in Appendix C.5]

The above bound is relevant in this thesis due to two main factors that characterize its behavior w.r.t. the data set size $n$ and the generalization error gap $a$: on one hand, Chernoff's bound is known to be quite loose in the mean of the variable (when $a \approx 0$) but tight on the tail (when $a \approx L(\boldsymbol{\theta}) - m_{\boldsymbol{\theta}}$). As a result, the bound is specially useful when talking about interpolators, where the value of $L(\boldsymbol{\theta}) - \hat{L}(D,\boldsymbol{\theta})$ is close to its maximum possible value ($a = L(\boldsymbol{\theta}) - m_{\boldsymbol{\theta}}$), as stated in the following result.

**Proposition 5.15.** *For any fixed* $\boldsymbol{\theta} \in \boldsymbol{\Theta}$ *and* $n > 0$*, it satisfies*

$$\lim_{a\to L(\boldsymbol{\theta})-m_{\boldsymbol{\theta}}} \mathbb{P}_{D\sim\nu^n}\Big(L(\boldsymbol{\theta}) - \hat{L}(D,\boldsymbol{\theta}) \geq a\Big) = \lim_{a\to L(\boldsymbol{\theta})-m_{\boldsymbol{\theta}}} e^{-n\mathcal{I}_{\boldsymbol{\theta}}(a)}\,. \tag{5.60}$$

[Proof in Appendix C.5]

On the other hand, Cramér's Theorem (Cramér, 1938) states that Chernoff's bound is exponentially tight for *large* $n$. Formally, this statement is written as follows,

**Theorem 5.16.** [Cramér, 1938; Ellis, 2012] *For any fixed* $\boldsymbol{\theta} \in \boldsymbol{\Theta}$ *and any* $a > 0$*, it satisfies*

$$\lim_{n\to\infty} -\frac{1}{n}\log \mathbb{P}_{D\sim\nu^n}\Big(L(\boldsymbol{\theta}) - \hat{L}(D,\boldsymbol{\theta}) \geq a\Big) = \mathcal{I}_{\boldsymbol{\theta}}(a)\,. \tag{5.61}$$

[Proof in Appendix C.5]

In LDT, the above asymptotic result is intuitively interpreted using the following equality,

$$\mathbb{P}_{D\sim\nu^n}\Big(L(\boldsymbol{\theta}) - \hat{L}(D,\boldsymbol{\theta}) \geq a\Big) = e^{-n\mathcal{I}_{\boldsymbol{\theta}}(a)+o(n,a)}\,, \tag{5.62}$$

which shows that the exact expression of $\mathbb{P}_{D\sim\nu^n}(L(\boldsymbol{\theta}) - \hat{L}(D,\boldsymbol{\theta}) \geq a)$ is defined by the rate function, up to a sub-exponential term, that is negligible when $n$ is *large*, because $\lim_{n\to\infty}\frac{o(n,a)}{n} = 0$ and, in consequence, it does not have any meaningful effect. Then, according to LDT, when $n$ is large, the rate function would be the key quantity describing the generalization error of a model, that is, statistical behavior of the difference between the expected $L(\boldsymbol{\theta})$ and the empirical loss $\hat{L}(D, \boldsymbol{\theta})$.

Summarizing the preceding analysis, *the Chernoff bound, governed by the rate function* $\mathcal{I}_{\boldsymbol{\theta}}(\cdot)$*, is tighter for larger datasets* ($\mathit{when}\ n \to \infty$) *and for models interpolating the data* ($a \approx L(\boldsymbol{\theta}) - m_{\boldsymbol{\theta}}$)*, which are the settings of modern machine learning.*

The Chernoff bound asserts that a model's generalization error is governed by its rate function. This rate function is an *oracle*, distribution-dependent quantity because it depends on the data-generating distribution $\nu$. Although $\nu$ is unknown in typical machine-learning settings, this dependence is informative: it helps explain why data augmentation is effective, why over-parameterized model families are beneficial, and why invariant architectures (e.g., convolutional neural networks) are desirable. Moreover, the rate function can be estimated from an independent dataset serving as a proxy for $\nu$. This aligns with standard practice that uses a separate *validation dataset* to estimate $L(\boldsymbol{\theta})$ while avoiding *data snooping*. As shown in Section 5.3.9, the rate function $\mathcal{I}_{\boldsymbol{\theta}}(\cdot)$ can be estimated by evaluating the model once on the validation set and applying log-sum-exp operations to compute $J_{\boldsymbol{\theta}}(\lambda)$, followed by a simple grid search to optimize $\lambda$ in Definition 5.10. The practical consequence—that the cumulant and rate functions can be readily estimated and plotted (see Figure 5.9)—is noteworthy, as it enables empirical illustration and validation of the framework's theoretical predictions.

Let us briefly illustrate why the rate function is useful to understand the generalization of interpolators. Figure 5.9 displays an estimation of the rate function $\mathcal{I}_{\boldsymbol{\theta}}(\cdot)$ for some neural network models used in C. Zhang et al., 2017. This work demonstrated that, within the same model class, interpolators that merely memorize the training data (as represented by "Random" in the figure, which has been learned using a random-labeled training data set) can co-exist with others that generalize exceptionally well, as the ones learned using weight-norm regularization (labeled as "L2") and/or data-augmentation techniques (random-cropping labeled as "Crop"). According to Chernoff's bound, models with a larger rate function are less likely to have significant disparities between their expected and empirical losses; in other words, the empirical loss $\hat{L}(D, \boldsymbol{\theta})$ is more concentrated around its mean $L(\boldsymbol{\theta})$. Figure 5.10 illustrates this fact for three of these models by plotting histograms that showcase the distribution of $\hat{L}(D, \boldsymbol{\theta})$ across various data sets of size $n = 50$ (retrieved from the test set). From the histograms, it is clear that the concentration of $\hat{L}(D, \boldsymbol{\theta})$ varies among the models; the initial model, defined by *Kaiming or He initialization* (Goodfellow et al., 2016), has a prominent rate function. For this model, $\hat{L}(D, \boldsymbol{\theta})$ is tightly concentrated around its mean, $L(\boldsymbol{\theta}) = \log 10$, observable through the minute scale of its x-axis. Comparatively, the *Standard* model, trained using SGD and characterized by a smaller rate function, has a more dispersed distribution around its mean $L(\boldsymbol{\theta}) = 0.65$. This dispersion is notably wider than that of the *L2-Crop* model. The latter, also trained with SGD but incorporating both $\ell_2$ regularization and data-augmentation, has a larger rate function than the *Standard* model.

| Inception | Crop | L2 | Train Acc. | Test Acc. | Test NLL | $\ell_2$-norm |
|---|---|---|---|---|---|---|
| Standard | no | no | 99.99% | 84.36% | 0.65 | 304 |
| Crop | yes | no | 99.94% | 86.89% | 0.58 | 309 |
| L2 | no | yes | 100.0% | 86.60% | 0.49 | 200 |
| L2-Crop | yes | yes | 99.98% | 88.45% | 0.42 | 130 |
| Random | no | no | 100.0% | 10.13% | 5.52 | 311 |
| Initial | - | - | 10.00% | 10.00% | 2.30 | 593 |

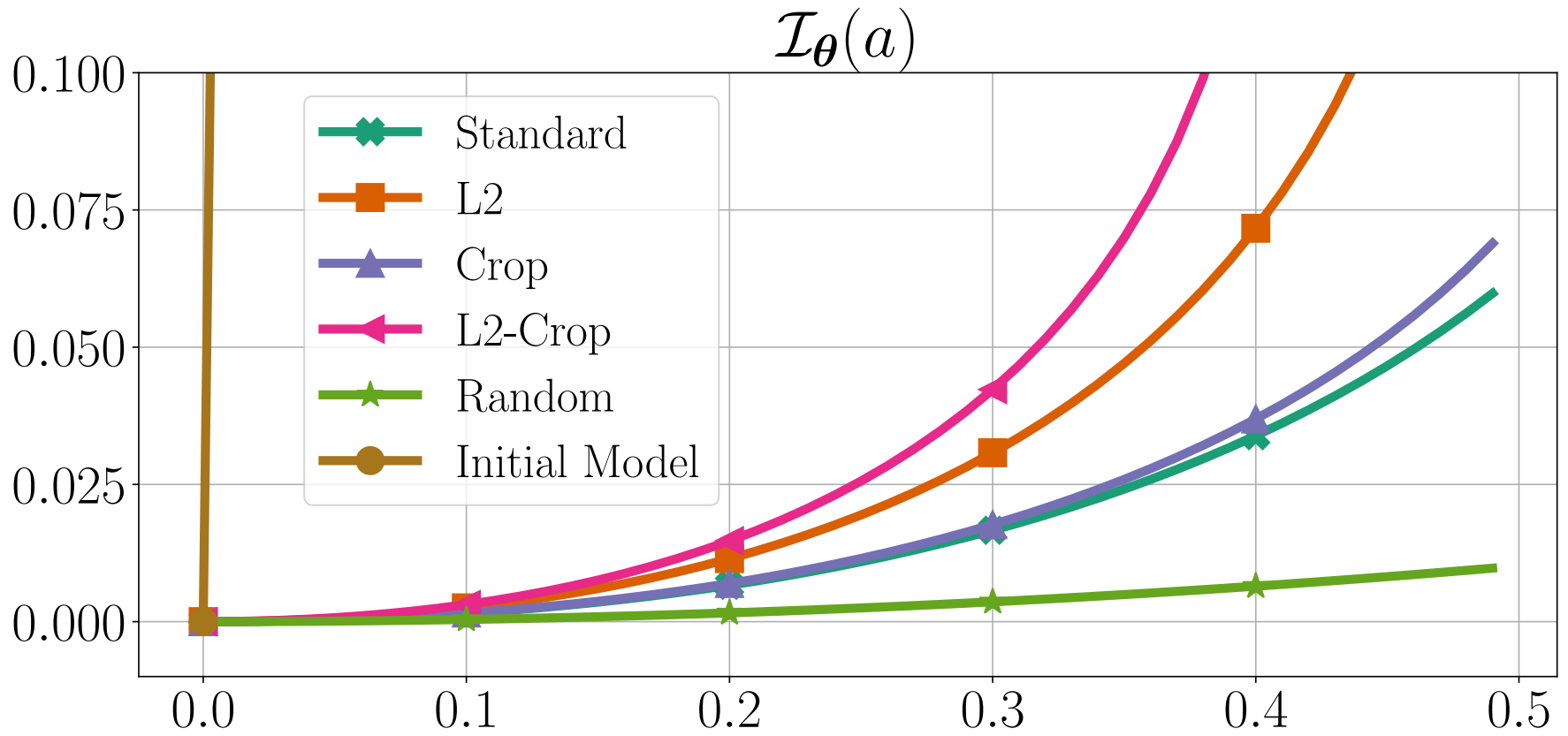

**Figure 5.9:** ***Metrics and rate-function curves for Inception variants on CIFAR-10 with random cropping and $\ell_2$ regularization.*** *The top panel is a table that, for each model variant—*Standard, Crop (*random cropping*), L2 (*$\ell_2$ weight decay*), L2–Crop (*both*), Random (*training on randomly sampled labels*) *and* Initial (*untrained*)*—records whether cropping and/or $\ell_2$ are used, together with training accuracy, test accuracy, test negative log-likelihood (NLL) and the parameter $\ell_2$-norm. The bottom panel plots the corresponding estimated rate functions $\mathcal{I}_{\boldsymbol{\theta}}(a)$ for the same variants (legend as in the table), with $a$ on the horizontal axis.*

The main purpose of Figures 5.9 and 5.10 is to illustrate that $\hat{L}(\boldsymbol{\theta}, D)$ is a random variable whose concentration around its mean $L(\boldsymbol{\theta})$ can vary substantially, a property captured by the model's rate function. When a model interpolates the training data, the observed value of $\hat{L}(\boldsymbol{\theta}, D)$ should be interpreted as one realization of this random variable for the particular dataset $D$. Consequently, when comparing two models that both interpolate, preference should be given to the one whose empirical loss is more tightly concentrated around its mean (i.e., exhibits a larger rate function), as it is more likely to achieve a smaller expected loss on new data. For example, the left and center panels of Figure 5.10 depict the distributions of $\hat{L}(\boldsymbol{\theta}, D)$ for the *Standard* and *L2-Crop* models, respectively; since both interpolate the training set but *L2-Crop* exhibits a more concentrated empirical-loss distribution, *L2-Crop* is preferable.

### 5.3.4 Generalization of Interpolators

This section presents the main results—Theorems 5.17 and 5.23—which demonstrate that the (inverse) rate function characterizes the generalization performance of interpolating models. In addition, a new notion of model smoothness is introduced that aligns with the principle that, under suitable conditions, a sufficiently smoother inter-polator generalizes better.

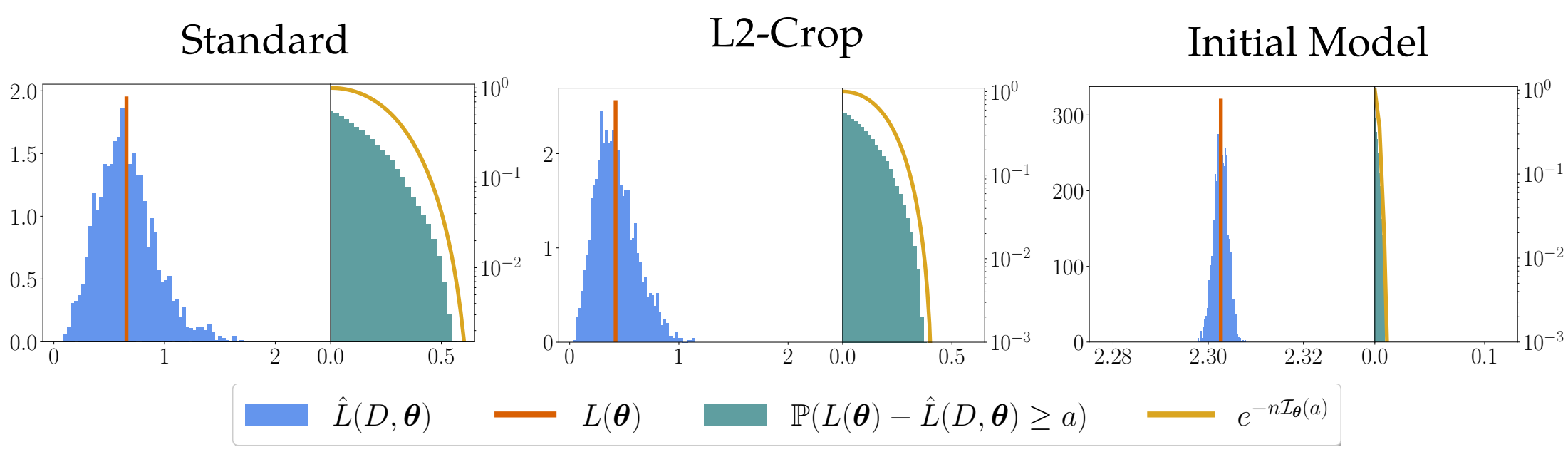

**Figure 5.10: *Distribution of empirical loss, tail probabilities and Chernoff bound for three Inception models on CIFAR-10* ($n = 50$).** *Panels titled* Standard, L2–Crop *and* Initial Model *display, for dataset size $n = 50$: a histogram of the empirical loss $\hat{L}(D, \theta)$ over repeated test-set subsamples (blue, left axis); a vertical line at the generalization error $L(\theta)$ computed on the full test set (orange); and, on the right sub-axes with logarithmic probability scale, the empirical tail probability $\mathbb{P}(L(\theta) - \hat{L}(D, \theta) \geq a)$ as a function of $a \in [0, L(\theta)]$ (green) together with the Chernoff bound $e^{-n\mathcal{I}_\theta(a)}$ derived from the corresponding rate function (gold). All quantities are approximated using the CIFAR-10 test set for the three Inception variants introduced in Figure 5.9.*

#### 5.3.4.1 Distribution-Dependent Bounds for Over-parameterized Interpolators

As discussed in the introduction, current high-probability generalization bounds in the form of Equation (5.49) have been unable so far to explain the generalization of over-parameterized interpolators. Different works (Gastpar et al., 2024; Nagarajan, 2021; Nagarajan and Kolter, 2019b; Y. Wang et al., 2024) have also shown that the problem lies in the impossibility of having tight generalization bounds solely depending on the training data for over-parameterized model classes. Here, a *tight* bound refers to one whose value closely matches the true generalization error, leaving little or no unexplained gap between the theoretical performance and the bound itself. Gastpar et al., 2024 even concludes that *"a bound without explicit distributional assumptions is likely to be not tight"*. Thus, an open question arises:

**Open Question 1.** Are there tight distribution-dependent bounds for over-parameterized models?

The answer to this question is not straightforward given that Nagarajan and Kolter, 2019b showed examples of distribution-dependent bounds (Nagarajan and Kolter, 2019b's Definition 3.3) which are not tight. However, the following result shows a uniform-convergence bound which applies simultaneously over the model class,

**Theorem 5.17.** [**PAC-Chernoff Bound**] *With h.p. $1 - \delta$ over $D \sim \nu^n$, for all $\theta \in \Theta$, simultaneously,*

$$L(\theta) \leq \hat{L}(D, \theta) + \mathcal{I}_\theta^{-1}\left(\tfrac{1}{n} \log \tfrac{k^p}{\delta}\right). \tag{5.63}$$

[Proof in Appendix C.5]

In this bound, $\mathcal{I}_\theta^{-1}(\frac{1}{n} \log \frac{k^p}{\delta})$ defines a complexity measure for the model $\theta$ in the context of a model class defined by $p$ parameters, the training data set of $n$ samples and, according to the own definition of $\mathcal{I}_{\boldsymbol{\theta}}^{-1}(\cdot)$ (see Definition 5.11), it also depends on the data-generating distribution $\nu$. From Proposition 5.13, this complexity measure $\mathcal{I}_{\boldsymbol{\theta}}^{-1}(\frac{1}{n}\log\frac{k^p}{\delta})$ monotonically grows with the size of the model class and monotonically decreases with the level of confidence $\delta$ and the size of the training data.

It is common to find models, within the same model class, that define the same loss function, $\ell(\cdot,\cdot,\boldsymbol{\theta}_1) = \ell(\cdot,\cdot,\boldsymbol{\theta}_2)$. For instance, in a Multi-Layer Perceptron (MLP), weights can be permuted in specific ways without altering model predictions, and models with zeros in the last layer produce the same predictions regardless of the weights in other layers. Consequently, when calculating the size of the model class, models that define the same empirical loss across different datasets can be effectively "excluded". Let $\bar{\boldsymbol{\Theta}} \subseteq \boldsymbol{\Theta}$ be a subset of the model class where, if $\boldsymbol{\theta}, \boldsymbol{\theta}' \in \bar{\boldsymbol{\Theta}}$, there exists a dataset $D \sim \nu^n$ such that $\hat{L}(D,\boldsymbol{\theta}) \neq \hat{L}(D,\boldsymbol{\theta}')$. The following result provides a refined bound, based on the size of $\bar{\boldsymbol{\Theta}}$.

**Corollary 5.18.** *With h.p.* $1-\delta$ *over* $D \sim \nu^n$*, for all* $\boldsymbol{\theta} \in \boldsymbol{\Theta}$*, simultaneously,*

$$L(\boldsymbol{\theta}) \leq \hat{L}(D,\boldsymbol{\theta}) + \mathcal{I}_{\boldsymbol{\theta}}^{-1}(\tfrac{1}{n}\log\tfrac{|\bar{\boldsymbol{\Theta}}|}{\delta})\,. \tag{5.64}$$

[Proof in Appendix C.5]

For the remainder of this work, $k^p$ is used in all the presented results. However, in each case, it may be replaced $k^p$ with the typically smaller term $|\bar{\boldsymbol{\Theta}}|$ by applying the result above. The following result shows that, when $n$ goes to infinity, the proposed complexity measure converges with rate $1/\sqrt{n}$ to the (scaled) standard deviation of the loss function.

**Theorem 5.19.** *For any* $\delta \in (0,1)$ *and any* $\boldsymbol{\theta} \in \boldsymbol{\Theta}$*, it verifies that*

$$\lim_{n\to\infty} \sqrt{n}\,\mathcal{I}_{\boldsymbol{\theta}}^{-1}(\tfrac{1}{n}\log\tfrac{k^p}{\delta}) = \sqrt{2\mathbb{V}_\nu(\ell(\mathbf{y},\mathbf{x},\boldsymbol{\theta}))\log\tfrac{k^p}{\delta}}\,. \tag{5.65}$$

[Proof in Appendix C.5]

The relevant property of this novel complexity measure is that it is a *perfectly tight proxy* of the expected loss $L(\boldsymbol{\theta})$ for models that interpolate the training data, $\hat{L}(D,\boldsymbol{\theta}) \leq \epsilon$, even if the model class is over-parameterized,

**Proposition 5.20.** *With h.p.* $1-\delta$ *over* $D \sim \nu^n$*, for all* $\boldsymbol{\theta} \in \boldsymbol{\Theta}$*, simultaneously,*

$$\textit{if} \quad \hat{L}(D,\boldsymbol{\theta}) \leq \epsilon \quad \textit{then} \quad 0 \leq L(\boldsymbol{\theta}) - \mathcal{I}_{\boldsymbol{\theta}}^{-1}(\tfrac{1}{n}\log\tfrac{k^p}{\delta}) \leq \epsilon\,. \tag{5.66}$$

[Proof in Appendix C.5]

Using those results, there exists an algorithm-distribution-dependent bound that is *perfectly tight* for any data distribution and any learning algorithm; even for over-parameterized model classes. However, for this bound to be tight, it is imperative that the model interpolates the training data. Formally, for any fixed $\epsilon > 0$, with $2^{\mathcal{X}\times\mathcal{Y}}$ denoting the power set of all possible data sets of any size; let $\mathcal{A}_\epsilon : 2^{\mathcal{X}\times\mathcal{Y}} \to \boldsymbol{\Theta}$ be an algorithm that takes any data set and returns a model $\boldsymbol{\theta} \in \boldsymbol{\Theta}$ such that its training loss is lower than $\epsilon$. That is, for any training data set $D \in 2^{\mathcal{X}\times\mathcal{Y}}$, $\mathcal{A}_\epsilon$ verifies that $\hat{L}(\mathcal{A}_\epsilon(D), D) \leq \epsilon$. This kind of algorithm is quite common in machine learning where the model space $\boldsymbol{\Theta}$ is big enough that there exist models capable of memorizing random labels (C. Zhang et al., 2017, Theorem 1). Then, for any $\mathcal{A}_\epsilon$ algorithm, with h.p. $1-\delta$ over $D \sim \nu^n$,

$$L(\mathcal{A}_\epsilon(D)) \leq \hat{L}(\mathcal{A}_\epsilon(D), D) + \mathcal{I}^{-1}_{\mathcal{A}_\epsilon(D)}\left(\tfrac{1}{n}\log\tfrac{kp}{\delta}\right) \leq L(\mathcal{A}_\epsilon(D)) + \epsilon\,. \tag{5.67}$$

This provides an answer to the Open Question 1 using this newly presented bound: *PAC-Chernoff bounds are perfectly tight for (over-parameterized) interpolators.*

The question now is whether this distribution-dependent bound is useful to understand the generalization of algorithms retrieving (over-parameterized) interpolators. A first insight from the above bound is that the (inverse) rate function of the retrieved interpolator defines its generalization error. This is the first (distribution-dependent) complexity measure characterizing the generalization error of an interpolator even in the context of an over-parameterized model class.

#### 5.3.4.2 Smoother Interpolators Generalize Better

As mentioned in the introduction, an open question in machine learning is the following:

**Open Question 2.** Given two models $\boldsymbol{\theta} \in \boldsymbol{\Theta}$ and $\boldsymbol{\theta}' \in \boldsymbol{\Theta}'$, both successfully interpolating the training data, which of them generalizes better?

In this section, a formal result is introduced which provides the following answer to the above question: *the smoother interpolator is the one that achieves better generalization, given that it is sufficiently smoother*. The condition of being *smoother* is defined in terms of the rate function of the models. A model $\boldsymbol{\theta}$ is *smoother* than a model $\boldsymbol{\theta}'$ if, for any $a > 0$, observing a deviation larger than $a$ between the expected (test) loss and the empirical (train) loss is consistently smaller for $\boldsymbol{\theta}$ than for $\boldsymbol{\theta}'$. This is formalized as follows.

**Definition 5.21.** Given a data-generating distribution $\nu$ and a loss function $\ell$, a model $\boldsymbol{\theta} \in \boldsymbol{\Theta}$ is $\beta$-smoother than a model $\boldsymbol{\theta}' \in \boldsymbol{\Theta}'$ if

$$\forall a \in (0, \beta] \quad \mathcal{I}_{\boldsymbol{\theta}}(a) \geq \mathcal{I}_{\boldsymbol{\theta}'}(a)\,. \tag{5.68}$$

Notice that if $\boldsymbol{\theta}$ is $\beta$-smoother than $\boldsymbol{\theta}'$, it is $\beta'$-smoother for any $\beta' \in (0, \beta]$. Furthermore, if $\boldsymbol{\theta}$ is $\beta$-smoother than $\boldsymbol{\theta}'$, it verifies that $\boldsymbol{\theta}'$ cannot be $\beta'$-smoother than $\boldsymbol{\theta}$ for any $\beta' > 0$. This property can be used to reasonably compare degrees of smoothness between models in the same or different model spaces. It is important to notice that the concept of smoothness is defined in the context of a given data-generating distribution and loss function.

According to the above definition, the higher the rate function, the smoother the model when comparing it to others. Furthermore, according to Chernoff's bound (see Theorem 5.14 and Equation (5.62)), smoother models will have an empirical loss $\hat{L}(D, \boldsymbol{\theta})$ more concentrated around its expected value $L(\boldsymbol{\theta})$. In consequence, it is more unlikely to observe higher differences between $\hat{L}(D, \boldsymbol{\theta})$ and $L(\boldsymbol{\theta})$ in smoother models. In fact, the following result states that there is a correspondence between the variance of the model's loss and the newly introduced notion of smoothness.

**Proposition 5.22.** *For any $\boldsymbol{\theta} \in \boldsymbol{\Theta}$ and $\boldsymbol{\theta}' \in \boldsymbol{\Theta}'$,*

$$\exists \beta > 0 \text{ s.t. } \boldsymbol{\theta} \text{ is } \beta\text{-smoother than } \boldsymbol{\theta}' \iff \mathbb{V}_\nu(\ell(\mathbf{y}, \mathbf{x}, \boldsymbol{\theta})) \leq \mathbb{V}_\nu(\ell(\mathbf{y}, \mathbf{x}, \boldsymbol{\theta}')) \,. \tag{5.69}$$

[Proof in Appendix C.5]

This result shows the relation between the smoothness and the variance of the loss function between a pair of models $\boldsymbol{\theta}$ and $\boldsymbol{\theta}'$. As a result, this notion of smoothness can be *intuitively* understood as a *generalization of the variance of the loss of a model*.

Using this definition of smoothness and the results presented in Section 5.3.4.1, the following result provides an answer to which interpolator generalizes better.

**Theorem 5.23.** *For any $\epsilon \geq 0$, with h.p. $1-\delta$ over $D \sim \nu^n$, for all $\boldsymbol{\theta} \in \boldsymbol{\Theta} \subset \mathbb{R}^p$ and $\boldsymbol{\theta}' \in \boldsymbol{\Theta}'$, simultaneously,*

$$\text{if } \hat{L}(D, \boldsymbol{\theta}) \leq \epsilon \text{ and } \boldsymbol{\theta} \text{ is } \mathcal{I}_{\boldsymbol{\theta}}^{-1}(\tfrac{1}{n} \log \tfrac{kp}{\delta})\text{-smoother than } \boldsymbol{\theta}', \text{ then, } L(\boldsymbol{\theta}) \leq L(\boldsymbol{\theta}') + \epsilon \,. \tag{5.70}$$

[Proof in Appendix C.5]

Note that, due to Proposition 5.22, the smoothness condition given by "$\mathcal{I}_{\boldsymbol{\theta}}^{-1}(\frac{1}{n} \log \frac{kp}{\delta})$-smoother" in Theorem 5.23 can be understood as an inequality between the variances of the loss function under the hypothesis that $n$ is sufficiently large.

In fact, if $\mathbb{V}_\nu(\ell(\mathbf{y}, \mathbf{x}, \boldsymbol{\theta})) \leq \mathbb{V}_\nu(\ell(\mathbf{y}, \mathbf{x}, \boldsymbol{\theta}'))$, there exists $\beta > 0$ such that $\boldsymbol{\theta}$ is $\beta$-smoother than $\boldsymbol{\theta}'$. Then, if $n$ is large enough to verify that $\mathcal{I}_{\boldsymbol{\theta}}^{-1}(\frac{1}{n} \log \frac{kp}{\delta}) \leq \beta$, it verifies that, with h.p., if $\hat{L}(D, \boldsymbol{\theta}) \leq \epsilon$, then, $L(\boldsymbol{\theta}) \leq L(\boldsymbol{\theta}') + \epsilon$. In short, Theorem 5.23 is a generalization of the following result: if $n$ is large enough, with h.p.,

$$\text{if } \hat{L}(D, \boldsymbol{\theta}) \leq \epsilon \text{ and } \mathbb{V}_\nu(\ell(\mathbf{y}, \mathbf{x}, \boldsymbol{\theta})) \leq \mathbb{V}_\nu(\ell(\mathbf{y}, \mathbf{x}, \boldsymbol{\theta}')), \text{ then, } L(\boldsymbol{\theta}) \leq L(\boldsymbol{\theta}') + \epsilon \,. \tag{5.71}$$

Theorem 5.23 states that an interpolator generalizes better than another (with h.p.) if it is *sufficiently smoother* in terms of its rate function. Figure 5.11 illustrates the premise of this theorem. Notice that the above result holds even for the *log-loss*, which is the *default* loss used for *training*, and for over-parameterized model classes. This result is specially useful when $\epsilon$ is very small or null, as it states, with h.p., that *smooth interpolators* generalizes better, up to an $\epsilon$, than other less smooth models; independently of whether these interpolate the data or not. Furthermore, the above result verifies that the higher the probability $1-\delta$ or the number of parameters $p$, the stronger the smoothness condition needs to be; and the opposite for larger $n$. In that sense, there might exist interpolators which are $\beta$-smoother than others but have worse generalization performance, because they are not smooth enough in order to apply Theorem 5.23. In summary, *interpolators with a larger rate function $\mathcal{I}_{\boldsymbol{\theta}}(\cdot)$ or, equivalently, smoother interpolators, are the ones that better generalize.*

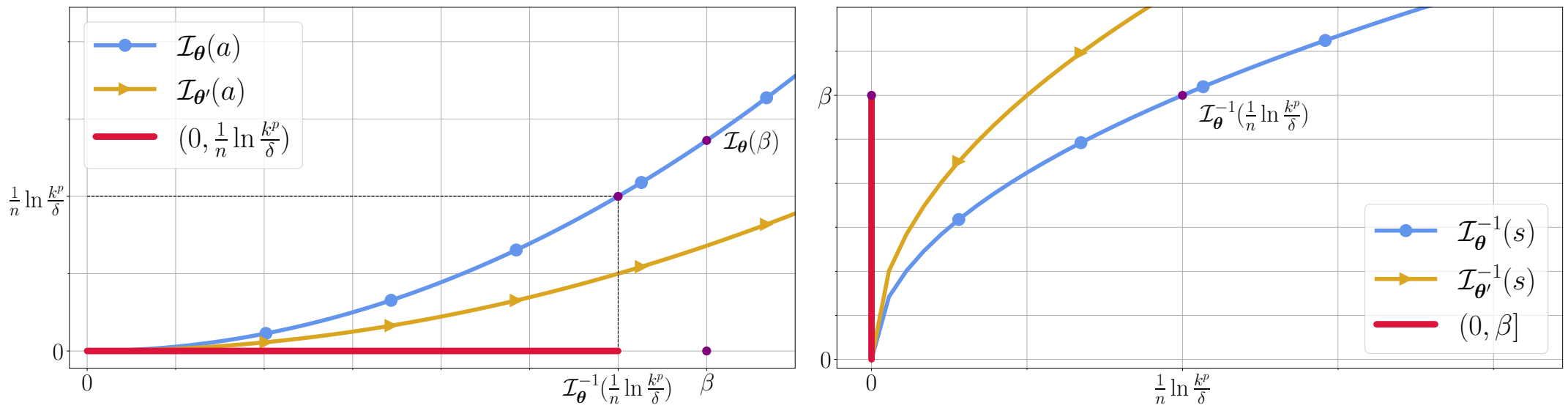

**Figure 5.11:** ***Rate and inverse–rate plots with annotated critical regions for two models.*** *The left panel shows the rate functions $\mathcal{I}_{\boldsymbol{\theta}}(a)$ (blue) and $\mathcal{I}_{\boldsymbol{\theta}'}(a)$ (yellow) as functions of the deviation $a$. A horizontal dashed line is drawn at $s = \frac{1}{n}\log\frac{k^p}{\delta}$; its intersection with $\mathcal{I}_{\boldsymbol{\theta}}$ defines $\mathcal{I}_{\boldsymbol{\theta}}^{-1}(s)$ (black mark), and the interval $[0, \mathcal{I}_{\boldsymbol{\theta}}^{-1}(s)]$ is highlighted. The right panel plots the inverse rate functions $\mathcal{I}_{\boldsymbol{\theta}}^{-1}(s)$ (blue) and $\mathcal{I}_{\boldsymbol{\theta}'}^{-1}(s)$ (yellow) versus $s$. A vertical red segment marks $[0, \beta]$ on the y-axis, and a magenta dot marks $s = \frac{1}{n}\log\frac{k^p}{\delta}$.*

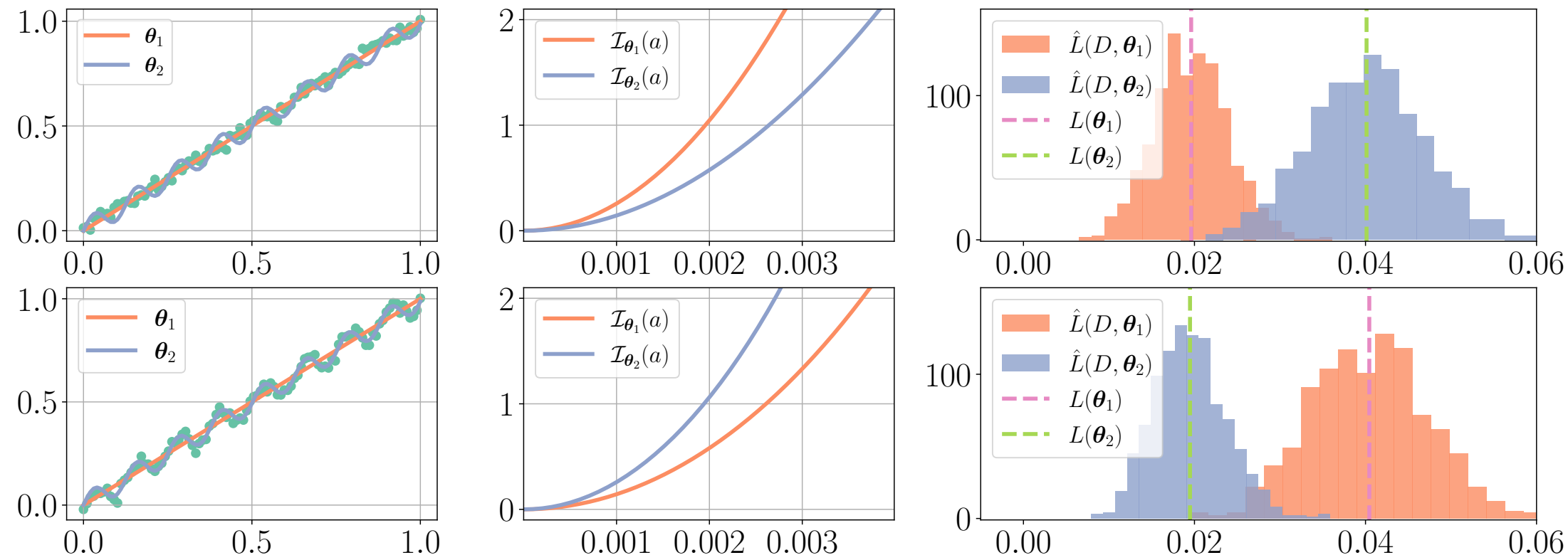

**Figure 5.12:** ***Visualization of two models and their rate functions across two data-generating distributions.*** *Rows correspond to distributions $\nu_1$ (top) and $\nu_2$ (bottom). Left column: observed data with predictions of a linear model $\boldsymbol{\theta}_1$ (orange) and a more complex model $\boldsymbol{\theta}_2$ (blue) over $x \in [0, 1]$. Middle column: rate functions $\mathcal{I}_{\boldsymbol{\theta}_1}(a)$ and $\mathcal{I}_{\boldsymbol{\theta}_2}(a)$ plotted as functions of the deviation $a$. Right column: histograms of empirical losses $\hat{L}(D, \boldsymbol{\theta}_1)$ (orange) and $\hat{L}(D, \boldsymbol{\theta}_2)$ (blue) computed from repeated subsamples; vertical dashed lines mark the corresponding generalization errors $L(\boldsymbol{\theta}_1)$ and $L(\boldsymbol{\theta}_2)$. Colors and line styles are consistent across panels.*

Figure 5.12 shows a synthetical example to highlight how the notion of smoothness depends on the *specific data-generating distribution*, and why one model is *smoother* than another only relative to such distribution. The example considers two different data generating distributions $\nu_1$ (first row, which adds random noise to a linear function) and $\nu_2$ (second row, adds random noise to a complex sinusoidal function). The considered loss is the mean squared error. The introduced notion of smoothness using the rate function shows how $\boldsymbol{\theta}_1$ (a linear model) is *smoother* than $\boldsymbol{\theta}_2$ (a more complex model) under $\nu_1$. In fact, under $\nu_1$, the distribution of $\hat{L}(D, \boldsymbol{\theta}_1)$ is more concentrated around a smaller mean value. However, the second row of this example also shows how the more complex model $\boldsymbol{\theta}_2$ can be smoother than a linear model $\boldsymbol{\theta}_1$ under a different data-generating distribution $\nu_2$ for exactly the same reasons. With this example, it is highlighted that a good notion of smoothness (as the one presented here) should consider the data in which the models are being evaluated, rather than just the complexity of the function they induce.

The results from this section clearly indicate that the generalization error of an interpolator is defined by its level of smoothness and its rate function. The question is whether this theoretical characterization, which relies on distribution-dependent quantities, is useful for understanding the inner workings of current learning techniques and complex phenomena appearing in deep learning. In Section 5.3.5, it is shown how the PAC-Chernoff bound and the novel smoothness criteria are powerful enough to analyze the so-called double-descent phenomenon (Belkin et al., 2019). In Section 5.3.6, the relationship between the smoothness of a model and widely used regularization techniques—such as the parameter norm of a model, distance from initialization, input-gradient norm, and Lipschitz constant—is examined. Many of these quantities have also been previously used as measures of smoothness or complexity of a model (Neyshabur et al., 2017a). In Section 5.3.7, the fact that invariant architectures and data augmentation induce smoother interpolators is studied. Finally, in Section 5.3.8, over-parameterization is revisited as a necessary condition for having smoother interpolators.

### 5.3.5 Understanding Double-Descent with PAC-Chernoff Bounds

The existing literature consistently demonstrates that interpolators with a larger number of parameters tend to perform better. A finding that defies traditional beliefs from classical statistical learning theory. Traditionally, it was assumed that increasing the number of parameters in a model would lead to higher overfitting and, consequently, poor generalization performance. However, this perspective has been challenged by the phenomenon of the double-descent curve (Belkin et al., 2019), which exemplifies a shift in dynamics as models enter the interpolation regime. In this regime, an increase in parameters paradoxically leads to improved performance. This counter-intuitive behavior suggests that once models begin to interpolate, their generalization performance improves as they grow larger.

Figure 5.13 illustrates this phenomenon by showing the training and the test loss of a sequence of convolutional networks with a growing number of parameters (implemented by adding more channels to the layers of a simple convolutional model). More precisely, models with a number of parameters ranging from $4k$ to $1\,161k$. All models were found by running stochastic gradient descent on CIFAR10's training data, until the training loss reaches $0.01$ or until it did not improve in two consecutive epochs of training.

Figure 5.13 also displays the evolution of the PAC-Chernoff bound for the whole range of models. As theoretically shown, this bound is perfectly tight for interpola-tors. In the classical regime, the bound also presents the so-called *double descent phenomena*, even though its tightness is not guaranteed. This highlights how distribution-dependent bounds are powerful enough to capture the complex dynamics emerging in the interpolation regime.

| Size | Train Acc. | Test Acc. | Test NLL | Bound |
|---|---|---|---|---|
| 4k | 43.82% | 42.74% | 1.56 | 2.93 |
| 32k | 70.84% | 59.86% | 1.17 | 2.01 |
| 85k | 86.69% | 62.73% | 1.35 | 1.77 |
| 163k | 97.65% | 63.97% | 1.82 | 1.94 |
| 266k | 99.84% | 65.19% | 2.15 | 2.19 |

| Size | Train Acc. | Test Acc. | Test NLL | Bound |
|---|---|---|---|---|
| 395k | 100.0% | 65.74% | 2.28 | 2.30 |
| 548k | 100.0% | 67.27% | 2.10 | 2.11 |
| 727k | 100.0% | 69.88% | 1.84 | 1.86 |
| 931k | 100.0% | 69.11% | 1.83 | 1.84 |
| 1161k | 100.0% | 69.82% | 1.72 | 1.74 |

**Figure 5.13:** ***Train/test losses and PAC–Chernoff bound versus model size with regimes and interpolation threshold annotated.*** Top: *table with selected networks showing parameter count ("Size"), training accuracy, test accuracy, test negative log-likelihood (Test NLL), and the corresponding bound value (two blocks of five models each).* Bottom: *plot of training loss, test loss, and the PAC–Chernoff bound against the number of parameters (horizontal axis, $\times 10^6$); dots denote individual models and thin lines show smoothed curves. A vertical black line marks the interpolation threshold; the shaded band indicates the "Critical Region", and the plot is divided into "Classical Regime" and "Modern Regime" as labeled.*

**Open Question 3.** Why does the generalization performance of interpolators improve with an increasing number of parameters?

The complexity term of the PAC-Chernoff bound $\mathcal{I}_{\boldsymbol{\theta}}^{-1}(\frac{1}{n}\log\frac{k^p}{\delta})$ monotonically increases with the size of the mode class, which, in principle, would contradict the *double-descent phenomena* on the bound and the fact that generalization error is reduced as the number of models increases. The following result, viewed as an extension of Theorem 5.23, offers a partial explanation for this puzzling phenomena. More precisely, it demonstrates that if the interpolators become progressively *smoother* (higher rate function), their generalization error will be reduced, despite being part of a model class characterized by a greater number of parameters.

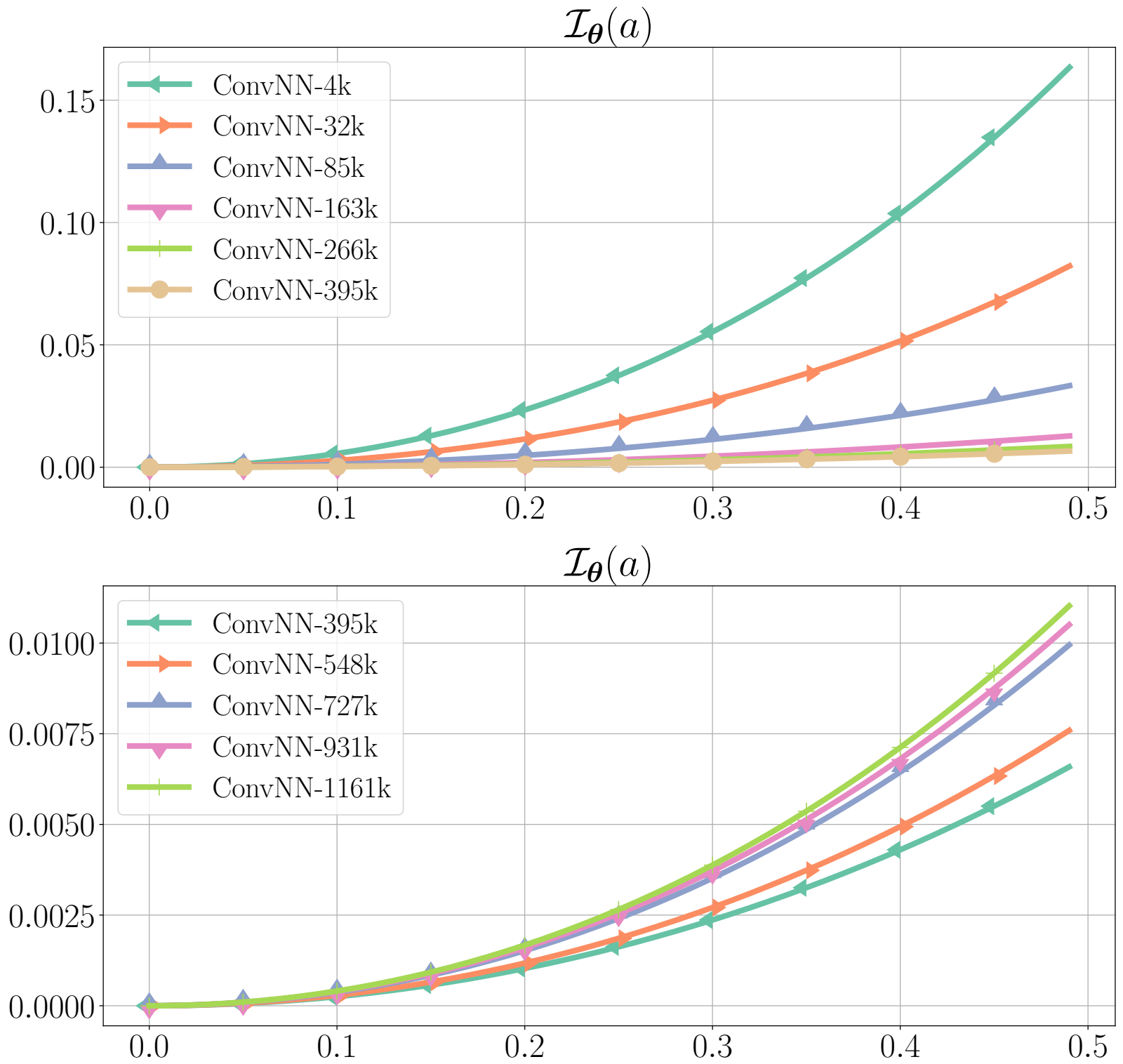

**Figure 5.14:** ***Rate-function curves $\mathcal{I}_{\boldsymbol{\theta}}(a)$ for convolutional networks of varying size on CIFAR-10.*** *Two panels plot $\mathcal{I}_{\boldsymbol{\theta}}(a)$ versus the deviation $a \in [0, 0.5]$. The upper panel shows models with $4k$, $32k$, $85k$, $163k$, $266k$, and $395k$ parameters; and the lower panel shows models with $395k$, $548k$, $727k$, $931k$ and $1\,161k$ parameters (as labeled in the legends). Each colored line corresponds to one CNN obtained from the same architecture family by increasing the number of channels.*

**Theorem 5.24.** *Let $\boldsymbol{\Theta} \subset \boldsymbol{\Theta}'$ be two nested model classes with $p < p'$ parameters respectively. For any $\epsilon > 0$, with h.p. $1-\delta$ over $D \sim \nu^n$, for any $\boldsymbol{\theta}' \in \boldsymbol{\Theta}'$, $\boldsymbol{\theta} \in \boldsymbol{\Theta}$, simultaneously*

$$\begin{gathered} \text{if } \hat{L}(D,\boldsymbol{\theta}') \leq \epsilon\,,\ \hat{L}(D,\boldsymbol{\theta}) \leq \epsilon \text{ and } \boldsymbol{\theta}' \text{ is } \mathcal{I}^{-1}_{\boldsymbol{\theta}'}(\tfrac{1}{n}\log\tfrac{kp'}{\delta})\text{-smoother than } \boldsymbol{\theta} \\ \Downarrow \\ L(\boldsymbol{\theta}') \leq L(\boldsymbol{\theta}) + \epsilon\,. \end{gathered} \tag{5.72}$$

[Proof in Appendix C.5]

Figure 5.14 (right) illustrates the rate functions of the interpolators from Figure 5.13, highlighting how *larger interpolators* become increasingly smoother, meaning their rate functions are progressively higher (i.e., they are progressively smoother). This explains how it is possible to obtain larger interpolators with smaller generalization error. In fact, by reversing the implication of Theorem 5.24, if a larger interpolator exhibits better generalization performance, then it cannot be *less smooth* than a smaller interpolator.

**Corollary 5.25.** *Let $\boldsymbol{\Theta} \subset \boldsymbol{\Theta}'$ be two nested model classes with $p < p'$ parameters respectively.*

*For any* $\epsilon > 0$*, with h.p.* $1-\delta$ *over* $D \sim \nu^n$*, for any* $\boldsymbol{\theta}' \in \boldsymbol{\Theta}'$*,* $\boldsymbol{\theta} \in \boldsymbol{\Theta}$*, simultaneously,*

$$\hat{L}(D,\boldsymbol{\theta}') \leq \epsilon,\ \hat{L}(D,\boldsymbol{\theta}) \leq \epsilon \text{ and } L(\boldsymbol{\theta}') + \epsilon < L(\boldsymbol{\theta}) \\ \Downarrow \\ \boldsymbol{\theta} \text{ is not } \mathcal{I}_{\boldsymbol{\theta}}^{-1}\left(\tfrac{1}{n}\log\tfrac{kp'}{\delta}\right)\text{-smoother than } \boldsymbol{\theta}'\,. \tag{5.73}$$

[Proof in Appendix C.5]

Consequently, Open Question 3 can be reformulated in terms of model smoothness rather than generalization performance.

**Open Question 4.** Why does the smoothness of interpolators improve with an increasing number of parameters?

Based on Theorem 5.24 and Corollary 5.25, addressing Open Question 4 is equivalent to answering Open Question 3. In other words, by answering why the rate function of the interpolators is increasingly higher as the number of parameters in nested model classes increases, Theorem 5.24 will effectively explain why the double descent phenomenon, and in particular, why the generalization performance of interpolators enhances as the number of parameters grows. However, to the best of current knowledge, there is no direct answer to this behavior without incurring additional hypotheses on the matter.

The proposed *rationale* for explaining this phenomenon posits that, by increasing the number of parameters, larger neural networks are better at capturing invariances in the data, a notion that has been supported both theoretically (Bronstein et al., 2021) and empirically (Goodfellow et al., 2009) by the machine learning community. Additionally, as will be shown in Section 5.3.7, this capability of capturing invariances leads to *smoother* models with higher rate functions. Combining these two factors, Theorem 5.24 can help to elucidate why the generalization performance of interpolators improves with an increase in parameters. The experimental setup presented in Figures 5.13 and 5.14 supports this understanding.

Finally, it is also quite interesting to look at the models on the *left part* of the *interpolation threshold* of Figure 5.13. Surprisingly, the sequence of rate functions, displayed in Figure 5.14 (left), gets increasingly lower as the size of the mode class increases, just the opposite of what happens in Figure 5.14 (right). In the classical regime, increasing the number of parameters leads to less smooth models with higher generalization error. However, explaining this behavior in the non-interpolation regime is out of the scope of this work, as the presented PAC-Chernoff bound is *not tight* in that regime and no theoretical guarantees can be derived in such framework. However, it is worth noticing that, even with the lack of tightness guarantees, the proposed bound is *close enough* to showcase the double descent phenomenon itself.

In the following section, the proposed oracle bound is connected with many existing regularization techniques, allowing to explain, in a unified framework, the effectiveness in terms of generalization.

### 5.3.6 Explicit Regularization

Structural risk minimization (Shawe-Taylor et al., 1998) is a learning principle based on *regularizers* that penalizes the complexity of a model when minimizing the training loss. Let $r(\boldsymbol{\theta})$ denote a regularizing function, the learning objective is then given by,

$$\min_{\boldsymbol{\theta}\in\boldsymbol{\Theta}} \hat{L}(D,\boldsymbol{\theta}) + r(\boldsymbol{\theta})\,. \tag{5.74}$$

High-probability bounds of the form $L(\boldsymbol{\theta}) \leq \hat{L}(D,\boldsymbol{\theta}) + \mathcal{C}(\boldsymbol{\theta}, D, \delta)$, have been traditionally used to justify the use of regularizers and to derive novel ones. According to these bounds, an obvious choice for regularizers is the complexity term, $r(\boldsymbol{\theta}) = \mathcal{C}(\boldsymbol{\theta}, D, \delta)$. Then, the structural risk minimization problem given in Equation (5.74) would correspond to minimizing a high-probability upper bound over the expected loss $L(\boldsymbol{\theta})$. Many regularizing methods have been studied from this perspective. One of the most common ones is the $\ell^2$-norm regularization, which recurrently shows up within the complexity term of many generalization bounds (Bartlett et al., 2017). However, as discussed in the introduction, these bounds are known to be loose in the context of over-parameterized model classes interpolating the data. In consequence, the rationale of using their associated complexity measures $\mathcal{C}(\boldsymbol{\theta}, D, \delta)$ as regularizers is no longer as strong as it used to be in learning setups where the model class is not over-parameterized.

**Open Question 5.** Is there a regularizer $r(\boldsymbol{\theta})$ that ensures near-optimal performance for over-parameterized model classes interpolating the training data?

The following result provides an answer to Open Question 5. It shows that the complexity measure of the PAC-Chernoff bound given in Theorem 5.17 can be used as a regularizer. That is, using the inverse rate function $\mathcal{I}_{\boldsymbol{\theta}}^{-1}(\frac{1}{n}\log\frac{kp}{\delta})$, an *optimal regularizer for over-parameterized interpolators* is obtained. More precisely, for any $\epsilon > 0$, let $\boldsymbol{\theta}^\star_\epsilon$ and $\boldsymbol{\theta}^\times_\epsilon$ denote the interpolator with the best generalization performance and the interpolator with the smallest inverse rate function, respectively,

$$\boldsymbol{\theta}^\star_\epsilon = \operatorname*{arg\,min}_{\boldsymbol{\theta}\,:\,\hat{L}(D,\boldsymbol{\theta})\leq\epsilon} L(\boldsymbol{\theta})\,, \qquad \boldsymbol{\theta}^\times_\epsilon = \operatorname*{arg\,min}_{\boldsymbol{\theta}\,:\,\hat{L}(D,\boldsymbol{\theta})\leq\epsilon} \hat{L}(D,\boldsymbol{\theta}) + \mathcal{I}_{\boldsymbol{\theta}}^{-1}(\tfrac{1}{n}\log\tfrac{kp}{\delta})\,. \tag{5.75}$$

With h.p., the expected loss of the interpolator with the smallest inverse rate function $\boldsymbol{\theta}^\times_\epsilon$ is very close to the expected loss of the optimal interpolator, $\boldsymbol{\theta}^\star_\epsilon$.

**Theorem 5.26.** *For any $\epsilon > 0$, with h.p. $1-\delta$ over $D \sim \nu^n$, $|L(\boldsymbol{\theta}^\star_\epsilon) - L(\boldsymbol{\theta}^\times_\epsilon)| \leq \epsilon$.* [Proof in Appendix C.5]

In simpler terms, favoring models with a small inverse rate function is the same as favoring models with a larger rate function, which essentially means choosing *smoother models*. This finding highlights that the *smoothest interpolator* not only fits the data but also performs nearly optimally when it comes to generalization. *The inverse rate function $\mathcal{I}_{\boldsymbol{\theta}}^{-1}(\frac{1}{n}\log\frac{kp}{\delta})$ is an optimal regularizer for over-parameterized interpolators.*

#### 5.3.6.1 Connecting the Inverse Rate with Existing Regularization Techniques

As shown in the previous sections, the inverse rate function, a distribution-dependent and oracle element, is an *optimal regularizer* for interpolators. However, in this section, the discussion will focus on how a wide range of existing regularizers are tightly connected to the inverse rate and, in consequence, to the proposed definition of smoothness. The aim of this thesis is not to fully explore these connections, nor to claim that the newly derived bounds are better than existing ones. The main focus is to show how the (inverse) rate function of a model can be used to unify many existing regularization techniques which were previously assumed to be unrelated.

**Norm $\ell_2$ Regularization** This regularizer, also known as weight decay, is widely used by the machine learning community and it is known to improve generalization even in over-parameterized model classes interpolating the training data (Goodfellow et al., 2016). The following result defines a connection between the inverse rate of a model and its norm, under the widely used assumption that the loss function $\ell(\mathbf{y}, \mathbf{x}, \boldsymbol{\theta})$ is Lipschitz with respect to the parameters of the model $\boldsymbol{\theta}$ (X. Li and Orabona, 2019).

**Proposition 5.27.** *If the loss function $\ell(\mathbf{y}, \mathbf{x}, \boldsymbol{\theta})$ is Lipschitz w.r.t. $\boldsymbol{\theta}$ with constant $M > 0$, then, for any $\boldsymbol{\theta}_0 \in \boldsymbol{\Theta}_0 = \{\boldsymbol{\theta} \in \boldsymbol{\Theta} \mid \mathbb{V}_\nu(\ell(\mathbf{y}, \mathbf{x}, \boldsymbol{\theta})) = 0\}$, it verifies that*

$$\mathcal{I}_{\boldsymbol{\theta}}^{-1}(\tfrac{1}{n} \log \tfrac{k^p}{\delta}) \leq \sqrt{2Ma}\, \|\boldsymbol{\theta} - \boldsymbol{\theta}_0\|_2 \,, \tag{5.76}$$

*where $a = min(1, \frac{1}{n} \log \frac{k^p}{\delta})$.* [Proof in Appendix C.5]

In many common machine learning models, the null vector is a model with null variance, that is, $\mathbf{0} \in \boldsymbol{\Theta}_0$. For example, in supervised classification problems with $K$ labels, a neural network with null weights has a constant loss equal to $\log K$. In these cases, the above result shows that by promoting models with small parameter norm, models with small inverse rate are achieved. In consequence, the $\ell_2$-norm is a proxy to minimize the inverse rate and works as a regularizer. Figure 5.9 illustrates an example of how the use of $\ell_2$-norm regularization leads to models with small parameter norm and larger rate function. This aligns with the existing evidence in the literature about the use of $\ell_2$-norm for successfully regularizing interpolators.

However, in the context of over-parameterized models interpolating the training data, it is well known that the $\ell_2$-norm of the parameters of the model does not correlate well with the generalization error (Y. Jiang et al., 2020). Consider that the $\ell_2$-norm just defines an upper bound over the inverse rate, which, as shown in Section 5.3.4.1 is an optimal proxy for the generalization of interpolators. The gap of this upper bound is the reason behind the failure of $\ell_2$-norm as a proxy for the generalization error of an interpolator. In fact, an open question in machine learning is:

**Open Question 6.** When does the norm of an interpolator correlate with its generalization error?

The following result provides a characterization of a learning setup where these two elements are correlated.

**Corollary 5.28.** *Considering the log-loss, if it belongs to the exponential family with a constant base measure, that is, there exist* $s : \mathcal{X} \times \mathcal{Y} \to \mathbb{R}^p$, $a : \mathbf{\Theta} \to \mathbb{R}$ *and* $k \in \mathbb{R}$ *such that*

$$\ell(\mathbf{y}, \mathbf{x}, \boldsymbol{\theta}) := -\log p(\mathbf{y}|\mathbf{x}, \boldsymbol{\theta}) = \boldsymbol{\theta}^T s(\mathbf{x}, \mathbf{y}) - a(\boldsymbol{\theta}) + k \quad \forall (\mathbf{x}, \mathbf{y}) \sim \nu \,. \tag{5.77}$$

*Then,* $\forall \epsilon > 0$ *exists* $n_0 > 0$ *such that* $\forall n > n_0$*:*

$$\left| \mathcal{I}_{\boldsymbol{\theta}}^{-1}\left(\tfrac{1}{n}\log\tfrac{k^p}{\delta}\right) - \sqrt{2\tfrac{1}{n}\log\tfrac{k^p}{\delta}}\sqrt{\boldsymbol{\theta}^T Cov_\nu(s(\mathbf{y}, \mathbf{x}))\boldsymbol{\theta}} \right| \leq \epsilon \,, \tag{5.78}$$

*where* $Cov_\nu(\cdot)$ *is the covariance w.r.t.* $\nu$ *of the sufficient statistics of each* $(\mathbf{x}, \mathbf{y})$ *sample.* [Proof in Appendix C.5]

This result shows that under the exponential family and a large enough number of data samples, the inverse rate is close to the scaled Riemannian norm of the model parameters under the positive definite matrix $\text{Cov}_\nu(s(\mathbf{y}, \mathbf{x}))$ that defines the Riemannian metric. Note that, in this case, the Riemannian norm will be an optimal regularizer as Theorem 5.26 applies and the inverse rate is a perfectly tight proxy for the generalization error of interpolators as shown in Section 5.3.4.1. The $\ell_2$-norm is obtained when the covariance matrix is a diagonal matrix with constant entries, a condition that requires the statistical independence of the components of the sufficient statistics $s(\mathbf{y}, \mathbf{x})$ with respect to the data-generating distribution.

In short, Proposition 5.27 shows how the $\ell_2$-norm of the parameters of the model upper bounds its inverse rate and, then, it shows why biasing the optimizer towards models with smaller $\ell_2$-norm tends to reduce the generalization error of the interpolators. However, Corollary 5.28 exemplifies a general setting to understand the gap in the inequality of Proposition 5.27 and how this gap can manifest if $\text{Cov}_\nu(s(\mathbf{y}, \mathbf{x}))$, a distribution-dependent quantity, is very different from a diagonal matrix.

**Distance From Initialization** The distance from initialization is known to be related to the generalization error of an interpolator (Nagarajan and Kolter, 2017). In fact, it has been successfully used as a regularizer in Hu et al., 2020. In these works, an implicit assumption was that the initial models are produced through one of the well-established initialization schemes used in neural networks, such as Xavier or He initialization (Goodfellow et al., 2016). These parameter initialization schemes provide an initial set of weights which guarantee that the variance of the activations across the different layers remains constant. In consequence, the randomly initialized model makes almost constant predictions. If $\boldsymbol{\theta}_0$ denotes the initial parameter, then, $\mathbb{V}_\nu(\ell(\mathbf{y}, \mathbf{x}, \boldsymbol{\theta}_0))$ should be very small. Figure 5.10 (right) shows how this is the case for a randomly initialized Inception model using He initialization.

Under the assumption that the initial model obtained by common initializers belongs to $\mathbf{\Theta}_0$, Proposition 5.27 also shows a link between distance from initialization and the inverse rate function and, in consequence, with the generalization error of a model. The same reasoning used with $\ell_2$-norm applies here: by promoting models with small distance from initialization, models with small generalization error are promoted.

Furthermore, as happens with the $\ell_2$-norm, in the context of over-parameterized interpolators, distance from initialization is not a reliable proxy for the generalization error (Y. Jiang et al., 2020). Again, this is easy to explain considering that distance from initialization is just an upper bound over the inverse rate. In consequence, an open question in machine learning is:

**Open Question 7.** When does the distance from initialization of an interpolator perfectly correlate with its generalization error?

When the cumulant generating function of the model $J_{\boldsymbol{\theta}}(\lambda)$ is a quadratic function or, equivalently, its second-order Taylor approximation around $\boldsymbol{\theta}_0 \in \boldsymbol{\Theta}_0$ is exact, as shown in the proof of Proposition 5.29, the cumulant generating function can be expressed as

$$J_{\boldsymbol{\theta}}(\lambda) = \frac{1}{2}\lambda^2 (\boldsymbol{\theta} - \boldsymbol{\theta}_0)^T \mathrm{Cov}_\nu (\nabla_{\boldsymbol{\theta}} \log p(\mathbf{y}|\mathbf{x}, \boldsymbol{\theta}_0)) (\boldsymbol{\theta} - \boldsymbol{\theta}_0)\,, \tag{5.79}$$

where $\mathrm{Cov}_\nu(\cdot)$ is the covariance w.r.t. $\nu$ of the gradient of the log-likelihood of each $(\mathbf{x}, \mathbf{y})$ sample. Under this hypothesis, the inverse rate can be expressed as stated next.

**Proposition 5.29.** *For any $\boldsymbol{\theta} \in \boldsymbol{\Theta}$, if the equality of Equation (5.79) holds, then*

$$\mathcal{I}_{\boldsymbol{\theta}}^{-1}(\tfrac{1}{n}\log\tfrac{kp}{\delta}) = \sqrt{2\tfrac{1}{n}\log\tfrac{kp}{\delta}}\sqrt{(\boldsymbol{\theta} - \boldsymbol{\theta}_0)^T Cov_\nu (\nabla_{\boldsymbol{\theta}} \log p(\mathbf{y}|\mathbf{x}, \boldsymbol{\theta}_0)) (\boldsymbol{\theta} - \boldsymbol{\theta}_0)}\,. \tag{5.80}$$

[Proof in Appendix C.5]

When training highly over-parameterized models using gradient descent, it is well known that parameters remain close to their initial values. This phenomenon is called lazy training and is a standard assumption in many theoretical analyses of deep neural networks (Jacot et al., 2018). However, when lazy training takes place, the second-order Taylor approximation of the cumulant could become highly accurate. The above proposition then suggests how the Riemannian distance, induced by the covariance matrix $\mathrm{Cov}_\nu(\nabla_{\boldsymbol{\theta}} \log p(\mathbf{y}|\mathbf{x}, \boldsymbol{\theta}_0))$, would be an optimal proxy for the generalization error of an interpolator in the lazy training regime. This reasoning also helps to understand the limitations in this regard of the Euclidean distance.

**Input-Gradient and Lipschitz Regularization.** Input-gradient regularization is a technique used in deep learning to improve the robustness of neural network models (Gouk et al., 2021). This approach focuses on the gradients of the model's output with respect to its input, which are indicative of how sensitive the model's predictions are to changes in the input data.

The following result shows how the inverse rate function and the expected norm of the input-gradient of a model are also related. This result is based on the *log-sobolev inequalities* (Chafaï, 2004) and relies on the assumption that $\nu(\mathbf{y}|\mathbf{x})$ is deterministic, to simplify gradients with respect to the target variable, and that $\nu(\mathbf{x}, \mathbf{y})$ is a uniformly strictly log-concave density, an assumption satisfied by a wide range of distributions.

**Proposition 5.30.** *If $\nu(\mathbf{x}, \mathbf{y})$ is a uniformly strictly log-concave density and $\nu(\mathbf{y}|\mathbf{x})$ is deterministic, then $\exists M > 0$, such that, for any $\boldsymbol{\theta} \in \boldsymbol{\Theta}$,*

$$\mathcal{I}_{\boldsymbol{\theta}}^{-1}(\tfrac{1}{n}\log\tfrac{k^p}{\delta}) \leq \sqrt{\tfrac{1}{n}\log\tfrac{k^p}{\delta}}\sqrt{M\mathbb{E}_\nu\left[\|\nabla_{\mathbf{x}}\ell(\mathbf{y}, \mathbf{x}, \boldsymbol{\theta})\|_2^2\right]}\,. \tag{5.81}$$

[Proof in Appendix C.5]

The above result explains why models with a small input-gradient norm tend to have a small generalization error. And why a small generalization error may coexist with a large input-gradient norm if the gap in the above inequality is large.

From the above result, it is straightforward to derive a connection between the generalization of a model and its Lipschitz constant. If a model $\boldsymbol{\theta}$ is Lipschitz with respect to the input data $\mathbf{x}$ with Lipschitz constant denoted $\mathrm{Lip}(\boldsymbol{\theta})$, then, by definition, the following inequality is verified:

$$\forall(\mathbf{x}, \mathbf{y}) \in \mathrm{supp}(\nu) \quad \|\nabla_{\mathbf{x}}\ell(\mathbf{y}, \mathbf{x}, \boldsymbol{\theta})\|_2^2 \leq \mathrm{Lip}(\boldsymbol{\theta})\,. \tag{5.82}$$

The above inequality can be combined with Proposition 5.30 to derive a new upper bound on the inverse of the rate function and on the generalization error of a model:

$$\mathcal{I}_{\boldsymbol{\theta}}^{-1}(\tfrac{1}{n}\log\tfrac{k^p}{\delta}) \leq \sqrt{\tfrac{1}{n}\log\tfrac{k^p}{\delta}}\sqrt{M\mathrm{Lip}(\boldsymbol{\theta})}\,. \tag{5.83}$$

In consequence, models with a small Lipschitz constant will tend to have a small generalization error. This aligns fully with the existing literature (Neyshabur et al., 2017b) and, also, explains the rationale behind regularization techniques that prevent overfitting by controlling the Lipschitz constant (Gouk et al., 2021).

**Summary.** In this section, an optimal (distribution-dependent) regularizer for interpolators has been theoretically characterized, even for over-parameterized model classes. This theoretical characterization can be used to understand why a wide range of commonly used regularization techniques promotes interpolators with a smaller generalization error.

In some specific but not less relevant situations, it has been analyzed when some of these common regularization methods could be optimal for interpolators; even in over-parameterized model classes. In this regard, this thesis complements other related theoretical results, like the ones proposed by Bartlett et al., 2020, where the Euclidean norm was shown to be an optimal regularizer for linear regression interpolators under some specific assumptions.

This section draws connections between the proposed definition of smoothness, as outlined in Section 5.3.4, and other existing definitions in the literature. The *smoothest* *interpolator* is shown to have near-optimal performance in Proposition 5.26. The clearest connection with existing measures of smoothness arises from the characterization of the Lipschitz constant, as demonstrated by combining Equation (5.82) and Proposition 5.30. The Lipschitz constant is frequently cited in the literature as a measure of a model's *smoothness* (Gouk et al., 2021). In this thesis, it has been shown how models with a small Lipschitz constant have a small inverse rate and, in consequence, are smoother according to the proposed definition.

### 5.3.7 Invariances

In many machine learning settings, the observed inputs $\mathbf{x}$ are transformed versions of underlying signals, typically as a consequence of the measurement process. For instance, sensor readings may be corrupted by random noise from imperfect hardware, and images may be rotated, blurred, or otherwise distorted during acquisition (Bronstein et al., 2021). This section interprets two widely used strategies—data augmentation (Shorten and Khoshgoftaar, 2019) and invariant architectures (Bronstein et al., 2021)—through the lens of PAC–Chernoff bounds and the rate function, viewing them as complementary mechanisms for improving learning under transformed inputs.

Let $G$ denote a set of input-data transformations such that any $g \in G$ denotes a function, $g : \mathcal{X} \to \mathcal{X}$, defining a specific transformation. Assume there exists an *unknown distribution* $h$ over $G$. In image classification, $G$ could be the set of all possible rotations, translations and/or reflections.

The central hypothesis in this section is that the generative process of the data follows the framework defined in the previous paragraphs.

**Assumption 5.31.** The data-generating distribution has the following structure:

$$(i)\ \mathbf{x}_0 \sim \nu_0(\mathbf{x}_0), \quad (ii)\ y \sim \nu(\mathbf{y}|\mathbf{x}_0), \quad (iii)\ g \sim h(g), \quad (iv)\ \mathbf{x} = g(\mathbf{x}_0)\,, \tag{5.84}$$

where $\mathbf{x}_0$ denotes the untransformed and unobserved input, which is distributed according to $\nu_0$, $\mathbf{x}$ denotes the observed (transformed) input. See Figure 5.15 (left) for a graph representation.

Under this assumption, the target value of an input is sampled *before* its transformation, that is, *transforming the inputs does not affect its label*. In statistical terms, under Assumption 5.31, the target random variable $\mathbf{y}$ is conditionally independent of $\mathbf{x}$ given $\mathbf{x}_0$. This fact perfectly aligns with many common settings, where, for example, rotating, reflecting or cropping an image does not alter the label of the object being represented in it (Bronstein et al., 2021). This assumption is weaker than the usual ones used in many related works (Bronstein et al., 2021; S. Chen et al., 2020), where the conditional generating distribution is assumed to be invariant under any transformation. These works assume that $\forall g \in G$, $\nu(\mathbf{y}|\mathbf{x}) = \nu(\mathbf{y}|g(\mathbf{x}))$. This assumption does not hold, for example, when considering rotation over digits, as the *true label* of the image of a *six* rotated 180 degrees *changes to a nine*.

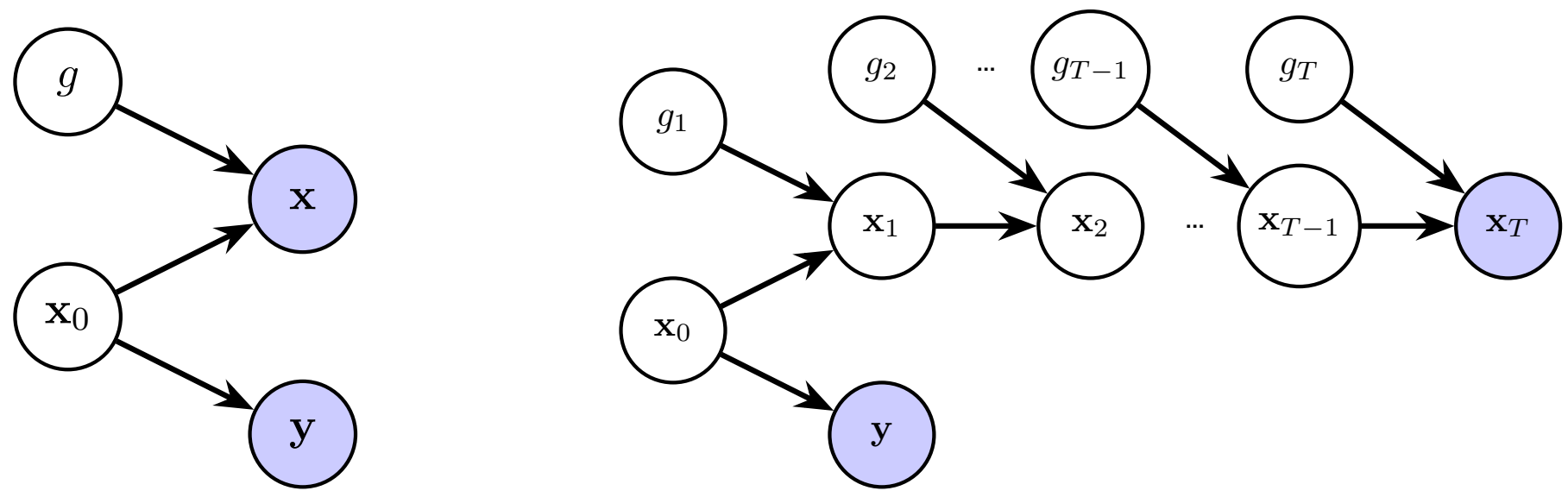

**Figure 5.15:** ***Graph representation of Assumption 5.31.*** *Left: original case with a single transformation $g$ producing the observable $\mathbf{x}$ ($g \to \mathbf{x}$), and a target $\mathbf{y}$ receiving inputs from $x$ and a nuisance variable $\mathbf{x}_0$ ($\mathbf{x} \to \mathbf{y}$, $\mathbf{x}_0 \to \mathbf{y}$). Right: generalized case with a sequence of transformations $g_1, \ldots, g_T$ generating intermediates $\mathbf{x}_1, \ldots, \mathbf{x}_T$ via $\mathbf{x}_{t-1} \to \mathbf{x}_t$ and $g_t \to \mathbf{x}_t$; the target $\mathbf{y}$ depends on $\mathbf{x}_0$ ($\mathbf{x}_0 \to \mathbf{y}$). Observed variables are highlighted with a blue background ($\mathbf{x}, \mathbf{y}$ on the left; $\mathbf{x}_T, \mathbf{y}$ on the right); unobserved variables are unshaded.*

Consider the generalized scenario where several transformations can be applied, that is, $g \in G$ is the composition of other transformations $g(\mathbf{x}_0) = g_T \circ \cdots \circ g_1(\mathbf{x}_0) = \mathbf{x}$. Let $G_1, \ldots, G_T$ denote the sets of all possible individual transformations, with $G = \cup_t G_t$. Notice that the value of $T$ is arbitrary and can differ from one transformation $g \in G$ to another; however, to simplify the notation, the same $T$ will be used consistently. where one may consider that some $g_i \in G_i$ are the identity. Then, consider the non-fully-transformed inputs $\mathbf{x}_t = g_t \circ \cdots \circ g_1(\mathbf{x}_0)$, with $0 < t < T$. The following result shows that the mutual information between the targets and the inputs, denoted $MI(\mathbf{y}\,;\mathbf{x})$, decreases the further the input is transformed.

**Proposition 5.32.** *Under Assumption 5.31, let $\mathbf{x}_t$ denote the random variable denoting a transformation of an input $\mathbf{x}_0$ using a composition of $t$ transformations, then*

$$MI(\mathbf{y}\,;\mathbf{x}_t) \leq MI(\mathbf{y}\,;\mathbf{x}_{t-1}) \quad \forall t \in \mathbb{N}^+ . \tag{5.85}$$

*As a consequence, $MI(\mathbf{y};\mathbf{x}) = MI(\mathbf{y};\mathbf{x}_T) \leq MI(\mathbf{y};\mathbf{x}_0)$.* [Proof in Appendix C.5]

The rate function and PAC-Chernoff bounds can be directly applied to explain why learning on transformed data $\mathbf{x}$ makes interpolators have worse generalization error compared to "ideally" learning on the untransformed data $\mathbf{x}_0$, which is never observed. In the following sections, this argument is used to explain the reason why invariant architectures and data augmentation promote models with better generalization errors by exploiting the fact that the inputs are transformed.

**Open Question 8.** Why do models interpolating transformed data, as described in Assumption 5.31, have a higher generalization error?

For a fixed data-generating distribution, the law of $\hat{L}(D, \boldsymbol{\theta})$ governs the generalization error of a model $\boldsymbol{\theta} \in \boldsymbol{\Theta}$. Proposition 5.32 shows that learning on transformed inputs reduces the information that $\mathbf{x}$ carries about its target $\mathbf{y}$. Consequently, under transformed inputs, $\hat{L}(D, \boldsymbol{\theta})$ becomes less concentrated and its expectation increases. Figure 5.16 illustrates this phenomenon for a multi-layer perceptron and an Inception network on CIFAR-10. The left panel displays the distributions of $\hat{L}(D_0, \boldsymbol{\theta})$, $\hat{L}(D_1, \boldsymbol{\theta})$, and $\hat{L}(D_2, \boldsymbol{\theta})$ for datasets of size 50, where $D_0$ is untransformed, $D_1$ applies random translations to $D_0$, and $D_2$ further applies random rotations to $D_1$; these correspond to draws from $\nu_0, \nu_1$, and $\nu_2$, respectively. Transforming the inputs once (translations) and twice (translations + rotations) increases both the mean loss and its dispersion, visible in the widening box-plots. As shown in the center and right panels, the reduced concentration of $\hat{L}(\cdot, \boldsymbol{\theta})$ under transformed inputs is mirrored by smaller rate functions. Note that a fixed model can exhibit different rate functions across data-generating distributions, since the rate function depends on both the model and the distribution. The figure also indicates that the adverse effect of input transformations can be partially mitigated by invariant architectures: for example, the Inception network's convolutional layers confer translation invariance (improving behavior under $\nu_1$) but not rotation invariance (leaving performance under $\nu_2$ relatively degraded). The next subsection elaborates on this point.

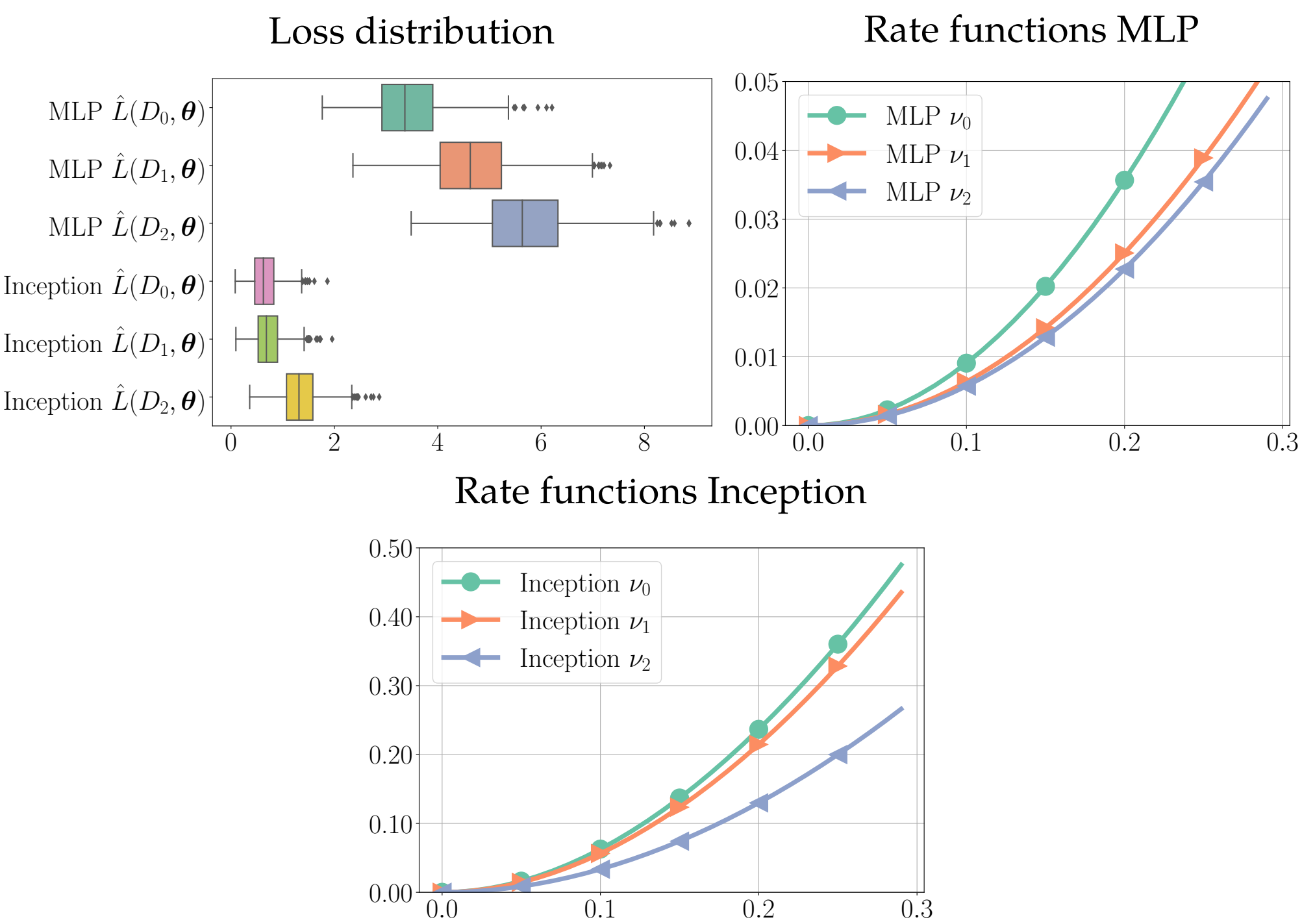

**Figure 5.16:** ***Loss distributions and rate-function curves for a fixed MLP and a fixed Inception network under input transformations.*** *The figure has three panels.* Left: *box-plots of the empirical loss $\hat{L}(D, \theta)$ for samples of size $n = 50$ drawn from three input distributions: $D_0 \sim \nu_0^{50}$ (CIFAR-10 test inputs), $D_1 \sim \nu_1^{50}$ (random translations by up to 3 pixels), and $D_2 \sim \nu_2^{50}$ (the same translations plus rotations up to $20°$). Results are shown for both models with comparable parameter counts.* Right: *estimated rate functions $\mathcal{I}_{\boldsymbol{\theta}}(a)$ for the MLP under $\nu_0, \nu_1, \nu_2$.* Bottom: *the corresponding $\mathcal{I}_{\boldsymbol{\theta}}(a)$ for the Inception model. Legends identify the distribution; in the rate plots the horizontal axis is the deviation $a$ and the vertical axis is $\mathcal{I}_{\boldsymbol{\theta}}(a)$.*

The changes observed in the distribution of the empirical loss $\hat{L}(\cdot, \boldsymbol{\theta})$ under transformed inputs can be formalized within the present framework by introducing suitable assumptions. Although these assumptions are not generally verifiable in practice, the obtained results provide *theoretical justification* for why input transformations make the distribution of $\hat{L}(\cdot, \boldsymbol{\theta})$ less concentrated and increase its expectation. Formally, let $\nu_t$ denote the marginal distribution of inputs transformed $t$ times, $\mathbf{x}_t = g_t \circ \cdots \circ g_1(\mathbf{x}_0)$, and write $L^{\nu_t}(\boldsymbol{\theta})$ and $L^{\nu_{t+1}}(\boldsymbol{\theta})$ for the expected loss of model $\boldsymbol{\theta}$ under $\nu_t$ and $\nu_{t+1}$, respectively. Likewise, let $\mathcal{I}_{\boldsymbol{\theta}}^{\nu_t}(a)$ and $\mathcal{I}_{\boldsymbol{\theta}}^{\nu_{t+1}}(a)$ denote the corresponding rate functions.

**Proposition 5.33.** *Under Assumption 5.31, if $\ell(\mathbf{y}, \mathbf{x}, \boldsymbol{\theta})$ is convex under $\mathbf{x}$ and $\mathbb{E}_{g_t}[g_t(\mathbf{x})] = \mathbf{x}$, then, $\forall \boldsymbol{\theta} \in \boldsymbol{\Theta},\ L^{\nu_{t+1}}(\boldsymbol{\theta}) \geq L^{\nu_t}(\boldsymbol{\theta})$.* [Proof in Appendix C.5]

The next result explains why the empirical loss becomes less concentrated when inputs are transformed as in Assumption 5.31. For a fixed datum $\mathbf{x}$, a model incurs loss $\ell(\mathbf{y}, \mathbf{x}, \boldsymbol{\theta})$; after applying a random transformation $g \sim h$ to the same input, the incurred loss is $\ell(\mathbf{y}, g(\mathbf{x}), \boldsymbol{\theta})$, which may be larger or smaller than the original value. Define the relative change

$$\Delta(\mathbf{y}, \mathbf{x}, g, \boldsymbol{\theta}) := \ell(\mathbf{y}, g(\mathbf{x}), \boldsymbol{\theta}) - \ell(\mathbf{y}, \mathbf{x}, \boldsymbol{\theta}). \tag{5.86}$$

If $\Delta(\mathbf{y}, \mathbf{x}, g, \boldsymbol{\theta})$ is statistically independent of the baseline loss $\ell(\mathbf{y}, \mathbf{x}, \boldsymbol{\theta})$, then the empirical loss under transformed inputs is less concentrated. Intuitively, the magnitude of the loss change induced by the transformation does not depend on the original (untransformed) loss, thereby increasing variability in the aggregate.

**Proposition 5.34.** *Under Assumption 5.31, if the change observed in the loss after a transformation $\Delta(\mathbf{y}, \mathbf{x}, g, \boldsymbol{\theta}) = \ell(\mathbf{y}, g(\mathbf{x}), \boldsymbol{\theta}) - \ell(\mathbf{y}, \mathbf{x}, \boldsymbol{\theta})$ is statistically independent of the loss itself, $\Delta(\mathbf{y}, \mathbf{x}, g, \boldsymbol{\theta}) \perp \ell(\mathbf{y}, \mathbf{x}, \boldsymbol{\theta})$, then*

$$\mathcal{I}_{\boldsymbol{\theta}}^{\nu_{t+1}}(a) \leq \mathcal{I}_{\boldsymbol{\theta}}^{\nu_t}(a) \quad \forall a > 0\,. \tag{5.87}$$

[Proof in Appendix C.5]

Summarizing, when input-data is transformed (Assumption 5.31), the input-data provides less information about the target variable (Proposition 5.32). This causes the distribution of the empirical loss $\hat{L}(\cdot, \boldsymbol{\theta})$ to be less concentrated and with higher expected loss $L(\boldsymbol{\theta})$ (as empirically shown in Figure 5.16 and theoretically argued in Propositions 5.33 and 5.34). *Transformed input-data makes the expected loss of the model $L(\boldsymbol{\theta})$ higher and the distribution of $\hat{L}(\cdot, \boldsymbol{\theta})$ less concentrated.*

The distribution-dependent bound given in Theorem 5.17 can be extended to the notation used in this section. With high probability over random draws $D_t \sim \nu_t^n$, simultaneously for all $\boldsymbol{\theta} \in \boldsymbol{\Theta}$, then:

$$\text{if } \hat{L}(D_t, \boldsymbol{\theta}) \leq \epsilon \text{ then } \left(\mathcal{I}_{\boldsymbol{\theta}}^{\nu_t}\right)^{-1}\left(\tfrac{1}{n} \log \tfrac{k^p}{\delta}\right) \leq L^{\nu_t}(\boldsymbol{\theta}) \leq \left(\mathcal{I}_{\boldsymbol{\theta}}^{\nu_t}\right)^{-1}\left(\tfrac{1}{n} \log \tfrac{k^p}{\delta}\right) + \epsilon\,. \tag{5.88}$$

The bound above clarifies why transformed inputs lead interpolators to exhibit larger generalization error. Transformations reduce the concentration of $\hat{L}(D_t, \boldsymbol{\theta})$ and thereby decrease the rate function (see Figure 5.16 and Proposition 5.34). Consequently, the inverse rate function increases, and so does $(\mathcal{I}_{\boldsymbol{\theta}}^{\nu_t})^{-1}(\frac{1}{n} \log \frac{k^p}{\delta})$. Since $\epsilon$ is assumed to be very small, the generalization error of the interpolator increases accordingly. *In* *summary, the PAC–Chernoff bound explains why interpolators trained on transformed inputs incur higher generalization error.*

#### 5.3.7.1 Invariant Architectures

Invariant architectures refer to a kind of neural networks that remain unaffected by certain transformations in the input data (Bronstein et al., 2021). In the case of images, these transformations might include scaling, rotation, translation, or other geometrical changes.

**Definition 5.35.** Given a set of transformations $G$, a model $\boldsymbol{\theta} \in \boldsymbol{\Theta}$ is said $G$-invariant if

$$P(\mathbf{y}|\mathbf{x}, \boldsymbol{\theta}) = P(\mathbf{y}|g(\mathbf{x}), \boldsymbol{\theta}) \quad \forall g \in G \;\; \forall (\mathbf{x}, \mathbf{y}) \in \mathcal{X} \times \mathcal{Y}\,. \tag{5.89}$$

Invariant architectures—such as convolutional neural networks (CNNs) (LeCun et al., 1998)—are crucial because they recognize and process patterns independently of position, scale, or orientation. This property is especially important in applications like image and speech recognition, where the salient structure must be captured despite such variations. Empirically, invariant architectures have consistently outperformed non-invariant counterparts (Goodfellow et al., 2016).

The generalization behavior of invariant neural networks has been extensively studied. Several bounds include complexity terms that shrink when the model architecture is invariant (Behboodi et al., 2022; Elesedy, 2022; Sokolic et al., 2017). While these results offer theoretical support for improved generalization with invariance, they primarily establish reductions in complexity surrogates rather than demonstrably tighter bounds on the generalization error itself. Moreover, as argued in Section 5.3, there is accumulating empirical (C. Zhang et al., 2017) and theoretical evidence (Gastpar et al., 2024; Nagarajan and Kolter, 2019b; Y. Wang et al., 2024) that many such bounds are non-tight—or even vacuous—and can violate basic desiderata (e.g., failing to decrease with larger sample sizes, as in Behboodi et al., 2022). Consequently, to the best of current knowledge, there is no widely accepted, specific generalization bound that comprehensively captures the effect of architectural invariance on an interpolator's generalization error, particularly in over-parameterized regimes. This raises the following open question:

**Open Question 9.** Why do (over-parameterized) invariant interpolators generalize better?

While a model's final generalization error is influenced by additional factors (e.g., optimization dynamics), a partial answer to the foregoing open question can be obtained by analyzing the rate function of invariant architectures. Under transformed inputs (Assumption 5.31), the key advantage of invariant models is that they are *unaffected* by input transformations and thus evade the issues identified in the preceding subsection. This observation can be formalized in the following result.

| Model | Train Acc. | Test Acc. | Test NLL |
|---|---|---|---|
| Inception | 100.0% | 74.08% | 1.00 |
| Inception-Shuffle | 100.0% | 42.46% | 2.45 |
| MLP | 99.99% | 51.69% | 3.29 |
| MLP-Shuffle | 99.99% | 51.12% | 3.29 |
| Initial Inception | 10.00% | 10.00% | 2.30 |
| Initial MLP | 10.00% | 9.96% | 2.30 |

**Figure 5.17: *Metrics and rate-function curves for a random input–shuffling experiment with MLP and Inception models.*** Left: *a table reporting training accuracy, test accuracy and test negative log-likelihood (NLL) for six models: Inception, Inception–Shuffle, MLP, MLP–Shuffle, and the corresponding untrained baselines Initial Inception and Initial MLP. "Shuffle" denotes the use of a fixed random shuffling of the inputs in both the training and test sets.* Right: *plots of the estimated rate functions $\mathcal{I}_{\boldsymbol{\theta}}(a)$ for the same models with respect to their matching (shuffled or un-shuffled) data distributions; the horizontal axis is the deviation $a$ and the vertical axis is $\mathcal{I}_{\boldsymbol{\theta}}(a)$. Legends identify the curves.*

**Proposition 5.36.** *Under Assumption 5.31, if a model $\boldsymbol{\theta} \in \boldsymbol{\Theta}$ is $G_t$-invariant then,*

$$L^{\nu_{t+1}}(\boldsymbol{\theta}) = L^{\nu_t}(\boldsymbol{\theta}) \quad \textit{and} \quad \mathcal{I}_{\boldsymbol{\theta}}^{\nu_{t+1}}(a) = \mathcal{I}_{\boldsymbol{\theta}}^{\nu_t}(a) \quad \forall a > 0\,. \tag{5.90}$$

[Proof in Appendix C.5]

As argued above, transforming the inputs makes the empirical loss $\hat{L}(\cdot, \boldsymbol{\theta})$ less concentrated (i.e., with a smaller rate function) and increases its expectation. The preceding result shows that invariant architectures circumvent this issue: for such models, the distribution of the empirical loss is unaffected by transformed inputs.

Figure 5.16 illustrates this effect. Recall that $D_0 \sim \nu_0^{50}$ denotes the original input distribution, and $D_1 \sim \nu_1^{50}$ the distribution obtained by applying random translations to $D_0$. As expected, for a fixed MLP—whose architecture is not translation invariant—the expected loss increases markedly and the rate function decreases substantially when moving from $D_0$ to $D_1$. By contrast, for a fixed *Inception* model, the changes are minimal: the model does not suffer the pronounced degradation observed for the MLP. The small residual difference between results on $D_0$ and $D_1$ for *Inception*, despite Proposition 5.36, arises because the experimental translations were implemented via padding (e.g., zero-padding), which introduces slight output differences even for architectures with convolution and max-pooling.

The PAC–Chernoff bound in Equation (5.88) captures this phenomenon precisely. As argued above, the presence of transformed inputs implies

$$(\mathcal{I}_{\boldsymbol{\theta}}^{\nu_1})^{-1}\left(\tfrac{1}{n}\log\tfrac{k^p}{\delta}\right) \geq (\mathcal{I}_{\boldsymbol{\theta}}^{\nu_0})^{-1}\left(\tfrac{1}{n}\log\tfrac{k^p}{\delta}\right). \tag{5.91}$$

However, if the model $\boldsymbol{\theta}$ is invariant to the transformations present under $\nu_1$, then, by Proposition 5.36, the complexity term in the bound does not increase. Combined with the fact that the distribution-dependent PAC–Chernoff bound is tight for interpolators, this provides, to the best of current knowledge, the most compelling theoretical expla-nation for why invariant models that interpolate the training data achieve lower generalization error. *The PAC-Chernoff bound explains why invariant interpolators have smaller generalization errors than non-invariant interpolators under transformed inputs.*

Figure 5.17 illustrates this behavior from a complementary perspective. Although the full set of invariances of an Inception network is not precisely known, convolutional layers confer at least translation invariance. To ablate this property, the experiment applies a fixed random permutation of image pixels—interpretable as an additional input transformation $g_{T+1}$ composed on top of the unknown sequence of transformations that generate CIFAR-10, $\nu_T = g_T \circ \cdots \circ g_1(\nu_0)$. Under the resulting distribution $\nu_{T+1}$, a convolutional architecture such as *Inception* ceases to be translation invariant. According to the analysis at the beginning of this section, the empirical loss should become less concentrated under $\nu_{T+1}$, and the rate function should therefore decrease relative to $\nu_T$. This effect is visible in the figure: the learned Inception models exhibit larger rate functions under $\nu_T$ (label *Inception*) than under $\nu_{T+1}$ (label *Inception–Shuffle*). Rate functions are computed with respect to the log-loss or negative log-likelihood (NLL).

The same procedure is applied to an MLP, which possesses no local invariances. As expected, its rate function is essentially unaffected by the pixel permutation. Although a given MLP is not invariant to shuffling, the *architecture* is equivariant in the sense that, for any MLP $\boldsymbol{\theta}$ and permutation $g_{T+1}$, there exists another MLP $\boldsymbol{\theta}'$ (obtained by permuting the first-layer weights) that yields the same predictions on $\nu_{T+1}$ as $\boldsymbol{\theta}$ does on $\nu_T$. Consequently, a learning algorithm can recover "the same" MLP under $\nu_T$ and $\nu_{T+1}$, leading to (nearly) identical rate functions in the experiment. The accompanying table further shows that non-invariant models incur substantially higher generalization error, in agreement with the theoretical predictions.

#### 5.3.7.2 Data Augmentation

Data augmentation is a widely-used technique in machine learning that enhances the robustness and generalization ability of models, particularly in the field of computer vision (Shorten and Khoshgoftaar, 2019). By applying various transformations like rotations, reflections, scaling, and translations to existing data samples, data augmentation artificially expands the dataset. This process helps the learned models to become invariant or less sensitive to these transformations, leading to improved performance on unseen data.

Recent research has delved deeply into the theoretical foundations of data augmentation (DA), with significant contributions coming from studies such as S. Chen et al., 2020 and Lyle et al., 2020. These works build on the assumption that the set of transformations $G$ forms a *group* and that these transformations do not alter the data-generating distribution, also assuming *an equality in distribution assumption*. Within this context, these studies illustrate how DA can decrease variance and improve bounds on generalization error. Lyle et al., 2020 further presents a PAC-Bayes bound based on these premises, showing that data augmentation can lower the bound's complexity mea-sure. However, these studies encounter a critical challenge: the *equality in distribution* assumption does not hold in reality, as it implies that any model $\boldsymbol{\theta}$ would yield identical expected losses on both transformed and non-transformed input data, a premise contradicted by evidence shown in Figure 5.16, where two models exhibit deteriorating performance as the data is increasingly transformed. Additionally, the PAC-Bayes bound introduced by Lyle et al., 2020 is not proven to be tight for interpolators. Thus, as highlighted in the introduction, merely demonstrating that data augmentation reduces the complexity term of a bound, which might be vacuous, does not assure a lower generalization error for an interpolator.

In this section, a more nuanced account is provided within the proposed framework. First, it is shown that data augmentation induces a *more concentrated* empirical loss, aligning with prior observations that augmentation reduces the variance of empirical risk (S. Chen et al., 2020; Lyle et al., 2020). Second, it is established that the complexity term in the PAC–Chernoff bound of Theorem 5.17 decreases under augmentation. Because this bound is perfectly tight for interpolators trained on augmented data, it follows that data augmentation reduces the generalization error of interpolators—by contrast with earlier analyses based on non-tight bounds (Lyle et al., 2020).

As argued, for example, by S. Chen et al., 2020, data augmentation is equivalent to optimizing, using Monte-Carlo estimates, the so-called *data-augmented loss*, denoted as $\ell_G$, which is the result of averaging the standard loss over all transformations obtained from a given input $\mathbf{x}$,

$$\ell_G(\mathbf{y}, \mathbf{x}, \boldsymbol{\theta}) = \mathbb{E}_{g\sim h}\left[\ell(\mathbf{y}, g(\mathbf{x}), \boldsymbol{\theta})\right] . \tag{5.92}$$

A natural question is whether the augmented loss $\ell_G$ yields a more concentrated empirical loss (i.e., a larger rate function), since greater concentration implies reduced generalization error for interpolators. The answer is nuanced. To streamline the discussion, the principal points are summarized below:

1. Each transformed dataset $D_T$ corresponds to an underlying, unaltered dataset $D_0$, from which $D_T$ is obtained via random transformations $g_t \sim h_t$ for $t = 1, \dots, T-1$. Two empirical losses are considered:
   (a) The conventional empirical loss on transformed inputs, denoted $\hat{L}^{\ell}(D_T, \boldsymbol{\theta})$ to emphasize that $D_T \sim \nu_T^n$ and the standard loss $\ell$ is used.
   (b) The augmented empirical loss on the original, untransformed inputs, denoted $\hat{L}^{\ell_G}(D_0, \boldsymbol{\theta})$, where $D_0 \sim \nu_0^n$ and

   $$\hat{L}^{\ell_G}(D_0, \boldsymbol{\theta}) = \frac{1}{n} \sum_{(\mathbf{x}_i, \mathbf{y}_i) \in D_0} \ell_G(\mathbf{y}_i, \mathbf{x}_i, \boldsymbol{\theta}), \tag{5.93}$$

   with $\ell_G$ defined in Equation (5.92).

   Note that $\hat{L}^{\ell_G}(D_0, \boldsymbol{\theta})$ is unobserved because $D_0$ is not available, yet it plays a central role in explaining the mechanics of data augmentation.
2. Theorem 5.37 shows—via the rate function—that the augmented empirical loss

on untransformed inputs, $\hat{L}^{\ell_G}(D_0,\boldsymbol{\theta})$, is more concentrated than the conventional empirical loss on transformed inputs, $\hat{L}^{\ell}(D_T,\boldsymbol{\theta})$. Since their expectations (population risks) coincide, minimizing $\hat{L}^{\ell_G}(D_0,\boldsymbol{\theta})$ constitutes a *better* optimization objective: by Chernoff's bound (Theorem 5.14), lower generalization errors are more likely.

In practice, $D_0$ is unavailable, and data augmentation optimizes $\hat{L}^{\ell_G}(D_T,\boldsymbol{\theta})$ rather than $\hat{L}^{\ell_G}(D_0,\boldsymbol{\theta})$. Therefore, it is necessary to relate $\hat{L}^{\ell_G}(D_T,\boldsymbol{\theta})$ to $\hat{L}^{\ell_G}(D_0,\boldsymbol{\theta})$.

3. Proposition 5.39 establishes that $\hat{L}^{\ell_G}(D_T,\boldsymbol{\theta}) = \hat{L}^{\ell_G}(D_0,\boldsymbol{\theta})$ when $G$ is a group and $h$ is uniform. Consequently, any interpolator of $\hat{L}^{\ell_G}(D_T,\boldsymbol{\theta})$ also interpolates $\hat{L}^{\ell_G}(D_0,\boldsymbol{\theta})$, which is more concentrated than $\hat{L}^{\ell}(D_T,\boldsymbol{\theta})$. Many practical augmentation schemes, however, fall outside this group setting (e.g., rotations within a finite range, random cropping).
4. Proposition 5.40 shows that $\hat{L}^{\ell_G}(D_T,\boldsymbol{\theta})$ and $\hat{L}^{\ell_G}(D_0,\boldsymbol{\theta})$ share the same optimal interpolators (i.e., minimizers with zero empirical error). Thus, if $\boldsymbol{\theta}$ perfectly interpolates $\hat{L}^{\ell_G}(D_T,\boldsymbol{\theta})$, then it also perfectly interpolates $\hat{L}^{\ell_G}(D_0,\boldsymbol{\theta})$. Such an interpolator is likely to achieve a lower generalization error than one minimizing $\hat{L}^{\ell}(D_T,\boldsymbol{\theta})$, since $\hat{L}^{\ell_G}(D_0,\boldsymbol{\theta})$ is more concentrated than $\hat{L}^{\ell}(D_T,\boldsymbol{\theta})$. By Chernoff's bound, an interpolator minimizing $\hat{L}^{\ell_G}(D_T,\boldsymbol{\theta})$ is therefore more likely to generalize better.

In what follows, $L^{\nu_0,\ell_G}(\boldsymbol{\theta})$ and $\mathcal{I}_{\boldsymbol{\theta}}^{\nu_0,\ell_G}(a)$ denote, respectively, the expected loss and the rate function associated with the augmented loss $\ell_G$ under the untransformed distribution $\nu_0$. Under this notation,

$$L(\boldsymbol{\theta}) = L^{\nu_T,\ell}(\boldsymbol{\theta}) \quad \text{and} \quad \mathcal{I}_{\boldsymbol{\theta}}(a) = \mathcal{I}_{\boldsymbol{\theta}}^{\nu_T,\ell}(a), \tag{5.94}$$

since these quantities are implicitly defined with respect to the standard loss $\ell$ and the observed (transformed) distribution $\nu_T$ from which the training sets are drawn.

**Theorem 5.37.** *Under Assumption 5.31, it verifies that*

$$\forall \boldsymbol{\theta} \in \boldsymbol{\Theta}, \quad L^{\nu_T,\ell}(\boldsymbol{\theta}) = L^{\nu_0,\ell_G}(\boldsymbol{\theta}) \quad \textit{and} \quad \mathcal{I}_{\boldsymbol{\theta}}^{\nu_T,\ell}(a) \leq \mathcal{I}_{\boldsymbol{\theta}}^{\nu_0,\ell_G}(a) \;\; \forall a > 0\,. \tag{5.95}$$

[Proof in Appendix C.5]

Figure 5.18 illustrates these findings for a fixed MLP and a fixed Inception model on CIFAR-10. The left panel shows that, for both architectures, the distribution of the empirical loss on transformed data, $\hat{L}^{\ell}(D_1,\boldsymbol{\theta})$ with $D_1 \sim \nu_1^{50}$, is more dispersed than the distribution on untransformed data, $\hat{L}^{\ell}(D_0,\boldsymbol{\theta})$ with $D_0 \sim \nu_0^{50}$. It also shows that the augmented empirical loss on untransformed data, $\hat{L}^{\ell_G}(D_0,\boldsymbol{\theta})$, is more concentrated while sharing the same expectation as $\hat{L}^{\ell}(D_1,\boldsymbol{\theta})$. The center panel further demonstrates that the rate functions of the Inception models across the three setups vary in the anticipated manner. Note that the models are fixed throughout; changes in the empirical-loss distributions arise solely from differences in the data-generating distribution and/or the loss function.

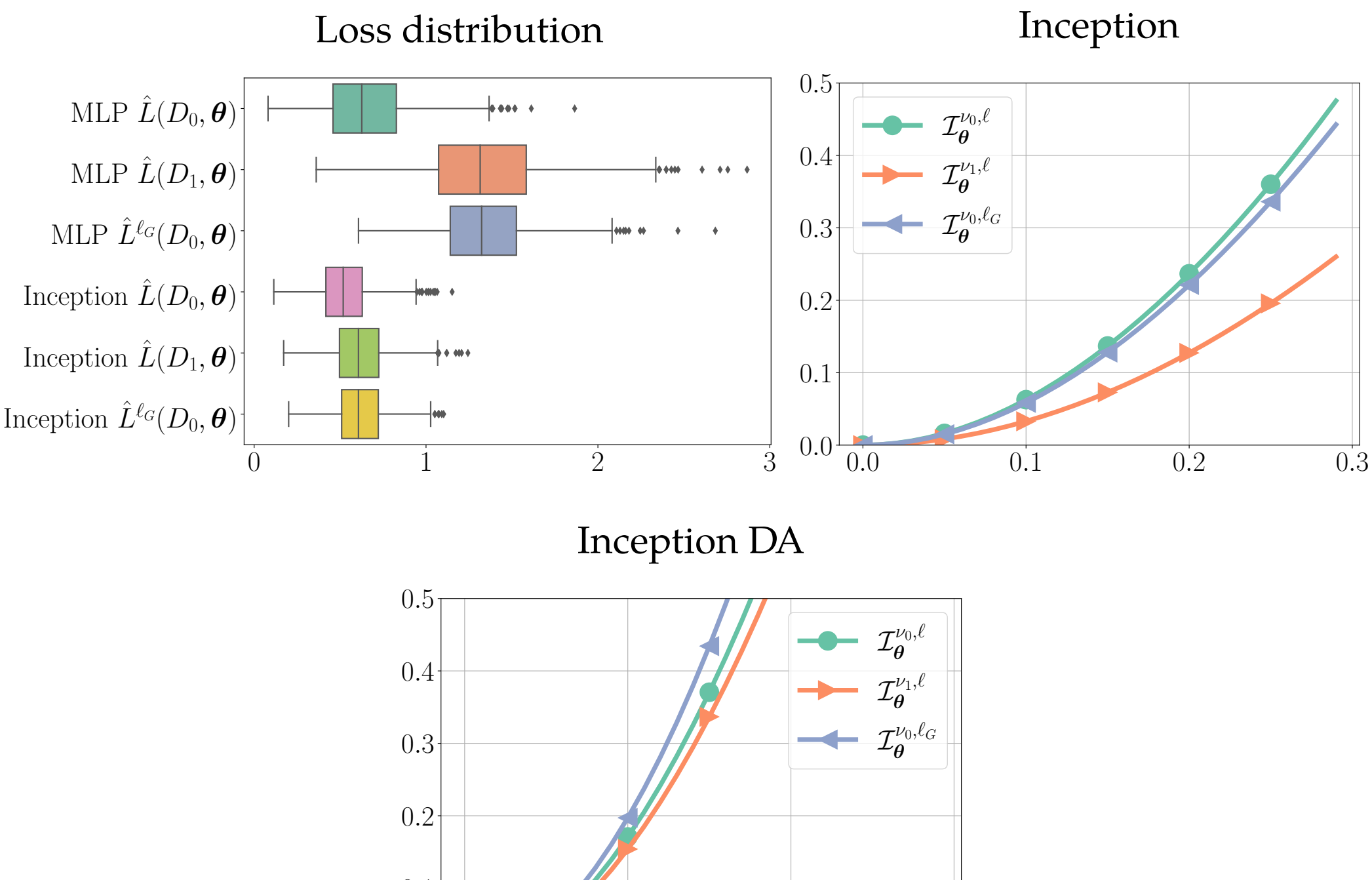

**Figure 5.18:** ***Loss distributions and rate–function curves under raw, transformed and augmentation–aware settings for MLP and Inception.*** *The figure has three panels.* Left: *box-plots of empirical losses for two fixed models (MLP and Inception) with comparable parameter counts under three sampling/criterion choices: $\hat{L}(D_0, \theta)$ with $D_0 \sim \nu_0^{50}$ (CIFAR-10 test inputs), $\hat{L}(D_1, \theta)$ with $D_1 \sim \nu_1^{50}$ obtained by translating inputs up to 3 pixels and rotating up to 20°, and $\hat{L}^{\ell_G}(D_0, \theta)$ where the augmented loss $\ell_G$ uses the same transformations as those generating $D_1$.* Right: *estimated rate functions for an Inception network evaluated with respect to $\nu_0$, $\nu_1$ and $\ell_G$.* Bottom: *the corresponding rate functions for an Inception network trained with the augmentation-aware loss $\ell_G$ (labeled "Inception DA").*

The following result is an adaptation of the PAC-Chernoff bound given in Theorem 5.17 for this setup, which describes the effect of using the data-augmented loss on the generalization error of interpolators.

**Corollary 5.38.** *With h.p. $1-\delta$ over $D_0 \sim \nu_0^n$, for all $\boldsymbol{\theta} \in \boldsymbol{\Theta}$, simultaneously,*

$$\text{if } \hat{L}^{\ell_G}(D_0, \boldsymbol{\theta}) \leq \epsilon \quad \text{then} \quad (\mathcal{I}_{\boldsymbol{\theta}}^{\nu_0,\ell_G})^{-1}(\tfrac{1}{n}\log\tfrac{kp}{\delta}) \leq L^{\nu_T,\ell}(\boldsymbol{\theta}) \leq (\mathcal{I}_{\boldsymbol{\theta}}^{\nu_0,\ell_G})^{-1}(\tfrac{1}{n}\log\tfrac{kp}{\delta}) + \epsilon\,. \tag{5.96}$$

[Proof in Appendix C.5]

Reformulating the bound for interpolators under the standard loss yields

$$\text{if } \hat{L}^{\ell}(D_T, \boldsymbol{\theta}) \leq \epsilon, \text{ then } (\mathcal{I}_{\boldsymbol{\theta}}^{\nu_T,\ell})^{-1}\left(\tfrac{1}{n}\log\tfrac{kp}{\delta}\right) \leq L^{\nu_T,\ell}(\boldsymbol{\theta}) \leq (\mathcal{I}_{\boldsymbol{\theta}}^{\nu_T,\ell})^{-1}\left(\tfrac{1}{n}\log\tfrac{kp}{\delta}\right) + \epsilon. \tag{5.97}$$

Together with Theorem 5.37, which asserts that

$$(\mathcal{I}_{\boldsymbol{\theta}}^{\nu_T,\ell})^{-1}(s) \;\geq\; (\mathcal{I}_{\boldsymbol{\theta}}^{\nu_0,\ell_G})^{-1}(s) \quad \forall\, s > 0, \tag{5.98}$$

it follows that an interpolator trained with the augmented loss attains a smaller generalization error, and hence better generalization. In particular, minimizing $\hat{L}^{\ell_G}(D_0, \boldsymbol{\theta})$ is preferable to minimizing $\hat{L}^{\ell}(D_T, \boldsymbol{\theta})$, since (by Theorem 5.37) the augmented loss induces a more concentrated empirical loss and a smaller inverse rate function; consequently, by Corollary 5.38, the interpolator's generalization error decreases.

In practice, data augmentation does not minimize $\hat{L}^{\ell_G}(D_0, \boldsymbol{\theta})$ because the unaltered dataset $D_0$ is unavailable; instead, one minimizes $\hat{L}^{\ell_G}(D_T, \boldsymbol{\theta})$, the augmented loss evaluated on the observed, transformed dataset. If the transformation set $G$ forms a group and its sampling distribution $h$ is uniform—assumptions common in data augmentation—then the next result shows that minimizing $\hat{L}^{\ell_G}(D_T, \boldsymbol{\theta})$ is effectively equivalent to minimizing $\hat{L}^{\ell_G}(D_0, \boldsymbol{\theta})$.

**Proposition 5.39.** *Under Assumption 5.31, if the transformations $G$ define a group and its probability distribution $h$ is uniform, then*

$$\hat{L}^{\ell_G}(D_T, \boldsymbol{\theta}) = \hat{L}^{\ell_G}(D_0, \boldsymbol{\theta}), \tag{5.99}$$

*where $D_0$ is any un-transformed dataset and $D_T$ is the corresponding transformed dataset obtained by applying a transformation $g \sim h$ to each of the samples in $D_0$.* [Proof in Appendix C.5]

When multiple transformations $g_1, \ldots, g_T$ are applied, the analysis extends directly as a sequence of equalities provided that each transformation set $G_1, \ldots, G_T$ forms a group and is sampled from a uniform distribution $h_1, \ldots, h_T$. From this perspective, the approach can be combined with the insights on invariant architectures from the previous section: the two techniques can address different subsets of a chain of transformations. For example, convolutional neural networks naturally handle translations, while data augmentation can target other types of transformations.

Theorem 5.37 establishes that, for a dataset $D_T \sim \nu_T^n$, the augmented empirical loss $\hat{L}^{\ell_G}(D_T, \boldsymbol{\theta})$ is more concentrated than the standard empirical loss $\hat{L}^{\ell}(D_T, \boldsymbol{\theta})$. Furthermore, by the Chernoff bound in Corollary 5.38, interpolators trained with the augmented loss exhibit reduced generalization error. To the best of current knowledge, this provides the strongest explanation of why data augmentation improves the generalization of interpolators, achieved via distribution-dependent bounds. In brief: *Theorem 5.37 shows that the augmented loss increases concentration of the empirical loss, and Corollary 5.38 then implies a smaller generalization error for interpolators under the augmented loss.*

Not all data-augmentation transformations form a group. For instance, rotations restricted to $[-20^\circ, 20^\circ]$ do not constitute a group, since composing $20^\circ$ with $10^\circ$ yields $30^\circ$, which lies outside the range. A rotation group would require the full range (e.g., $[-360^\circ, 360^\circ]$), which is rarely used in practice. Similarly, random cropping does not yield a group. In such cases, the analysis can proceed under the approximation

$$\hat{L}^{\ell_G}(D_T, \boldsymbol{\theta}) \approx \hat{L}^{\ell_G}(D_0, \boldsymbol{\theta}). \tag{5.100}$$

Consequently, minimizing $\hat{L}^{\ell_G}(D_T, \boldsymbol{\theta})$ effectively also minimizes $\hat{L}^{\ell_G}(D_0, \boldsymbol{\theta})$, which is more concentrated and therefore admits interpolators with smaller generalization error.

The next result provides an alternative treatment for transformations that do not form a group. It requires the existence of inverses within $G$: for every $g \in G$ there is $g^{-1} \in G$. Rotations in $[-20°, 20°]$ satisfy this condition because the inverse of any angle $a \in [-20, 20]$ is $-a \in [-20, 20]$. Under this assumption, if $\boldsymbol{\theta}$ is a *perfect interpolator* for $\hat{L}^{\ell_G}(D_T, \boldsymbol{\theta})$—the empirical objective optimized in practice—then $\boldsymbol{\theta}$ is also a *perfect interpolator* for $\hat{L}^{\ell}(D_0, \boldsymbol{\theta})$. Here, "perfect interpolator" denotes a model that attains the minimal achievable empirical loss.

**Proposition 5.40.** *Given an untransformed dataset $D_0$, for any transformed dataset $D_T$ derived from $D_0$ under Assumption 5.31, if*

$$m_{\boldsymbol{\theta}} := \operatorname*{ess\,inf}_{(\mathbf{x},\mathbf{y})\sim\nu_T} \ell(\mathbf{y}, g(\mathbf{x}), \boldsymbol{\theta}) \quad \forall g \in G\,, \tag{5.101}$$

*and $\forall g \in G$, $\exists g^{-1} \in G$. Then, it verifies that*

$$\forall \boldsymbol{\theta} \in \boldsymbol{\Theta} \quad \hat{L}^{\ell_G}(D_T, \boldsymbol{\theta}) = m_{\boldsymbol{\theta}} \iff \hat{L}^{\ell}(D_0, \boldsymbol{\theta}) = m_{\boldsymbol{\theta}}\,. \tag{5.102}$$

[Proof in Appendix C.5]

This result implies that if a model achieves perfect interpolation on the transformed dataset $D_T$ using the augmented loss, it concurrently achieves perfect interpolation on the original, untransformed dataset $D_0$ under the standard loss. This means that when the learning algorithm identifies an interpolator utilizing the augmented loss (as typically occurs in practice), it is essentially also identifying an interpolator for the untransformed dataset $D_0$. This scenario parallels the one encountered with invariant architectures, where the algorithm retrieves interpolators for the untransformed dataset. As highlighted at the outset of Section 5.3.7, interpolators for the untransformed dataset $D_0$ are associated with a lower generalization error compared to those for the transformed dataset $D_T$.

Figure 5.18 (right) illustrates the rate function of an inception model trained with data augmentation achieving almost perfect interpolation, where $\hat{L}^{\ell_G}(D_T, \boldsymbol{\theta}) \approx m_{\boldsymbol{\theta}}$. It's particularly noteworthy to observe how this model's various rate functions are nearly indistinguishable, indicating that the model has nearly become invariant to rotations too. According to Proposition 5.36, if a model is invariant to rotations, then $\mathcal{I}^{\nu_0,\ell}(a) = \mathcal{I}^{\nu_1,\ell}(a)$, which aligns with the outcome of this experiment.

Within this framework—unlike prior methodologies—it is possible to unify invariant architectures, data augmentation, and the explicit regularization strategies of Section 5.3.6. In particular, this perspective explains why explicit regularization, when combined with invariance or augmentation, leads to interpolators with smaller generalization error, a relationship already observed empirically in Figure 5.9. The key step is to extend the analysis of Section 5.3.6 to the inverse rate functions $(\mathcal{I}^{\nu_0}_{\boldsymbol{\theta}})^{-1}(\frac{1}{n}\log\frac{k^p}{\delta})$

and $(\mathcal{I}_{\boldsymbol{\theta}}^{\nu_0,\ell_G})^{-1}(\frac{1}{n}\log\frac{kp}{\delta})$, associated with invariant architectures and data augmentation, respectively. For example, applying Proposition 5.27 to these inverse rate functions clarifies the benefit of $\ell_2$-norm or distance-from-initialization regularization in the presence of invariance and/or augmentation; likewise, Proposition 5.30 elucidates the advantage of input-gradient normalization in these settings.

### 5.3.8 Over-Parameterization and Smooth Interpolation

Many theoretical works have established a direct link between over-parameterization and the strong performance of interpolators (Bubeck et al., 2021; Neyshabur et al., 2019). In particular, the study of Bubeck and Sellke (2023), under an isoperimetry assumption, shows that any model capable of interpolating the training data below the noise threshold must possess a (Euclidean) Lipschitz constant on the order of at least $\sqrt{nd/p}$, where $d$ is the ambient data dimension. Using the present notation, say that a model $\boldsymbol{\theta}$ interpolates below the noise threshold if $\hat{L}(D,\boldsymbol{\theta}) \leq L^\star := \min_{\boldsymbol{\theta}\in\boldsymbol{\Theta}} L(\boldsymbol{\theta})$, and let $Lip(\boldsymbol{\theta})$ denote the Lipschitz constant as in Equation (5.82). Then Theorem 1 of Bubeck and Sellke (2023) can be informally stated as follows: under an isoperimetry assumption, for any $\epsilon \in (0, L^\star)$ and $\delta \in (0,1)$, with probability at least $1-\delta$ over $D \sim \nu^n$, for all $\boldsymbol{\theta} \in \boldsymbol{\Theta}$,

$$\text{if} \quad \hat{L}(D,\boldsymbol{\theta}) \leq \epsilon \quad \text{then} \quad Lip(\boldsymbol{\theta}) \;\geq\; \Omega\Big((L^\star - \epsilon)\sqrt{\tfrac{nd}{p}}\Big). \tag{5.103}$$

This implies that, to maintain $O(1)$ Lipschitz constants for models that continue to interpolate as $n$ grows, the number of parameters $p$ must also increase. In Bubeck and Sellke (2023), the Lipschitz constant serves as a proxy for *smoothness*; their findings therefore identify over-parameterization as a key prerequisite for interpolators to achieve small Lipschitz constants and, consequently, improved generalization.

The PAC–Chernoff perspective affords a more refined understanding of the role of over-parameterization in the generalization of interpolators. In fact, the following result shows that over-parameterization emerges naturally from the PAC–Chernoff bound of Theorem 5.17. Specifically, a lower bound on the number of parameters of a model class is derived in terms of the rate function of a model that interpolates below the noise threshold.

**Theorem 5.41.** *For any $\epsilon \in (0, L^\star)$ and any $\delta \in (0,1)$, with high probability $1-\delta$ over $D \sim \nu^n$, for all $\boldsymbol{\theta} \in \boldsymbol{\Theta}$, simultaneously,*

$$\textit{if} \quad \hat{L}(D,\boldsymbol{\theta}) \leq \epsilon \quad \textit{then} \quad p \geq \frac{n\mathcal{I}_{\boldsymbol{\theta}}(L^\star - \epsilon) + \log\delta}{\log k}. \tag{5.104}$$

[Proof in Appendix C.5]

The above bound connects the smoothness of an interpolator, measured by the rate function as discussed in Section 5.3.4, and the minimum number of parameters in the model class. As the size of the training dataset increases, the number of parameters must also increase linearly to maintain the same degree of *smoothness* in the model. This result generalizes the results of Bubeck and Sellke, 2023 and *it does not require a isoperimetry assumption*.

The principal advantage of these findings is that they forge a connection between over-parameterization, interpolation, and the preceding analyses of model smoothness and rate functions. This linkage shows that the *smooth interpolation* perspective of Bubeck and Sellke (2023)—framed via isoperimetry and Lipschitz continuity—constitutes one among several approximations to the characterization in Theorem 5.41. As discussed in Section 5.3.6, parameter norms, distance from initialization, input-gradient norms, and the Lipschitz constant serve as proxies for the (inverse) rate function and, by extension, for smoothness. In particular, Bubeck and Sellke (2023)'s characterization of smooth interpolation can be recovered directly from Theorem 5.41 under the same isoperimetry assumption; analogous derivations follow under log-concavity by the techniques used in Proposition 5.30 together with Equation (5.82).

**Corollary 5.42.** *If $\ell(\mathbf{y},\mathbf{x},\boldsymbol{\theta})$ is Lipschitz w.r.t. $\mathbf{x}$ with constant $Lip(\boldsymbol{\theta})$ and satisfies a $c$-isoperimetry assumption, then for any $\epsilon \in (0, L^\star)$ and any $\delta \in (0,1)$, with high probability $1-\delta$ over $D \sim \nu^n$, for all $\boldsymbol{\theta} \in \boldsymbol{\Theta}$, simultaneously,*

$$\textit{if} \quad \hat{L}(D,\boldsymbol{\theta}) \leq \epsilon \quad \textit{then} \quad Lip(\boldsymbol{\theta}) \geq \sqrt{\tfrac{nd}{2c(p\log k - \log\delta)}}(L^\star - \epsilon)\,. \tag{5.105}$$

[Proof in Appendix C.5]

Similarly, a bound with a giving norm or distance-from-initialization can de derived.

**Corollary 5.43.** *If the loss function $\ell(\mathbf{y},\mathbf{x},\boldsymbol{\theta})$ is Lipschitz w.r.t. $\boldsymbol{\theta}$ with constant $M > 0$, then, for any $\boldsymbol{\theta}_0 \in \boldsymbol{\Theta}_0 = \{\boldsymbol{\theta} \in \boldsymbol{\Theta} \mid \mathbb{V}_\nu(\ell(\mathbf{y},\mathbf{x},\boldsymbol{\theta})) = 0\} \subset \boldsymbol{\Theta}$, any $\epsilon \in (0, L^\star)$ and any $\delta \in (0,1)$, with high probability $1-\delta$ over $D \sim \nu^n$, for all $\boldsymbol{\theta} \in \boldsymbol{\Theta}$, simultaneously,*

$$\textit{if} \quad \hat{L}(D,\boldsymbol{\theta}) \leq \epsilon \quad \textit{then} \quad \|\boldsymbol{\theta} - \boldsymbol{\theta}_0\|_2 \geq \sqrt{\tfrac{n}{8M(p\log k - \log\delta)}}(L^\star - \epsilon)\,. \tag{5.106}$$

[Proof in Appendix C.5]

Corollary 5.42 and 5.43 imply that, as the number of samples $n$ increases, to have *smooth* interpolators with a small Lipschitz constant, or a small parameter norm, or a small distance from initialization, the number of parameters of the model class must be increased too. In any case, all these results are approximations of the general result given in Theorem 5.41. Note that a similar result could be derived in the context of input-gradient norms using Proposition 5.30. *Interpolating with a small parameter norm, or a small distance from initialization, or a small input-gradient norm requires over-parameterization.*

Remarkably, Theorem 5.41 can also be connected to invariant architectures and data augmentation (Section 5.3.7). In particular, combining Theorem 5.37—which asserts that the data-augmented loss induces a higher rate function—with Theorem 5.41 implies that interpolation under data augmentation will, beyond a certain point, neces-sitate a larger number of parameters. An analogous conclusion holds for more invariant architectures, which likewise yield models with higher rate functions. *Consequently, interpolating with data augmentation and/or an invariant architecture requires over-parameterization.*

To the best of current knowledge, no prior results establish this type of relationship between over-parameterization and smooth interpolators.

### 5.3.9 Experimental Settings

This appendix elaborates on the experimental setup, detailing the model architectures, hyper-parameters, and training procedures used for each figure.

#### 5.3.9.1 Learning Settings for Figures

The code accompanying these experiments is available at https://github.com/Ludvins/2024_PAC-Chernoff-Bound. Unless otherwise noted, experiments employ a compact InceptionV3 architecture (Szegedy et al., 2016) following C. Zhang et al. (2017), trained on the CIFAR-10 dataset (Krizhevsky, Hinton, et al., 2009). Before presenting the exact specification of this "small Inception" model (as adapted from C. Zhang et al., 2017), the constituent modules are described in detail below.

1. Convolutional module: Convolutional layer, batch-normalization and ReLU activation.
2. Inception module with output channels $o_{1\times1}$ and $o_{3\times3}$: Consists on 2 different convolutional layers, one with kernel $1 \times 1$ and $o_{1\times1}$ output channels and another with kernel $3 \times 3$ and $o_{3\times3}$ output channels. The output of this layers is then concatenated, so the total number of output channels is $o_{1\times1} + o_{3\times3}$.
3. Downsample module: Convolutional module with kernel size 3, stride 2 and padding 0 and MaxPooling with kernel size of 3 and stride 2. The outputs of these two layers is concatenated.

With these elements, the architecture of small InceptionV3 network is

1. Convolutional module with 96 output channels, kernel size 3, stride 1 and padding 0.
2. Inception Module with $o_{1\times1} = 32$ and $o_{3\times3} = 32$.
3. Inception Module with $o_{1\times1} = 32$ and $o_{3\times3} = 48$.
4. DownSample Module with $o_{3\times3} = 80$.
5. Inception Module with $o_{1\times1} = 112$ and $o_{3\times3} = 48$.
6. Inception Module with $o_{1\times1} = 96$ and $o_{3\times3} = 64$.
7. Inception Module with $o_{1\times1} = 80$ and $o_{3\times3} = 80$.
8. Inception Module with $o_{1\times1} = 48$ and $o_{3\times3} = 96$.
9. DownSample Module with $o_{3\times3} = 96$.
10. Inception Module with $o_{1\times1} = 176$ and $o_{3\times3} = 160$.

11. Inception Module with $o_{1\times 1} = 176$ and $o_{3\times 3} = 160$.
12. Adaptative Average Pooling layer with kernel $7 \times 7$.
13. Fully connected layer from $16464$ to the number of classes (i.e, $10$).

Where the total number of parameters of this model is $1.814.106$.

**Figure 5.9.** For this experiment, all Inception models were trained using SGD with momentum $0.9$ and learning rate $0.01$ with exponential decay of $0.95$. All models are trained for $30.000$ iterations of batches of size $200$ or until the train loss is under $0.005$. These settings are selected to ensure that the random label model converges to an interpolator. Random cropping is employed using `RandomResizeCrop` function of `torchvision` with scale $(0.8, 1.0)$ and ratio $(0.9, 1.1)$. For $\ell_2$ regularization, the multiplicative factor is $0.01$.

**Figure 5.10.** For this figure, Standard, L2-Crop, and Initial model from Figure 5.9 are used. Subsets of size $n = 50$ of CIFAR10's test split are used to approximate samples of the data generating distribution and build the histograms.

**Figure 5.13.** For this figure, a generalization of LeNet5 is used (three convolutional layers and two fully connected with ReLu activation and average pooling), where the number of channels of the convolutional layers was parameterized by $k$. Precisely, the first layer had 3 input and $\lfloor 6k \rceil$ output channels; the second layer $\lfloor 6k \rceil$ input and $\lfloor 16k \rceil$ output channels; and the last layer $\lfloor 16k \rceil$ input and $\lfloor 120k \rceil$ output channels. The set of models is created ranging $k$ from $0.2$ to $4.9$ every $0.1$; raising models from $7k$ parameters to models with $1.2M$ parameters. Each of these models is then trained until the train loss is lower than $0.01$ or until the train loss has not lowered in two epochs (this only happens in the smallest models).

**Figure 5.14.** The rate function of a subset of all the models in Figure 5.13 is computed here.

**Figure 5.16.** The batch size is fixed to $250$ and images are standardized (this was necessary to improve learning in the MLP model). The precise MLP has $3$ hidden layers with $512$ units, with a total of $1.735.178$ parameters. All models are trained until the interpolation regime, that is, until the train loss is under $0.015$, which, in the worst case where $20.000$ iterations for the MLP. Inception models are trained using a learning rate of $0.001$ whereas MLP models use $0.1$, both with $0.9$ momentum and $0.95$ exponential decay. Regarding the data, $D_0$ is CIFAR10's test set, $D_1$ is the result of performing random translations of $5\%$ and $D_2$ considers random translations of $5\%$ and rotations of up to $20\%$. Both transformations are computed using `RandomAffine` function of `torchvision`.

**Figure 5.17.** The model's specification and training setup is the same as in Figure 5.16. Regarding the data, the random shuffling of the pixels was performed using a random permutation using `Numpy`; the dataset was fully permuted and stored as a new dataset.

**Figure 5.18.** The model's specification and training setup is the same as in Figure 5.16. $D_1$ considers random translations of $5\%$ and rotations of up to $20\%$ (the same as $D_2$ in Figure 5.16). Data augmentation was produced using the same transformations as those that define $D_1$.

#### 5.3.9.2 Estimating the Cumulant and Rate Function

From the definition of the cumulant function $J_{\boldsymbol{\theta}}(\lambda)$,

$$J_{\boldsymbol{\theta}}(\lambda) = \log \mathbb{E}_\nu \left[ e^{\lambda(L(\boldsymbol{\theta}) - \ell(\mathbf{y},\mathbf{x},\boldsymbol{\theta}))} \right] = \log \mathbb{E}_\nu \left[ p(\mathbf{y}|\mathbf{x},\boldsymbol{\theta})^\lambda \right] - \mathbb{E}_\nu [\log p(\mathbf{y}|\mathbf{x},\boldsymbol{\theta})^\lambda] , \quad (5.107)$$

it is clear that computing its true value requires access to the true data generation distribution $\nu$. However, in real-world problems, this distribution is unknown and inaccessible.

The Machine Learning community is used to approximate this kind of quantities (such as the expected loss $L(\boldsymbol{\theta})$) using separate validation datasets $D^{val}$. In fact, due to the large amount of data available in nowadays's problems, using this approach is perfectly doable, leading to

$$J_{\boldsymbol{\theta}}(\lambda) \approx \log \left( \frac{1}{M} \sum_{(\mathbf{x},\mathbf{y}) \in D^{val}} p(\mathbf{y}|\mathbf{x},\boldsymbol{\theta})^\lambda \right) - \frac{1}{M} \sum_{(\mathbf{x},\mathbf{y}) \in D^{val}} \log p(\mathbf{y}|\mathbf{x},\boldsymbol{\theta})^\lambda . \quad (5.108)$$

It is important to notice that the above estimator is biased due to the first term and Jensen's Inequality. In fact,

$$\begin{aligned} \mathbb{E}_{D^{val}} \left[ \log \left( \frac{1}{M} \sum_{(\mathbf{x},\mathbf{y}) \in D^{val}} p(\mathbf{y}|\mathbf{x},\boldsymbol{\theta})^\lambda \right) \right] &\leq \log \left( \mathbb{E}_{D^{val}} \left[ \frac{1}{M} \sum_{(\mathbf{x},\mathbf{y}) \in D^{val}} p(\mathbf{y}|\mathbf{x},\boldsymbol{\theta})^\lambda \right] \right) \\ &= \log \mathbb{E}_\nu \left[ p(\mathbf{y}|\mathbf{x},\boldsymbol{\theta})^\lambda \right] . \end{aligned} \quad (5.109)$$

As a result, if the size of $D^{val}$ is not large enough, the cumulant function might be underestimated.

In regard to computational stability, computing the estimation in Equation (5.108) can be computationally unstable due to the use of probabilities. The use of log-probabilities and log-sum-exp operations is encouraged as,

$$J_{\boldsymbol{\theta}}(\lambda) \approx \log \left( \sum_{(\mathbf{x},\mathbf{y}) \in D^{val}} \exp(\lambda \log p(\mathbf{y}|\mathbf{x},\boldsymbol{\theta})) \right) - \log M - \frac{1}{M} \sum_{(\mathbf{x},\mathbf{y}) \in D^{val}} \lambda \log p(\mathbf{y}|\mathbf{x},\boldsymbol{\theta}) . \quad (5.110)$$

From this, it is straightforward to compute the log-probabilities of the model (for ex-ample, skipping the softmax layer of a NN), multiply them by $\lambda$ and compute the mean and log-sum-exp of these quantities.

Once the cumulant function has been approximated, computing the rate function relies on computing the optimal value of $\lambda$,

$$\mathcal{I}_{\boldsymbol{\theta}}(a) = \sup_{\lambda>0} \lambda a - J_{\boldsymbol{\theta}}(\lambda)\,. \tag{5.111}$$

In this regard, trying to optimize the value of $\lambda$ doing automatic optimization resulted in a very unstable method in the conducted experiments. Thus, the recommended and used method is using a *binary search* algorithm. Fixed a range in which to optimize lambda $[\lambda_{min}, \lambda_{max}]$, a binary search algorithm has complexity $\mathcal{O}(\log_2(\lambda_{max} - \lambda_{min}))$. In fact, if (due to the nature of the problem) the needed value of $\lambda_{max}$ is too large, one might perform the binary search in $[\log(\lambda_{min}), \log(\lambda_{max})]$, which has the same complexity but makes it easier to consider larger values of $\lambda$.

It is clear that computing the rate function is more complex than computing only the cumulant function (as the former requires the latter). In fact, the next result shows that it might not be necessary to compute the rate function, as the cumulant might be enough to characterize smoothness.

**Proposition 5.44.** *If* $\forall\lambda \geq 0$, $J_{\boldsymbol{\theta}}(\lambda) \leq J_{\boldsymbol{\theta}'}(\lambda)$, *then* $\forall a \geq 0$, *it verifies* $\mathcal{I}_{\boldsymbol{\theta}}(a) \geq \mathcal{I}_{\boldsymbol{\theta}'}(a)$.

*Proof.* Direct consequence of the $\mathcal{I}_{\boldsymbol{\theta}}(a)$ being the Legendre transform of $J_{\boldsymbol{\theta}}(\lambda)$. □

From this result, if a model $\boldsymbol{\theta}$ has a higher cumulant function than another model $\boldsymbol{\theta}'$, then $\boldsymbol{\theta}$ is smoother than $\boldsymbol{\theta}'$ and many results apply. Figure 5.19 clearly illustrates this case. *Just plotting the cumulants is enough to understand which models are smoother*.

### 5.3.10 Discussion and Limitations

This thesis has examined a growing body of evidence (Gastpar et al., 2024; Nagarajan and Kolter, 2019b; Y. Wang et al., 2024; C. Zhang et al., 2017) indicating that bounds depending solely on the training sample are provably vacuous for over-parameterized model classes, motivating alternative approaches to understanding generalization in deep learning. In response, distribution-dependent bounds—those that depend explicitly on the data-generating distribution—were advocated. A distribution-dependent PAC–Chernoff bound (Theorem 5.17) was introduced and shown to be perfectly tight for any interpolator. Building on its complexity measure, a new notion of *smoothness* was proposed (Definition 5.21), providing a principled answer to the previously unresolved question of which interpolators generalize more effectively.

Section 5.3.5 demonstrated that PAC–Chernoff bounds capture the double-descent phenomenon, explaining how interpolators can achieve smaller generalization error even as the number of parameters increase.

Section 5.3.6 established that the complexity term of the PAC–Chernoff bound induces a regularizer that yields near-optimal performance for over-parameterized models interpolating the training data (Theorem 5.26). The remainder of the section showed that a wide array of existing regularizers act as proxies for this optimal construction; in some instances, refined variants were identified as effective surrogates for an interpolator's generalization error. This analysis clarifies both the scope and the limitations of standard techniques such as parameter norms, distance from initialization, and input-gradient penalties.

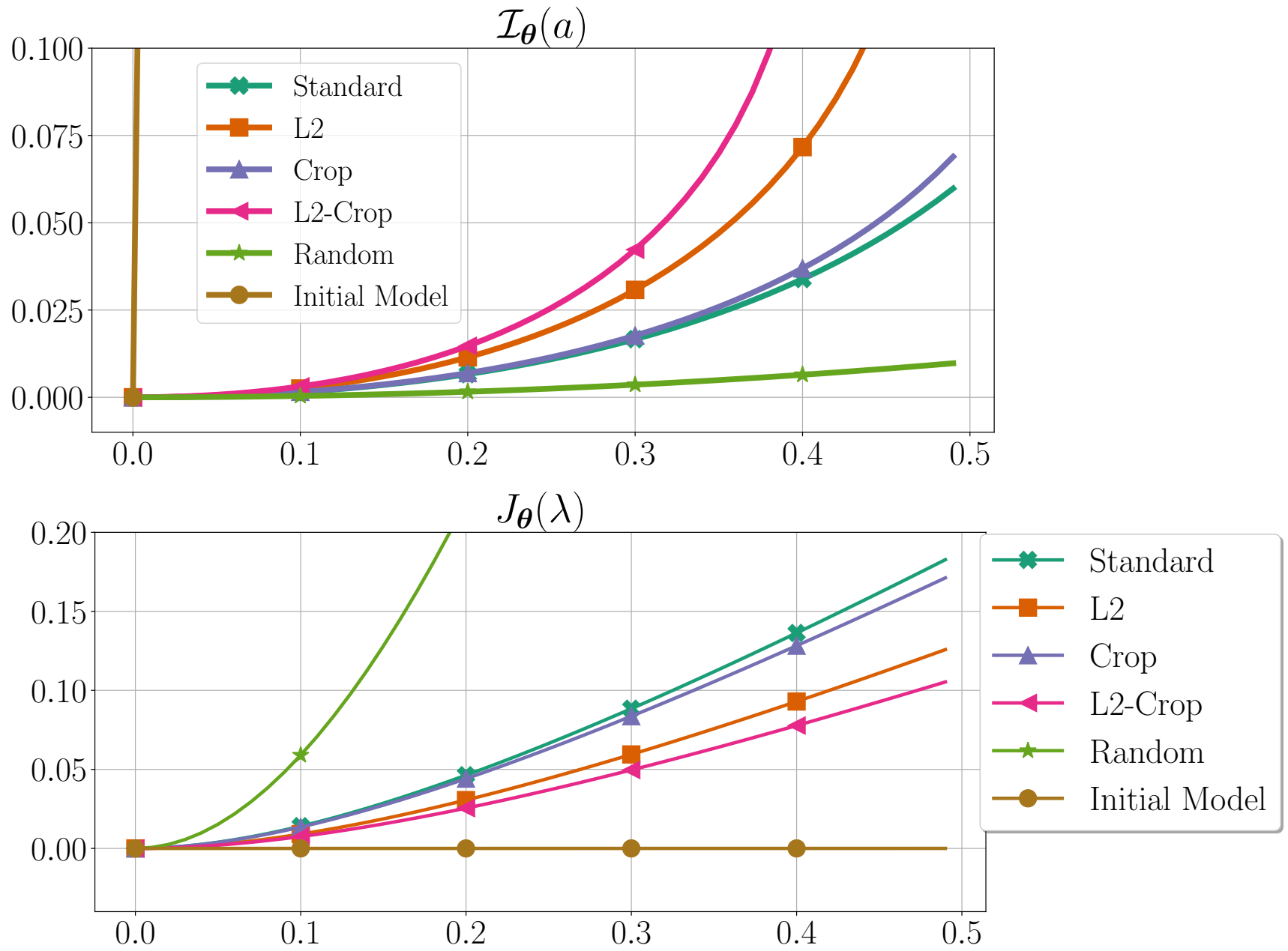

**Figure 5.19:** ***Illustration on the relationship between the rate function $\mathcal{I}_{\theta}(a)$ (top) and the cumulant function $J_{\theta}(\lambda)$ (bottom).*** *They display the rate and the cumulant functions for the same set of models (from Figure 5.9).*

Section 5.3.7 analyzed the impact of transformed inputs (Assumption 5.31), showing via the PAC–Chernoff bound in Equation 5.88 that interpolation becomes harder due to increased generalization error. The same framework, combined with additional results, explains why invariant architectures and data augmentation yield interpolators with smaller generalization error. The section concluded by unifying invariant architectures, data augmentation, and the explicit regularizers from Section 5.3.6.

Finally, Section 5.3.8 established over-parameterization as a necessary condition for achieving smooth interpolators, where smoothness can be characterized through parameter norms, Lipschitz constants, or the use of invariant architectures and data augmentation. This perspective reveals connections among approaches previously regarded as unrelated, and it clarifies how larger model classes can accommodate smooth interpolators with superior generalization.

In summary, distribution-dependent PAC–Chernoff bounds constitute a powerful framework for understanding—and improving—the generalization behavior of (over-parameterized) interpolators.

**Discussion of Limitations**

A principal limitation of this study is the assumption of a finite model class. It is anticipated that this restriction can be relaxed by leveraging recently proposed PAC–Bayes–Chernoff bounds (Casado et al., 2024), which are likewise distribution-dependent and rely on an analogous rate function. This suggests that the results may extend to infinite model classes, although a full development lies beyond the present scope. Additionally, the analysis does not address the role of stochastic gradient descent (SGD) in identifying interpolators with minimal generalization error.

A second limitation concerns the absence of explicit links between the smoothness notion introduced here and commonly used notions in the literature. Examples include characterizations via the largest Hessian eigenvalue (Nesterov, 2003), loss variation within a perturbation radius (Keskar et al., 2017), and Lipschitz continuity of the loss (Shalev-Shwartz and Ben-David, 2014). While the concept of smoothness adopted in this work is grounded in the PAC–Chernoff framework through its distribution-dependent complexity term, future research could investigate alignments and divergences with these established definitions. Such connections could clarify how the present framework complements or generalizes existing theory. The focus here has been on the implications of the proposed definition for generalization, leaving broader comparisons to subsequent work; this underscores an opportunity for follow-up studies to bridge the present notion of smoothness with the wider landscape of smoothness concepts in machine learning theory.

## 5.4 The Implicit Bias of Stochastic Gradient Descent

Stochastic Gradient Descent (SGD) has become the standard method for training modern neural networks, enabling large-scale models that underpin many of today's applications (Bottou, 2010). Beyond its efficiency as an optimization algorithm, SGD plays a central role in determining how well models generalize, particularly in highly overparameterized regimes where numerous parameter settings can achieve zero training error (C. Zhang et al., 2017). A long-standing observation is that SGD is not neutral among these solutions: it often converges to models with stronger generalization performance, an implicit bias that remains only partly understood.

One explanation attributes this bias to the randomness introduced by mini-batch sampling. The stochasticity of SGD has been linked to a tendency toward simpler or lower-complexity models, effectively acting as a form of implicit regularization during learning (Neyshabur et al., 2015b; Zou et al., 2021). Empirical studies even suggest that this implicit effect can outperform explicit regularizers in certain cases, such as linear regression (Zou et al., 2021). While such results highlight the importance of algorithmic regularization, existing theoretical approaches—often based on concentration inequalities or uniform convergence—tend to produce loose bounds or fail to capture the distributional nuances of deep models (Nagarajan and Kolter, 2019b). This moti-vates the exploration of alternative frameworks capable of describing the mechanisms behind SGD's bias in greater detail.

In this thesis, the use of Large Deviation Theory (LDT) (Ellis, 2012; Touchette, 2009) is investigated as such a framework. A decomposition of the generalization error is derived, which consists of three parts: the expected loss, a concentration term that quantifies how tightly the empirical loss is distributed around its mean, and an abnormality term that reflects the rarity of observed deviations. This perspective reveals a sharp contrast between optimization methods: full-batch Gradient Descent (GD) is prone to models with poor concentration and abnormal deviations—explaining its overfitting behavior—while mini-batch SGD reduces both effects, favoring models that achieve lower generalization error. The main contributions of this thesis in this field of study can be summarized as follows:

1. The introduction of an LDT-based decomposition of the generalization error that distinguishes the roles of concentration and abnormal deviations.
2. The fact that GD and SGD exhibit fundamentally different biases is explored: GD tends to exploit poorly concentrated and abnormal empirical losses, while SGD avoids them.
3. The theory is validated using deep convolutional networks on CIFAR-10, showing that batch size and $\ell_2$ regularization systematically influence concentration and abnormality in line with our predictions.

This approach represents an initial, but first, step in applying LDT to the study of SGD's implicit bias, a central open problem in machine learning. LDT has long been used in physics, finance, and telecommunications to characterize fluctuations and rare events by quantifying deviations from typical behavior (Ellis, 2012; Touchette, 2009). Rather than offering a complete explanation of SGD's implicit bias, this thesis shows that LDT provides a complementary perspective that clarifies why SGD tends to favor models with smaller generalization error.

Understanding the implicit regularization of SGD is of clear practical significance, as modern deep learning relies almost exclusively on SGD and its variants to train highly overparameterized models. Despite their ability to interpolate training data, these models often generalize well in practice, largely due to the biases induced by SGD. The proposed LDT-based perspective provides a principled explanation of this phenomenon that has the potential to guide the design of novel optimization strategies that could enhance reliability and robustness in real-world applications.

### 5.4.1 Preliminaries

As in the previous sections of this chapter, let $D$ be an i.i.d. sample of size $n$ from an unknown distribution $\nu(\mathbf{y}, \mathbf{x})$. For each $\boldsymbol{\theta} \in \boldsymbol{\Theta}$, the loss function $\ell(\mathbf{y}, \mathbf{x}, \boldsymbol{\theta})$ is positive, yielding an expected loss $L(\boldsymbol{\theta}) = \mathbb{E}_\nu[\ell(\mathbf{y}, \mathbf{x}, \boldsymbol{\theta})]$ and an empirical loss $\hat{L}(D, \boldsymbol{\theta}) = \frac{1}{n}\sum_{i=1}^n \ell(\mathbf{y}_i, \mathbf{x}_i, \boldsymbol{\theta})$. Since the dataset $D$ is random, $\hat{L}(D, \boldsymbol{\theta})$ fluctuates around $L(\boldsymbol{\theta})$, and the tightness of this concentration depends on the model. Figure 5.20 (left) illustrates this behavior with histograms of empirical losses for three InceptionV3 models (Szegedy et al., 2016). The *Initial* model (with Kaiming initialization) shows empirical losses tightly concentrated around its mean $\ln 10$, the $\ell_2$-regularized model exhibits similarly strong concentration, while the *Standard* model displays a wider spread.

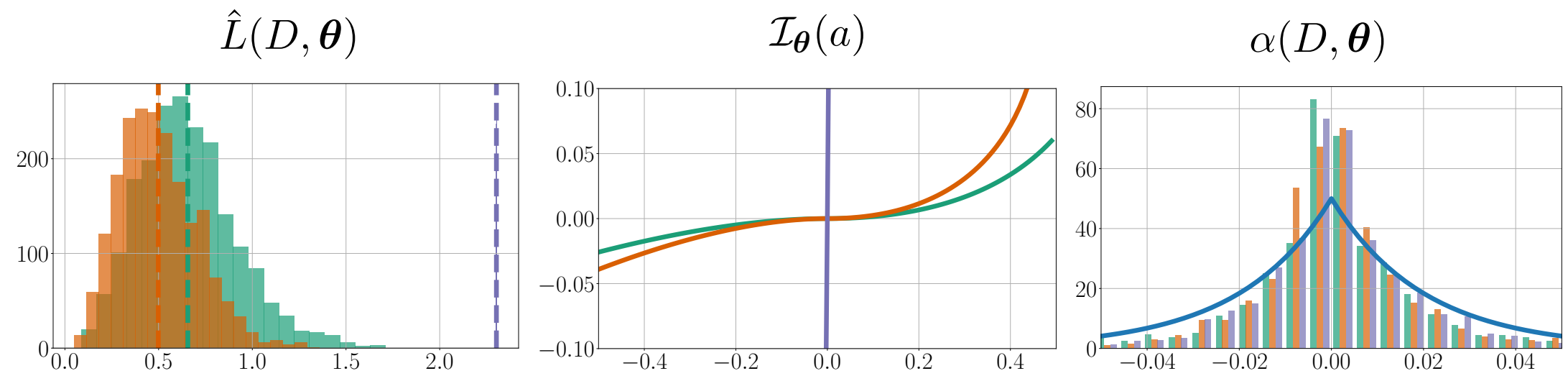

**Figure 5.20:** ***Empirical loss, rate function, and abnormality for InceptionV3 on CIFAR-10*** ($n = 50$) ***from Figure 5.9*** (***Section 5.3***). *Left: distributions of the empirical loss $\hat{L}(D, \theta)$ over draws $D \sim \nu^n$ for three models: a standard SGD-trained network (green), an $\ell_2$–regularized network (orange), and the initial (untrained) network (gray). Center: corresponding estimates of the rate function $\mathcal{I}_{\boldsymbol{\theta}}(a)$, showing how training and $\ell_2$ regularization increase concentration. Right: distributions of the abnormality $\alpha(D, \boldsymbol{\theta})$ for the same models; the blue curve overlays the* $\mathrm{Exp}(n)$ *law (plotted on both sides of zero) to illustrate Theorem 5.46.*

Empirical risk minimization (ERM) seeks $\min_{\boldsymbol{\theta}} \hat{L}(D, \boldsymbol{\theta})$. The main challenge is ensuring that $\hat{L}(D, \boldsymbol{\theta})$ remains close to $L(\boldsymbol{\theta})$ (i.e., small generalization error). Two factors shape the difference between $L(\boldsymbol{\theta})$ and a realization of $\hat{L}(D, \boldsymbol{\theta})$: (i) the *level of concentration* of the distribution of $\hat{L}(D, \boldsymbol{\theta})$ around $L(\boldsymbol{\theta})$ (a low empirical loss can arise from a high-mean model if the distribution of the empirical loss $\hat{L}(D, \boldsymbol{\theta})$ is wide), and (ii) the *level of abnormality*, i.e., whether a small observed empirical loss stems from a rare left-tail outcome from the distribution of $\hat{L}(D, \boldsymbol{\theta})$. Models whose empirical loss is both well-concentrated around its mean $L(\boldsymbol{\theta})$ and is not the result of a rare left-tail realization will have smaller generalization error.

In order to mathematically formalize these two factors, we use the so-called rate function, the central function in LDT, which is denoted by $\mathcal{I}_{\boldsymbol{\theta}}(a) : \mathbb{R} \to \mathbb{R}$, and it is defined as the *Legendre transform* of the *cumulant generating function*, denoted by $J_{\boldsymbol{\theta}}(\lambda) : \mathbb{R} \to \mathbb{R}^+$. In this section, we introduced a signed version of the rate function and consider the *cumulant generating function* of the model's centered loss. These two functions are defined as

$$J_{\boldsymbol{\theta}}(\lambda) := \ln \mathbb{E}_\nu \left[ e^{\lambda(L(\boldsymbol{\theta}) - \ell(\mathbf{y}, \mathbf{x}, \boldsymbol{\theta}))} \right], \tag{5.112}$$

and

$$\mathcal{I}_{\boldsymbol{\theta}}(a) := sign(a) \cdot \sup_{\lambda \in \mathbb{R}} \lambda a - J_{\boldsymbol{\theta}}(\lambda), \tag{5.113}$$

where $\mathcal{I}_{\boldsymbol{\theta}}(a)$ is a *signed* rate function to make it invertible in $\mathbb{R}$. The rate $\mathcal{I}_{\boldsymbol{\theta}}(a)$ and the cumulant $J_{\boldsymbol{\theta}}(\lambda)$ are well-defined, positive, and strictly monotonic real-valued functions, satisfying $\mathcal{I}_{\boldsymbol{\theta}}(0) = 0$ and $J_{\boldsymbol{\theta}}(0) = 0$ (Rockafellar, 1970).

Figure 5.20 (center) presents the rate functions for the three previously discussed InceptionV3 (Szegedy et al., 2016) neural networks. The rate functions clearly reflect the varying levels of concentration in the empirical losses, as depicted by the histograms in Figure 5.20 (left). The *Initial* model exhibits a prominent rate function, while the *Standard* model has a smaller rate function compared to the $\ell_2$-regularized model.

### 5.4.2 Gradient Descent

In this section, a novel decomposition of a model's generalization error is introduced, formalizing the concept of *abnormality* in the generalization error, and demonstrating how full-batch GD is biased toward finding models with poorly concentrated empirical losses and whose realized empirical loss deviates abnormally from the expected loss.

#### 5.4.2.1 Decomposing the empirical loss

The following result provides a decomposition of the empirical loss in terms of the expected loss $L(\boldsymbol{\theta})$, the inverse of the signed rate function $\mathcal{I}_{\boldsymbol{\theta}}^{-1}(s)$, and a novel function $\alpha : \mathcal{D} \times \boldsymbol{\theta} \to \mathbb{R}$. As discussed in the next section, $\alpha(D, \boldsymbol{\theta})$ quantifies the *degree of abnormality* of the observed generalization error $L(\boldsymbol{\theta}) - \hat{L}(D, \boldsymbol{\theta})$ for the model $\boldsymbol{\theta}$.

**Proposition 5.45.** *For any $D \sim \nu^n$ and any $\boldsymbol{\theta} \in \boldsymbol{\Theta}$, it verifies that*

$$\hat{L}(D, \boldsymbol{\theta}) = L(\boldsymbol{\theta}) - \mathcal{I}_{\boldsymbol{\theta}}^{-1}(\alpha(D, \boldsymbol{\theta}))\,, \tag{5.114}$$

*where $\alpha : \mathcal{D} \times \boldsymbol{\Theta} \to \mathbb{R}$ is defined as $\alpha(D, \boldsymbol{\theta}) := \mathcal{I}_{\boldsymbol{\theta}}(L(\boldsymbol{\theta}) - \hat{L}(D, \boldsymbol{\theta}))$.* [Proof in Appendix C.6]

Although the decomposition above is technically straightforward, it separates the empirical loss into two distinct components with clear interpretations. The first component is the expected loss $L(\boldsymbol{\theta})$. The second component, which captures the deviation of the observed empirical loss from its expectation, corresponds to the generalization error. Within this term, the function $\mathcal{I}_{\boldsymbol{\theta}}^{-1}(s)$ characterizes the level of concentration of the distribution of $\hat{L}(D, \boldsymbol{\theta})$ around its mean $L(\boldsymbol{\theta})$. As shown in Section 5.3.2, models with a larger rate function $\mathcal{I}_{\boldsymbol{\theta}}(\cdot)$ exhibit stronger concentration, and thus their inverse rate function $\mathcal{I}_{\boldsymbol{\theta}}^{-1}(s)$ is smaller. In fact, a second-order Taylor expansion of $\mathcal{I}_{\boldsymbol{\theta}}^{-1}(s)$ around $s = 0$ reveals a close connection with the standard deviation of the model loss, denoted by $\sigma(\ell_{\boldsymbol{\theta}})$:

$$\mathcal{I}_{\boldsymbol{\theta}}^{-1}(s) \approx \text{sign}(s)\sqrt{2|s|}\,\sigma(\ell_{\boldsymbol{\theta}}) \quad \text{where} \quad \sigma(\ell_{\boldsymbol{\theta}}) := \sqrt{\mathbb{E}_\nu\left[(\ell(\mathbf{y}, \mathbf{x}, \boldsymbol{\theta}) - L(\boldsymbol{\theta}))^2\right]}. \tag{5.115}$$

Finally, note that both $L(\boldsymbol{\theta})$ and $\mathcal{I}_{\boldsymbol{\theta}}^{-1}(s)$ are deterministic; according to Proposition 5.45, all the randomness in $\hat{L}(D, \boldsymbol{\theta})$ arises from the abnormality value $\alpha(D, \boldsymbol{\theta})$. In the following section, the value of $\alpha(D, \boldsymbol{\theta})$ is shown to represent the degree of abnormality in the magnitude of the generalization error. $\alpha(D, \boldsymbol{\theta})$ will be higher when the observed $\hat{L}(D, \boldsymbol{\theta})$ comes from the tails of the distribution of the empirical loss and small otherwise. Since $\mathcal{I}_{\boldsymbol{\theta}}^{-1}(s)$ *increases monotonically with* $s$ (Rockafellar, 1970), a larger $\alpha(D, \boldsymbol{\theta})$ value leads to a greater difference between $L(\boldsymbol{\theta})$ and $\hat{L}(D, \boldsymbol{\theta})$.

#### 5.4.2.2 The Abnormality of the Generalization Error

A large gap between $\hat{L}(D, \boldsymbol{\theta})$ and $L(\boldsymbol{\theta})$ is considered highly unlikely, or *abnormal*, when the distribution of the empirical loss is tightly concentrated around its mean $L(\boldsymbol{\theta})$, in which case $\hat{L}(D, \boldsymbol{\theta})$ must originate from the tails of this distribution. In contrast, the same gap may *not be abnormal* for a model whose empirical loss is only weakly concentrated.

Using the approximation in Equation (5.115), the following approximation for $\alpha(D, \boldsymbol{\theta})$ can be derived:

$$\alpha(D, \boldsymbol{\theta}) \approx \frac{1}{2} sign\Big(L(\boldsymbol{\theta}) - \hat{L}(D, \boldsymbol{\theta})\Big) \left(\frac{L(\boldsymbol{\theta}) - \hat{L}(D, \boldsymbol{\theta})}{\sigma(\ell_{\boldsymbol{\theta}})}\right)^2 . \tag{5.116}$$

From this approximation, it is clear that large values of $\alpha(D, \boldsymbol{\theta})$ correspond to situations where $\hat{L}(D, \boldsymbol{\theta})$ lie abnormally far from its mean $L(\boldsymbol{\theta})$. In general, Cramér's Theorem (Theorem 5.16) shows that $\alpha(D, \boldsymbol{\theta})$ asymptotically equals the (normalized) log-probability of observing a generalization error at least as large as $L(\boldsymbol{\theta}) - \hat{L}(D, \boldsymbol{\theta})$:

$$\alpha(D, \boldsymbol{\theta}) \asymp -\frac{1}{n} \ln \mathbb{P}_{S \sim \nu^n}\Big(L(\boldsymbol{\theta}) - \hat{L}(S, \boldsymbol{\theta}) \geq L(\boldsymbol{\theta}) - \hat{L}(D, \boldsymbol{\theta})\Big) . \tag{5.117}$$

Given a fixed dataset $D$, the abnormality rate $\alpha(D, \boldsymbol{\theta})$ *quantifies the log-probability that, for another dataset* $S$*, the generalization error exceeds the one observed with* $D$. When comparing two models, $\boldsymbol{\theta}$ and $\boldsymbol{\theta}'$, if $\alpha(D, \boldsymbol{\theta}) \geq \alpha(D, \boldsymbol{\theta}')$, then the likelihood of observing a generalization error at least as large as that of $D$, for another dataset $S \sim \nu^n$, is lower for $\boldsymbol{\theta}$ than for $\boldsymbol{\theta}'$. In this case, *the observed generalization error for* $D$ *was more abnormal under* $\boldsymbol{\theta}$ *than under* $\boldsymbol{\theta}'$.

The following result shows how $\alpha(D, \boldsymbol{\theta})$, as a random variable over $D \sim \nu^n$, is highly related to an exponential distribution of parameter $n$:

**Theorem 5.46.** *For any* $\boldsymbol{\theta} \in \boldsymbol{\Theta}$, $n > 0$, *and* $D \sim \nu^n$, *the cumulative distribution of* $\alpha(D, \boldsymbol{\theta})$ *satisfies*

$$\forall s > 0 \quad \mathbb{P}_{D \sim \nu^n}(\alpha(D, \boldsymbol{\theta}) \geq s) \leq e^{-n|s|} , \tag{5.118}$$

$$\forall s < 0 \quad \mathbb{P}_{D \sim \nu^n}(\alpha(D, \boldsymbol{\theta}) \leq s) \leq e^{-n|s|} , \tag{5.119}$$

*and both inequalities are asymptotically tight,*

$$\forall s > 0 \quad \mathbb{P}_{D \sim \nu^n}(\alpha(D, \boldsymbol{\theta}) \geq s) \asymp e^{-n|s|} , \tag{5.120}$$

$$\forall s < 0 \quad \mathbb{P}_{D \sim \nu^n}(\alpha(D, \boldsymbol{\theta}) \leq s) \asymp e^{-n|s|} . \tag{5.121}$$

[Proof in Appendix C.6]

Theorem 5.46 shows that the tails of the distribution of $\alpha(D,\boldsymbol{\theta})$ are always *thinner* than those of an exponential distribution with rate $n$, denoted by $\mathrm{Exp}(n)$. Importantly, this property holds *regardless of the model or the data-generating distribution*. This insight makes it possible to quantify the degree of abnormality in a model's generalization error by locating the corresponding $\alpha(D,\boldsymbol{\theta})$ value within the tail of an exponential distribution. For example, in a dataset of size $50\,000$, if $\alpha(D,\boldsymbol{\theta}) \geq \frac{1}{50\,000}\ln\frac{1}{0.01} \approx 0.0001$, the probability of observing such an event is less than $1\%$. This provides a *universal* cut-off, valid for any model and any data-generating distribution.

The second result in Theorem 5.46 shows that for *large datasets*, $\alpha(D,\boldsymbol{\theta})$ closely approximates a zero-centered double-exponential distribution, or Laplace distribution, regardless of the model or the data-generating distribution. This indicates that for *large datasets, the stochasticity associated with $\hat{L}(D,\boldsymbol{\theta})$ can be effectively represented by a Laplace distribution, independently of the model family or the underlying data-generating process*. Figure 5.20 (right) illustrates this point with surprising accuracy. The figure shows how the empirical distribution of $\alpha(D,\boldsymbol{\theta})$ for three very different InceptionV3 models trained on CIFAR10, where $D \sim \nu^{50}$, closely resembles a double-exponential or Laplace distribution, even with such a small $n$ value. In conclusion, the distribution of empirical loss for *large* $n$ values can be expressed as:

$$\hat{L}(D,\boldsymbol{\theta}) \approx L(\boldsymbol{\theta}) - \mathcal{I}_{\boldsymbol{\theta}}^{-1}(s), \quad s \sim \text{Laplace}(0, n^{-1})\,, \tag{5.122}$$

where the Laplace distribution is parameterized using its location $0$ and scale $n^{-1}$. The above equation resembles the reparameterization of a Gaussian distribution, particularly when considering the approximation given in Equation (5.115). This perspective highlights a novel asymptotic approximation of the generalization error offered by LDT (Ellis, 2012), that, at first, differs from the one provided by the Central Limit Theorem.

#### 5.4.2.3 Analysis of the Implicit Biases of Gradient Descent

Proposition 5.45 shows that the empirical loss $\hat{L}(D,\boldsymbol{\theta})$ can be expressed as $\hat{L}(D,\boldsymbol{\theta}) = L(\boldsymbol{\theta}) - \mathcal{I}_{\boldsymbol{\theta}}^{-1}(\alpha(D,\boldsymbol{\theta}))$. Minimizing $\hat{L}(D,\boldsymbol{\theta})$ requires balancing two competing objectives. The first is reducing $L(\boldsymbol{\theta})$, since models with lower expected loss naturally tend to achieve lower empirical loss. The second is increasing $\mathcal{I}_{\boldsymbol{\theta}}^{-1}(\alpha(D,\boldsymbol{\theta}))$, which, due to the strict monotonicity of $\mathcal{I}_{\boldsymbol{\theta}}^{-1}(\cdot)$, involves two aspects. One is maximizing $\alpha(D,\boldsymbol{\theta})$, which promotes more *abnormal* left-tail realizations, leading to empirical losses that deviate more significantly from their mean. The other is maximizing $\mathcal{I}_{\boldsymbol{\theta}}^{-1}(\cdot)$ for a given $\alpha(D,\boldsymbol{\theta})$, which implicitly favors models whose empirical loss distributions are less concentrated around $L(\boldsymbol{\theta})$—a larger $\mathcal{I}_{\boldsymbol{\theta}}^{-1}(\cdot)$ implies a lower $\mathcal{I}_{\boldsymbol{\theta}}(\cdot)$ which corresponds to a less concentrated distribution.

Although it could be expected that an empirical-loss minimizer focuses solely on achieving a small $L(\boldsymbol{\theta})$, Proposition 5.45 highlights a competing incentive: a model whose empirical loss distribution is more spread out can "luck into" a dataset $D$ in the left tail, yielding a small $\hat{L}(D,\boldsymbol{\theta})$. In practice, minimizing $\hat{L}(D,\boldsymbol{\theta})$ by using GD will lean toward models that (i) reduce $L(\boldsymbol{\theta})$ and (ii) inflate $L(\boldsymbol{\theta}) - \hat{L}(D,\boldsymbol{\theta})$. Due to the second effect, the models minimizing $\hat{L}(D,\boldsymbol{\theta})$ often exhibit significant generalization error.

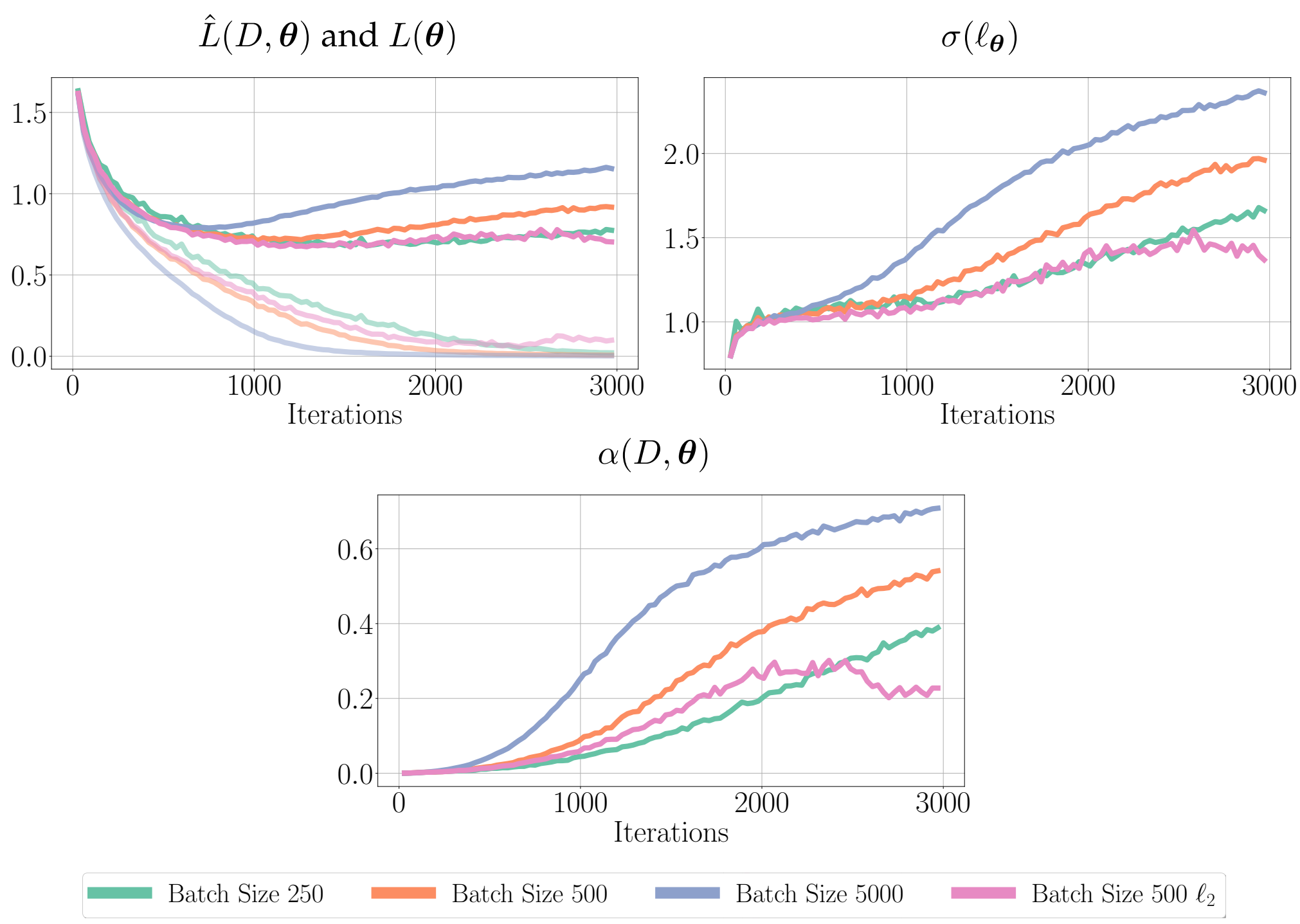

**Figure 5.21:** ***Training/test loss, loss dispersion, and abnormality under varying batch sizes (InceptionV3, CIFAR-10).*** *Top left: evolution of training empirical loss $\hat{L}(D,\theta)$ and test loss $L(\theta)$ across iterations for batch sizes $m \in \{250, 500, 5000\}$ and for an $\ell_2$-regularized run at $m = 500$. Top right: corresponding standard deviation of the per-sample loss, $\sigma(\ell_\theta)$, showing increased dispersion for larger batches and its mitigation by $\ell_2$ regularization. Bottom: abnormality rate $\alpha(D,\theta)$ over training, likewise reduced by regularization at fixed batch size $m = 500$.*

Figure 5.21 illustrates these dynamics, where GD is approximated by using SGD with an exceptionally large batch size of $5\ 000$. In Figure 5.21 (top left), $\hat{L}(D,\boldsymbol{\theta})$ decreases monotonically, while $L(\boldsymbol{\theta})$ initially decreases but later rises slightly. Figure 5.21 (top right) shows the variance of $\hat{L}(D,\boldsymbol{\theta})$, reflecting its increasing concentration over time. Figure 5.21 (bottom) presents the abnormality rate $\alpha(D,\boldsymbol{\theta})$, which steadily grows. Notably, GD leads to models where $\hat{L}(D,\boldsymbol{\theta})$ deviates significantly from $L(\boldsymbol{\theta})$. Actually, this difference is a highly unlikely event. Using Theorem 5.46, the probability of observing $\alpha(D,\boldsymbol{\theta}) = 0.7$ with $n = 50\,000$ is smaller than $e^{-50\,000 \cdot 0.7} \approx 10^{-8\,000}$, an astronomically small value. This underscores how GD explores a vast space of model realizations, some exhibiting extreme deviations from $L(\boldsymbol{\theta})$.

Figure 5.21 also depicts the dynamics of SGD for small mini-batches, revealing a distinct optimization behavior. In contrast to full-batch GD, SGD converges to models that exhibit a different trade-off: the concentration of the distribution of $\hat{L}(D,\boldsymbol{\theta})$ is higher, while the abnormality rate $\alpha(D,\boldsymbol{\theta})$ remains lower. This results in models with better generalization error, as the empirical loss aligns more closely with the expected loss.

### 5.4.3 Stochastic Gradient Descent

SGD operates by selecting mini-batches $B \subseteq D$ of size $m$ from the training dataset without replacement. Informally, this process is denoted as $B \sim D$, representing the distribution over mini-batches of size $m$ sampled without replacement from $D$.

The full empirical loss $\hat{L}(D, \boldsymbol{\theta})$ can be expressed as the *expected* empirical loss over all mini-batches $B$ sampled from $D$, given by:

$$\hat{L}(D, \boldsymbol{\theta}) = \mathbb{E}_{B \sim D}[\hat{L}(B, \boldsymbol{\theta})], \tag{5.123}$$

where $\hat{L}(B, \boldsymbol{\theta})$ denotes the empirical loss computed *only* on the samples within the batch $B$.

SGD minimizes this expected empirical loss $\mathbb{E}_{B \sim D}[\hat{L}(B, \boldsymbol{\theta})]$ by iteratively updating parameters using noisy gradient estimates. These estimates are unbiased approximations of the true gradient of $\mathbb{E}_{B \sim D}[\hat{L}(B, \boldsymbol{\theta})]$. Since this expectation is equal to $\hat{L}(D, \boldsymbol{\theta})$, SGD effectively minimizes the same empirical loss as full-batch GD.

By Proposition 5.45, each batch loss $\hat{L}(B, \boldsymbol{\theta})$ also admits a decomposition in terms of its expected loss $L(\boldsymbol{\theta})$, the inverse rate function, and an "abnormality" term:

$$\hat{L}(B, \boldsymbol{\theta}) \;=\; L(\boldsymbol{\theta}) \;-\; \mathcal{I}_{\boldsymbol{\theta}}^{-1}(\alpha(B, \boldsymbol{\theta})). \tag{5.124}$$

As discussed in Section 5.4.2.2, $\alpha(B, \boldsymbol{\theta})$ measures again the degree to which the empirical loss on $B$ abnormally deviates from $L(\boldsymbol{\theta})$. Taking expectations over mini-batches by combining Equations (5.123) and (5.124):

$$\hat{L}(D, \boldsymbol{\theta}) \;=\; L(\boldsymbol{\theta}) \;-\; \mathbb{E}_{B \sim D}\Big[\mathcal{I}_{\boldsymbol{\theta}}^{-1}(\alpha(B, \boldsymbol{\theta}))\Big]. \tag{5.125}$$

To gain further insight, let $Q$ denote a distribution over the $\alpha$ values induced by mini-batches, where $\alpha_{\boldsymbol{\theta}}^{B}$ compactly denotes $\alpha(B, \boldsymbol{\theta})$:

$$\alpha_{\boldsymbol{\theta}}^{B} \sim Q(\alpha_{\boldsymbol{\theta}}^{B} | D, \boldsymbol{\theta})\,. \tag{5.126}$$

In other words, $Q(\alpha_{\boldsymbol{\theta}}^{B} | D, \boldsymbol{\theta})$ represents the distribution of *the abnormality scores* $\alpha$ that arises when sampling mini-batches $B$ from $D$. With this notation, using Equation (5.125), we can rewrite $\hat{L}(D, \boldsymbol{\theta})$ as

$$\hat{L}(D, \boldsymbol{\theta}) = L(\boldsymbol{\theta}) - \mathbb{E}_{\alpha_{\boldsymbol{\theta}}^{B} \sim Q(\cdot \,|\, D, \boldsymbol{\theta})}\Big[\mathcal{I}_{\boldsymbol{\theta}}^{-1}(\alpha_{\boldsymbol{\theta}}^{B})\Big]. \tag{5.127}$$

Equation (5.127) starts to reveal how the learning objective in SGD differs from the full-batch GD objective. While both objectives aim to minimize $\hat{L}(D, \boldsymbol{\theta})$, the dynamics of SGD are shaped by the distribution $Q(\alpha_{\boldsymbol{\theta}}^{B} | D, \boldsymbol{\theta})$, which is directly influenced by the variability in batch sampling. Extending the trade-offs discussed in Section 5.4.2.3 to this case, minimizing Equation (5.127) involves (i) reducing the expected loss $L(\boldsymbol{\theta})$, (ii) maximizing $\mathcal{I}_{\boldsymbol{\theta}}^{-1}(\cdot)$, and (iii) favoring models where $Q(\alpha_{\boldsymbol{\theta}}^{B} | D, \boldsymbol{\theta})$ is concentrated around large $\alpha_{\boldsymbol{\theta}}^B$ values. Points (i) and (ii) were already illustrated in Figure 5.21, and Figure 5.22 (top left) illustrates point (iii) by plotting the evolution of the mean of $Q(\alpha_{\boldsymbol{\theta}}^B|D,\boldsymbol{\theta})$, denoted $\mu(\alpha_{\boldsymbol{\theta}}^B) := \mathbb{E}_{Q(\cdot|D,\boldsymbol{\theta})}[\alpha_{\boldsymbol{\theta}}^B]$. Importantly, the distribution of $Q(\alpha_{\boldsymbol{\theta}}^B|D,\boldsymbol{\theta})$ is determined by the size of the batches $m$, which, as we will see later, introduces different biases in the optimization dynamics.

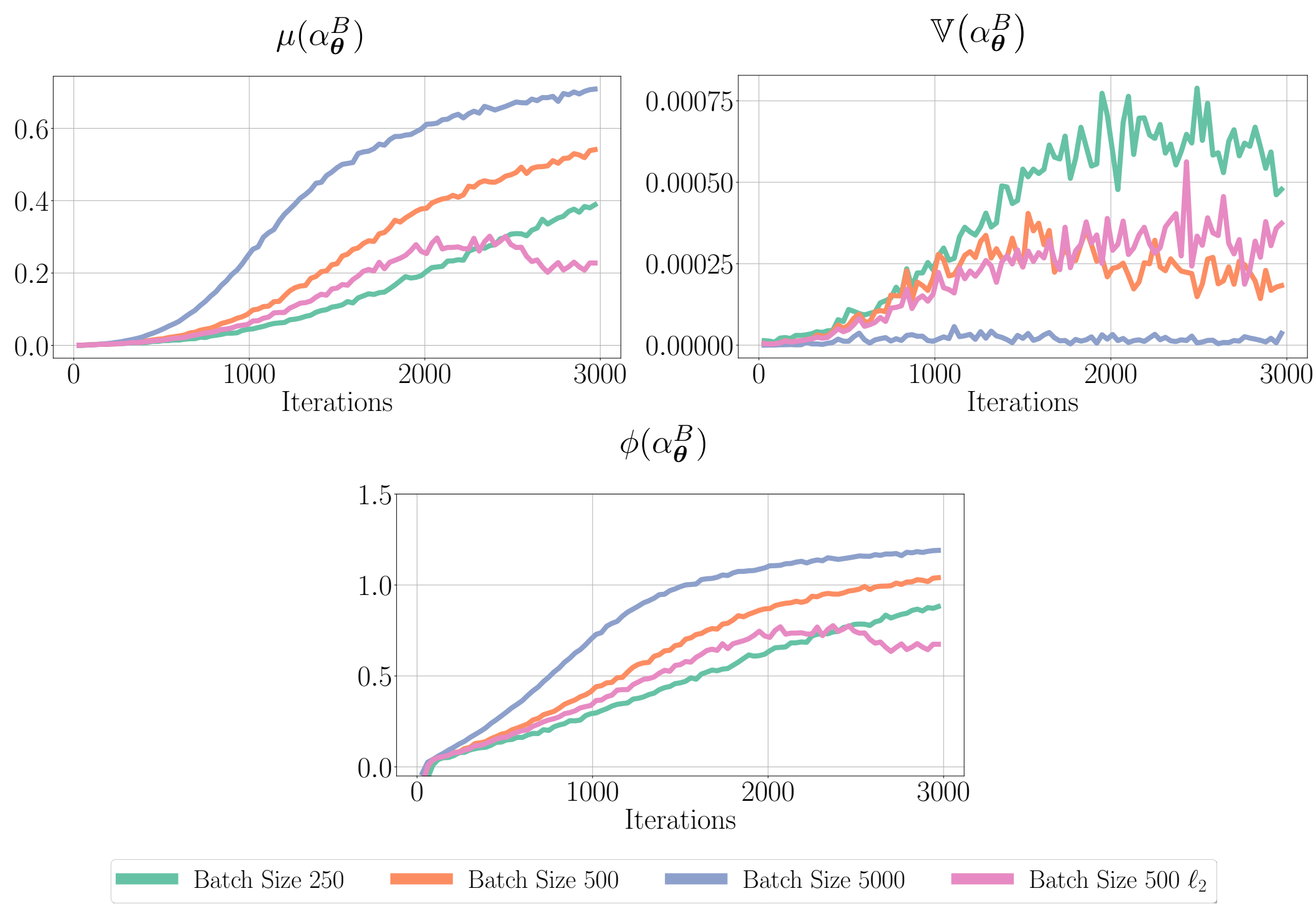

**Figure 5.22:** ***Batch-size effects on abnormality statistics (InceptionV3, CIFAR-10).*** *Evolution across iterations of the mean $\mu(\alpha_\theta^B)$ (top left), the variance $\mathbb{V}(\alpha_\theta^B)$ (top right), and the function $\phi(\alpha_\theta^B)$ (bottom) for batch sizes $m \in \{250, 500, 5000\}$ and for an $\ell_2$-regularized run at $m = 500$. The regularized model exhibits smaller mean/variance of abnormality relative to the unregularized counterpart at the same batch size. Definitions of these quantities are given in Equations (5.130) and (5.131).*

To gain deeper insight into these new biases in the optimization dynamics of SGD, $\mathcal{I}_{\boldsymbol{\theta}}^{-1}(\alpha_{\boldsymbol{\theta}}^B)$ is approximated by employing a second-order Taylor expansion of $\mathcal{I}_{\boldsymbol{\theta}}^{-1}(\alpha)$ around the mean $\mu(\alpha_{\boldsymbol{\theta}}^B)$. This yields:

$$\begin{aligned}&\mathbb{E}_{\alpha_{\boldsymbol{\theta}}^B\sim Q(\cdot\,|\,D,\boldsymbol{\theta})}[\mathcal{I}_{\boldsymbol{\theta}}^{-1}(\alpha_{\boldsymbol{\theta}}^B)]\\ &\quad\approx\ \mathcal{I}_{\boldsymbol{\theta}}^{-1}(\mu(\alpha_{\boldsymbol{\theta}}^B))\ +\ \tfrac{1}{2}\,\nabla_\alpha^2\,\mathcal{I}_{\boldsymbol{\theta}}^{-1}(\alpha)\,\big|_{\alpha=\mu(\alpha_{\boldsymbol{\theta}}^B)}\,\mathbb{V}(\alpha_{\boldsymbol{\theta}}^B).\end{aligned} \tag{5.128}$$

where $\mathbb{V}(\alpha_{\boldsymbol{\theta}}^B)$ denotes the variance of $\alpha_{\boldsymbol{\theta}}^B$, defined as $\mathbb{V}(\alpha_{\boldsymbol{\theta}}^B) := \mathbb{E}_{B\sim D}[(\alpha_{\boldsymbol{\theta}}^B - \mu(\alpha_{\boldsymbol{\theta}}^B))^2]$.

From standard properties of the rate function (Rockafellar, 1970), $\mathcal{I}_{\boldsymbol{\theta}}(a)$ is convex for $a \geq 0$. As a result, its inverse, $\mathcal{I}_{\boldsymbol{\theta}}^{-1}(\alpha)$, is concave for $\alpha \geq 0$, which implies $\nabla_\alpha^2\mathcal{I}_{\boldsymbol{\theta}}^{-1}(\alpha) \leq 0$ for $\alpha \geq 0$. Using the approximation in Equation (5.115), when $\mu(\alpha_{\boldsymbol{\theta}}^B) \geq 0$:

$$\nabla_\alpha^2\mathcal{I}_{\boldsymbol{\theta}}^{-1}(\alpha)\Big|_{\alpha=\mu(\alpha_{\boldsymbol{\theta}}^B)}\ \approx\ -\,(2\,\mu(\alpha_{\boldsymbol{\theta}}^B))^{-\frac{3}{2}}\,\sigma(\ell_{\boldsymbol{\theta}}), \tag{5.129}$$

where $\sigma(\ell_{\boldsymbol{\theta}})$ is the standard deviation of the individual losses $\ell(\mathbf{y}, \mathbf{x}, \boldsymbol{\theta})$ under the data-generating distribution $\nu$. Note that the condition $\mu(\alpha_{\boldsymbol{\theta}}^B) \geq 0$ is not restrictive, as SGD's bias towards large $\alpha_{\boldsymbol{\theta}}^B$ values ensures this condition after a few iterations, as empirically shown in Figure 5.22 (top left). Substituting Equation (5.129) into Equation (5.128) and then back into Equation (5.127) yields

$$\hat{L}(D, \boldsymbol{\theta}) \approx L(\boldsymbol{\theta}) \; - \; \mathcal{I}_{\boldsymbol{\theta}}^{-1}(\mu(\alpha_{\boldsymbol{\theta}}^B)) \; + \; \frac{0.5\mathbb{V}(\alpha_{\boldsymbol{\theta}}^B)}{\left(2\,\mu(\alpha_{\boldsymbol{\theta}}^B)\right)^{3/2}}\sigma(\ell_{\boldsymbol{\theta}})\,. \tag{5.130}$$

Equations (5.127) and (5.130) further reveal how the SGD objective differs from the full-batch GD objective. When the batch size $m$ equals the entire training set $n$, the distribution $Q(\alpha_{\boldsymbol{\theta}}^B|D, \boldsymbol{\theta})$ collapses to a Dirac delta at $\alpha(D, \boldsymbol{\theta})$, since all sampled batches $B \sim D$ coincide with $D$. Consequently, $\mathbb{V}(\alpha_{\boldsymbol{\theta}}^B)$ vanishes. For smaller batch sizes, however, $\mathbb{V}(\alpha_{\boldsymbol{\theta}}^B)$ remains nonzero due to the intrinsic randomness of mini-batch sampling. As illustrated in Figure 5.22 (top right), this variance is notably higher for smaller batches.

#### 5.4.3.1 Analysis of the Implicit Biases in SGD

Prior to analyzing the implicit biases of SGD, Equation (5.130) is expressed via the approximation in Equation (5.115), yielding the following, more interpretable, decomposition of $\hat{L}(D, \boldsymbol{\theta})$:

$$\hat{L}(D, \boldsymbol{\theta}) \approx L(\boldsymbol{\theta}) - \underbrace{\left(\sqrt{2\,\mu(\alpha_{\boldsymbol{\theta}}^B)} - \frac{0.5\,\mathbb{V}(\alpha_{\boldsymbol{\theta}}^B)}{\left(2\,\mu(\alpha_{\boldsymbol{\theta}}^B)\right)^{3/2}}\right)}_{\phi(\alpha_{\boldsymbol{\theta}}^B)} \sigma(\ell_{\boldsymbol{\theta}}). \tag{5.131}$$

Here, $\phi(\alpha_{\boldsymbol{\theta}}^B)$ aggregates the effects of the mean and variance of $\alpha_{\boldsymbol{\theta}}^B$ on the bias term in the empirical–loss decomposition; specifically, $\phi$ increases monotonically with $\mu(\alpha_{\boldsymbol{\theta}}^B)$ and decreases with $\mathbb{V}(\alpha_{\boldsymbol{\theta}}^B)$. Empirical support for the accuracy of Equations (5.130) and (5.131) is provided in Figure 5.23.

Consider the minimization of composite objectives of the form

$$f(\boldsymbol{\theta}) \; = \; g(\boldsymbol{\theta}) \; + \; \gamma\, h(\boldsymbol{\theta}), \tag{5.132}$$

with $\gamma > 0$. Optimization entails a trade-off between the primary term $g(\boldsymbol{\theta})$ and the bias term $h(\boldsymbol{\theta})$ (Boyd and Vandenberghe, 2004). Smaller values of $\gamma$ emphasize $g$, whereas larger values increase the influence of $h$, shifting solutions toward minimizers of $h$. The weight $\gamma$ may depend on $\boldsymbol{\theta}$.

Equation (5.131) can be mapped to Equation (5.132) by identifying $g(\boldsymbol{\theta}) = L(\boldsymbol{\theta})$, $h(\boldsymbol{\theta}) = \sigma(\ell_{\boldsymbol{\theta}})$, and $\gamma = \phi(\alpha_{\boldsymbol{\theta}}^B)$, which depend on $\boldsymbol{\theta}$ and on the batch size $m$. A smaller value of $\phi(\alpha_{\boldsymbol{\theta}}^B)$ weakens the preference for solutions with small $\sigma(\ell_{\boldsymbol{\theta}})$, whereas a larger value strengthens it. Mini-batch stochasticity increases $\mathbb{V}(\alpha_{\boldsymbol{\theta}}^B)$ for smaller $m$ (Figure 5.22, top right); since $\phi$ decreases with $\mathbb{V}(\alpha^B_{\boldsymbol{\theta}})$, smaller batches yield smaller $\phi(\alpha^B_{\boldsymbol{\theta}})$. This is consistent with Figure 5.21 (top right), where smaller batch sizes are associated with models exhibiting lower $\sigma(\ell_{\boldsymbol{\theta}})$, and with Figure 5.22 (bottom), which shows $\phi(\alpha^B_{\boldsymbol{\theta}})$ decreasing as $m$ decreases. The analysis therefore indicates a bias of SGD toward models whose empirical losses are more concentrated.

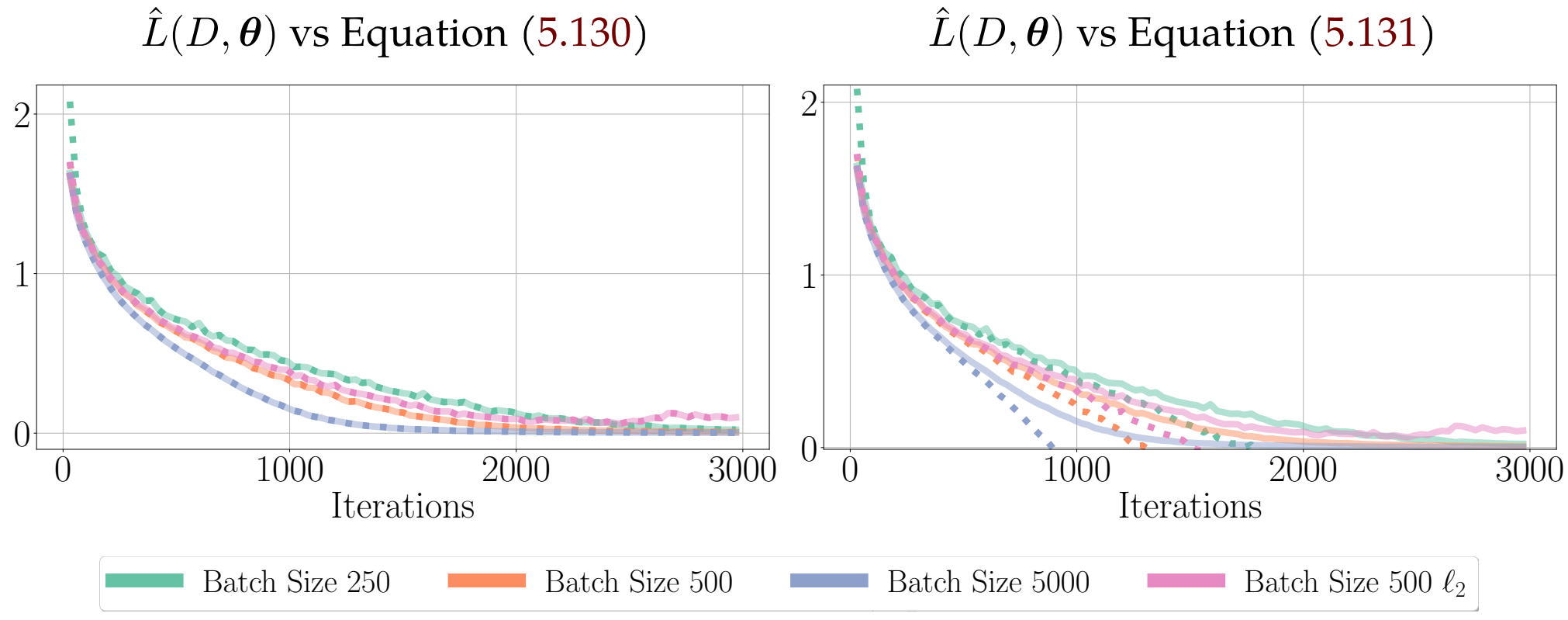

**Figure 5.23:** ***Empirical loss versus analytical approximations under varying batch sizes.*** *Solid curves show the evolution of the empirical loss $\hat{L}(D, \theta)$ across training iterations for batch sizes $m \in \{250, 500, 5000\}$ and for an $\ell_2$-regularized run at $m = 500$. Dashed curves display the corresponding approximations from Equation (5.130) (left) and Equation(5.131) (right). The panels illustrate the quality of each approximation throughout training for the different batch-size regimes.*

Equation (5.131) can also be cast into the template in Equation (5.132) by taking $g(\boldsymbol{\theta}) = L(\boldsymbol{\theta})$, $h(\boldsymbol{\theta}) = \phi(\alpha^B_{\boldsymbol{\theta}})$, and $\gamma = \sigma(\ell_{\boldsymbol{\theta}})$. In this view, larger values of $\mu(\alpha^B_{\boldsymbol{\theta}})$ directly increase the bias term $\phi(\alpha^B_{\boldsymbol{\theta}})$, thereby favoring models for which the distribution $Q(\alpha^B_{\boldsymbol{\theta}}|D, \boldsymbol{\theta})$ is centered at higher $\alpha^B_{\boldsymbol{\theta}}$. However, smaller batch sizes are associated with smaller $\sigma(\ell_{\boldsymbol{\theta}})$ (Figure 5.21, top right), which reduces the effective weight $\gamma$ and weakens this preference. Consistently, Figure 5.22 (top left) shows that decreasing the batch size shifts the optimization toward models with lower $\mu(\alpha^B_{\boldsymbol{\theta}})$.

The preceding arguments indicate that SGD exhibits an implicit bias toward solutions with smaller generalization error, driven by two effects: (i) a preference for models whose empirical loss is more concentrated around its expectation (smaller $\sigma(\ell_{\boldsymbol{\theta}})$); and (ii) an aversion to models exhibiting highly abnormal left–tail deviations (smaller $\mu(\alpha^B_{\boldsymbol{\theta}})$). Both effects strengthen as the mini-batch size decreases, so that small-batch SGD acts as an implicit regularizer that discourages convergence to poorly generalizing solutions. The empirical trends in Figures 5.21 and 5.22 accord with this interpretation.

#### 5.4.3.2 SGD and $\ell_2$ Regularization

To assess the theoretical claims from the preceding section, an ablation study is conducted on the effect of introducing $\ell_2$ regularization into SGD. The $\ell_2$ penalty (weight decay) adds a term proportional to the squared Euclidean norm of the parameters to the objective, yielding the surrogate loss $\hat{L}(D, \boldsymbol{\theta}) + \gamma\|\boldsymbol{\theta}\|_2^2$ with regularization strength $\gamma > 0$. By constraining the parameter norm, $\ell_2$ regularization alters the optimization

trajectory of SGD and systematically biases it toward models of smaller norm, which are typically associated with improved generalization.

A theoretical justification for this behavior is provided in Section 5.3.6, where an upper bound on the inverse rate function is given in terms of the $\ell_2$ norm of the parameters. Informally, for $s > 0$,

$$\mathcal{I}_{\boldsymbol{\theta}}^{-1}(s) \;\leq\; \sqrt{2Ms\,\|\boldsymbol{\theta}\|_2^2}, \tag{5.133}$$

for a model–dependent constant $M > 0$. Hence, smaller parameter norms entail a smaller inverse rate, i.e., greater concentration of the empirical loss.

Under this analysis, $\ell_2$ regularization is expected to *reinforce* the SGD bias toward models with more concentrated losses. This explicit bias is independent of the implicit mini-batch–induced bias discussed earlier, which arises because smaller batches increase $\mathbb{V}(\alpha_{\boldsymbol{\theta}}^B)$ and thereby modify the trade-off term in Equation (5.131). Consistent with this separation, Figure 5.22 (top right) shows that $\mathbb{V}(\alpha_{\boldsymbol{\theta}}^B)$ is essentially unchanged when comparing plain SGD at $m = 500$ with $\ell_2$-regularized SGD at the same batch size. Consequently, the two effects compound: $\ell_2$ regularization further pushes the optimizer toward models with higher loss concentration (lower variance). This pattern is observed in Figure 5.21 (top right): at $m = 500$, $\ell_2$-regularized SGD explores models with greater concentration than plain SGD; moreover, the concentration level resembles that achieved by plain SGD with a smaller batch size $m = 250$, which inherently induces stronger implicit regularization.

A second consequence follows from the analysis of the bias toward models with less abnormal deviations. Because $\ell_2$ regularization reduces the standard deviation $\sigma(\ell_{\boldsymbol{\theta}})$, the effective trade-off parameter $\gamma$ in the mapping $g(\boldsymbol{\theta}) = L(\boldsymbol{\theta})$, $h(\boldsymbol{\theta}) = \phi(\alpha_{\boldsymbol{\theta}}^B)$, $\gamma = \sigma(\ell_{\boldsymbol{\theta}})$ (Section 5.4.3.1) decreases, which weakens the tendency to favor models with large $\mu(\alpha_{\boldsymbol{\theta}}^B)$. Empirically, Figure 5.22 (left) shows that $\ell_2$-regularized SGD with $m = 500$ visits models with smaller abnormality levels (lower $\mu(\alpha_{\boldsymbol{\theta}}^B)$) than plain SGD at the same batch size, and achieves abnormality levels comparable to those of plain SGD with $m = 250$. Thus, $\ell_2$ regularization simultaneously reduces the dispersion of the empirical loss and mitigates the propensity to select models exhibiting large abnormal deviations, aligning with the theoretical predictions.

### 5.4.4 Related Work

The generalization properties of stochastic gradient descent (SGD) have been investigated from multiple angles. A prominent line of work attributes improved generalization to the tendency of SGD to converge to *flat* minima in the loss landscape. The notion of flat minima was introduced by Hochreiter and Schmidhuber (1997), who argued that solutions lying in wide, low-curvature regions exhibit better out-of-sample performance. Subsequent empirical and theoretical studies reported that small-batch SGD more frequently identifies flatter minima, whereas large-batch training is prone to sharper minima that correlate with degraded generalization (Keskar et al., 2017). Although the present analysis focuses on concentration properties of the empirical loss rather than the local geometry of the parameter space, the two perspectives are naturally connected: flatter minima are plausibly associated with predictors whose empirical losses display higher concentration and fewer abnormal deviations, thus aligning geometric flatness with distributional stability.

A complementary perspective emphasizes the *implicit regularization* provoked by SGD. Neyshabur et al. (2015c) argued that SGD implicitly biases solutions toward smaller parameter norms, in agreement with capacity-control principles that link norm-based complexity to generalization. Related work formalized spectral- and norm-based measures that correlate with test performance in deep networks (Bartlett et al., 2017). This line of research connects to concentration-based accounts through results showing that models with smaller norms admit tighter distribution-dependent control of generalization error; for instance, Section 5.3 established bounds that tie reduced parameter norms to smaller inverse rate functions, implying greater concentration of the empirical loss.

Another body of literature approaches SGD through concentration inequalities and stability arguments, treating the empirical loss of each hypothesis as a random variable (Bartlett et al., 2017; Golowich et al., 2018; Kawaguchi et al., 2022; Liang et al., 2019; Neyshabur et al., 2017a). While illuminating, many such results rely on uniform upper bounds that are known to be loose or even vacuous in highly over-parameterized regimes (Gastpar et al., 2024; Nagarajan and Kolter, 2019b). In addition, these approaches often do not differentiate the *model-specific* concentration behavior within the hypothesis class, potentially obscuring performance differences driven by distribution-dependent effects (Casado et al., 2024). By contrast, the analysis developed here employs a decomposition that isolates terms governing concentration and abnormal deviations of the empirical loss for each candidate model, thereby offering a more granular, distribution-dependent account of when SGD favors predictors with lower generalization error.

### 5.4.5 Discussion and Limitations

A perspective based on LDT has been developed to study the implicit bias of stochastic gradient descent (SGD). The generalization error was decomposed into two distribution-dependent components—*loss concentration* and *abnormality*—which together account for the tendency of SGD to favor solutions with smaller test error. In contrast to full-batch gradient descent, which is biased toward poorly concentrated and more abnormal empirical losses, mini-batch SGD implicitly promotes models with tighter concentration and fewer abnormal deviations, thereby acting as an implicit regularizer. Empirical results with deep convolutional networks support these claims: smaller batch sizes are consistently associated with more concentrated empirical-loss distributions and reduced abnormality. The introduction of explicit $\ell_2$ regularization further strengthens this tendency, indicating that implicit and explicit regularization mechanisms can be complementary.

The present analysis is grounded in standard assumptions (e.g., i.i.d. sampling and smooth losses), while modern deep learning often involves non-convex objectives and non-i.i.d. data. Extending the LDT framework to account for dependence, heavy tails, or non-stationarity constitutes an important direction. Another avenue is to compare the predictive utility of the LDT-based decomposition against traditional generalization bounds and to relate it more directly to geometric and norm-based explanations (e.g., flat minima and spectral complexity).

The analysis presented here is subject to several limitations. First, the Large Deviation Theory (LDT) framework presupposes i.i.d. sampling, whereas real-world datasets may exhibit correlations, distributional shifts, or temporal dependence that violate this assumption. Second, LDT yields asymptotic guarantees as the dataset size grows ($n \to \infty$); in finite-sample regimes typical of deep learning, sub-exponential correction terms $o(n, a)$ may be non-negligible and can affect the tightness of the resulting characterizations. Third, several theoretical insights rely on approximations (e.g., Equations (5.126) and (5.128)) that introduce simplifying assumptions; while the empirical evidence reported here indicates that these approximations produce consistent predictions, their validity may weaken in settings with highly non-smooth losses, heavy-tailed noise, or strong dependence, and results should be interpreted accordingly.

## 5.5 Conclusions

This chapter develops a unified theoretical and empirical framework to explain why and how modern neural networks generalize, even in regimes of over-parameterization where classical statistical learning theory fails. Across Sections 5.1–5.4, the chapter connects three seemingly distinct mechanisms—diversity, smoothness, and stochasticity—under a common probabilistic foundation rooted in PAC–Bayesian and large-deviation principles.

**Diversity and Ensemble Generalization.** This chapter first establishes that diversity among predictors is essential for generalization in ensemble learning. Using a PAC–Bayesian formulation, it introduces a unified diversity measure that decomposes the ensemble generalization error into two interpretable terms:

- the average individual model error, and
- a diversity-dependent correction that captures correlations among predictors.

This framework rigorously explains the empirical success of deep ensembles: randomization mechanisms such as independent initialization or stochastic optimization encourage functional diversity, improving test performance. The theoretical analysis (Theorem 5.1 and Theorem 5.7) and experiments confirm that ensembles with greater diversity achieve lower generalization error, even when their individual networks operate in the interpolation regime.

**Generalization Error and PAC–Chernoff Bounds.** The next section extends classical generalization bounds by deriving distribution-dependent PAC–Chernoff bounds that remain meaningful at interpolation. By expressing the generalization error in terms of a rate function from large-deviation theory, the analysis captures how model smoothness, regularization, and invariances determine the tightness of the bounds. This approach:

- explains the double-descent phenomenon through the shape of the rate function,
- shows that explicit regularizers (e.g., weight decay, distance-from-initialization, or gradient penalties) improve generalization by enlarging the rate function, and
- unifies regularization, data augmentation, and architectural invariances under a single smoothness-based framework.

**Implicit Bias of Stochastic Gradient Descent.** The final section analyzes SGD as a stochastic process using the same large-deviation perspective. It demonstrates that mini-batch stochasticity introduces an implicit regularization bias favoring flatter minima and models with higher loss concentration—hence, better generalization. A decomposition of the empirical loss in terms of the expected loss, rate function, and an "abnormality" term clarifies why full-batch gradient descent tends to overfit while SGD achieves robustness.

### Overall Synthesis

Together, these sections present a coherent picture of generalization in modern deep learning:

- **Diversity** explains how ensembles reduce variance through error de-correlation.
- **Smoothness and rate functions** explain why certain interpolating models generalize better than others.
- **Stochastic dynamics** explain how optimization implicitly regularizes without explicit penalties.

In essence, **generalization arises from the interplay of diversity, smoothness, and concentration**, all governed by probabilistic laws derived from PAC–Bayesian and large-deviation theory. This synthesis transcends classical VC and uniform-convergence analyses, offering a distribution-dependent understanding of deep learning generalization that bridges theory and practice.

# 6
# Conclusions and Future Work

This dissertation investigates the intersection between Bayesian inference and modern deep learning, with the central goal of improving our understanding of how neural networks generalize and how their uncertainty can be more accurately quantified and improved.

## 6.1 Conclusions

**Deep Variational Implicit Processes.** Chapter 3 developed the *Deep Variational Implicit Process* (DVIP), a scalable Bayesian model that extends the idea of implicit processes into deep architectures. The model is built upon the notion of *implicit processes*, which defines function distributions that are easy to sample from but lack tractable likelihoods. By composing these processes hierarchically, the DVIP framework produces non-Gaussian predictive distributions that are both expressive and amenable to variational inference in function space. Experimentation across regression and classification tasks showed that DVIPs perform on par with, or better than, deep Gaussian processes (DGPs) while being considerably more efficient. The experiments further illustrated how hierarchical structure, prior adaptation, and domain-specific priors (e.g., convolutional priors) contribute to improved expressivity and calibration. Overall, DVIP offers a principled and practical route toward uncertainty-aware deep learning.

**Post-hoc Uncertainty Estimation.** Whereas DVIP integrates uncertainty estimation during training, Chapter 4 focused on methods to equip pre-trained deterministic networks with Bayesian uncertainty estimates *after* training. Two distinct approaches were introduced: the *Variational Linearized Laplace Approximation* (VaLLA) and the *Fixed-Mean Gaussian Process* (FMGP). Both derive from a sparse GP formulation that allows fixing the predictive mean to that of a pre-trained network, effectively transforming it into a calibrated probabilistic predictor without retraining. VaLLA extends the classical Laplace approximation to function space, offering scalable uncertainty estimates whose computational cost is independent of dataset size. FMGP, in contrast, treats the deterministic model as a fixed-mean GP, learning only the covariance structure—resulting in minimal overhead and broad architectural compatibility. Experiments demonstrated that both methods provide well-calibrated predictions on large-scale benchmarks, including ImageNet and molecular property datasets for FMGPs, illustrating practical ways to incorporate Bayesian uncertainty into existing models.

**Generalization in Neural Networks.** The final methodological component, developed in Chapter 5, turned to one of the most fundamental open problems in modern deep learning: understanding why over-parameterized models generalize. The chapter proposed a unified theoretical framework linking three key factors—*diversity*, *smoothness*, and *stochasticity*—under a probabilistic formulation inspired by PAC–Bayes and large-deviation theory.

Section 5.2 demonstrated that functional diversity among ensemble models is crucial for good generalization in ensembles. A decomposition of the generalization error in terms of predictor diversity explained why mechanisms such as random initialization or stochastic training often lead to better test performance.

Section 5.3 extended these results through PAC–Chernoff bounds that remain informative even in interpolation regimes. These bounds express generalization error through a *rate function* describing concentration of empirical loss, offering a probabilistic explanation for phenomena like double descent. Classical regularization methods—including weight decay and distance-from-initialization constraints—emerged naturally within this framework as means of enlarging the rate function, thereby improving smoothness and generalization.

Section 5.4 examined stochastic gradient descent (SGD) through the same lens, showing that stochasticity itself functions as an implicit regularizer. By decomposing the empirical loss into its expected value, a rate function, and an abnormality term, the analysis clarified how mini-batch SGD tends to prefer flatter minima and more concentrated loss distributions, whereas full-batch optimization is more prone to overfitting. Collectively, these insights established a probabilistic, distribution-dependent view of generalization that ties together theory, optimization dynamics, and empirical findings.

In summary, this dissertation shows that probabilistic reasoning—when extended beyond its traditional scope—offers both practical tools for scalable uncertainty quantification and theoretical insight into generalization in deep learning. By bringing together Bayesian inference, large-deviation theory, and function-space modeling, this work contributes to a unified probabilistic perspective on learning that bridges theory, algorithms, and empirical behavior.

## 6.2 Future Work

Several natural extensions stem from this work.

**Advances in Deep Variational Implicit Processes.** Future research on DVIP could explore alternative approximate inference techniques that do not rely on Gaussian or GP-based approximations. For instance, normalizing-flow or diffusion-based variational families could be used to model richer posterior dependencies in function space. Furthermore, richer base models beyond Bayesian linear approximations could be explored; Sparse implicit processes (SIP) offer a promising direction for improving posterior approximations. Beyond inference, DVIPs could benefit from more structured priors, such as recurrent or attention-based priors for sequence modeling, or convolutional priors tailored to spatial data. Integrating DVIP into autoencoder or generative architectures also represents a compelling opportunity to unify representation learning and uncertainty quantification within a single probabilistic framework.

**Extensions of Post-hoc Uncertainty Methods.** For post-hoc methods such as VaLLA and FMGP, several improvements are conceivable. In VaLLA, more flexible variational approximations could replace the Gaussian assumption on the linearized posterior, potentially improving calibration in highly non-linear regions. For FMGP, future work could focus on designing task-specific kernels—particularly for images or spatio-temporal data—where structure-aware kernels can capture meaningful correlations between feature activations. Another interesting direction is the approximation of the Jacobian kernel, which could provide computationally efficient uncertainty estimates consistent with the local geometry of the network. Together, these developments would broaden the applicability of post-hoc Bayesianization methods to a wider range of architectures and modalities.

**Future Theoretical Directions.** The theoretical framework in Chapter 5 also opens several avenues for further investigation. One direction involves relaxing some of the simplifying assumptions used in the large-deviation analysis, such as independence between samples, to better capture the behavior of modern training setups. Extending the analysis to structured data distributions, correlated noise, or non-i.i.d. regimes could provide a more realistic description of generalization in deep networks. Another promising line of work is connecting the PAC–Chernoff framework with complementary approaches, such as information-theoretic generalization bounds or algorithmic stability theory. Finally, the rate-function formalism could be applied to study other phenomena in deep learning—such as transfer learning, continual learning, or the behavior of large language models—where distribution-dependent generalization remains poorly understood.

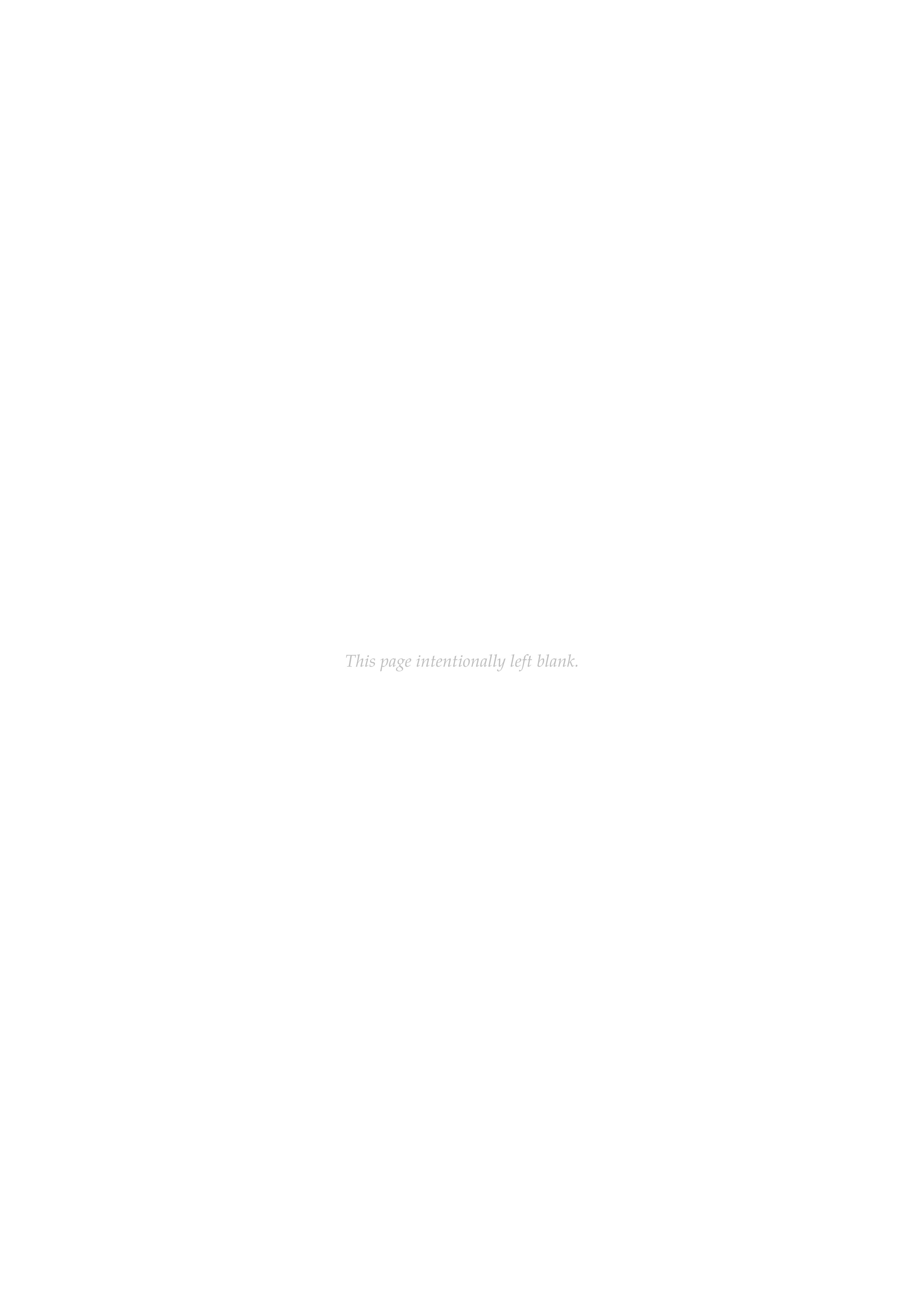

*This page intentionally left blank.*

# Bibliography

Antorán, J., Padhy, S., Barbano, R., Nalisnick, E. T., Janz, D., & Hernández-Lobato, J. M. (2023). Sampling-based inference for large linear models, with application to linearised Laplace. *International Conference on Learning Representations*.

Aronszajn, N. (1950). Theory of reproducing kernels. *Transactions of the American Mathematical Society*, *68*(3), 337–404.

Arora, S., Cohen, N., & Hazan, E. (2018a). On the optimization of deep networks: Implicit acceleration by overparameterization. *International Conference on Machine Learning*, 244–253.

Arora, S., Ge, R., Neyshabur, B., & Zhang, Y. (2018b). Stronger generalization bounds for deep nets via a compression approach. *International Conference on Machine Learning*, 254–263.

Arora, S., Liang, Y., & Ma, T. (2016). Why are deep nets reversible: A simple theory, with implications for training. https://arxiv.org/abs/1511.05653

Athreya, K. B., & Lahiri, S. N. (2006). *Measure theory and probability theory*. Springer.

Bach, F. (2017). On the equivalence between quadrature rules and random features. *Advances in Neural Information Processing Systems 30*, 456–467.

Baldi, P., Sadowski, P., & Whiteson, D. (2014). Searching for exotic particles in high-energy physics with deep learning. *Nature communications*, *5*(1), 1–9.

Balog, M., Salakhutdinov, R., & Ghahramani, Z. (2016). Mondrian forests for large-scale regression when uncertainty matters. *International Conference on Artificial Intelligence and Statistics* (*AISTATS*), 1119–1127.

Barber, D. (2012). *Bayesian reasoning and machine learning*. Cambridge University Press.

Bartlett, P. L., Foster, D. J., & Telgarsky, M. J. (2017). Spectrally-normalized margin bounds for neural networks. *Advances in neural information processing systems*, *30*.

Bartlett, P. L., Harvey, N., Liaw, C., & Mehrabian, A. (2019). Nearly-tight VC-dimension and pseudodimension bounds for piecewise linear neural networks. *The Journal of Machine Learning Research*, *20*(1), 2285–2301.

Bartlett, P. L., Long, P. M., Lugosi, G., & Tsigler, A. (2020). Benign overfitting in linear regression. *Proceedings of the National Academy of Sciences*, *117*(48), 30063–30070.

Bauer, M., van der Wilk, M., & Rasmussen, C. E. (2016). Understanding probabilistic sparse Gaussian process approximations. *Advances in Neural Information Processing Systems*, *29*, 1533–1541.

Becker, R. A. (2012). The variance drain and Jensen's inequality. *CAEPR Working Paper No. 2012-004*. http://dx.doi.org/10.2139/ssrn.2027471

Behboodi, A., Cesa, G., & Cohen, T. S. (2022). A PAC-Bayesian generalization bound for equivariant networks. *Advances in Neural Information Processing Systems*, *35*, 5654–5668.

Belkin, M., Hsu, D., Ma, S., & Mandal, S. (2019). Reconciling modern machine-learning practice and the classical bias–variance trade-off. *Proceedings of the National Academy of Sciences*, *116*(32), 15849–15854.

Berend, D., & Kontorovich, A. (2016). A finite sample analysis of the naive Bayes classifier. *Journal of Machine Learning Research*.

Bergamin, F., Moreno-Muñoz, P., Hauberg, S., & Arvanitidis, G. (2023). Riemannian Laplace approximations for Bayesian neural networks. *Advances in Neural Information Processing Systems*.

Bertin-Mahieux, T. (2011). Year Prediction MSD [DOI: https://doi.org/10.24432/C50K61].

Billingsley, P. (1995). *Probability and measure* (3rd). John Wiley & Sons.

Bishop, C. M. (1995). *Neural networks for pattern recognition*. Oxford University Press.

Bishop, C. M. (2006). *Pattern recognition and machine learning*. Springer.

Blei, D. M., Kucukelbir, A., & McAuliffe, J. D. (2017). Variational inference: A review for statisticians. *Journal of the American Statistical Association*, *112*(518), 859–877. https://doi.org/10.1080/01621459.2017.1285773

Blundell, C., Cornebise, J., Kavukcuoglu, K., & Wierstra, D. (2015). Weight uncertainty in neural networks. *International Conference on Machine Learning*, 1613–1622.

Bogachev, V. I. (1998). *Gaussian measures* (Vol. 62). American Mathematical Society.

Bogachev, V. I. (2007). *Measure theory*. Springer.

Botev, A., Ritter, H., & Barber, D. (2017). Practical Gauss-Newton optimisation for deep learning. *International Conference on Machine Learning*, 557–565.

Bottou, L. (2010). Large-scale machine learning with stochastic gradient descent. In *Proceedings of compstat'2010* (pp. 177–186). Springer.

Boyd, S., & Vandenberghe, L. (2004). *Convex optimization*. Cambridge university press.

Breiman, L. (1996). Bagging predictors. *Machine Learning*, *24*(2).

Breiman, L. (2001). Random forests. *Machine learning*, *45*(1), 5–32.

Bronstein, M. M., Bruna, J., Cohen, T., & Veličković, P. (2021). Geometric deep learning: Grids, groups, graphs, geodesics, and gauges. *arXiv preprint arXiv:2104.13478*.

Brown, G. (2009). An information theoretic perspective on multiple classifier systems. *International Workshop on Multiple Classifier Systems*, 344–353.

Brown, G., Wyatt, J. L., & Tiňo, P. (2005). Managing diversity in regression ensembles. *Journal of Machine Learning Research*, *6*(Sep), 1621–1650.

Bubeck, S., Li, Y., & Nagaraj, D. M. (2021). A law of robustness for two-layers neural networks. *Conference on Learning Theory*, 804–820.

Bubeck, S., & Sellke, M. (2023). A universal law of robustness via isoperimetry. *Journal of the ACM*, *70*(2), 1–18.

Bui, T., Hernández-Lobato, D., Hernandez-Lobato, J., Li, Y., & Turner, R. (2016). Deep Gaussian processes for regression using approximate expectation propagation. *International conference on machine learning*, 1472–1481.

Bui, T. D., Yan, J., & Turner, R. E. (2017). A unifying framework for Gaussian process pseudo-point approximations using power expectation propagation. *Journal of Machine Learning Research*, *18*, 1–72.

Buntine, W., & Jaakkola, T. (2022). Alpha–divergences, expectation propagation and Bayesian neural networks. *International Conference on Artificial Intelligence and Statistics (AISTATS)*, 1234–1242.

Burt, D. R., & Rasmussen, C. E. (2020). Convergence of sparse variational inference in Gaussian processes. *Journal of Machine Learning Research*, *21*(131), 1–63.

Buschjäger, S., Pfahler, L., & Morik, K. (2020). Generalized negative correlation learning for deep ensembling. *arXiv preprint arXiv:2011.02952*.

Casado, I., Ortega, L. A., Masegosa, A. R., & Pérez, A. (2024). PAC-Bayes-Chernoff bounds for unbounded losses. *Proceedings of the 38th Conference on Neural Information Processing Systems*.

Catoni, O. (2007). PAC-Bayesian supervised classification: The thermodynamics of statistical learning. *arXiv preprint arXiv:0712.0248*.

Chafaï, D. (2004). Entropies, convexity, and functional inequalities, on Φ-entropies and Φ-sobolev inequalities. *Journal of Mathematics of Kyoto University*, *44*(2), 325–363.

Chandra, A., & Yao, X. (2004). Divace: Diverse and accurate ensemble learning algorithm. *International Conference on Intelligent Data Engineering and Automated Learning*, 619–625.

Chen, S., Dobriban, E., & Lee, J. H. (2020). A group-theoretic framework for data augmentation. *The Journal of Machine Learning Research*, *21*(1), 9885–9955.

Chen, T., Fox, E., & Guestrin, C. (2014). Stochastic gradient hamiltonian monte carlo. *International Conference on Machine Learning*, 1683–1691.

Chen, T., & Guestrin, C. (2016). Xgboost. *Proceedings of the 22nd ACM SIGKDD International Conference on Knowledge Discovery and Data Mining*. https://doi.org/10.1145/2939672.2939785

Cheng, C.-A., & Boots, B. (2016). Incremental variational sparse Gaussian process regression. *Advances in Neural Information Processing Systems*, *29*, 4403–4411.

Cheng, C.-A., & Boots, B. (2017). Variational inference for Gaussian process models with linear complexity. *Advances in Neural Information Processing Systems*, *30*, 5184–5194.

Chernoff, H. (1952). A measure of asymptotic efficiency for tests of a hypothesis based on the sum of observations. *The Annals of Mathematical Statistics*, 493–507.

Cho, Y., & Saul, L. (2009). Kernel methods for deep learning. *Advances in neural information processing systems*, *22*.

Conway, J. B. (1990). *A course in functional analysis* (2nd). Springer.

Cortez, P., Cerdeira, A., Almeida, F., Matos, T., & Reis, J. (2009). *Modeling wine preferences by data mining from physicochemical properties*. https://archive.ics.uci.edu/ml/datasets/wine+quality

Cramér, H. (1938). Sur un nouveau théoreme-limite de la théorie des probabilités. *Actual. Sci. Ind.*, *736*, 5–23.

Cunningham, P., & Carney, J. (2000). Diversity versus quality in classification ensembles based on feature selection. *European Conference on Machine Learning*, 109–116.

Cutajar, K., Bonilla, E. V., Michiardi, P., & Filippone, M. (2017). Random feature expansions for deep Gaussian processes. *International Conference on Machine Learning*, 884–893.

Da Prato, G., & Zabczyk, J. (2014). *Stochastic equations in infinite dimensions* (2nd). Cambridge University Press.

Damianou, A., & Lawrence, N. D. (2013). Deep Gaussian processes. *Artificial intelligence and statistics*, 207–215.

Daxberger, E., Kristiadi, A., Immer, A., Eschenhagen, R., Bauer, M., & Hennig, P. (2021a). Laplace Redux - Effortless Bayesian deep learning. *Advances in Neural Information Processing Systems*, *34*, 20089–20103.

Daxberger, E., Nalisnick, E., Allingham, J. U., Antoran, J., & Hernandez-Lobato, J. M. (2021b). Bayesian deep learning via subnetwork inference. *International Conference on Machine Learning*, 2510–2521.

Deisenroth, M. P., Fox, D., & Rasmussen, C. E. (2015). *Gaussian processes for data-efficient learning in robotics and control* (Vol. 118). Springer.

Deng, L. (2012). The mnist database of handwritten digit images for machine learning research. *IEEE Signal Processing Magazine*, *29*, 141–142.

Deng, Z., Zhou, F., & Zhu, J. (2022). Accelerated linearized Laplace approximation for Bayesian deep learning. *Advances in Neural Information Processing Systems*, *35*, 2695–2708.

Deng, Z., & Zhu, J. (2023). Bayesadapter: Being Bayesian, inexpensively and reliably, via Bayesian fine-tuning. *Asian Conference on Machine Learning*, 280–295.

Dietterich, T. G. (2000). Ensemble methods in machine learning. *International workshop on multiple classifier systems*, 1–15.

Dua, D., & Graff, C. (2019). UCI machine learning repository. http://archive.ics.uci.edu/ml

Durrett, R. (2019). *Probability: Theory and examples* (5th). Cambridge University Press.

Dutordoir, V., Durrande, N., & Hensman, J. (2020). Sparse Gaussian processes with spherical harmonic features. *International Conference on Machine Learning*, 2793–2802.

Duvenaud, D., Rippel, O., Adams, R., & Ghahramani, Z. (2014). Avoiding pathologies in very deep networks. *Artificial Intelligence and Statistics*, 202–210.

Elesedy, B. (2022). Group symmetry in PAC learning. *ICLR 2022 Workshop on Geometrical and Topological Representation Learning*.

Ellis, R. S. (2012). *Entropy, large deviations, and statistical mechanics* (Vol. 271). Springer Science & Business Media.

Feldman, V., & Zhang, C. (2020). What neural networks memorize and why: Discovering the long tail via influence estimation. *Advances in Neural Information Processing Systems*, *33*, 2881–2891.

Fey, M., & Lenssen, J. E. (2019). Fast graph representation learning with pytorch geometric. https://arxiv.org/abs/1903.02428

Filippone, M., & Engler, U. (2015). Pseudo-marginal Bayesian inference for Gaussian processes. *IEEE Transactions on Pattern Analysis and Machine Intelligence*, *37*(3), 546–560.

Flam-Shepherd, D., Requeima, J., & Duvenaud, D. (2017). Mapping Gaussian process priors to Bayesian neural networks. *NIPS Bayesian deep learning workshop*, *3*.

Foong, A. Y., Li, Y., Hernández-Lobato, J. M., & Turner, R. E. (2019). 'In-Between' Uncertainty in Bayesian neural networks. *ICML Workshop on Uncertainty and Robustness in Deep Learning*.

Fort, S., Hu, H., & Lakshminarayanan, B. (2019). Deep ensembles: A loss landscape perspective. *arXiv preprint arXiv:1912.02757*.

Freund, Y., & Schapire, R. E. (1996). Experiments with a new boosting algorithm. *Proceedings of the International Conference on Machine Learning (ICML)*.

Gal, Y. (2016). *Uncertainty in deep learning* [Doctoral dissertation, University of Cambridge].

Gal, Y., & Turner, R. E. (2015). Improving the Gaussian process sparse spectrum approximation by representing uncertainty in frequency inputs. *International Conference on Machine Learning (ICML)*, 655–664.

Gardner, J. R., Pleiss, G., Bindel, D., Weinberger, K. Q., & Wilson, A. G. (2018). GPyTorch: Blackbox matrix-matrix Gaussian process inference with GPU acceleration. *Advances in Neural Information Processing Systems 31*.

Gastpar, M., Nachum, I., Shafer, J., & Weinberger, T. (2024). Fantastic generalization measures are nowhere to be found. *International Conference on Learning Representations*.

Gelman, A., Carlin, J. B., Stern, H. S., Dunson, D. B., Vehtari, A., & Rubin, D. B. (2013). *Bayesian data analysis* (3rd). CRC Press.

Geman, S., Bienenstock, E., & Doursat, R. (1992). Neural networks and the bias/variance dilemma. *Neural computation*, *4*(1), 1–58.

Germain, P., Bach, F., Lacoste, A., & Lacoste-Julien, S. (2016). PAC-Bayesian theory meets Bayesian inference. *Advances in Neural Information Processing Systems*, 1884–1892.

Germain, P., Lacasse, A., Laviolette, F., Marchand, M., & Roy, J.-F. (2015). Risk bounds for the majority vote: From a PAC-Bayesian analysis to a learning algorithm. *Journal of Machine Learning Research*, *16*.

Gilbert, A. C., Zhang, Y., Lee, K., Zhang, Y., & Lee, H. (2017). Towards understanding the invertibility of convolutional neural networks, 1703–1710. https://doi.org/10.24963/ijcai.2017/236

Gilks, W. R., Richardson, S., & Spiegelhalter, D. (1995). *Markov chain monte carlo in practice*. CRC press.

Girard, A., Rasmussen, C. E., Quiñonero-Candela, J., & Murray-Smith, R. (2003). Gaussian process priors with uncertain inputs — application to multiple-step ahead time series forecasting. *Advances in Neural Information Processing Systems 15* (*NIPS 2002*), 529–536.

Girshick, R., Donahue, J., Darrell, T., & Malik, J. (2014). Rich feature hierarchies for accurate object detection and semantic segmentation. *Proceedings of the IEEE conference on computer vision and pattern recognition*, 580–587.

Gneiting, T., & Raftery, A. E. (2007). Strictly proper scoring rules, prediction, and estimation. *Journal of the American statistical Association*, *102*, 359–378.

Golowich, N., Rakhlin, A., & Shamir, O. (2018). Size-independent sample complexity of neural networks. *Conference On Learning Theory*, 297–299.

Goodfellow, I., Bengio, Y., Courville, A., & Bengio, Y. (2016). *Deep learning* (Vol. 1). MIT press.

Goodfellow, I., Lee, H., Le, Q., Saxe, A., & Ng, A. (2009). Measuring invariances in deep networks. *Advances in neural information processing systems*, *22*.

Gouk, H., Frank, E., Pfahringer, B., & Cree, M. J. (2021). Regularisation of neural networks by enforcing lipschitz continuity. *Machine Learning*, *110*, 393–416.

Graves, A. (2011). Practical variational inference for neural networks. *Advances in Neural Information Processing Systems*, *24*, 2348–2356.

Grimmett, G. R., & Stirzaker, D. R. (2001). *Probability and random processes* (3rd). Oxford University Press.

Gross, L. (1967). Abstract wiener spaces.

Gunasekar, S., Lee, J. D., Soudry, D., & Srebro, N. (2018). Implicit bias of gradient descent on linear convolutional networks. *Advances in neural information processing systems*, *31*.

Guo, C., Pleiss, G., Sun, Y., & Weinberger, K. Q. (2017). On calibration of modern neural networks. *International Conference on Machine Learning*, 1321–1330.

Hansen, L. K., & Salamon, P. (1990). Neural network ensembles. *IEEE Transactions on Pattern Analysis and Machine Intelligence*, *12*(10).

Hardt, M., & Ma, T. (2017). Identity matters in deep learning. *International Conference on Learning Representations*.

Hastie, T., Tibshirani, R., & Friedman, J. (2001). *The elements of statistical learning: Data mining, inference, and prediction*. Springer.

Hastie, T., Tibshirani, R., & Friedman, J. (2009). *The elements of statistical learning: Data mining, inference, and prediction* (Second). Springer.

Havasi, M., Hernández-Lobato, J. M., & Murillo-Fuentes, J. J. (2018). Inference in deep Gaussian processes using stochastic gradient Hamiltonian Monte Carlo. *Advances in Neural Information Processing Systems*, *31*.

He, K., Zhang, X., Ren, S., & Sun, J. (2016a). Deep residual learning for image recognition. *Proceedings of the IEEE conference on computer vision and pattern recognition*, 770–778.

He, K., Zhang, X., Ren, S., & Sun, J. (2016b). Identity mappings in deep residual networks. *European conference on computer vision*, 630–645.

Hensman, J., Durrande, N., Solin, A., et al. (2017). Variational fourier features for Gaussian processes. *Journal of Machine Learning Research*, *18*(1), 5537–5588.

Hensman, J., Fusi, N., & Lawrence, N. D. (2013). Gaussian processes for big data. *Conference on Uncertainty in Artificial Intelligence* (*UAI*), 282–290.

Herdegen, M. (2008). The theorem of bahadur and rao and large portfolio losses. *Journal of Applied Mathematics*, *2011*.

Hernandez-Lobato, J. M., Li, Y., Rowland, M., Bui, T., Hernández-Lobato, D., & Turner, R. (2016). Black-box alpha divergence minimization. *International Conference on Machine Learning*, 1511–1520.

Hernández-Lobato, D., Hernández-Lobato, J., & Dupont, P. (2011). Robust multi-class Gaussian process classification. *Advances in Neural Information Processing Systems*, *24*.

Hernández-Lobato, J. M., & Adams, R. (2015). Probabilistic backpropagation for scalable learning of Bayesian neural networks. *International conference on machine learning*, 1861–1869.

Hoch, T. (2015). An ensemble learning approach for the kaggle taxi travel time prediction challenge. *Proceedings of the 2015th International Conference on ECML PKDD Discovery Challenge-Volume 1526*, 52–62.

Hochreiter, S., & Schmidhuber, J. (1997). Flat minima. *Neural Computation*, *9*(1), 1–42.

Hoeffding, W. (1963). Probability inequalities for sums of bounded random variables. *Journal of the American Statistical Association*, *58*(301), 13–30.

Hu, W., Li, Z., & Yu, D. (2020). Simple and effective regularization methods for training on noisily labeled data with generalization guarantee. *International Conference on Learning Representations*.

Hytönen, T., Van Neerven, J., Veraar, M., & Weis, L. (2018). *Analysis in banach spaces: Volume ii: Probabilistic methods and operator theory* (Vol. 67). Springer.

Immer, A., Korzepa, M., & Bauer, M. (2021). Improving predictions of Bayesian neural nets via local linearization. *International Conference on Artificial Intelligence and Statistics*, 703–711.

Izmailov, P., Maddox, W. J., Kirichenko, P., Garipov, T., Vetrov, D., & Wilson, A. G. (2020). Subspace inference for Bayesian deep learning. *Uncertainty in Artificial Intelligence*, 1169–1179.

Jaakkola, T. S. (2001). Tutorial on variational approximation methods. In M. Opper & D. Saad (Eds.), *Advanced mean field methods: Theory and practice* (pp. 129–160). MIT Press.

Jacot, A., Gabriel, F., & Hongler, C. (2018). Neural tangent kernel: Convergence and generalization in neural networks. *Advances in neural information processing systems*, *31*.

Jain, S., Liu, G., Mueller, J., & Gifford, D. (2020). Maximizing overall diversity for improved uncertainty estimates in deep ensembles. *Proceedings of the AAAI Conference on Artificial Intelligence*, *34*(04), 4264–4271.

Jiang, Y., Neyshabur, B., Mobahi, H., Krishnan, D., & Bengio, S. (2020). Fantastic generalization measures and where to find them. *International Conference on Learning Representations*.

Jiang, Z., Liu, H., Fu, B., & Wu, Z. (2017). Generalized ambiguity decompositions for classification with applications in active learning and unsupervised ensemble pruning. *Proceedings of the AAAI Conference on Artificial Intelligence*, *31*(1).

Jordan, M. I. (1999). *An introduction to variational methods for graphical models*. Springer.

Kawaguchi, K., Kaelbling, L. P., & Bengio, Y. (2022). Generalization in deep learning. In *Mathematical aspects of deep learning*. Cambridge University Press. https://doi.org/10.1017/9781009025096.003

Kendall, A., & Gal, Y. (2017). What uncertainties do we need in Bayesian deep learning for computer vision? *Advances in Neural Information Processing Systems*, 5574–5584.

Keskar, N. S., Nocedal, J., Tang, P., Mudigere, D., & Smelyanskiy, M. (2017). On large-batch training for deep learning: Generalization gap and sharp minima. *Proceedings of ICLR 2017*.

Khan, M. E. E., Immer, A., Abedi, E., & Korzepa, M. (2019). Approximate inference turns deep networks into Gaussian processes. *Advances in Neural Information Processing Systems*, *32*, 3094–3104.

Kingma, D. P., & Ba, J. (2015). Adam: A method for stochastic optimization. *International Conference for Learning Representations*.

Kingma, D. P., & Welling, M. (2014). Auto-encoding variational Bayes. *International Conference on Learning Representations*.

Klenke, A. (2013). *Probability theory: A comprehensive course* (2nd). Springer.

Knoblauch, J., Jewson, J., & Damoulas, T. (2022). An optimization-centric view on Bayes' rule: Reviewing and generalizing variational inference. *Journal of Machine Learning Research*, *23*(132), 1–109.

Kolmogorov, A. N. (1956). *Foundations of the theory of probability* (2nd English ed.) [Translated from the 2nd Russian edition by Nathan Morrison]. Chelsea / Addison–Wesley.

Krizhevsky, A., Hinton, G., et al. (2009). Learning multiple layers of features from tiny images.

Krogh, A., & Vedelsby, J. (1994). Neural network ensembles, cross validation and active learning. *Proceedings of the 7th International Conference on Neural Information Processing Systems*, 231–238.

Kuncheva, L. I., & Whitaker, C. J. (2003). Measures of diversity in classifier ensembles and their relationship with the ensemble accuracy. *Machine learning*, *51*(2), 181–207.

Kuo, H.-H. (2006). Gaussian measures in banach spaces. In *Gaussian measures in banach spaces* (pp. 1–109). Springer.

Lakshminarayanan, B., Pritzel, A., & Blundell, C. (2017). Simple and scalable predictive uncertainty estimation using deep ensembles. *Advances in Neural Information Processing Systems*, 6402–6413.

Langford, J., & Shawe-Taylor, J. (2002). PAC-Bayes & margins. *Advances in Neural Information Processing Systems (NeurIPS)*.

Lauritzen, S. L. (1996). *Graphical models*. Clarendon Press (Oxford University Press).

Laviolette, F., Marchand, M., & Roy, J.-F. (2011). From PAC-Bayes bounds to quadratic programs for majority votes. *Proceedings of the International Conference on Machine Learning (ICML)*.

Lawrence, N. D. (2001). *Variational inference in probabilistic models* [Doctoral dissertation, Citeseer].

Lawrence, N. D., & Moore, A. J. (2007). Hierarchical Gaussian process latent variable models. *Proceedings of the 24th international conference on Machine learning*, 481–488.

Le, Q., Sarlós, T., & Smola, A. (2013). Fastfood—approximating kernel expansions in loglinear time. *International Conference on Machine Learning*, 244–252.

LeCun, Y., Boser, B., Denker, J. S., Henderson, D., Howard, R. E., Hubbard, W., & Jackel, L. D. (1989). Backpropagation applied to handwritten zip code recognition. *Neural computation*, *1*(4), 541–551.

LeCun, Y., Bottou, L., Bengio, Y., & Haffner, P. (1998). Gradient-based learning applied to document recognition. *Proceedings of the IEEE*, *86*(11), 2278–2324.

Lee, J. [Jaehoon], Bahri, Y., Novak, R., Schoenholz, S. S., Pennington, J., & Sohl-Dickstein, J. (2018). Deep neural networks as Gaussian processes. *6th International Conference on Learning Representations, ICLR 2018, Vancouver, BC, Canada, April 30 - May 3, 2018, Conference Track Proceedings*.

Lee, J. [Jongseok], Feng, J., Humt, M., Müller, M. G., & Triebel, R. (2022). Trust your robots! predictive uncertainty estimation of neural networks with sparse Gaussian processes. *Conference on Robot Learning*, 1168–1179.

Lee, S., Purushwalkam, S., Cogswell, M., Ranjan, V., Crandall, D. J., & Batra, D. (2016). Stochastic multiple choice learning for training diverse deep ensembles. *Advances in Neural Information Processing Systems (NeurIPS)*.

Leibig, C., Allken, V., Ayhan, M. S., Berens, P., & Wahl, S. (2017). Leveraging uncertainty information from deep neural networks for disease detection. *Scientific reports*, *7*, 1–14.

Li, X., & Orabona, F. (2019). On the convergence of stochastic gradient descent with adaptive stepsizes. *The 22nd international conference on artificial intelligence and statistics*, 983–992.

Li, Y., & Gal, Y. (2017). Dropout inference in Bayesian neural networks with alpha-divergences. *International conference on machine learning*, 2052–2061.

Liang, T., Poggio, T., Rakhlin, A., & Stokes, J. (2019). Fisher-rao metric, geometry, and complexity of neural networks. *The 22nd international conference on artificial intelligence and statistics*, 888–896.

Liao, J., & Berg, A. (2019). Sharpening Jensen's inequality. *The American Statistician*, *73*(3), 278–281.

Lin, J. A., Antorán, J., Padhy, S., Janz, D., Hernández-Lobato, J. M., & Terenin, A. (2024). Sampling from Gaussian process posteriors using stochastic gradient descent. *Advances in Neural Information Processing Systems*, *36*.

Liu, J. Z., Padhy, S., Ren, J., Lin, Z., Wen, Y., Jerfel, G., Nado, Z., Snoek, J., Tran, D., & Lakshminarayanan, B. (2023). A simple approach to improve single-model deep uncertainty via distance-awareness. *Journal of Machine Learning Research*, *24*, 1–63.

Liu, Y., & Yao, X. (1999). Ensemble learning via negative correlation. *Neural networks*, *12*(10), 1399–1404.

Livni, R., Shalev-Shwartz, S., & Shamir, O. (2014). On the computational efficiency of training neural networks. *Advances in neural information processing systems*, *27*.

Lloyd, S. (1982). Least squares quantization in pcm. *IEEE transactions on information theory*, *28*(2), 129–137.

Lu, Z., Wu, X., Zhu, X., & Bongard, J. (2010). Ensemble pruning via individual contribution ordering. *Proceedings of the 16th ACM SIGKDD international conference on Knowledge discovery and data mining*, 871–880.

Lyle, C., van der Wilk, M., Kwiatkowska, M., Gal, Y., & Bloem-Reddy, B. (2020). On the benefits of invariance in neural networks. *arXiv preprint arXiv:2005.00178*.

Ma, C., & Hernández-Lobato, J. M. (2021). Functional variational inference based on stochastic process generators. *Advances in Neural Information Processing Systems*, *34*, 21795–21807.

Ma, C., Li, Y., & Hernández-Lobato, J. M. (2019). Variational implicit processes. *International Conference on Machine Learning (ICML)*, 4222–4233.

MacKay, D. J. C. (1992a). Bayesian interpolation. *Neural Computation*, *4*(3), 415–447. https://doi.org/10.1162/neco.1992.4.3.415

MacKay, D. J. C. (1992b). The evidence framework applied to classification networks. *Neural computation*, *4*(5), 720–736.

MacKay, D. J. C. (1992c). A practical Bayesian framework for backpropagation networks. *Neural computation*, *4*, 448–472.

MacKay, D. J. C. (2003). *Information theory, inference, and learning algorithms*. Cambridge University Press.

Maddox, W. J., Izmailov, P., Garipov, T., Vetrov, D. P., & Wilson, A. G. (2019). A simple baseline for Bayesian uncertainty in deep learning. *Advances in Neural Information Processing Systems*, *32*, 13153–13164.

Maintainers, T., & Contributors. (2016). Torchvision: Pytorch's computer vision library.

Martens, J. (2020). New insights and perspectives on the natural gradient method. *The Journal of Machine Learning Research*, *21*, 5776–5851.

Martens, J., & Grosse, R. (2015). Optimizing neural networks with kronecker-factored approximate curvature. *International Conference on Machine Learning*, 2408–2417.

Masegosa, A. R. (2020). Learning under model misspecification: Applications to variational and ensemble methods. *Advances in Neural Information Processing Systems*.

Masegosa, A. R., Lorenzen, S., Igel, C., & Seldin, Y. (2020). Second order pac-Bayesian bounds for the weighted majority vote. *Advances in Neural Information Processing Systems*, *33*, 5263–5273.

McAllester, D. A. (1998). Some PAC-Bayesian theorems. *Proceedings of the Conference on Learning Theory (COLT)*.

McAllester, D. A. (1999). PAC-Bayesian model averaging. *Proceedings of the twelfth annual conference on Computational learning theory*, 164–170.

Mercer, J. (1909). Functions of positive and negative type, and their connection with the theory of integral equations. *Philosophical Transactions of the Royal Society A*, *209*, 415–446.

Minka, T. P. [Thomas P.]. (2001). *Expectation propagation for approximate Bayesian inference* (tech. rep. No. MSR–TR–2001–41). Microsoft Research. Redmond, WA.

Minka, T. P. [Thomas P]. (2005). *Divergence measures and message passing* (tech. rep. No. MSR-TR-2005-173).

Mohri, M., Rostamizadeh, A., & Talwalkar, A. (2018). *Foundations of machine learning*. MIT press.

Mucsányi, B., Kirchhof, M., & Oh, S. J. (2024). Benchmarking uncertainty disentanglement: Specialized uncertainties for specialized tasks. *ICML 2024 Workshop on Structured Probabilistic Inference & Generative ModelingL*.

Murphy, K. P. (2012). *Machine learning: A probabilistic perspective*. MIT Press.

Nagarajan, V. (2021). Explaining generalization in deep learning: Progress and fundamental limits. *arXiv preprint arXiv:2110.08922*.

Nagarajan, V., & Kolter, J. Z. (2017). Generalization in deep networks: The role of distance from initialization. *NeurIPS Workshop on Deep Learning: Bridging Theory and Practice*.

Nagarajan, V., & Kolter, J. Z. (2019a). Deterministic PAC-Bayesian generalization bounds for deep networks via generalizing noise-resilience. *International Conference on Learning Representations*.

Nagarajan, V., & Kolter, J. Z. (2019b). Uniform convergence may be unable to explain generalization in deep learning. *Advances in Neural Information Processing Systems*, 11611–11622.

Nakkiran, P., Kaplun, G., Bansal, Y., Yang, T., Barak, B., & Sutskever, I. (2020). Deep double descent: Where bigger models and more data hurt [Originally released in 2019; key empirical study of the double-descent phenomenon]. *arXiv preprint arXiv:1912.02292*.

Neal, R. M. (1993). *Probabilistic inference using markov chain monte carlo methods* (tech. rep. No. CRG–TR–93–1). Department of Computer Science, University of Toronto. Toronto.

Neal, R. M. (2012). *Bayesian learning for neural networks* (Vol. 118). Springer Science & Business Media.

Negrea, J., Dziugaite, G. K., & Roy, D. (2020). In defense of uniform convergence: Generalization via derandomization with an application to interpolating predictors. *International Conference on Machine Learning*, 7263–7272.

Nesterov, Y. (2003). *Introductory lectures on convex optimization: A basic course*. Springer.

Netzer, Y., Wang, T., Coates, A., Bissacco, A., Wu, B., & Ng, A. Y. (2011). Reading digits in natural images with unsupervised feature learning. *NIPS Workshop on Deep Learning and Unsupervised Feature Learning*. http://ufldl.stanford.edu/housenumbers

Neyshabur, B., Bhojanapalli, S., McAllester, D., & Srebro, N. (2017a). Exploring generalization in deep learning. *Advances in neural information processing systems*, *30*.

Neyshabur, B., Bhojanapalli, S., & Srebro, N. (2017b). A PAC-Bayesian approach to spectrally-normalized margin bounds for neural networks. *International Conference on Learning Representations*.

Neyshabur, B., Li, Z., Bhojanapalli, S., LeCun, Y., & Srebro, N. (2019). The role of overparametrization in generalization of neural networks. *International Conference on Learning Representations*. https://openreview.net/forum?id=BygfghAcYX

Neyshabur, B., Salakhutdinov, R. R., & Srebro, N. (2015a). Path-sgd: Path-normalized optimization in deep neural networks. *Advances in neural information processing systems*, *28*.

Neyshabur, B., Tomioka, R., & Srebro, N. (2015b). In search of the real inductive bias: On the role of implicit regularization in deep learning. *arXiv preprint arXiv:1412.6614*.

Neyshabur, B., Tomioka, R., & Srebro, N. (2015c). Norm-based capacity control in neural networks. *Proceedings of the 28th International Conference on Learning Theory (COLT)*, 1376–1401.

Nocedal, J., & Wright, S. J. (2006). *Numerical optimization* (2nd). Springer.

Novak, R., Sohl-Dickstein, J., & Schoenholz, S. S. (2022). Fast finite width neural tangent kernel. *International Conference on Machine Learning*, 17018–17044.

Novikov, A., & Izmailov, P. (2018). Tensor train kernel trick. *Neural Networks*, *104*, 1–19.

Ortega, L. A. [Luis A.], Rodriguez-Santana, S., & Hernández-Lobato, D. (2024a). Variational linearized Laplace approximation for Bayesian deep learning. *International Conference on Machine Learning*, 38815–38836.

Ortega, L. A. [Luis A.], Rodriguez-Santana, S., & Hernández-Lobato, D. (2023). Deep variational implicit processes. *International Conference of Learning Representations*.

Ortega, L. A. [Luis A], Rodríguez-Santana, S., & Hernández-Lobato, D. (2024b). Fixed-mean Gaussian processes for post-hoc Bayesian deep learning. *arXiv preprint arXiv:2412.04177*.

Ovadia, Y., Fertig, E., Ren, J., Nado, Z., Sculley, D., Nowozin, S., Dillon, J., Lakshminarayanan, B., & Snoek, J. (2019). Can you trust your model's uncertainty? evaluating predictive uncertainty under dataset shift? *Advances in Neural Information Processing Systems*, 13969–13980.

Pang, T., Xu, K., Du, C., Chen, N., & Zhu, J. (2019). Improving adversarial robustness via promoting ensemble diversity. *International Conference on Machine Learning*, 4970–4979.

Patel, A. B., Nguyen, M. T., & Baraniuk, R. (2016). A probabilistic framework for deep learning. *Advances in neural information processing systems*, *29*.

Poole, B., Lahiri, S., Raghu, M., Sohl-Dickstein, J., & Ganguli, S. (2016). Exponential expressivity in deep neural networks through transient chaos. *Advances in neural information processing systems*, *29*.

Puurula, A., Read, J., & Bifet, A. (2014). Kaggle lshtc4 winning solution. *arXiv preprint arXiv:1405.0546*.

Quiñonero-Candela, J., & Rasmussen, C. E. (2005). A unifying view of sparse approximate Gaussian process regression. *The Journal of Machine Learning Research*, *6*, 1939–1959.

Rahimi, A., & Recht, B. (2007). Random features for large-scale kernel machines. *Advances in Neural Information Processing Systems 20*, 1177–1184.

Rasmussen, C. E., & Williams, C. K. I. (2006). *Gaussian processes for machine learning*. MIT Press.

Reed, M., & Simon, B. (1980). *Methods of modern mathematical physics. vol. i: Functional analysis*. Academic Press.

Rényi, A. (1961). On measures of entropy and information. *Proceedings of the Fourth Berkeley Symposium on Mathematical Statistics and Probability, Volume 1: Contributions to the Theory of Statistics*, *4*, 547–562.

Ritter, H., Botev, A., & Barber, D. (2018). A scalable Laplace approximation for neural networks. *International Conference on Learning Representations*, *6*.

Rockafellar, R. T. (1970). *Convex analysis*. Princeton University Press. https://doi.org/doi:10.1515/9781400873173

Rodríguez-Santana, S., & Hernández-Lobato, D. (2022). Adversarial $\alpha$-divergence minimization for Bayesian approximate inference. *Neurocomputing*, *471*, 260–274.

Roli, F., Giacinto, G., & Vernazza, G. (2001). Methods for designing multiple classifier systems. *International Workshop on Multiple Classifier Systems*, 78–87.

Rubin, D., & Stein, M. (2016). Spatially adaptive Bayesian covariance tapering. *International Conference on Artificial Intelligence and Statistics (AISTATS)*, 650–658.

Ruddigkeit, L., Van Deursen, R., Blum, L. C., & Reymond, J.-L. (2012). Enumeration of 166 billion organic small molecules in the chemical universe database gdb-17. *Journal of chemical information and modeling*, *52*(11), 2864–2875.

Rudin, W. (1991). *Functional analysis* (2nd). McGraw–Hill.

Russakovsky, O., Deng, J., Su, H., Krause, J., Satheesh, S., Ma, S., Huang, Z., Karpathy, A., Khosla, A., Bernstein, M., Berg, A. C., & Fei-Fei, L. (2015). ImageNet Large Scale Visual Recognition Challenge. *International Journal of Computer Vision (IJCV)*, *115*(3), 211–252. https://doi.org/10.1007/s11263-015-0816-y

Salimbeni, H., & Deisenroth, M. (2017). Doubly stochastic variational inference for deep Gaussian processes. *Advances in Neural Information Processing Systems*, *30*.

Santana, S. R., Zaldivar, B., & Hernández-Lobato, D. (2021). Sparse implicit processes for approximate inference. *arXiv preprint arXiv:2110.07618*.

Scannell, A., Mereu, R., Chang, P., Tamir, E., Pajarinen, J., & Solin, A. (2024). Function-space parameterization of neural networks for sequential learning. *International Conference on Learning Representations*.

Schoenholz, S. S., Gilmer, J., Ganguli, S., & Sohl-Dickstein, J. (2017). Deep information propagation. *International Conference on Learning Representations*. https://openreview.net/forum?id=H1W1UN9gg

Seeger, M. (2002). PAC-Bayesian generalisation error bounds for gaussian process classification. *Journal of Machine Learning Research*, *3*, 233–269.

Seeger, M. (2003). Fast forward selection to speed up sparse Gaussian process regression. *Proceedings of the 9th International Workshop on Artificial Intelligence and Statistics*.

Shalev-Shwartz, S., & Ben-David, S. (2014). *Understanding machine learning: From theory to algorithms*. Cambridge University Press.

Shawe-Taylor, J., Bartlett, P. L., Williamson, R. C., & Anthony, M. (1998). Structural risk minimization over data-dependent hierarchies. *IEEE transactions on Information Theory*, *44*(5), 1926–1940.

Shi, J., Sun, S., & Zhu, J. (2018). A spectral approach to gradient estimation for implicit distributions. *International Conference on Machine Learning*, 4644–4653.

Shorten, C., & Khoshgoftaar, T. M. (2019). A survey on image data augmentation for deep learning. *Journal of big data*, *6*(1), 1–48.

Smola, A. J., & Schölkopf, B. (2000). Sparse greedy matrix approximation for machine learning. *Proceedings of the 17th International Conference on Machine Learning*, 911–918.

Snelson, E., & Ghahramani, Z. (2006). Sparse Gaussian processes using pseudo-inputs. *Advances in Neural Information Processing Systems 18*, 1257–1264.

Snoek, J., Larochelle, H., & Adams, R. P. (2012). Practical Bayesian optimization of machine learning algorithms. *Advances in Neural Information Processing Systems 25*, 2951–2959.

Snoek, J., Ovadia, Y., Fertig, E., Lakshminarayanan, B., Nowozin, S., Sculley, D., Dillon, J., Ren, J., & Nado, Z. (2019). Can you trust your model's uncertainty? Evaluating predictive uncertainty under dataset shift. *Advances in Neural Information Processing Systems*, 13969–13980.

Sokolic, J., Giryes, R., Sapiro, G., & Rodrigues, M. (2017). Generalization error of invariant classifiers. *Artificial Intelligence and Statistics*, 1094–1103.

Soudry, D., Hoffer, E., Nacson, M. S., Gunasekar, S., & Srebro, N. (2018). The implicit bias of gradient descent on separable data. *The Journal of Machine Learning Research*, *19*(1), 2822–2878.

Stallkamp, J., Schlipsing, M., Salmen, J., & Igel, C. (2012). Man vs. computer: Benchmarking machine learning algorithms for traffic sign recognition. *Neural networks*, *32*, 323–332.

Stein, M. L. (1999). *Interpolation of spatial data: Some theory for kriging*. Springer.

Stroock, D. W. (2010). *Probability theory: An analytic view*. Cambridge university press.

Sun, S., Zhang, G., Shi, J., & Grosse, R. (2019). Functional variational Bayesian neural networks. *International Conference on Learning Representations*.

Szegedy, C., Vanhoucke, V., Ioffe, S., Shlens, J., & Wojna, Z. (2016). Rethinking the inception architecture for computer vision. *Proceedings of the IEEE conference on computer vision and pattern recognition*, 2818–2826.

Tang, E. K., Suganthan, P. N., & Yao, X. (2006). An analysis of diversity measures. *Machine learning*, *65*(1), 247–271.

Tipping, M. E. (2001). Sparse Bayesian learning and the relevance vector machine. *Journal of Machine Learning Research*, *1*, 211–244.

Tishby, N., & Zaslavsky, N. (2015). Deep learning and the information bottleneck principle. *2015 ieee information theory workshop (itw)*, 1–5.

Titsias, M. (2009). Variational learning of inducing variables in sparse Gaussian processes. *Artificial intelligence and statistics*, 567–574.

Touchette, H. (2009). The large deviation approach to statistical mechanics. *Physics Reports*, *478*(1-3), 1–69.

Van Neerven, J., et al. (2010). $\gamma$-radonifying operators—a survey. *The AMSI-ANU workshop on spectral theory and harmonic analysis*, *44*, 1–61.

Vapnik, V. N. (1998). *Statistical learning theory*. Wiley.

Vapnik, V. N., & Chervonenkis, A. Y. (2015). On the uniform convergence of relative frequencies of events to their probabilities. In *Measures of complexity: Festschrift for alexey chervonenkis* (pp. 11–30). Springer.

Vaswani, A., Shazeer, N., Parmar, N., Uszkoreit, J., Jones, L., Gomez, A. N., Kaiser, Ł., & Polosukhin, I. (2017). Attention is All you need. *Advances in Neural Information Processing Systems*, 5998–6008.

Vidal, R., Bruna, J., Giryes, R., & Soatto, S. (2020). Mathematics of deep learning. *Mexican Conference on Pattern Recognition*.

Villacampa-Calvo, C., & Hernández-Lobato, D. (2020). Alpha divergence minimization in multi-class Gaussian process classification. *Neurocomputing*, *378*, 210–227.

Wang, J., Liu, Z., Wu, Y., & Yuan, J. (2012). Mining actionlet ensemble for action recognition with depth cameras. *2012 IEEE Conference on Computer Vision and Pattern Recognition*, 1290–1297.

Wang, Y., Sonthalia, R., & Hu, W. (2024). Near-interpolators: Rapid norm growth and the trade-off between interpolation and generalization. *International Conference on Artificial Intelligence and Statistics*, 4483–4491.

Wen, Y., Tran, D., & Ba, J. (2019). Batchensemble: An alternative approach to efficient ensemble and lifelong learning. *International Conference on Learning Representations*.

Wen, Y., Vicol, P., Ba, J., Tran, D., & Grosse, R. (2018). Flipout: Efficient pseudo-independent weight perturbations on mini-batches. *International Conference on Learning Representations*. https://openreview.net/forum?id=rJNpifWAb

Wenzel, F., Roth, K., Veeling, B., Swiatkowski, J., Tran, L., Mandt, S., Snoek, J., Salimans, T., Jenatton, R., & Nowozin, S. (2020a). How good is the Bayes posterior in deep neural networks really? *International Conference on Machine Learning*, 10248–10259.

Wenzel, F., Snoek, J., Tran, D., & Jenatton, R. (2020b). Hyperparameter ensembles for robustness and uncertainty quantification. *arXiv preprint arXiv:2006.13570*.

Wilson, A. G., & Nickisch, H. (2015). Kernel interpolation for scalable structured Gaussian processes (kiss-gp). *International Conference on Machine Learning (ICML)*, 1775–1784.

Ykhlef, H., & Bouchaffra, D. (2017). An efficient ensemble pruning approach based on simple coalitional games. *Information Fusion*, *34*, 28–42.

Yu, H., Chen, Y., Low, B. K. H., Jaillet, P., & Dai, Z. (2019). Implicit posterior variational inference for deep Gaussian processes. *Advances in Neural Information Processing Systems*, *32*, 14475–14486.

Yu, Y., Li, Y.-F., & Zhou, Z.-H. (2011). Diversity regularized machine. *Twenty-Second International Joint Conference on Artificial Intelligence*.

Zhang, C., Bengio, S., Hardt, M., Recht, B., & Vinyals, O. (2017). Understanding deep learning requires rethinking generalization. *Proceedings of the International Conference on Learning Representations (ICLR)*.

Zhang, R., Li, C., Zhang, J., Chen, C., & Wilson, A. G. (2019). Cyclical stochastic gradient mcmc for Bayesian deep learning. *International Conference on Learning Representations*.

Zhang, T. (2006). Information-theoretic upper and lower bounds for statistical estimation. *IEEE Transactions on Information Theory*, *52*(4), 1307–1321.

Zhou, W., Veitch, V., Austern, M., Adams, R. P., & Orbanz, P. (2019). Non-vacuous generalization bounds at the imagenet scale: A PAC-Bayesian compression approach. *International Conference on Learning Representations*. https://openreview.net/forum?id=BJgqqsAct7

Zhou, X., Xie, L., Zhang, P., & Zhang, Y. (2014). An ensemble of deep neural networks for object tracking. *2014 IEEE International Conference on Image Processing (ICIP)*, 843–847.

Zhou, Z.-H. (2012). *Ensemble methods: Foundations and algorithms*. CRC press.

Zhou, Z.-H., & Li, N. (2010). Multi-information ensemble diversity. *International Workshop on Multiple Classifier Systems*, 134–144.

Zhu, S., & Rohwer, R. (1995). Information geometry and prior construction. *Entropy*, *1*, 3–22.

Zou, D., Wu, J., Braverman, V., Gu, Q., Foster, D. P., & Kakade, S. (2021). The benefits of implicit regularization from sgd in least squares problems. *Advances in Neural Information Processing Systems*.

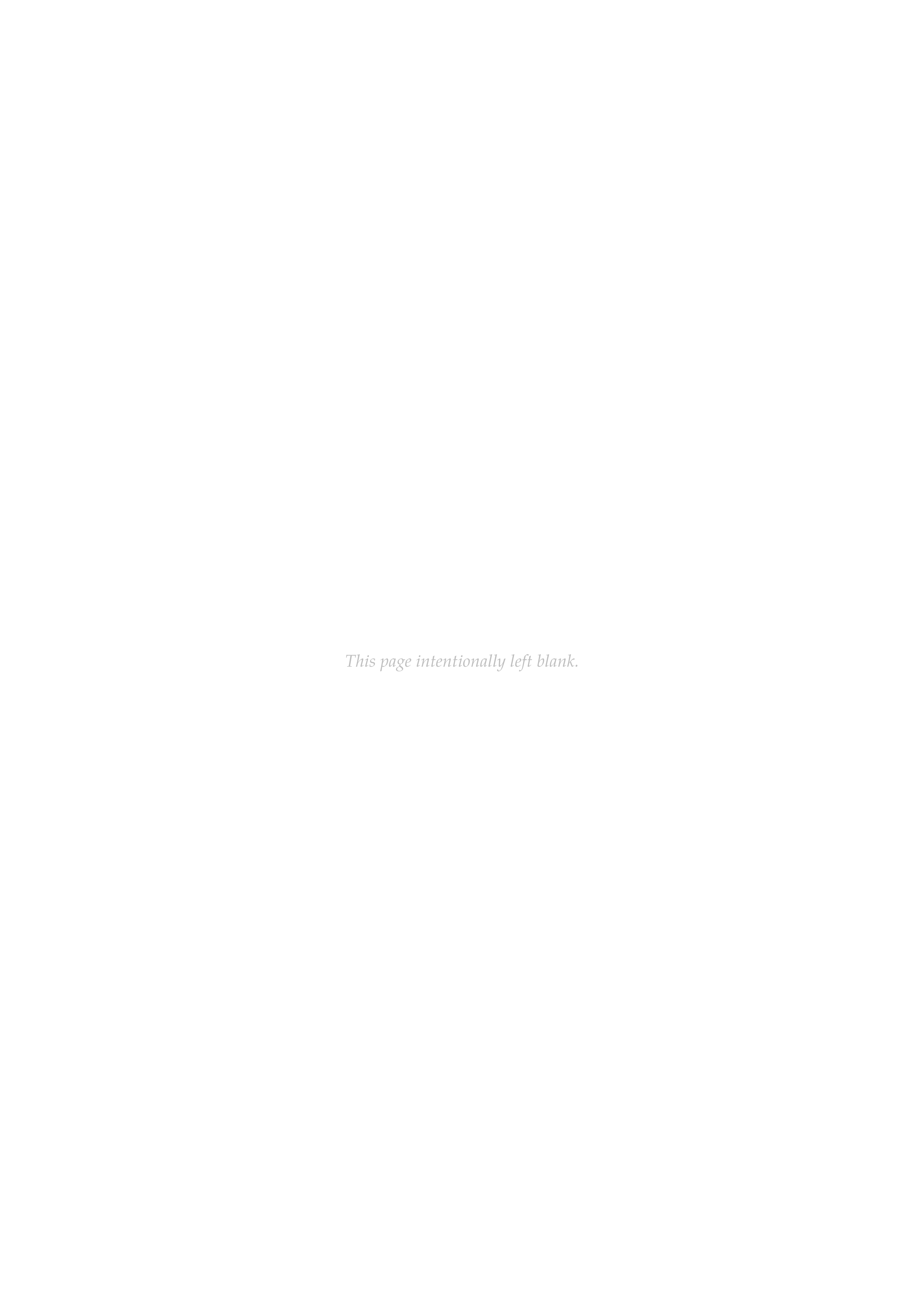

This page intentionally left blank.

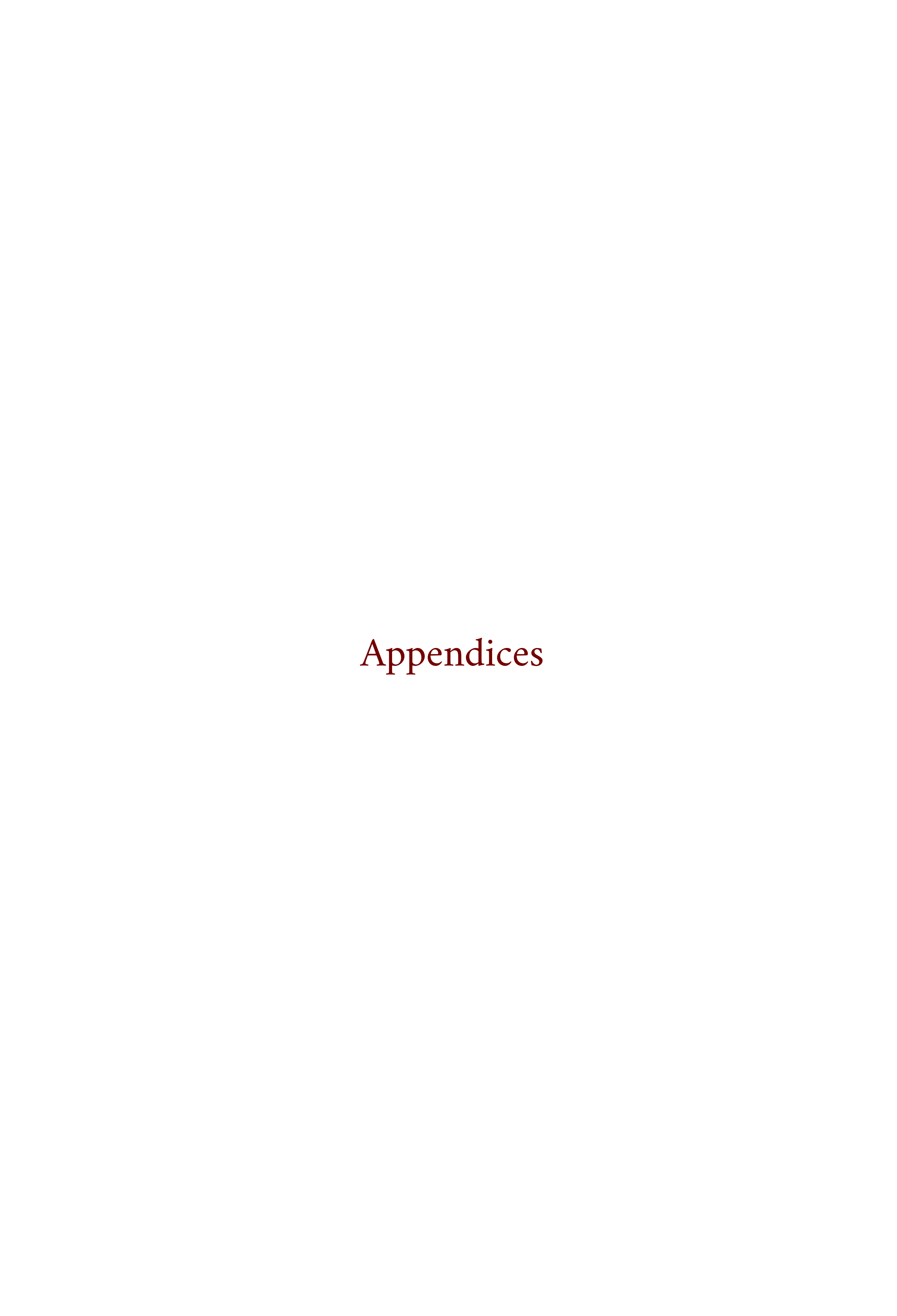

# Appendices

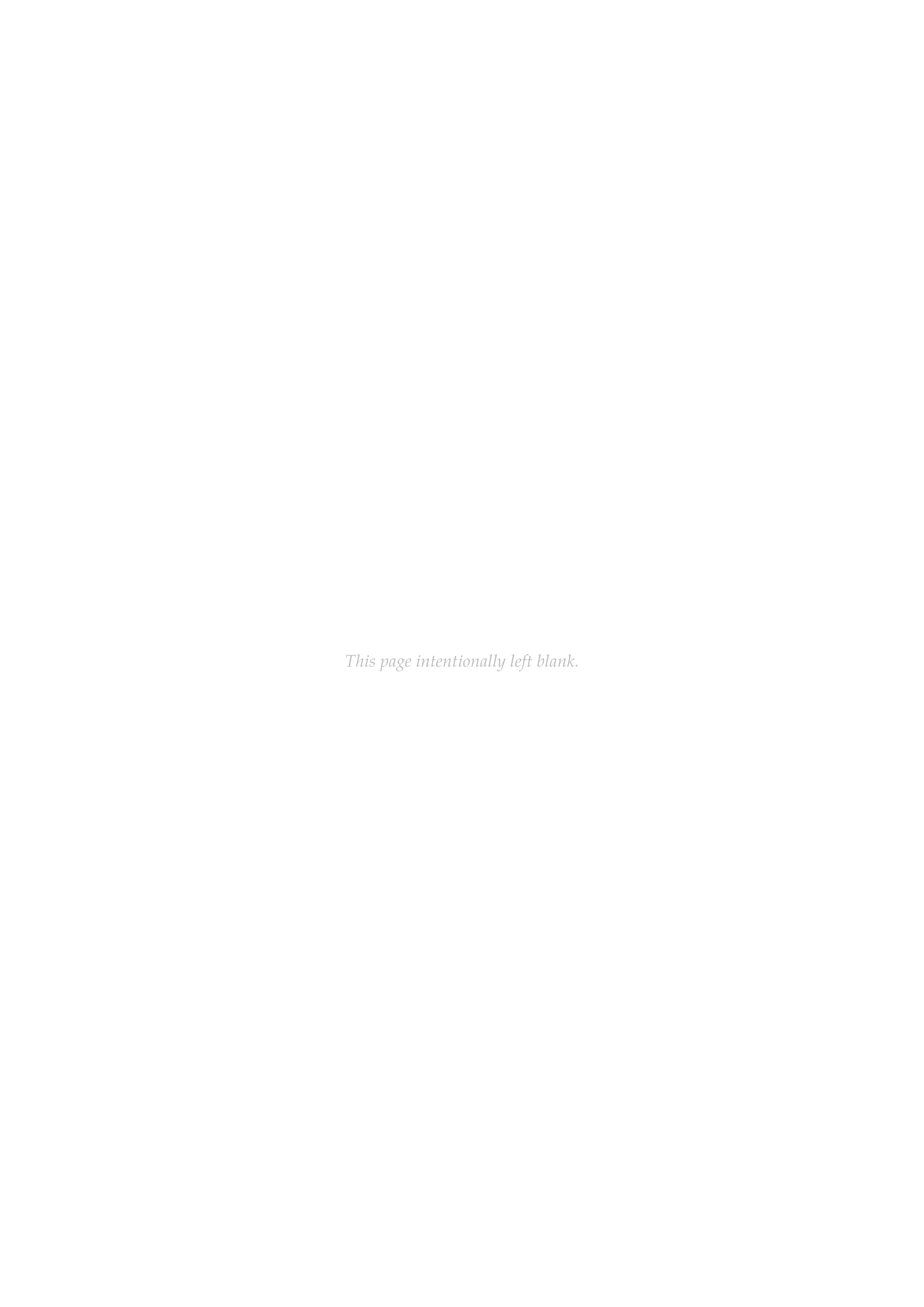
This page intentionally left blank.

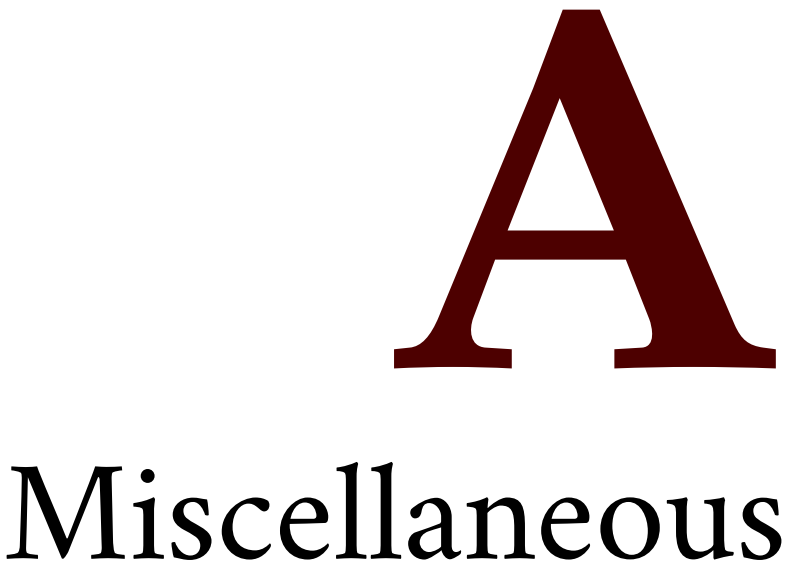

# Miscellaneous

## A.1 Synthetic Experiments For the Capacity Landscape

This appendix derives the closed-form formulas used for Figure 1.3.

### A.1.1 Classical Bias–Variance Curve

In each Monte Carlo replicate, draw training inputs $x_i \overset{\text{iid}}{\sim} \mathcal{U}(0,1)$ and corresponding targets $t_i = \sin(2\pi x_i) + \varepsilon_i, \varepsilon_i \sim \mathcal{N}(0, \sigma^2), \sigma = 0.1$. Use $N_{\text{train}} = 40$ samples for training and evaluate performance on $N_{\text{test}} = 1\,000$ fresh points drawn from the same distribution.

For each replicate, fit ordinary least-squares polynomials of degree $m = 0, \dots, 10$. The test mean squared error (MSE) of degree $m$ is given by

$$\frac{1}{N_{\text{test}}} \sum_{j=1}^{N_{\text{test}}} \left(y_m(x_j) - t_j^{\text{test}}\right)^2, \tag{A.1}$$

where $y_m$ denotes the fitted polynomial of degree $m$. The curve shown in the figure is obtained by averaging this quantity over $1\,000$ independent replicates, yielding a smooth estimate of the expected risk.

### A.1.2 Double-descent Experiment

The purpose of this experiment is to illustrate the *double-descent phenomenon* in linear regression, where the test error first decreases, then increases sharply around the interpolation threshold, and finally decreases again as the model becomes increasingly overparameterized. Fix the number of training samples to

$$N_{\text{train}} = 50. \tag{A.2}$$

The data is generated from a low-dimensional linear model

$$y = X_{1:d_0} w_\star + \varepsilon, \tag{A.3}$$

where

- $d_0 = 10$ is the true intrinsic dimensionality,
- $w_\star \sim \mathcal{N}(0, I)$ are the ground-truth regression weights,
- $\varepsilon \sim \mathcal{N}(0, \sigma^2)$ with $\sigma^2 = 2$ is additive Gaussian noise.

The design matrix

$$X \in \mathbb{R}^{N_{\text{train}} \times p} \tag{A.4}$$

has entries $X_{ij} \sim \mathcal{N}(0, 1)$. The feature dimension $p$ is varied from $p = 1$ to $p = 100$. Within a single run, the same training set $(X, y)$ is reused across all values of $p$, progressively increasing the number of available features. For each $p$, compute the minimum-norm least-squares estimator

$$\hat{w} = X^\top (X X^\top)^{-1} y, \tag{A.5}$$

which coincides with the standard least-squares solution when $p < N_{\text{train}}$, and interpolates the training data exactly when $p \geq N_{\text{train}}$.

Generalization performance is assessed using the mean squared error (MSE) on 10 000 fresh test points, sampled independently from the same distribution as the training data. To reduce variance and highlight the characteristic shape of the error curve, results are averaged over 200 independent repetitions of the experiment, each with newly drawn $(X, w_\star, \varepsilon)$.

## A.2 Kullback–Leibler Divergence Between Multivariate Gaussians

In this appendix, the Kullback–Leibler (KL) divergence between two multivariate Gaussian distributions is derived. This quantity admits a closed-form expression, which will prove useful in subsequent sections. Let $P$ and $Q$ denote two $k$-dimensional Gaussian distributions over the same random variable $\boldsymbol{a}$, defined as

$$P(\boldsymbol{a}) = \mathcal{N}(\boldsymbol{a} \,|\, \boldsymbol{\mu}_1, \boldsymbol{\Sigma}_1), \quad Q(\boldsymbol{a}) = \mathcal{N}(\boldsymbol{a} \,|\, \boldsymbol{\mu}_2, \boldsymbol{\Sigma}_2). \tag{A.6}$$

The KL divergence from $P$ to $Q$ is defined as

$$\mathrm{KL}(P \,|\, Q) = \mathbb{E}_{P(\boldsymbol{a})} \left[ \log P(\boldsymbol{a}) - \log Q(\boldsymbol{a}) \right]. \tag{A.7}$$

The log-density of a multivariate Gaussian distribution can be expressed as

$$\begin{aligned} \log P(\boldsymbol{a}) &= -\frac{k}{2} \log(2\pi) - \frac{1}{2} \log\det(\boldsymbol{\Sigma}_1) - \frac{1}{2} (\boldsymbol{a} - \boldsymbol{\mu}_1)^\top \boldsymbol{\Sigma}_1^{-1} (\boldsymbol{a} - \boldsymbol{\mu}_1) \\ &= -\frac{k}{2} \log(2\pi) - \frac{1}{2} \log\det(\boldsymbol{\Sigma}_1) - \frac{1}{2} \operatorname{tr}\left[ \boldsymbol{\Sigma}_1^{-1} (\boldsymbol{a} - \boldsymbol{\mu}_1)(\boldsymbol{a} - \boldsymbol{\mu}_1)^\top \right]. \end{aligned} \tag{A.8}$$

Since the trace operator is linear, its expectation under $P$ simplifies as follows:

$$\begin{aligned}\mathbb{E}_{P(\boldsymbol{a})}\left[\operatorname{tr}\left(\boldsymbol{\Sigma}_1^{-1}(\boldsymbol{a}-\boldsymbol{\mu}_1)(\boldsymbol{a}-\boldsymbol{\mu}_1)^\top\right)\right] &= \operatorname{tr}\left(\boldsymbol{\Sigma}_1^{-1}\mathbb{E}_{P(\boldsymbol{a})}\left[(\boldsymbol{a}-\boldsymbol{\mu}_1)(\boldsymbol{a}-\boldsymbol{\mu}_1)^\top\right]\right) \\ &= \operatorname{tr}\left(\boldsymbol{\Sigma}_1^{-1}\boldsymbol{\Sigma}_1\right) = k.\end{aligned} \tag{A.9}$$

The corresponding quadratic term in the expression for $\log Q(\boldsymbol{a})$. Then,

$$\begin{aligned}\mathbb{E}_{P(\boldsymbol{a})}\left[\operatorname{tr}\left(\boldsymbol{\Sigma}_2^{-1}(\boldsymbol{a}-\boldsymbol{\mu}_2)(\boldsymbol{a}-\boldsymbol{\mu}_2)^\top\right)\right] &= \operatorname{tr}\left(\boldsymbol{\Sigma}_2^{-1}\mathbb{E}_{P(\boldsymbol{a})}\left[(\boldsymbol{a}-\boldsymbol{\mu}_2)(\boldsymbol{a}-\boldsymbol{\mu}_2)^\top\right]\right) \\ &= \operatorname{tr}\left(\boldsymbol{\Sigma}_2^{-1}\left(\boldsymbol{\Sigma}_1+(\boldsymbol{\mu}_1-\boldsymbol{\mu}_2)(\boldsymbol{\mu}_1-\boldsymbol{\mu}_2)^\top\right)\right) \\ &= \operatorname{tr}(\boldsymbol{\Sigma}_2^{-1}\boldsymbol{\Sigma}_1)+(\boldsymbol{\mu}_1-\boldsymbol{\mu}_2)^\top\boldsymbol{\Sigma}_2^{-1}(\boldsymbol{\mu}_1-\boldsymbol{\mu}_2).\end{aligned} \tag{A.10}$$

Combining the above components, the KL divergence becomes

$$\mathrm{KL}(P\,|\,Q) = \frac{1}{2}\left[\log\frac{\det(\boldsymbol{\Sigma}_2)}{\det(\boldsymbol{\Sigma}_1)} - k + \operatorname{tr}(\boldsymbol{\Sigma}_2^{-1}\boldsymbol{\Sigma}_1) + (\boldsymbol{\mu}_1-\boldsymbol{\mu}_2)^\top\boldsymbol{\Sigma}_2^{-1}(\boldsymbol{\mu}_1-\boldsymbol{\mu}_2)\right]. \tag{A.11}$$

In the special case where $\boldsymbol{\mu}_2 = \mathbf{0}$ and $\boldsymbol{\Sigma}_2 = \boldsymbol{I}$, the expression simplifies further to

$$\mathrm{KL}(P\,|\,\mathcal{N}(\mathbf{0},\boldsymbol{I})) = \frac{1}{2}\left[-\log\det(\boldsymbol{\Sigma}_1) - k + \operatorname{tr}(\boldsymbol{\Sigma}_1) + \boldsymbol{\mu}_1^\top\boldsymbol{\mu}_1\right]. \tag{A.12}$$

## A.3 Gaussian Expectation of Powered Likelihoods

In this section, the closed-form expression for the expectation of a powered Gaussian likelihood under a Gaussian distribution is derived; which is relevant for evaluating generalized divergences such as the $\alpha$-divergence. Specifically, let $Q(f) = \mathcal{N}(\mu, \Sigma)$ be a Gaussian distribution over the latent variable $f$, and let the likelihood be Gaussian as well, $P(y|f) = \mathcal{N}(y|f, \sigma^2)$. Consider the case where the likelihood is raised to a power $\alpha \in (0, 1)$.

Begin by observing that raising a Gaussian density to the power $\alpha$ yields an unnormalized Gaussian:

$$P(y|f)^\alpha = \left(\frac{1}{\sqrt{2\pi}\sigma}\exp\left\{-\frac{1}{2}\frac{(y-f)^2}{\sigma^2}\right\}\right)^\alpha = \frac{1}{(2\pi\sigma^2)^{\alpha/2}}\exp\left\{-\frac{1}{2}\frac{(y-f)^2}{\sigma^2/\alpha}\right\}. \tag{A.13}$$

This expression corresponds to the density of a Gaussian distribution with variance $\sigma^2/\alpha$, up to a normalization constant. That is,

$$P(y|f)^\alpha = \left(\frac{2\pi\sigma^2}{\alpha}\right)^{1/2}\cdot\frac{1}{(2\pi\sigma^2)^{\alpha/2}}\cdot\mathcal{N}(y|f,\sigma^2/\alpha). \tag{A.14}$$

Compute the expectation of this unnormalized expression under the distribution $Q(f)$:

$$\mathbb{E}_{Q(f)}\left[P(y|f)^\alpha\right] = \left(\frac{2\pi\sigma^2}{\alpha}\right)^{1/2}\cdot\frac{1}{(2\pi\sigma^2)^{\alpha/2}}\cdot\mathbb{E}_{Q(f)}\left[\mathcal{N}(y|f,\sigma^2/\alpha)\right]. \tag{A.15}$$

The remaining expectation involves the convolution of two Gaussian densities. Since the convolution of Gaussians is Gaussian:

$$\mathbb{E}_{Q(f)}\left[\mathcal{N}(y|f,\sigma^2/\alpha)\right] = \mathcal{N}(y|\mu,\Sigma+\sigma^2/\alpha). \tag{A.16}$$

Hence, the final expression becomes

$$\mathbb{E}_{Q(f)}\left[P(y|f)^{\alpha}\right] = \left(\frac{2\pi\sigma^2}{\alpha}\right)^{1/2} \cdot \frac{1}{(2\pi\sigma^2)^{\alpha/2}} \cdot \mathcal{N}(y|\mu,\Sigma+\sigma^2/\alpha). \tag{A.17}$$

# B
# Deep Variational Implicit Processes Full Regression Results

**Table B.1:** *Experiment results on Deep Variational Implicit Processes (Chapter 3). Log Likelihood, Root Mean Squared Error and Continuous Ranked Probability Score results on regression UCI benchmark datasets. Eight datasets are considered: Boston, Energy, Concrete, WineRed, Power, Naval, Protein, and Kin8nm. Results are averaged over 20 random train-test splits. Best value is highlighted in* ***purple*** *and second-to-best in* ***teal****.*

| NLL | Single-layer | | | DVIP | | | | Salimbeni | | | |
|---|---|---|---|---|---|---|---|---|---|---|---|
| | SGP | VIP | VIP 200 | DVIP 2 | DVIP 3 | DVIP 4 | DVIP 5 | DGP 2 | DGP 3 | DGP 4 | DGP 5 |
| Boston | **2.62 ± 0.05** | 2.76 ± 0.05 | 2.69 ± 0.03 | 2.85 ± 0.09 | **2.59 ± 0.06** | 2.67 ± 0.09 | 2.66 ± 0.08 | 2.63 ± 0.05 | 2.63 ± 0.05 | 2.64 ± 0.05 | 2.65 ± 0.05 |
| Energy | 1.54 ± 0.02 | 2.07 ± 0.02 | 2.07 ± 0.02 | 0.76 ± 0.02 | **0.70 ± 0.01** | **0.70 ± 0.01** | 0.73 ± 0.01 | **0.72 ± 0.01** | 0.74 ± 0.01 | **0.72 ± 0.01** | 0.73 ± 0.01 |
| Concrete | 3.16 ± 0.01 | 3.45 ± 0.02 | 3.48 ± 0.01 | 3.24 ± 0.04 | 3.20 ± 0.05 | **3.03 ± 0.02** | **3.06 ± 0.02** | 3.17 ± 0.01 | 3.20 ± 0.01 | 3.13 ± 0.01 | 3.12 ± 0.01 |
| Winered | **0.93 ± 0.01** | **0.94 ± 0.01** | 0.96 ± 0.01 | **0.94 ± 0.01** | **0.94 ± 0.01** | **0.94 ± 0.01** | 0.95 ± 0.01 | **0.94 ± 0.01** | **0.94 ± 0.01** | **0.94 ± 0.01** | **0.93 ± 0.01** |
| Power | 2.84 ± 0.00 | 2.85 ± 0.00 | 2.86 ± 0.00 | 2.82 ± 0.01 | 2.81 ± 0.00 | **2.79 ± 0.01** | **2.79 ± 0.01** | 2.81 ± 0.01 | **2.80 ± 0.00** | **2.80 ± 0.00** | **2.80 ± 0.01** |
| Protein | 2.93 ± 0.00 | 3.03 ± 0.00 | 3.03 ± 0.00 | 2.93 ± 0.00 | 2.89 ± 0.00 | 2.88 ± 0.00 | 2.86 ± 0.00 | 2.84 ± 0.00 | **2.79 ± 0.00** | **2.79 ± 0.00** | **2.80 ± 0.00** |
| Naval | −6.11 ± 0.06 | −4.50 ± 0.02 | −4.31 ± 0.00 | −5.89 ± 0.02 | −5.98 ± 0.01 | −5.90 ± 0.01 | −5.92 ± 0.01 | **−6.35 ± 0.09** | −6.21 ± 0.04 | **−6.27 ± 0.06** | −6.21 ± 0.08 |
| Kin8nm | −0.91 ± 0.00 | −0.31 ± 0.00 | −0.25 ± 0.00 | −1.00 ± 0.00 | −1.13 ± 0.00 | −1.15 ± 0.00 | −1.16 ± 0.00 | −1.29 ± 0.00 | **−1.32 ± 0.00** | **−1.33 ± 0.00** | 1.30 ± 0.00 |

| RMSE | Single-layer | | | DVIP | | | | Salimbeni | | | |
|---|---|---|---|---|---|---|---|---|---|---|---|
| | SGP | VIP | VIP 200 | DVIP 2 | DVIP 3 | DVIP 4 | DVIP 5 | DGP 2 | DGP 3 | DGP 4 | DGP 5 |
| Boston | **3.48 ± 0.17** | 4.78 ± 0.28 | 4.49 ± 0.28 | 3.87 ± 0.19 | **3.50 ± 0.20** | 3.60 ± 0.19 | 3.66 ± 0.21 | 3.51 ± 0.18 | 3.53 ± 0.19 | 3.55 ± 0.20 | 3.56 ± 0.20 |
| Energy | 1.07 ± 0.03 | 2.57 ± 0.08 | 2.68 ± 0.07 | 0.52 ± 0.01 | **0.47 ± 0.01** | **0.46 ± 0.01** | **0.47 ± 0.01** | **0.46 ± 0.01** | **0.47 ± 0.01** | **0.46 ± 0.01** | **0.46 ± 0.01** |
| Concrete | 5.84 ± 0.12 | 7.75 ± 0.15 | 8.06 ± 0.16 | 6.01 ± 0.16 | 5.68 ± 0.18 | **5.13 ± 0.12** | **5.27 ± 0.13** | 5.86 ± 0.12 | 6.01 ± 0.12 | 5.54 ± 0.11 | 5.52 ± 0.12 |
| Winered | **0.61 ± 0.00** | **0.62 ± 0.00** | 0.63 ± 0.00 | **0.62 ± 0.00** | **0.62 ± 0.00** | **0.62 ± 0.00** | **0.62 ± 0.00** | **0.62 ± 0.00** | **0.62 ± 0.00** | **0.62 ± 0.00** | **0.62 ± 0.00** |
| Power | 4.15 ± 0.03 | 4.21 ± 0.03 | 4.22 ± 0.03 | 4.06 ± 0.04 | 4.01 ± 0.04 | 3.97 ± 0.04 | **3.95 ± 0.04** | 4.00 ± 0.04 | 3.98 ± 0.03 | 3.99 ± 0.03 | **3.96 ± 0.04** |
| Protein | 4.56 ± 0.01 | 5.05 ± 0.01 | 5.04 ± 0.01 | 4.54 ± 0.01 | 4.40 ± 0.01 | 4.33 ± 0.01 | 4.26 ± 0.01 | 4.17 ± 0.01 | **4.00 ± 0.01** | **4.01 ± 0.01** | 4.02 ± 0.01 |
| Naval | **0.00 ± 0.00** | **0.00 ± 0.00** | **0.00 ± 0.00** | **0.00 ± 0.00** | **0.00 ± 0.00** | **0.00 ± 0.00** | **0.00 ± 0.00** | **0.00 ± 0.00** | **0.00 ± 0.00** | **0.00 ± 0.00** | **0.00 ± 0.00** |
| Kin8nm | 0.09 ± 0.00 | 0.17 ± 0.00 | 0.18 ± 0.00 | 0.08 ± 0.00 | **0.07 ± 0.00** | **0.07 ± 0.00** | **0.07 ± 0.00** | **0.06 ± 0.00** | **0.06 ± 0.00** | **0.06 ± 0.00** | **0.06 ± 0.00** |

| CRPS | Single-layer | | | DVIP | | | | Salimbeni | | | |
|---|---|---|---|---|---|---|---|---|---|---|---|
| | SGP | VIP | VIP 200 | DVIP 2 | DVIP 3 | DVIP 4 | DVIP 5 | DGP 2 | DGP 3 | DGP 4 | DGP 5 |
| Boston | 1.79 ± 0.05 | 2.25 ± 0.08 | 2.13 ± 0.08 | 1.91 ± .06 | **1.76 ± 0.07** | 1.81 ± 0.07 | **1.78 ± 0.06** | 1.79 ± 0.05 | 1.80 ± 0.06 | 1.80 ± 0.06 | 1.81 ± 0.06 |
| Energy | 0.62 ± 0.01 | 1.27 ± 0.04 | 1.30 ± 0.03 | **0.28 ± 0.00** | **0.26 ± 0.00** | **0.26 ± 0.00** | **0.26 ± 0.00** | **0.26 ± 0.00** | **0.26 ± 0.00** | **0.26 ± 0.00** | **0.26 ± 0.00** |
| Concrete | 3.20 ± 0.05 | 4.29 ± 0.08 | 4.43 ± 0.08 | 3.26 ± 0.07 | 3.03 ± 0.09 | **2.74 ± 0.05** | **2.83 ± 0.05** | 3.21 ± 0.05 | 3.31 ± 0.05 | 3.05 ± 0.05 | 3.04 ± 0.05 |
| Winered | **0.34 ± 0.00** | **0.34 ± 0.00** | **0.35 ± 0.00** | **0.34 ± 0.00** | **0.34 ± 0.00** | **0.34 ± 0.00** | **0.34 ± 0.00** | **0.34 ± 0.00** | **0.34 ± 0.00** | **0.34 ± 0.00** | **0.34 ± 0.00** |
| Power | 2.27 ± 0.01 | 2.31 ± 0.01 | 2.31 ± 0.01 | 2.21 ± 0.01 | 2.18 ± 0.01 | **2.14 ± 0.01** | **2.14 ± 0.01** | 2.17 ± 0.01 | 2.16 ± 0.01 | 2.17 ± 0.01 | **2.15 ± 0.01** |
| Protein | 2.56 ± 0.00 | 2.87 ± 0.00 | 2.86 ± 0.01 | 2.54 ± 0.00 | 2.43 ± 0.00 | 2.38 ± 0.00 | 2.33 ± 0.00 | 2.31 ± 0.00 | **2.19 ± 0.00** | **2.19 ± 0.00** | **2.20 ± 0.00** |
| Naval | **0.00 ± 0.00** | **0.00 ± 0.00** | **0.00 ± 0.00** | **0.00 ± 0.00** | **0.00 ± 0.00** | **0.00 ± 0.00** | **0.00 ± 0.00** | **0.00 ± 0.00** | **0.00 ± 0.00** | **0.00 ± 0.00** | **0.00 ± 0.00** |
| Kin8nm | 0.05 ± 0.00 | 0.09 ± 0.0 | 0.10 ± 0.00 | **0.04 ± 0.00** | **0.04 ± 0.00** | **0.04 ± 0.00** | **0.04 ± 0.00** | **0.03 ± 0.00** | **0.03 ± 0.00** | **0.03 ± 0.00** | **0.03 ± 0.00** |

# C

# Mathematical Proofs

## C.1 Bayesian Inference, Gaussian Processes and Generalization Bounds

**Proposition 2.9** [Return to statement]. The Kullback-Leibler divergence is always non-negative.

*Proof.* As the logarithm is bounded by $x - 1$,

$$\log x \leq x - 1 \implies \frac{P(x)}{Q(x)} - 1 \geq \log \frac{P(x)}{Q(x)}. \tag{C.1}$$

Since probabilities are non-negative, multiply by $Q(x)$ in the last inequality

$$P(x) - Q(x) \geq Q(x) \log \frac{P(x)}{Q(x)} = Q(x) \log P(x) - Q(x) \log Q(x). \tag{C.2}$$

Now, integrating both sides

$$0 \geq \mathbb{E}_Q[\log P(x) - \log Q(x)] \implies \mathbb{E}_Q[\log Q(x) - \log P(x)] \geq 0. \tag{C.3}$$

Moreover, the Kullback-Leibler divergence is 0 if and only if the two distributions are equal almost everywhere. □

**Proposition 2.10** [Return to statement]. The $\alpha$–divergence interpolates between the two asymmetric KLs:

$$\lim_{\alpha \to 0} D_\alpha(P|Q) = \mathrm{KL}(Q|P), \quad \text{and,} \quad \lim_{\alpha \to 1} D_\alpha(P|Q) = \mathrm{KL}(P|Q). \tag{C.4}$$

*Proof.* Write $I(\alpha) := \int P(\boldsymbol{\theta})^{\alpha} Q(\boldsymbol{\theta})^{1-\alpha}\,\mathrm{d}\boldsymbol{\theta}$. The $\alpha$–divergence in Equation (2.18) can be expressed as

$$D_\alpha(P|Q) = \frac{1 - I(\alpha)}{\alpha(1-\alpha)}\,. \tag{C.5}$$

Because $\int Q(\boldsymbol{\theta})\,\mathrm{d}\boldsymbol{\theta} = 1$, it verifies that $I(0) = 1$. L'Hôpital's rule therefore applies:

$$\lim_{\alpha\to 0} D_\alpha(P|Q) = \lim_{\alpha\to 0} \frac{-I'(\alpha)}{1-2\alpha}\,, \tag{C.6}$$

where $I'(\alpha)$ denotes the derivative of $I$ with respect to $\alpha$. Differentiate under the integral sign (permitted by the positivity of $P, Q$):

$$I'(\alpha) = \int P(\boldsymbol{\theta})^{\alpha} Q(\boldsymbol{\theta})^{1-\alpha} \log \frac{P(\boldsymbol{\theta})}{Q(\boldsymbol{\theta})}\,\mathrm{d}\boldsymbol{\theta}. \tag{C.7}$$

Evaluating at $\alpha = 0$ yields $I'(0) = \int Q(\boldsymbol{\theta}) \log \frac{P(\boldsymbol{\theta})}{Q(\boldsymbol{\theta})}\,\mathrm{d}\boldsymbol{\theta}$. Consequently

$$\lim_{\alpha\to 0} D_\alpha(P|Q) = -\int Q(\boldsymbol{\theta}) \log \frac{P(\boldsymbol{\theta})}{Q(\boldsymbol{\theta})}\,\mathrm{d}\boldsymbol{\theta} = \mathrm{KL}(Q|P). \tag{C.8}$$

Using the symmetry $D_\alpha(P|Q) = D_{1-\alpha}(Q|P)$, substitute $\tilde{\alpha} = 1 - \alpha$ and apply the first part:

$$\lim_{\alpha\to 1} D_\alpha(P|Q) = \lim_{\tilde{\alpha}\to 0} D_{\tilde{\alpha}}(Q|P) = \mathrm{KL}(P|Q)\,. \tag{C.9}$$

□

**Theorem 2.19** [Return to statement]. Let $K : \mathcal{X} \times \mathcal{X} \to \mathbb{R}$ be a positive-definite kernel, $\mathcal{H}$ its RKHS, and $\phi_x := K(\,\cdot\,, x) \in \mathcal{H}$. Define the centred cylindrical Gaussian measure

$$P_{\text{cyl}} \;=\; \text{Gauss}_{\text{cyl}}(0, I_{\mathcal{H}}) \quad \text{on the cylindrical } \sigma\text{-algebra } \mathscr{C}(\mathcal{H}). \tag{C.10}$$

For every finite index set $X = \{x_1, \dots, x_m\} \subset \mathcal{X}$,

$$(\delta_{x_1}, \dots, \delta_{x_m})_{\#} P_{\text{cyl}} \;=\; \mathcal{N}(0,\, K(X, X))\,, \tag{C.11}$$

so $P_{\text{cyl}}$ reproduces exactly the finite-dimensional marginals of the process $f \sim \mathcal{GP}(0, K)$.

*Proof.* For each $x \in \mathcal{X}$, the evaluation map $\delta_x : \mathcal{H} \to \mathbb{R}$, $\delta_x(h) = h(x)$, is a continuous linear functional on $\mathcal{H}$ because, by the RKHS property with $\phi_x := K(\cdot, x)$,

$$\delta_x(h) = \langle h, \phi_x \rangle_{\mathcal{H}} \quad \text{and} \quad |\delta_x(h)| \le \|h\|_{\mathcal{H}}\, \|\phi_x\|_{\mathcal{H}} = \|h\|_{\mathcal{H}} \sqrt{K(x,x)}. \tag{C.12}$$

Hence $\delta_x = L_{\phi_x}$. For a finite set $X = \{x_1, \dots, x_m\}$, by the push-forward definition and the previous display,

$$(\delta_{x_1}, \dots, \delta_{x_m})_{\sharp} P_{\text{cyl}} = \mathcal{N}(0,\, [\langle \phi_{x_j},\, \phi_{x_i} \rangle_{\mathcal{H}}]_{i,j=1}^{m}). \tag{C.13}$$

The reproducing property yields $\langle \phi_{x_j}, \phi_{x_i} \rangle_{\mathcal{H}} = K(x_i, x_j)$, so

$$(\delta_{x_1}, \ldots, \delta_{x_m})_\sharp P_{\text{cyl}} \;=\; \mathcal{N}(0,\, K(X, X)). \tag{C.14}$$

Therefore $P_{\text{cyl}}$ reproduces exactly the finite-dimensional marginals of the Gaussian process $f \sim \mathcal{GP}(0, K)$, as claimed. □

**Theorem 2.20** [Return to statement]. Let $(\mathcal{H}, \langle \cdot, \cdot \rangle_{\mathcal{H}})$ be a separable infinite-dimensional Hilbert space. Then there exist:

1. a separable Banach space $(\mathcal{B}, \| \cdot \|_{\mathcal{B}})$ that contains $\mathcal{H}$ as a dense subspace under $\| \cdot \|_{\mathcal{B}}$,
2. a continuous, injective embedding $i : \mathcal{H} \hookrightarrow \mathcal{B}$ such that the push-forward of the canonical cylindrical Gaussian on $\mathcal{H}$ with covariance $I_{\mathcal{H}}$, $\text{Gauss}_{\text{cyl}}(0, I_{\mathcal{H}})$, under $i$ extends *uniquely* to a centred Gaussian Radon probability measure $\gamma = \mathcal{N}_{\mathcal{B}}(0, J)$ on $\mathcal{B}$ with covariance operator

$$J = i \circ i^* : \mathcal{B}^* \longrightarrow \mathcal{B}, \tag{C.15}$$

   where $i^* : \mathcal{B}^* \to \mathcal{H}$ is the adjoint operator of $i$.

such that the triple $(\mathcal{B}, \gamma, \mathcal{H})$ is an *abstract Wiener space*; that is, $\mathcal{H}$ is the Cameron–Martin space of $\gamma$.

*Proof.* **Step 1: Construct a larger (Hilbert) Banach space and the embedding.** Let $(e_k)_{k \geq 1}$ be an orthonormal basis of $\mathcal{H}$. Choose weights $(\beta_k)_{k \geq 1}$ with $\beta_k \geq 1$ and

$$\sum_{k=1}^{\infty} \beta_k^{-2} \;<\; \infty, \quad \text{for example, } \beta_k = k\,. \tag{C.16}$$

Define, for $h \in \mathcal{H}$,

$$\|h\|_{\mathcal{B}}^2 \;:=\; \sum_{k=1}^{\infty} \beta_k^{-2}\, |\langle h, e_k \rangle_{\mathcal{H}}|^2. \tag{C.17}$$

Since $\beta_k^{-2} \leq 1$, there holds $\|h\|_{\mathcal{B}} \leq \|h\|_{\mathcal{H}}$ for all $h \in \mathcal{H}$; hence, the identity map

$$\text{id} : (\mathcal{H}, \| \cdot \|_{\mathcal{H}}) \longrightarrow (\mathcal{H}, \| \cdot \|_{\mathcal{B}}) \tag{C.18}$$

is continuous. Let $\mathcal{B}$ be the completion of $\mathcal{H}$ with respect to $\| \cdot \|_{\mathcal{B}}$. The sesquilinear form

$$\langle h, g \rangle_{\mathcal{B}} \;:=\; \sum_{k=1}^{\infty} \beta_k^{-2}\, \langle h, e_k \rangle_{\mathcal{H}}\, \langle g, e_k \rangle_{\mathcal{H}}, \qquad h, g \in \mathcal{H}, \tag{C.19}$$

extends by continuity to an inner product on $\mathcal{B}$; therefore $\mathcal{B}$ is a separable Hilbert space (in particular, a Banach space), and $\mathcal{H}$ is dense in $\mathcal{B}$. Let

$$i : \mathcal{H} \hookrightarrow \mathcal{B} \tag{C.20}$$

be the canonical inclusion. From Equation (C.17),

$$\|i(e_k)\|_{\mathcal{B}}^2 \;=\; \beta_k^{-2}, \qquad \sum_{k=1}^{\infty} \|i(e_k)\|_{\mathcal{B}}^2 \;=\; \sum_{k=1}^{\infty} \beta_k^{-2} \;<\; \infty, \tag{C.21}$$

so $i$ is Hilbert–Schmidt. This will guarantee the square-summability of the Gaussian series below.

**Step 2: Build a $\mathcal{B}$-valued Gaussian by an $L^2$-convergent series.** Let $(\xi_k)_{k\geq 1}$ be independent standard normal variables on some $(\Omega, \mathscr{F}, \mathbb{P})$ and define

$$X_N \;:=\; \sum_{k=1}^{N} \xi_k \, i(e_k) \quad \in \mathcal{B}, \qquad N \geq 1. \tag{C.22}$$

For integers $M < N$, by the bilinearity of $\langle \cdot, \cdot \rangle_{\mathcal{B}}$ and their independence,

$$\mathbb{E}\|X_N - X_M\|_{\mathcal{B}}^2 = \mathbb{E}\Big\langle \sum_{k=M+1}^{N} \xi_k \, i(e_k), \; \sum_{j=M+1}^{N} \xi_j \, i(e_j) \Big\rangle_{\mathcal{B}} \tag{C.23}$$

$$= \sum_{k=M+1}^{N} \sum_{j=M+1}^{N} \mathbb{E}[\xi_k \xi_j] \; \langle i(e_k), i(e_j) \rangle_{\mathcal{B}} \tag{C.24}$$

$$= \sum_{k=M+1}^{N} \|i(e_k)\|_{\mathcal{B}}^2 \;=\; \sum_{k=M+1}^{N} \beta_k^{-2}. \tag{C.25}$$

Equation (C.24) utilises the linearity of expectation; Equation (C.25) uses $\mathbb{E}[\xi_k \xi_j] = \delta_{kj}$. By Equation (C.16), the right-hand side tends to 0 as $M, N \to \infty$. Thus $(X_N)$ is Cauchy in $L^2(\Omega; \mathcal{B})$; hence, there exists a $\mathcal{B}$-valued random element $X$ with

$$X_N \xrightarrow[N \to \infty]{L^2(\Omega; \mathcal{B})} X. \tag{C.26}$$

Define the probability measure

$$\gamma \;:=\; \mathbb{P} \circ X^{-1} \tag{C.27}$$

on the Borel $\sigma$-algebra of $\mathcal{B}$. Since $X$ is an $L^2$-limit of finite Gaussian sums, $\gamma$ is a centred Gaussian probability measure on $\mathcal{B}$. To justify this, observe that for every $\ell \in \mathcal{B}^*$ we have

$$\ell(X_N) = \sum_{k=1}^{N} \xi_k \, \ell(i(e_k)), \tag{C.28}$$

a finite linear combination of independent $\mathcal{N}(0, 1)$ variables; hence, a centred real Gaussian. Since $\ell$ is continuous,

$$\mathbb{E}\,|\ell(X_N) - \ell(X)|^2 \leq \|\ell\|^2 \, \mathbb{E}\|X_N - X\|^2 \xrightarrow[N \to \infty]{} 0, \tag{C.29}$$

so $\ell(X_N) \to \ell(X)$ in $L^2(\Omega)$. Moreover, with $a_k := \ell(i(e_k))$ we have

$$\sum_{k\geq 1} a_k^2 = \sum_{k\geq 1} |\ell(i(e_k))|^2 \leq \|\ell\|^2 \sum_{k\geq 1} \|i(e_k)\|_{\mathcal{B}}^2 = \|\ell\|^2 \sum_{k\geq 1} \beta_k^{-2} < \infty, \tag{C.30}$$

so $\ell(X) = \sum_{k\geq 1} a_k \xi_k$ in $L^2$, hence $\ell(X) \sim \mathcal{N}(0, \sum a_k^2)$. By the standard characterization (a $\mathcal{B}$–valued law is Gaussian if and only if all its continuous linear functionals are real Gaussian), the push-forward $\gamma = \mathbb{P} \circ X^{-1}$ is a centred Gaussian probability measure on $\mathcal{B}$.

**Step 3: Identify the covariance operator as $J = i\, i^* : \mathcal{B} \to \mathcal{B}$.** Let $i^* : \mathcal{B} \to \mathcal{H}$ denote the *Hilbert adjoint* of $i$, that is,

$$\langle i(h),\, v\rangle_{\mathcal{B}} \;=\; \langle h,\, i^*(v)\rangle_{\mathcal{H}} \qquad \text{for all } h \in \mathcal{H},\; v \in \mathcal{B}. \tag{C.31}$$

Fix $v, w \in \mathcal{B}$. Using Equation (C.26) and the Cauchy–Schwarz inequality in $L^2$, the mapping $Y \mapsto \mathbb{E}[\langle Y, v\rangle_{\mathcal{B}} \langle Y, w\rangle_{\mathcal{B}}]$ is continuous for $L^2$-convergence; hence,

$$\mathbb{E}\big[\langle X, v\rangle_{\mathcal{B}}\, \langle X, w\rangle_{\mathcal{B}}\big] \;=\; \lim_{N\to\infty} \mathbb{E}\big[\langle X_N, v\rangle_{\mathcal{B}}\, \langle X_N, w\rangle_{\mathcal{B}}\big]. \tag{C.32}$$

Expanding $X_N$ from Equation (C.22) and using bilinearity,

$$\mathbb{E}\big[\langle X_N, v\rangle_{\mathcal{B}}\, \langle X_N, w\rangle_{\mathcal{B}}\big] = \mathbb{E}\Big[\Big\langle \sum_{k=1}^N \xi_k i(e_k),\, v\Big\rangle_{\mathcal{B}} \Big\langle \sum_{j=1}^N \xi_j i(e_j),\, w\Big\rangle_{\mathcal{B}}\Big] \tag{C.33}$$

$$= \sum_{k=1}^N \sum_{j=1}^N \mathbb{E}[\xi_k \xi_j]\, \langle i(e_k), v\rangle_{\mathcal{B}}\, \langle i(e_j), w\rangle_{\mathcal{B}} \tag{C.34}$$

$$= \sum_{k=1}^N \langle i(e_k), v\rangle_{\mathcal{B}}\, \langle i(e_k), w\rangle_{\mathcal{B}}. \tag{C.35}$$

Equation (C.34) uses the linearity of expectation; Equation (C.35) uses $\mathbb{E}[\xi_k \xi_j] = \delta_{kj}$. Now use the definition of the adjoint, Equation (C.31), to rewrite

$$\langle i(e_k), v\rangle_{\mathcal{B}} \;=\; \langle e_k,\, i^*(v)\rangle_{\mathcal{H}}. \tag{C.36}$$

Substituting Equation (C.36) into Equation (C.35) and letting $N \to \infty$, the Parseval identity in $\mathcal{H}$ yields

$$\mathbb{E}\big[\langle X, v\rangle_{\mathcal{B}}\, \langle X, w\rangle_{\mathcal{B}}\big] = \sum_{k=1}^\infty \langle e_k,\, i^*(v)\rangle_{\mathcal{H}}\, \langle e_k,\, i^*(w)\rangle_{\mathcal{H}} \tag{C.37}$$

$$= \langle i^*(v),\, i^*(w)\rangle_{\mathcal{H}} \;=\; \langle i\,(i^*(v)),\, w\rangle_{\mathcal{B}}. \tag{C.38}$$

Thus, the covariance operator of $\gamma$ is

$$J \;=\; i \circ i^* : \mathcal{B} \longrightarrow \mathcal{B}. \tag{C.39}$$

**Step 4: $J$ is trace class; explicit diagonalisation.** Define

$$g_k \;:=\; \beta_k\, i(e_k) \in \mathcal{B}. \tag{C.40}$$

From Equation (C.19) and Equation (C.21),

$$\langle g_k, g_j\rangle_{\mathcal{B}} \;=\; \beta_k\beta_j\, \langle i(e_k), i(e_j)\rangle_{\mathcal{B}} \;=\; \beta_k\beta_j\, \beta_k^{-2}\delta_{kj} \;=\; \delta_{kj}, \tag{C.41}$$

so $(g_k)_{k\geq 1}$ is an orthonormal basis of $\mathcal{B}$. Moreover, for $e_k \in \mathcal{H}$,

$$\langle (i^* \circ i)e_k,\, e_j\rangle_{\mathcal{H}} \;=\; \langle i(e_k),\, i(e_j)\rangle_{\mathcal{B}} \;=\; \beta_k^{-2}\delta_{kj}, \tag{C.42}$$

hence $(i^* \circ i)e_k = \beta_k^{-2} e_k$. Using Equation (C.39),

$$J(g_k) = i \circ i^*(\beta_k i(e_k)) \;=\; \beta_k\, i((i^* \circ i)e_k) \;=\; \beta_k\, i(\beta_k^{-2} e_k) \;=\; \beta_k^{-2}\, g_k. \tag{C.43}$$

Therefore $J$ is diagonal in the orthonormal basis $(g_k)$ with eigenvalues $(\beta_k^{-2})$, and

$$\operatorname{tr} J \;=\; \sum_{k=1}^{\infty} \langle J(g_k),\, g_k\rangle_{\mathcal{B}} \;=\; \sum_{k=1}^{\infty} \beta_k^{-2} \;<\; \infty, \tag{C.44}$$

so $J$ is trace class. In particular, $\gamma = \mathcal{N}_{\mathcal{B}}(0, J)$ is a centred Gaussian probability measure on the separable Hilbert space $\mathcal{B}$, hence a Radon measure (every Borel probability on a separable Hilbert space is Radon).

**Step 5: The push-forward of the cylindrical Gaussian equals $\gamma$.** Let $\mathrm{Gauss}_{\mathrm{cyl}}(0, I_{\mathcal{H}})$ denote the standard cylindrical Gaussian pre-measure on $\mathcal{H}$. Push it forward to $\mathcal{B}$ by $i : \mathcal{H} \to \mathcal{B}$ and set

$$\mu \;:=\; i_{\sharp}\mathrm{Gauss}_{\mathrm{cyl}}(0, I_{\mathcal{H}}). \tag{C.45}$$

a cylindrical pre-measure on $\mathcal{B}$ defined on the cylindrical $\sigma$-algebra generated by $\mathcal{B}^*$. Because $\mathcal{B}$ is a Hilbert space, the Riesz map $R_{\mathcal{B}} : \mathcal{B} \to \mathcal{B}^*$, $R_{\mathcal{B}}(v) = \langle\,\cdot\,, v\rangle_{\mathcal{B}}$, is an isometric isomorphism. For each $\ell \in \mathcal{B}^*$ let $v_\ell := R_{\mathcal{B}}^{-1}(\ell) \in \mathcal{B}$. Then for $x \in \mathcal{H}$,

$$(\ell \circ i)(x) = \ell(i(x)) = \langle i(x), v_\ell\rangle_{\mathcal{B}} = \langle x,\, i^*(v_\ell)\rangle_{\mathcal{H}}. \tag{C.46}$$

Thus $L_\ell := \ell \circ i \in \mathcal{H}^*$ is represented by $h_\ell := i^*(v_\ell) \in \mathcal{H}$. Fix $\ell_1, \ldots, \ell_m \in \mathcal{B}^*$. By definition of $\mu$ and of the cylindrical Gaussian on $\mathcal{H}$,

$$(\ell_1, \ldots, \ell_m)_{\sharp}\mu \;=\; (L_{\ell_1}, \ldots, L_{\ell_m})_{\sharp}\mathrm{Gauss}_{\mathrm{cyl}}(0, I_{\mathcal{H}}) \;=\; \mathcal{N}(0,\, [\,\langle h_{\ell_p}, h_{\ell_q}\rangle_{\mathcal{H}}\,]_{p,q}). \tag{C.47}$$

By Step 2, for every $\ell \in \mathcal{B}^*$,

$$\ell(X) = \sum_{k\geq 1} \xi_k\, \ell(i(e_k)) \quad \text{in } L^2(\Omega), \qquad \sum_{k\geq 1} |\ell(i(e_k))|^2 \leq \|\ell\|^2 \sum_{k\geq 1} \beta_k^{-2} < \infty, \tag{C.48}$$

so $\ell(X)$ is centred Gaussian with variance $\sum_{k\geq 1} \ell(i(e_k))^2$. For $\ell_p, \ell_q \in \mathcal{B}^*$,

$$\mathbb{E}[\ell_p(X)\,\ell_q(X)] = \sum_{k\geq 1} \ell_p(i(e_k))\,\ell_q(i(e_k)) = \sum_{k\geq 1} \langle e_k, h_{\ell_p}\rangle_{\mathcal{H}} \langle e_k, h_{\ell_q}\rangle_{\mathcal{H}} = \langle h_{\ell_p}, h_{\ell_q}\rangle_{\mathcal{H}}, \tag{C.49}$$

where we used $\ell(i(e_k)) = \langle i(e_k), v_\ell\rangle_{\mathcal{B}} = \langle e_k, i^*(v_\ell)\rangle_{\mathcal{H}}$ and Parseval in $\mathcal{H}$. Hence

$$(\ell_1, \ldots, \ell_m)_\sharp \gamma = \mathcal{N}(0,\, [\,\langle h_{\ell_p}, h_{\ell_q}\rangle_{\mathcal{H}}\,]_{p,q}). \tag{C.50}$$

Comparing Equations (C.47) and (C.50) shows that $\mu$ and $\gamma$ agree on all finite-dimensional distributions determined by $\mathcal{B}^*$. Let $\mathcal{C}(\mathcal{B}^*)$ be the cylindrical $\sigma$-algebra generated by $\{\ell : \mathcal{B} \to \mathbb{R} : \ell \in \mathcal{B}^*\}$. The previous paragraph implies $\mu = \gamma$ on $\mathcal{C}(\mathcal{B}^*)$. Since $\mathcal{B}$ is separable, $\mathcal{C}(\mathcal{B}^*) = \mathscr{B}(\mathcal{B})$ (Bogachev, 1998; Kuo, 2006); hence, $\mu = \gamma$ on $\mathscr{B}(\mathcal{B})$.

**Step 6: Identify the Cameron–Martin space.** Let $J^{1/2}$ be the unique positive square root of $J = i\,i^*$. The Cameron–Martin space $\mathcal{H}_\gamma$ is $\operatorname{Ran} J^{1/2}$ with the inner product

$$\langle u, v\rangle_{\mathcal{H}_\gamma} := \langle J^{-1/2}(u),\, J^{-1/2}(v)\rangle_{\mathcal{B}}. \tag{C.51}$$

Using Equation (C.43), $J^{1/2}(g_k) = \beta_k^{-1} g_k$. For $h = \sum_k a_k e_k \in \mathcal{H}$,

$$i(h) = \sum_{k=1}^{\infty} a_k\, i(e_k) = \sum_{k=1}^{\infty} a_k\, \beta_k^{-1} g_k, \qquad J^{-1/2} i(h) = \sum_{k=1}^{\infty} a_k\, g_k. \tag{C.52}$$

Hence, for $h, g \in \mathcal{H}$,

$$\langle i(h),\, i(g)\rangle_{\mathcal{H}_\gamma} = \Big\langle \sum_k a_k g_k,\, \sum_k b_k g_k \Big\rangle_{\mathcal{B}} = \sum_{k=1}^{\infty} a_k b_k = \langle h, g\rangle_{\mathcal{H}}. \tag{C.53}$$

Thus $i : \mathcal{H} \to \mathcal{H}_\gamma$ is an isometric isomorphism onto its range and

$$\operatorname{Ran} J^{1/2} = i(\mathcal{H}). \tag{C.54}$$

Therefore $\mathcal{H}$ is the Cameron–Martin space of $\gamma$.

The construction provides a separable Banach space $\mathcal{B}$ containing $\mathcal{H}$ densely, a continuous injective map $i : \mathcal{H} \hookrightarrow \mathcal{B}$ such that $i_\sharp \mathrm{Gauss}_{\mathrm{cyl}}(0, I_{\mathcal{H}})$ extends to the centred Gaussian Radon measure $\gamma = \mathcal{N}_{\mathcal{B}}(0, J)$ with covariance $J = i\,i^*$, and the Cameron–Martin space of $\gamma$ is $\mathcal{H}$. This proves the theorem. □

**Theorem 2.21** [Return to statement]. Let $f \sim \mathcal{GP}(m, K)$, with $m : \mathcal{X} \to \mathbb{R}$ and $K : \mathcal{X} \times \mathcal{X} \to \mathbb{R}$. Let $m \in \mathcal{H}$, where $\mathcal{H}$ is the RKHS generated by $K$ with sections $\phi_x := K(\cdot\,, x)$. Let $(\mathcal{B}, \gamma, \mathcal{H})$ be the abstract Wiener space constructed in Theorem 2.20; write $i : \mathcal{H} \hookrightarrow \mathcal{B}$ for the embedding and let $J = i \circ i^*$ be its covariance operator so that $\gamma = \mathcal{N}_{\mathcal{B}}(0, J)$. Let $Z \sim \gamma$ and define $F := i(m) + Z$. Then:

1. $F$ is a $\mathcal{B}$-valued Gaussian random element with law

$$F \sim \mathcal{N}_{\mathcal{B}}(i(m),\ J). \tag{C.55}$$

2. Let $\mathcal{W} : \mathcal{H} \to L^2(\gamma)$ denote the Paley–Wiener map associated with $(\mathcal{B}, \gamma, \mathcal{H})$, i.e. for $h \in \mathcal{H}$, $\mathcal{W}(h)$ is the $\gamma$-measurable linear functional on $\mathcal{B}$ satisfying $\mathcal{W}(h)(i(g)) = \langle h, g\rangle_{\mathcal{H}}$ for all $g \in \mathcal{H}$. Then, for every finite set $X = \{x_1, \dots, x_n\} \subset \mathcal{X}$,

$$(\mathcal{W}(\phi_{x_1}), \dots, \mathcal{W}(\phi_{x_n}))_\sharp \mathcal{N}_{\mathcal{B}}(i(m),\ J) \ = \ \mathcal{N}(m(X),\, K(X, X)). \tag{C.56}$$

   Equivalently, $(\mathcal{W}(\phi_{x_1})(F), \dots, \mathcal{W}(\phi_{x_n})(F)) \sim \mathcal{N}(m(X),\, K(X, X))$.

*Remark.* If, in addition, the point evaluations $\delta_x$ are continuous on $\mathcal{B}$ (i.e. $\delta_x \in \mathcal{B}^*$ for all $x \in \mathcal{X}$), then $\mathcal{W}(\phi_x) = \delta_x$ $\gamma$-a.s., and the display in (2) coincides with the push-forward by $(\delta_{x_1}, \dots, \delta_{x_n})$.

*Proof.* Let $(\mathcal{B}, \gamma, \mathcal{H})$ be the abstract Wiener space produced by Theorem 2.20. Thus $\mathcal{B}$ is a separable Hilbert space, $i : \mathcal{H} \hookrightarrow \mathcal{B}$ is continuous and injective, and $\gamma = \mathcal{N}_{\mathcal{B}}(0, J)$ is a centred Gaussian Radon measure with covariance $J$.

**Step 1: $F = i(m) + Z$ is Gaussian with mean $i(m)$ and covariance $J$.** Let $Z \sim \gamma = \mathcal{N}_{\mathcal{B}}(0, J)$ and set

$$F \ := \ i(m) + Z. \tag{C.57}$$

For any $v \in \mathcal{B}$, $\langle F, v\rangle_{\mathcal{B}} = \langle i(m), v\rangle_{\mathcal{B}} + \langle Z, v\rangle_{\mathcal{B}}$ is the sum of a constant and a centred real Gaussian; hence $\langle F, v\rangle_{\mathcal{B}}$ is real Gaussian with mean $\langle i(m), v\rangle_{\mathcal{B}}$ and variance $\langle Jv, v\rangle_{\mathcal{B}}$. Since this holds simultaneously for all finite families $v_1, \dots, v_n$ by linearity, $F$ is a $\mathcal{B}$–valued Gaussian random element with

$$F \sim \mathcal{N}_{\mathcal{B}}(i(m),\ J)\,, \tag{C.58}$$

which proves item (1).

**Step 2: Paley–Wiener map and its basic identities.** Write $\mathcal{W} : \mathcal{H} \to L^2(\gamma)$ for the Paley–Wiener map associated with $(\mathcal{B}, \gamma, \mathcal{H})$. By definition, for every $h, g \in \mathcal{H}$,

$$\mathcal{W}(h)(i(g)) \ = \ \langle h,\, g\rangle_{\mathcal{H}}. \tag{C.59}$$

Moreover, $\mathcal{W}$ is linear and isometric into $L^2(\gamma)$:

$$\mathbb{E}_\gamma[\mathcal{W}(h)\, \mathcal{W}(g)] \ = \ \langle h,\, g\rangle_{\mathcal{H}}, \qquad \mathbb{E}_\gamma[\mathcal{W}(h)] = 0. \tag{C.60}$$

Since $F$ has the translated law $T_{i(m)\,\sharp}\gamma$, with $T_{i(m)}(x) := x + i(m)$, and translations by $i(m)$ with $m \in \mathcal{H}$ being quasi-invariant (Cameron–Martin Theorem), $\mathcal{W}(h)$ is defined $F$-a.s. as well. This means that any set where $W(h)$ fails to be defined has $\gamma$-measure zero, and by quasi-invariance, it also has zero measure under the translated law of $F$. Therefore, $W(h)(F)$ exists with probability one.

**Step 3: Mean and covariance of $\mathcal{W}(h)(F)$.** By linearity of $\mathcal{W}(h)$ and Equation (C.57),

$$\mathcal{W}(h)(F) \;=\; \mathcal{W}(h)(i(m)) \;+\; \mathcal{W}(h)(Z). \tag{C.61}$$

Using Equation (C.59) with $g = m$,

$$\mathbb{E}[\mathcal{W}(h)(F)] \;=\; \langle h,\, m\rangle_{\mathcal{H}}. \tag{C.62}$$

For $h, g \in \mathcal{H}$, Equation (C.60) and the fact that $Z$ is centred yield

$$\operatorname{Cov}(\mathcal{W}(h)(F),\, \mathcal{W}(g)(F)) \;=\; \operatorname{Cov}(\mathcal{W}(h)(Z),\, \mathcal{W}(g)(Z)) \;=\; \langle h,\, g\rangle_{\mathcal{H}}. \tag{C.63}$$

**Step 4: Finite-dimensional laws for the sections $h = \phi_x$.** Let $X = \{x_1, \ldots, x_n\} \subset \mathcal{X}$ and set $h_j := \phi_{x_j} = K(\cdot, x_j) \in \mathcal{H}$. By the reproducing property of the RKHS,

$$\langle \phi_{x_i},\, m\rangle_{\mathcal{H}} \;=\; m(x_i), \qquad \langle \phi_{x_i},\, \phi_{x_j}\rangle_{\mathcal{H}} \;=\; K(x_i, x_j). \tag{C.64}$$

Combining Equations (C.62), (C.63), and (C.64) gives

$$\mathbb{E}\begin{bmatrix}\mathcal{W}(\phi_{x_1})(F)\\ \vdots \\ \mathcal{W}(\phi_{x_n})(F)\end{bmatrix} \;=\; \begin{bmatrix}m(x_1)\\ \vdots \\ m(x_n)\end{bmatrix}, \qquad \operatorname{Cov}\begin{bmatrix}\mathcal{W}(\phi_{x_1})(F)\\ \vdots \\ \mathcal{W}(\phi_{x_n})(F)\end{bmatrix} \;=\; K(X, X). \tag{C.65}$$

Finally, since $F$ is Gaussian in $\mathcal{B}$ and each $\mathcal{W}(\phi_{x_j})$ is linear, the vector

$$(\mathcal{W}(\phi_{x_1})(F), \ldots, \mathcal{W}(\phi_{x_n})(F))$$

is multivariate normal with mean and covariance as in Equation (C.65). Hence

$$(\mathcal{W}(\phi_{x_1}), \ldots, \mathcal{W}(\phi_{x_n}))_{\sharp}\mathcal{N}_{\mathcal{B}}(i(m), J) \;=\; \mathcal{N}(m(X),\, K(X, X)), \tag{C.66}$$

which proves item (2).

**Remark on evaluations.** If, in addition, every point-evaluation $\delta_x$ is continuous on $\mathcal{B}$ (equivalently $\delta_x \in \mathcal{B}^*$), then the Paley–Wiener functional coincides $\gamma$-a.s. with evaluation:

$$\mathcal{W}(\phi_x) \;=\; \delta_x \quad \gamma\text{-a.s.} \tag{C.67}$$

In that case, Equation (C.66) is the stated identity with $(\delta_{x_1}, \ldots, \delta_{x_n})$. □

**Theorem 2.22** [Return to statement]. Let $K$ be a positive-definite kernel with RKHS $\mathcal{H}$ (sections $\phi_x := K(\cdot, x)$). Let $(\mathcal{B}, \gamma, \mathcal{H})$ be an abstract Wiener space with continuous embedding $i : \mathcal{H} \hookrightarrow \mathcal{B}$. Fix $m \in \mathcal{H}$ and $\Sigma \in \mathcal{L}^+(\mathcal{H})$.

1. $P := \mathcal{N}_{\mathcal{B}}(i(m),\ i\Sigma i^*)$ is a well-defined (Radon) Gaussian measure on $\mathcal{B}$ with mean $i(m)$ and covariance operator $C = i\Sigma i^*$.
2. Let $\mathcal{W} : \mathcal{H} \to L^2(\gamma)$ denote the Paley–Wiener map associated with $(\mathcal{B}, \gamma, \mathcal{H})$, i.e. for $h \in \mathcal{H}$, $\mathcal{W}(h)$ is the $\gamma$-measurable linear functional on $\mathcal{B}$ satisfying $\mathcal{W}(h)(i(g)) = \langle h, g\rangle_{\mathcal{H}}$ for all $g \in \mathcal{H}$. Then, for every finite set $X = \{x_1, \ldots, x_n\} \subset \mathcal{X}$,

$$(\mathcal{W}(\phi_{x_1}), \ldots, \mathcal{W}(\phi_{x_n}))_{\sharp} P \;=\; \mathcal{N}(m(X),\, K_\Sigma(X, X)). \tag{C.68}$$

*Remark.* If, in addition, the point evaluations $\delta_x$ are continuous on $\mathcal{B}$ (i.e. $\delta_x \in \mathcal{B}^*$ for all $x \in \mathcal{X}$), then $\mathcal{W}(\phi_x) = \delta_x$ $\gamma$-a.s., and the display in (2) coincides with the push-forward by $(\delta_{x_1}, \ldots, \delta_{x_n})$.

*Proof.* Let $(\mathcal{B}, \gamma, \mathcal{H})$ be an abstract Wiener space with continuous injection $i : \mathcal{H} \hookrightarrow \mathcal{B}$. Fix $m \in \mathcal{H}$ and a bounded, self-adjoint, positive operator $\Sigma \in \mathcal{L}^+(\mathcal{H})$. Set

$$C \;:=\; i\,\Sigma\, i^* \;:\; \mathcal{B}^* \longrightarrow \mathcal{B}. \tag{C.69}$$

**(1) Existence and form of the Gaussian measure.** Because $i$ is the abstract Wiener space embedding, it is $\gamma$-radonifying (Bogachev, 1998; Hytönen et al., 2018); hence $i\Sigma^{1/2}$ is also $\gamma$-radonifying (Hytönen et al., 2018). Let $\xi$ be an isonormal Gaussian process over $\mathcal{H}$ (i.e. a centred Gaussian family with $\mathbb{E}\langle h, \xi\rangle = 0$ and $\mathrm{Cov}(\langle h, \xi\rangle, \langle g, \xi\rangle) = \langle h, g\rangle_{\mathcal{H}}$). Define the $\mathcal{B}$-valued random elements

$$Y \;:=\; (i\Sigma^{1/2})\,\xi, \qquad F \;:=\; i(m) + Y. \tag{C.70}$$

Standard properties of radonifying operators (Bogachev, 1998; Hytönen et al., 2018) imply that $Y$ is a centred *Radon* Gaussian in $\mathcal{B}$ with covariance operator $C = i\Sigma i^*$, so $F$ is Gaussian Radon with mean $i(m)$ and covariance $C$. Equivalently, $F \sim \mathcal{N}_{\mathcal{B}}(i(m),\, i\Sigma i^*) =: P$.

**(2) GP marginals under Paley–Wiener evaluations.** Let $\mathcal{W} : \mathcal{H} \to L^2(\gamma)$ be the Paley–Wiener map of the AWS; for $h \in \mathcal{H}$, $\mathcal{W}(h)$ is a (measurable) linear functional on $\mathcal{B}$ satisfying $\mathcal{W}(h)(i(g)) = \langle h, g\rangle_{\mathcal{H}}$ for all $g \in \mathcal{H}$. (In particular, these linear functionals admit representatives that are defined $P$-a.s.) For the RKHS sections $\phi_x := K(\cdot, x)$, the reproducing property gives

$$\mathbb{E}[\mathcal{W}(\phi_x)(F)] = \mathcal{W}(\phi_x)(i(m)) = \langle \phi_x, m\rangle_{\mathcal{H}} = m(x). \tag{C.71}$$

For the covariance, the shift $i(m)$ drops out and we compute with $Y$:

$$\mathcal{W}(\phi_x)(Y) = \mathcal{W}(\phi_x)\Big(i\Big(\Sigma^{1/2}\xi\Big)\Big) = \langle \phi_x,\ \Sigma^{1/2}\xi\rangle_{\mathcal{H}} = \langle \Sigma^{1/2}\phi_x,\ \xi\rangle, \tag{C.72}$$

so

$$\mathrm{Cov}(\mathcal{W}(\phi_x)(F), \mathcal{W}(\phi_{x'})(F)) = \left\langle \Sigma^{1/2}\phi_x,\ \Sigma^{1/2}\phi_{x'} \right\rangle_{\mathcal{H}} = \langle \phi_x,\ \Sigma\phi_{x'} \rangle_{\mathcal{H}} =: K_\Sigma(x, x'). \tag{C.73}$$

Therefore, for any finite $X = \{x_1, \ldots, x_n\} \subset \mathcal{X}$,

$$(\mathcal{W}(\phi_{x_1})(F), \ldots, \mathcal{W}(\phi_{x_n})(F)) \sim \mathcal{N}(m(X),\ K_\Sigma(X, X)), \tag{C.74}$$

which is exactly the GP law with kernel $K_\Sigma$.

**Remark (point evaluations).** If, in addition, each $\delta_x \in \mathcal{B}^*$, then for all $g \in \mathcal{H}$,

$$\delta_x(i(g)) = g(x) = \langle \phi_x, g \rangle_{\mathcal{H}} = \mathcal{W}(\phi_x)(i(g)), \tag{C.75}$$

so $\delta_x$ and $\mathcal{W}(\phi_x)$ agree on the dense subspace $i(\mathcal{H})$ and hence coincide $\gamma$-a.s. (and thus $P$-a.s.). The push-forward statement can then be written with $(\delta_{x_1}, \ldots, \delta_{x_n})$ in place of $(\mathcal{W}(\phi_{x_1}), \ldots, \mathcal{W}(\phi_{x_n}))$. □

**Theorem 2.23** [Return to statement]. Let $\boldsymbol{\Theta}$ be a finite hypothesis class with cardinality $M$ and let $\ell$ be a loss bounded in $[0, C]$. Then for any $\delta \in (0, 1)$, with probability at least $1 - \delta$ over the random draw of the sample $D \sim \nu^n$, the following *simultaneous* concentration inequality holds:

$$\forall \boldsymbol{\theta} \in \boldsymbol{\Theta}: \qquad L(\boldsymbol{\theta}) \ \le\ \hat{L}(D, \boldsymbol{\theta}) + C\sqrt{\frac{\log(M/\delta)}{2n}}. \tag{C.76}$$

*Proof.* Since $\ell$ takes values in $[0, C]$, the random variables $X_i := \ell(f_{\boldsymbol{\theta}}(\mathbf{x}_i), y_i) - L(\boldsymbol{\theta})$ are independent, centered, and bounded by $C$. Hoeffding's inequality (Hoeffding, 1963) therefore gives

$$\mathbb{P}\left(L(\boldsymbol{\theta}) - \hat{L}(D, \boldsymbol{\theta}) > \varepsilon\right) \le \exp\left(-\frac{2n\varepsilon^2}{C^2}\right). \tag{C.77}$$

Equivalently, with probability at least $1 - \delta_{\boldsymbol{\theta}}$,

$$L(\boldsymbol{\theta}) \ \le\ \hat{L}(D, \boldsymbol{\theta}) + C\sqrt{\frac{\log(1/\delta_{\boldsymbol{\theta}})}{2n}}. \tag{C.78}$$

Setting $\delta_{\boldsymbol{\theta}} := \delta/M$ in Equation (C.78) and applying the union bound yields

$$\mathbb{P}\left(\exists \boldsymbol{\theta} \in \boldsymbol{\Theta} : L(\boldsymbol{\theta}) - \hat{L}(D, \boldsymbol{\theta}) > C\sqrt{\frac{\log(M/\delta)}{2n}}\right) \le \sum_{\boldsymbol{\theta} \in \boldsymbol{\Theta}} \frac{\delta}{M} = \delta\,. \tag{C.79}$$

Taking the complement proves the first equation of the theorem. □

**Theorem 2.24** [Return to statement]. [Catoni, 2007] Let the loss satisfy $0 \le \ell(\cdot,\cdot) \le C$ and fix $\delta \in (0,1)$. For every $\lambda > 0$ the following event holds with probability at least $1-\delta$ over the draw of $D \sim \nu^n$:

$$\forall \rho \in \mathcal{P}(\Theta): \quad \mathbb{E}_{\boldsymbol{\theta}\sim\rho}[L(\boldsymbol{\theta})] \;\le\; \mathbb{E}_{\boldsymbol{\theta}\sim\rho}[\hat{L}(D,\boldsymbol{\theta})] + \frac{\lambda C^2}{8n} + \frac{\mathrm{KL}(\rho|\pi) + \log(1/\delta)}{\lambda}. \tag{C.80}$$

*Proof.* Fix an arbitrary $\lambda > 0$. For each $\boldsymbol{\theta} \in \Theta$ define the centered random variable

$$Z_{\boldsymbol{\theta}}(D) := \exp\left\{\lambda\left(L(\boldsymbol{\theta}) - \hat{L}(D,\boldsymbol{\theta})\right) - \frac{\lambda^2 C^2}{8n}\right\}. \tag{C.81}$$

Because $L(\boldsymbol{\theta})$ is deterministic with respect to $D$, it verifies that

$$L(\boldsymbol{\theta}) - \hat{L}(D,\boldsymbol{\theta}) = \frac{1}{n}\sum_{i=1}^{n}\left[\mathbb{E}_{(\mathbf{x},y)\sim\nu}\,\ell\left(f_{\boldsymbol{\theta}}(\mathbf{x}), y\right) - \ell\left(f_{\boldsymbol{\theta}}(\mathbf{x}_i), y_i\right)\right]. \tag{C.82}$$

Each summand is independent, has mean 0, and lies in $[-C, C]$. Hoeffding's lemma therefore yields

$$\mathbb{E}_D\left[\exp\left(\lambda\left(L(\boldsymbol{\theta}) - \hat{L}(D,\boldsymbol{\theta})\right)\right)\right] \le \exp\left(\tfrac{\lambda^2 C^2}{8n}\right), \tag{C.83}$$

so $\mathbb{E}_D\left[Z_{\boldsymbol{\theta}}(D)\right] \le 1$ for every $\boldsymbol{\theta}$.

Define the (random) moment-generating function $M(D) := \mathbb{E}_{\boldsymbol{\theta}\sim\pi}\left[Z_{\boldsymbol{\theta}}(D)\right]$. By Fubini's theorem and the previous bound, $\mathbb{E}_D[M(D)] \le 1$. Applying Markov's inequality gives

$$\mathbb{P}_D\left(M(D) > 1/\delta\right) \le \delta. \tag{C.84}$$

With probability at least $1-\delta$ the sample therefore falls in the "good event" $\mathcal{E} := \{D : M(D) \le 1/\delta\}$.

Fix $D \in \mathcal{E}$ and any posterior $\rho \in \mathcal{P}(\boldsymbol{\Theta})$. Using Jensen's inequality (the Donsker–Varadhan variational formula),

$$\begin{aligned} \lambda\,\mathbb{E}_\rho\left[L(\boldsymbol{\theta}) - \hat{L}(D,\boldsymbol{\theta})\right] - \frac{\lambda^2 C^2}{8n} - \mathrm{KL}(\rho|\pi) &= \mathbb{E}_\rho[\log Z_{\boldsymbol{\theta}}(D)] && \text{(C.85)}\\ &\le \log \mathbb{E}_\rho\left[Z_{\boldsymbol{\theta}}(D)\right] && \text{(C.86)}\\ &\le \log M(D) \;\le\; \log(1/\delta). && \text{(C.87)} \end{aligned}$$

Rearranging and dividing by $\lambda$ yields

$$\mathbb{E}_\rho\left[L(\boldsymbol{\theta})\right] \;\le\; \mathbb{E}_\rho\left[\hat{L}(D,\boldsymbol{\theta})\right] + \frac{\lambda C^2}{8n} + \frac{\mathrm{KL}(\rho|\pi) + \log(1/\delta)}{\lambda}. \tag{C.88}$$

Because the argument did not rely on the specific choice of $\rho$, the bound holds *simultaneously for all posteriors.* □

## C.2 Deep Variational Implicit Processes

**Lemma 3.3** [Return to statement]. The family $\{P_{\mathcal{J}}\}_{\mathcal{J}\subset\mathcal{X}}$ satisfies the Kolmogorov Extension Theorem; that is, for every pair of finite index sets $\mathcal{I} \subset \mathcal{J}$ it verifies that $\pi_{\mathcal{I}\mathcal{J}\,\sharp}P_{\mathcal{J}} = P_{\mathcal{I}}$.

*Proof.* For a given finite set $\mathcal{J} = \{\mathbf{x}_1, \dots, \mathbf{x}_k\} \subset \mathcal{X}$, recall that

$$\mathbf{f}_{\mathcal{J}} : \Omega \longrightarrow \mathbb{R}^{\mathcal{J}}, \qquad \mathbf{f}_{\mathcal{J}}(\omega) = \big(g(\omega, \mathbf{x}_1), \dots, g(\omega, \mathbf{x}_k)\big). \tag{C.89}$$

By construction $g$ is measurable in $\omega$, so $\mathbf{f}_{\mathcal{J}}$ is a measurable map between $(\Omega, \mathcal{F})$ and $(\mathbb{R}^{\mathcal{J}}, \mathcal{B}(\mathbb{R})^{\otimes\mathcal{J}})$. Set $P_{\mathcal{J}} := P \circ \mathbf{f}_{\mathcal{J}}^{-1}$. For a fixed finite set $\mathcal{I} \subset \mathcal{J}$, the canonical projection $\pi_{\mathcal{I}\mathcal{J}} : \mathbb{R}^{\mathcal{J}} \to \mathbb{R}^{\mathcal{I}}$ verifies that for any $y = (y_x)_{x\in\mathcal{J}} \in \mathbb{R}^{\mathcal{J}}$,

$$\pi_{\mathcal{I}\mathcal{J}}(y) := (y_x)_{x\in\mathcal{I}} \in \mathbb{R}^{\mathcal{I}}. \tag{C.90}$$

Observe that applying $\pi_{\mathcal{I}\mathcal{J}}$ to $\mathbf{f}_{\mathcal{J}}(\omega)$ simply discards the components with indices in $\mathcal{J} \setminus \mathcal{I}$:

$$\pi_{\mathcal{I}\mathcal{J}}\big(\mathbf{f}_{\mathcal{J}}(\omega)\big) = \big(g(\omega, \mathbf{x})\big)_{\mathbf{x}\in\mathcal{I}} = \mathbf{f}_{\mathcal{I}}(\omega), \qquad \forall\, \omega \in \Omega. \tag{C.91}$$

Hence, as pointwise maps,

$$\pi_{\mathcal{I}\mathcal{J}} \circ \mathbf{f}_{\mathcal{J}} = \mathbf{f}_{\mathcal{I}}. \tag{C.92}$$

Using the definition of push-forward measures ($f_\sharp Q := Q \circ f^{-1}$), it verifies that

$$\pi_{\mathcal{I}\mathcal{J}\,\sharp}P_{\mathcal{J}} = \pi_{\mathcal{I}\mathcal{J}\,\sharp}(P \circ \mathbf{f}_{\mathcal{J}}^{-1}) = P \circ (\mathbf{f}_{\mathcal{J}})^{-1} \circ \pi_{\mathcal{I}\mathcal{J}}^{-1} = P \circ (\pi_{\mathcal{I}\mathcal{J}} \circ \mathbf{f}_{\mathcal{J}})^{-1} = P \circ (\mathbf{f}_{\mathcal{I}})^{-1} = P_{\mathcal{I}},. \tag{C.93}$$

which is exactly the desired consistency identity. Since the argument holds for *every* finite $\mathcal{I} \subset \mathcal{J} \subset \mathcal{X}$, the family $\{P_{\mathcal{J}}\}$ is consistent, completing the proof. □

**Proposition 3.5** [Return to statement]. Let $\boldsymbol{\Omega} = \{\boldsymbol{m}_h^l, \boldsymbol{S}_h^l : l = 1, \dots, L\,, h = 1, \dots, H_l\}$ denote the parameters of $Q$ and $\boldsymbol{\Theta} = \{\boldsymbol{\theta}_h^l : l = 1, \dots, L\,, h = 1, \dots, H_l\}$ the deep IP prior parameters. Then, a variational lower bound can be derived; at whose maximum the KL divergence between $Q(\{\mathbf{f}^{\,l}, \boldsymbol{a}^l\}_{l=1}^L)$ and $P\left(\{\mathbf{f}^{\,l}, \boldsymbol{a}^l\}_{l=1}^L|\mathcal{D}\right)$ is minimized

$$\mathcal{L}\left(\boldsymbol{\Omega}, \boldsymbol{\Theta}, \{\boldsymbol{\sigma}_l^2\}_{l=1}^{L-1}\right) = \sum_{n=1}^N \mathbb{E}_{Q(\mathbf{f}_{\cdot,n}^{\,L})}\big[\log P(y_n|\mathbf{f}_{\cdot,n}^{\,L})\big] - \sum_{l=1}^L \sum_{h=1}^{H_l} \mathrm{KL}(Q(\boldsymbol{a}_h^l)|P(\boldsymbol{a}_h^l))\,. \tag{C.94}$$

*Proof.* Consider the general definition of the ELBO in variational inference, i.e.

$$\mathcal{L}\left(\boldsymbol{\Omega}, \boldsymbol{\Theta}, \{\boldsymbol{\sigma}_l^2\}_{l=1}^{L-1}\right) = \mathbb{E}_{Q(\{\mathbf{F}^{\,l}, \mathbf{a}^l\}_{l=1}^L)}\left[\log \frac{P\left(\mathbf{y}, \{\mathbf{F}^{\,l}, \mathbf{a}^l\}_{l=1}^L\right)}{Q(\{\mathbf{F}^{\,l}, \mathbf{a}^l\}_{l=1}^L)}\right], \tag{C.95}$$

using DVIPs model specification:

$$P(\mathbf{y},\{\mathbf{F}^l,\mathbf{a}^l\}_{l=1}^L) = \prod_{n=1}^N P(y_n|\mathbf{f}_{\cdot,n}^L)\prod_{n=1}^N\prod_{l=1}^L\prod_{h=1}^{H_l} P(f_{h,n}^l|\mathbf{a}_h^l)P(\mathbf{a}_h^l)\,, \tag{C.96}$$

$$Q(\{\mathbf{F}^l,\mathbf{a}^l\}_{l=1}^L) = \prod_{n=1}^N\prod_{l=1}^L\prod_{h=1}^{H_l} P(f_{h,n}^l|\mathbf{a}_h^l)Q(\mathbf{a}_h^l)\,, \tag{C.97}$$

the variational lower bound takes the following form:

$$\mathcal{L} = \mathbb{E}_{Q(\{\mathbf{F}^l,\mathbf{a}^l\}_{l=1}^L)}\left[\log\frac{P\left(\mathbf{y},\{\mathbf{F}^l,\mathbf{a}^l\}_{l=1}^L\right)}{Q(\{\mathbf{F}^l,\mathbf{a}^l\}_{l=1}^L)}\right] \tag{C.98}$$

$$= \mathbb{E}_{Q(\{\mathbf{F}^l,\mathbf{a}^l\}_{l=1}^L)}\left[\log\frac{\prod_{n=1}^N P(y_n|\mathbf{f}_{\cdot,n}^L)\prod_{n=1}^N\prod_{l=1}^L\prod_{h=1}^{H_l}\cancel{P(f_{h,n}^l|\mathbf{a}_h^l)}P(\mathbf{a}_h^l)}{\prod_{n=1}^N\prod_{l=1}^L\prod_{h=1}^{H_l}\cancel{P(f_{h,n}^l|\mathbf{a}_h^l)}Q(\mathbf{a}_h^l)}\right] \tag{C.99}$$

$$= \mathbb{E}_{Q(\{\mathbf{F}^l,\mathbf{a}^l\}_{l=1}^L)}\left[\log\frac{\prod_{n=1}^N P(y_n|\mathbf{f}_{\cdot,n}^L)\prod_{l=1}^L\prod_{h=1}^{H_l}P(\mathbf{a}_h^l)}{\prod_{l=1}^L\prod_{h=1}^{H_l}Q(\mathbf{a}_h^l)}\right]. \tag{C.100}$$

From this expression, the expectation can be split into two terms

$$\mathcal{L} = \mathbb{E}_{Q(\{\mathbf{F}^l,\mathbf{a}^l\}_{l=1}^L)}\left[\log\prod_{n=1}^N P(y_n|\mathbf{f}_{\cdot,n}^L)\right] + \mathbb{E}_{Q(\{\mathbf{F}^l,\mathbf{a}^l\}_{l=1}^L)}\left[\log\frac{\prod_{l=1}^L\prod_{h=1}^{H_l}P(\mathbf{a}_h^l)}{\prod_{l=1}^L\prod_{h=1}^{H_l}Q(\mathbf{a}_h^l)}\right]. \tag{C.101}$$

The logarithm in the first term does not depend on the regression coefficients $\{\boldsymbol{A}^l\}_{l=1}^L$ and neither on $\{\boldsymbol{F}^l\}_{l=1}^{L-1}$. On the other hand, the logarithm in the second term does not depend on $\{\boldsymbol{F}^l\}_{l=1}^L$. Thus,

$$\mathcal{L} = \mathbb{E}_{Q(\mathbf{F}^L)}\left[\log\prod_{n=1}^N P(y_n|\mathbf{f}_{\cdot,n}^L)\right] + \mathbb{E}_{Q(\{\mathbf{a}^l\}_{l=1}^L)}\left[\log\frac{\prod_{l=1}^L\prod_{h=1}^{H_l}P(\mathbf{a}_h^l)}{\prod_{l=1}^L\prod_{h=1}^{H_l}Q(\mathbf{a}_h^l)}\right] \tag{C.102}$$

$$= \sum_{n=1}^N \mathbb{E}_q\left[\log P(y_n|\mathbf{f}_{\cdot,n}^L)\right] - \sum_{l=1}^L\sum_{h=1}^{H_l}\mathrm{KL}(Q(\mathbf{a}_h^l)|P(\mathbf{a}_h^l))\,. \tag{C.103}$$

□

## C.3 Variational Linearized Laplace Approximation

**Theorem 4.1** [Return to statement]. For any kernel function $K$ and its corresponding RKHS $\mathcal{H}$ —with feature maps defined as $\phi_{\mathbf{x}} := K(\cdot,\mathbf{x}) \in \mathcal{H}$—, any $\boldsymbol{a} \in \mathbb{R}^M$, $\boldsymbol{A} \in \mathbb{R}^{M\times M}$ with $\boldsymbol{A} \succeq 0$ and $\mathbf{Z} = (\mathbf{z}_1,\dots,\mathbf{z}_M) \in \mathcal{X}^M$, let $\tilde{\mu} \in \mathcal{H}$ and $\tilde{\Sigma} : \mathcal{H} \to \mathcal{H}$ be defined as

$$\tilde{\mu} = \sum_{m=1}^M a_m\phi_{\mathbf{z}_m}, \qquad \text{and} \qquad \tilde{\Sigma} = \left(I + \sum_{i=1}^M\sum_{j=1}^M \phi_{\mathbf{z}_i}A_{i,j}\phi_{\mathbf{z}_j}^T\right)^{-1}, \tag{C.104}$$

then, $\tilde{\Sigma} \in L^+(\mathcal{H})$, and the Gaussian measure defined by $(\tilde{\mu}, \tilde{\Sigma})$ is equivalent to a SVGP with variational distribution $Q(\mathbf{u}) = \mathcal{N}(\mathbf{u}|\boldsymbol{m}, \boldsymbol{S})$, defined as

$$\boldsymbol{m} = K(\mathbf{Z}, \mathbf{Z})\boldsymbol{a}, \quad \text{and} \quad \boldsymbol{S} = K(\mathbf{Z}, \mathbf{Z}) - K(\mathbf{Z}, \mathbf{Z})(\boldsymbol{A}^{-1} + K(\mathbf{Z}, \mathbf{Z}))^{-1} K(\mathbf{Z}, \mathbf{Z}). \quad \text{(C.105)}$$

*Proof.* Write $\Psi = [\phi_{\mathbf{z}_1} \cdots \phi_{\mathbf{z}_M}] : \mathbb{R}^M \to \mathcal{H}$. Because $\boldsymbol{A} = \boldsymbol{A}^\top$ it verifies that $\Psi \boldsymbol{A} \Psi^\top = (\Psi \boldsymbol{A} \Psi^\top)^*$, hence $(\tilde{\Sigma}^{-1})^* = \tilde{\Sigma}^{-1}$. For any $f \in \mathcal{H}$,

$$\langle f, \tilde{\Sigma}^{-1}(f) \rangle_{\mathcal{H}} = \|f\|^2_{\mathcal{H}} + \langle \Psi^\top f, \boldsymbol{A} \Psi^\top f \rangle_{\mathbb{R}^M} \geq 0, \quad \text{(C.106)}$$

because $\boldsymbol{A} \succeq 0$. Thus $\tilde{\Sigma}^{-1} \succeq 0$ and $\tilde{\Sigma}^{-1} \in \mathcal{L}^+(\mathcal{H})$. Thus $\tilde{\Sigma} \in \mathcal{L}^+(\mathcal{H})$. Recall that from the dual formulation of GPs, the mean function of the GP is defined as $m^\star(\mathbf{x}) := \langle \phi_{\mathbf{x}}, \mu \rangle$. Thus, it verifies that

$$m^\star(\mathbf{x}) = \langle \phi_{\mathbf{x}}, \tilde{\mu} \rangle = \langle \phi_{\mathbf{x}}, \sum_{m=1}^{M} a_m \phi_{\mathbf{z}_m} \rangle = \sum_{m=1}^{M} a_m \langle \phi_{\mathbf{x}}, \phi_{\mathbf{z}_m} \rangle = K(\mathbf{x}, \mathbf{Z})\boldsymbol{a}. \quad \text{(C.107)}$$

Using that $\boldsymbol{a} = K(\mathbf{Z}, \mathbf{Z})^{-1}\boldsymbol{m}$, the mean function from SVGPs (Titsias, 2009) is recovered. The same procedure can be used for the covariance matrix, using that $\tilde{\Sigma}^{-1} = I + \Psi \boldsymbol{A} \Psi^\top$, applying Woodbury's matrix identity:

$$\tilde{\Sigma} = I + \Psi \tilde{\boldsymbol{A}} \Psi^\top, \quad \text{where } \tilde{\boldsymbol{A}} = -(\boldsymbol{A}^{-1} + K(\mathbf{Z}, \mathbf{Z}))^{-1}. \quad \text{(C.108)}$$

Thus, the kernel $K^\star(\mathbf{x}, \mathbf{x}') := \langle \phi_{\mathbf{x}}, \tilde{\Sigma}(\phi_{\mathbf{x}'}) \rangle$:

$$K^\star(\mathbf{x}, \mathbf{x}') = \langle \phi_{\mathbf{x}}, \tilde{\Sigma}(\phi_{\mathbf{x}'}) \rangle = \langle \phi_{\mathbf{x}}, \phi_{\mathbf{x}'} + \sum_{i=1}^{M} \sum_{j=1}^{M} \phi_{\mathbf{z}_i} \tilde{A}_{i,j} \langle \phi_{\mathbf{z}_j}, \phi_{\mathbf{x}'} \rangle \rangle \quad \text{(C.109)}$$

$$= \langle \phi_{\mathbf{x}}, \phi_{\mathbf{x}'} \rangle + \sum_{i=1}^{M} \sum_{j=1}^{M} \langle \phi_{\mathbf{x}}, \phi_{\mathbf{z}_i} \rangle \tilde{A}_{i,j} \langle \phi_{\mathbf{z}_j}, \phi_{\mathbf{x}'} \rangle \quad \text{(C.110)}$$

$$= K(\mathbf{x}, \mathbf{x}') + \sum_{i=1}^{M} \sum_{j=1}^{M} K(\mathbf{x}, \mathbf{z}_i) \tilde{A}_{i,j} K(\mathbf{z}_j, \mathbf{x}') \quad \text{(C.111)}$$

$$= K(\mathbf{x}, \mathbf{x}') + K(\mathbf{x}, \mathbf{Z}) \tilde{\boldsymbol{A}} K(\mathbf{Z}, \mathbf{x}'). \quad \text{(C.112)}$$

From the definition of $\boldsymbol{S}$:

$$K(\mathbf{Z}, \mathbf{Z})^{-1} - K(\mathbf{Z}, \mathbf{Z})^{-1} \boldsymbol{S} K(\mathbf{Z}, \mathbf{Z})^{-1} = (\boldsymbol{A}^{-1} + K(\mathbf{Z}, \mathbf{Z}))^{-1} = -\tilde{\boldsymbol{A}}. \quad \text{(C.113)}$$

Exchanging $\tilde{\boldsymbol{A}}$ with this quantity recovers the covariance function of SVGPs and completes the proof. □

**Proposition 4.2** [Return to statement]. For any kernel $K : \mathcal{X} \times \mathcal{X} \to \mathbb{R}$ and its corresponding RKHS $\mathcal{H}$, if $g(\cdot, \hat{\boldsymbol{\theta}}) \in \mathcal{H}$, then $\forall \epsilon > 0$, $\exists M_\alpha \in \mathbb{N}^+$, $\mathbf{Z}_\alpha \in \mathcal{X}^{M_\alpha}$, $\boldsymbol{a} \in \mathbb{R}^{M_\alpha}$ such that, for any $M_\beta \in \mathbb{N}$, $\mathbf{Z}_\beta \in \mathcal{X}^{\mathcal{M}_\beta}$, $\boldsymbol{A} \in \mathbb{R}^{M_\beta \times M_\beta}$ with $\boldsymbol{A} \succeq 0$, the Gaussian measure $Q(f) = \mathcal{N}(f|\tilde{\mu}_{\alpha,\boldsymbol{a}}, \tilde{\Sigma}_{\beta,\boldsymbol{A}}) \in \mathcal{Q}^+$ has a corresponding GP with mean and covariance functions defined as

$$m^\star(\mathbf{x}) = \tilde{\mu}_{\alpha,\boldsymbol{a}}(\mathbf{x})\,, \quad K^\star(\mathbf{x},\mathbf{x}') = K(\mathbf{x},\mathbf{x}') - K_{\mathbf{x},\mathbf{Z}_\beta}(\boldsymbol{A}^{-1} + \boldsymbol{K}_\beta)^{-1}K_{\mathbf{Z}_\beta,\mathbf{x}'}\,, \tag{C.114}$$

where $d_\mathcal{H}(g(\cdot,\hat{\boldsymbol{\theta}}), \tilde{\mu}_{\alpha,\boldsymbol{a}}) \leq \epsilon$, with $d_\mathcal{H}(\cdot,\cdot)$.

*Proof.* If $g(\cdot,\hat{\boldsymbol{\theta}}) \in \mathcal{H}$, the reproducing property of the RKHS verifies that $\forall \epsilon > 0$ there exists $\mathbf{Z}_\alpha \subset \mathcal{X}$ and $\{\boldsymbol{a}_i\}_{i\in\mathbb{N}}$ such that $\tilde{\mu}_{\alpha,\boldsymbol{a}} = \sum_{i=1}^{M_\alpha} a_i \phi_{\mathbf{z}_i}$ verifies

$$d_\mathcal{H}(g(\cdot,\hat{\boldsymbol{\theta}}), \tilde{\mu}_{\alpha,\boldsymbol{a}}) \leq \epsilon\,. \tag{C.115}$$

As a result, the mean function of the SVGP is

$$m^\star(\mathbf{x}) = \langle \phi_\mathbf{x}, \tilde{\mu}_{\alpha,\boldsymbol{a}} \rangle = \tilde{\mu}_{\alpha,\boldsymbol{a}}(\mathbf{x}) \approx g(\mathbf{x},\hat{\boldsymbol{\theta}})\,. \tag{C.116}$$

On the other hand, using that the covariance function is

$$K^\star(\mathbf{x},\mathbf{x}') = \langle \phi_\mathbf{x}, \tilde{\Sigma}(\phi_{\mathbf{x}'}) \rangle = \langle \phi_\mathbf{x}, \phi_{\mathbf{x}'} \rangle - \langle \phi_\mathbf{x}, \Phi_{\mathbf{Z}_\beta}(\boldsymbol{A}^{-1} + \Phi_{\mathbf{Z}_\beta}^T \Phi_{\mathbf{Z}_\beta})^{-1} \Phi_{\mathbf{Z}_\beta}^T \phi_{\mathbf{x}'} \rangle \tag{C.117}$$

$$= K(\mathbf{x},\mathbf{x}') - K_{\mathbf{x},\mathbf{Z}_\beta}(\boldsymbol{A}^{-1} - K_{\mathbf{Z}_\beta})^{-1} K_{\mathbf{Z}_\beta,\mathbf{x}'}\,. \tag{C.118}$$

□

**Proposition 4.3** [Return to statement]. The value of $\boldsymbol{A}$ that minimizes Equation (4.9) is

$$\boldsymbol{A} = \frac{1}{\sigma^2}\boldsymbol{K}_\beta^{-1}\boldsymbol{K}_{\boldsymbol{Z}_\beta,\boldsymbol{X}}\boldsymbol{K}_{\boldsymbol{X},\boldsymbol{Z}_\beta}\boldsymbol{K}_\beta^{-1}\,, \tag{C.119}$$

where $\sigma^2$ is the noise variance and $\boldsymbol{K}_{\boldsymbol{X},\boldsymbol{Z}_\beta}$ is a matrix with the prior covariances between $f(\mathbf{X})$ and $f(\mathbf{Z}_\beta)$. If $\boldsymbol{Z}_\beta = \mathbf{X}$, the covariance function of the predictive distribution in Equation (4.24) is equal to that of the full GP.

*Proof.* The objective is to maximize the ELBO over the covariance parameter of the decoupled variational family under a Gaussian likelihood $p(\mathbf{y}|\mathbf{f}) = \mathcal{N}(\mathbf{y}|\mathbf{f}, \sigma^2\mathbf{I})$. Let $\boldsymbol{u}_\beta := f(\mathbf{Z}_\beta)$ with prior $p(\boldsymbol{u}_\beta) = \mathcal{N}(0, \boldsymbol{K}_\beta)$, where $(\boldsymbol{K}_\beta)_{ij} = K(\mathbf{z}_{\beta,i}, \mathbf{z}_{\beta,j})$. Consider a variational distribution $q(\boldsymbol{u}_\beta) = \mathcal{N}(0, \boldsymbol{S})$ with some PSD matrix $S$. Only the covariance part matters for optimizing $\mathbf{A}$. Using Gaussian conditioning identities,

$$\mathrm{Cov}_q(\mathbf{f}) = \boldsymbol{K}_{\mathbf{X},\mathbf{X}} - \boldsymbol{K}_{\mathbf{X},\mathbf{Z}_\beta}\boldsymbol{K}_\beta^{-1}\boldsymbol{K}_{\mathbf{Z}_\beta,\mathbf{X}} + \boldsymbol{K}_{\mathbf{X},\mathbf{Z}_\beta}\boldsymbol{K}_\beta^{-1}\boldsymbol{S}\boldsymbol{K}_\beta^{-1}\boldsymbol{K}_{\mathbf{Z}_\beta,\mathbf{X}}. \tag{C.120}$$

The expected log-likelihood contributes (up to a constant independent of $\boldsymbol{S}$):

$$-\frac{1}{2\sigma^2}\,\mathrm{tr}\left(\boldsymbol{K}_{\mathbf{X},\mathbf{Z}_\beta}\boldsymbol{K}_\beta^{-1}\boldsymbol{S}\boldsymbol{K}_\beta^{-1}\boldsymbol{K}_{\mathbf{Z}_\beta,\mathbf{X}}\right). \tag{C.121}$$

Define

$$\boldsymbol{C} := \boldsymbol{K}_\beta^{-1}\boldsymbol{K}_{\mathbf{Z}_\beta,\mathbf{X}}\boldsymbol{K}_{\mathbf{X},\mathbf{Z}_\beta}\boldsymbol{K}_\beta^{-1}. \tag{C.122}$$

Then the term simplifies to

$$-\frac{1}{2\sigma^2}\operatorname{tr}(\boldsymbol{S}\boldsymbol{C}). \tag{C.123}$$

The KL divergence between $q(u_\beta) = \mathcal{N}(0, \boldsymbol{S})$ and $p(u_\beta) = \mathcal{N}(0, \boldsymbol{K}_\beta)$ is

$$\mathrm{KL}(q|p) = \tfrac{1}{2}\Big(\operatorname{tr}(\boldsymbol{K}_\beta^{-1}\boldsymbol{S}) - \log\det(\boldsymbol{K}_\beta^{-1}\boldsymbol{S}) - m_\beta\Big), \tag{C.124}$$

where $m_\beta = |\mathbf{Z}_\beta|$. Thus the $\boldsymbol{S}$–dependent part of the ELBO is

$$\mathcal{L}(S) \;=\; -\frac{1}{2\sigma^2}\operatorname{tr}(\boldsymbol{S}\boldsymbol{C}) \;-\; \tfrac{1}{2}\Big(\operatorname{tr}(\boldsymbol{K}_\beta^{-1}\boldsymbol{S}) - \log\det(\boldsymbol{K}_\beta^{-1}\boldsymbol{S}) - m_\beta\Big). \tag{C.125}$$

Taking derivatives w.r.t. $\boldsymbol{S}$ and setting to zero:

$$-\frac{1}{2\sigma^2}\boldsymbol{C} - \tfrac{1}{2}\boldsymbol{K}_\beta^{-1} + \tfrac{1}{2}\boldsymbol{S}^{-1} = 0. \tag{C.126}$$

Therefore

$$\boldsymbol{S}^{-1} = \boldsymbol{K}_\beta^{-1} + \tfrac{1}{\sigma^2}\boldsymbol{C}\,. \tag{C.127}$$

Applying Woodbury matrix identity:

$$\boldsymbol{S} = \boldsymbol{K}_\beta - \boldsymbol{K}_\beta(\sigma^2\boldsymbol{C}^{-1} + \boldsymbol{K}_\beta)^{-1}\boldsymbol{K}_\beta\,. \tag{C.128}$$

In the dual parameterization, $\boldsymbol{S}$ and $\boldsymbol{A}$ are related via:

$$\boldsymbol{S} \;=\; \boldsymbol{K}_\beta - \boldsymbol{K}_\beta(\boldsymbol{A}^{-1} + \boldsymbol{K}_\beta)^{-1}\boldsymbol{K}_\beta. \tag{C.129}$$

As a result, it verifies that

$$\boldsymbol{A}^{-1} = \sigma^2\boldsymbol{C}^{-1} \implies \boldsymbol{A} = \tfrac{1}{\sigma^2}\boldsymbol{K}_\beta^{-1}\boldsymbol{K}_{\mathbf{Z}_\beta,\mathbf{X}}\boldsymbol{K}_{\mathbf{X},\mathbf{Z}_\beta}\boldsymbol{K}_\beta^{-1}\,. \tag{C.130}$$

If $\mathbf{Z}_\beta = \mathbf{X}$, then $\boldsymbol{K}_{\mathbf{Z}_\beta,\mathbf{X}} = \boldsymbol{K}_{\mathbf{X},\mathbf{X}}$ and $\boldsymbol{A}^* = \frac{1}{\sigma^2}\boldsymbol{I}$, which recovers the full GP posterior covariance.

□

**Proposition 4.5** [Return to statement]. Let $\mathcal{Z} \subset \mathcal{X}$ be compact and let $K\colon \mathcal{X}\times\mathcal{X} \to \mathbb{R}$ be a *universal* kernel in the sense of Definition 4.4. Then, for every function $f \in C(\mathcal{Z})$ and for every $\varepsilon > 0$ there exist

$$M_\alpha \in \mathbb{N}, \qquad \{\mathbf{z}_1, \dots, \mathbf{z}_{M_\alpha}\} \subset \mathcal{Z}, \quad \text{and} \quad \{a_1, \dots, a_{M_\alpha}\} \subset \mathbb{R}, \tag{C.131}$$

such that the finite kernel expansion

$$m(\mathbf{x}) \;:=\; \langle \phi_{\mathbf{x}}, \tilde{\mu}_{\alpha,\boldsymbol{a}} \rangle_{\mathcal{H}_K} \;=\; \sum_{m=1}^{M_\alpha} a_m\, K(\mathbf{x}, \mathbf{z}_m), \qquad \mathbf{x} \in \mathcal{Z}, \tag{C.132}$$

satisfies the uniform approximation bound

$$\|f - m\|_{\infty,\mathcal{Z}} \;\leq\; \varepsilon. \tag{C.133}$$

*Proof.* Fix an arbitrary function $f \in C(\mathcal{Z})$ together with $\varepsilon > 0$. Because $K$ is universal, Definition 4.4 guarantees the existence of a function $g_{\varepsilon/2} \in K(\mathcal{Z})$ such that

$$\|f - g_{\varepsilon/2}\|_{\infty,\mathcal{Z}} \;\leq\; \frac{\varepsilon}{2}. \tag{C.134}$$

By construction, the kernel section admits the representation

$$K(\mathcal{Z}) \;=\; \overline{\Big\{\sum_{i=1}^{n} a_i\, K(\,\cdot\,, \mathbf{z}_i) : n \in \mathbb{N},\; a_i \in \mathbb{R},\; \mathbf{z}_i \in \mathcal{Z}\Big\}}^{\|\cdot\|_{\infty,\mathcal{Z}}}. \tag{C.135}$$

Consequently, there exists a *finite* sum

$$g(\mathbf{x}) \;=\; \sum_{m=1}^{M_\alpha} a_m\, K\big(\mathbf{x}, \mathbf{z}_m\big), \qquad \mathbf{x} \in \mathcal{Z}, \tag{C.136}$$

such that

$$\|g_{\varepsilon/2} - g\|_{\infty,\mathcal{Z}} \;\leq\; \frac{\varepsilon}{2}. \tag{C.137}$$

Combining the two preceding inequalities yields

$$\|f - g\|_{\infty,\mathcal{Z}} \;\leq\; \|f - g_{\varepsilon/2}\|_{\infty,\mathcal{Z}} + \|g_{\varepsilon/2} - g\|_{\infty,\mathcal{Z}} \;\leq\; \frac{\varepsilon}{2} + \frac{\varepsilon}{2} \;=\; \varepsilon. \tag{C.138}$$

□

## C.4 Diversity and Generalization in Deep Neural Network Ensembles

**Theorem 5.1** [Return to statement]. For any distribution $\rho$ over $\boldsymbol{\Theta}$, and any of the three considered loss functions for ensembles, there exists a function $\mathbb{D}(\rho)$ such that

$$L(\rho) \leq \alpha(\mathbb{E}_\rho[L(\boldsymbol{\theta})] - \mathbb{D}(\rho)), \tag{C.139}$$

where $\alpha$ equals $1$ if the *sq*-loss or the *ce*-loss are considered, and $4$ for the $0/1$-loss. Furthermore, for the *sq*-loss, this inequality becomes an equality. The expression of the diversity measure for each of these loss functions is:

$$\mathbb{D}_{sq}(\rho) := \mathbb{E}_\nu\Big[\mathbb{V}_\rho(h_R(\mathbf{x};\boldsymbol{\theta}))\Big], \tag{C.140}$$

$$\mathbb{D}_{ce}(\rho) := \mathbb{E}_\nu\left[\mathbb{V}_\rho\left(\frac{P(y|\mathbf{x},\boldsymbol{\theta})}{\sqrt{2}\max_{\boldsymbol{\theta}} P(y|\mathbf{x},\boldsymbol{\theta})}\right)\right], \tag{C.141}$$

$$\mathbb{D}_{0/1}(\rho) := \mathbb{E}_\nu\Big[\mathbb{V}_\rho\Big(\mathbb{I}(h_W(\mathbf{x};\boldsymbol{\theta}) \neq y)\Big)\Big], \tag{C.142}$$

where $\mathbb{V}_\rho(\cdot)$ denotes the variance of a function w.r.t. $\rho$, and for $\mathbb{D}_{ce}(\rho)$ to be well-defined it must verify that $0 < \max_{\boldsymbol{\theta}\in\boldsymbol{\Theta}} P(y|\mathbf{x},\boldsymbol{\theta}) \leq 1$ for every $(\mathbf{x}, y) \in supp(\nu)$. Finally, note that all diversity terms described above can be written as

$$\mathbb{D}(\rho) = \mathbb{E}_\nu\Big[\mathbb{V}_\rho\left(f(y,\mathbf{x};\boldsymbol{\theta})\right)\Big], \tag{C.143}$$

with a specific function $f$ for each of the loss functions.

*Proof.* Using the fact that the variance of the classifiers can be decomposed as:

$$\mathbb{V}_\rho(f(\boldsymbol{\theta})) = \mathbb{E}_\rho[f(\boldsymbol{\theta})^2] - \mathbb{E}_\rho[f(\boldsymbol{\theta})]^2 = \mathbb{E}_\rho[f(\boldsymbol{\theta})^2] - \mathbb{E}_{\rho^2}[f(\boldsymbol{\theta})f(\boldsymbol{\theta}')] \tag{C.144}$$

$$= \mathbb{E}_{\rho^2}\Big[f(\boldsymbol{\theta})^2 - f(\boldsymbol{\theta})f(\boldsymbol{\theta}')\Big], \tag{C.145}$$

where $f$ is determined by the considered loss function: for the $sq$-loss, $f(\boldsymbol{\theta}) = h_R(\boldsymbol{x};\boldsymbol{\theta})$, the $ce$-loss, $f(\boldsymbol{\theta}) = p(y|\boldsymbol{x},\boldsymbol{\theta})$, and the $0/1$-loss $f(\boldsymbol{\theta}) = \mathbb{I}(h(\boldsymbol{x};\boldsymbol{\theta}) \neq y)$. The diversity terms $\mathbb{D}(\rho)$ defined in Theorem 5.1 can be written as,

$$\mathbb{D}_{sq}(\rho) = \mathbb{E}_{\rho^2}\Big[\mathbb{E}_\nu\Big[h_R(\boldsymbol{x};\boldsymbol{\theta})^2 - h_R(\boldsymbol{x};\boldsymbol{\theta})h_R(\boldsymbol{x};\boldsymbol{\theta}')\Big]\Big] \tag{C.146}$$

$$\mathbb{D}_{0/1}(\rho) = \mathbb{E}_{\rho^2}\Big[\mathbb{E}_\nu\Big[\mathbb{I}(h(\boldsymbol{x};\boldsymbol{\theta}) = y)\mathbb{I}(h(\boldsymbol{x};\boldsymbol{\theta}') \neq y)\Big]\Big] \tag{C.147}$$

$$\mathbb{D}_{ce}(\rho) = \mathbb{E}_{\rho^2}\left[\mathbb{E}_\nu\left[\frac{p(y|\boldsymbol{x},\boldsymbol{\theta})^2 - p(y \mid \boldsymbol{x},\boldsymbol{\theta})p(y|\boldsymbol{x},\boldsymbol{\theta}')}{2\max\limits_{\boldsymbol{\theta}\in\boldsymbol{\Theta}} p(y|\boldsymbol{x},\boldsymbol{\theta})^2}\right]\right] \tag{C.148}$$

where $\rho^2$ is a shorthand for the product distribution $\rho \times \rho$ over $\boldsymbol{\Theta} \times \boldsymbol{\Theta}$ and the shorthand $\mathbb{E}_{\rho^2}[f(\boldsymbol{\theta},\boldsymbol{\theta}')] = \mathbb{E}_{\boldsymbol{\theta}\sim\rho,\boldsymbol{\theta}'\sim\rho}[f(\boldsymbol{\theta},\boldsymbol{\theta}')]$. Recall the definition of the expected mean squared error of a regression ensemble reparameterized by the distribution $\rho$, that is,

$$L_{sq}(\rho) = \mathbb{E}_\nu\left[(y - h_R(\boldsymbol{x};\rho))^2\right], \tag{C.149}$$

where $h_R(\boldsymbol{x};\rho) = \mathbb{E}_\rho\left[h_R(\boldsymbol{x};\boldsymbol{\theta})\right]$ with $h_R(\boldsymbol{x};\boldsymbol{\theta})$ an individual regression model. The desired result can be obtained by expanding the square in each of the elements on the right-hand side of the equation, that is:

$$\mathbb{E}\left[L_{sq}(\boldsymbol{\theta})\right] = \mathbb{E}_{\rho,\nu}\left[(y - h_R(\boldsymbol{x};\boldsymbol{\theta}))^2\right] = \mathbb{E}_{\rho,\nu}\left[y^2 - 2yh_R(\boldsymbol{x};\boldsymbol{\theta}) + h_R(\boldsymbol{x};\boldsymbol{\theta})^2\right] \tag{C.150}$$

$$= \mathbb{E}_\nu\left[y^2 - 2yh_R(\boldsymbol{x};\rho) + \mathbb{E}_\rho[h_R(\boldsymbol{x};\boldsymbol{\theta})^2]\right], \tag{C.151}$$

where the fact that $y$ is constant under $\mathbb{E}_\rho$ is used. On the other hand,

$$\mathbb{D}_{sq}(\rho) = \mathbb{E}_{\nu,\rho}\left[(h_R(\boldsymbol{x},\boldsymbol{\theta}) - \mathbb{E}_\rho[h_R(\boldsymbol{x};\boldsymbol{\theta})])^2\right] = \mathbb{E}_{\nu,\rho}\left[(h_R(\boldsymbol{x},\boldsymbol{\theta}) - h_R(\boldsymbol{x};\rho))^2\right] \quad \text{(C.152)}$$

$$= \mathbb{E}_{\nu,\rho}\left[h_R(\boldsymbol{x},\boldsymbol{\theta})^2 - 2h_R(\boldsymbol{x},\boldsymbol{\theta})h_R(\boldsymbol{x},\rho) + h_R(\boldsymbol{x};\rho)^2\right] \quad \text{(C.153)}$$

$$= \mathbb{E}_{\nu}\left[\mathbb{E}_\rho[h_R(\boldsymbol{x},\boldsymbol{\theta})^2] - 2h_R(\boldsymbol{x},\rho)^2 + h_R(\boldsymbol{x};\rho)^2\right]. \quad \text{(C.154)}$$

Finally, subtracting both expressions:

$$\mathbb{E}_\rho[L_{sq}(\boldsymbol{\theta})] - \mathbb{D}_{sq}(\rho) = \mathbb{E}_\nu\left[y^2 - 2yh_R(\boldsymbol{x};\rho) + h_R(\boldsymbol{x};\rho)^2\right] \quad \text{(C.155)}$$

$$= \mathbb{E}_\nu\left[(y - h_R(\boldsymbol{x};\rho))^2\right] \quad \text{(C.156)}$$

$$= L_{sq}(\rho). \quad \text{(C.157)}$$

Continuing with the cross-entropy error, the use of Taylor's theorem with a remainder of second order over the logarithm function is needed. That is, given $\log x$ and a fixed value $a > 0$,

$$\log x = \log a + \frac{1}{a}(x-a) - \frac{1}{2\xi^2}(x-a)^2, \quad \xi \in (x,a). \quad \text{(C.158)}$$

Applying this to $p(y|\boldsymbol{x},\boldsymbol{\theta})$ centered at $\mathbb{E}_\rho[p(y|\boldsymbol{x},\boldsymbol{\theta})] > 0$,

$$\log p(y|\boldsymbol{x},\boldsymbol{\theta}) = \log \mathbb{E}_\rho[p(y|\boldsymbol{x},\boldsymbol{\theta})] + \frac{1}{\mathbb{E}_\rho[p(y|\boldsymbol{x},\boldsymbol{\theta})]}\left(p(y|\boldsymbol{x},\boldsymbol{\theta}) - \mathbb{E}_\rho[p(y|\boldsymbol{x},\boldsymbol{\theta})]\right) \quad \text{(C.159)}$$

$$- \frac{1}{2\xi^2}\left(p(y|\boldsymbol{x},\boldsymbol{\theta}) - \mathbb{E}_\rho[p(y|\boldsymbol{x},\boldsymbol{\theta})]\right)^2. \quad \text{(C.160)}$$

Taking expectation over $\rho$ at both sides,

$$\mathbb{E}_\rho[\log p(y|\boldsymbol{x},\boldsymbol{\theta})] = \log \mathbb{E}_\rho[p(y|\boldsymbol{x},\boldsymbol{\theta})] - \mathbb{E}_\rho\left[\frac{1}{2\xi^2}\left(p(y|\boldsymbol{x},\boldsymbol{\theta}) - \mathbb{E}_\rho[p(y|\boldsymbol{x},\boldsymbol{\theta})]\right)^2\right]. \quad \text{(C.161)}$$

Rearranging terms,

$$-\log \mathbb{E}_\rho[p(y|\boldsymbol{x},\boldsymbol{\theta})] = -\mathbb{E}_\rho[\log p(y|\boldsymbol{x},\boldsymbol{\theta})] - \mathbb{E}_\rho\left[\frac{1}{2\xi^2}\left(p(y|\boldsymbol{x},\boldsymbol{\theta}) - \mathbb{E}_\rho[p(y|\boldsymbol{x},\boldsymbol{\theta})]\right)^2\right]. \quad \text{(C.162)}$$

The desired inequality arises from the fact that $\xi$ is between $p(y|\boldsymbol{x},\boldsymbol{\theta})$ and $\mathbb{E}_\rho[p(y|\boldsymbol{x},\boldsymbol{\theta})]$, and hence, is upper bounded by $max_{\boldsymbol{\theta}\in\boldsymbol{\Theta}} p(y|\boldsymbol{x},\boldsymbol{\theta})$. Additionally, the square in the last term is always positive, implying the whole term is positive. Using these two properties,

$$-\log \mathbb{E}_\rho[p(y|\boldsymbol{x},\boldsymbol{\theta})] \leq -\mathbb{E}_\rho[\log p(y|\boldsymbol{x},\boldsymbol{\theta})] \quad \text{(C.163)}$$

$$- \mathbb{E}_\rho\left[\frac{1}{2\max_{\boldsymbol{\theta}} p(y|\boldsymbol{x},\boldsymbol{\theta})^2}\left(p(y|\boldsymbol{x},\boldsymbol{\theta}) - \mathbb{E}_\rho[p(y|\boldsymbol{x},\boldsymbol{\theta})]\right)^2\right]. \quad \text{(C.164)}$$

Finally, taking expectations w.r.t. $\nu$ on both sides raises the desired result. Lastly, consider the $0/1$-error. In order to prove this result, use Markov's inequality for monotonically increasing functions, in this case, for $\psi(a) = a^2$. That is, for a given random variable $X$,

$$\mathbb{P}(|x| \geq a) \leq \frac{\mathbb{E}[\psi(|x|)]}{\psi(a)} = \frac{\mathbb{E}[|x|^2]}{a^2}, \quad \text{for any } a > 0. \tag{C.165}$$

Applying this theorem to $\mathbb{E}_\rho\left[\mathbb{I}(h_W(\boldsymbol{x};\boldsymbol{\theta}) \neq y)\right]$, it verifies that

$$\mathbb{P}(\mathbb{E}_\rho\left[\mathbb{I}\left(h_W(\boldsymbol{x};\boldsymbol{\theta}) \neq y\right)\right] \geq 0.5) \leq 4\mathbb{E}_\nu\left[\mathbb{E}_\rho\left[\mathbb{I}\left(h_W(\boldsymbol{x};\boldsymbol{\theta}) \neq y\right)\right]^2\right]. \tag{C.166}$$

Notice that when majority vote makes an error, at least half ($\rho$-weighted) of the classifiers are wrong, that is,

$$\mathbb{I}(h_W(\boldsymbol{x};\rho) \neq y) \leq \mathbb{I}[\mathbb{E}_\rho[\mathbb{I}h_W(\boldsymbol{x};\boldsymbol{\theta}) \neq y] \geq 0.5], \tag{C.167}$$

which implies

$$\mathbb{E}_\nu\left[\mathbb{I}\left(h_W(\boldsymbol{x};\rho) \neq y\right)\right] \leq \mathbb{E}_\nu\left[\mathbb{I}\left(\mathbb{E}_\rho\left[\mathbb{I}\left(h_W(\boldsymbol{x};\boldsymbol{\theta}) \neq y\right)\right] \geq 0.5\right)\right] = \tag{C.168}$$

$$= \mathbb{P}\left(\mathbb{E}_\rho\left[\mathbb{I}\left(h_W(\boldsymbol{x};\boldsymbol{\theta}) \neq y\right)\right] \geq 0.5\right). \tag{C.169}$$

Using the derived inequality of the last term,

$$L_{0/1}(\rho) = \mathbb{E}_\nu\left[\mathbb{I}\left(h_W(\boldsymbol{x};\rho) \neq y\right)\right] \leq 4\mathbb{E}_\nu\left[\mathbb{E}_\rho\left[\mathbb{I}\left(h_W(\boldsymbol{x};\boldsymbol{\theta}) \neq y\right)\right]^2\right]. \tag{C.170}$$

To conclude the proof, it remains to show that the right-hand side of the inequality is the desired upper bound on the 0/1 loss. Using that $\mathrm{Var}(X) = \mathbb{E}\left[(X - \mathbb{E}[X])^2\right] = \mathbb{E}[X^2] - \mathbb{E}[X]^2$ and applying this identity to the indicator $X = \mathbb{I}(h_W(\boldsymbol{x};\boldsymbol{\theta}) \neq y)$, it follows that

$$\mathbb{D}_{0/1}(\rho)) = \mathbb{E}_\nu\left[\mathbb{E}_\rho\left[(\mathbb{I}(h_W(\boldsymbol{x};\boldsymbol{\theta}) \neq y) - \mathbb{E}_\rho\left[\mathbb{I}(h_W(\boldsymbol{x};\boldsymbol{\theta}) \neq y)\right])^2\right]\right]) \tag{C.171}$$

$$= \mathbb{E}_\nu\left[\mathbb{E}_\rho\left[\mathbb{I}(h_W(\boldsymbol{x};\boldsymbol{\theta}) \neq y)\right] - \mathbb{E}_\rho\left[\mathbb{I}(h_W(\boldsymbol{x};\boldsymbol{\theta}) \neq y)\right]^2\right] \tag{C.172}$$

$$= \mathbb{E}_\rho\Big[\mathbb{E}_\nu\left[\mathbb{I}(h_W(\boldsymbol{x};\boldsymbol{\theta}) \neq y)\right]\Big] - \mathbb{E}_\nu\left[\mathbb{E}_\rho\left[\mathbb{I}(h_W(\boldsymbol{x};\boldsymbol{\theta}) \neq y)\right]^2\right]. \tag{C.173}$$

Which implies that

$$4\left(\mathbb{E}_\rho[L_{0/1}(\boldsymbol{\theta})] - \mathbb{D}_{0/1}(\rho)\right) = 4\Big(\mathbb{E}_\rho\left[\mathbb{E}_\nu\left[\mathbb{I}(h_W(\boldsymbol{x};\boldsymbol{\theta}) \neq y)\right]\right] - \mathbb{D}_{0/1}(\rho)\Big) \tag{C.174}$$

$$= 4\mathbb{E}_\nu\left[\mathbb{E}_\rho\left[\mathbb{I}(h_W(\boldsymbol{x};\boldsymbol{\theta}) \neq y)\right]^2\right]. \tag{C.175}$$

□

**Theorem 5.2** [Return to statement]. Any distribution $\rho$ over $\boldsymbol{\Theta}$ satisfies the following inequality,

$$L_{ce}(\rho) \leq \mathbb{E}_\rho[L_{ce}(\boldsymbol{\theta})] - \mathbb{V}^T_{ce}(\rho), \tag{C.176}$$

where $\mathbb{V}_{ce}^{T}(\rho)$ is the normalized variance of $P(y|\mathbf{x},\boldsymbol{\theta})$ w.r.t. $\rho(\boldsymbol{\theta})$,

$$\mathbb{V}_{ce}^{T}(\rho) := \mathbb{E}_{\nu}\Big[h(m,\mu)\mathbb{E}_{\rho}\Big[(P(y|\mathbf{x},\boldsymbol{\theta}) - P(y))^2\Big]\Big]. \tag{C.177}$$

Where $\mu := \mathbb{E}_{\rho}[P(y|\mathbf{x},\boldsymbol{\theta})]$, $m := \max_{\boldsymbol{\theta}} P(y|\mathbf{x},\boldsymbol{\theta})$ and $h(m,\mu) := \frac{\ln\mu - \ln m}{(m-\mu)^2} + \frac{1}{\mu(m-\mu)}$.

*Proof.* Apply Liao and Berg, 2019's result to the random variable $p(\boldsymbol{x}|\boldsymbol{\theta})$, following the same strategy used in the proof of Theorem 5.1. □

**Lemma 5.3** [Return to statement]. The diversity terms $\mathbb{D}(\rho)$ defined in Theorem 5.1 satisfy the following properties:

1. If all the ensemble members provide the same predictions, or if $\rho$ outs all its probability mass on a single predictor, then $\mathbb{D}(\rho)$ is null.
2. $0 \leq \mathbb{D}(\rho) \leq \mathbb{E}_{\rho}[L(\boldsymbol{\theta})]$.
3. $\mathbb{D}(\rho)$ is invariant to reparametrizations.

*Proof.*
1. Notice that if all models make the same prediction, $h(\boldsymbol{x};\boldsymbol{\theta}) = \mathbb{E}_{\rho}[h(\boldsymbol{x};\boldsymbol{\theta})]$ and $p(y|\boldsymbol{x},\boldsymbol{\theta}) = \mathbb{E}_{\rho}[p(y|\boldsymbol{x},\boldsymbol{\theta})]$, which nullifies all diversity definitions from Theorem 5.1.
2. Using that every considered loss function and variance are positive, the given inequality is trivial.
3. Let $\phi : \boldsymbol{\Omega} \rightarrow \boldsymbol{\Theta}$ be an injective differentiable function with continuous partial derivatives, with non-zero Jacobian at any point. This result follows from the fact that all considered a probability distributions (ensembles) are a finite mixture of delta distributions, as a result, the distribution is compactly supported and the variable change theorem can be applied to a continuous function $f : \boldsymbol{\Theta} \rightarrow \mathbb{R}$:

$$\mathbb{E}_{\rho}[f(\boldsymbol{\theta})] = \int_{\boldsymbol{\Theta}} \rho(\boldsymbol{\theta}) f(\boldsymbol{\theta}) = \int_{\boldsymbol{\Omega}} \rho \circ \phi(\boldsymbol{\omega})\ f \circ \phi(\boldsymbol{\omega})\ |det(D\phi)(\boldsymbol{\omega})| \tag{C.178}$$

$$= \mathbb{E}_{\rho'}[f \circ \phi(\boldsymbol{\omega})] \tag{C.179}$$

where

$$\rho'(\boldsymbol{\omega}) = \rho \circ \phi(\boldsymbol{\omega})\ |det(D\phi)(\boldsymbol{\omega})|. \tag{C.180}$$

The result follows from taking $f(\boldsymbol{\theta}) = (\boldsymbol{\theta} - \mathbb{E}_{\rho}(\boldsymbol{\theta}))^2$.

□

**Theorem 5.4** [Return to statement]. The diversity terms $\mathbb{D}(\rho)$ defined in Theorem 5.1 can be written as

$$\mathbb{D}(\rho) = \mathbb{V}_{\nu\times\rho}(f(y,\mathbf{x};\boldsymbol{\theta})) - \mathbb{E}_{\rho\times\rho}[Cov_{\nu}(f(y,\mathbf{x};\boldsymbol{\theta}), f(y,\mathbf{x};\boldsymbol{\theta}'))], \tag{C.181}$$

where $\rho \times \rho$ denotes the joint distribution over $\boldsymbol{\theta} \times \boldsymbol{\theta}$, $\rho \times \nu$ denotes the joint distribution over $\boldsymbol{\theta} \times (\mathcal{X}, \mathcal{Y})$, and $Cov_\nu(\cdot, \cdot)$ is the covariance between two models with respect to the data generating distribution $\nu$.

*Proof.* This result can be easily shown as follows. Begin with the definition of variance:

$$\mathbb{D}(\rho) = \mathbb{E}_\nu\Big[\mathbb{V}_\rho[f(y, \boldsymbol{x}; \boldsymbol{\theta})]\Big] = \mathbb{E}_\nu\Big[\mathbb{E}_{\rho^2}[f(y, \boldsymbol{x}; \boldsymbol{\theta})^2 - f(y, \boldsymbol{x}; \boldsymbol{\theta}) f(y, \boldsymbol{x}; \boldsymbol{\theta}')]\Big]. \tag{C.182}$$

Split the expectation in two

$$\mathbb{D}(\rho) = \mathbb{E}_\nu\Big[\mathbb{V}_\rho[f(y, \boldsymbol{x}; \boldsymbol{\theta})]\Big] = \mathbb{E}_\nu\Big[\mathbb{E}_\rho[f(y, \boldsymbol{x}; \boldsymbol{\theta})^2] - \mathbb{E}_{\rho^2}[f(y, \boldsymbol{x}; \boldsymbol{\theta}) f(y, \boldsymbol{x}; \boldsymbol{\theta}')]\Big]. \tag{C.183}$$

On one hand, use the definition of variance again, getting that:

$$\mathbb{E}_\nu\Big[\mathbb{E}_\rho[f(y, \boldsymbol{x}; \boldsymbol{\theta})^2] = \mathbb{V}_{\nu \times \rho}[f(y, \boldsymbol{x}; \boldsymbol{\theta})] + \mathbb{E}_{\nu \times \rho}[f(y, \boldsymbol{x}; \boldsymbol{\theta})]^2. \tag{C.184}$$

On the other hand, the definition of covariance gives:

$$\mathbb{E}_{\rho \times \rho}\Big[Cov_\nu(f(y, \boldsymbol{x}; \boldsymbol{\theta}), f(y, \boldsymbol{x}; \boldsymbol{\theta}'))\Big] = \mathbb{E}_{\nu \times \rho^2}[f(y, \boldsymbol{x}; \boldsymbol{\theta}) f(y, \boldsymbol{x}; \boldsymbol{\theta}')]\Big] - \mathbb{E}_{\nu \times \rho}[f(y, \boldsymbol{x}; \boldsymbol{\theta})]^2. \tag{C.185}$$

Using both terms, the proof is completed:

$$\mathbb{D}(\rho) = \mathbb{E}_\nu\Big[\mathbb{V}_\rho[f(y, \boldsymbol{x}; \boldsymbol{\theta})]\Big] = \mathbb{V}_{\nu \times \rho}[f(y, \boldsymbol{x}; \boldsymbol{\theta})] - \mathbb{E}_{\rho \times \rho}\Big[Cov_\nu(f(y, \boldsymbol{x}; \boldsymbol{\theta}), f(y, \boldsymbol{x}; \boldsymbol{\theta}'))\Big]. \tag{C.186}$$

□

**Theorem 5.7** [Return to statement]. For any prior distribution $\pi$ over $\boldsymbol{\theta}$ independent of $D$, any $\xi \in (0, 1)$, and any $\lambda > 0$, with probability at least $1 - \xi$ over draws of training data $D \sim \nu^n$, for all distributions $\rho$ over $\boldsymbol{\theta}$, simultaneously,

$$L(\rho) \leq \alpha\left(\mathbb{E}_\rho[\hat{L}(\boldsymbol{\theta}, D)] - \hat{\mathbb{D}}(\rho, D) + \frac{2\mathrm{KL}(\rho|\pi) + \epsilon(\nu, \pi, \lambda, n, \xi)}{\lambda n}\right) \tag{C.187}$$

where $\alpha$ is equal to $1$ for the $sq$-loss or the $ce$-loss, and $4$ for the $0/1$-loss. Furthermore, $\epsilon$ is a positive function that is independent of $\rho$ but also depends on the specific loss (the functional forms of these $\epsilon$ terms can be found in the proof).

*Proof.* In order to prove this theorem, the r.h.s. term is shown to be an upper bound for $\alpha(\mathbb{E}_\rho[L(\boldsymbol{\theta})] - \mathbb{D}(\rho))$, which, using Theorem 5.1 concludes the proof. First of all, consider the following *tandem* losses:

$$L_{sq}(\boldsymbol{\theta},\boldsymbol{\theta}') = L_{sq}(\boldsymbol{\theta}) - \mathbb{E}_{\nu}\left[h_R(\boldsymbol{x};\boldsymbol{\theta})^2 - h_R(\boldsymbol{x};\boldsymbol{\theta})h_R(\boldsymbol{x};\boldsymbol{\theta}')\right], \quad (C.188)$$

$$L_{0/1}(\boldsymbol{\theta},\boldsymbol{\theta}') = L_{0/1}(\boldsymbol{\theta}) - \mathbb{E}_{\nu}\left[\mathbb{I}(h(\boldsymbol{x};\boldsymbol{\theta}) = y)\mathbb{I}(h(\boldsymbol{x};\boldsymbol{\theta}') \neq y)\right], \quad (C.189)$$

$$L_{ce}(\boldsymbol{\theta},\boldsymbol{\theta}') = L_{ce}(\boldsymbol{\theta}) - \mathbb{E}_{\nu}\left[\frac{p(y|\boldsymbol{x},\boldsymbol{\theta})^2 - p(y|\boldsymbol{x},\boldsymbol{\theta})p(y|\boldsymbol{x},\boldsymbol{\theta}')}{2\max_{\boldsymbol{\theta}\in\boldsymbol{\Theta}} p(y|\boldsymbol{x},\boldsymbol{\theta})^2}\right], \quad (C.190)$$

which verify

$$\mathbb{E}_{\rho^2}[L(\boldsymbol{\theta},\boldsymbol{\theta}')] = \mathbb{E}_{\rho}[L(\boldsymbol{\theta})] - \mathbb{D}(\rho). \quad (C.191)$$

This can be shown using the fact that the variance of the classifiers can be decomposed as

$$\mathrm{Var}_{\rho}(f(\boldsymbol{\theta})) = \mathbb{E}_{\rho}\left[f(\boldsymbol{\theta})^2\right] - \mathbb{E}_{\rho}\left[f(\boldsymbol{\theta})\right]^2 = \mathbb{E}_{\rho}\left[f(\boldsymbol{\theta})^2\right] - \mathbb{E}_{\rho^2}\left[f(\boldsymbol{\theta})f(\boldsymbol{\theta}')\right] \quad (C.192)$$

$$= \mathbb{E}_{\rho^2}\left[f(\boldsymbol{\theta})^2 - f(\boldsymbol{\theta})f(\boldsymbol{\theta}')\right], \quad (C.193)$$

where the function $f$ is determined by the loss under consideration: for the squared loss, $f(\boldsymbol{\theta}) = h_R(\mathbf{x};\boldsymbol{\theta})$; for the cross-entropy loss, $f(\boldsymbol{\theta}) = p(y|\mathbf{x},\boldsymbol{\theta})$; and for the $0/1$-loss, $f(\boldsymbol{\theta}) = \mathbb{I}(h(\mathbf{x};\theta) \neq y)$.

Applying Germain et al. (2016, Theorem 3) to the tandem loss functionals described above with a prior distribution $\pi(\boldsymbol{\theta},\boldsymbol{\theta}') = \pi(\boldsymbol{\theta})\pi(\boldsymbol{\theta}')$ raises that for any $\lambda n > 0$ and $\delta \in (0,1]$, with probability at least $1-\delta$:

$$\mathbb{E}_{\rho}[L(\boldsymbol{\theta})] - \mathbb{D}(\rho) \leq \mathbb{E}_{\rho(\boldsymbol{\theta},\boldsymbol{\theta}')}[\hat{L}(\boldsymbol{\theta},\boldsymbol{\theta}')] + \frac{1}{\lambda n}\left[\mathrm{KL}(\rho(\boldsymbol{\theta},\boldsymbol{\theta}')|\pi(\boldsymbol{\theta},\boldsymbol{\theta}')) + \epsilon(\nu,\pi,\lambda,n,\xi)\right]. \quad (C.194)$$

Where

$$\epsilon(\nu,\pi,\lambda,n,\xi) := \log \mathbb{E}_{\pi(\boldsymbol{\theta},\boldsymbol{\theta}')}\left[\mathbb{E}_{\nu}\left[\exp\left(\lambda\left(L(\boldsymbol{\theta},\boldsymbol{\theta}') - \hat{L}(\boldsymbol{\theta},\boldsymbol{\theta}',D)\right)\right)\right]\right] + \log\frac{1}{\delta}, \quad (C.195)$$

and $\mathrm{KL}(\rho(\boldsymbol{\theta},\boldsymbol{\theta}')|\pi(\boldsymbol{\theta},\boldsymbol{\theta}')) = 2\mathrm{KL}(\rho|\pi)$. In short:

$$L(\rho) \leq \alpha\left(\mathbb{E}_{\rho}[\hat{L}(\boldsymbol{\theta},D)] - \hat{\mathbb{D}}(\rho,D) + \frac{1}{\lambda n}\left[2KL(\rho|\pi) + \epsilon(\nu,\pi,\lambda,n,\xi)\right]\right). \quad (C.196)$$

□

## C.5 Generalization Error and Chernoff Bounds

**Proposition 5.12** [Return to statement]. Under Assumption 5.9, $\forall\boldsymbol{\theta} \in \boldsymbol{\Theta}$, $\mathcal{I}_{\boldsymbol{\theta}}(\cdot)$ and $\mathcal{I}_{\boldsymbol{\theta}}^{-1}(\cdot)$, are well defined. That is, $\forall a \in [0, L(\boldsymbol{\theta}) - m_{\boldsymbol{\theta}})$, $\mathcal{I}_{\boldsymbol{\theta}}(a) < \infty$ and $\forall s \in \mathbb{R}_0^+$, $\mathcal{I}_{\boldsymbol{\theta}}^{-1}(s) < \infty$.

*Proof.* First, from Assumption 5.9, it verifies that $m_{\boldsymbol{\theta}} \geq 0$, then $\ell(\mathbf{y}, \mathbf{x}, \boldsymbol{\theta}) \geq 0 \; \forall (\mathbf{x}, \mathbf{y}) \in \mathcal{X} \times \mathcal{Y}$. Then, $\lambda(L(\boldsymbol{\theta}) - \ell(\mathbf{y}, \mathbf{x}, \boldsymbol{\theta}))) \leq \lambda L(\boldsymbol{\theta})$. Taking exponential and expectations,

$$\mathbb{E}_\nu \left[ e^{\lambda(L(\boldsymbol{\theta}) - \ell(\mathbf{y},\mathbf{x},\boldsymbol{\theta}))} \right] \leq \mathbb{E}_\nu \left[ e^{\lambda L(\boldsymbol{\theta})} \right] . \tag{C.197}$$

Lastly, the expectation on the r.h.s. is constant, leading to

$$J_{\boldsymbol{\theta}}(\lambda) = \ln \mathbb{E}_\nu \left[ e^{\lambda(L(\boldsymbol{\theta}) - \ell(\mathbf{y},\mathbf{x},\boldsymbol{\theta}))} \right] \leq \ln \mathbb{E}_\nu \left[ e^{\lambda L(\boldsymbol{\theta})} \right] = \lambda L(\boldsymbol{\theta}) . \tag{C.198}$$

As, by Assumption 5.9, $\forall \boldsymbol{\theta} \in \boldsymbol{\Theta}$, $L(\boldsymbol{\theta}) < \infty$ and the function $J_{\boldsymbol{\theta}}(\lambda)$ is well-defined for $\lambda > 0$. It is left to show that the supremum over $\lambda$ is reached in the definition of the rate function. For that, it will be shown that it is actually a maximum. Firstly,

$$\frac{\partial}{\partial \lambda}(\lambda a - J_{\boldsymbol{\theta}}(\lambda)) = a - \frac{\partial}{\partial \lambda} J_{\boldsymbol{\theta}}(\lambda) , \tag{C.199}$$

where the second derivative is negative as $\frac{\partial^2}{\partial \lambda^2} J_{\boldsymbol{\theta}}(\lambda) \geq 0$ (the cumulant is convex). As a result, when the previous derivative is zero, the maximum is reached. In fact the optimum of $\lambda$ is a $\lambda^\star$ such that $a = \frac{\partial}{\partial \lambda} J_{\boldsymbol{\theta}}(\lambda^\star)$. Then, $\forall a \in (0, L(\boldsymbol{\theta}) - m_{\boldsymbol{\theta}})$, it is necessary to show $\exists \lambda^\star \in \mathbb{R}^+$. It verifies that $\frac{\partial}{\partial \lambda} J_{\boldsymbol{\theta}}(\lambda)$ is a continuous function, as it is combination of continuous functions:

$$\frac{\partial}{\partial \lambda} J_{\boldsymbol{\theta}}(\lambda) = \frac{\mathbb{E}_\nu[p(\mathbf{y}|\mathbf{x}, \boldsymbol{\theta})^\lambda \ln p(\mathbf{y}|\mathbf{x}, \boldsymbol{\theta})]}{\mathbb{E}_\nu[p(\mathbf{y}|\mathbf{x}, \boldsymbol{\theta})^\lambda]} - \mathbb{E}_\nu[\ln p(\mathbf{y}|\mathbf{x}, \boldsymbol{\theta})] . \tag{C.200}$$

By standard properties of the cumulant generating function, $\frac{\partial}{\partial \lambda} J_{\boldsymbol{\theta}}(0) = 0$. On the other had, also by standard properties of the cummulant Herdegen, 2008, Lemma 1,

$$\lim_{\lambda \to \infty} \frac{\partial}{\partial \lambda} J_{\boldsymbol{\theta}}(\lambda) = ess \sup_{(\mathbf{x},\mathbf{y})} \; L(\boldsymbol{\theta}) - \ell(\mathbf{y}, \mathbf{x}, \boldsymbol{\theta}) = L(\boldsymbol{\theta}) - m_{\boldsymbol{\theta}} . \tag{C.201}$$

Thus, if $\frac{\partial}{\partial \lambda} J_{\boldsymbol{\theta}}(\lambda)$ is continuous, $\frac{\partial}{\partial \lambda} J_{\boldsymbol{\theta}}(0) = 0$ and $\lim_{\lambda \to \infty} \frac{\partial}{\partial \lambda} J_{\boldsymbol{\theta}}(\lambda) = L(\boldsymbol{\theta}) - m_{\boldsymbol{\theta}}$, then $\forall a \in [0, L(\boldsymbol{\theta}) - m_{\boldsymbol{\theta}})$ there always exist a $\lambda^\star \in \mathbb{R}^+$ such that $a = \frac{\partial}{\partial \lambda} J_{\boldsymbol{\theta}}(\lambda^\star)$. Thus, for such values of $a$ the rate function is finite and well defined.

For $a \geq L(\boldsymbol{\theta}) - m_{\boldsymbol{\theta}}$, it verifies that the supremum is reached when $\lambda \to \infty$, because $J_{\boldsymbol{\theta}}(\lambda)$ is monotonically increasing. Every $a \geq L(\boldsymbol{\theta}) - m_{\boldsymbol{\theta}}$, can be written as $a = L(\boldsymbol{\theta}) - b$, where $b \leq m_{\boldsymbol{\theta}}$. Then the limit when $\lambda \to \infty$ for any $a \geq L(\boldsymbol{\theta}) - m_{\boldsymbol{\theta}}$ can be written as follows,

$$\lim_{\lambda\to\infty}\lambda(L(\boldsymbol{\theta})-b)-J_{\boldsymbol{\theta}}(\lambda)=\lim_{\lambda\to\infty}-\lambda b-\ln\mathbb{E}_\nu\left[p(\mathbf{y}|\mathbf{x},\boldsymbol{\theta})^\lambda\right]=\lim_{\lambda\to\infty}-\ln\mathbb{E}_\nu\left[\left(\frac{p(\mathbf{y}|\mathbf{x},\boldsymbol{\theta})}{e^{-b}}\right)^\lambda\right] \tag{C.202}$$

$$=-\ln\mathbb{E}_\nu\left[\lim_{\lambda\to\infty}\left(\frac{p(\mathbf{y}|\mathbf{x},\boldsymbol{\theta})}{e^{-b}}\right)^\lambda\right]=-\ln\mathbb{E}_\nu\left[\mathbb{I}(p(\mathbf{y}|\mathbf{x},\boldsymbol{\theta})=e^{-b}\right] \tag{C.203}$$

$$=-\ln\mathbb{E}_\nu[\mathbb{I}(\ell(\mathbf{y},\mathbf{x},\boldsymbol{\theta})=b)]=-\ln\mathbb{P}_\nu(\ell(\mathbf{y},\mathbf{x},\boldsymbol{\theta})=b)\,. \tag{C.204}$$

If $b<m_{\boldsymbol{\theta}}$ or, equivalently, $a>L(\boldsymbol{\theta})-m_{\boldsymbol{\theta}}$, then $\mathcal{I}_{\boldsymbol{\theta}}(a)=\infty$. Moreover, if $a=L-m_{\boldsymbol{\theta}}$, then $\mathcal{I}_{\boldsymbol{\theta}}(a)=-\ln\mathbb{P}_\nu(\ell(\mathbf{y},\mathbf{x},\boldsymbol{\theta})=m_{\boldsymbol{\theta}})$, which may be well-defined or equal to $\infty$. By definition of the inverse rate function,

$$\mathcal{I}_{\boldsymbol{\theta}}^{-1}(s)=\inf_{\lambda\geq 0}\frac{s+J_{\boldsymbol{\theta}}(\lambda)}{\lambda}\leq\lim_{\lambda\to\infty}\frac{s+J_{\boldsymbol{\theta}}(\lambda)}{\lambda}=\lim_{\lambda\to\infty}\frac{J_{\boldsymbol{\theta}}(\lambda)}{\lambda} \tag{C.205}$$

Then, by L'Hôpital's rule and Equation Equation (C.201), it verifies that

$$\lim_{\lambda\to\infty}\frac{J_{\boldsymbol{\theta}}(\lambda)}{\lambda}=\lim_{\lambda\to\infty}\nabla_\lambda J_{\boldsymbol{\theta}}(\lambda)=L(\boldsymbol{\theta})-m_{\boldsymbol{\theta}} \tag{C.206}$$

From Assumption 5.9, $L(\boldsymbol{\theta})<\infty$ and $m_{\boldsymbol{\theta}}\geq 0$, then, $\forall s\geq 0$, $\mathcal{I}_{\boldsymbol{\theta}}^{-1}(s)<\infty$. □

**Theorem 5.14** [Return to statement]. [Chernoff, 1952] For any fixed $\boldsymbol{\theta}\in\boldsymbol{\Theta}$ and $a>0$, it satisfies

$$\mathbb{P}_{D\sim\nu^n}\left(L(\boldsymbol{\theta})-\hat{L}(D,\boldsymbol{\theta})\geq a\right)\leq e^{-n\mathcal{I}_{\boldsymbol{\theta}}(a)}\,. \tag{C.207}$$

*Proof.* Cramér-Chernoff's bound states that for any random variable $X$, it verifies that $P(X\geq s)\leq\inf_{t>0}\mathbb{E}[e^{tX}]e^{-ts}$. Applying this result to the random variable over possible datasets $\hat{L}(D,\boldsymbol{\theta})-L(\boldsymbol{\theta})$, for a fixed $\boldsymbol{\theta}\in\boldsymbol{\Theta}$, leads to

$$P(L(\boldsymbol{\theta})-\hat{L}(D,\boldsymbol{\theta})\geq s)\leq\inf_{t>0}\mathbb{E}\left[e^{t(L(\boldsymbol{\theta})-\hat{L}(D,\boldsymbol{\theta}))}\right]e^{-ts}\,. \tag{C.208}$$

The expectation in the r.h.s. can be transformed as

$$\mathbb{E}\left[e^{t(L(\boldsymbol{\theta})-\hat{L}(D,\boldsymbol{\theta}))}\right]=\mathbb{E}\left[e^{t(\frac{1}{n}\ln p(D|\boldsymbol{\theta})-\mathbb{E}_\nu[\ln p(\mathbf{y}|\mathbf{x},\boldsymbol{\theta})]}\right]=\mathbb{E}\left[e^{\frac{t}{n}\ln p(D|\boldsymbol{\theta})}\right]e^{-t\mathbb{E}_\nu[\ln p(\mathbf{y}|\mathbf{x},\boldsymbol{\theta})]}\,. \tag{C.209}$$

Moreover, the first expectation in this last term is

$$\mathbb{E}\left[e^{\frac{t}{n}\ln p(D|\boldsymbol{\theta})}\right]=\mathbb{E}_{\nu^n}\left[p(D|\boldsymbol{\theta})^{\frac{t}{n}}\right]=\mathbb{E}_\nu\left[P(\mathbf{y}|\mathbf{x},\boldsymbol{\theta})^{\frac{t}{n}}\right]^n=\mathbb{E}\left[e^{\frac{t}{n}\ln p(\mathbf{y}|\mathbf{x},\boldsymbol{\theta})}\right]^n\,. \tag{C.210}$$

Using this, and parameterizing $t$ as $\lambda n$, with $\lambda > 0$,

$$P(L(\boldsymbol{\theta}) - \hat{L}(D, \boldsymbol{\theta}) \geq s) \leq \inf_{\lambda > 0} \mathbb{E}\left[e^{\lambda(L(\boldsymbol{\theta}) - \ell(\mathbf{y}, \mathbf{x}, \boldsymbol{\theta}))}\right]^n e^{-\lambda n s}. \tag{C.211}$$

Taking exponential and logarithm on the r.h.s, and using the definition of the smoothness function $J_{\boldsymbol{\theta}}(\lambda)$ and the rate function $\mathcal{I}_{\boldsymbol{\theta}}(a)$:

$$P(L(\boldsymbol{\theta}) - \hat{L}(D, \boldsymbol{\theta}) \geq s) \leq \inf_{\lambda > 0} e^{n \ln \mathbb{E}\left[e^{\lambda(L(\boldsymbol{\theta}) - \ell(\mathbf{y}, \mathbf{x}, \boldsymbol{\theta}))}\right] - \lambda n s} \leq \inf_{\lambda > 0} e^{n J_{\boldsymbol{\theta}}(\lambda) - \lambda n s} = e^{-n \mathcal{I}_{\boldsymbol{\theta}}(s)}. \tag{C.212}$$

□

**Proposition 5.15** [Return to statement]. For any fixed $\boldsymbol{\theta} \in \boldsymbol{\Theta}$ and $n > 0$, it satisfies

$$\lim_{a \to L(\boldsymbol{\theta}) - m_{\boldsymbol{\theta}}} \mathbb{P}_{D \sim \nu^n}\Big(L(\boldsymbol{\theta}) - \hat{L}(D, \boldsymbol{\theta}) \geq a\Big) = \lim_{a \to L(\boldsymbol{\theta}) - m_{\boldsymbol{\theta}}} e^{-n \mathcal{I}_{\boldsymbol{\theta}}(a)}. \tag{C.213}$$

*Proof.* It is clear that the limit on both sides is

$$\mathbb{P}_{D \sim \nu_n}(\hat{L}(D, \boldsymbol{\theta}) = m_{\boldsymbol{\theta}}) = n \mathbb{P}_{(\mathbf{y}, \mathbf{x}) \sim \nu}(\ell(\mathbf{y}, \mathbf{x}, \boldsymbol{\theta}) = m_{\boldsymbol{\theta}}). \tag{C.214}$$

□

**Theorem 5.16** [Return to statement]. [Cramér, 1938; Ellis, 2012] For any fixed $\boldsymbol{\theta} \in \boldsymbol{\Theta}$ and any $a > 0$, it satisfies

$$\lim_{n \to \infty} -\frac{1}{n} \log \mathbb{P}_{D \sim \nu^n}\Big(L(\boldsymbol{\theta}) - \hat{L}(D, \boldsymbol{\theta}) \geq a\Big) = \mathcal{I}_{\boldsymbol{\theta}}(a). \tag{C.215}$$

*Proof.* Direct application of Cramér's Theorem (Cramér, 1938; Ellis, 2012), over the random variable $X = L(\boldsymbol{\theta}) - L(D, \boldsymbol{\theta})$, for a fixed $\boldsymbol{\theta}$, where the randomness comes from $D \sim \nu^n$. Similar to the application of Chernoff's bound in Theorem 5.14. □

**Theorem 5.17** [Return to statement]. [**PAC-Chernoff Bound**] With h.p. $1 - \delta$ over $D \sim \nu^n$, for all $\boldsymbol{\theta} \in \boldsymbol{\Theta}$, simultaneously,

$$L(\boldsymbol{\theta}) \leq \hat{L}(D, \boldsymbol{\theta}) + \mathcal{I}_{\boldsymbol{\theta}}^{-1}(\tfrac{1}{n} \log \tfrac{k^p}{\delta}). \tag{C.216}$$

*Proof.* By Chernoff's Theorem 5.14, for a given $\boldsymbol{\theta}$, it verifies that $\mathbb{P}\Big(L(\boldsymbol{\theta}) - \hat{L}(D, \boldsymbol{\theta}) \geq a\Big) \leq e^{-n \mathcal{I}_{\boldsymbol{\theta}}(a)}$. Naming $\delta' = e^{-n \mathcal{I}_{\boldsymbol{\theta}}(a)}$ and re-arranging terms, $a = \mathcal{I}_{\boldsymbol{\theta}}^{-1}(-\frac{1}{n} \ln \delta')$. As a result:

$$\mathbb{P}\Big(L(\boldsymbol{\theta}) - \hat{L}(D, \boldsymbol{\theta}) \geq \mathcal{I}_{\boldsymbol{\theta}}^{-1}(\tfrac{1}{n} \ln \tfrac{1}{\delta'})\Big) \leq \delta'. \tag{C.217}$$

Taking the complementary probability and using the union bound over the set of models,

$$\mathbb{P}\Big(\bigcup_{\boldsymbol{\theta}\in\boldsymbol{\Theta}} L(\boldsymbol{\theta}) - \hat{L}(D,\boldsymbol{\theta}) \geq \mathcal{I}_{\boldsymbol{\theta}}^{-1}(\tfrac{1}{n}\ln\tfrac{1}{\delta'})\Big) \leq \sum_{\boldsymbol{\theta}\in\boldsymbol{\Theta}} \mathbb{P}\Big(L(\boldsymbol{\theta}) - \hat{L}(D,\boldsymbol{\theta}) \geq \mathcal{I}_{\boldsymbol{\theta}}^{-1}(\tfrac{1}{n}\ln\tfrac{1}{\delta'})\Big)\,. \tag{C.218}$$

Given that the model space considers $k^p$ different models, the r.h.s. can be rewritten as

$$\mathbb{P}\Big(\bigcup_{\boldsymbol{\theta}\in\boldsymbol{\Theta}} L(\boldsymbol{\theta}) - \hat{L}(D,\boldsymbol{\theta}) \geq \mathcal{I}_{\boldsymbol{\theta}}^{-1}(\tfrac{1}{n}\ln\tfrac{1}{\delta'})\Big) \leq k^p\delta'\,. \tag{C.219}$$

By reparameterizing the above inequality with $\delta' = \delta k^{-p}$:

$$\mathbb{P}\Big(\bigcup_{\boldsymbol{\theta}\in\boldsymbol{\Theta}} L(\boldsymbol{\theta}) - \hat{L}(D,\boldsymbol{\theta}) \geq \mathcal{I}_{\boldsymbol{\theta}}^{-1}(\tfrac{1}{n}\ln\tfrac{k^p}{\delta})\Big) \leq \delta\,. \tag{C.220}$$

Which verifies,

$$1-\mathbb{P}\Big(\bigcup_{\boldsymbol{\theta}\in\boldsymbol{\Theta}} L(\boldsymbol{\theta}) - \hat{L}(D,\boldsymbol{\theta}) \geq \mathcal{I}_{\boldsymbol{\theta}}^{-1}(\tfrac{1}{n}\ln\tfrac{k^p}{\delta})\Big) \geq 1-\delta\,. \tag{C.221}$$

Which is equivalent to,

$$\mathbb{P}\Big(\bigcap_{\boldsymbol{\theta}\in\boldsymbol{\Theta}} L(\boldsymbol{\theta}) - \hat{L}(D,\boldsymbol{\theta}) \leq \mathcal{I}_{\boldsymbol{\theta}}^{-1}(\tfrac{1}{n}\ln\tfrac{k^p}{\delta})\Big) \geq 1-\delta\,. \tag{C.222}$$

□

**Corollary 5.18** [Return to statement]. With h.p. $1-\delta$ over $D \sim \nu^n$, for all $\boldsymbol{\theta} \in \boldsymbol{\Theta}$, simultaneously,

$$L(\boldsymbol{\theta}) \leq \hat{L}(D,\boldsymbol{\theta}) + \mathcal{I}_{\boldsymbol{\theta}}^{-1}(\tfrac{1}{n}\log\tfrac{|\bar{\boldsymbol{\Theta}}|}{\delta})\,. \tag{C.223}$$

*Proof.* The proof follows the same approach as Theorem 5.17, with the union bound applied to the models in $\bar{\boldsymbol{\Theta}}$ rather than directly to the models in $\boldsymbol{\Theta}$. This adjustment is valid because, when applying the union bound, only the total number of distinct random variables needs to be considered; in this case, $L(D,\boldsymbol{\theta})$ with $D \sim \nu^n$. Models in $\boldsymbol{\Theta}$ that are not in $\bar{\boldsymbol{\Theta}}$ define random variables that are duplicated by definition and therefore do not need to be accounted for in the application of the union bound. □

**Theorem 5.19** [Return to statement]. For any $\delta \in (0,1)$ and any $\boldsymbol{\theta} \in \boldsymbol{\Theta}$, it verifies that

$$\lim_{n\to\infty} \sqrt{n}\,\mathcal{I}_{\boldsymbol{\theta}}^{-1}(\tfrac{1}{n}\log\tfrac{k^p}{\delta}) = \sqrt{2\mathbb{V}_\nu(\ell(\mathbf{y},\mathbf{x},\boldsymbol{\theta}))\log\tfrac{k^p}{\delta}}\,. \tag{C.224}$$

*Proof.* Let $\bar{Z}_n := \sqrt{n}(L(\boldsymbol{\theta}) - L(D,\boldsymbol{\theta}))$ and $J_{\bar{Z}_n}, \mathcal{I}_{\bar{Z}_n}$ and $\mathcal{I}_{\bar{Z}_n}^{-1}$ denote its cumulant, rate and inverse rate function. Then, by properties of the cumulant, it verifies that $J_{\bar{Z}_n}(\lambda) =$

$nJ_{\boldsymbol{\theta}}(\lambda/\sqrt{n})$. Using the definition of the rate function:

$$\mathcal{I}_{\bar{Z}_n}(a) = \sup_{\lambda}\; a\lambda + nJ_{\boldsymbol{\theta}}(\lambda/\sqrt{n}) = n\sup_{\lambda}\; \frac{\lambda}{\sqrt{n}}\frac{a}{\sqrt{n}} + J_{\boldsymbol{\theta}}(\lambda/\sqrt{n}) = n\mathcal{I}_{\boldsymbol{\theta}}(a/\sqrt{n})\,. \quad \text{(C.225)}$$

Using a second order Taylor expansion over $\mathcal{I}_{\boldsymbol{\theta}}(a)$ around $a = 0$, where $\mathcal{I}_{\boldsymbol{\theta}}(0) = 0$ and $\frac{\partial}{\partial a}\mathcal{I}_{\boldsymbol{\theta}}(0) = 0$:

$$\mathcal{I}_{\boldsymbol{\theta}}(a/\sqrt{n}) = \frac{1}{2}\frac{\partial^2}{\partial a^2}\mathcal{I}_{\boldsymbol{\theta}}(0)\left(\frac{a}{\sqrt{n}}\right)^2 + \left(\frac{a}{\sqrt{n}}\right)^2 h(a/\sqrt{n})\,, \quad \text{(C.226)}$$

where $\lim_{t\to 0} h(t) = 0$. Furthermore, it verifies that (Ellis, 2012)

$$\frac{\partial^2}{\partial a^2}\mathcal{I}_{\boldsymbol{\theta}}(0) = \left(\frac{\partial^2}{\partial \lambda^2}J_{\boldsymbol{\theta}}(0)\right)^{-1}. \quad \text{(C.227)}$$

From there:

$$\begin{aligned}
\frac{\partial^2}{\partial \lambda^2}J_{\boldsymbol{\theta}}(\lambda) &= -\frac{\partial}{\partial \lambda}\frac{\mathbb{E}_\nu[e^{-\lambda\ell(\mathbf{y},\mathbf{x},\boldsymbol{\theta})}\ell(\mathbf{y},\mathbf{x},\boldsymbol{\theta})]}{\mathbb{E}_\nu[e^{-\lambda\ell(\mathbf{y},\mathbf{x},\boldsymbol{\theta})}]} && \text{(C.228)}\\
&= \frac{\mathbb{E}_\nu[e^{-\lambda\ell(\mathbf{y},\mathbf{x},\boldsymbol{\theta})}(\ell(\mathbf{y},\mathbf{x},\boldsymbol{\theta}))^2]}{\mathbb{E}_\nu[e^{-\lambda\ell(\mathbf{y},\mathbf{x},\boldsymbol{\theta})}]} - \frac{\mathbb{E}_\nu[e^{-\lambda\ell(\mathbf{y},\mathbf{x},\boldsymbol{\theta})}\ell(\mathbf{y},\mathbf{x},\boldsymbol{\theta})]^2}{\mathbb{E}_\nu[e^{-\lambda\ell(\mathbf{y},\mathbf{x},\boldsymbol{\theta})}]^2}. && \text{(C.229)}
\end{aligned}$$

Which evaluated at $\lambda = 0$, gives

$$\frac{\partial^2}{\partial \lambda^2}J_{\boldsymbol{\theta}}(0) = \mathbb{V}_\nu(\ell(\mathbf{y},\mathbf{x},\boldsymbol{\theta}))\,. \quad \text{(C.230)}$$

Let this last quantity be written as $\sigma^2 := \mathbb{V}_\nu(\ell(\mathbf{y},\mathbf{x},\boldsymbol{\theta}))$. Then,

$$\begin{aligned}
\lim_{n\to\infty}\mathcal{I}_{\bar{Z}_n}(a) = \lim_{n\to\infty} n\mathcal{I}_{\boldsymbol{\theta}}(a/\sqrt{n}) &= \lim_{n\to\infty} n\frac{1}{2}\sigma^{-2}\left(\frac{a}{\sqrt{n}}\right)^2 + n\left(\frac{a}{\sqrt{n}}\right)^2 h(a/\sqrt{n}) && \text{(C.231)}\\
&= \frac{1}{2}\sigma^{-2}a^2\,. && \text{(C.232)}
\end{aligned}$$

Then, as both the rate and its inverse are continuous functions, it verifies that

$$\lim_{n\to\infty}\mathcal{I}^{-1}_{\bar{Z}_n}(s) = \left(\lim_{n\to\infty}\mathcal{I}_{\bar{Z}_n}\right)^{-1}(s) = \sqrt{2\sigma^2 s} \quad \text{(C.233)}$$

By definition of the inverse rate:

$$\begin{aligned}
\mathcal{I}^{-1}_{\bar{Z}_n}(s) &= \inf_{\lambda}\frac{s + J_{\bar{Z}_n}(\lambda)}{\lambda} = \inf_{\lambda}\frac{s + nJ_{\boldsymbol{\theta}}(\lambda/\sqrt{n})}{\lambda} && \text{(C.234)}\\
&= \sqrt{n}\inf_{\lambda}\frac{\frac{s}{n} + nJ_{\boldsymbol{\theta}}(\lambda/\sqrt{n})}{\frac{\lambda}{\sqrt{n}}} = \sqrt{n}\mathcal{I}^{-1}_{\boldsymbol{\theta}}(s/n)\,. && \text{(C.235)}
\end{aligned}$$

As a result,

$$\lim_{n\to\infty}\sqrt{n}\mathcal{I}^{-1}_{\boldsymbol{\theta}}(s/n) = \lim_{n\to\infty}\mathcal{I}^{-1}_{\bar{Z}_n}(s) = \sqrt{2\sigma^2 s}\,. \quad \text{(C.236)}$$

□

**Proposition 5.20** [Return to statement]. With h.p. $1-\delta$ over $D \sim \nu^n$, for all $\boldsymbol{\theta} \in \boldsymbol{\Theta}$, simultaneously,

$$\text{if} \quad \hat{L}(D,\boldsymbol{\theta}) \leq \epsilon \quad \text{then} \quad 0 \leq L(\boldsymbol{\theta}) - \mathcal{I}_{\boldsymbol{\theta}}^{-1}(\tfrac{1}{n}\log\tfrac{k_P}{\delta}) \leq \epsilon\,. \tag{C.237}$$

*Proof.* By Theorem 5.17, $L(\boldsymbol{\theta}) \leq L(D,\boldsymbol{\theta}) + \mathcal{I}_{\boldsymbol{\theta}}^{-1}(\frac{1}{n}\ln\frac{k_P}{\delta})$, and, by Proposition 5.12, $\mathcal{I}_{\boldsymbol{\theta}}(a)$ is well defined $\forall a \in [0, L(\boldsymbol{\theta}) - m_{\boldsymbol{\theta}})$; and, $\forall b > 0$, it verifies that $\mathcal{I}_{\boldsymbol{\theta}}^{-1}(b) \in [0, L(\boldsymbol{\theta}) - m_{\boldsymbol{\theta}})$. In consequence:

$$L(\boldsymbol{\theta}) \leq L(D,\boldsymbol{\theta}) + \mathcal{I}_{\boldsymbol{\theta}}^{-1}(\tfrac{1}{n}\ln\tfrac{k_P}{\delta}) \leq L(D,\boldsymbol{\theta}) + L(\boldsymbol{\theta}) - m_{\boldsymbol{\theta}}\,. \tag{C.238}$$

The result follows from $m_{\boldsymbol{\theta}} \geq 0$, $L(D,\boldsymbol{\theta}) \leq \epsilon$, and rearranging terms. □

**Proposition 5.22** [Return to statement]. For any $\boldsymbol{\theta} \in \boldsymbol{\Theta}$ and $\boldsymbol{\theta}' \in \boldsymbol{\Theta}'$,

$$\exists \beta > 0 \text{ s.t. } \boldsymbol{\theta} \text{ is } \beta\text{-smoother than } \boldsymbol{\theta}' \iff \mathbb{V}_\nu(\ell(\mathbf{y},\mathbf{x},\boldsymbol{\theta})) \leq \mathbb{V}_\nu(\ell(\mathbf{y},\mathbf{x},\boldsymbol{\theta}'))\,. \tag{C.239}$$

*Proof.* Let $\bar{Z}_n := \sqrt{n}(L(\boldsymbol{\theta}) - L(D,\boldsymbol{\theta}))$ and $J_{\bar{Z}_n}, \mathcal{I}_{\bar{Z}_n}$ and $\mathcal{I}_{\bar{Z}_n}^{-1}$ denote its cumulant, rate, and inverse rate function. Then, by properties of the cumulant, it verifies that $J_{\bar{Z}_n}(\lambda) = nJ_{\boldsymbol{\theta}}(\frac{\lambda}{\sqrt{n}})$. Using the definition of the rate function:

$$\mathcal{I}_{\bar{Z}_n}(a) = \sup_\lambda\; a\lambda + nJ_{\boldsymbol{\theta}}(\tfrac{\lambda}{\sqrt{n}}) = n\sup_\lambda\; \tfrac{\lambda}{\sqrt{n}}\tfrac{a}{\sqrt{n}} + J_{\boldsymbol{\theta}}(\tfrac{\lambda}{\sqrt{n}}) = n\mathcal{I}_{\boldsymbol{\theta}}(\tfrac{a}{\sqrt{n}})\,. \tag{C.240}$$

Using a second order Taylor expansion over $\mathcal{I}_{\boldsymbol{\theta}}(a)$ around $a=0$, where $\mathcal{I}_{\boldsymbol{\theta}}(0) = 0$ and $\frac{\partial}{\partial a}\mathcal{I}_{\boldsymbol{\theta}}(0) = 0$, it verifies that

$$\mathcal{I}_{\boldsymbol{\theta}}(\tfrac{a}{\sqrt{n}}) = \tfrac{1}{2}\tfrac{\partial^2}{\partial a^2}\mathcal{I}_{\boldsymbol{\theta}}(0)(\tfrac{a}{\sqrt{n}})^2 + (\tfrac{a}{\sqrt{n}})^2 h(\tfrac{a}{\sqrt{n}})\,, \tag{C.241}$$

where $\lim_{t\to 0} h(t) = 0$. Furthermore, it verifies that

$$\tfrac{\partial^2}{\partial a^2}\mathcal{I}_{\boldsymbol{\theta}}(0) = \left(\tfrac{\partial^2}{\partial\lambda^2}J_{\boldsymbol{\theta}}(0)\right)^{-1} = \mathbb{V}_\nu(\ell(\mathbf{y},\mathbf{x},\boldsymbol{\theta}))^{-1} =: \sigma^{-2}\,. \tag{C.242}$$

Then,

$$\lim_{n\to\infty}\mathcal{I}_{\bar{Z}_n}(a) = \lim_{n\to\infty} n\mathcal{I}_{\boldsymbol{\theta}}(\tfrac{a}{\sqrt{n}}) = \lim_{n\to\infty} n\frac{1}{2}\sigma^{-2}(\tfrac{a}{\sqrt{n}})^2 + n(\tfrac{a}{\sqrt{n}})^2 h(\tfrac{a}{\sqrt{n}}) = \frac{1}{2}\sigma^{-2}a^2\,. \tag{C.243}$$

Let us assume that $\mathbb{V}_\nu(\ell(\mathbf{y},\mathbf{x},\boldsymbol{\theta})) \leq \mathbb{V}_\nu(\ell(\mathbf{y},\mathbf{x},\boldsymbol{\theta}'))$. Then, the aim is to show that there exists $\beta > 0$ such that $\mathcal{I}_{\boldsymbol{\theta}}(a) \geq \mathcal{I}_{\boldsymbol{\theta}'}(a) \quad \forall a \leq \beta$. For a fixed value of $a$, by the limit in Equation C.243, it verifies that for any $\epsilon > 0$, there exists $n_0(a), n_0'(a) > 0$ such that

$$|n\mathcal{I}_{\boldsymbol{\theta}}(\tfrac{a}{\sqrt{n}}) - \tfrac{1}{2}\mathbb{V}_\nu(\ell(\mathbf{y},\mathbf{x},\boldsymbol{\theta}))^{-2}a^2| < \epsilon \quad \forall n > n_0(a)\,, \tag{C.244}$$

and

$$\left|n\mathcal{I}_{\boldsymbol{\theta}'}(\tfrac{a}{\sqrt{n}}) - \tfrac{1}{2}\mathbb{V}_\nu(\ell(\mathbf{y},\mathbf{x},\boldsymbol{\theta}'))^{-2}a^2\right| < \epsilon \quad \forall n > n_0'(a)\,. \tag{C.245}$$

Let $\epsilon = \mathbb{V}_\nu(\ell(\mathbf{y},\mathbf{x},\boldsymbol{\theta}))^{-2}a^2 - \mathbb{V}_\nu(\ell(\mathbf{y},\mathbf{x},\boldsymbol{\theta}'))^{-2}a^2 \geq 0$ and $n_0^\star(a) = \max\{n_0(a), n_0'(a)\}$. Then,

$$\mathcal{I}_{\boldsymbol{\theta}}(\tfrac{a}{\sqrt{n}}) \geq \mathcal{I}_{\boldsymbol{\theta}'}(\tfrac{a}{\sqrt{n}}) \quad \forall n > n_0^\star(a)\,. \tag{C.246}$$

This implies that $\mathcal{I}_{\boldsymbol{\theta}}(c) \geq \mathcal{I}_{\boldsymbol{\theta}'}(c)$, for any $c \leq a/\sqrt{n_0^\star(a)}$, as there exists $n = (a/c)^2$ verifying $c = a/\sqrt{n}$. Consider then $\beta := \sup_{a>0}\{a/\sqrt{n_0^\star(a)}\} > 0$, which might be infinite. It verifies that

$$\mathcal{I}_{\boldsymbol{\theta}}(a) \geq \mathcal{I}_{\boldsymbol{\theta}'}(a) \quad \forall a \leq \beta\,. \tag{C.247}$$

Let us now assume that there exists $\beta > 0$ such that $\boldsymbol{\theta}$ is $\beta$-smoother than $\boldsymbol{\theta}'$, or, equivalently $\mathcal{I}_{\boldsymbol{\theta}}(a) \geq \mathcal{I}_{\boldsymbol{\theta}'}(a) \quad \forall a \leq \beta$. Then,

$$\lim_{n\to\infty} n\mathcal{I}_{\boldsymbol{\theta}}(\tfrac{a}{\sqrt{n}}) \geq \lim_{n\to\infty} n\mathcal{I}_{\boldsymbol{\theta}'}(\tfrac{a}{\sqrt{n}})\,. \tag{C.248}$$

Using the limit in Equation C.243,

$$\mathbb{V}_\nu(\ell(\mathbf{y},\mathbf{x},\boldsymbol{\theta}))^{-2}a^2 \geq \mathbb{V}_\nu(\ell(\mathbf{y},\mathbf{x},\boldsymbol{\theta}'))^{-2}a^2\,, \tag{C.249}$$

leading to

$$\mathbb{V}_\nu(\ell(\mathbf{y},\mathbf{x},\boldsymbol{\theta})) \leq \mathbb{V}_\nu(\ell(\mathbf{y},\mathbf{x},\boldsymbol{\theta}'))\,. \tag{C.250}$$

□

**Theorem 5.23** [Return to statement]. For any $\epsilon \geq 0$, with h.p. $1-\delta$ over $D \sim \nu^n$, for all $\boldsymbol{\theta} \in \boldsymbol{\Theta} \subset \mathbb{R}^p$ and $\boldsymbol{\theta}' \in \boldsymbol{\Theta}'$, simultaneously,

$$\text{if } \hat{L}(D,\boldsymbol{\theta}) \leq \epsilon \text{ and } \boldsymbol{\theta} \text{ is } \mathcal{I}_{\boldsymbol{\theta}}^{-1}(\tfrac{1}{n}\log\tfrac{kp}{\delta})\text{-smoother than } \boldsymbol{\theta}'\text{, then, } L(\boldsymbol{\theta}) \leq L(\boldsymbol{\theta}') + \epsilon\,. \tag{C.251}$$

*Proof.* If $\boldsymbol{\theta}$ is $\beta$-smoother than $\boldsymbol{\theta}'$, by Definition 5.21, $\forall a \in (0,\beta] \quad \mathcal{I}_{\boldsymbol{\theta}}(a) \geq \mathcal{I}_{\boldsymbol{\theta}'}(a)$, where $\beta = \mathcal{I}_{\boldsymbol{\theta}}^{-1}(\frac{1}{n}\ln\frac{kp}{\delta})$. Then,

$$\mathcal{I}_{\boldsymbol{\theta}}^{-1}(\tfrac{1}{n}\ln\tfrac{kp}{\delta}) \leq \mathcal{I}_{\boldsymbol{\theta}'}^{-1}(\tfrac{1}{n}\ln\tfrac{kp}{\delta})\,. \tag{C.252}$$

As the rate function $\mathcal{I}_{\boldsymbol{\theta}'}(a)$ is invertible and its image lies in $[0, L(\boldsymbol{\theta}')-m_{\boldsymbol{\theta}'})$, where $m_{\boldsymbol{\theta}'} \geq 0$, due to Assumption 5.9, it verifies that $\mathcal{I}_{\boldsymbol{\theta}'}^{-1}(s) \in [0, L(\boldsymbol{\theta}') - m_{\boldsymbol{\theta}'})$. In consequence,

$$\mathcal{I}_{\boldsymbol{\theta}'}^{-1}(\tfrac{1}{n}\ln\tfrac{kp}{\delta}) \leq L(\boldsymbol{\theta}')\,. \tag{C.253}$$

As $\mathcal{I}_{\boldsymbol{\theta}}^{-1}(\frac{1}{n}\ln\frac{kp}{\delta}) \leq \mathcal{I}_{\boldsymbol{\theta}'}^{-1}(\frac{1}{n}\ln\frac{kp}{\delta})$, it verifies $\mathcal{I}_{\boldsymbol{\theta}}^{-1}(\frac{1}{n}\ln\frac{kp}{\delta}) \leq L(\boldsymbol{\theta}')$. By the PAC-Chernoff bound of Theorem 5.17 and because $L(D,\boldsymbol{\theta}) \leq \epsilon$, with h.p. $1-\delta$ over $D \sim \nu^n$,

$$L(\boldsymbol{\theta}) \leq L(D,\boldsymbol{\theta}) + \mathcal{I}_{\boldsymbol{\theta}}^{-1}(\tfrac{1}{n}\ln\tfrac{kp}{\delta}) \leq \epsilon + \mathcal{I}_{\boldsymbol{\theta}}^{-1}(\tfrac{1}{n}\ln\tfrac{kp}{\delta})\,. \tag{C.254}$$

Combining the last two inequalities,

$$L(\boldsymbol{\theta}) \leq \epsilon + \mathcal{I}_{\boldsymbol{\theta}}^{-1}(\tfrac{1}{n}\ln\tfrac{kp}{\delta}) \leq \epsilon + L(\boldsymbol{\theta}')\,. \tag{C.255}$$

The statement of the theorem directly derives from the above inequality. □

**Theorem 5.26** [Return to statement]. For any $\epsilon > 0$, with h.p. $1-\delta$ over $D \sim \nu^n$, $|L(\boldsymbol{\theta}^\star_\epsilon) - L(\boldsymbol{\theta}^\times_\epsilon)| \leq \epsilon$.

*Proof.* From the definitions of $\boldsymbol{\theta}^\times_\epsilon$ and $\boldsymbol{\theta}^\star_\epsilon$, given by

$$\boldsymbol{\theta}^\star_\epsilon = \underset{\boldsymbol{\theta}\,:\,L(D,\boldsymbol{\theta})\leq\epsilon}{\arg\min}\; L(\boldsymbol{\theta})\,, \qquad \boldsymbol{\theta}^\times_\epsilon = \underset{\boldsymbol{\theta}\,:\,L(D,\boldsymbol{\theta})\leq\epsilon}{\arg\min}\; L(D,\boldsymbol{\theta}) + \mathcal{I}_{\boldsymbol{\theta}}^{-1}(\tfrac{1}{n}\ln\tfrac{kp}{\delta})\,, \tag{C.256}$$

it is clear that $L(\boldsymbol{\theta}^\star_\epsilon) \leq L(\boldsymbol{\theta}^\times_\epsilon)$. On the other hand, because for any $s \geq 0$, $\mathcal{I}_{\boldsymbol{\theta}}^{-1}(s) \in [0, L(\boldsymbol{\theta}) - m_{\boldsymbol{\theta}})$ (Proposition 5.13), it verifies that $\forall \boldsymbol{\theta} \in \boldsymbol{\Theta}$:

$$\mathcal{I}_{\boldsymbol{\theta}}^{-1}(\tfrac{1}{n}\ln\tfrac{kp}{\delta}) + L(D,\boldsymbol{\theta}) \leq L(\boldsymbol{\theta}) + L(D,\boldsymbol{\theta})\,. \tag{C.257}$$

By definition of $\boldsymbol{\theta}^\times_\epsilon$ and $\boldsymbol{\theta}^\star_\epsilon$,

$$\mathcal{I}_{\boldsymbol{\theta}^\times_\epsilon}^{-1}(\tfrac{1}{n}\ln\tfrac{kp}{\delta}) + \hat{L}(D,\boldsymbol{\theta}^\times_\epsilon) \leq \mathcal{I}_{\boldsymbol{\theta}^\star_\epsilon}^{-1}(\tfrac{1}{n}\ln\tfrac{kp}{\delta}) + \hat{L}(D,\boldsymbol{\theta}^\star_\epsilon)\,, \tag{C.258}$$

which gives

$$\mathcal{I}_{\boldsymbol{\theta}^\times_\epsilon}^{-1}(\tfrac{1}{n}\ln\tfrac{kp}{\delta}) + \hat{L}(D,\boldsymbol{\theta}^\times_\epsilon) \leq L(\boldsymbol{\theta}^\star_\epsilon) + \hat{L}(D,\boldsymbol{\theta}^\star_\epsilon) \leq L(\boldsymbol{\theta}^\star_\epsilon) + \epsilon\,. \tag{C.259}$$

This, in combination with the PAC-Chernoff bound of Theorem 5.17 gives

$$L(\boldsymbol{\theta}^\times_\epsilon) \leq \mathcal{I}_{\boldsymbol{\theta}^\times_\epsilon}^{-1}(\tfrac{1}{n}\ln\tfrac{kp}{\delta}) + \hat{L}(D,\boldsymbol{\theta}^\times_\epsilon) \leq L(\boldsymbol{\theta}^\star_\epsilon) + \epsilon\,. \tag{C.260}$$

From this, $L(\boldsymbol{\theta}^\times_\epsilon) \leq L(\boldsymbol{\theta}^\star_\epsilon) + \epsilon$. Thus, $L(\boldsymbol{\theta}^\star_\epsilon) \leq L(\boldsymbol{\theta}^\times_\epsilon)$ and $L(\boldsymbol{\theta}^\times_\epsilon) \leq L(\boldsymbol{\theta}^\star_\epsilon) + \epsilon$, finishing the proof. □

**Proposition 5.27** [Return to statement]. If the loss function $\ell(\mathbf{y},\mathbf{x},\boldsymbol{\theta})$ is Lipschitz w.r.t. $\boldsymbol{\theta}$ with constant $M > 0$, then, for any $\boldsymbol{\theta}_0 \in \boldsymbol{\Theta}_0 = \{\boldsymbol{\theta} \in \boldsymbol{\Theta} \mid \mathbb{V}_\nu(\ell(\mathbf{y},\mathbf{x},\boldsymbol{\theta})) = 0\}$, it verifies that

$$\mathcal{I}_{\boldsymbol{\theta}}^{-1}(\tfrac{1}{n}\log\tfrac{kp}{\delta}) \leq \sqrt{2Ma}\,\|\boldsymbol{\theta}-\boldsymbol{\theta}_0\|_2\,, \tag{C.261}$$

where $a = min(1, \frac{1}{n}\log\frac{kp}{\delta})$.

*Proof.* If the loss is Lipschitz continuous with constant $M$, $\forall y, x, \boldsymbol{\theta}\quad \|\nabla_{\boldsymbol{\theta}}\ell(\mathbf{y},\mathbf{x},\boldsymbol{\theta})\|_2^2 \leq$

$M$. Then, $J_{\boldsymbol{\theta}}(\lambda)$ verifies

$$\|\nabla_{\boldsymbol{\theta}} J_{\boldsymbol{\theta}}(\lambda)\|_2^2 = || - \lambda \mathbb{E}_{\nu p^\lambda} \left[\nabla_{\boldsymbol{\theta}} \ell(\mathbf{y}, \mathbf{x}, \boldsymbol{\theta})\right] + \lambda \mathbb{E}_\nu \left[\nabla_{\boldsymbol{\theta}} \ell(\mathbf{y}, \mathbf{x}, \boldsymbol{\theta})\right] ||_2^2 \tag{C.262}$$

$$\leq \lambda^2 \mathbb{E}_{\nu p^\lambda} \left[\|\nabla_{\boldsymbol{\theta}} \ell(\mathbf{y}, \mathbf{x}, \boldsymbol{\theta})\|_2^2\right] + \lambda^2 \mathbb{E}_\nu \left[\|\nabla_{\boldsymbol{\theta}} \ell(\mathbf{y}, \mathbf{x}, \boldsymbol{\theta})\|_2^2\right] \tag{C.263}$$

$$\leq 2M\lambda^2 , \tag{C.264}$$

where $\mathbb{E}_{\nu p^\lambda} \left[\nabla_{\boldsymbol{\theta}} \ell(\mathbf{y}, \mathbf{x}, \boldsymbol{\theta})\right] = \frac{\mathbb{E}_\nu [p(\mathbf{y}|\mathbf{x},\boldsymbol{\theta})^\lambda \ell(\mathbf{y},\mathbf{x},\boldsymbol{\theta})]}{\mathbb{E}_\nu [p(\mathbf{y}|\mathbf{x},\boldsymbol{\theta})^\lambda]}$. With this,

$$|J_{\boldsymbol{\theta}}(\lambda) - J_{\boldsymbol{\theta}_0}(\lambda)| \leq 2M\lambda^2 \|\boldsymbol{\theta} - \boldsymbol{\theta}_0\|_2^2 \implies J_{\boldsymbol{\theta}}(\lambda) \leq 2M\lambda^2 \|\boldsymbol{\theta} - \boldsymbol{\theta}_0\|_2^2 . \tag{C.265}$$

Then, for any $a \geq 0$, it verifies $\frac{a + J_{\boldsymbol{\theta}}(\lambda)}{\lambda} \leq \frac{a + 2M\lambda^2 \|\boldsymbol{\theta} - \boldsymbol{\theta}_0\|_2^2}{\lambda}$; where by definition of the inverse rate,

$$\mathcal{I}_{\boldsymbol{\theta}}^{-1}(a) \leq \frac{a + J_{\boldsymbol{\theta}}(\lambda)}{\lambda} \leq \frac{a + 2M\lambda^2 \|\boldsymbol{\theta} - \boldsymbol{\theta}_0\|_2^2}{\lambda} . \tag{C.266}$$

As the inequality holds for any $\lambda \geq 0$, take the one that minimizes the r.h.s, leading to $\mathcal{I}_{\boldsymbol{\theta}}^{-1}(a) \leq \sqrt{2Ma} \|\boldsymbol{\theta} - \boldsymbol{\theta}_0\|_2$. On the other hand, $\mathcal{I}_{\boldsymbol{\theta}}^{-1}(a)$ is Lipschitz with constant $M$ as

$$\|\nabla_{\boldsymbol{\theta}} \mathcal{I}_{\boldsymbol{\theta}}^{-1}(a)\|_2^2 = \left\|\frac{\nabla_{\boldsymbol{\theta}} J_{\boldsymbol{\theta}}(\lambda_a^\star)}{\lambda_a^\star}\right\|_2^2 \leq \frac{\|\nabla_{\boldsymbol{\theta}} J_{\boldsymbol{\theta}}(\lambda_a^\star)\|_2^2}{\lambda_a^{\star,2}} \leq 2M . \tag{C.267}$$

In consequence, $(\mathcal{I}_{\boldsymbol{\theta}}^{-1}(\frac{1}{n} \ln \frac{k^p}{\delta}) - \mathcal{I}_{\boldsymbol{\theta}_0}^{-1}(\frac{1}{n} \ln \frac{k^p}{\delta}))^2 \leq 2M||\boldsymbol{\theta} - \boldsymbol{\theta}_0||^2$. Which implies that $\mathcal{I}_{\boldsymbol{\theta}}^{-1}(\frac{1}{n} \ln \frac{k^p}{\delta}) \leq \sqrt{2M}||\boldsymbol{\theta} - \boldsymbol{\theta}_0||_2$. Thus, it simultaneously holds that

$$\mathcal{I}_{\boldsymbol{\theta}}^{-1}(\tfrac{1}{n} \ln \tfrac{k^p}{\delta}) \leq \sqrt{2M}||\boldsymbol{\theta} - \boldsymbol{\theta}_0||_2 \quad \text{and} \quad \mathcal{I}_{\boldsymbol{\theta}}^{-1}(\tfrac{1}{n} \ln \tfrac{k^p}{\delta}) \leq \sqrt{2M \tfrac{1}{n} \ln \tfrac{k^p}{\delta}} \|\boldsymbol{\theta} - \boldsymbol{\theta}_0\|_2 . \tag{C.268}$$

□

**Corollary 5.28** [Return to statement]. Considering the log-loss, if it belongs to the exponential family with a constant base measure, that is, there exist $s : \mathcal{X} \times \mathcal{Y} \to \mathbb{R}^p$, $a : \boldsymbol{\Theta} \to \mathbb{R}$ and $k \in \mathbb{R}$ such that

$$\ell(\mathbf{y}, \mathbf{x}, \boldsymbol{\theta}) := -\log p(\mathbf{y}|\mathbf{x}, \boldsymbol{\theta}) = \boldsymbol{\theta}^T s(\mathbf{x}, \mathbf{y}) - a(\boldsymbol{\theta}) + k \quad \forall (\mathbf{x}, \mathbf{y}) \sim \nu . \tag{C.269}$$

Then, $\forall \epsilon > 0$ exists $n_0 > 0$ such that $\forall n > n_0$:

$$\left|\mathcal{I}_{\boldsymbol{\theta}}^{-1}(\tfrac{1}{n} \log \tfrac{k^p}{\delta}) - \sqrt{2 \tfrac{1}{n} \log \tfrac{k^p}{\delta}} \sqrt{\boldsymbol{\theta}^T \text{Cov}_\nu(s(\mathbf{y}, \mathbf{x}))\boldsymbol{\theta}}\right| \leq \epsilon , \tag{C.270}$$

where $\text{Cov}_\nu(\cdot)$ is the covariance w.r.t. $\nu$ of the sufficient statistics of each $(\mathbf{x}, \mathbf{y})$ sample.

*Proof.* Using Theorem 5.19, we got that $\lim_{n\to\infty} \sqrt{n} \mathcal{I}_{\boldsymbol{\theta}}^{-1}(\frac{1}{n} \ln \frac{k^p}{\delta}) = \sqrt{2\mathbb{V}_\nu(\ell(\mathbf{y}, \mathbf{x}, \boldsymbol{\theta})) \ln \frac{k^p}{\delta}}$. The proof concludes from the fact that using the exponential family,

$$\mathbb{V}_\nu(\ell(\mathbf{y}, \mathbf{x}, \boldsymbol{\theta})) = \mathbb{V}_\nu(\boldsymbol{\theta}^T s(\mathbf{y}, \mathbf{x}) - a(\boldsymbol{\theta}) + k) = \boldsymbol{\theta}^T \text{Cov}_\nu(s(\mathbf{y}, \mathbf{x}))\boldsymbol{\theta} . \tag{C.271}$$

□

**Proposition 5.29** [Return to statement]. For any $\boldsymbol{\theta} \in \boldsymbol{\Theta}$, if the equality of Equation (5.79) holds, then

$$\mathcal{I}_{\boldsymbol{\theta}}^{-1}(\tfrac{1}{n}\log\tfrac{k^p}{\delta}) = \sqrt{2\tfrac{1}{n}\log\tfrac{k^p}{\delta}}\sqrt{(\boldsymbol{\theta}-\boldsymbol{\theta}_0)^T\mathrm{Cov}_\nu(\nabla_{\boldsymbol{\theta}}\log p(\mathbf{y}|\mathbf{x},\boldsymbol{\theta}_0))(\boldsymbol{\theta}-\boldsymbol{\theta}_0)}\,. \tag{C.272}$$

*Proof.* A second-order Taylor expansion of $J_{\boldsymbol{\theta}}(\lambda)$ w.r.t. to $\boldsymbol{\theta}$ centered around $\boldsymbol{\theta}_0$ is

$$J_{\boldsymbol{\theta}_0}(\lambda) + \nabla_{\boldsymbol{\theta}} J_{\boldsymbol{\theta}_0}(\lambda)(\boldsymbol{\theta}-\boldsymbol{\theta}_0) + \frac{1}{2}(\boldsymbol{\theta}-\boldsymbol{\theta}_0)^T\nabla_{\boldsymbol{\theta}\boldsymbol{\theta}} J_{\boldsymbol{\theta}_0}(\lambda)(\boldsymbol{\theta}-\boldsymbol{\theta}_0)\,. \tag{C.273}$$

By standard properties of the cumulant generating function over centered random variables, it verifies that $J_{\boldsymbol{\theta}_0}(\lambda) = 0$. While the $\nabla_{\boldsymbol{\theta}} J_{\boldsymbol{\theta}}(\lambda)$ can be expressed as,

$$\nabla_{\boldsymbol{\theta}} J_{\boldsymbol{\theta}}(\lambda) = \lambda\mathbb{E}_{\nu p^\lambda}\left[\nabla_{\boldsymbol{\theta}}\ln p(\mathbf{y}|\mathbf{x},\boldsymbol{\theta})\right] - \lambda\mathbb{E}_{\nu}\left[\nabla_{\boldsymbol{\theta}}\ln p(\mathbf{y}|\mathbf{x},\boldsymbol{\theta})\right]\,, \tag{C.274}$$

where $\nu p^\lambda$ denotes $\nu p^\lambda(\mathbf{y},\mathbf{x}) = \frac{\nu(\mathbf{y},\mathbf{x})p(\mathbf{y}|\mathbf{x},\boldsymbol{\theta})^\lambda}{\mathbb{E}_\nu[p(\mathbf{y}|\mathbf{x},\boldsymbol{\theta})^\lambda]}$. At $\boldsymbol{\theta}_0$, it verifies that $\nu p^\lambda = \nu$, because, by definition, $p(\mathbf{y}|\mathbf{x},\boldsymbol{\theta}_0)$ is constant for any $(\mathbf{y},\mathbf{x})$. In consequence, the gradient at $\boldsymbol{\theta}_0$ simplifies as,

$$\nabla_{\boldsymbol{\theta}} J_{\boldsymbol{\theta}_0}(\lambda) = \lambda\mathbb{E}_{\nu}\left[\nabla_{\boldsymbol{\theta}}\ln p(\mathbf{y}|\mathbf{x},\boldsymbol{\theta})\right] - \lambda\mathbb{E}_{\nu}\left[\nabla_{\boldsymbol{\theta}}\ln p(\mathbf{y}|\mathbf{x},\boldsymbol{\theta})\right] = 0\,. \tag{C.275}$$

The Hessian of the cumulant $J_{\boldsymbol{\theta}}(\lambda)$ w.r.t. to $\boldsymbol{\theta}$ can be written as follows:

$$\nabla_{\boldsymbol{\theta}\boldsymbol{\theta}} J_{\boldsymbol{\theta}}(\lambda) = \lambda^2 Cov_{\nu p^\lambda}(\nabla_{\boldsymbol{\theta}}\ln p(\mathbf{y}|\mathbf{x},\boldsymbol{\theta})) + \lambda\mathbb{E}_{\nu p^\lambda}\left[\nabla_{\boldsymbol{\theta}\boldsymbol{\theta}}\ln p(\mathbf{y}|\mathbf{x},\boldsymbol{\theta})\right] \tag{C.276}$$

$$- \lambda\mathbb{E}_{\nu}\left[\nabla_{\boldsymbol{\theta}\boldsymbol{\theta}}\ln p(\mathbf{y}|\mathbf{x},\boldsymbol{\theta})\right]\,. \tag{C.277}$$

Again, at $\boldsymbol{\theta}_0$, it verifies that $\nu p^\lambda = \nu$, so the Hessian of the cumulant $J_{\boldsymbol{\theta}}(\lambda)$ at $\boldsymbol{\theta}_0$ simplifies as, $\nabla_{\boldsymbol{\theta}\boldsymbol{\theta}} J_{\boldsymbol{\theta}_0}(\lambda) = \lambda^2 Cov_\nu(\nabla_{\boldsymbol{\theta}}\ln p(\mathbf{y}|\mathbf{x},\boldsymbol{\theta}_0))$. With this, the second order Taylor expansion of $J_{\boldsymbol{\theta}}(\lambda)$ evaluated on $\boldsymbol{\theta}_0$ is $\frac{\lambda^2}{2}(\boldsymbol{\theta}-\boldsymbol{\theta}_0)^T\mathrm{Cov}_\nu(\nabla_{\boldsymbol{\theta}}\ln p(\mathbf{y}|\mathbf{x},\boldsymbol{\theta}_0))(\boldsymbol{\theta}-\boldsymbol{\theta}_0)$.

On the other hand, if, at the definition of $\mathcal{I}_{\boldsymbol{\theta}}^{-1}(s)$ given in Equation Equation (5.55), replace $J_{\boldsymbol{\theta}}(\lambda)$ by the above Taylor expansion, leading to

$$\mathcal{I}_{\boldsymbol{\theta}}^{-1}(s) = \inf_{\lambda>0}\ \frac{s}{\lambda} + \frac{\lambda}{2}(\boldsymbol{\theta}-\boldsymbol{\theta}_0)^T\mathrm{Cov}_\nu(\nabla_{\boldsymbol{\theta}}\ln p(\mathbf{y}|\mathbf{x},\boldsymbol{\theta}_0))(\boldsymbol{\theta}-\boldsymbol{\theta}_0)\,. \tag{C.278}$$

The optimal value of $\lambda$ is acquired taking derivatives on the above expression w.r.t. $\lambda$, giving,

$$\frac{-s}{\lambda^2} + \frac{1}{2}(\boldsymbol{\theta}-\boldsymbol{\theta}_0)^T\mathrm{Cov}_\nu(\nabla_{\boldsymbol{\theta}}\ln p(\mathbf{y}|\mathbf{x},\boldsymbol{\theta}_0))(\boldsymbol{\theta}-\boldsymbol{\theta}_0) = 0\,. \tag{C.279}$$

One may show that it is a minimum by taking the second derivative. From this, the optimal value of $\lambda$ is $\lambda^\star = \left(\frac{2s}{(\boldsymbol{\theta}-\boldsymbol{\theta}_0)^T \mathrm{Cov}_\nu(\nabla_{\boldsymbol{\theta}} \ln p(\mathbf{y}|\mathbf{x},\boldsymbol{\theta}_0))(\boldsymbol{\theta}-\boldsymbol{\theta}_0)}\right)^{\frac{1}{2}}$. Using this expression,

$$\mathcal{I}_{\boldsymbol{\theta}}^{-1}(s) = 2s\sqrt{\frac{1}{2s}(\boldsymbol{\theta}-\boldsymbol{\theta}_0)^T \mathrm{Cov}_\nu(\nabla_{\boldsymbol{\theta}} \ln p(\mathbf{y}|\mathbf{x},\boldsymbol{\theta}_0))(\boldsymbol{\theta}-\boldsymbol{\theta}_0)}\,. \tag{C.280}$$

Evaluating the expression on $s = \frac{1}{n}\ln\frac{k^p}{\delta}$ and $\boldsymbol{\theta} = \mathbf{0}$ concludes the proof. □

**Proposition 5.30** [Return to statement]. If $\nu(\mathbf{x},\mathbf{y})$ is a uniformly strictly log-concave density and $\nu(\mathbf{y}|\mathbf{x})$ is deterministic, then $\exists M > 0$, such that, for any $\boldsymbol{\theta} \in \boldsymbol{\Theta}$,

$$\mathcal{I}_{\boldsymbol{\theta}}^{-1}(\tfrac{1}{n}\log\tfrac{k^p}{\delta}) \le \sqrt{\tfrac{1}{n}\log\tfrac{k^p}{\delta}}\sqrt{M\mathbb{E}_\nu\left[\|\nabla_{\mathbf{x}}\ell(\mathbf{y},\mathbf{x},\boldsymbol{\theta})\|_2^2\right]}\,. \tag{C.281}$$

*Proof.* Using Chafaï, 2004's Corollary 2.1 on $\phi(f) = -\ln f$ and $f_{\boldsymbol{\theta}}(y,\mathbf{x}) = e^{-\lambda\ell(\mathbf{y},\mathbf{x},\boldsymbol{\theta})}$; it verifies that

$$J_{\boldsymbol{\theta}}(\lambda) \le M\lambda^2\,\mathbb{E}_\nu\left[|\nabla_x\ell(\mathbf{y},\mathbf{x},\boldsymbol{\theta})|_2^2\right]. \tag{C.282}$$

Then,

$$\mathcal{I}_{\boldsymbol{\theta}}^{-1}(s) = \inf_{\lambda>0}\frac{s+J_{\boldsymbol{\theta}}(\lambda)}{\lambda} \le \inf_{\lambda>0}\frac{s+M\lambda^2\mathbb{E}_\nu[|\nabla_x\ell(\mathbf{y},\mathbf{x},\boldsymbol{\theta})|^2]}{\lambda}\,. \tag{C.283}$$

Deriving w.r.t. $\lambda$ to compute the optimal value, gives $\lambda_{inf} = \sqrt{\frac{s}{M\mathbb{E}_\nu[|\nabla_x\ell(\mathbf{y},\mathbf{x},\boldsymbol{\theta})|^2]}}$. Using this,

$$\mathcal{I}_{\boldsymbol{\theta}}^{-1}(s) \le \sqrt{s}\sqrt{M\mathbb{E}_\nu[|\nabla_x\ell(\mathbf{y},\mathbf{x},\boldsymbol{\theta})|^2]}\,. \tag{C.284}$$

□

**Proposition 5.32** [Return to statement]. Under Assumption 5.31, let $\mathbf{x}_t$ denote the random variable denoting a transformation of an input $\mathbf{x}_0$ using a composition of $t$ transformations, then

$$MI(\mathbf{y}\,;\mathbf{x}_t) \le MI(\mathbf{y}\,;\mathbf{x}_{t-1}) \quad \forall t \in \mathbb{N}^+\,. \tag{C.285}$$

As a consequence, $MI(\mathbf{y};\mathbf{x}) = MI(\mathbf{y};\mathbf{x}_T) \le MI(\mathbf{y};\mathbf{x}_0)$.

*Proof.* According to Assumption 5.31, the targets $\mathbf{y}$ are independent of the input $\mathbf{x}_t$ given $\mathbf{x}_{t-1}$, that is, $\mathbf{y} \perp \mathbf{x}_t|\mathbf{x}_{t-1}$. This is a result of the fact that $\mathbf{x}_{t-1}$ d-connects $\mathbf{x}_t$ and $\mathbf{y}$ in the graph representation of Assumption 5.31. The result follows then by the data processing inequality. □

**Proposition 5.33** [Return to statement]. Under Assumption 5.31, if $\ell(\mathbf{y},\mathbf{x},\boldsymbol{\theta})$ is convex under $\mathbf{x}$ and $\mathbb{E}_{g_t}[g_t(\mathbf{x})] = \mathbf{x}$, then, $\forall\boldsymbol{\theta} \in \boldsymbol{\Theta},\ L^{\nu_{t+1}}(\boldsymbol{\theta}) \ge L^{\nu_t}(\boldsymbol{\theta})$.

*Proof.* By definition, $L^{\nu_{t+1}}(\boldsymbol{\theta}) = \mathbb{E}_{\nu_{t+1}}[\ell(\mathbf{y}, \mathbf{x}_{t+1}), \boldsymbol{\theta})] = \mathbb{E}_{\nu_t}\mathbb{E}_{g_{t+1}}[\ell(\mathbf{y}, g_{t+1}(\mathbf{x}_t), \boldsymbol{\theta})]$. As $\ell(\mathbf{y}, \mathbf{x}, \boldsymbol{\theta})$ is convex in $\mathbf{x}$, the expectation under $g_{t+1}$ can be moved inside $\ell$:

$$L^{\nu_{t+1}}(\boldsymbol{\theta}) \geq \mathbb{E}_{\nu_t}[\ell(\mathbf{y}, \mathbb{E}_{g_{t+1}}[g_{t+1}(\mathbf{x}_t)], \boldsymbol{\theta})] \,. \tag{C.286}$$

Given that $\mathbb{E}_{g_{t+1}}[g_{t+1}(\mathbf{x})] = \mathbf{x}$, it verifies that $L^{\nu_{t+1}}(\boldsymbol{\theta}) \geq \mathbb{E}_{\nu_t}[\ell(\mathbf{y}, \mathbf{x}_t, \boldsymbol{\theta})] = L^{\nu_t}(\boldsymbol{\theta})$. □

**Proposition 5.34** [Return to statement]. Under Assumption 5.31, if the change observed in the loss after a transformation $\Delta(\mathbf{y}, \mathbf{x}, g, \boldsymbol{\theta}) = \ell(\mathbf{y}, g(\mathbf{x}), \boldsymbol{\theta}) - \ell(\mathbf{y}, \mathbf{x}, \boldsymbol{\theta})$ is statistically independent of the loss itself, $\Delta(\mathbf{y}, \mathbf{x}, g, \boldsymbol{\theta}) \perp \ell(\mathbf{y}, \mathbf{x}, \boldsymbol{\theta})$, then

$$\mathcal{I}_{\boldsymbol{\theta}}^{\nu_{t+1}}(a) \leq \mathcal{I}_{\boldsymbol{\theta}}^{\nu_t}(a) \quad \forall a > 0 \,. \tag{C.287}$$

*Proof.* By definition, it verifies that $J_{\boldsymbol{\theta}}^{\nu_{t+1}}(\lambda) = \lambda L^{\nu_{t+1}}(\boldsymbol{\theta}) + \ln \mathbb{E}_{\nu_{t+1}}[e^{-\ell(\mathbf{y}, \mathbf{x}_{t+1}, \boldsymbol{\theta})}]$, where $\nu_{t+1}$ can be expanded leading to

$$J_{\boldsymbol{\theta}}^{\nu_{t+1}}(\lambda) = \lambda L^{\nu_{t+1}}(\boldsymbol{\theta}) + \ln \mathbb{E}_{\nu_t}\mathbb{E}_{g_{t+1}}\left[e^{-\lambda\ell(\mathbf{y}, g_{t+1}(\mathbf{x}_t), \boldsymbol{\theta})}\right] \,. \tag{C.288}$$

Using the decomposition of the loss,

$$J_{\boldsymbol{\theta}}^{\nu_{t+1}}(\lambda) = \lambda L^{\nu_{t+1}}(\boldsymbol{\theta}) + \ln \mathbb{E}_{\nu_t}\mathbb{E}_{g_{t+1}}\left[e^{-\lambda(\ell(\mathbf{y}, \mathbf{x}_t, \boldsymbol{\theta}) + \Delta(\mathbf{y}, \mathbf{x}_t, g_{t+1}, \boldsymbol{\theta}))}\right] \,, \tag{C.289}$$

which can be written as

$$J_{\boldsymbol{\theta}}^{\nu_{t+1}}(\lambda) = \lambda L^{\nu_{t+1}}(\boldsymbol{\theta}) + \ln \mathbb{E}_{\nu_t}\left[e^{-\lambda\ell(\mathbf{y}, \mathbf{x}_t, \boldsymbol{\theta})}\right] + \ln \mathbb{E}_{\nu_t}\mathbb{E}_{g_{t+1}}\left[e^{-\lambda\Delta(\mathbf{y}, \mathbf{x}_t, g_{t+1}, \boldsymbol{\theta})}\right] . \tag{C.290}$$

Using Jensen's inequality in the last term,

$$J_{\boldsymbol{\theta}}^{\nu_{t+1}}(\lambda) \geq \lambda L^{\nu_{t+1}}(\boldsymbol{\theta}) + \ln \mathbb{E}_{\nu_t}\left[e^{-\lambda\ell(\mathbf{y}, \mathbf{x}_t, \boldsymbol{\theta})}\right] - \lambda\mathbb{E}_{\nu_t}\mathbb{E}_{g_{t+1}}[\Delta(\mathbf{y}, \mathbf{x}_t, g_{t+1}, \boldsymbol{\theta})] \,, \tag{C.291}$$

and using the definition of $\Delta(\mathbf{y}, \mathbf{x}_t, g_{t+1}, \boldsymbol{\theta})$,

$$J_{\boldsymbol{\theta}}^{\nu_{t+1}}(\lambda) \geq \lambda L^{\nu_{t+1}}(\boldsymbol{\theta}) + \ln \mathbb{E}_{\nu_t}\left[e^{-\lambda\ell(\mathbf{y}, \mathbf{x}_t, \boldsymbol{\theta})}\right] - \lambda\mathbb{E}_{\nu_t}\mathbb{E}_{g_{t+1}}[\ell(\mathbf{y}, g_{t+1}(\mathbf{x}_t), \boldsymbol{\theta})] + \lambda\mathbb{E}_{\nu_t}[\ell(\mathbf{y}, \mathbf{x}_t, \boldsymbol{\theta})]] \,. \tag{C.292}$$

By definition of the expected loss,

$$J_{\boldsymbol{\theta}}^{\nu_{t+1}}(\lambda) \geq \lambda L^{\nu_{t+1}}(\boldsymbol{\theta}) + \ln \mathbb{E}_{\nu_t}[e^{-\lambda\ell(\mathbf{y}, \mathbf{x}_t, \boldsymbol{\theta})}] - \lambda L^{\nu_{t+1}}(\boldsymbol{\theta}) + \lambda L^{\nu_t}(\boldsymbol{\theta}) \tag{C.293}$$

$$= \ln \mathbb{E}_{\nu_t}[e^{-\lambda\ell(\mathbf{y}, \mathbf{x}_t, \boldsymbol{\theta})}] + \lambda L^{\nu_t}(\boldsymbol{\theta}) = J_{\boldsymbol{\theta}}^{\nu_t}(\lambda) \,. \tag{C.294}$$

From this inequality and by considering standard properties of the Legendre transform, it verifies that $J_{\boldsymbol{\theta}}^{\nu_{t+1}}(\lambda) \geq J_{\boldsymbol{\theta}}^{\nu_t}(\lambda)$, which implies that $\mathcal{I}_{\boldsymbol{\theta}}^{\nu_{t+1}}(a) \leq \mathcal{I}_{\boldsymbol{\theta}}^{\nu_t}(a)$. □

**Proposition 5.36** [Return to statement]. Under Assumption 5.31, if a model $\boldsymbol{\theta} \in \boldsymbol{\Theta}$ is $G_t$-invariant then,

$$L^{\nu_{t+1}}(\boldsymbol{\theta}) = L^{\nu_t}(\boldsymbol{\theta}) \quad \text{and} \quad \mathcal{I}_{\boldsymbol{\theta}}^{\nu_{t+1}}(a) = \mathcal{I}_{\boldsymbol{\theta}}^{\nu_t}(a) \quad \forall a > 0\,. \tag{C.295}$$

*Proof.* It immediately follows from the definition of model invariant (Definition 5.35) and the definitions of $L^{\nu_{t+1}}(\boldsymbol{\theta})$, $L^{\nu_t}(\boldsymbol{\theta})$, $\mathcal{I}_{\boldsymbol{\theta}}^{\nu_{t+1}}(a)$ and $\mathcal{I}_{\boldsymbol{\theta}}^{\nu_t}$. □

**Theorem 5.37** [Return to statement]. Under Assumption 5.31, it verifies that

$$\forall \boldsymbol{\theta} \in \boldsymbol{\Theta}, \quad L^{\nu_T,\ell}(\boldsymbol{\theta}) = L^{\nu_0,\ell_G}(\boldsymbol{\theta}) \quad \textit{and} \quad \mathcal{I}_{\boldsymbol{\theta}}^{\nu_T,\ell}(a) \leq \mathcal{I}_{\boldsymbol{\theta}}^{\nu_0,\ell_G}(a) \;\; \forall a > 0\,. \tag{C.296}$$

*Proof.* Under Assumption 5.31, it is clear that

$$L^{\nu_0,\ell_G}(\boldsymbol{\theta}) = \mathbb{E}_{\nu_0}[\ell_G(\mathbf{y}, \mathbf{x}, \boldsymbol{\theta}))] = \mathbb{E}_{\nu_0}\mathbb{E}_{g\sim h}[\ell(\mathbf{y}, g(\mathbf{x}), \boldsymbol{\theta}))] = \mathbb{E}_{\nu}[\ell(\mathbf{y}, \mathbf{x}, \boldsymbol{\theta}))] = L^{\nu_T,\ell}(\boldsymbol{\theta})\,. \tag{C.297}$$

On the other hand, from the definition of the cummulant function, $J_{\boldsymbol{\theta}}(\lambda) = \ln \mathbb{E}_{\nu_T}[e^{-\lambda\ell(\mathbf{y},\mathbf{x},\boldsymbol{\theta})}] + \lambda\mathbb{E}_{\nu_T}[\ell(\mathbf{y}, \mathbf{x}, \boldsymbol{\theta})]$. Where by definition

$$J_{\boldsymbol{\theta}}(\lambda) = \ln \mathbb{E}_{\nu_0}\mathbb{E}_{g\sim h}[e^{-\lambda\ell(\mathbf{y},g(\mathbf{x}),\boldsymbol{\theta})}] + \lambda\mathbb{E}_{\nu_0}\mathbb{E}_{g\sim h}[\ell(\mathbf{y}, g(\mathbf{x}), \boldsymbol{\theta})]\,, \tag{C.298}$$

where expectations can be exchanged as

$$J_{\boldsymbol{\theta}}(\lambda) = \ln \mathbb{E}_{\nu_0(\mathbf{x})}\mathbb{E}_{\nu(\mathbf{y}|\mathbf{x})}\mathbb{E}_{g}[e^{-\lambda\ell(\mathbf{y},g(\mathbf{x}),\boldsymbol{\theta})}] + \lambda\mathbb{E}_{\nu_0(\mathbf{x})}\mathbb{E}_{\nu(\mathbf{y}|\mathbf{x})}\mathbb{E}_{g}[\ell(\mathbf{y}, g(\mathbf{x}), \boldsymbol{\theta})]\,. \tag{C.299}$$

Applying Jensen's inequality to the exponential,

$$J_{\boldsymbol{\theta}}(\lambda) \geq \ln \mathbb{E}_{\nu_0(\mathbf{x})}\mathbb{E}_{\nu(\mathbf{y}|\mathbf{x})}\left[e^{-\lambda\mathbb{E}_g[\ell(\mathbf{y},g(\mathbf{x}),\boldsymbol{\theta})]}\right] + \lambda\mathbb{E}_{\nu_0(\mathbf{x})}\mathbb{E}_{\nu(\mathbf{y}|\mathbf{x})}\mathbb{E}_{g}[\ell(\mathbf{y}, g(\mathbf{x}), \boldsymbol{\theta})]\,. \tag{C.300}$$

Where by definition of $\ell_G$,

$$J_{\boldsymbol{\theta}}(\lambda) \geq \ln \mathbb{E}_{\nu_0}[e^{-\lambda[\ell_G(\mathbf{y},\mathbf{x},\boldsymbol{\theta})]}] + \lambda\mathbb{E}_{\nu_0}[\ell_G(\mathbf{y}, \mathbf{x}, \boldsymbol{\theta})] = J_{\boldsymbol{\theta}}^{\nu_0,\ell_G}(\lambda)\,. \tag{C.301}$$

From this point, the inequality regarding the inverse rate is clear from the definition. □

**Corollary 5.38** [Return to statement]. With h.p. $1-\delta$ over $D_0 \sim \nu_0^n$, for all $\boldsymbol{\theta} \in \boldsymbol{\Theta}$, simultaneously,

$$\text{if } \hat{L}^{\ell_G}(D_0, \boldsymbol{\theta}) \leq \epsilon \quad \text{then} \quad (\mathcal{I}_{\boldsymbol{\theta}}^{\nu_0,\ell_G})^{-1}(\tfrac{1}{n}\log\tfrac{k^p}{\delta}) \leq L^{\nu_T,\ell}(\boldsymbol{\theta}) \leq (\mathcal{I}_{\boldsymbol{\theta}}^{\nu_0,\ell_G})^{-1}(\tfrac{1}{n}\log\tfrac{k^p}{\delta}) + \epsilon\,. \tag{C.302}$$

*Proof.* Under Assumption 5.31, it is clear that

$$L^{v_0,\ell_G}(\boldsymbol{\theta}) = \mathbb{E}_{\nu_0}[\ell_G(\mathbf{y},\mathbf{x},\boldsymbol{\theta}))] = \mathbb{E}_{\nu_0}\mathbb{E}_{g\sim h}[\ell(\mathbf{y},g(\mathbf{x}),\boldsymbol{\theta}))] = \mathbb{E}_{\nu}[\ell(\mathbf{y},\mathbf{x},\boldsymbol{\theta}))] = L^{\nu_T,\ell}(\boldsymbol{\theta})\,. \tag{C.303}$$

Where it verifies that $L(\boldsymbol{\theta}) := L^{\nu_T,\ell}(\boldsymbol{\theta})$. Thus, the result comes from the application of Theorem 5.17.

□

**Proposition 5.39** [Return to statement]. Under Assumption 5.31, if the transformations $G$ define a group and its probability distribution $h$ is uniform, then

$$\hat{L}^{\ell_G}(D_T,\boldsymbol{\theta}) = \hat{L}^{\ell_G}(D_0,\boldsymbol{\theta})\,, \tag{C.304}$$

where $D_0$ is any un-transformed dataset and $D_T$ is the corresponding transformed dataset obtained by applying a transformation $g \sim h$ to each of the samples in $D_0$.

*Proof.* On one hand, by definition of $\ell_G$ it verifies that

$$\hat{L}^{\ell_G}(D_T,\boldsymbol{\theta}) = \frac{1}{n}\sum_i \mathbb{E}_{g'\sim h}[\ell(\mathbf{y}_i,g'(\mathbf{x}_i),\boldsymbol{\theta})] = \frac{1}{n}\sum_i \int h(g')\ell(\mathbf{y}_i,g'(g_i(x_{0,i})),\boldsymbol{\theta})\;dg'\,. \tag{C.305}$$

As the set of transformations define a group, for each transformation $g_i$ that generated each input, there exists $g_i^{-1}$. Furthermore, there exists $l := g' \circ g_i \in G_1$ such that $g' = l \circ g_i^{-1}$. Thus,

$$\hat{L}^{\ell_G}(D_T,\boldsymbol{\theta}) = \frac{1}{n}\sum_i \int_{g'=l\circ g_i^{-1}} h(l\circ g_i^{-1})\ell(\mathbf{y}_i,l\circ g_i^{-1}\circ g_i(x_{0,i}),\boldsymbol{\theta})\;dg'\,. \tag{C.306}$$

Then, using a change of variables

$$\hat{L}^{\ell_G}(D_T,\boldsymbol{\theta}) = \frac{1}{n}\sum_i \int_l h(l\circ g_i^{-1})\ell(\mathbf{y}_i,l(x_{0,i}),\boldsymbol{\theta})\;dl\,. \tag{C.307}$$

As $h$ is uniform $h(l\circ g_i^{-1}) = h(l)$,

$$\hat{L}^{\ell_G}(D_T,\boldsymbol{\theta}) = \frac{1}{n}\sum_i \int_l h(l)\ell(\mathbf{y}_i,l(x_{0,i}),\boldsymbol{\theta})dl = \hat{L}^{\ell_G}(D_0,\boldsymbol{\theta})\,. \tag{C.308}$$

□

**Proposition 5.40** [Return to statement]. Given an untransformed dataset $D_0$, for any transformed dataset $D_T$ derived from $D_0$ under Assumption 5.31, if

$$m_{\boldsymbol{\theta}} := \operatorname*{ess\,inf}_{(\mathbf{x},\mathbf{y})\sim\nu_T} \ell(\mathbf{y},g(\mathbf{x}),\boldsymbol{\theta}) \quad \forall g\in G\,, \tag{C.309}$$

and $\forall g \in G, \exists g^{-1} \in G$. Then, it verifies that

$$\forall \boldsymbol{\theta} \in \boldsymbol{\Theta} \quad \hat{L}^{\ell_G}(D_T, \boldsymbol{\theta}) = m_{\boldsymbol{\theta}} \iff \hat{L}^{\ell}(D_0, \boldsymbol{\theta}) = m_{\boldsymbol{\theta}} \,. \tag{C.310}$$

*Proof.* First of all, by definition we got that $m_{\boldsymbol{\theta}} = ess \inf_{(\mathbf{x},\mathbf{y})\sim\nu_T} \ell(\mathbf{y}, \mathbf{x}, \boldsymbol{\theta})$. Then, it verifies that

$$\hat{L}^{\ell_G}(D_T, \boldsymbol{\theta}) = m_{\boldsymbol{\theta}} \iff \ell_G(\mathbf{y}, \mathbf{x}, \boldsymbol{\theta}) = ess \inf_{(\mathbf{x},\mathbf{y})\sim\nu_T} \ell(\mathbf{y}, \mathbf{x}, \boldsymbol{\theta}) \quad \forall (\mathbf{x}, \mathbf{y}) \in D_T \,. \tag{C.311}$$

Thus, by definition of $\ell_G$:

$$\hat{L}^{\ell_G}(D_T, \boldsymbol{\theta}) = m_{\boldsymbol{\theta}} \iff \mathbb{E}_g[\ell(\mathbf{y}, g(\mathbf{x}), \boldsymbol{\theta})] = ess \inf_{(\mathbf{x},\mathbf{y})\sim\nu_T} \ell(\mathbf{y}, \mathbf{x}, \boldsymbol{\theta}) \quad \forall (\mathbf{x}, \mathbf{y}) \in D_T \,. \tag{C.312}$$

Now, using that $ess \inf_{(\mathbf{x},\mathbf{y})\sim\nu_T} \ell(\mathbf{y}, g(\mathbf{x}), \boldsymbol{\theta}) = m_{\boldsymbol{\theta}} \; \forall g \in G$, reaching the essential infimum in expectation means all the losses inside the expectation reach such infimum:

$$\hat{L}^{\ell_G}(D_T, \boldsymbol{\theta}) = m_{\boldsymbol{\theta}} \iff \ell(\mathbf{y}, g(\mathbf{x}), \boldsymbol{\theta}) = ess \inf_{(\mathbf{x},\mathbf{y})\sim\nu_T} \ell(\mathbf{y}, \mathbf{x}, \boldsymbol{\theta}) \quad \forall g \in G, \forall (\mathbf{x}, \mathbf{y}) \in D_T \,. \tag{C.313}$$

Using that for every $(\mathbf{x}_0, \mathbf{y}) \in D_0$ exists $g' \in G$ associated to that input such that $\mathbf{x} = g'(\mathbf{x}_0)$ and that

$$\hat{L}^{\ell_G}(D_T, \boldsymbol{\theta}) = m_{\boldsymbol{\theta}} \iff \ell(\mathbf{y}, g \circ g'(\mathbf{x}_0), \boldsymbol{\theta}) = ess \inf_{(\mathbf{x},\mathbf{y})\sim\nu_T} \ell(\mathbf{y}, \mathbf{x}, \boldsymbol{\theta}) \quad \forall g \in G, \forall (\mathbf{x}_0, \mathbf{y}) \in D_0 \,. \tag{C.314}$$

Then, using that there exists $g'^{-1} \in G$, we got that

$$\hat{L}^{\ell_G}(D_T, \boldsymbol{\theta}) = m_{\boldsymbol{\theta}} \iff \ell(\mathbf{y}, \mathbf{x}_0, \boldsymbol{\theta}) = ess \inf_{(\mathbf{x},\mathbf{y})\sim\nu_T} \ell(\mathbf{y}, \mathbf{x}, \boldsymbol{\theta}) \quad \forall (\mathbf{x}_0, \mathbf{y}) \in D_0 \,. \tag{C.315}$$

As a result, the empirical loss $\hat{L}^{\ell}(D_0, \boldsymbol{\theta})$ is equal to $m_{\boldsymbol{\theta}}$ too. □

**Theorem 5.41** [Return to statement]. For any $\epsilon \in (0, L^\star)$ and any $\delta \in (0, 1)$, with high probability $1 - \delta$ over $D \sim \nu^n$, for all $\boldsymbol{\theta} \in \boldsymbol{\Theta}$, simultaneously,

$$\text{if} \quad \hat{L}(D, \boldsymbol{\theta}) \leq \epsilon \quad \text{then} \quad p \geq \frac{n\mathcal{I}_{\boldsymbol{\theta}}(L^\star - \epsilon) + \log \delta}{\log k} \,. \tag{C.316}$$

*Proof.* Due to fact that $L(D, \boldsymbol{\theta}) \leq \epsilon$ and to Theorem 5.17, we have that with high probability $1 - \delta$ over $D \sim \nu^n(y, x)$,

$$L(\boldsymbol{\theta}) \leq \epsilon + \mathcal{I}_{\boldsymbol{\theta}}^{-1}(\tfrac{1}{n} \ln \tfrac{k^p}{\delta}) \,. \tag{C.317}$$

Then, just rearranging terms as follows, we get $L(\boldsymbol{\theta}) - \epsilon \leq \mathcal{I}_{\boldsymbol{\theta}}^{-1}(\frac{1}{n} \ln \frac{k^p}{\delta})$. Applying the rate function at both sides,

$$\mathcal{I}_{\boldsymbol{\theta}}(L(\boldsymbol{\theta})-\epsilon)\leq \tfrac{1}{n}\ln\tfrac{k^p}{\delta}=\frac{1}{n}\left(\ln k^p-\ln\delta\right)\,. \tag{C.318}$$

Then, we got $\ln k^p \geq n\mathcal{I}_{\boldsymbol{\theta}}(L(\boldsymbol{\theta})-\epsilon)+\ln\delta$ and

$$p\geq \frac{n\mathcal{I}_{\boldsymbol{\theta}}(L(\boldsymbol{\theta})-\epsilon)+\ln\delta}{\ln k}\,. \tag{C.319}$$

As $L^\star \leq L(\boldsymbol{\theta})$ and the rate function is monotonically increasing, then

$$p\geq \frac{n\mathcal{I}_{\boldsymbol{\theta}}(L(\boldsymbol{\theta})-\epsilon)+\ln\delta}{\ln k}\geq \frac{n\mathcal{I}_{\boldsymbol{\theta}}(L^\star-\epsilon)+\ln\delta}{\ln k}\,. \tag{C.320}$$

□

**Corollary 5.42** [Return to statement]. If $\ell(\mathbf{y},\mathbf{x},\boldsymbol{\theta})$ is Lipschitz w.r.t. $\mathbf{x}$ with constant $Lip(\boldsymbol{\theta})$ and satisfies a $c$-isoperimetry assumption, then for any $\epsilon\in(0,L^\star)$ and any $\delta\in(0,1)$, with high probability $1-\delta$ over $D\sim\nu^n$, for all $\boldsymbol{\theta}\in\boldsymbol{\Theta}$, simultaneously,

$$\text{if}\quad \hat{L}(D,\boldsymbol{\theta})\leq\epsilon \quad\text{then}\quad Lip(\boldsymbol{\theta})\geq\sqrt{\tfrac{nd}{2c(p\log k-\log\delta)}}(L^\star-\epsilon)\,. \tag{C.321}$$

*Proof.* Under the isoperimetry assumption, we have $\mathcal{I}_{\boldsymbol{\theta}}(a)\leq\frac{da^2}{2cLip(\boldsymbol{\theta})^2}$. Due to fact that $L(D,\boldsymbol{\theta})\leq\epsilon$ and to Theorem 5.17, we have that with high probability $1-\delta$ over $D\sim\nu^n(y,x)$,

$$L(\boldsymbol{\theta})\leq\epsilon+\mathcal{I}_{\boldsymbol{\theta}}^{-1}(\tfrac{1}{n}\ln\tfrac{k^p}{\delta})\,. \tag{C.322}$$

Then, just rearranging terms as follows, we get $L(\boldsymbol{\theta})-\epsilon\leq\mathcal{I}_{\boldsymbol{\theta}}^{-1}(\frac{1}{n}\ln\frac{k^p}{\delta})$. Applying the rate function at both sides,

$$\mathcal{I}_{\boldsymbol{\theta}}(L(\boldsymbol{\theta})-\epsilon)\leq \tfrac{1}{n}\ln\tfrac{k^p}{\delta}=\frac{1}{n}\left(\ln k^p-\ln\delta\right)\,. \tag{C.323}$$

Then, we got $\ln k^p \geq n\mathcal{I}_{\boldsymbol{\theta}}(L(\boldsymbol{\theta})-\epsilon)+\ln\delta$ and

$$p\geq \frac{n\mathcal{I}_{\boldsymbol{\theta}}(L(\boldsymbol{\theta})-\epsilon)+\ln\delta}{\ln k}\,. \tag{C.324}$$

As $L^\star \leq L(\boldsymbol{\theta})$ and the rate function is monotonically increasing, then

$$p\geq \frac{n\mathcal{I}_{\boldsymbol{\theta}}(L(\boldsymbol{\theta})-\epsilon)+\ln\delta}{\ln k}\geq \frac{n\mathcal{I}_{\boldsymbol{\theta}}(L^\star-\epsilon)+\ln\delta}{\ln k}\,. \tag{C.325}$$

Using the upper bound over the rate function we obtained before, we got that

$$p\geq \frac{n\frac{d(L^\star-\epsilon)^2}{2cLip(\boldsymbol{\theta})^2}+\ln\delta}{\ln k}\,. \tag{C.326}$$

Re-arranging terms,

$$\sqrt{\frac{nd(L^\star-\epsilon)^2}{2c(p\ln k-\ln\delta)}}\leq Lip(\boldsymbol{\theta})\,. \tag{C.327}$$

□

**Corollary 5.43** [Return to statement]. If the loss function $\ell(\mathbf{y},\mathbf{x},\boldsymbol{\theta})$ is Lipschitz w.r.t. $\boldsymbol{\theta}$ with constant $M>0$, then, for any $\boldsymbol{\theta}_0\in\boldsymbol{\Theta}_0=\{\boldsymbol{\theta}\in\boldsymbol{\Theta}\mid\mathbb{V}_\nu(\ell(\mathbf{y},\mathbf{x},\boldsymbol{\theta}))=0\}\subset\boldsymbol{\Theta}$, any $\epsilon\in(0,L^\star)$ and any $\delta\in(0,1)$, with high probability $1-\delta$ over $D\sim\nu^n$, for all $\boldsymbol{\theta}\in\boldsymbol{\Theta}$, simultaneously,

$$\text{if}\quad \hat{L}(D,\boldsymbol{\theta})\leq\epsilon\quad\text{then}\quad \|\boldsymbol{\theta}-\boldsymbol{\theta}_0\|_2\geq\sqrt{\tfrac{n}{8M(p\log k-\log\delta)}}(L^\star-\epsilon)\,. \tag{C.328}$$

*Proof.* If the loss is Lipschitz continuous with constant $M$, $\forall y,x,\boldsymbol{\theta}\quad \|\nabla_{\boldsymbol{\theta}}\ell(\mathbf{y},\mathbf{x},\boldsymbol{\theta})\|_2^2\leq M$. Then, $J_{\boldsymbol{\theta}}(\lambda)$ verifies

$$\|\nabla_{\boldsymbol{\theta}}J_{\boldsymbol{\theta}}(\lambda)\|_2^2=||-\lambda\mathbb{E}_{\nu p^\lambda}\left[\nabla_{\boldsymbol{\theta}}\ell(\mathbf{y},\mathbf{x},\boldsymbol{\theta})\right]+\lambda\mathbb{E}_\nu\left[\nabla_{\boldsymbol{\theta}}\ell(\mathbf{y},\mathbf{x},\boldsymbol{\theta})\right]||_2^2\leq 2M\lambda^2\,. \tag{C.329}$$

where $\mathbb{E}_{\nu p^\lambda}\left[\nabla_{\boldsymbol{\theta}}\ell(\mathbf{y},\mathbf{x},\boldsymbol{\theta})\right]=\frac{\mathbb{E}_\nu[p(\mathbf{y}|\mathbf{x},\boldsymbol{\theta})^\lambda\ell(\mathbf{y},\mathbf{x},\boldsymbol{\theta})]}{\mathbb{E}_\nu[p(\mathbf{y}|\mathbf{x},\boldsymbol{\theta})^\lambda]}$. With this, we have

$$|J_{\boldsymbol{\theta}}(\lambda)-J_{\boldsymbol{\theta}_0}(\lambda)|\leq 2M\lambda^2\|\boldsymbol{\theta}-\boldsymbol{\theta}_0\|_2^2\implies J_{\boldsymbol{\theta}}(\lambda)\leq 2M\lambda^2\|\boldsymbol{\theta}-\boldsymbol{\theta}_0\|_2^2\,. \tag{C.330}$$

Then, for any $a\geq 0$, we have by definition of the rate function that

$$\mathcal{I}_{\boldsymbol{\theta}}(a)\geq a\lambda-J_{\boldsymbol{\theta}}(\lambda)\geq a\lambda-2M\lambda^2\|\boldsymbol{\theta}-\boldsymbol{\theta}_0\|_2^2\,. \tag{C.331}$$

As the inequality holds for any $\lambda>0$, maximizing it raises $\lambda^\star=\frac{a}{4M\|\boldsymbol{\theta}-\boldsymbol{\theta}_0\|_2^2}$. Thus,

$$\mathcal{I}_{\boldsymbol{\theta}}(a)\geq a\lambda-J_{\boldsymbol{\theta}}(\lambda)\geq a\frac{a}{4M\|\boldsymbol{\theta}-\boldsymbol{\theta}_0\|_2^2}-2M\frac{a^2}{(4M\|\boldsymbol{\theta}-\boldsymbol{\theta}_0\|_2^2)^2}\|\boldsymbol{\theta}-\boldsymbol{\theta}_0\|_2^2=\frac{a^2}{8M\|\boldsymbol{\theta}-\boldsymbol{\theta}_0\|_2^2}\,. \tag{C.332}$$

Due to fact that $L(D,\boldsymbol{\theta})\leq\epsilon$ and to Theorem 5.17, we have that with high probability $1-\delta$ over $D\sim\nu^n(\mathbf{x},\mathbf{y})$, $L(\boldsymbol{\theta})\leq\epsilon+\mathcal{I}_{\boldsymbol{\theta}}^{-1}(\frac{1}{n}\ln\frac{k^p}{\delta})$. Then, just rearranging terms as follows, we get $L(\boldsymbol{\theta})-\epsilon\leq\mathcal{I}_{\boldsymbol{\theta}}^{-1}(\frac{1}{n}\ln\frac{k^p}{\delta})$. Applying the rate function at both sides,

$$\mathcal{I}_{\boldsymbol{\theta}}(L(\boldsymbol{\theta})-\epsilon)\leq\tfrac{1}{n}\ln\tfrac{k^p}{\delta}=\frac{1}{n}\left(\ln k^p-\ln\delta\right)\,. \tag{C.333}$$

Then, we got $\ln k^p\geq n\mathcal{I}_{\boldsymbol{\theta}}(L(\boldsymbol{\theta})-\epsilon)+\ln\delta$ and $p\geq\frac{n\mathcal{I}_{\boldsymbol{\theta}}(L(\boldsymbol{\theta})-\epsilon)+\ln\delta}{\ln k}$. As $L^\star\leq L(\boldsymbol{\theta})$ and the rate function is monotonically increasing, then

$$p\geq\frac{n\mathcal{I}_{\boldsymbol{\theta}}(L(\boldsymbol{\theta})-\epsilon)+\ln\delta}{\ln k}\geq\frac{n\mathcal{I}_{\boldsymbol{\theta}}(L^\star-\epsilon)+\ln\delta}{\ln k}\,. \tag{C.334}$$

Using the upper bound over the rate function we obtained before, we got that

$$p \geq \frac{n\frac{(L^\star-\epsilon)^2}{8M\|\boldsymbol{\theta}-\boldsymbol{\theta}_0\|_2^2} + \ln\delta}{\ln k} \implies (L^\star - \epsilon)\sqrt{\frac{n}{8M(p\ln k - \ln\delta)}} \leq \|\boldsymbol{\theta} - \boldsymbol{\theta}_0\|_2 \,. \tag{C.335}$$

□

**Corollary 5.25** [Return to statement]. Let $\boldsymbol{\Theta} \subset \boldsymbol{\Theta}'$ be two nested model classes with $p < p'$ parameters respectively. For any $\epsilon > 0$, with h.p. $1-\delta$ over $D \sim \nu^n$, for any $\boldsymbol{\theta}' \in \boldsymbol{\Theta}'$, $\boldsymbol{\theta} \in \boldsymbol{\Theta}$, simultaneously,

$$\begin{gathered} \hat{L}(D,\boldsymbol{\theta}') \leq \epsilon,\ \hat{L}(D,\boldsymbol{\theta}) \leq \epsilon \text{ and } L(\boldsymbol{\theta}') + \epsilon < L(\boldsymbol{\theta}) \\ \Downarrow \\ \boldsymbol{\theta} \text{ is not } \mathcal{I}_{\boldsymbol{\theta}}^{-1}(\tfrac{1}{n}\log\tfrac{k^{p'}}{\delta})\text{-smoother than } \boldsymbol{\theta}' \,. \end{gathered} \tag{C.336}$$

*Proof.* We can apply Theorem 5.17 on $\boldsymbol{\theta}$ assuming that $\boldsymbol{\theta} \in \boldsymbol{\Theta}'$, because we have that $\boldsymbol{\Theta} \subset \boldsymbol{\Theta}'$. In consequence, with h.p., we have

$$L(\boldsymbol{\theta}) \leq L(D,\boldsymbol{\theta}) + \mathcal{I}_{\boldsymbol{\theta}}^{-1}(\tfrac{1}{n}\ln\tfrac{k^{p'}}{\delta}) \tag{C.337}$$

Using the fact that $L(D,\boldsymbol{\theta}) \leq \epsilon$, and that the inverse rate is bounded by the expected loss, we arrive to the following inequality

$$L(\boldsymbol{\theta}) \leq \epsilon + \mathcal{I}_{\boldsymbol{\theta}}^{-1}(\tfrac{1}{n}\ln\tfrac{k^{p'}}{\delta}) \leq \epsilon + L(\boldsymbol{\theta}) \tag{C.338}$$

The same reasoning applies for $\boldsymbol{\theta}'$:

$$L(\boldsymbol{\theta}') \leq \epsilon + \mathcal{I}_{\boldsymbol{\theta}'}^{-1}(\tfrac{1}{n}\ln\tfrac{k^{p'}}{\delta}) \leq \epsilon + L(\boldsymbol{\theta}') \tag{C.339}$$

Using the theorem's premise, $L(\boldsymbol{\theta}') + \epsilon < L(\boldsymbol{\theta})$, we can chain the last two h.p. upper bounds. And note that we can do that *with no extra cost*, as both apply on the same bigger model class $\boldsymbol{\Theta}'$. That is, it is the same upper bound, which holds simultaneously for all models within $\boldsymbol{\Theta}'$, used on two different models $\boldsymbol{\theta}$ and $\boldsymbol{\theta}'$. Then, we have,

$$\mathcal{I}_{\boldsymbol{\theta}'}^{-1}(\tfrac{1}{n}\ln\tfrac{k^{p'}}{\delta}) < \mathcal{I}_{\boldsymbol{\theta}}^{-1}(\tfrac{1}{n}\ln\tfrac{k^{p'}}{\delta}) \,. \tag{C.340}$$

Naming $a = \mathcal{I}_{\boldsymbol{\theta}}^{-1}(\tfrac{1}{n}\ln\tfrac{k^{p'}}{\delta})$, such that $\tfrac{1}{n}\ln\tfrac{k^{p'}}{\delta} = \mathcal{I}_{\boldsymbol{\theta}}(a)$, we got that

$$\mathcal{I}_{\boldsymbol{\theta}'}^{-1}(\mathcal{I}_{\boldsymbol{\theta}}(a)) < \mathcal{I}_{\boldsymbol{\theta}}^{-1}(\mathcal{I}_{\boldsymbol{\theta}}(a)) = a \,. \tag{C.341}$$

Applying $\mathcal{I}_{\boldsymbol{\theta}'}(\cdot)$ at both sides:

$$\mathcal{I}_{\boldsymbol{\theta}'}(\mathcal{I}_{\boldsymbol{\theta}'}^{-1}(\mathcal{I}_{\boldsymbol{\theta}}(a))) = \mathcal{I}_{\boldsymbol{\theta}}(a) < \mathcal{I}_{\boldsymbol{\theta}'}(a) \,. \tag{C.342}$$

Then, we can negate the inverse statement using $\neg$ notation as $\neg\Big[\mathcal{I}_{\boldsymbol{\theta}}(a) \geq \mathcal{I}_{\boldsymbol{\theta}'}(a)\Big]$. Then we have that $\boldsymbol{\theta}$ is not $\beta$-smoother than $\boldsymbol{\theta}'$ for any $\beta \geq a$. Which is equivalent to say that, $\boldsymbol{\theta}$ is not $\beta$-smoother than $\boldsymbol{\theta}'$ for any $\beta$ such that $\mathcal{I}_{\boldsymbol{\theta}}(\beta) \geq \frac{1}{n}\ln\frac{kp'}{\delta}$. □

**Theorem 5.24** [Return to statement]. Let $\boldsymbol{\Theta} \subset \boldsymbol{\Theta}'$ be two nested model classes with $p < p'$ parameters respectively. For any $\epsilon > 0$, with h.p. $1-\delta$ over $D \sim \nu^n$, for any $\boldsymbol{\theta}' \in \boldsymbol{\Theta}'$, $\boldsymbol{\theta} \in \boldsymbol{\Theta}$, simultaneously

$$\begin{gathered}\text{if } \hat{L}(D,\boldsymbol{\theta}') \leq \epsilon\,,\ \hat{L}(D,\boldsymbol{\theta}) \leq \epsilon \text{ and } \boldsymbol{\theta}' \text{ is } \mathcal{I}^{-1}_{\boldsymbol{\theta}'}(\tfrac{1}{n}\log\tfrac{kp'}{\delta})\text{-smoother than } \boldsymbol{\theta} \\ \Downarrow \\ L(\boldsymbol{\theta}') \leq L(\boldsymbol{\theta}) + \epsilon\,. \end{gathered} \tag{C.343}$$

*Proof.* If $\boldsymbol{\theta}'$ is $\beta$-smoother than $\boldsymbol{\theta}$, by Definition 5.21, $\forall a \in (0,\beta]$ $\quad \mathcal{I}_{\boldsymbol{\theta}'}(a) \geq \mathcal{I}_{\boldsymbol{\theta}}(a)$ where $\beta = \mathcal{I}^{-1}_{\boldsymbol{\theta}'}(\frac{1}{n}\ln\frac{kp'}{\delta})$. Then, we have that $\mathcal{I}^{-1}_{\boldsymbol{\theta}'}(s) \leq \mathcal{I}^{-1}_{\boldsymbol{\theta}}(s)$ for $s = \mathcal{I}_{\boldsymbol{\theta}'}(\beta) = \mathcal{I}_{\boldsymbol{\theta}'}\Big(\mathcal{I}^{-1}_{\boldsymbol{\theta}'}(\frac{1}{n}\ln\frac{kp'}{\delta})\Big) = \frac{1}{n}\ln\frac{kp'}{\delta}$. Thus,

$$\mathcal{I}^{-1}_{\boldsymbol{\theta}'}(\tfrac{1}{n}\ln\tfrac{kp'}{\delta}) \leq \mathcal{I}^{-1}_{\boldsymbol{\theta}}(\tfrac{1}{n}\ln\tfrac{kp'}{\delta}) \leq L(\boldsymbol{\theta})\,. \tag{C.344}$$

Where we used that the inverse rate is upper bounded by $L(\boldsymbol{\theta})$. By Theorem 5.17, because $L(D,\boldsymbol{\theta}) \leq \epsilon$, we have with h.p. $1-\delta$ over $D \sim \nu^n$,

$$L(\boldsymbol{\theta}') \leq L(D,\boldsymbol{\theta}') + \mathcal{I}^{-1}_{\boldsymbol{\theta}'}(\tfrac{1}{n}\ln\tfrac{kp'}{\delta}) \leq \epsilon + \mathcal{I}^{-1}_{\boldsymbol{\theta}}(\tfrac{1}{n}\ln\tfrac{kp'}{\delta})\,. \tag{C.345}$$

By combining the last two inequalities, we have

$$L(\boldsymbol{\theta}') \leq \epsilon + \mathcal{I}^{-1}_{\boldsymbol{\theta}'}(\tfrac{1}{n}\ln\tfrac{kp'}{\delta}) \leq \epsilon + L(\boldsymbol{\theta})\,. \tag{C.346}$$

□

## C.6 The Implicit Bias of Stochastic Gradient Descent

**Proposition 5.45** [Return to statement]. For any $D \sim \nu^n$ and any $\boldsymbol{\theta} \in \boldsymbol{\Theta}$, it verifies that

$$\hat{L}(D,\boldsymbol{\theta}) = L(\boldsymbol{\theta}) - \mathcal{I}^{-1}_{\boldsymbol{\theta}}(\alpha(D,\boldsymbol{\theta}))\,, \tag{C.347}$$

where $\alpha : \mathcal{D} \times \boldsymbol{\Theta} \to \mathbb{R}$ is defined as $\alpha(D,\boldsymbol{\theta}) := \mathcal{I}_{\boldsymbol{\theta}}(L(\boldsymbol{\theta}) - \hat{L}(D,\boldsymbol{\theta}))$.

*Proof.* Direct consequence of the *signed* rate function being invertible in $\mathbb{R}$. □

**Theorem 5.46** [Return to statement]. For any $\boldsymbol{\theta} \in \boldsymbol{\Theta}$, $n > 0$, and $D \sim \nu^n$, the cumulative distribution of $\alpha(D, \boldsymbol{\theta})$ satisfies

$$\forall s > 0 \quad \mathbb{P}_{D\sim\nu^n}(\alpha(D,\boldsymbol{\theta}) \geq s) \leq e^{-n|s|}\,, \tag{C.348}$$

$$\forall s < 0 \quad \mathbb{P}_{D\sim\nu^n}(\alpha(D,\boldsymbol{\theta}) \leq s) \leq e^{-n|s|}\,, \tag{C.349}$$

and both inequalities are asymptotically tight,

$$\forall s > 0 \quad \mathbb{P}_{D\sim\nu^n}(\alpha(D,\boldsymbol{\theta}) \geq s) \asymp e^{-n|s|}\,, \tag{C.350}$$

$$\forall s < 0 \quad \mathbb{P}_{D\sim\nu^n}(\alpha(D,\boldsymbol{\theta}) \leq s) \asymp e^{-n|s|}\,. \tag{C.351}$$

*Proof.* From Chernoff's bound, it verifies that

$$\forall a \geq 0, \quad \mathbb{P}_{D\sim\nu^n}(L(\boldsymbol{\theta}) - \hat{L}(D,\boldsymbol{\theta}) \geq a) \leq e^{-n|\mathcal{I}_{\boldsymbol{\theta}}(a)|}, \tag{C.352}$$

$$\forall a \leq 0, \quad \mathbb{P}_{D\sim\nu^n}(L(\boldsymbol{\theta}) - \hat{L}(D,\boldsymbol{\theta}) \leq a) \leq e^{-n|\mathcal{I}_{\boldsymbol{\theta}}(a)|}\,. \tag{C.353}$$

As a result, for any value of $s \in \mathbb{R}$, taking $a = \mathcal{I}_{\boldsymbol{\theta}}^{-1}(s)$, we got

$$\forall s \geq 0, \quad \mathbb{P}_{D\sim\nu^n}(L(\boldsymbol{\theta}) - \hat{L}(D,\boldsymbol{\theta}) \geq \mathcal{I}_{\boldsymbol{\theta}}^{-1}(s)) \leq e^{-n|s|}, \tag{C.354}$$

$$\forall s \leq 0, \quad \mathbb{P}_{D\sim\nu^n}(L(\boldsymbol{\theta}) - \hat{L}(D,\boldsymbol{\theta}) \leq \mathcal{I}_{\boldsymbol{\theta}}^{-1}(s)) \leq e^{-n|s|}\,. \tag{C.355}$$

As $\mathcal{I}_{\boldsymbol{\theta}}(\cdot)$ is a strictly monotonic and increasing function, we can apply it at both sides of the inequality inside the probability, giving us:

$$\forall s \geq 0, \quad \mathbb{P}_{D\sim\nu^n}(\mathcal{I}_{\boldsymbol{\theta}}(L(\boldsymbol{\theta}) - \hat{L}(D,\boldsymbol{\theta})) \geq s) \leq e^{-n|s|}, \tag{C.356}$$

$$\forall s \leq 0, \quad \mathbb{P}_{D\sim\nu^n}(\mathcal{I}_{\boldsymbol{\theta}}(L(\boldsymbol{\theta}) - \hat{L}(D,\boldsymbol{\theta})) \leq s) \leq e^{-n|s|}\,. \tag{C.357}$$

The asymptotic inequalities can be obtained by applying the same reasoning to Equation (5.62), which is a direct consequence of Cramér's Theorem. □

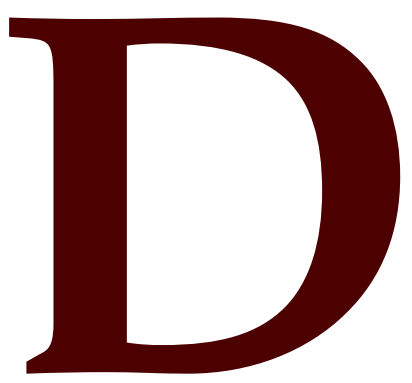

# BayesiPy: Post-hoc Bayesian Inference for Pre-trained Neural Networks

## D.1 Introduction

In contemporary deep learning, deterministic neural networks often achieve remarkable accuracy but remain overconfident in their predictions, especially under distributional shift. Estimating *epistemic* uncertainty—the uncertainty due to limited knowledge of model parameters—has therefore become a key research focus. Traditional Bayesian Neural Networks (BNNs) model a posterior over parameters $P(\boldsymbol{\theta}|\mathcal{D})$, yet this is computationally prohibitive for large-scale architectures.

**BayesiPy** is a Python library (available at www.github.com/Ludvins/BayesiPy) providing a unified framework for *post-hoc uncertainty estimation*. Rather than retraining or modifying the backbone network, BayesiPy wraps a pre-trained model $f(\mathbf{x}; \boldsymbol{\theta}_{\text{MAP}})$ with a Bayesian posterior approximation. This converts a deterministic predictor into a calibrated probabilistic model whose output includes both a predictive mean and an uncertainty estimate.

BayesiPy consolidates several state-of-the-art post-hoc Bayesian approximations, each offering distinct trade-offs between fidelity and scalability:

1. **Linearized Laplace Approximations (LLA)** – including full, layerwise, subnetwork, and last-layer variants with different curvature approximations.
2. **Accelerated Linearized Laplace Approximation (ELLA)** – a scalable, low-rank extension of LLA (Z. Deng et al., 2022).
3. **Variational Linearized Laplace Approximation (VaLLA)** – a variational Gaussian-process reinterpretation of LLA (Luis A. Ortega et al., 2024a).
4. **Mean-Field Variational Inference (MFVI)** – a classic weight-space mean-field variational approximation.

5. **Spectral-Normalized Gaussian Process (SNGP)** – a deterministic yet distance-aware GP head for OOD robustness (Bartlett et al., 2017).
6. **Fixed-Mean Gaussian Process (FMGP)** – a sparse variational GP with the DNN output as fixed mean (Luis A Ortega et al., 2024b).

Together, these methods span the main axis of post-hoc Bayesian deep learning: from *local linearizations in weight-space* (Laplace family) to *function-space Gaussian processes* (VaLLA, FMGP, SNGP).

## D.2 Considered Approaches

In this section, the different methodologies implemented in BayesiPy are reviewed. Please refer to the original works for a better understanding of implementation matters. Table D.1 includes a summarised comparison of the methods included in the library.

### D.2.1 Linearized Laplace Approximations (LLA)

The **Linearized Laplace Approximation (LLA)** (Immer et al., 2021) linearizes the neural network around its maximum-a-posteriori (MAP) parameters $\boldsymbol{\theta}_{\text{MAP}}$. Let $f(\mathbf{x};\boldsymbol{\theta})$ denote the network output and $\ell(\boldsymbol{\theta}) = -\log P(\mathcal{D}|\boldsymbol{\theta}) - \log P(\boldsymbol{\theta})$ its negative log-posterior. A second-order Taylor expansion yields a local Gaussian posterior:

$$P(\boldsymbol{\theta}|\mathcal{D}) \approx \mathcal{N}(\boldsymbol{\theta}_{\text{MAP}},\, H^{-1}), \quad H = \nabla^2_{\boldsymbol{\theta}}\ell(\boldsymbol{\theta})\big|_{\boldsymbol{\theta}=\boldsymbol{\theta}_{\text{MAP}}}, \tag{D.1}$$

where $H$ is the Hessian or a curvature surrogate (e.g., Gauss–Newton). Predictive uncertainty for an unseen input $\mathbf{x}_*$ follows a Gaussian distribution:

$$f(\mathbf{x}_*;\boldsymbol{\theta}) \sim \mathcal{N}(f(\mathbf{x}_*;\boldsymbol{\theta}), J_{\boldsymbol{\theta}_{\text{MAP}}}(\mathbf{x}_*)H^{-1}J_{\boldsymbol{\theta}_{\text{MAP}}}(\mathbf{x}_*)^\top). \tag{D.2}$$

### D.2.2 Accelerated Linearized Laplace Approximation (ELLA)

**ELLA** (Z. Deng et al., 2022) accelerates inference by approximating the *neural tangent kernel* (NTK) with a low-rank Nyström decomposition. Taking a subset of size $M$ of the training data, ELLA constructs approximated features of size $K$, much lower dimensional than the true Jacobian. BayesiPy implements this procedure, enabling near-real-time uncertainty calibration for large architectures.

### D.2.3 Variational Linearized Laplace Approximation (VaLLA)

**VaLLA** (Luis A. Ortega et al., 2024a) recasts LLA in *function space* using sparse Gaussian-process (GP) inference. Rather than storing or inverting $H$, VaLLA uses the GP equivalence of LLA:

$$f(x) \sim \mathcal{GP}(f(x;\boldsymbol{\theta}_{\text{MAP}}),\, J_{\boldsymbol{\theta}_{\text{MAP}}}(x)\, H^{-1}\, J_{\boldsymbol{\theta}_{\text{MAP}}}(x)^\top) \tag{D.3}$$

and performs a sparse variational approximation to the GP using inducing variables $u = f(Z)$ only for the covariance of the GP. From a theoretical point of view, two different sets of inducing locations are used, where the first set is used to fix the GP posterior mean to the pretrained model. The variational posterior $Q(\mathbf{u}) = \mathcal{N}(\mathbf{m}, \mathbf{S})$ minimizes

$$\mathcal{L}_{\text{ELBO}} = \mathbb{E}_{Q(f)}[\log P(y|f)] - \text{KL}(Q(\mathbf{u})|P(\mathbf{u})). \tag{D.4}$$

This yields calibrated uncertainties with sub-linear dependence on data size, while retaining interpretability. The pre-trained model is kept as the posterior GP mean using the dual formulation of GPs in Hilbert spaces; as a result, $\mathbf{m}$ is fixed and not optimized.

Given a test input $x_*$, the *predictive distribution* under VaLLA follows directly from the sparse variational GP formulation:

$$Q(f_*) = \mathcal{N}(f(\mathbf{x}_*; \boldsymbol{\theta}_{\text{MAP}}),\, \sigma_*^2), \tag{D.5}$$

$$\sigma_*^2 = K_{\text{NTK}}(\mathbf{x}_*, \mathbf{x}_*) + K_{\text{NTK}}(\mathbf{x}_*, \mathbf{Z})K_{\text{NTK}}(\mathbf{Z}, \mathbf{Z})^{-1}K_{\text{NTK}}(\mathbf{Z}, \mathbf{x}_*) \tag{D.6}$$

$$- K_{\text{NTK}}(\mathbf{x}_*, \mathbf{Z})K_{\text{NTK}}(\mathbf{Z}, \mathbf{Z})^{-1}\mathbf{S}K_{\text{NTK}}(\mathbf{Z}, \mathbf{Z})^{-1}K_{\text{NTK}}(\mathbf{Z}, \mathbf{x}_*), \tag{D.7}$$

where $K_{\text{NTK}}$ denotes the NTK kernel $K_{\text{NTK}}(\mathbf{x}, \mathbf{x}') = J_{\boldsymbol{\theta}_{\text{MAP}}}(\mathbf{x})\, J_{\boldsymbol{\theta}_{\text{MAP}}}(\mathbf{x}')^\top$, and $K_{ZZ}$ its Gram matrix over the inducing inputs $Z$. This expression provides both the mean correction relative to the deterministic network and a closed-form posterior variance, making VaLLA a scalable Bayesianization of neural networks with predictive uncertainty that is consistent with the local curvature of the loss landscape.

### D.2.4 Fixed-Mean Gaussian Process (FMGP)

The **FMGP** (Luis A Ortega et al., 2024b) uses the dual formulation of GPs in Hilbert spaces to generalize VaLLA to any kernel function (given some caveats). As a result, only kernel hyper-parameters and variational parameters of a sparse GP are optimized. Because the pre-trained model is kept as the mean of the posterior GP approximation, the base network's accuracy is preserved while uncertainty is learned post-hoc.

### D.2.5 Mean-Field Variational Inference (MFVI)

Given a pre-trained neural network with parameters $\boldsymbol{\mu} = \{\mu_1, \ldots, \mu_P\}$, **MFVI** (Z. Deng and Zhu, 2023) approximates the posterior over weights by a fully factorized Gaussian:

$$P(\boldsymbol{\theta}|\mathcal{D}) \approx Q(\boldsymbol{\theta}) := \prod_i \mathcal{N}(\theta_i|\mu_i, \sigma_i^2), \tag{D.8}$$

optimizing the evidence lower bound (ELBO):

$$\mathcal{L} = \mathbb{E}_{Q(\boldsymbol{\theta})}[\log P(\mathcal{D}|\boldsymbol{\theta})] - \text{KL}(Q(\boldsymbol{\theta})|P(\boldsymbol{\theta})). \tag{D.9}$$

Although it underestimates posterior covariance, MFVI provides a cheap Bayesian correction and serves as a baseline for comparison. The set of mean parameters $\boldsymbol{\mu}$ can be either fixed or finetuned. The Flip-out trick is employed for variance-reduced gradient estimation (Wen et al., 2018).

The predictive distribution under MFVI is therefore given by

$$P(y_*|\mathbf{x}_*, \mathcal{D}) = \int P(y_*|\mathbf{x}_*, \boldsymbol{\theta})Q(\boldsymbol{\theta})\, d\boldsymbol{\theta}, \tag{D.10}$$

which is intractable in closed form for general neural networks. In practice, it is approximated by Monte Carlo integration:

$$P(y_*|x_*, \mathcal{D}) \approx \frac{1}{S}\sum_{s=1}^{S} P\left(y_*|\mathbf{x}_*, \boldsymbol{\theta}^{(s)}\right), \quad \boldsymbol{\theta}^{(s)} \sim Q(\boldsymbol{\theta}). \tag{D.11}$$

This corresponds to averaging the model outputs over multiple stochastic forward passes, each with independently sampled weights from the variational posterior. For regression with Gaussian likelihood, the predictive mean and variance can be estimated as

$$\mathbb{E}[f_*] \approx \frac{1}{S}\sum_{s=1}^{S} f(\mathbf{x}_*; \boldsymbol{\theta}^{(s)}), \tag{D.12}$$

$$\mathrm{Var}[f_*] \approx \frac{1}{S}\sum_{s=1}^{S} \left(f(\mathbf{x}_*; \boldsymbol{\theta}^{(s)}) - \mathbb{E}[f_*]\right)^2. \tag{D.13}$$

These expressions yield a practical, sampling-based approximation to Bayesian model averaging. Despite its strong independence assumptions, MFVI provides an efficient and widely used baseline for uncertainty estimation in neural networks.

### D.2.6 Spectral-Normalized Gaussian Process (SNGP)

**SNGP** (J. Z. Liu et al., 2023) augments a deterministic neural network with a *distance-aware Gaussian process* (GP) output layer, enabling calibrated uncertainty estimation and robust out-of-distribution (OOD) detection at deterministic inference cost. The method combines two key components: *spectral normalization* to enforce Lipschitz continuity of the feature extractor, and a *random-feature approximation* of a GP head that models predictive uncertainty.

Let $h(\mathbf{x})$ denote the hidden representation from the penultimate layer. SNGP replaces the final linear layer with a GP approximated by random Fourier features:

$$\Phi(\mathbf{x}) = \sqrt{\tfrac{2\sigma^2}{D_L}} \cos(-\boldsymbol{W}_L h(\mathbf{x}) + \boldsymbol{b}_L), \tag{D.14}$$

where $\boldsymbol{W}_L \sim \mathcal{N}(0, \tau \boldsymbol{I})$ and $\boldsymbol{b}_L \sim \mathcal{U}(0, 2\pi)$. This random-feature expansion approximates a radial basis function (RBF) kernel in feature space. The predictive model is a

Bayesian linear regression over $\Phi(\mathbf{x})$:

$$P(y|\mathbf{x}) = \mathcal{N}(\Phi(\mathbf{x})^\top \boldsymbol{\beta},\ \Phi(\mathbf{x})^\top \boldsymbol{\Sigma}_\beta \Phi(x)), \tag{D.15}$$

with prior $\boldsymbol{\beta} \sim \mathcal{N}(0, I)$ and posterior $P(\boldsymbol{\beta}|\mathcal{D})$ obtained via a Laplace approximation. Thus

$$P(\boldsymbol{\beta}|\mathcal{D}) \approx \mathcal{N}(\boldsymbol{\beta}|\hat{\boldsymbol{\beta}}, \hat{\boldsymbol{\Sigma}}) \tag{D.16}$$

where $\hat{\boldsymbol{\beta}} = \arg\max_{\boldsymbol{\beta}} P(\boldsymbol{\beta}|\mathcal{D})$, and $\hat{\boldsymbol{\Sigma}}_{(i,j)} = -\frac{\partial^2}{\partial \beta_i \partial \beta_j} \log P(\boldsymbol{\beta}|\mathcal{D})_{|\boldsymbol{\beta}=\hat{\boldsymbol{\beta}}}$.

During training, spectral normalization is applied to each weight matrix $\{\boldsymbol{W}_\ell\}_{\ell=1}^{L-1}$ in the model to constrain the feature extractor's Lipschitz constant and preserve distance information in $h(\mathbf{x})$. Given minibatches $\{(\mathbf{x}_m, y_m)\}$, the parameters of both the feature extractor and the random-feature GP head are updated using standard SGD steps.

For a test input $\mathbf{x}_*$, SNGP computes:

$$\Phi_* = \sqrt{\tfrac{2\sigma^2}{D_L}} \cos(\boldsymbol{W}_L h(\mathbf{x}_*) + \boldsymbol{b}_L), \tag{D.17}$$

$$\mu_* = \Phi_*^\top \hat{\boldsymbol{\beta}}, \tag{D.18}$$

$$\sigma_*^2 = \Phi_*^\top \hat{\boldsymbol{\Sigma}}_\beta \Phi_*, \tag{D.19}$$

and outputs the predictive distribution

$$p(y_*|\mathbf{x}_*) = \int \mathrm{softmax}(m)\, \mathcal{N}(m \,|\, \mu_*,\, \sigma_*^2)\, \mathrm{d}m\,, \tag{D.20}$$

for classification problems and the corresponding Gaussian distribution for regression problems.

## D.3 Example Usage

Below is an example of applying BayesiPy's FMGP wrapper to a pre-trained PyTorch model. The used kernel is the squared exponential, with inducing locations initialized using K-Means.

```
1  import copy
2  import torch
3  import numpy as np
4  from bayesipy.fmgp import FMGP
5
6  # Suppose f is your pre-trained PyTorch model
7  fmgp = FMGP(
8      model=copy.deepcopy(f),
9      likelihood="regression",
10     kernel="RBF",
11     inducing_locations="kmeans",
```

```
    num_inducing=50,
)

fmgp.fit(iterations=3000, lr=1e-3, train_loader=train_loader)
mean, var = fmgp.predict(torch.tensor(X_test, dtype=torch.float32))
```

For Linearized Laplace methods, the library allows importing methods from the Laplace Library (Immer et al., 2021), ELLA and VaLLA.

```
from bayesipy.laplace import Laplace, ELLA, VaLLA

lap = Laplace(model=f,
              approximation="kron",
              subset_of_weights="last_layer",
              likelihood="classification")

lap.fit(train_loader)
pred_mean, pred_covariance = lap.predict(test_loader)
```

**Table D.1:** *Comparison of techniques. Insights adapted from the Laplace approximation library (Immer et al., 2021) and recent works such as (Luis A Ortega et al., 2024b).*

| Method | Pros | Cons | Use Case |
|---|---|---|---|
| **Full Laplace** | Most faithful local Gaussian approximation to MAP parameters; captures correlations in all parameters. | Extremely memory- and computation-heavy; not feasible for large networks. | Good for small-to-medium networks or when maximizing fidelity is paramount. |
| **Layerwise / Subnet** | Balances coverage (whole network or a subset) with computational feasibility; can approximate key layers. | More complex to configure (which layers to include? which blocks?); some approximation needed for the Hessian. | For medium-to-large networks if you need more coverage than last-layer but cannot afford full Laplace. |
| **Last-Layer Laplace** | Very fast post-hoc correction; no retraining, just a Hessian-based Gaussian around the final layer. | Focuses only on last-layer uncertainty; can miss uncertainties originating in earlier layers. | A quick Bayesian "upgrade" that often fixes overconfidence on moderate tasks. |
| **ELLA** | Scalable variant of Laplace using low-rank or Nyström approximations. | Hyperparameters for kernel approximation need tuning. | Large-scale scenarios where even standard LLA is too costly. |
| **VaLLA** | Excellent calibration (function-space GP perspective) with sub-linear complexity in data size. | Requires iterative variational optimization; can be slower to converge; more complicated inference step. | High-fidelity uncertainty for large datasets or critical applications. |
| **MFVI** | Classic fully Bayesian approach over weights; easy to implement (Bayes by Backprop). | Mean-field assumption often underestimates uncertainty; can be very slow or memory-heavy for large networks. | For those wanting a "full BNN" approach or partial Bayesian layers with factorized posteriors. |
| **SNGP** | Distance-aware; single forward pass at inference; strong OOD detection. | Typically requires training from scratch or at least heavy fine-tuning with spectral normalization; not purely post-hoc. | Production-friendly if you can integrate spectral norms and a GP head early on. |
| **FMGP** | High-quality calibration; scalable to large data; easy to wrap any pre-trained model (fixed mean). | Extra training to fit the GP's variational parameters; number of inducing points is a hyperparameter that can affect memory usage. | Post-hoc method offering advanced Bayesian-quality uncertainty for large-scale tasks. |

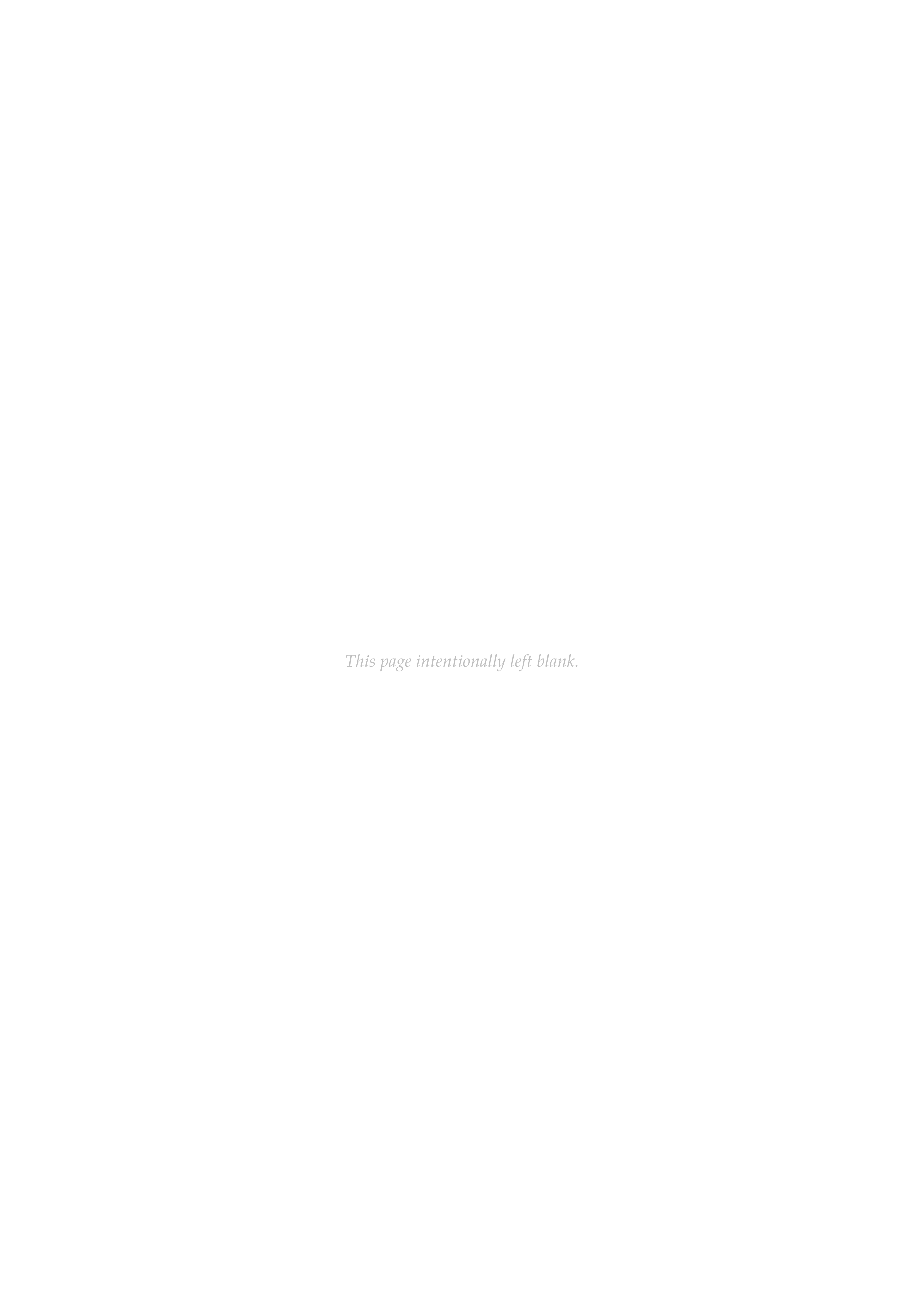

*This page intentionally left blank.*

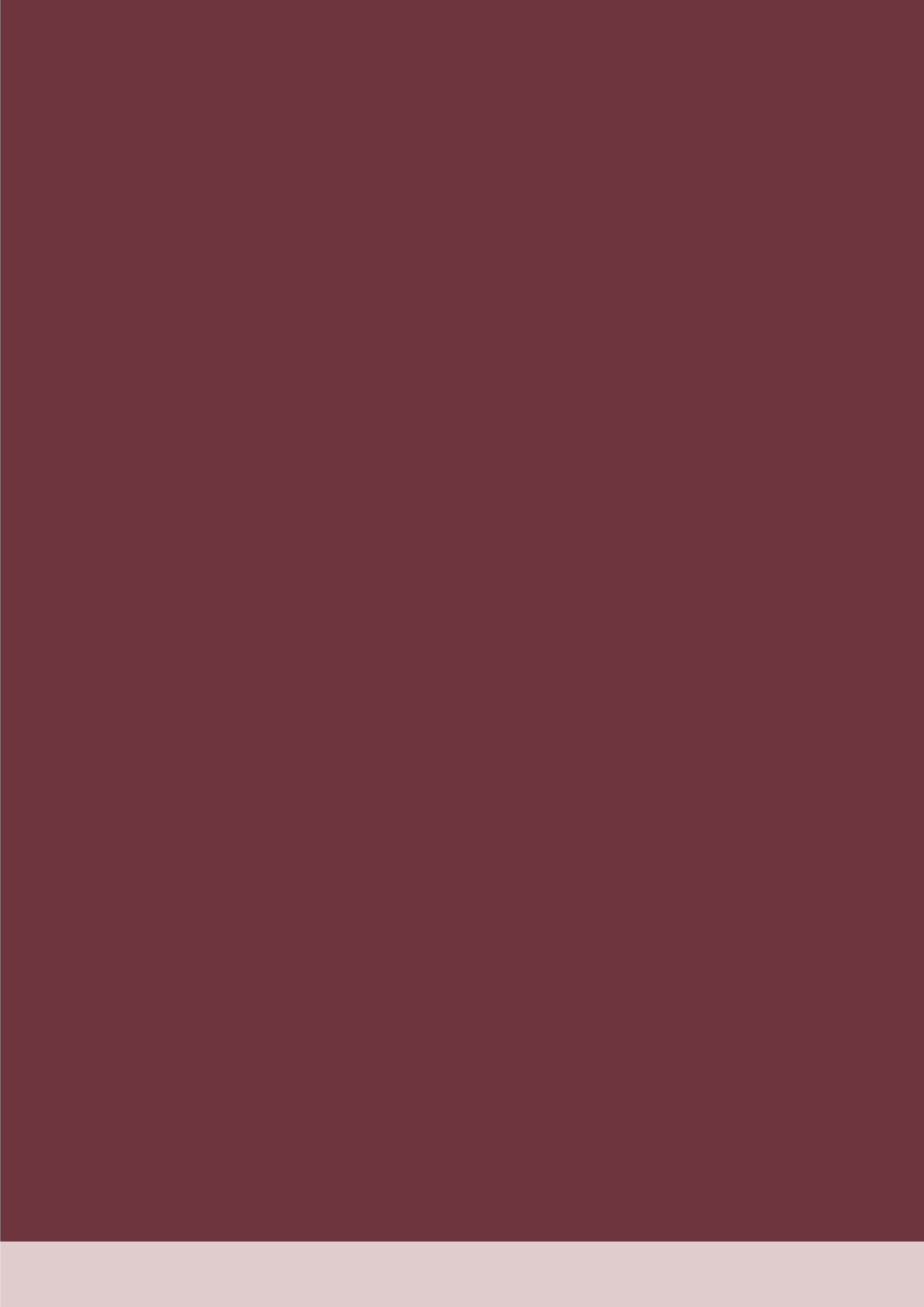